\documentclass[justified,nobib,twoside]{tufte-book}

\usepackage{tikz}
\usetikzlibrary{calc, arrows}
\usetikzlibrary{shapes}
\usetikzlibrary{positioning}
\usepackage{booktabs}
\usepackage{tabularx}
\usepackage{multirow}
\usepackage{longtable}
\usepackage{makecell}
\usepackage{amsmath,amsfonts,amsthm,amssymb,mathrsfs,bbm,mathtools,nicefrac}
\usepackage{enumitem}
\usepackage[noabbrev,capitalize]{cleveref}
\usepackage[italic]{derivative}
\usepackage{xparse}
\usepackage{upgreek}
\usepackage{setspace}
\usepackage[most]{tcolorbox}
\usepackage[parfill]{parskip}
\usepackage{centernot}
\usepackage{needspace}

\usepackage{ifthen}
\newboolean{manuscript}

\usepackage{natbib}
\setcitestyle{authoryear}
\usepackage{usebib}
\bibinput{sources}
\NewDocumentCommand{\pcite}{m}{\citeauthor*{#1} (\citeyear{#1}).\\\emph{\usebibentry{#1}{title}}.}
\NewDocumentCommand{\icite}{m}{``\usebibentry{#1}{title}'' \citep{#1}}

\usepackage{graphicx}

\usepackage[list-style=longtable]{acro}
\usepackage{imakeidx}
\usepackage{hyperref}
\makeindex[intoc]
\indexsetup{headers={\indexname}{\indexname}}
\acsetup{index/use=true}

\NewDocumentCommand{\pidx}{mo}{\index{\IfValueTF{#2}{#1|#2}{#1|idxpage}}}
\NewDocumentCommand{\idx}{d<>mo}{\emph{\IfValueTF{#1}{#1}{#2}}\pidx{#2}[#3]}
\NewDocumentCommand{\midx}{d<>omo}{\idx<#1>{#3}[#4]}

\newtheorem{thm}{Theorem}[chapter]
\newtheorem{cor}[thm]{Corollary}
\newtheorem{lem}[thm]{Lemma}
\newtheorem{fct}[thm]{Fact}

\theoremstyle{definition}
\newtheorem{defn}[thm]{Definition}

\numberwithin{equation}{chapter}

\Crefname{thm}{Theorem}{Theorems}

\definecolor{blue}{RGB}{02,106,253}
\definecolor{red}{RGB}{245,51,30}
\definecolor{green}{RGB}{96,189,69}
\definecolor{purple}{RGB}{200,0,240}
\def\b{\textcolor{blue}}
\def\r{\textcolor{red}}
\def\g{\textcolor{green}}
\def\purple{\textcolor{purple}}

\usepackage{xcolor,amsmath}
\usepackage[linesnumbered,ruled,vlined,algochapter]{algorithm2e}
\DontPrintSemicolon
\makeatletter
    \let\c@algocf\c@thm
\makeatother
\crefname{algocf}{Algorithm}{Algorithms}
\Crefname{algocf}{Algorithm}{Algorithms}
\SetAlgoSkip{bigskip}

\SetKwComment{Comment}{\color{green!50!black}// }{}

\SetKwProg{Function}{function}{}{}

\tcbuselibrary{theorems}
\newtcbtheorem
  [use counter*=thm,number within=chapter,crefname={Example}{Examples},Crefname={Example}{Examples}]%
  {ex}
  {Example}
  {%
    before skip=10pt,after skip=10pt,
    left=0.2cm,right=0.2cm,top=0cm,
    toptitle=0.2cm,bottomtitle=0cm,
    breakable,
    toprule at break=0.2cm,
    sharp corners,
    colback=blue!10,
    coltitle=black,
    colframe=blue!10,
    fonttitle=\bfseries,
    parbox=false,
    halign=justify, %
  }%
  {ex}%

\newtcbtheorem
  [use counter*=thm,number within=chapter,crefname={Remark}{Remarks},Crefname={Remark}{Remarks}]%
  {rmk}
  {Remark}
  {%
    before skip=10pt,after skip=10pt,
    left=0.2cm,right=0.2cm,top=0cm,
    toptitle=0.2cm,bottomtitle=0cm,
    breakable,
    toprule at break=0.2cm,
    sharp corners,
    colback=gray!20,
    coltitle=black,
    colframe=gray!20,
    fonttitle=\bfseries,
    parbox=false,
    halign=justify,
  }%
  {rmk}%

\newtcbtheorem
  [use counter*=thm,number within=chapter,crefname={Problem}{Problems},Crefname={Problem}{Problems}]%
  {exc}
  {Problem}
  {%
    before skip=10pt,after skip=10pt,
    left=0.2cm,right=0.2cm,top=0cm,
    toptitle=0.2cm,bottomtitle=0cm,
    breakable,
    toprule at break=0.2cm,
    sharp corners,
    colback=purple!10,
    coltitle=black,
    colframe=purple!10,
    fonttitle=\bfseries,
    parbox=false,
    halign=justify,
  }%
  {exercise}%

\newtcolorbox{readings}{%
    before skip=10pt,after skip=10pt,
    left=0.2cm,right=0.2cm,top=0cm,
    toptitle=0.2cm,bottomtitle=0cm,
    breakable,
    toprule at break=0.2cm,
    sharp corners,
    colback=green!30,
    coltitle=black,
    colframe=green!30,
    fonttitle=\bfseries,
    title={Readings},
    parbox=false,
    halign=justify,
}
\newtcolorbox{oreadings}{%
    before skip=10pt,after skip=10pt,
    left=0.2cm,right=0.2cm,top=0cm,
    toptitle=0.2cm,bottomtitle=0cm,
    breakable,
    toprule at break=0.2cm,
    sharp corners,
    colback=green!20,
    coltitle=black,
    colframe=green!20,
    fonttitle=\bfseries,
    title={Optional Readings},
    parbox=false,
    halign=justify,
}

\newtcolorbox{thmb}{%
    before skip=10pt,after skip=10pt,
    left=0.2cm,right=0.2cm, top=0cm,
    toptitle=0cm,bottomtitle=0cm,
    breakable,
    toprule at break=0.2cm,
    sharp corners,
    colback=gray!10,
    coltitle=black,
    colframe=gray!10,
    fonttitle=\bfseries,
    title={},
    parbox=false,
    halign=justify,
}

\let\marginnote\relax
\usepackage{marginnote}
\NewDocumentCommand{\margintag}{O{0\baselineskip}m}{%
  \checkoddpage%
  \ifoddpage%
    {\marginnote{\footnotesize #2}[#1]}%
  \else%
    {\reversemarginpar\marginnote{\footnotesize #2}[#1]}%
  \fi}%
\NewDocumentCommand{\safefootnote}{om}{\footnotemark\margintag[#1]{\textsuperscript{\tiny\arabic{footnote}} \normalfont\justifying #2}}

\NewDocumentEnvironment{marginbox}{O{0\baselineskip}m}{\begin{marginfigure}[#1]{\textbf{#2}}\quad}{\end{marginfigure}}

\makeatletter
\NewDocumentCommand{\embeq}{m}{%
  \leavevmode\hfill\refstepcounter{equation}\textup{\tagform@{\theequation}}\label{#1}%
}
\makeatother

\makeatletter
\NewDocumentCommand{\algeq}{m}{%
  \leavevmode\Comment*[r]{\refstepcounter{equation}\textup{\tagform@{\theequation}}\label{#1}}%
}
\makeatother

\usepackage{etoolbox}
\makeatletter
\patchcmd{\@algocf@start}%
  {-1.5em}%
  {0pt}%
  {}{}%
\makeatother

\allowdisplaybreaks

\setlist[enumerate]{noitemsep, topsep=-6pt, leftmargin=16pt}
\setlist[itemize]{noitemsep, topsep=-6pt}

\usepackage{import}
\usepackage{xifthen}
\usepackage{pdfpages}
\usepackage{transparent}

\NewDocumentCommand{\incfig}{om}{%
    \IfValueTF{#1}{%
        \def\svgwidth{#1}%
    }{%
        \def\svgwidth{\columnwidth}%
    }%
    \centering\import{./figures/}{#2.pdf_tex}%
}

\usepackage{titlesec}
\titleclass{\part}{top} %
\titleformat{\part}
[display]
{\centering\normalfont}
{\vspace{3pt}\Large\smallcaps{\partname} \thepart}
{0pt}
{\vspace{1pc}\Huge\normalfont\textit}
\titlespacing*{\part}{0pt}{0pt}{20pt}

\NewDocumentEnvironment{exercise}{mm}{\begin{exc}{#1}{#2}}{\par\textit{\hyperref[solution:#2]{$\triangleright$ Solution}}
\end{exc}}

\newtheorem{nexc}{}[chapter]
\crefname{nexc}{Problem}{Problems}
\Crefname{nexc}{Problem}{Problems}

\NewDocumentCommand{\excheading}{}{\needspace{6\baselineskip}\section*{Problems}}

\NewDocumentEnvironment{nexercise}{mm}{%
  \needspace{2\baselineskip}%
  \begin{nexc}%
  \hyperref[solution:#2]{\textbf{#1.}}\label{exercise:#2}%
  \par\nobreak
}{%
  \end{nexc}%
}

\NewDocumentCommand{\exerciseref}{mo}{\margintag{\normalfont\textbf{\Cref{exercise:#1} \IfValueT{#2}{({#2})}{}}}}
\newcommand*\circled[1]{\tikz[baseline=(char.base)]{
            \node[shape=circle,draw,inner sep=1pt] (char) {#1};}}
\NewDocumentCommand{\exerciserefmark}{mo}{\hyperref[exercise:#1]{\circled{\normalfont\textbf{?}}}\exerciseref{#1}[#2]}

\NewDocumentEnvironment{solution}{m}{\paragraph{\normalfont{\textbf{Solution to \cref{exercise:#1}.}}}\label{solution:#1}}{}

\newcommand{\course}{\ifthenelse{\boolean{manuscript}}{manuscript}{course}\xspace}

\NewDocumentCommand{\ceil}{m}{\left\lceil #1 \right\rceil}
\NewDocumentCommand{\ip}{m}{\left\langle #1 \right\rangle}

\newcommand*{\abs}[1]{\left| #1 \right|}
\newcommand*{\card}[1]{\left| #1 \right|}
\NewDocumentCommand{\norm}{sm}{\IfBooleanTF{#1}{\|#2\|}{\left\| #2 \right\|}}

\newcommand*{\const}{\mathrm{const}}

\newcommand*{\defeq}{\overset{.}{=}}
\newcommand*{\eqdef}{\overset{.}{=}}

\DeclareMathOperator*{\argmax}{arg\,max}
\DeclareMathOperator*{\argmin}{arg\,min}
\DeclareMathOperator*{\iid}{\smash{\overset{\mathrm{iid}}{\sim}}}

\DeclareMathOperator*{\almostsurely}{\smash{\overset{a.s.}{\rightarrow}}}
\DeclareMathOperator*{\eqalmostsurely}{\smash{\overset{a.s.}{=}}}
\DeclareMathOperator*{\convp}{\smash{\overset{\fnPr}{\rightarrow}}}
\DeclareMathOperator*{\convd}{\smash{\overset{\mathcal{D}}{\rightarrow}}}

\DeclarePairedDelimiter\parentheses{(}{)}
\DeclarePairedDelimiter\brackets{[}{]}
\DeclarePairedDelimiter\braces{\{}{\}}

\newcommand{\R}{\mathbb{R}}
\newcommand{\Rzero}{\mathbb{R}_{\geq 0}}
\newcommand{\Nat}{\mathbb{N}}
\newcommand{\NatZ}{\mathbb{N}_0}

\newcommand{\MSE}{\mathrm{mse}}
\newcommand{\MLE}{\mathrm{MLE}}
\newcommand{\MAP}{\mathrm{MAP}}
\newcommand{\ls}{\mathrm{ls}}
\newcommand{\ridge}{\mathrm{ridge}}
\newcommand{\lasso}{\mathrm{lasso}}
\newcommand{\train}{\mathrm{train}}
\newcommand{\val}{\mathrm{val}}
\newcommand{\id}{\mathrm{id}}
\newcommand{\init}{\mathrm{init}}

\renewcommand{\vec}[1]{\mathbold{#1}}
\newcommand{\mat}[1]{\mathbold{#1}}
\newcommand{\rvec}[1]{\mathbf{#1}}
\newcommand{\set}[1]{#1}
\newcommand{\spa}[1]{\mathcal{#1}}

\newcommand{\mean}[1]{\overline{#1}}
\newcommand{\compl}[1]{\overline{#1}}
\newcommand{\old}[1]{#1^{\mathrm{old}}}
\newcommand{\opt}[1]{#1^\star}

\newcommand{\altpi}{\Pi} %

\NewDocumentCommand{\Ind}{m}{\mathbb{1}\{{#1}\}}
\NewDocumentCommand{\fnPr}{}{\mathbb{P}}
\RenewDocumentCommand{\Pr}{om}{\fnPr\IfValueT{#1}{_{#1}}\parentheses*{#2}}
\RenewDocumentCommand{\H}{mo}{\mathrm{H}\IfValueTF{#2}{\!\left[#1\ \middle|\ #2\right]}{\brackets*{#1}}}
\NewDocumentCommand{\Hsm}{mo}{\mathrm{H}\IfValueTF{#2}{[#1 \mid #2]}{\brackets{#1}}}
\NewDocumentCommand{\I}{mmo}{\mathrm{I}\IfValueTF{#3}{\!\left(#1;#2\ \middle|\ #3\right)}{\parentheses*{#1; #2}}}
\NewDocumentCommand{\Ism}{mmo}{\mathrm{I}\IfValueTF{#3}{(#1;#2 \mid #3)}{\parentheses{#1; #2}}}
\NewDocumentCommand{\fnS}{}{\mathrm{S}}
\RenewDocumentCommand{\S}{m}{\fnS\brackets*{#1}}
\NewDocumentCommand{\E}{somo}{\ensuremath{\mathbb{E}\IfValueT{#2}{_{#2}}{} \IfBooleanTF{#1}{#3}{\IfValueTF{#4}{\!\left[#3\ \middle|\ #4\right]}{\brackets*{#3}}}}}
\NewDocumentCommand{\Var}{somo}{\mathrm{Var}\IfValueT{#2}{_{#2}}{} \IfBooleanTF{#1}{#3}{\IfValueTF{#4}{\!\left[#3\ \middle|\ #4\right]}{\brackets*{#3}}}}
\NewDocumentCommand{\Cov}{som}{\mathrm{Cov}\IfValueT{#2}{_{#2}}{} \IfBooleanTF{#1}{#3}{\brackets*{#3}}}
\NewDocumentCommand{\Cor}{som}{\mathrm{Cor}\IfValueT{#2}{_{#2}}{} \IfBooleanTF{#1}{#3}{\brackets*{#3}}}
\NewDocumentCommand{\SD}{m}{\sigma\brackets*{#1}}
\NewDocumentCommand{\KL}{mm}{\mathrm{KL}\parentheses*{#1 \| #2}}
\NewDocumentCommand{\KLsm}{mm}{\mathrm{KL}\parentheses{#1 \| #2}}
\NewDocumentCommand{\Fisher}{mm}{\mathrm{J}\parentheses*{#1 \| #2}}
\NewDocumentCommand{\crH}{mm}{\mathrm{H}\brackets*{#1 \| #2}}
\NewDocumentCommand{\crHsm}{mm}{\mathrm{H}\brackets{#1 \| #2}}

\NewDocumentCommand{\fnv}{oo}{v\IfValueT{#2}{_{#2}}\IfValueT{#1}{^{#1}}}
\RenewDocumentCommand{\v}{somo}{\IfBooleanTF{#1}{\fnv[\star][#4]\parentheses{#3}}{\fnv[#2][#4]\parentheses{#3}}}
\NewDocumentCommand{\fnq}{oo}{q\IfValueT{#2}{_{#2}}\IfValueT{#1}{^{#1}}}
\NewDocumentCommand{\q}{sommo}{\IfBooleanTF{#1}{\fnq[\star][#5]\parentheses{#3,#4}}{\fnq[#2][#5]\parentheses{#3,#4}}}
\NewDocumentCommand{\fnV}{oo}{V\IfValueT{#2}{_{#2}}\IfValueT{#1}{^{#1}}}
\NewDocumentCommand{\V}{somo}{\IfBooleanTF{#1}{\fnV[\star][#4]\parentheses{#3}}{\fnV[#2][#4]\parentheses{#3}}}
\NewDocumentCommand{\fnQ}{oo}{Q\IfValueT{#2}{_{#2}}\IfValueT{#1}{^{#1}}}
\NewDocumentCommand{\Q}{sommo}{\IfBooleanTF{#1}{\fnQ[\star][#5]\parentheses{#3,#4}}{\fnQ[#2][#5]\parentheses{#3,#4}}}
\NewDocumentCommand{\fna}{oo}{a\IfValueT{#2}{_{#2}}\IfValueT{#1}{^{#1}}}
\RenewDocumentCommand{\a}{sommo}{\IfBooleanTF{#1}{\fna[\star][#5]\parentheses{#3,#4}}{\fna[#2][#5]\parentheses{#3,#4}}}
\NewDocumentCommand{\fnA}{oo}{A\IfValueT{#2}{_{#2}}\IfValueT{#1}{^{#1}}}
\NewDocumentCommand{\fnAhat}{oo}{\hat{A}\IfValueT{#2}{_{#2}}\IfValueT{#1}{^{#1}}}
\NewDocumentCommand{\A}{sommo}{\IfBooleanTF{#1}{\fnA[\star][#5]\parentheses{#3,#4}}{\fnA[#2][#5]\parentheses{#3,#4}}}
\NewDocumentCommand{\fnj}{o}{J\IfValueT{#1}{_{#1}}}
\RenewDocumentCommand{\j}{mo}{\fnj[#2]\parentheses{#1}}
\NewDocumentCommand{\fnJ}{o}{\widehat{J}\IfValueT{#1}{_{#1}}}
\NewDocumentCommand{\J}{mo}{\fnJ[#2]\parentheses{#1}}

\NewDocumentCommand{\pset}{m}{\mathcal{P}\parentheses*{#1}}

\NewDocumentCommand{\pf}{mm}{{#1}_\sharp #2}

\NewDocumentCommand{\grad}{e_}{\boldsymbol{\nabla}\IfValueT{#1}{_{\!\!#1}\,}}
\NewDocumentCommand{\jac}{}{\mD}
\NewDocumentCommand{\hes}{}{\mH}
\NewDocumentCommand{\dive}{}{\grad\cdot}
\NewDocumentCommand{\lapl}{}{\Delta}

\NewDocumentCommand{\BigO}{m}{O\parentheses*{#1}}
\NewDocumentCommand{\BigOTil}{m}{\widetilde{O}\parentheses*{#1}}

\NewDocumentCommand{\transpose}{m}{#1^\top}
\NewDocumentCommand{\inv}{m}{#1^{-1}}
\RenewDocumentCommand{\det}{m}{\mathrm{det}\parentheses*{#1}}
\NewDocumentCommand{\tr}{m}{\mathrm{tr}\parentheses*{#1}}
\NewDocumentCommand{\diag}{om}{\mathrm{diag}\IfValueT{#1}{_{#1}}{}\braces{#2}}
\NewDocumentCommand{\msqrt}{m}{#1^{\nicefrac{1}{2}}}
\NewDocumentCommand{\vecop}{m}{\mathrm{vec}\brackets{#1}}

\NewDocumentCommand{\N}{somm}{\mathcal{N}\IfBooleanTF{#1}{\left(}{(}\IfValueT{#2}{#2;}{} #3, #4\IfBooleanTF{#1}{\right)}{)}}
\NewDocumentCommand{\SN}{o}{\mathcal{N}(\IfValueT{#1}{#1;}{} \vzero, \mI)}
\NewDocumentCommand{\uSN}{o}{\mathcal{N}(\IfValueT{#1}{#1;}{} 0, 1)}
\NewDocumentCommand{\KF}{ommmmm}{\mathcal{KF}(\IfValueT{#1}{#1;}{} #2, #3, #4, #5, #6)}
\NewDocumentCommand{\GP}{omm}{\mathcal{GP}(\IfValueT{#1}{#1;}{} #2, #3)}
\NewDocumentCommand{\Unif}{om}{\mathrm{Unif}(\IfValueT{#1}{#1;}{} #2)}
\NewDocumentCommand{\Bern}{om}{\mathrm{Bern}(\IfValueT{#1}{#1;}{} #2)}
\NewDocumentCommand{\Bin}{omm}{\mathrm{Bin}(\IfValueT{#1}{#1;}{} #2, #3)}
\NewDocumentCommand{\Beta}{omm}{\mathrm{Beta}(\IfValueT{#1}{#1;}{} #2, #3)}
\NewDocumentCommand{\Laplace}{omm}{\mathrm{Laplace}(\IfValueT{#1}{#1;}{} #2, #3)}
\NewDocumentCommand{\Cauchy}{omm}{\mathrm{Cauchy}(\IfValueT{#1}{#1;}{} #2, #3)}
\NewDocumentCommand{\GammaDistr}{omm}{\mathrm{Gamma}(\IfValueT{#1}{#1;}{} #2, #3)}
\NewDocumentCommand{\Pareto}{omm}{\mathrm{Pareto}(\IfValueT{#1}{#1;}{} #2, #3)}

\newcommand{\sigman}{\sigma_{\mathrm{n}}}
\newcommand{\sigmap}{\sigma_{\mathrm{p}}}

\newcommand{\vzero}{\vec{0}}

\newcommand{\va}{\vec{a}}
\newcommand{\vap}{\vec{a'}}
\newcommand{\vas}{\vec{\opt{a}}}
\newcommand{\vb}{\vec{b}}
\newcommand{\vc}{\vec{c}}
\newcommand{\vd}{\vec{d}}
\newcommand{\ve}{\vec{e}}
\newcommand{\vf}{\vec{f}}
\newcommand{\vfhat}{\vec{\hat{f}}}
\newcommand{\vg}{\vec{g}}
\newcommand{\vh}{\vec{h}}
\newcommand{\vk}{\vec{k}}
\newcommand{\vm}{\vec{m}}
\newcommand{\vp}{\vec{p}}
\newcommand{\vq}{\vec{q}}
\newcommand{\vr}{\vec{r}}
\newcommand{\vs}{\vec{s}}
\newcommand{\vt}{\vec{t}}
\newcommand{\vu}{\vec{u}}
\newcommand{\vv}{\vec{v}}
\newcommand{\vvp}{\vec{v'}}
\newcommand{\vvs}{\vec{\opt{v}}}
\newcommand{\vw}{\vec{w}}
\newcommand{\vwhat}{\vec{\hat{w}}}
\newcommand{\vx}{\vec{x}}
\newcommand{\vxp}{\vec{x'}}
\newcommand{\vxs}{\vec{\opt{x}}}
\newcommand{\vy}{\vec{y}}
\newcommand{\vyp}{\vec{y'}}
\newcommand{\vz}{\vec{z}}
\newcommand{\valpha}{\vec{\alpha}}
\newcommand{\valphahat}{\vec{\hat{\alpha}}}
\newcommand{\vdelta}{\vec{\delta}}
\newcommand{\vDelta}{\vec{\Delta}}

\newcommand{\vvarepsilon}{\vec{\varepsilon}}
\newcommand{\veta}{\vec{\eta}}
\newcommand{\vlambda}{\vec{\lambda}}
\newcommand{\vmu}{\vec{\mu}}
\newcommand{\vmuhat}{\vec{\hat{\mu}}}
\newcommand{\vmup}{\vec{\mu'}}
\newcommand{\vnu}{\vec{\nu}}
\newcommand{\vomega}{\vec{\omega}}
\newcommand{\vphi}{\vec{\phi}}
\newcommand{\vpi}{\vec{\pi}}
\newcommand{\vpsi}{\vec{\psi}}
\newcommand{\vvarphi}{\vec{\varphi}}
\newcommand{\vvarphihat}{\vec{\hat{\varphi}}}
\newcommand{\vtheta}{\vec{\theta}}
\newcommand{\vthetahat}{\vec{\hat{\theta}}}
\newcommand{\vxi}{\vec{\xi}}
\newcommand{\mzero}{\mat{0}}
\newcommand{\mA}{\mat{A}}
\newcommand{\mB}{\mat{B}}
\newcommand{\mBs}{\mat{\opt{B}}}
\newcommand{\mC}{\mat{C}}
\newcommand{\mD}{\mat{D}}
\newcommand{\mF}{\mat{F}}
\newcommand{\mH}{\mat{H}}
\newcommand{\mI}{\mat{I}}
\newcommand{\mK}{\mat{K}}

\newcommand{\mCalL}{\mat{\mathcal{L}}}
\newcommand{\mM}{\mat{M}}
\newcommand{\mP}{\mat{P}}
\newcommand{\mQ}{\mat{Q}}
\newcommand{\mS}{\mat{S}}

\newcommand{\mU}{\mat{U}}
\newcommand{\mV}{\mat{V}}
\newcommand{\mW}{\mat{W}}
\newcommand{\mX}{\mat{X}}
\newcommand{\mLambda}{\mat{\Lambda}}
\newcommand{\mPhi}{\mat{\Phi}}
\newcommand{\mPi}{\mat{\Pi}}
\newcommand{\mSigma}{\mat{\Sigma}}
\newcommand{\mSigmap}{\mat{\Sigma'}}
\newcommand{\rG}{\rvec{G}}

\newcommand{\rW}{\rvec{W}}
\newcommand{\rX}{\rvec{X}}
\newcommand{\rXp}{\rvec{X'}}
\newcommand{\rY}{\rvec{Y}}
\newcommand{\rZ}{\rvec{Z}}
\newcommand{\sA}{\set{A}}
\newcommand{\sB}{\set{B}}

\newcommand{\sI}{\set{I}}
\newcommand{\sM}{\set{M}}
\newcommand{\sS}{\set{S}}
\newcommand{\sU}{\set{U}}
\newcommand{\sX}{\set{X}}
\newcommand{\sY}{\set{Y}}
\newcommand{\sZ}{\set{Z}}
\newcommand{\spA}{\spa{A}}
\newcommand{\spB}{\spa{B}}
\newcommand{\spC}{\spa{C}}
\newcommand{\spD}{\spa{D}}
\newcommand{\spF}{\spa{F}}
\newcommand{\spH}{\spa{H}}
\newcommand{\spL}{\spa{L}}
\newcommand{\spM}{\spa{M}}
\newcommand{\spO}{\spa{O}}
\newcommand{\spP}{\spa{P}}
\newcommand{\spQ}{\spa{Q}}
\newcommand{\spT}{\spa{T}}
\newcommand{\spW}{\spa{W}}
\newcommand{\spX}{\spa{X}}
\newcommand{\spY}{\spa{Y}}

\newcommand{\fs}{\opt{f}}

\newcommand{\qs}{\opt{q}}
\newcommand{\xs}{\opt{x}}
\newcommand{\ys}{\opt{y}}

\newcommand{\sSs}{\opt{\sS}}

\newcommand{\pis}{\opt{\pi}}

\newcommand{\vF}{\rvec{F}}

\newcommand{\vT}{\rvec{T}}

\setboolean{manuscript}{true}

\DeclareAcronym{iff}{short=iff, long=if and only if}
\DeclareAcronym{wrt}{short=w.r.t., long=with respect to}
\DeclareAcronym{wlog}{short=w.l.o.g., long=without loss of generality}
\DeclareAcronym{iid}{short=i.i.d., long=independent and identically distributed}
\DeclareAcronym{as}{short=a.s., long={almost surely, with high probability, with probability 1}}

\DeclareAcronym{A2C}{short=A2C, long=advantage actor-critic}
\DeclareAcronym{BALD}{short=BALD, long=Bayesian active learning by disagreement}
\DeclareAcronym{BLR}{short=BLR, long=Bayesian linear regression}
\DeclareAcronym{CDF}{short=CDF, long=cumulative distribution function}
\DeclareAcronym{CLT}{short=CLT, long=central limit theorem}
\DeclareAcronym{DDPG}{short=DDPG, long=deep deterministic policy gradients}
\DeclareAcronym{DDQN}{short=DDQN, long=double deep Q-networks}
\DeclareAcronym{DPO}{short=DPO, long=direct preference optimization}
\DeclareAcronym{DQN}{short=DQN, long=deep Q-networks}
\DeclareAcronym{ECE}{short=ECE, long=expected calibration error}
\DeclareAcronym{EI}{short=EI, long=expected improvement}
\DeclareAcronym{ELBO}{short=ELBO, long=evidence lower bound}
\DeclareAcronym{ES}{short=ES, long=entropy search}
\DeclareAcronym{FITC}{short=FITC, long=fully independent training conditional}
\DeclareAcronym{GAE}{short=GAE, long=generalized advantage estimation}
\DeclareAcronym{GD}{short=GD, long=gradient descent}
\DeclareAcronym{GLIE}{short=GLIE, long=greedy in the limit with infinite exploration}
\DeclareAcronym{GP}{short=GP, long=Gaussian process}
\DeclareAcronym{GPC}{short=GPC, long=Gaussian process classification}
\DeclareAcronym{GRPO}{short=GRPO, long=group relative policy optimization}
\DeclareAcronym{GRV}{short=GRV, long=Gaussian random vector}
\DeclareAcronym{HMC}{short=HMC, long=Hamiltonian Monte Carlo}
\DeclareAcronym{HMM}{short=HMM, long=hidden Markov model}
\DeclareAcronym{H-UCRL}{short=H-UCRL, long=hallucinated upper confidence reinforcement learning}
\DeclareAcronym{IDS}{short=IDS, long=information-directed sampling}
\DeclareAcronym{JES}{short=JES, long=joint entropy search}
\DeclareAcronym{KL}{short=KL, long=Kullback-Leibler}
\DeclareAcronym{LAMBDA}{short=LAMBDA, long=Lagrangian model-based agent}
\DeclareAcronym{LASSO}{short=LASSO, long=least absolute shrinkage and selection operator}
\DeclareAcronym{LD}{short=LD, long=Langevin dynamics}
\DeclareAcronym{LITE}{short=LITE, long=linear-time independence-based estimators (of probability of maximality)}
\DeclareAcronym{LMC}{short=LMC, long=Langevin Monte Carlo}
\DeclareAcronym{LOTE}{short=LOTE, long=law of total expectation}
\DeclareAcronym{LOTP}{short=LOTP, long=law of total probability}
\DeclareAcronym{LOTV}{short=LOTV, long=law of total variance}
\DeclareAcronym{LOTUS}{short=LOTUS, long=law of the unconscious statistician}
\DeclareAcronym{LSI}{short=LSI, long=log-Sobolev inequality}
\DeclareAcronym{MAB}{short=MAB, long=multi-armed bandits}
\DeclareAcronym{MALA}{short=MALA, long=Metropolis adjusted Langevin algorithm}
\DeclareAcronym{MAP}{short=MAP, long=maximum a posteriori}
\DeclareAcronym{MC}{short=MC, long=Monte Carlo}
\DeclareAcronym{MCE}{short=MCE, long=maximum calibration error}
\DeclareAcronym{MCMC}{short=MCMC, long=Markov chain Monte Carlo}
\DeclareAcronym{MCTS}{short=MCTS, long=Monte Carlo tree search}
\DeclareAcronym{MDP}{short=MDP, long=Markov decision process}
\DeclareAcronym{MES}{short=MES, long=max-value entropy search}
\DeclareAcronym{MERL}{short=MERL, long=maximum entropy reinforcement learning}
\DeclareAcronym{MGF}{short=MGF, long=moment-generating function}
\DeclareAcronym{MI}{short=MI, long=mutual information}
\DeclareAcronym{MLE}{short=MLE, long=maximum likelihood estimate}
\DeclareAcronym{MLL}{short=MLL, long=marginal log likelihood}
\DeclareAcronym{MPC}{short=MPC, long=model predictive control}
\DeclareAcronym{MSE}{short=MSE, long=mean squared error}
\DeclareAcronym{NLL}{short=NLL, long=negative log likelihood}
\DeclareAcronym{ODE}{short=ODE, long=ordinary differential equation}
\DeclareAcronym{OPES}{short=OPES, long=output-space predictive entropy search}
\DeclareAcronym{PBPI}{short=PBPI, long=point-based policy iteration}
\DeclareAcronym{PBVI}{short=PBVI, long=point-based value iteration}
\DeclareAcronym{PDF}{short=PDF, long=probability density function}
\DeclareAcronym{PES}{short=PES, long=predictive entropy search}
\DeclareAcronym{PETS}{short=PETS, long=probabilistic ensembles with trajectory sampling}
\DeclareAcronym{PI1}{short=PI, long=policy iteration}
\DeclareAcronym{PI2}{short=PI, long=probability of improvement}
\DeclareAcronym{PILCO}{short=PILCO, long=probabilistic inference for learning control}
\DeclareAcronym{PL}{short=PL, long=Polyak-Łojasiewicz}
\DeclareAcronym{PlaNet}{short=PlaNet, long=deep planning network}
\DeclareAcronym{PMF}{short=PMF, long=probability mass function}
\DeclareAcronym{POMDP}{short=POMDP, long=partially observable Markov decision process}
\DeclareAcronym{PPO}{short=PPO, long=proximal policy optimization}
\DeclareAcronym{RBF}{short=RBF, long=radial basis function}
\DeclareAcronym{ReLU}{short=ReLU, long=rectified linear unit}
\DeclareAcronym{RHC}{short=RHC, long=receding horizon control}
\DeclareAcronym{RKHS}{short=RKHS, long=reproducing kernel Hilbert space}
\DeclareAcronym{RL}{short=RL, long=reinforcement learning}
\DeclareAcronym{RLHF}{short=RLHF, long=reinforcement learning from human feedback}
\DeclareAcronym{RM}{short=RM, long=Robbins-Monro}
\DeclareAcronym{SAA}{short=SAA, long=stochastic average approximation}
\DeclareAcronym{SAC}{short=SAC, long=soft actor-critic}
\DeclareAcronym{SARSA}{short=SARSA, long=state-action-reward-state-action}
\DeclareAcronym{SDE}{short=SDE, long=stochastic differential equation}
\DeclareAcronym{SGD}{short=SGD, long=stochastic gradient descent}
\DeclareAcronym{SG-HMC}{short=SG-HMC, long=stochastic gradient Hamiltonian Monte Carlo}
\DeclareAcronym{SGLD}{short=SGLD, long=stochastic gradient Langevin dynamics}
\DeclareAcronym{SLLN}{short=SLLN, long=strong law of large numbers}
\DeclareAcronym{SoR}{short=SoR, long=subset of regressors}
\DeclareAcronym{SVG}{short=SVG, long=stochastic value gradients}
\DeclareAcronym{SVGD}{short=SVGD, long=Stein variational gradient descent}
\DeclareAcronym{SWA}{short=SWA, long=stochastic weight averaging}
\DeclareAcronym{SWAG}{short=SWAG, long=stochastic weight averaging-Gaussian}
\DeclareAcronym{Tanh}{short=Tanh, long=hyperbolic tangent}
\DeclareAcronym{TD}{short=TD, long=temporal difference}
\DeclareAcronym{TD3}{short=TD3, long=twin delayed DDPG}
\DeclareAcronym{TRPO}{short=TRPO, long=trust-region policy optimization}
\DeclareAcronym{UCB}{short=UCB, long=upper confidence bound}
\DeclareAcronym{ULA}{short=ULA, long=unadjusted Langevin algorithm}
\DeclareAcronym{VI1}{short=VI, long=value iteration}
\DeclareAcronym{VI2}{short=VI, long=variational inference}
\DeclareAcronym{WLLN}{short=WLLN, long=weak law of large numbers}

\acuseall

\title{Probabilistic \\ Artificial Intelligence}
\author{Andreas Krause, Jonas Hübotter}
\date{Fall 2024}
\publisher{Institute for Machine Learning \\ Department of Computer Science}
\input{pages/titlepage}

\begin{document}
  \frontmatter\pagenumbering{roman}

  \maketitle

  \begin{fullwidth}
~\vfill
\thispagestyle{empty}
\setlength{\parindent}{0pt}
\setlength{\parskip}{\baselineskip}

Compiled on \textbf{\today}.

\ifthenelse{\boolean{manuscript}}{
  \small\par This manuscript is based on the course \textsc{Probabilistic Artificial Intelligence}
  (263-5210-00L) at ETH Z\"urich.
}{
  \small\par This set of notes was written for the course \textsc{Probabilistic Artificial Intelligence}
  (263-5210-00L) at ETH Z\"urich.
  Distribution of these notes without the
  permission of the authors is prohibited.
}

\ifthenelse{\boolean{manuscript}}{
  \begin{small}
    \includegraphics[scale=0.5]{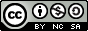}\\
    This work is licensed under a \href{http://creativecommons.org/licenses/by-nc-sa/4.0/}{Creative Commons Attribution-NonCommercial-ShareAlike 4.0 International License}.
  \end{small}
}{}

© \the\year{} ETH Zürich. All rights reserved.

\end{fullwidth}
\cleardoublepage

  \chapter*{Preface}

Artificial intelligence commonly refers to the science and engineering of artificial systems that can carry out tasks generally associated with requiring aspects of human intelligence, such as playing games, translating languages, and driving cars.
In recent years, there have been exciting advances in learning-based, data-driven approaches towards AI, and machine learning and deep learning have enabled computer systems to perceive the world in unprecedented ways.
Reinforcement learning has enabled breakthroughs in complex games such as Go and challenging robotics tasks such as quadrupedal locomotion.

A key aspect of intelligence is to not only make predictions, but reason about the {\em uncertainty} in these predictions, and to consider this uncertainty when making decisions.
This is what ``Probabilistic Artificial Intelligence'' is about.
The first part covers probabilistic approaches to machine learning.
We discuss the differentiation between ``epistemic'' uncertainty due to lack of data and ``aleatoric'' uncertainty, which is irreducible and stems, e.g., from noisy observations and outcomes.
We discuss concrete approaches towards probabilistic inference, such as Bayesian linear regression, Gaussian process models and Bayesian neural networks.
Often, inference and making predictions with such models is intractable, and we discuss modern approaches to efficient approximate inference.

The second part of the \course is about taking uncertainty into account in sequential decision tasks.
We consider active learning and Bayesian optimization --- approaches that collect data by proposing experiments that are informative for reducing the epistemic uncertainty.
We then consider reinforcement learning, a rich formalism for modeling agents that learn to act in uncertain environments.
After covering the basic formalism of Markov Decision Processes, we consider modern deep RL approaches that use neural network function approximation.
We close by discussing modern approaches in model-based RL, which harness epistemic and aleatoric uncertainty to guide exploration, while also reasoning about safety.

\section*{Guide to the Reader}

The material covered in this manuscript may support a one semester graduate introduction to probabilistic machine learning and sequential decision-making.
We welcome readers from all backgrounds.
However, we assume familiarity with basic concepts in probability, calculus, linear algebra, and machine learning (e.g., neural networks) as covered in a typical introductory course to machine learning.
In \cref{sec:fundamentals}, we give a gentle introduction to probabilistic inference, which serves as the foundation for the rest of the manuscript.
As part of this first chapter, we also review key concepts from probability theory.
We provide a chapter reviewing key concepts of further mathematical background in the back of the manuscript.

Throughout the manuscript, we focus on key concepts and ideas rather than their historical development.
We encourage you to consult the provided references for further reading and historical context to delve deeper into the covered topics.

\ifthenelse{\boolean{manuscript}}{}{
  Accompanying this manuscript, we provide an extensive set of curated examples as Jupyter notebooks which you can run and play with.
  You can find them at: \\
  \url{https://gitlab.inf.ethz.ch/OU-KRAUSE/pai-demos}.
}

Finally, we have included a set of exercises at the end of each chapter.
When we highlight an exercise throughout the text, we use this question mark: \exerciserefmark{properties_of_probability} --- so don't be surprised when you stumble upon it.
You will find solutions to all exercises in the back of the manuscript.

We hope you will find this resource useful.

\section*{Contributing}

We encourage you to raise issues and suggest fixes for anything you think can be improved.
We are thankful for any such feedback! \\
\textsc{Contact:} \href{mailto:pai-script@lists.inf.ethz.ch}{pai-script@lists.inf.ethz.ch}
\ifthenelse{\boolean{manuscript}}{}{\\ \textsc{Repository:} \url{https://gitlab.inf.ethz.ch/OU-KRAUSE/pai-script}}

\section*{Acknowledgements}

We are grateful to Sebastian Curi for creating the original Jupyter notebooks that accompany the course at ETH Zürich and which were instrumental in the creation of many figures.
We thank Hado van Hasselt for kindly contributing \cref{fig:maximization_bias}, and thank Tuomas Haarnoja~\citep{haarnoja2018soft} and Roberto Calandra~\citep{chua2018deep} for kindly agreeing to have their figures included in this manuscript.
Furthermore, many of the exercises in these notes are adapted from iterations of the course at ETH Zürich.
Special thanks to all instructors that contributed to the course material over the years.
We also thank all students of the course in the Fall of 2022, 2023, and 2024 who provided valuable feedback on various iterations of this manuscript and corrected many mistakes.
Finally, we thank Zhiyuan Hu, Shyam Sundhar Ramesh, Leander Diaz-Bone, Nicolas Menet, and Ido Hakimi for proofreading parts of various drafts of this text.

  \tableofcontents

  \mainmatter\setcounter{page}{1}\pagenumbering{arabic}

  \chapter{Fundamentals of Inference}\label{sec:fundamentals}

Boolean logic is the algebra of statements which are either true or false.
Consider, for example, the statements \begin{center}
  ``If it is raining, the ground is wet.'' \quad and \quad ``It is raining.''
\end{center}
A quite remarkable property of Boolean logic is that we can combine these premises to draw logical inferences which are \emph{new} (true) statements.
In the above example, we can conclude that the ground must be wet.
This is an example of logical reasoning which is commonly referred to as \midx{logical inference}, and the study of artificial systems that are able to perform logical inference is known as \midx{symbolic artificial intelligence}.\looseness=-1

But is it really raining?
Perhaps it is hard to tell by looking out of the window.
Or we have seen it rain earlier, but some time has passed since we have last looked out of the window.
And is it really true that if it rains, the ground is wet?
Perhaps the rain is just light enough that it is absorbed quickly, and therefore the ground still appears dry.

This goes to show that in our experience, the real world is rarely black and white.
We are frequently (if not usually) uncertain about the truth of statements, and yet we are able to reason about the world and make predictions.
We will see that the principles of Boolean logic can be extended to reason in the face of uncertainty.
The mathematical framework that allows us to do this is probability theory, which --- as we will find in this first chapter --- can be seen as a natural extension of Boolean logic from the domain of certainty to the domain of uncertainty.
In fact, in the 20th century, Richard Cox and Edwin Thompson Jaynes have done early work to formalize probability theory as the ``logic under uncertainty''~\citep{cox2001algebra,jaynes2003probability}.

In this first chapter, we will briefly recall the fundamentals of probability theory, and we will see how \midx{probabilistic inference} can be used to reason about the world.
In the remaining chapters, we will then discuss how probabilistic inference can be performed efficiently given limited computational resources and limited time, which is the key challenge in \midx{probabilistic artificial intelligence}.

\section{Probability}\label{sec:fundamentals:probability}

Probability is commonly interpreted in two different ways.
In the frequentist interpretation, one interprets the probability of an event (say a coin coming up ``heads'' when flipping it) as the limit of relative frequencies in repeated independent experiments. That is, \begin{align*}
    \text{Probability} = \lim_{N \to \infty} \frac{\text{\# events happening in $N$ trials}}{N}.
\end{align*}
This interpretation is natural, but has a few issues.
It is not very difficult to conceive of settings where repeated experiments do not make sense.
Consider the outcome: \begin{center}
  ``Person X will live for at least 80 years.''
\end{center}
There is no way in which we could conduct multiple independent experiments in this case.
Still, this statement is going to turn out either true or false, as humans we are just not able to determine its truth value beforehand.
Nevertheless, humans commonly have \emph{beliefs} about statements of this kind.
We also commonly reason about statements such as \begin{center}
  ``The Beatles were more groundbreaking than The Monkees.''
\end{center}
This statement does not even have an objective truth value, and yet we as humans tend to have opinions about it.

While it is natural to consider the relative frequency of the outcome in repeated experiments as our belief, if we are not able to conduct repeated experiments, our notion of probability is simply a subjective measure of uncertainty about outcomes.
In the early 20th century, Bruno De Finetti has done foundational work to formalize this notion which is commonly called \midx{Bayesian reasoning} or the Bayesian interpretation of probability \citep{de2017theory}.

We will see that modern approaches to probabilistic inference often lend themselves to a Bayesian interpretation, even if such an interpretation is not strictly necessary.
For our purposes, probabilities will be a means to an end: the end usually being solving some task.
This task may be to make a prediction or to take an action with an uncertain outcome, and we can evaluate methods according to how well they perform on this task.
No matter the interpretation, the mathematical framework of probability theory which we will formally introduce in the following is the same.

\subsection{Probability Spaces}

A probability space is a mathematical model for a random experiment.
The set of all possible outcomes of the experiment $\Omega$ is called \midx{sample space}.
An \midx{event} $\sA \subseteq \Omega$ of interest may be any combination of possible outcomes.
The set of all events $\spA \subseteq \pset{\Omega}$ that we are interested in is often called the \midx{event space} of the experiment.\footnote{We use $\pset{\Omega}$ to denote the \emph{power set} (set of all subsets) of $\Omega$.}
This set of events is required to be a $\sigma$-algebra over the sample space.

\begin{defn}[$\sigma$-algebra]\pidx{$\sigma$-algebra}
  Given the set $\Omega$, the set $\spA \subseteq \pset{\Omega}$ is a \emph{$\sigma$-algebra} over $\Omega$ if the following properties are satisfied: \begin{enumerate}
    \item $\Omega \in \spA$;
    \item if $\sA \in \spA$, then $\compl{\sA} \in \spA$ (\emph{closedness under complements}); and
    \item if we have $\sA_i \in \spA$ for all $i$, then $\bigcup_{i=1}^\infty \sA_i \in \spA$ (\emph{closedness under countable unions}).
  \end{enumerate}
\end{defn}

Note that the three properties of $\sigma$-algebras correspond to characteristics we universally expect when working with random experiments.
Namely, that we are able to reason about the event $\Omega$ that any of the possible outcomes occur, that we are able to reason about an event not occurring, and that we are able to reason about events that are composed of multiple (smaller) events.

\begin{ex}{Event space of throwing a die}{}
  The event space $\spA$ can also be thought of as ``how much information is available about the experiment''.
  For example, if the experiment is a throw of a die and $\Omega$ is the set of possible values on the die: ${\Omega = \{1, \dots, 6\}}$, then the following $\spA$ implies that the observer cannot distinguish between $1$ and $3$: \begin{align*}
    \spA \defeq \{\emptyset, \Omega, \{1, 3, 5\}, \{2, 4, 6\}\}.
  \end{align*}
  Intuitively, the observer only understands the parity of the face of the die.
\end{ex}

\begin{defn}[Probability measure]\pidx{probability measure}
  Given the set $\Omega$ and the $\sigma$-algebra $\spA$ over $\Omega$, the function \begin{align*}
    \fnPr : \spA \to \R
  \end{align*} is a \emph{probability measure} on $\spA$ if the \midx{Kolmogorov axioms} are satisfied: \begin{enumerate}
    \item $0 \leq \Pr{A} \leq 1$ for any $A \in \spA$;
    \item $\Pr{\Omega} = 1$; and
    \item $\Pr{\bigcup_{i=1}^\infty \sA_i} = \sum_{i=1}^\infty \Pr{\sA_i}$ for any countable set of mutually disjoint events $\{\sA_i \in \spA\}_i$.\footnote{We say that a set of sets $\{\sA_i\}_i$ is disjoint if for all $i \neq j$ we have $\sA_i \cap \sA_j = \emptyset$.}
  \end{enumerate}
\end{defn}
Remarkably, all further statements about probability follow from these three natural axioms.
For an event $\sA \in \spA$, we call $\Pr{\sA}$ the \midx{probability} of $\sA$.
We are now ready to define a probability space.

\begin{defn}[Probability space]\pidx{probability space}
  A \idx{probability space} is a triple $(\Omega, \spA, \fnPr)$ where \begin{itemize}
    \item $\Omega$ is a sample space,
    \item $\spA$ is a $\sigma$-algebra over $\Omega$, and
    \item $\fnPr$ is a probability measure on $\spA$.
  \end{itemize}
\end{defn}

\begin{ex}{Borel $\sigma$-algebra over $\R$}{}
  In our context, we often have that $\Omega$ is the set of real numbers $\R$ or a compact subset of it.
  In this case, a natural event space is the $\sigma$-algebra generated by the set of events \begin{align*}
    \sA_x \defeq \{x '\in \Omega : x' \leq x\}.
  \end{align*}
  The smallest $\sigma$-algebra $\spA$ containing all sets $A_x$ is called the \midx{Borel $\sigma$-algebra}.
  $\spA$ contains all ``reasonable'' subsets of $\Omega$ (except for some pathological examples).
  For example, $\spA$ includes all singleton sets $\{x\}$, as well as all countable unions of intervals.

  In the case of discrete $\Omega$, in fact $\spA=\pset{\Omega}$, i.e., the Borel $\sigma$-algebra contains {\em all} subsets of $\Omega$.
\end{ex}

\subsection{Random Variables}

The set $\Omega$ is often rather complex.
For example, take $\Omega$ to be the set of all possible graphs on $n$ vertices.
Then the outcome of our experiment is a graph.
Usually, we are not interested in a specific graph but rather a property such as the number of edges, which is shared by many graphs.
A function that maps a graph to its number of edges is a random variable.

\begin{defn}[Random variable]\pidx{random variable}
  A \idx{random variable} $X$ is a function \begin{align*}
    X : \Omega \to \spT
  \end{align*} where $\spT$ is called \midx{target space} of the random variable,\footnote{For a random variable that maps a graph to its number of edges, $\spT = \NatZ$. For our purposes, you can generally assume $\spT \subseteq \R$.} and where $X$ respects the information available in the $\sigma$-algebra $\spA$. That is,\footnote{In our example of throwing a die, $X$ should assign the same value to the outcomes $1,3,5$.} \begin{align}
    \forall S \subseteq \spT :\quad \{\omega \in \Omega : X(\omega) \in S\} \in \spA.
  \end{align}
\end{defn}

Concrete values $x$ of a random variable $X$ are often referred to as \midx<states>{state of a random variable} or \midx<realizations>{realizations of a random variable} of $X$.
The probability that $X$ takes on a value in $S \subseteq \spT$ is \begin{align}
  \Pr{X \in S} = \Pr{\{\omega \in \Omega : X(\omega) \in S\}}.
\end{align}

\subsection{Distributions}

Consider a random variable $X$ on a probability space $(\Omega, \spA, \fnPr)$, where $\Omega$ is a compact subset of $\R$, and $\spA$ the Borel $\sigma$-algebra.

In this case, we can refer to the probability that $X$ assumes a particular state or set of states by writing
\begin{align}
  p_X(x) &\defeq \Pr{X = x} \quad \text{(in the discrete setting)}, \\
  P_X(x) &\defeq \Pr{X \leq x}.
\end{align}
Note that ``$X = x$'' and ``$X \leq x$'' are merely events (that is, they characterize subsets of the sample space $\Omega$ satisfying this condition) which are in the Borel $\sigma$-algebra, and hence their probability is well-defined.

Hereby, $p_X$ and $P_X$ are referred to as the probability mass function (PMF) and cumulative distribution function (CDF) of $X$, respectively.
Note that we can also \emph{implicitly} define probability spaces through random variables and their associated PMF/CDF, which is often very convenient.\looseness=-1

We list some common examples of discrete distributions in \cref{sec:background:probability:common_discrete_distributions}.
Further, note that for continuous variables, $\Pr{X=x}=0$.
Here, instead we typically use the probability density function (PDF), to which we (with slight abuse of notation) also refer with $p_X$. We discuss densities in greater detail in \cref{sec:fundamentals:continuous_distributions}.\looseness=-1

We call the subset $\sS \subseteq \spT$ of the domain of a PMF or PDF $p_X$ such that all elements $x \in \sS$ have positive probability, $p_X(x) > 0$, the \midx{support} of the distribution $p_X$. This quantity is denoted by $X(\Omega)$.

\subsection{Continuous Distributions}\label{sec:fundamentals:continuous_distributions}

As mentioned, a continuous random variable can be characterized by its \midx{probability density function} (PDF).
But what is a density? We can derive some intuition from physics.

Let $\set{M}$ be a (non-homogeneous) physical object, e.g., a rock.
We commonly use $m(\set{M})$ and $\mathrm{vol}(\set{M})$ to refer to its mass and volume, respectively. Now, consider for a point $\vx \in \set{M}$ and a ball $B_r(\vx)$ around $\vx$ with radius $r$ the following quantities: \begin{align*}
  \lim_{r\to0} \mathrm{vol}(B_r(\vx)) = 0 \quad \lim_{r\to0} m(B_r(\vx)) = 0.
\end{align*} They appear utterly uninteresting at first, yet, if we divide them, we get what is called the \midx{density} of $\set{M}$ at $\vx$.
\begin{align*}
  \lim_{r\to0} \frac{m(B_r(\vx))}{\mathrm{vol}(B_r(\vx))} \eqdef \rho(\vx).
\end{align*}
We know that the relationship between density and mass is described by the following formula: \begin{align*}
  m(\set{M}) = \int_\set{M} \rho(\vx) \,d\vx.
\end{align*}
In other words, the density is to be integrated. For a small region $\set{I}$ around $\vx$, we can approximate $m(\set{I}) \approx \rho(\vx) \cdot \mathrm{vol}(\set{I})$.

Crucially, observe that even though the mass of any particular point $\vx$ is zero, i.e., ${m(\{\vx\}) = 0}$, assigning a density $\rho(\vx)$ to $\vx$ is useful for integration and approximation.
The same idea applies to continuous random variables, only that volume corresponds to intervals on the real line and mass to probability.
Recall that probability density functions are normalized such that their probability mass across the entire real line integrates to one.

\begin{marginfigure}[10\baselineskip]
  \begin{center}
    \import{./plots/output/}{normal_pdf.pgf}
  \end{center}

	\caption{PDF of the standard normal distribution. Observe that the PDF is symmetric around the mode.}
\end{marginfigure}

\begin{ex}{Normal distribution / Gaussian}{}\pidx{normal distribution}\pidx{Gaussian}%
  A famous example of a continuous distribution is the \emph{normal distribution}, also called \emph{Gaussian}.
  We say, a random variable $X$ is \emph{normally distributed}, ${X \sim \N{\mu}{\sigma^2}}$, if its PDF is \begin{align}
    \N[x]{\mu}{\sigma^2} \defeq \frac{1}{\sqrt{2\pi\sigma^2}} \exp\left(-\frac{(x - \mu)^2}{2 \sigma^2}\right). \label{eq:univ_normal}
  \end{align}
  We have ${\E{X} = \mu}$ and ${\Var{X} = \sigma^2}$. If ${\mu = 0}$ and ${\sigma^2 = 1}$, this distribution is called the \midx{standard normal distribution}.
  The Gaussian CDF cannot be expressed in closed-form.

  Note that the mean of a Gaussian distribution coincides with the maximizer of its PDF, also called \midx{mode} of a distribution.
\end{ex}

We will focus in the remainder of this chapter on continuous distributions, but the concepts we discuss extend mostly to discrete distributions simply by ``replacing integrals by sums''.

\subsection{Joint Probability}

A joint probability (as opposed to a marginal probability) is the probability of two or more events occurring simultaneously: \begin{align}
  \Pr{\sA, \sB} \defeq \Pr{\sA \cap \sB}.
\end{align}
In terms of random variables, this concept extends to joint distributions.
Instead of characterizing a single random variable, a \midx{joint distribution} is a function $p_{\rX} : \R^n \to \R$, characterizing a \midx{random vector} $\rX \defeq \transpose{[X_1 \; \cdots \;X_n]}$.
For example, if the $X_i$ are discrete, the joint distribution characterizes joint probabilities of the form \begin{align*}
  \Pr{\rX = [x_1, \dots, x_n]} = \Pr{X_1 = x_1, \dots, X_n = x_n},
\end{align*} and hence describes the relationship among all variables $X_i$.
For this reason, a joint distribution is also called a \midx{generative model}.
We use $X_{i:j}$ to denote the random vector $\transpose{[X_i \; \cdots \; X_j]}$.

We can ``sum out'' (respectively ``integrate out'') variables from a joint distribution in a process called ``marginalization''\pidx{marginalization}:

\begin{thmb}
  \begin{fct}[Sum rule]\pidx{sum rule} We have that
    \begin{align}
      p(x_{1:i-1}, x_{i+1:n}) &= \int_{X_i(\Omega)} p(x_{1:i-1}, x_i, x_{i+1:n}) \,dx_i. \label{eq:sum_rule}
    \end{align}
  \end{fct}
\end{thmb}

\subsection{Conditional Probability}

Conditional probability updates the probability of an event $\sA$ given some new information, for example, after observing the event $\sB$.

\begin{defn}[Conditional probability]\pidx{conditional probability}
  Given two events $\sA$ and $\sB$ such that $\Pr{\sB} > 0$, the probability of $\sA$ conditioned on $\sB$ is given as \begin{align}
    \Pr{\sA \mid \sB} \defeq \frac{\Pr{\sA, \sB}}{\Pr{\sB}}. \label{eq:cond_prob}
  \end{align}
\end{defn}

\begin{marginfigure}
  \incfig{conditional_probability}
	\caption{Conditioning an event $\sA$ on another event $\sB$ can be understood as replacing the universe of all possible outcomes $\Omega$ by the observed outcomes $\sB$.
  Then, the conditional probability is simply expressing the likelihood of $\sA$ given that $\sB$ occurred.}
\end{marginfigure}

Simply rearranging the terms yields, \begin{align}
  \Pr{\sA, \sB} = \Pr{\sA \mid \sB} \cdot \Pr{\sB} = \Pr{\sB \mid \sA} \cdot \Pr{\sA}. \label{eq:simple_product_rule}
\end{align}
Thus, the probability that both $\sA$ and $\sB$ occur can be calculated by multiplying the probability of event $\sA$ and the probability of $\sB$ conditional on $\sA$ occurring.

We say $\rZ \sim \rX \mid \rY = \vy$ (or simply $\rZ \sim \rX \mid \vy$) if $\rZ$ follows the \midx{conditional distribution} \begin{align}
  p_{\rX\mid\rY}(\vx \mid \vy) \defeq \frac{p_{\rX,\rY}(\vx, \vy)}{p_\rY(\vy)}. \label{eq:cond_distr}
\end{align}
If $\rX$ and $\rY$ are discrete, we have that $p_{\rX \mid \rY}(\vx \mid \vy) = \Pr{\rX = \vx \mid \rY = \vy}$ as one would naturally expect.

Extending \cref{eq:simple_product_rule} to arbitrary random vectors yields the product rule (also called the \midx{chain rule of probability}):

\begin{thmb}
  \begin{fct}[Product rule]\pidx{product rule}
    Given random variables $X_{1:n}$, \begin{align}
      p(x_{1:n}) = p(x_1) \cdot \prod_{i=2}^n p(x_i \mid x_{1:i-1}). \label{eq:product_rule}
    \end{align}
  \end{fct}
\end{thmb}

Combining sum rule and product rule, we can compute marginal probabilities too: \begin{align}
  p(\vx) = \int_{\rY(\Omega)} p(\vx, \vy) \,d\vy = \int_{\rY(\Omega)} p(\vx \mid \vy) \cdot p(\vy) \,d\vy \label{eq:lotp} \margintag{first using the sum rule \eqref{eq:sum_rule} then the product rule \eqref{eq:product_rule}}
\end{align}
This is called the \midx{law of total probability} (LOTP), which is colloquially often referred to as \midx{conditioning} on $\rY$.
If it is difficult to compute $p(\vx)$ directly, conditioning can be a useful technique when $\rY$ is chosen such that the densities $p(\vx \mid \vy)$ and $p(\vy)$ are straightforward to understand.

\subsection{Independence}

Two random vectors $\rX$ and $\rY$ are \midx<independent>{independence} (denoted $\rX \perp \rY$) if and only if knowledge about the state of one random vector does not affect the distribution of the other random vector, namely if their conditional CDF (or in case they have a joint density, their conditional PDF) simplifies to \begin{align}
  P_{\rX\mid\rY}(\vx \mid \vy) = P_\rX(\vx),\quad p_{\rX\mid\rY}(\vx \mid \vy) = p_\rX(\vx).
\end{align}
For the conditional probabilities to be well-defined, we need to assume that $p_\rY(\vy) > 0$.

The more general characterization of independence is that $\rX$ and $\rY$ are independent if and only if their joint CDF (or in case they have a joint density, their joint PDF) can be decomposed as follows: \begin{align}
  P_{\rX,\rY}(\vx,\vy) = P_\rX(\vx) \cdot P_\rY(\vy), \quad p_{\rX,\rY}(\vx,\vy) = p_\rX(\vx) \cdot p_\rY(\vy).
\end{align}
The equivalence of the two characterizations (when $p_\rY(\vy) > 0$) is easily proven using the product rule: $p_{\rX,\rY}(\vx,\vy) = p_\rY(\vy) \cdot p_{\rX\mid\rY}(\vx \mid \vy)$.

A ``weaker'' notion of independence is conditional independence.\footnote{We discuss in \cref{rmk:common_causes} how ``weaker'' is to be interpreted in this context.}
Two random vectors $\rX$ and $\rY$ are \midx<conditionally independent>{conditional independence} given a random vector $\rZ$ (denoted $\rX \perp \rY \mid \rZ$) iff, given $\rZ$, knowledge about the value of one random vector $\rY$ does not affect the distribution of the other random vector $\rX$, namely if \begin{subequations}\begin{align}
  P_{\rX \mid \rY, \rZ}(\vx \mid \vy, \vz) &= P_{\rX \mid \rZ}(\vx \mid \vz),\\
  p_{\rX \mid \rY, \rZ}(\vx \mid \vy, \vz) &= p_{\rX \mid \rZ}(\vx \mid \vz).
\end{align}\end{subequations}
Similarly to independence, we have that $\rX$ and $\rY$ are conditionally independent given $\rZ$ if and only if their joint CDF or joint PDF can be decomposed as follows: \begin{subequations}\begin{align}
  P_{\rX,\rY \mid \rZ}(\vx,\vy \mid \vz) &= P_{\rX \mid \rZ}(\vx \mid \vz) \cdot P_{\rY \mid \rZ}(\vy \mid \vz), \\
  p_{\rX,\rY \mid \rZ}(\vx,\vy \mid \vz) &= p_{\rX \mid \rZ}(\vx \mid \vz) \cdot p_{\rY \mid \rZ}(\vy \mid \vz).
\end{align}\end{subequations}

\begin{rmk}{Common causes}{common_causes}
  How can conditional independence be understood as a ``weaker'' notion of independence?
  Clearly, conditional independence does not imply independence: a trivial example is $X \perp X \mid X \centernot\implies X \perp X$.\safefootnote{$X \perp X \mid X$ is true trivially.}
  Neither does independence imply conditional independence: for example, $X \perp Y \centernot\implies X \perp Y \mid X + Y$.\safefootnote{Knowing $X$ and $X + Y$ already implies the value of $Y$, and hence, $X \not\perp Y \mid X + Y$.}

  When we say that conditional independence is a weaker notion we mean to emphasize that $X$ and $Y$ can be ``made'' (conditionally) independent by conditioning on the ``right'' $Z$ even if $X$ and $Y$ are dependent.
  This is known as \midx{Reichenbach's common cause principle} which says that for any two random variables $X \not\perp Y$ there exists a random variable $Z$ (which may be $X$ or $Y$) that causally influences both $X$ and $Y$, and which is such that $X \perp Y \mid Z$.
\end{rmk}

\subsection{Directed Graphical Models}\pidx{directed graphical model}%

Directed graphical models (also called \midx{Bayesian networks}) are often used to visually denote the (conditional) independence relationships of a large number of random variables.
They are a schematic representation of the factorization of the generative model into a product of conditional distributions as a directed acyclic graph.
Given the sequence of random variables $\{X_i\}_{i=1}^n$, their generative model can be expressed as \begin{align}
  p(x_{1:n}) = \prod_{i=1}^n p(x_i \mid \mathrm{parents}(x_i)) \label{eq:directed_graphical_model}
\end{align} where $\mathrm{parents}(x_i)$ is the set of parents of the vertex $X_i$ in the directed graphical model.
In other words, the parenthood relationship encodes a conditional independence of a random variable $X$ with a random variable $Y$ given their parents:\footnote[][-5\baselineskip]{More generally, vertices $u$ and $v$ are conditionally independent given a set of vertices $\sZ$ if $\sZ$ \idx<d-separates>{d-separation} $u$ and $v$, which we will not cover in depth here.} \begin{align}
    X \perp Y \mid \mathrm{parents}(X), \mathrm{parents}(Y).
\end{align}
\Cref{eq:directed_graphical_model} simply uses the product rule and the conditional independence relationships to factorize the generative model.
This can greatly reduce the model's complexity, i.e., the length of the product.

\begin{marginfigure}[-2\baselineskip]
  \incfig{directed_graphical_model}
  \caption{Example of a directed graphical model. The random variables $X_1, \dots, X_n$ are mutually independent given the random variable $Y$. The squared rectangular nodes are used to represent dependencies on parameters $c, a_1, \dots, a_n$.}\label{fig:directed_graphical_model}
\end{marginfigure}

An example of a directed graphical model is given in \cref{fig:directed_graphical_model}.
Circular vertices represent random quantities (i.e., random variables).
In contrast, square vertices are commonly used to represent deterministic quantities (i.e., parameters that the distributions depend on).
In the given example, we have that $X_i$ is conditionally independent of all other $X_j$ given $Y$.
\midx<Plate notation>{plate notation} is a condensed notation used to represent repeated variables of a graphical model.
An example is given in \cref{fig:plate_notation}.\looseness=-1

\begin{marginfigure}
  \incfig{plate_notation}
  \caption{The same directed graphical model as in \cref{fig:directed_graphical_model} using plate notation.}
  \label{fig:plate_notation}
\end{marginfigure}

\subsection{Expectation}

The \midx<expected value>{expectation} or \midx{mean} $\E{\rX}$ of a random vector $\rX$ is the (asymptotic) arithmetic mean of an arbitrarily increasing number of independent realizations of $\rX$. That is,\footnote{In infinite probability spaces, absolute convergence of $\E{\rX}$ is necessary for the existence of $\E{\rX}$.} \begin{align}
  \E{\rX} &\defeq \int_{\rX(\Omega)} \vx \cdot p(\vx) \,d\vx \label{eq:expectation}
\end{align}
A very special and often used property of expectations is their \midx<linearity>{linearity of expectation}, namely that for any random vectors $\rX$ and $\rY$ in $\R^n$ and any ${\mA \in \R^{m \times n}}, {\vb \in \R^m}$ it holds that \begin{align}
  \E{\mA \rX + \vb} = \mA \E{\rX} + \vb \quad\text{and}\quad \E{\rX + \rY} = \E{\rX} + \E{\rY}. \label{eq:linearity_expectation}
\end{align}
Note that $\rX$ and $\rY$ do not necessarily have to be independent!
Further, if $\rX$ and $\rY$ are independent then \begin{align}
  \E{\rX\transpose{\rY}} = \E{\rX} \cdot \transpose{\E{\rY}}. \label{eq:expectation_product}
\end{align}

The following intuitive lemma can be used to compute expectations of transformed random variables.

\begin{thmb}
  \begin{fct}[Law of the unconscious statistician, LOTUS]\pidx{law of the unconscious statistician}
    \begin{align}
      \E{\vg(\rX)} = \int_{\rX(\Omega)} \vg(\vx) \cdot p(\vx) \,d\vx \label{eq:lotus}
    \end{align} where $\vg : \rX(\Omega) \to \R^n$ is a ``nice'' function\safefootnote{$\vg$ being a continuous function, which is either bounded or absolutely integrable (i.e., $\int \abs{g(\vx)} p(\vx) \, d\vx < \infty$), is sufficient. This is satisfied in most cases.} and $\rX$ is a continuous random vector.
    The analogous statement with a sum replacing the integral holds for discrete random variables.
  \end{fct}
\end{thmb}

This is a nontrivial fact that can be proven using the change of variables formula which we discuss in \cref{sec:fundamentals:probability:cov}.

Similarly to conditional probability, we can also define conditional expectations.
The expectation of a continuous random vector $\rX$ given that $\rY = \vy$ is defined as \begin{align}
  \E{\rX \mid \rY = \vy} \defeq \int_{\rX(\Omega)} \vx \cdot p_{\rX \mid \rY}(\vx \mid \vy) \,d\vx.
\end{align}
Observe that $\E{\rX \mid \rY = {\cdot}}$ defines a deterministic mapping from $\vy$ to $\E{\rX \mid \rY = \vy}$.
Therefore, $\E{\rX \mid \rY}$ is itself a random vector: \begin{align}
  \E{\rX \mid \rY}(\omega) = \E{\rX \mid \rY = \rY(\omega)}
\end{align} where $\omega \in \Omega$.
This random vector $\E{\rX \mid \rY}$ is called the \midx{conditional expectation} of $\rX$ given $\rY$.

Analogously to the law of total probability \eqref{eq:lotp}, one can condition an expectation on another random vector.
This is known as the \midx{tower rule} or the \midx{law of total expectation} (LOTE):

\begin{thmb}
  \begin{thm}[Tower rule]
    Given random vectors $\rX$ and $\rY$, we have \begin{align}
      \E[\rY]{\E[\rX]{\rX \mid \rY}} = \E{\rX}. \label{eq:tower_rule}
    \end{align}
  \end{thm}
\end{thmb}
\begin{proof}[Proof sketch]
  We only prove the case where $\rX$ and $\rY$ have a joint density.
  We have\looseness=-1 \begin{align*}
    \E{\E{\rX \mid \rY}} &= \int \left(\int \vx \cdot p(\vx \mid \vy) \,d\vx\right) p(\vy) \,d\vy \\
    &= \int \int \vx \cdot p(\vx, \vy) \,d\vx \,d\vy \margintag{by definition of conditional densities \eqref{eq:cond_distr}} \\
    &= \int \vx \int p(\vx, \vy) \,d\vy \,d\vx \margintag{by Fubini's theorem} \\
    &= \int \vx \cdot p(\vx) \,d\vx \margintag{using the sum rule \eqref{eq:sum_rule}} \\
    &= \E{\rX}. \qedhere
  \end{align*}
\end{proof}

\subsection{Covariance and Variance}\label{ssec:cov}

Given two random vectors $\rX$ in $\R^n$ and $\rY$ in $\R^m$, their \midx{covariance} is defined as \begin{align}
  \Cov{\rX, \rY} &\defeq \E{(\rX - \E{\rX})\transpose{(\rY - \E{\rY})}} \label{eq:covariance} \\
  &= \E{\rX \transpose{\rY}} - \E{\rX} \cdot \transpose{\E{\rY}} \\
  &= \transpose{\Cov{\rY, \rX}} \in \R^{n \times m}.
\end{align}
Covariance measures the linear dependence between two random vectors since a direct consequence of its definition \eqref{eq:covariance} is that given linear maps $\mA \in \R^{n' \times n}, \mB \in \R^{m' \times m}$, vectors $\vc \in \R^{n'}, \vd \in \R^{m'}$ and random vectors $\rX$ in $\R^n$ and $\rY$ in $\R^m$, we have that \begin{align}
  \Cov{\mA\rX + \vc, \mB\rY + \vd} = \mA\Cov{\rX,\rY}\transpose{\mB}. \label{eq:linear_map_covariance}
\end{align}
Two random vectors $\rX$ and $\rY$ are said to be \midx{uncorrelated} if and only if $\Cov{\rX, \rY} = \mzero$.
Note that if $\rX$ and $\rY$ are independent, then \cref{eq:expectation_product} implies that $\rX$ and $\rY$ are uncorrelated.
The reverse does not hold in general.

\begin{rmk}{Correlation}{correlation}
  The \idx{correlation} of the random vectors $\rX$ and $\rY$ is a normalized covariance, \begin{align}
    \Cor{\rX,\rY}(i,j) \defeq \frac{\Cov{X_i,Y_j}}{\sqrt{\Var{X_i} \Var{Y_j}}} \in [-1,1]. \label{eq:correlation}
  \end{align}
  Two random vectors $\rX$ and $\rY$ are therefore uncorrelated if and only if $\Cor{\rX,\rY} = \mzero$.

  There is also a nice geometric interpretation of covariance and correlation.
  For zero mean random variables $X$ and $Y$, $\Cov{X,Y}$ is an inner product.\safefootnote{That is, \begin{itemize}
    \item $\Cov{X,Y}$ is symmetric,
    \item $\Cov{X,Y}$ is linear (here we use $\E*{X} = \E*{Y} = 0$), and
    \item $\Cov{X,X} \geq 0$.
  \end{itemize}}

  The cosine of the angle $\theta$ between $X$ and $Y$ (that are not deterministic) coincides with their correlation, \begin{align}
    \cos\theta = \frac{\Cov{X,Y}}{\norm{X}\norm{Y}} = \Cor{X,Y}. \margintag{using the Euclidean inner product formula, $\Cov{X, Y} = \norm{X} \norm{Y} \cos\theta$}
  \end{align}
  $\cos\theta$ is also called a \midx{cosine similarity}.
  Thus, \begin{align}
    \theta = \arccos\,\Cor{X,Y}.
  \end{align}
  For example, if $X$ and $Y$ are uncorrelated, then they are orthogonal in the inner product space.
  If $\Cor{X, Y} = -1$ then $\theta \equiv \pi$ (that is, $X$ and $Y$ ``point in opposite directions''), whereas if $\Cor{X, Y} = 1$ then $\theta \equiv 0$ (that is, $X$ and $Y$ ``point in the same direction'').
\end{rmk}

The covariance of a random vector $\rX$ in $\R^n$ with itself is called its \midx{variance}: \begin{align}
  \Var{\rX} &\defeq \Cov{\rX, \rX} \\
  &= \E{(\rX - \E{\rX})\transpose{(\rX - \E{\rX})}} \label{eq:variance} \\
  &= \E{\rX \transpose{\rX}} - \E{\rX} \cdot \transpose{\E{\rX}} \label{eq:variance2} \\
  &= \begin{bmatrix}
    \Cov{X_1, X_1} & \cdots & \Cov{X_1, X_n} \\
    \vdots & \ddots & \vdots \\
    \Cov{X_n, X_1} & \cdots & \Cov{X_n, X_n} \\
  \end{bmatrix} \label{eq:covariance_matrix}.
\end{align}
The scalar variance $\Var{X}$ of a random variable $X$ is a measure of uncertainty about the value of $X$ since it measures the average squared deviation from $\E{X}$.
We will see that the eigenvalue spectrum of a covariance matrix can serve as a measure of uncertainty in the multivariate setting.\footnote{The \emph{multivariate} setting (as opposed to the \emph{univariate} setting) studies the joint distribution of multiple random variables.}

\begin{rmk}{Standard deviation}{}
  The length of a random variable $X$ in the inner product space described in \cref{rmk:correlation} is called its \midx{standard deviation}, \begin{align}
    \norm{X} = \sqrt{\Cov{X,X}} = \sqrt{\Var{X}} \eqdef \SD{X}.
  \end{align}
  That is, the longer a random variable is in the inner product space, the more ``uncertain'' we are about its value.
  If a random variable has length $0$, then it is deterministic.
\end{rmk}

The variance of a random vector $\rX$ is also called the \midx{covariance matrix} of $\rX$ and denoted by $\mSigma_{\rX}$ (or $\mSigma$ if the correspondence to $\rX$ is clear from context).
A covariance matrix is symmetric by definition due to the symmetry of covariance, and is always positive semi-definite \exerciserefmark{cov_mat_pos_sd}.

Two useful properties of variance are the following: \begin{itemize}
  \item It follows from \cref{eq:linear_map_covariance} that for any linear map $\mA \in \R^{m \times n}$ and vector $\vb \in \R^m$, \begin{align}
    \Var{\mA\rX + \vb} = \mA\Var{\rX}\transpose{\mA}. \label{eq:linear_map_variance}
  \end{align}
  In particular, $\Var{-\rX} = \Var{\rX}$.

  \item It follows from the definition of variance \eqref{eq:variance} that for any two random vectors $\rX$ and $\rY$, \begin{align}
    \Var{\rX + \rY} = \Var{\rX} + \Var{\rY} + 2 \Cov{\rX, \rY}. \label{eq:sum_variance}
  \end{align}
  In particular, if $\rX$ and $\rY$ are independent then the covariance term vanishes and $\Var{\rX + \rY} = \Var{\rX} + \Var{\rY}$.
\end{itemize}

Analogously to conditional probability and conditional expectation, we can also define conditional variance. The \midx{conditional variance} of a random vector $\rX$ given another random vector $\rY$ is the random vector \begin{align}
  \Var{\rX \mid \rY} \defeq \E{(\rX - \E{\rX \mid \rY}) \transpose{(\rX - \E{\rX \mid \rY})}}[\rY].
\end{align}
Intuitively, the conditional variance is the remaining variance when we use $\E{\rX \mid \rY}$ to predict $\rX$ rather than if we used $\E{\rX}$.
One can also condition a variance on another random vector, analogously to the laws of total probability \eqref{eq:lotp} and expectation \eqref{eq:tower_rule}.

\begin{thmb}
  \begin{thm}[Law of total variance, LOTV]\pidx{law of total variance}\label{thm:lotv}
    \begin{align}
      \Var{\rX} = \E[\rY]{\Var[\rX]{\rX \mid \rY}} + \Var[\rY]{\E[\rX]{\rX \mid \rY}}. \label{eq:lotv}
    \end{align}
  \end{thm}
\end{thmb}

Here, the first term measures the average deviation from the mean of $\rX$ across realizations of $\rY$ and the second term measures the uncertainty in the mean of $\rX$ across realizations of $\rY$.
In \cref{sec:blr:uncertainty}, we will see that both terms have a meaningful characterization in the context of probabilistic inference.

\vspace{-10pt}\begin{proof}[Proof sketch of LOTV]
  To simplify the notation, we present only a proof for the univariate setting.
  \begin{align*}
    \Var{X} &= \E{X^2} - \E{X}^2 \\
    &= \E{\E{X^2 \mid Y}} - \E{\E{X \mid Y}}^2 \margintag{by the tower rule \eqref{eq:tower_rule}} \\
    &= \E{\Var{X \mid Y} + \E{X \mid Y}^2} - \E{\E{X \mid Y}}^2 \margintag{by the definition of variance \eqref{eq:variance2}} \\
    &= \E{\Var{X \mid Y}} + \parentheses*{\E{\E{X \mid Y}^2} - \E{\E{X \mid Y}}^2} \\
    &= \E{\Var{X \mid Y}} + \Var{\E{X \mid Y}}. \qedhere \margintag{by the definition of variance \eqref{eq:variance2}}
  \end{align*}
\end{proof}

\subsection{Change of Variables}\label{sec:fundamentals:probability:cov}

It is often useful to understand the distribution of a transformed random variable $Y = g(X)$ that is defined in terms of a random variable $X$, whose distribution is known.
Let us first consider the univariate setting. We would like to express the distribution of $Y$ in terms of the distribution of $X$, that is, we would like to find \begin{align}
    P_Y(y) = \Pr{Y \leq y} = \Pr{g(X) \leq y} = \Pr{X \leq \inv{g}(y)}.
\end{align}
When the random variables are continuous, this probability can be expressed as an integration over the domain of $X$.
We can then use the substitution rule of integration to ``change the variables'' to an integration over the domain of $Y$.
Taking the derivative yields the density $p_Y$.\footnote{The full proof of the change of variables formula in the univariate setting can be found in section 6.7.2 of \icite{mml}.}
There is an analogous change of variables formula for the multivariate setting.

\begin{thmb}
  \begin{fct}[Change of variables formula]\pidx{change of variables formula}
    Let $\rX$ be a random vector in $\R^n$ with density $p_\rX$ and let $\vg : \R^n \to \R^n$ be a differentiable and invertible function.
    Then $\rY = \vg(\rX)$ is another random variable, whose density can be computed based on $p_\rX$ and $\vg$ as follows: \begin{align}
      p_\rY(\vy) = p_\rX(\inv{\vg}(\vy)) \cdot \abs{\det{\jac \inv{\vg}(\vy)}} \label{eq:change_of_variables}
    \end{align} where $\jac \inv{\vg}(\vy)$ is the Jacobian of $\inv{\vg}$ evaluated at $\vy$.
  \end{fct}
\end{thmb}

Here, the term $\abs{\det{\jac \inv{\vg}(\vy)}}$ measures how much a unit volume changes when applying $\vg$.
Intuitively, the change of variables swaps the coordinate system over which we integrate.
The factor $\abs{\det{\jac \inv{\vg}(\vy)}}$ corrects for the change in volume that is caused by this change in coordinates.

Intuitively, you can think of the vector field $\vg$ as a perturbation to $\rX$, ``pushing'' the probability mass around.
The perturbation of a density $p_\rX$ by $\vg$ is commonly denoted by the \midx{pushforward} \begin{align}
  \pf{\vg}{p_\rX} \defeq p_\rY \quad\text{where $\rY = \vg(\rX)$}.
\end{align}
This concludes our quick tour of probability theory, and we are well-prepared to return to the topic of probabilistic inference.

\section{Probabilistic Inference}

Recall the logical implication ``If it is raining, the ground is wet.'' from the beginning of this chapter.
Suppose that we look outside a window and see that it is not raining: will the ground be dry?
Logical reasoning does not permit drawing an inference of this kind, as there might be reasons other than rain for which the ground could be wet (e.g., sprinklers).
However, intuitively, by observing that it is not raining, we have just excluded the possibility that the ground is wet because of rain, and therefore we would deem it ``more likely'' that the ground is dry than before.
In other words, if we were to walk outside now and the ground was wet, we would be more surprised than we would have been if we had not looked outside the window before.

As humans, we are constantly making such ``plausible'' inferences of our beliefs: be it about the weather, the outcomes of our daily decisions, or the behavior of others.
\midx<Probabilistic inference>{probabilistic inference}[idxpagebf] is the process of updating such a prior belief $\Pr{\overline{W}}$ to a posterior belief $\Pr{\overline{W} \mid \overline{R}}$ upon observing $\overline{R}$ where --- to reduce clutter --- we write $W$ for ``The ground is wet'' and $R$ for ``It is raining''.

The central principle of probabilistic inference is Bayes' rule:

\begin{thmb}
  \begin{thm}[Bayes' rule]\pidx{Bayes' rule}
    Given random vectors $\rX$ in $\R^n$ and $\rY$ in $\R^m$, we have for any $\vx \in \R^n, \vy \in \R^m$ that
    \begin{align}
      p(\vx \mid \vy) = \frac{p(\vy \mid \vx) \cdot p(\vx)}{p(\vy)}. \label{eq:bayes_rule}
    \end{align}
  \end{thm}
\end{thmb}
\begin{proof}
  Bayes' rule is a direct consequence of the definition of conditional densities \eqref{eq:cond_distr} and the product rule \eqref{eq:product_rule}.
\end{proof}

Let us consider the meaning of each term separately: \begin{itemize}
  \item the \midx{prior} $p(\vx)$ is the initial belief about $\vx$,
  \item the \midx<(conditional) likelihood>{likelihood}\pidx{conditional likelihood} $p(\vy \mid \vx)$ describes how likely the observations $\vy$ are under a given value $\vx$,
  \item the \midx{posterior} $p(\vx \mid \vy)$ is the updated belief about $\vx$ after observing $\vy$,
  \item the \midx{joint likelihood} $p(\vx, \vy) = p(\vy \mid \vx) p(\vx)$ combines prior and likelihood,
  \item the \midx{marginal likelihood} $p(\vy)$ describes how likely the observations $\vy$ are across all values of $\vx$.
\end{itemize}
The marginal likelihood can be computed using the sum rule \eqref{eq:sum_rule} or the law of total probability \eqref{eq:lotp}, \begin{align}
  p(\vy) = \int_{\rX(\Omega)} p(\vy \mid \vx) \cdot p(\vx) \,d\vx. \label{eq:marginal_likelihood}
\end{align}
Note, however, that the marginal likelihood is simply normalizing the conditional distribution to integrate to one, and therefore a constant with respect to $\vx$.
For this reason, $p(\vy)$ is commonly called the \midx{normalizing constant}.

\begin{ex}{Plausible inferences}{weak_probabilistic_inference}
  Let us confirm our intuition from the above example.
  The logical implication ``If it is raining, the ground is wet.'' (denoted $R \rightarrow W$) can be succinctly expressed as $\Pr{W \mid R} = 1$.
  Since $\Pr{W} \leq 1$, we know that \begin{align*}
    \Pr{R \mid W} = \frac{\Pr{W \mid R} \cdot \Pr{R}}{\Pr{W}} = \frac{\Pr{R}}{\Pr{W}} \geq \Pr{R}.
  \end{align*}
  That is, observing that the ground is wet makes it more likely to be raining.
  From $\Pr{R \mid W} \geq \Pr{R}$ we know $\Pr{\overline{R} \mid W} \leq \Pr{\overline{R}}$,\safefootnote{since $\Pr{\overline{X}} = 1 - \Pr{X}$} which leads us to follow that \begin{align*}
    \Pr{W \mid \overline{R}} = \frac{\Pr{\overline{R} \mid W} \cdot \Pr{W}}{\Pr{\overline{R}}} \leq \Pr{W},
  \end{align*} that is, having observed it not to be raining made the ground less likely to be wet.
\end{ex}

\Cref{ex:weak_probabilistic_inference} is called a \midx{plausible inference} because the observation of $\overline{R}$ does not completely determine the truth value of $\overline{W}$, and hence, does not permit logical inference.
In the case, however, that logical inference is permitted, it coincides with probabilistic inference.

\begin{ex}{Logical inferences}{strong_inference}
  For example, if we were to observe that the ground is not wet, then logical inference implies that it must not be raining: $\overline{W} \rightarrow \overline{R}$.
  This is called the \midx{contrapositive} of $R \rightarrow W$.

  Indeed, by probabilistic inference, we obtain analogously \begin{align*}
    \Pr{R \mid \overline{W}} = \frac{\Pr{\overline{W} \mid R} \cdot \Pr{R}}{\Pr{\overline{W}}} = \frac{(1 - \Pr{W \mid R}) \cdot \Pr{R}}{\Pr{\overline{W}}} = 0. \margintag{as $\Pr{W \mid R} = 1$}
  \end{align*}
\end{ex}

Observe that a logical inference does not depend on the prior $\Pr{R}$:
Even if the prior was $\Pr{R} = 1$ in \cref{ex:strong_inference}, after observing that the ground is not wet, we are forced to conclude that it is not raining to maintain logical consistency.
The examples highlight that while \emph{logical} inference does not require the notion of a prior, plausible (\emph{probabilistic}!) inference does.

\subsection{Where do priors come from?}\label{sec:fundamentals:inference:priors}

Bayes' rule necessitates the specification of a prior $p(\vx)$.
Different priors can lead to the deduction of dramatically different posteriors, as one can easily see by considering the extreme cases of a prior that is a point density at $\vx = \vx_0$ and a prior that is ``uniform'' over $\R^n$.\footnote{The latter is not a valid probability distribution, but we can still derive meaning from the posterior as we discuss in \cref{rmk:improper_priors}.}
In the former case, the posterior will be a point density at $\vx_0$ regardless of the likelihood.
In other words, no evidence can alter the ``prior belief'' the learner ascribed to $\vx$.
In the latter case, the learner has ``no prior belief'', and therefore the posterior will be proportional to the likelihood.
Both steps of probabilistic inference are perfectly valid, though one might debate which prior is more reasonable.

Someone who follows the Bayesian interpretation of probability might argue that everything is conditional, meaning that the prior is simply a posterior of all former observations.
While this might seem natural (``my world view from today is the combination of my world view from yesterday and the observations I made today''), this lacks an explanation for ``the first day''.
Someone else who is more inclined towards the frequentist interpretation might also object to the existence of a prior belief altogether, arguing that a prior is \emph{subjective} and therefore not a valid or desirable input to a learning algorithm.
Put differently, a frequentist ``has the belief not to have any belief''.
This is perfectly compatible with probabilistic inference, as long as the prior is chosen to be \midx<noninformative>{noninformative prior}\pidx{informative prior}:\begin{align}
  p(\vx) \propto \const.
\end{align}
Choosing a noninformative prior in the absence of any evidence is known as the \midx{principle of indifference} or the \midx{principle of insufficient reason}, which dates back to the famous mathematician Pierre-Simon Laplace.

\begin{ex}{Why be indifferent?}{}
  Consider a criminal trial with three suspects, A, B, and C.
  The collected evidence shows that suspect C can not have committed the crime, however it does not yield any information about suspects A and B.
  Clearly, any distribution respecting the data must assign zero probability of having committed the crime to suspect C.
  However, any distribution interpolating between $(1,0,0)$ and $(0,1,0)$ respects the data.
  The principle of indifference suggests that the desired distribution is $(\frac{1}{2}, \frac{1}{2}, 0)$, and indeed, any alternative distribution seems unreasonable.
\end{ex}

\begin{rmk}{Noninformative and improper priors}{improper_priors}
  It is not necessarily required that the prior $p(\vx)$ is a valid distribution (i.e., integrates to $1$).
  Consider for example, the noninformative prior $p(\vx) \propto \Ind{\vx \in I}$ where $I \subseteq \R^n$ is an infinitely large interval.
  Such a prior which is not a valid distribution is called an \midx{improper prior}.
  We can still derive meaning from the posterior of a given likelihood and (improper) prior as long as the posterior is a valid distribution.
\end{rmk}

Laplace's principle of indifference can be generalized to cases where \emph{some} evidence is available.
The \midx{maximum entropy principle}, originally proposed by \cite{jaynes1968prior}, states that one should choose as prior from all possible distributions that are \emph{consistent} with prior knowledge, the one that makes the \emph{least} ``additional assumptions'', i.e., is the least ``informative''.
In philosophy, this principle is known as \midx{Occam's razor} or the \midx{principle of parsimony}.
The ``informativeness'' of a distribution $p$ is quantified by its \idx{entropy} which is defined as \begin{align}
  \H{p} \defeq \E[\vx \sim p]{- \log p(\vx)}. \label{eq:mep_entropy}
\end{align}
The more concentrated $p$ is, the less is its entropy; the more diffuse $p$ is, the greater is its entropy.\footnote{We give a thorough introduction to entropy in \cref{sec:approximate_inference:information_theory}.}

In the absence of any prior knowledge, the uniform distribution has the highest entropy,\footnote{This only holds true when the set of possible outcomes of $\vx$ finite (or a bounded continuous interval), as in this case, the noninformative prior is a proper distribution --- the uniform distribution. In the ``infinite case'', there is no uniform distribution and the noninformative prior can be attained from the maximum entropy principle as the limiting solution as the number of possible outcomes of $\vx$ is increased.} and hence, the maximum entropy principle suggests a noninformative prior (as does Laplace's principle of indifference).
In contrast, if the evidence perfectly determines the value of~$\vx$, then the only consistent explanation is the point density at~$\vx$.
The maximum entropy principle characterizes a reasonable choice of prior for these two extreme cases and all cases in between.
Bayes' rule can in fact be derived as a consequence of the maximum entropy principle in the sense that the posterior is the least ``informative'' distribution among all distributions that are consistent with the prior and the observations \exerciserefmark{mep_and_posteriors}.

\subsection{Conjugate Priors}\label{sec:fundamentals:bayesian:conjugacy}

If the prior $p(\vx)$ and posterior $p(\vx \mid \vy)$ are of the same family of distributions, the prior is called a \midx{conjugate prior} to the likelihood $p(\vy \mid \vx)$.
This is a very desirable property, as it allows us to recursively apply the same learning algorithm implementing probabilistic inference.
We will see in \cref{sec:blr} that under some conditions the Gaussian is \midx<self-conjugate>{self-conjugacy}.
That is, if we have a Gaussian prior and a Gaussian likelihood then our posterior will also be Gaussian.
This will provide us with the first \emph{efficient} implementation of probabilistic inference.

\begin{ex}{Conjugacy of beta and binomial distribution}{}
  As an example for conjugacy, we will show that the beta distribution is a conjugate prior to a binomial likelihood.
  Recall the PMF of the binomial distribution \begin{align}
    \Bin[k]{n}{\theta} = {n \choose k} \theta^k (1-\theta)^{n-k} \label{eq:binomial_distr}
  \end{align} and the PDF of the beta distribution, \begin{align}
    \Beta[\theta]{\alpha}{\beta} \propto \theta^{\alpha-1} (1-\theta)^{\beta-1}, \label{eq:beta_distr}
  \end{align}
  We assume the prior ${\theta \sim \Beta{\alpha}{\beta}}$ and likelihood ${k \mid \theta \sim \Bin{n}{\theta}}$.
  Let $n_H = k$ be the number of heads and $n_T = n - k$ the number of tails in the binomial trial $k$.
  Then, \begin{align*}
    p(\theta \mid k) &\propto p(k \mid \theta) p(\theta) \margintag{using Bayes' rule \eqref{eq:bayes_rule}} \\
    &\propto \theta^{n_H} (1-\theta)^{n_T} \theta^{\alpha-1} (1-\theta)^{\beta-1} \\
    &= \theta^{\alpha+n_H-1} (1-\theta)^{\beta+n_T-1}.
  \end{align*}
  Thus, $\theta \mid k \sim \Beta{\alpha + n_H}{\beta + n_T}$.

  This same conjugacy can be shown for the multivariate generalization of the beta distribution, the \midx{Dirichlet distribution}, and the multivariate generalization of the binomial distribution, the \midx{multinomial distribution}.
\end{ex}

\subsection{Tractable Inference with the Normal Distribution}\label{sec:fundamentals:gaussians}

Using arbitrary distributions for learning and inference is computationally very expensive when the number of dimensions is large --- even in the discrete setting.
For example, computing marginal distributions using the sum rule yields an exponentially long sum in the size of the random vector.
Similarly, the normalizing constant of the conditional distribution is a sum of exponential length.
Even to represent any discrete joint probability distribution requires space that is exponential in the number of dimensions (cf. \cref{tab:joint_distribution}).

\begin{marginfigure}
  \begin{tabular}{ccccr}
    \toprule
    $X_1$ & $\cdots$ & $X_{n-1}$ & $X_n$ & $\Pr{X_{1:n}}$ \\
    \midrule
    $0$ & $\cdots$ & $0$ & $0$ & $0.01$ \\
    $0$ & $\cdots$ & $0$ & $1$ & $0.001$ \\
    $0$ & $\cdots$ & $1$ & $0$ & $0.213$ \\
    $\vdots$ & $\vdots$ & $\vdots$ & $\vdots$ & \\
    $1$ & $\cdots$ & $1$ & $1$ & $0.0003$ \\
    \bottomrule
  \end{tabular}\vspace{1em}
  \caption{A table representing a joint distribution of $n$ binary random variables. The table has $2^n$ rows. The number of parameters is $2^n - 1$ since the final probability is determined by all other probabilities as they must sum to one.}\label{tab:joint_distribution}
\end{marginfigure}

One strategy to get around this computational blowup is to restrict the class of distributions.
Gaussians are a popular choice for this purpose since they have extremely useful properties: they have a compact representation and --- as we will see in \cref{sec:blr} --- they allow for closed-form probabilistic inference.

In \cref{eq:univ_normal}, we have already seen the PDF of the univariate Gaussian distribution.
A random vector $\rX$ in $\R^n$ is \emph{normally distributed}, $\rX~\sim~\N{\vmu}{\mSigma}$, if its PDF is \begin{thmb}\begin{align}\pidx{normal distribution}[idxpagebf]\pidx{Gaussian}[idxpagebf]
  \N[\vx]{\vmu}{\mSigma} \defeq \frac{1}{\sqrt{\det{2\pi\mSigma}}} \exp\parentheses*{-\frac{1}{2}\transpose{(\vx-\vmu)} \inv{\mSigma} (\vx-\vmu)} \label{eq:normal}
\end{align}\end{thmb} where $\vmu \in \R^n$ is the mean vector and $\mSigma \in \R^{n \times n}$ the covariance matrix \exerciserefmark{expectation_and_variance_of_gaussians}.
We call $\mLambda \defeq \inv{\mSigma}$ the \midx{precision matrix}.
$\rX$ is also called a \midx{Gaussian random vector} (GRV).
$\SN$ is the multivariate \midx{standard normal distribution}[idxpagebf].
We call a Gaussian \midx<isotropic>{isotropic Gaussian} if its covariance matrix is of the form $\mSigma~=~\sigma^2\mI$ for some $\sigma^2 \in \R$.
In this case, the sublevel sets of the PDF are perfect spheres as can be seen in \cref{fig:multivariate_normal}.

\begin{figure}
  \begin{center}
    \import{./plots/output/}{multivariate_gaussian.pgf}
  \end{center}

  \caption{Shown are the PDFs of two-dimensional Gaussians with mean $\vzero$ and covariance matrices \begin{equation*}
    \mSigma_1 \defeq \begin{bmatrix}
      1 & 0 \\
      0 & 1
    \end{bmatrix}, \quad \mSigma_2 \defeq \begin{bmatrix}
      1 & 0.9 \\
      0.9 & 1
    \end{bmatrix}
  \end{equation*} respectively.}\label{fig:multivariate_normal}
\end{figure}

Note that a Gaussian can be represented using only $\BigO{n^2}$ parameters.
In the case of a diagonal covariance matrix, which corresponds to $n$ independent univariate Gaussians \exerciserefmark{grv_uncor_indep}, we just need $\BigO{n}$ parameters.

In \cref{eq:normal}, we assume that the covariance matrix $\mSigma$ is invertible, i.e., does not have the eigenvalue $0$.
This is not a restriction since it can be shown that a covariance matrix has a zero eigenvalue if and only if there exists a deterministic linear relationship between some variables in the joint distribution \exerciserefmark{zero_ev_of_cov_mats}.
As we have already seen that a covariance matrix does not have negative eigenvalues \exerciserefmark{cov_mat_pos_sd}, this ensures that $\mSigma$ and~$\mLambda$ are positive definite.\footnote{The inverse of a positive definite matrix is also positive definite.}

An important property of the normal distribution is that it is closed under marginalization and conditioning.

\begin{thmb}
  \begin{thm}[Marginal and conditional distribution]\label{fct:marginal_and_cond_gaussian}\exerciserefmark{marginal_and_cond_gaussian}
    Consider the Gaussian random vector $\rX$ and fix index sets $\sA \subseteq [n]$ and $\sB \subseteq [n]$.
    Then, we have that for any such \emph{marginal distribution},
    \begin{align}
      \rX_A \sim \N{\vmu_\sA}{\mSigma_{\sA\sA}}, \margintag{\normalfont By $\vmu_\sA$ we denote $[\mu_{i_1}, \dots, \mu_{i_k}]$ where $\sA = \{i_1, \dots i_k\}$. $\mSigma_{\sA\sA}$ is defined analogously.}
    \end{align} and that for any such \emph{conditional distribution},
    \begin{subequations}\begin{align}
      \rX_\sA \mid \rX_\sB &= \vx_\sB \sim \N{\vmu_{\sA \mid \sB}}{\mSigma_{\sA \mid \sB}} \quad\text{where} \\
      \vmu_{\sA \mid \sB} &\defeq \vmu_\sA + \mSigma_{\sA\sB}\inv{\mSigma_{\sB\sB}}(\vx_\sB - \vmu_\sB), \margintag{\normalfont Here, $\vmu_\sA$ characterizes the prior belief and \smash{$\mSigma_{\sA\sB}\inv{\mSigma_{\sB\sB}}(\vx_\sB - \vmu_\sB)$} represents ``how different'' $\vx_\sB$ is from what was expected.} \\
      \mSigma_{\sA \mid \sB} &\defeq \mSigma_{\sA\sA} - \mSigma_{\sA\sB}\inv{\mSigma_{\sB\sB}}\mSigma_{\sB\sA}.
    \end{align}\label{eq:cond_gaussian}\end{subequations}
  \end{thm}\vspace{-0.5cm}
\end{thmb}

\Cref{fct:marginal_and_cond_gaussian} provides a closed-form characterization of probabilistic inference for the case that random variables are jointly Gaussian.
We will discuss in \cref{sec:blr}, how this can be turned into an efficient inference algorithm.

Observe that upon inference, the variance can only shrink! Moreover, how much the variance is reduced depends purely on \emph{where} the observations are made (i.e., the choice of $\sB$) but not on \emph{what} the observations are.
In contrast, the posterior mean $\vmu_{\sA \mid \sB}$ depends affinely on $\vmu_\sB$.
These are special properties of the Gaussian and do not generally hold true for other distributions.

It can be shown that Gaussians are additive and closed under affine transformations \exerciserefmark{gaussian_closedness}.
The closedness under affine transformations \eqref{eq:gaussian_lin_trans} implies that a Gaussian $\rX \sim \N{\vmu}{\mSigma}$ is equivalently characterized as \begin{align}
  \rX = \msqrt{\mSigma} \rY + \vmu. \label{eq:gaussian_affine_transformation}
\end{align} where $\rY \sim \SN$ and $\msqrt{\mSigma}$ is the square root of $\mSigma$.\footnote{More details on the square root of a symmetric and positive definite matrix can be found in \cref{sec:fundamentals:qf}.}
Importantly, this implies together with \cref{fct:marginal_and_cond_gaussian} and additivity \eqref{eq:gaussian_additivity} that: \begin{center}
  \emph{Any affine transformation of a Gaussian random vector \\ is a Gaussian random vector.}
\end{center}
A consequence of this is that given any jointly Gaussian random vectors $\rX_\sA$ and $\rX_\sB$, $\rX_\sA$ can be expressed as an affine function of $\rX_\sB$ with added independent Gaussian noise.
Formally, we define \begin{subequations}\begin{align}
  \rX_\sA &\defeq \mA \rX_\sB + \vb + \vvarepsilon \quad\text{where} \\
  \mA &\defeq \mSigma_{\sA\sB}\inv{\mSigma_{\sB\sB}}, \\
  \vb &\defeq \vmu_A - \mSigma_{\sA\sB}\inv{\mSigma_{\sB\sB}} \vmu_B, \\
  \vvarepsilon &\sim \N{\vzero}{\mSigma_{\sA\mid\sB}}.
\end{align}\label{eq:cond_linear_gaussian}\end{subequations}
It directly follows from the closedness of Gaussians under affine transformations \eqref{eq:gaussian_lin_trans} that the characterization of $\rX_\sA$ via \cref{eq:cond_linear_gaussian} is equivalent to $\rX_\sA \sim \N{\vmu_\sA}{\mSigma_{\sA\sA}}$, and hence, \emph{any} Gaussian $\rX_\sA$ can be modeled as a so-called \midx{conditional linear Gaussian}, i.e., an affine function of another Gaussian $\rX_\sB$ with additional independent Gaussian noise.
We will use this fact frequently to represent Gaussians in a compact form.

\section{Supervised Learning and Point Estimates}\label{sec:fundamentals:supervised_learning}

Throughout the first part of this manuscript, we will focus mostly on the \midx{supervised learning}[idxpagebf] problem where we want to learn a function \begin{align*}
  \opt{f} : \spX \to \spY
\end{align*} from labeled training data.
That is, we are given a collection of labeled examples, $\spD_n \defeq \{(\vx_i, y_i)\}_{i=1}^n$, where the $\vx_i \in \spX$ are \midx{inputs} and the $y_i \in \spY$ are \midx{outputs} (called \midx{labels}), and we want to find a function $\hat{f}$ that best-approximates $\opt{f}$.
It is common to choose $\hat{f}$ from a parameterized \midx{function class} $\spF(\Theta)$, where each function $f_\vtheta$ is described by some parameters $\vtheta \in \Theta$.

\begin{marginfigure}
  \incfig{estimation_approximation_error}
  \caption{Illustration of \textbf{estimation error} and \textbf{\b{approximation error}}. $\fs$ denotes the true function and $\hat{f}$ is the best approximation from the function class $\spF$. We do not specify here, how one could quantify ``error''. For more details, see \cref{sec:fundamentals:supervised_learning:risk}.}\label{fig:estimation_approximation_error}
\end{marginfigure}

\begin{rmk}{What this manuscript is about and not about}{}
  As illustrated in \cref{fig:estimation_approximation_error}, the restriction to a function class leads to two sources of error: the \midx{estimation error} of having ``incorrectly'' determined $\hat{f}$ within the function class, and the \midx{approximation error} of the function class itself.
  Choosing a ``good'' function class / architecture with small approximation error is therefore critical for any practical application of machine learning.
  We will discuss various function classes, from linear models to deep neural networks, however, determining the ``right'' function class will not be the focus of this manuscript.
  To keep the exposition simple, we will assume in the following that $\fs \in \spF(\Theta)$ with parameters $\opt{\vtheta} \in \Theta$.

  Instead, we will focus on the problem of estimation/inference within a given function class.
  We will see that inference in smaller function classes is often more computationally efficient since the search space is smaller or --- in the case of Gaussians --- has a known tractable structure.
  On the other hand, larger function classes are more expressive and therefore can typically better approximate the ground truth $\opt{f}$.
\end{rmk}

We differentiate between the task of \midx{regression} where ${\spY \defeq \R^k}$,\footnote{The labels are usually scalar, so ${k = 1}$.} and the task of \midx{classification} where ${\spY \defeq \spC}$ and $\spC$ is an $m$-element set of classes.
In other words, regression is the task of predicting a continuous label, whereas classification is the task of predicting a discrete class label.
These two tasks are intimately related: in fact, we can think of classification tasks as a regression problem where we learn a probability distribution over class labels.
In this regression problem, ${\spY \defeq \Delta^{\spC}}$ where $\Delta^{\spC}$ denotes the set of all probability distributions over the set of classes $\spC$ which is an ${(m-1)}$-dimensional convex polytope in the $m$-dimensional space of probabilities $[0,1]^m$ (cf. \cref{sec:background:probability:probability_simplex}).

For now, let us stick to the regression setting.
We will assume that the observations are noisy, that is, $y_i \iid p(\cdot \mid \vx_i, \opt{\vtheta})$ for some \emph{known} conditional distribution $p(\cdot \mid \vx_i, \vtheta)$ but \emph{unknown} parameter $\opt{\vtheta}$.\footnote{The case where the labels are deterministic is the special case of $p(\cdot \mid \vx_i, \opt{\vtheta})$ being a point density at $\fs(\vx_i)$.}
Our assumption can equivalently be formulated as \begin{align}
  y_i = \underbrace{f_\vtheta(\vx_i)}_{\text{signal}} + \underbrace{\varepsilon_i(\vx_i)}_{\text{noise}} \label{eq:data}
\end{align} where $f_\vtheta(\vx_i)$ is the mean of $p(\cdot \mid \vx_i, \vtheta)$ and $\varepsilon_i(\vx_i) = y_i - f_\vtheta(\vx_i)$ is some independent zero-mean noise, for example (but not necessarily) Gaussian.\footnote{It is crucial that the assumed noise distribution accurately reflects the noise of the data. For example, using a (light-tailed) Gaussian noise model in the presence of heavy-tailed noise will fail! We discuss the distinction between light and heavy tails in \cref{sec:fundamentals:parameter_esitmation:heavy_tails}.}
When the noise distribution may depend on $\vx_i$, the noise is said to be \midx{heteroscedastic}[idxpagebf] and otherwise the noise is called \midx{homoscedastic}[idxpagebf].

\subsection{Maximum Likelihood Estimation}\label{sec:fundamentals:parameter_estimation:mle}

A common approach to finding $\hat{f}$ is to select the model $f \in \spF(\Theta)$ under which the training data is most likely.
This is called the \midx{maximum likelihood estimate} (or MLE): \begin{align}
  \vthetahat_\MLE &\defeq \argmax_{\vtheta \in \Theta} p(y_{1:n} \mid \vx_{1:n}, \vtheta) \label{eq:mle} \\
  &= \argmax_{\vtheta \in \Theta} \prod_{i=1}^n p(y_i \mid \vx_i, \vtheta). \margintag{using the independence of the training data \eqref{eq:data}} \nonumber
  \intertext{Such products of probabilities are often numerically unstable, which is why one typically takes the logarithm:}
  &= \argmax_{\vtheta \in \Theta} \underbrace{\sum_{i=1}^n \log p(y_i \mid \vx_i, \vtheta)}_{\text{\midx{log-likelihood}}}.
\end{align}
We will denote the \midx{negative log-likelihood} by $\ell_\mathrm{nll}(\vtheta; \spD_n)$.

The MLE is often used in practice due to its desirable asymptotic properties as the sample size $n$ increases.
We give a brief summary here and provide additional background and definitions in \cref{sec:background:parameter_estimation}.
To give any guarantees on the convergence of the MLE, we necessarily need to assume that $\opt{\vtheta}$ is identifiable.\footnote{That is, $\opt{\vtheta} \neq \vtheta \implies \fs \neq f_\vtheta$ for any $\vtheta \in \Theta$. In words, there is no other parameter $\vtheta$ that yields the same function $f_\vtheta$ as $\opt{\vtheta}$.}
If additionally, $\ell_\mathrm{nll}$ is ``well-behaved'' then standard results say that the MLE is \emph{consistent} and \emph{asymptotically normal} \citep{van2000asymptotic}: \begin{align}
  \vthetahat_\MLE \convp \opt{\vtheta} \quad\text{and}\quad \vthetahat_\MLE \convd \N{\opt{\vtheta}}{\mS_n} \quad\text{as $n \to \infty$}. \label{eq:mle_asymptotics}
\end{align}
Here, we denote by $\mS_n$ the asymptotic variance of the MLE which can be understood as measuring the ``quality'' of the estimate.\footnote{A ``smaller'' variance means that we can be more confident that the MLE is close to the true parameter.}
This implies in some sense that the MLE is asymptotically unbiased.
Moreover, the MLE can be shown to be \emph{asymptotically efficient} which is to say that there exists no other consistent estimator with a ``smaller'' asymptotic variance.\footnote{see \cref{sec:fundamentals:parameter_estimation:asymptotic_efficiency}.}

The situation is quite different in the finite sample regime.
Here, the MLE need not be unbiased, and it is susceptible to \midx{overfitting} to the (finite) training data as we discuss in more detail in \cref{sec:fundamentals:supervised_learning:risk}.

\subsection{Using Priors: Maximum a Posteriori Estimation}\label{sec:fundamentals:parameter_esitmation:map}

We can incorporate prior assumptions about the parameters $\opt{\vtheta}$ into the estimation procedure.
One approach of this kind is to find the mode of the posterior distribution, called the \midx{maximum a posteriori estimate} (or MAP estimate): \begin{align}
  \vthetahat_\MAP &\defeq \argmax_{\vtheta \in \Theta} p(\vtheta \mid \vx_{1:n}, y_{1:n}) \\
  &= \argmax_{\vtheta \in \Theta} p(y_{1:n} \mid \vx_{1:n}, \vtheta) \cdot p(\vtheta) \margintag{by Bayes' rule \eqref{eq:bayes_rule}} \\
  &= \argmax_{\vtheta \in \Theta} \log p(\vtheta) + \sum_{i=1}^n \log p(y_i \mid \vx_i, \vtheta) \margintag{taking the logarithm} \label{eq:map} \\
  &= \argmin_{\vtheta \in \Theta} \underbrace{- \log p(\vtheta)}_{\text{regularization}} + \underbrace{\ell_\mathrm{nll}(\vtheta; \spD_n)}_{\text{quality of fit}}.
\end{align}
Here, the \midx{log-prior} $\log p(\vtheta)$ acts as a regularizer.
Common regularizers are given, for example, by \begin{itemize}
  \item $p(\vtheta) = \N[\vtheta]{\vzero}{\inv{(2 \lambda)} \mI}$ which yields $-\log p(\vtheta) = \lambda \norm{\vtheta}_2^2 + \const$,
  \item $p(\vtheta) = \Laplace[\vtheta]{\vzero}{\inv{\lambda}}$ which yields $-\log p(\vtheta) = \lambda \norm{\vtheta}_1 + \const$,
  \item a uniform prior (cf. \cref{sec:fundamentals:inference:priors}) for which the MAP is equivalent to the MLE.
  In other words, the MLE is merely the mode of the posterior distribution under a uniform prior.
\end{itemize}

The Gaussian and Laplace regularizers act as simplicity biases, preferring simpler models over more complex ones, which empirically tends to reduce the risk of overfitting.
However, one may also encode more nuanced information about the (assumed) structure of $\opt{\vtheta}$ into the prior.

An alternative way of encoding a prior is by restricting the function class to some $\widetilde{\Theta} \subset \Theta$, for example to rotation- and translation-invariant models as often done when the inputs are images.
This effectively sets $p(\vtheta) = 0$ for all $\vtheta \in \Theta \setminus \widetilde{\Theta}$ but is better suited for numerical optimization than to impose this constraint directly on the prior.

Encoding prior assumptions into the function class or into the parameter estimation can accelerate learning and improve generalization performance dramatically, yet importantly, incorporating a prior can also inhibit learning in case the prior is ``wrong''.
For example, when the learning task is to differentiate images of cats from images of dogs, consider the (stupid) prior that only permits models that exclusively use the upper-left pixel for prediction.
No such model will be able to solve the task, and therefore starting from this prior makes the learning problem effectively unsolvable which illustrates that priors have to be chosen with care.

\subsection{When does the prior matter?}

We have seen that the MLE has desirable asymptotic properties, and that MAP estimation can be seen as a regularized MLE where the type of regularization is encoded by the prior.
Is it possible to derive similar asymptotic results for the MAP estimate?

To answer this question, we will look at the asymptotic effect of the prior on the posterior more generally.
\midx{Doob's consistency theorem} states that assuming parameters are identifiable,\footnote{This is akin to the assumption required for consistency of the MLE, cf. \cref{eq:mle_asymptotics}.} there exists $\widetilde{\Theta} \subseteq \Theta$ with $p(\widetilde{\Theta}) = 1$ such that the posterior is consistent for any $\opt{\vtheta} \in \widetilde{\Theta}$ \citep{doob1949application,miller2016lecture,miller2018detailed}: \begin{align}
  \vtheta \mid \spD_n \convp \opt{\vtheta} \qquad\text{as $n \to \infty$}.
\end{align}
In words, Doob's consistency theorem tells us that for \emph{any} prior distribution, the posterior is guaranteed to converge to a point density in the (small) neighborhood $\opt{\vtheta} \in B$ of the true parameter as long as $p(B) > 0$.\footnote{$B$ can for example be a ball of radius $\epsilon$ around $\opt{\vtheta}$ (with respect to some geometry of $\Theta$).}
We call such a prior a \midx{well-specified prior}.

\begin{rmk}{Cromwell's rule}{}
  In the case where $|\Theta|$ is finite, Doob's consistency theorem strongly suggests that the prior should not assign $0$ probability (or probability $1$ for that matter) to any individual parameter $\vtheta \in \Theta$, unless we know with certainty that $\opt{\vtheta} \neq \vtheta$.
  This is called \midx{Cromwell's rule}, and a prior obeying by this rule is always well-specified.
\end{rmk}

Under the same assumption that the prior is well-specified (and regularity conditions\footnote{These regularity conditions are akin to the assumptions required for asymptotic normality of the MLE, cf. \cref{eq:mle_asymptotics}.}), the \midx{Bernstein-von Mises theorem}, which was first discovered by Pierre-Simon Laplace in the early 19th century, establishes the asymptotic normality of the posterior distribution \citep{van2000asymptotic,miller2016lecture}: \begin{align}
  \vtheta \mid \spD_n \convd \N{\opt{\vtheta}}{\mS_n} \qquad\text{as $n \to \infty$}
\end{align} and where $\mS_n$ is the same as the asymptotic variance of the MLE.\footnote{This has also been called the ``Bayesian central limit theorem'', which is a bit of a misnomer since the theorem also applies to likelihoods (when the prior is noninformative) which are often used in frequentist statistics.}

These results link probabilistic inference to maximum likelihood estimation in the asymptotic limit of infinite data.
Intuitively, in the limit of infinite data, the prior is ``overwhelmed'' by the observations and the posterior becomes equivalent to the limiting distribution of the MLE.\footnote{More examples and discussion can be found in section 17.8 of \cite{le2012asymptotic}, chapter 8 of \cite{le2000asymptotics}, chapter 10 of \cite{van2000asymptotic}, and in \cite{tanner1993tools}.}
One can interpret the regime of infinite data as the regime where computational resources and time are unlimited and plausible inferences evolve into logical inferences.
This transition signifies a shift from the realm of uncertainty to that of certainty.
The importance of the prior surfaces precisely in the non-asymptotic regime where plausible inferences are necessary due to \emph{limited computational resources and limited time}.

\subsection{Estimation vs Inference}

You can interpret a single parameter vector $\vtheta \in \Theta$ as ``one possible explanation'' of the data.
Maximum likelihood and maximum a posteriori estimation are examples of \midx{estimation} algorithms which return a one such parameter vector --- called a \midx{point estimate}.
That is, given the training set $\spD_n$, they return a single parameter vector $\vthetahat_n$.
We give a more detailed account of estimation in \cref{sec:background:parameter_estimation}.

\begin{ex}{Point estimates and invalid logical inferences}{invalid_logical_inferences}
  To see why point estimates can be problematic, recall \cref{ex:weak_probabilistic_inference}.
  We have seen that the logical implication \begin{center}
    ``If it is raining, the ground is wet.''
  \end{center} can be expressed as $\Pr{W \mid R} = 1$.
  Observing that ``The ground is wet.'' does not permit logical inference, yet, the maximum likelihood estimate of $R$ is $\hat{R}_\MLE = 1$.
  This is logically inconsistent since there might be other \emph{explanations} for the ground to be wet, such as a sprinkler!
  With only a finite sample (say independently observing $n$ times that the ground is wet), we cannot rule out with certainty that the ground is wet for other reasons than rain.
\end{ex}

In practice, we never observe an infinite amount of data.
\Cref{ex:invalid_logical_inferences} demonstrates that on a finite sample, point estimates may perform invalid logical inferences, and can therefore lure us into a false sense of certainty.

\begin{rmk}{MLE and MAP are approximations of inference}{}
  The MLE and MAP estimate can be seen as a naive approximation of probabilistic inference, represented by a point density which ``collapses'' all probability mass at the mode of the posterior distribution.
  This can be a relatively decent --- even if overly simple --- approximation when the distribution is unimodal, symmetric, and light-tailed as in \cref{fig:point_estimates}, but is usually a very poor approximation for practical posteriors that are complex and multimodal.
\end{rmk}

\begin{marginfigure}
  \begin{center}
    \import{./plots/output/}{point_estimates.pgf}
  \end{center}

  \caption{A the MLE/MAP are point estimates at the mode $\hat{\theta}$ of the posterior distribution $p(\theta \mid \spD)$.}
  \label{fig:point_estimates}
\end{marginfigure}

In this manuscript, we will focus mainly on algorithms for \midx{probabilistic inference} which compute or approximate the distribution ${p(\vtheta \mid \vx_{1:n}, y_{1:n})}$ over parameters.
Returning a distribution over parameters is natural since this acknowledges that given a finite sample with noisy observations, more than one parameter vector can explain the data.

\subsection{Probabilistic Inference and Prediction}

The prior distribution $p(\vtheta)$ can be interpreted as the degree of our belief that the model parameterized by $\vtheta$ ``describes the (previously seen) data best''.
The likelihood captures how likely the training data is under a particular model: \begin{align}
  p(y_{1:n} \mid \vx_{1:n}, \vtheta) = \prod_{i=1}^n p(y_i \mid \vx_i, \vtheta). \label{eq:likelihood}
\end{align}
The posterior then represents our belief about the best model after seeing the training data.
Using Bayes' rule~\eqref{eq:bayes_rule}, we can write it as\footnote{We generally assume that \begin{align*}
  p(\vtheta \mid \vx_{1:n}) = p(\vtheta).
\end{align*} For our purposes, you can think of the inputs $\vx_{1:n}$ as fixed deterministic parameters, but one can also consider inputs drawn from a distribution over $\spX$.} \begin{subequations}\begin{align}
  p(\vtheta \mid \vx_{1:n}, y_{1:n})&= \frac{1}{Z} p(\vtheta) \prod_{i=1}^n p(y_i \mid \vx_i, \vtheta) \quad\text{where} \\
  Z &\defeq \int_\Theta p(\vtheta) \prod_{i=1}^n p(y_i \mid \vx_i, \vtheta) \, d\vtheta \label{eq:normalizing_constant}
\end{align}\end{subequations} is the \midx{normalizing constant}.
We refer to this process of learning a model from data as \midx{learning}.
We can then use our learned model for \midx{prediction} at a new input $\vxs$ by conditioning on $\vtheta$, \begin{align}
  p(\ys \mid \vxs, \vx_{1:n}, y_{1:n}) &= \int_\Theta p(\ys, \vtheta \mid \vxs, \vx_{1:n}, y_{1:n}) \,d\vtheta \nonumber \margintag{by the sum rule \eqref{eq:sum_rule}} \\
  &= \int_\Theta p(\ys \mid \vxs, \vtheta) \cdot p(\vtheta \mid \vx_{1:n}, y_{1:n}) \,d\vtheta. \label{eq:prediction} \margintag{by the product rule \eqref{eq:product_rule} and $\ys \perp \vx_{1:n}, y_{1:n} \mid \vtheta$}
\end{align}
Here, the distribution over models $p(\vtheta \mid \vx_{1:n}, y_{1:n})$ is called the \midx{posterior}[idxpagebf] and the distribution over predictions $p(\ys \mid \vxs, \vx_{1:n}, y_{1:n})$ is called the \midx{predictive posterior}.
The predictive posterior quantifies our posterior uncertainty about the ``prediction'' $\ys$, however, since this is typically a complex distribution, it is difficult to communicate this uncertainty to a human.
One statistic that can be used for this purpose is the smallest set $\spC_\delta(\vxs) \subseteq \R$ for a fixed $\delta \in (0,1)$ such that \begin{align}
  \Pr{\ys \in \spC_\delta(\vxs) \mid \vxs, \vx_{1:n}, y_{1:n}} \geq 1 - \delta. \label{eq:credible_interval}
\end{align}
That is, we believe with ``confidence'' at least $1 - \delta$ that the true value of $\ys$ lies in $\spC_\delta(\vxs)$.
Such a set $\spC_\delta(\vxs)$ is called a \midx{credible set}.

\begin{marginfigure}
  \begin{center}
    \import{./plots/output/}{credible_intervals.pgf}
  \end{center}

  \caption{Example of a $95\%$ credible set at $\vxs$ where the predictive posterior is Gaussian with mean $\mu(\vxs)$ and standard deviation $\sigma(\vxs)$. In this case, the gray area integrates to $\approx 0.95$ for \begin{align*}
    \spC_{0.05}(\vxs) = [\mu(\vxs) \pm 1.96 \sigma(\vxs)].
  \end{align*}}
  \label{fig:credible_intervals}
\end{marginfigure}

We have seen here that the tasks of learning and prediction are intimately related.
Indeed, ``prediction'' can be seen in many ways as a natural by-product of ``reasoning'' (i.e., probabilistic inference), where we evaluate the likelihood of outcomes given our learned explanations for the world.
This intuition can be read off directly from \cref{eq:prediction} where $p(\ys \mid \vxs, \vtheta)$ corresponds to the likelihood of an outcome given the explanation $\vtheta$ and $p(\vtheta \mid \vx_{1:n}, y_{1:n})$ corresponds to our inferred belief about the world.
We will see many more examples of this link between probabilistic inference and prediction throughout this manuscript.

The high-dimensional integrals of \cref{eq:normalizing_constant,eq:prediction} are typically intractable, and represent the main computational challenge in probabilistic inference.
Throughout the first part of this manuscript, we will describe settings where exact inference is tractable, as well as modern approximate inference algorithms that can be used when exact inference is intractable.

\subsection{Recursive Probabilistic Inference and Memory}\label{sec:fundamentals:recursive_inference}

We have already alluded to the fact that probabilistic inference has a recursive structure, which lends itself to continual learning and which often leads to efficient algorithms.
Let us denote by \begin{align}
  p^{(t)}(\vtheta) \defeq p(\vtheta \mid \vx_{1:t}, y_{1:t})
\end{align} the posterior after the first $t$ observations with $p^{(0)}(\vtheta) = p(\vtheta)$.
Now, suppose that we have already computed $p^{(t)}(\vtheta)$ and observe $y_{t+1}$.
We can recursively update the posterior as follows, \begin{align}
  p^{(t+1)}(\vtheta) &= p(\vtheta \mid y_{1:t+1}) \nonumber \\
  &\propto p(\vtheta \mid y_{1:t}) \cdot p(y_{t+1} \mid \vtheta, y_{1:t}) \margintag{using Bayes' rule \eqref{eq:bayes_rule}} \nonumber \\
  &= p^{(t)}(\vtheta) \cdot p(y_{t+1} \mid \vtheta). \margintag{using $y_{t+1} \perp y_{1:t} \mid \vtheta$, see \cref{fig:blr}} \label{eq:recursive_probabilistic_inference}
\end{align}
Intuitively, the posterior distribution at time $t$ ``absorbs'' or ``summarizes'' all seen data.

By unrolling the recursion of \cref{eq:recursive_probabilistic_inference}, we see that regardless of the philosophical interpretation of probability, probabilistic inference is a fundamental  mechanism of learning.
Even the MLE which performs naive approximate inference without a prior (i.e., a uniform prior), is based on $p^{(n)}(\vtheta) \propto p(y_{1:n} \mid \vx_{1:n}, \vtheta)$ which is the result of $n$ individual plausible inferences, where the $(t+1)$-st inference uses the posterior of the $t$-th inference as its prior.

So far we have been considering the supervised learning setting, where all data is available a-priori.
However, by sequentially obtaining the new posterior and replacing our prior, we can also perform probabilistic inference as data arrives online (i.e., in ``real-time'').
This is analogous to recursive logical inference where derived consequences are repeatedly added to the set of propositions to derive new consequences.
This also highlights the intimate connection between ``reasoning'' and ``memory''.
Indeed, the posterior distribution $p^{(t)}(\vtheta)$ can be seen as a form of memory that evolves with time $t$.

\section{Outlook: Decision Theory}\label{sec:decision_theory}

How can we use our predictions to make concrete decisions under uncertainty?
We will study this question extensively in \cref{part2} of this manuscript, but briefly introduce some fundamental concepts here.
Making decisions using a probabilistic model $p(y \mid \vx)$ of output $y \in \spY$ given input $\vx \in \spX$, such as the ones we have discussed in the previous section, is commonly formalized by \begin{itemize}
  \item a set of possible actions $\spA$, and
  \item a reward function $r(y, a) \in \R$ that computes the reward or utility of taking action ${a \in \spA}$, assuming the true output is ${y \in \spY}$.
\end{itemize}
Standard decision theory recommends picking the action with the largest expected utility: \begin{align}
  \opt{a}(\vx) \defeq \argmax_{a \in \spA} \E[y \mid \vx]{r(y, a)}. \label{eq:decision_theory}
\end{align}
Here, $\opt{a}$ is called the \midx{optimal decision rule} because, under the given probabilistic model, no other rule can yield a higher expected utility.

Let us consider some examples of reward functions and their corresponding optimal decisions:

\begin{ex}{Reward functions}{decision_theory}
  Under the decision rule from \cref{eq:decision_theory}, different reward functions~$r$ can lead to different decisions.
  Let us examine two reward functions for the case where $\spY = \spA = \R$ \exerciserefmark{decision_theory}.

  \begin{itemize}
    \item Alternatively to considering $r$ as a reward function, we can interpret $-r$ as the loss of taking action $a$ when the true output is $y$.
    If our goal is for our actions $a$ to ``mimic'' the output $y$, a natural choice is the squared loss, $- r(y, a) = (y - a)^2$.
    It turns out that under the squared loss, the optimal decision is simply the mean: $\opt{a}(\vx) = \E{y \mid \vx}$.
    \item To contrast this, we consider the asymmetric loss, \begin{align*}
      - r(y, a)= c_1 \underbrace{\max\{y-a, 0\}}_{\text{underestimation error}} \quad+\quad c_2 \underbrace{\max\{a-y, 0\}}_{\text{overestimation error}},
    \end{align*} which penalizes underestimation and overestimation differently.
    When $y \mid \vx \sim \N{\mu_\vx}{\sigma_\vx^2}$ then the optimal decision is \begin{align*}
      \opt{a}(\vx) = \mu_\vx + \underbrace{\sigma_\vx \cdot \inv{\Phi}\parentheses*{\frac{c_1}{c_1 + c_2}}}_{\text{pessimism / optimism}}
    \end{align*} where $\Phi$ is the CDF of the standard normal distribution.\safefootnote{Recall that the CDF~$\Phi$ of the standard normal distribution is a sigmoid with its inverse satisfying \begin{align*}
      \inv{\Phi}(u) \begin{cases}
        < 0 & \text{if } u < 0.5, \\
        = 0 & \text{if } u = 0.5, \\
        > 0 & \text{if } u > 0.5.
      \end{cases}
    \end{align*}}
    Note that if $c_1 = c_2$, then the second term vanishes and the optimal decision is the same as under the squared loss.
    If $c_1 > c_2$, the second term is positive (i.e., \midx<optimistic>{optimism in the face of uncertainty}) to avoid underestimation, and if $c_1 < c_2$, the second term is negative (i.e., \emph{pessimistic}) to avoid overestimation.
    We will find these notions of optimism and pessimism to be useful in many decision-making scenarios.
  \end{itemize}
\end{ex}

While \cref{eq:decision_theory} describes how to make optimal decisions given a (posterior) probabilistic model, it does not tell us how to learn or improve this model in the first place.
That is, these decisions are only optimal under the assumption that we cannot use their outcomes and our resulting observations to update our model and inform future decisions.
When we start to consider the effect of our decisions on future data and future posteriors, answering ``how do I make optimal decisions?'' becomes more complex, and we will study this in \cref{part2} on sequential decision-making.

\section*{Discussion}

In this chapter, we have learned about the fundamental concepts of probabilistic inference.
We have seen that probabilistic inference is the natural extension of logical reasoning to domains with uncertainty.
We have also derived the central principle of probabilistic inference, Bayes' rule, which is simple to state but often computationally challenging.
In the next part of this manuscript, we will explore settings where exact inference is tractable, as well as modern approaches to approximate probabilistic inference.

\section*{Overview of Mathematical Background}

We have included brief summaries of the fundamentals of \textbf{parameter estimation} (mean estimation in particular) and \textbf{optimization} in \cref{sec:background:parameter_estimation,sec:fundamentals:optimization}, respectively, which we will refer back to throughout the manuscript.
\Cref{sec:fundamentals:qf} discusses the \textbf{correspondence of Gaussians and quadratic forms}.
\Cref{sec:background:identities_and_inequalities} comprises a list of useful matrix identities and inequalities.

\excheading

\begin{nexercise}{Properties of probability}{properties_of_probability}
  Let $(\Omega, \spA, \fnPr)$ be a probability space.
  Derive the following properties of probability from the Kolmogorov axioms:
  \begin{enumerate}
    \item For any $A, B \in \spA$, if $A \subseteq B$ then $\Pr{A} \leq \Pr{B}$.
    \item For any $A \in \spA$, $\Pr{\compl{A}} = 1 - \Pr{A}$.
    \item For any countable set of events $\{\sA_i \in \spA\}_i$, \begin{align}
      \Pr{\bigcup_{i=1}^\infty \sA_i} \leq \sum_{i=1}^\infty \Pr{\sA_i}. \label{eq:union_bound}
    \end{align} which is called a \midx{union bound}.
  \end{enumerate}
\end{nexercise}

\begin{nexercise}{Random walks on graphs}{graph_random_walk}
  Let $G$ be a simple connected finite graph. We start at a vertex $u$ of $G$. At every step, we move to one of the neighbors of the current vertex uniformly at random, e.g., if the vertex has $3$ neighbors, we move to one of them, each with probability $1 / 3$. What is the probability that the walk visits a given vertex $v$ eventually?
\end{nexercise}

\begin{nexercise}{Law of total expectation}{lote}
  Show that if $\{\sA_i\}_{i=1}^k$ are a partition of $\Omega$ and $\rX$ is a random vector, \begin{align}
    \E{\rX} = \sum_{i=1}^k \E{\rX \mid \sA_i} \cdot \Pr{\sA_i}.
  \end{align}
\end{nexercise}

\begin{nexercise}{Covariance matrices are positive semi-definite}{cov_mat_pos_sd}
  Prove that a covariance matrix $\mSigma$ is always positive semi-definite.
  That is, all of its eigenvalues are greater or equal to zero, or equivalently, $\transpose{\vx}\mSigma\vx \geq 0$ for any $\vx \in \R^n$.
\end{nexercise}

\begin{nexercise}{Probabilistic inference}{bayes_rule}
  As a result of a medical screening, one of the tests revealed a serious disease in a person.
  The test has a high accuracy of $99\%$ (the probability of a positive response in the presence of a disease is $99\%$ and the probability of a negative response in the absence of a disease is also~$99\%$).
  However, the disease is quite rare and occurs only in one person per $10\,000$.
  Calculate the probability of the examined person having the identified disease.
\end{nexercise}

\begin{nexercise}{Zero eigenvalues of covariance matrices}{zero_ev_of_cov_mats}
  We say that a random vector $\rX$ in $\R^n$ is not linearly independent if for some $\valpha \in \R^n \setminus \{\vzero\}$, $\transpose{\valpha} \rX = 0$.
  \begin{enumerate}
    \item Show that if $\rX$ is not linearly independent, then $\Var{\rX}$ has a zero eigenvalue.

    \item Show that if $\Var{\rX}$ has a zero eigenvalue, then $\rX$ is not linearly independent. \\
    \textit{Hint: Consider the variance of $\transpose{\vlambda} \rX$ where $\vlambda$ is the eigenvector corresponding to the zero eigenvalue.}
  \end{enumerate}
  Thus, we have shown that $\Var{\rX}$ has a zero eigenvalue if and only if~$\rX$ is not linearly independent.
\end{nexercise}

\begin{nexercise}{Product of Gaussian PDFs}{gaussian_pdf_prod}
  Let $\vmu_1, \vmu_2 \in \R^n$ be mean vectors and $\mSigma_1, \mSigma_2 \in \R^{n \times n}$ be covariance matrices. Prove that \begin{align}
    \N[\vx]{\vmu}{\mSigma} \propto \N[\vx]{\vmu_1}{\mSigma_1} \cdot \N[\vx]{\vmu_2}{\mSigma_2} \label{eq:gaussian_pdf_product}
  \end{align} for some mean vector $\vmu \in \R^n$ and covariance matrix $\mSigma \in \R^{n \times n}$.
  That is, show that the product of two Gaussian PDFs is proportional to the PDF of a Gaussian.
\end{nexercise}

\begin{nexercise}{Independence of Gaussians}{grv_uncor_indep}
  Show that two jointly Gaussian random vectors, $\rX$ and $\rY$, are independent if and only if $\rX$ and $\rY$ are uncorrelated.
\end{nexercise}

\begin{nexercise}{Marginal / conditional distribution of a Gaussian}{marginal_and_cond_gaussian}
  Prove \cref{fct:marginal_and_cond_gaussian}. That is, show that \begin{enumerate}
    \item every marginal of a Gaussian is Gaussian; and
    \item conditioning on a subset of variables of a joint Gaussian is Gaussian
  \end{enumerate} by finding their corresponding PDFs.

  \textit{Hint: You may use that for matrices $\mSigma$ and $\mLambda$ such that $\inv{\mSigma} = \mLambda$, \begin{itemize}
    \item if $\mSigma$ and $\mLambda$ are symmetric, \begin{align*}
      &\transpose{\begin{bmatrix}
        \vx_\sA \\
        \vx_\sB \\
      \end{bmatrix}} \begin{bmatrix}
        \mLambda_{\sA\sA} & \mLambda_{\sA\sB} \\
        \mLambda_{\sB\sA} & \mLambda_{\sB\sB} \\
      \end{bmatrix} \begin{bmatrix}
        \vx_\sA \\
        \vx_\sB \\
      \end{bmatrix} \\
      &= \transpose{\vx_\sA}\mLambda_{\sA\sA}\vx_\sA + \transpose{\vx_\sA}\mLambda_{\sA\sB}\vx_\sB + \transpose{\vx_\sB}\mLambda_{\sB\sA}\vx_\sA + \transpose{\vx_\sB}\mLambda_{\sB\sB}\vx_\sB \\
      &= \begin{multlined}[t]
        \transpose{\vx_\sA}(\mLambda_{\sA\sA} - \mLambda_{\sA\sB}\inv{\mLambda_{\sB\sB}}\mLambda_{\sB\sA})\vx_\sA + \\ \transpose{(\vx_\sB + \inv{\mLambda_{\sB\sB}}\mLambda_{\sB\sA}\vx_\sA)}\mLambda_{\sB\sB}(\vx_\sB + \inv{\mLambda_{\sB\sB}}\mLambda_{\sB\sA}\vx_\sA),
      \end{multlined}
    \end{align*}
    \item $\inv{\mLambda_{\sB\sB}} = \mSigma_{\sB\sB} - \mSigma_{\sB\sA}\inv{\mSigma_{\sA\sA}}\mSigma_{\sA\sB}$,
    \item $\inv{\mLambda_{\sB\sB}}\mLambda_{\sB\sA} = -\mSigma_{\sB\sA}\inv{\mSigma_{\sA\sA}}$.
  \end{itemize}
  The final two equations follow from the general characterization of the inverse of a block matrix \citep[section 9.1.3]{petersen2008matrix}.}
\end{nexercise}

\begin{nexercise}{Closedness properties of Gaussians}{gaussian_closedness}
  Recall the notion of a \midx{moment-generating function} (MGF) of a random vector $\rX$ in $\R^n$ which is defined as \begin{align}
    \varphi_{\rX}(\vt) \defeq \E*{\exp\parentheses*{\transpose{\vt}\rX}}, \quad\text{for all $\vt \in \R^n$}. \label{eq:mgf}
  \end{align}
  An MGF uniquely characterizes a distribution.
  The MGF of the multivariate Gaussian $\rX \sim \N{\vmu}{\mSigma}$ is \begin{align}
    \varphi_\rX(\vt) = \exp\parentheses*{\transpose{\vt}\vmu + \frac{1}{2}\transpose{\vt}\mSigma\vt}. \label{eq:mgf_gaussian}
  \end{align}
  This generalizes the MGF of the univariate Gaussian from \cref{eq:mgf_univ_gaussian}.

  Prove the following facts.
  \begin{enumerate}
    \item \emph{Closedness under affine transformations:} Given an $n$-dimensional Gaussian $\rX \sim \N{\vmu}{\mSigma}$, and ${\mA \in \R^{m \times n}}$ and $\vb \in \R^m$, \begin{align}
      \mA \rX + \vb \sim \N{\mA \vmu + \vb}{\mA \mSigma \transpose{\mA}}. \label{eq:gaussian_lin_trans}
    \end{align}

    \item \emph{Additivity:} Given two independent Gaussian random vectors ${\rX \sim \N{\vmu}{\mSigma}}$ and ${\rXp \sim \N{\vmup}{\mSigmap}}$ in $\R^n$, \begin{align}
      \rX + \rXp \sim \N{\vmu + \vmup}{\mSigma + \mSigmap}. \label{eq:gaussian_additivity}
    \end{align}
  \end{enumerate}

  These properties are unique to Gaussians and a reason for why they are widely used for learning and inference.
\end{nexercise}

\begin{nexercise}{Expectation and variance of Gaussians}{expectation_and_variance_of_gaussians}
  Derive that $\E{\rX} = \vmu$ and $\Var{\rX} = \mSigma$ when $\rX \sim \N{\vmu}{\mSigma}$. \\
  \textit{Hint: First derive the expectation and variance of a univariate standard normal random variable.}
\end{nexercise}

\begin{nexercise}{Non-affine transformations of Gaussians}{non_affine_transformations_of_gaussians}
  Answer the following questions with \textbf{yes} or \textbf{no}.
  \begin{enumerate}
    \item Does there exist any non-affine transformation of a Gaussian random vector which is Gaussian? If yes, give an example.
    \item Let $X, Y, Z$ be independent standard normal random variables. Is $\frac{X + YZ}{\sqrt{1 + Z^2}}$ Gaussian?
  \end{enumerate}
\end{nexercise}

\begin{nexercise}{Decision theory}{decision_theory}
  Derive the optimal decisions under the squared loss and the asymmetric loss from \cref{ex:decision_theory}.
\end{nexercise}

  \part{Probabilistic Machine Learning}\label{part1}
  \chapter*{Preface to \Cref{part1}}

As humans, we constantly learn about the world around us.
We learn to interact with our physical surroundings.
We deepen our understanding of the world by establishing relationships between actors, objects, and events.
And we learn about ourselves by observing how we interact with the world and with ourselves.
We then continuously use this knowledge to make inferences and predictions, be it about the weather, the movement of a ball, or the behavior of a friend.

With limited computational resources, limited genetic information, and limited life experience, we are not able to learn everything about the world to complete certainty.
We saw in \cref{sec:fundamentals} that probability theory is the mathematical framework for reasoning with uncertainty in the same way that logic is the mathematical framework for reasoning with certainty.
We will discuss two kinds of uncertainty: ``aleatoric'' uncertainty which cannot be reduced under computational constraints, and ``epistemic'' uncertainty which can be reduced by observing more data.\looseness=-1

An important aspect of learning is that we do not just learn once, but continually.
Bayes' rule allows us to update our beliefs and reduce our uncertainty as we observe new data --- a process that is called \midx{probabilistic inference}.
By taking the former posterior as the new prior, probabilistic inference can be performed continuously and repeated indefinitely as we observe more and more data.

\begin{marginfigure}
  \incfig{perception_diagram}
	\caption{A schematic illustration of probabilistic inference in the context of the (supervised) learning of a model $\vtheta$ from perceived data $\spD$. The prior model $p(\vtheta)$ can equip the model with anything from substantial, to little, to no prior knowledge.\looseness=-1}
  \label{fig:perception_diagram}
\end{marginfigure}

Our sensory information is often noisy and imperfect, which is another source of uncertainty.
The same is true for machines, even if they can sometimes sense aspects of the world more accurately than humans.
We discuss how one can infer latent structure of the world from sensed data, such as the state of a dynamical system like a car, in a process that is called \midx{filtering}.

In this first part of the \course, we examine how we can build machines that are capable of (continual) learning and inference.
First, we introduce probabilistic inference in the context of linear models which make predictions based on fixed (often hand-designed) features.
We then discuss how probabilistic inference can be scaled to kernel methods and Gaussian processes which use a large (potentially infinite) number of features, and to deep neural networks which learn features dynamically from data.
In these models, exact inference is typically intractable, and we discuss modern methods for approximate inference such as variational inference and Markov chain Monte Carlo.
We highlight a tradeoff between curiosity (i.e., extrapolating beyond the given data) and conformity (i.e., fitting the given data), which surfaces as a fundamental principle of probabilistic inference in the regime where the data and our computational resources are limited.

  \chapter{Linear Regression}\label{sec:blr}\pidx{linear regression}

As a first example of probabilistic inference, we will study linear models for regression\footnote{As we have discussed in \cref{sec:fundamentals:supervised_learning}, regression models can also be used for classification. The canonical example of a linear model for classification is logistic regression, which we will discuss in \cref{sec:approximate_inference:bayesian_logistic_regression}.} which assume that the output $y \in \R$ is a linear function of the input $\vx \in \R^d$: \begin{align*}
  y \approx \transpose{\vw}\vx + w_0
\end{align*} where $\vw \in \R^d$ are the weights and $w_0 \in \R$ is the intercept.
Observe that if we define the extended inputs $\vx' \defeq (\vx, 1)$ and $\vw' \defeq (\vw, w_0)$, then $\transpose{\vw'}\vx' = \transpose{\vw}\vx~+~w_0$, implying that without loss of generality it suffices to study linear functions without the intercept term $w_0$.
We will therefore consider the following function class of linear models \begin{align*}
  f(\vx; \vw) \defeq \transpose{\vw} \vx.
\end{align*}

\begin{marginfigure}
  \begin{center}
    \import{./plots/output/}{linear_regression.pgf}
  \end{center}

  \caption{Example of linear regression with the least squares estimator (shown in blue).}\label{fig:lr}
\end{marginfigure}

We will consider the supervised learning task of learning weights $\vw$ from labeled training data $\{(\vx_i, y_i)\}_{i=1}^n$.
We define the \midx{design matrix}, \begin{align}
  \mX \defeq \begin{bmatrix}
    \transpose{\vx_1} \\
    \vdots \\
    \transpose{\vx_n} \\
  \end{bmatrix} \in \R^{n \times d},
\end{align} as the collection of inputs and the vector $\vy \defeq \transpose{[y_1 \cdots y_n]} \in \R^n$ as the collection of labels.
For each noisy observation $(\vx_i, y_i)$, we define the value of the approximation of our model, $f_i \defeq \transpose{\vw} \vx_i$.
Our model at the inputs $\mX$ is described by the vector $\vf \defeq \transpose{[f_1 \; \cdots \; f_n]}$ which can be expressed succinctly as $\vf = \mX \vw$.

The most common way of estimating $\vw$ from data is the \midx{least squares estimator},\looseness=-1 \begin{align}
  \vwhat_{\ls} \defeq \argmin_{\vw \in \R^d} \sum_{i=1}^n (y_i - \transpose{\vw} \vx_i)^2 = \argmin_{\vw \in \R^d} \norm{\vy - \mX\vw}_2^2, \label{eq:least_squares}
\end{align} minimizing the squared difference between the labels and predictions of the model.
A slightly different estimator is used for \midx{ridge regression}, \begin{align}
  \vwhat_{\ridge} \defeq \argmin_{\vw \in \R^d} \norm{\vy - \mX\vw}_2^2 + \lambda \norm{\vw}_2^2
\end{align} where $\lambda > 0$.
The squared $L_2$ regularization term $\lambda \norm{\vw}_2^2$ penalizes large $\vw$ and thus reduces the ``complexity'' of the resulting model.\footnote[][-\baselineskip]{Ridge regression is more robust to multicollinearity than standard linear regression. \idx<Multicollinearity>{multicollinearity} occurs when multiple independent inputs are highly correlated. In this case, their individual effects on the predicted variable cannot be estimated well. Classical linear regression is highly volatile to small input changes. The regularization of ridge regression reduces this volatility by introducing a bias on the weights towards $0$.}
It can be shown that the unique solutions to least squares and ridge regression are given by \begin{align}
  \vwhat_{\ls} &= \inv{(\transpose{\mX} \mX)} \transpose{\mX} \vy \label{eq:linear_regression} \qquad\text{and} \\
  \vwhat_{\ridge} &= \inv{(\transpose{\mX} \mX + \lambda \mI)} \transpose{\mX} \vy, \label{eq:ridge_regression}
\end{align} respectively if the Hessian of the loss is positive definite (i.e., the loss is strictly convex) \exerciserefmark{closed_form_linear_regression}[1] which is the case as long as the columns of $\mX$ are not linearly dependent.
Least squares regression can be seen as finding the orthogonal projection of $\vy$ onto the column space of $\mX$, as is illustrated in \cref{fig:linear_regression_projection} \exerciserefmark{closed_form_linear_regression}[2].

\begin{marginfigure}[9\baselineskip]
  \begin{center}
    %% Creator: Matplotlib, PGF backend
%%
%% To include the figure in your LaTeX document, write
%%   \input{<filename>.pgf}
%%
%% Make sure the required packages are loaded in your preamble
%%   \usepackage{pgf}
%%
%% Also ensure that all the required font packages are loaded; for instance,
%% the lmodern package is sometimes necessary when using math font.
%%   \usepackage{lmodern}
%%
%% Figures using additional raster images can only be included by \input if
%% they are in the same directory as the main LaTeX file. For loading figures
%% from other directories you can use the `import` package
%%   \usepackage{import}
%%
%% and then include the figures with
%%   \import{<path to file>}{<filename>.pgf}
%%
%% Matplotlib used the following preamble
%%   \usepackage{amsmath}\usepackage{amssymb}
%%   \makeatletter\@ifpackageloaded{underscore}{}{\usepackage[strings]{underscore}}\makeatother
%%
\begingroup%
\makeatletter%
\begin{pgfpicture}%
\pgfpathrectangle{\pgfpointorigin}{\pgfqpoint{2.100000in}{2.100000in}}%
\pgfusepath{use as bounding box, clip}%
\begin{pgfscope}%
\pgfsetbuttcap%
\pgfsetmiterjoin%
\pgfsetlinewidth{0.000000pt}%
\definecolor{currentstroke}{rgb}{0.000000,0.000000,0.000000}%
\pgfsetstrokecolor{currentstroke}%
\pgfsetstrokeopacity{0.000000}%
\pgfsetdash{}{0pt}%
\pgfpathmoveto{\pgfqpoint{0.000000in}{0.000000in}}%
\pgfpathlineto{\pgfqpoint{2.100000in}{0.000000in}}%
\pgfpathlineto{\pgfqpoint{2.100000in}{2.100000in}}%
\pgfpathlineto{\pgfqpoint{0.000000in}{2.100000in}}%
\pgfpathlineto{\pgfqpoint{0.000000in}{0.000000in}}%
\pgfpathclose%
\pgfusepath{}%
\end{pgfscope}%
\begin{pgfscope}%
\pgfsetbuttcap%
\pgfsetmiterjoin%
\pgfsetlinewidth{0.000000pt}%
\definecolor{currentstroke}{rgb}{0.000000,0.000000,0.000000}%
\pgfsetstrokecolor{currentstroke}%
\pgfsetstrokeopacity{0.000000}%
\pgfsetdash{}{0pt}%
\pgfpathmoveto{\pgfqpoint{0.000000in}{0.000000in}}%
\pgfpathlineto{\pgfqpoint{2.100000in}{0.000000in}}%
\pgfpathlineto{\pgfqpoint{2.100000in}{2.100000in}}%
\pgfpathlineto{\pgfqpoint{0.000000in}{2.100000in}}%
\pgfpathlineto{\pgfqpoint{0.000000in}{0.000000in}}%
\pgfpathclose%
\pgfusepath{}%
\end{pgfscope}%
\begin{pgfscope}%
\pgfpathrectangle{\pgfqpoint{0.000000in}{0.000000in}}{\pgfqpoint{2.100000in}{2.100000in}}%
\pgfusepath{clip}%
\pgfsetbuttcap%
\pgfsetroundjoin%
\pgfsetlinewidth{1.003750pt}%
\definecolor{currentstroke}{rgb}{0.941176,0.125490,0.000000}%
\pgfsetstrokecolor{currentstroke}%
\pgfsetdash{}{0pt}%
\pgfpathmoveto{\pgfqpoint{1.251262in}{0.690527in}}%
\pgfpathlineto{\pgfqpoint{0.552292in}{0.590989in}}%
\pgfusepath{stroke}%
\end{pgfscope}%
\begin{pgfscope}%
\pgfpathrectangle{\pgfqpoint{0.000000in}{0.000000in}}{\pgfqpoint{2.100000in}{2.100000in}}%
\pgfusepath{clip}%
\pgfsetbuttcap%
\pgfsetroundjoin%
\pgfsetlinewidth{1.003750pt}%
\definecolor{currentstroke}{rgb}{0.941176,0.125490,0.000000}%
\pgfsetstrokecolor{currentstroke}%
\pgfsetdash{}{0pt}%
\pgfpathmoveto{\pgfqpoint{1.251262in}{0.690527in}}%
\pgfpathlineto{\pgfqpoint{1.184474in}{0.656466in}}%
\pgfusepath{stroke}%
\end{pgfscope}%
\begin{pgfscope}%
\pgfpathrectangle{\pgfqpoint{0.000000in}{0.000000in}}{\pgfqpoint{2.100000in}{2.100000in}}%
\pgfusepath{clip}%
\pgfsetbuttcap%
\pgfsetroundjoin%
\pgfsetlinewidth{1.003750pt}%
\definecolor{currentstroke}{rgb}{0.941176,0.125490,0.000000}%
\pgfsetstrokecolor{currentstroke}%
\pgfsetdash{}{0pt}%
\pgfpathmoveto{\pgfqpoint{1.251262in}{0.690527in}}%
\pgfpathlineto{\pgfqpoint{1.184746in}{0.705667in}}%
\pgfusepath{stroke}%
\end{pgfscope}%
\begin{pgfscope}%
\definecolor{textcolor}{rgb}{0.000000,0.000000,0.000000}%
\pgfsetstrokecolor{textcolor}%
\pgfsetfillcolor{textcolor}%
\pgftext[x=1.293558in,y=1.426938in,left,base]{\color{textcolor}\rmfamily\fontsize{8.330000}{9.996000}\selectfont \(\displaystyle \boldsymbol{y}\)}%
\end{pgfscope}%
\begin{pgfscope}%
\definecolor{textcolor}{rgb}{0.941176,0.125490,0.000000}%
\pgfsetstrokecolor{textcolor}%
\pgfsetfillcolor{textcolor}%
\pgftext[x=1.286053in,y=0.729049in,left,base]{\color{textcolor}\rmfamily\fontsize{8.330000}{9.996000}\selectfont \(\displaystyle \boldsymbol{X}\boldsymbol{\hat{w}}_{\mathrm{ls}}\)}%
\end{pgfscope}%
\begin{pgfscope}%
\definecolor{textcolor}{rgb}{0.000000,0.000000,0.000000}%
\pgfsetstrokecolor{textcolor}%
\pgfsetfillcolor{textcolor}%
\pgftext[x=0.359036in,y=0.205232in,left,base]{\color{textcolor}\rmfamily\fontsize{8.330000}{9.996000}\selectfont \(\displaystyle \mathrm{span}\{\boldsymbol{X}\}\)}%
\end{pgfscope}%
\begin{pgfscope}%
\pgfpathrectangle{\pgfqpoint{0.000000in}{0.000000in}}{\pgfqpoint{2.100000in}{2.100000in}}%
\pgfusepath{clip}%
\pgfsetbuttcap%
\pgfsetroundjoin%
\pgfsetlinewidth{1.003750pt}%
\definecolor{currentstroke}{rgb}{0.000000,0.000000,0.000000}%
\pgfsetstrokecolor{currentstroke}%
\pgfsetdash{}{0pt}%
\pgfpathmoveto{\pgfqpoint{1.257502in}{1.386277in}}%
\pgfpathlineto{\pgfqpoint{0.552292in}{0.590989in}}%
\pgfusepath{stroke}%
\end{pgfscope}%
\begin{pgfscope}%
\pgfpathrectangle{\pgfqpoint{0.000000in}{0.000000in}}{\pgfqpoint{2.100000in}{2.100000in}}%
\pgfusepath{clip}%
\pgfsetbuttcap%
\pgfsetroundjoin%
\pgfsetlinewidth{1.003750pt}%
\definecolor{currentstroke}{rgb}{0.000000,0.000000,0.000000}%
\pgfsetstrokecolor{currentstroke}%
\pgfsetdash{}{0pt}%
\pgfpathmoveto{\pgfqpoint{1.257502in}{1.386277in}}%
\pgfpathlineto{\pgfqpoint{1.200958in}{1.283475in}}%
\pgfusepath{stroke}%
\end{pgfscope}%
\begin{pgfscope}%
\pgfpathrectangle{\pgfqpoint{0.000000in}{0.000000in}}{\pgfqpoint{2.100000in}{2.100000in}}%
\pgfusepath{clip}%
\pgfsetbuttcap%
\pgfsetroundjoin%
\pgfsetlinewidth{1.003750pt}%
\definecolor{currentstroke}{rgb}{0.000000,0.000000,0.000000}%
\pgfsetstrokecolor{currentstroke}%
\pgfsetdash{}{0pt}%
\pgfpathmoveto{\pgfqpoint{1.257502in}{1.386277in}}%
\pgfpathlineto{\pgfqpoint{1.175135in}{1.332547in}}%
\pgfusepath{stroke}%
\end{pgfscope}%
\begin{pgfscope}%
\pgfpathrectangle{\pgfqpoint{0.000000in}{0.000000in}}{\pgfqpoint{2.100000in}{2.100000in}}%
\pgfusepath{clip}%
\pgfsetbuttcap%
\pgfsetroundjoin%
\definecolor{currentfill}{rgb}{0.679477,0.679477,0.679477}%
\pgfsetfillcolor{currentfill}%
\pgfsetfillopacity{0.000000}%
\pgfsetlinewidth{1.003750pt}%
\definecolor{currentstroke}{rgb}{0.000000,0.000000,0.000000}%
\pgfsetstrokecolor{currentstroke}%
\pgfsetdash{}{0pt}%
\pgfsys@defobject{currentmarker}{\pgfqpoint{0.195235in}{0.172193in}}{\pgfqpoint{1.930745in}{1.098328in}}{%
\pgfpathmoveto{\pgfqpoint{0.195235in}{0.540142in}}%
\pgfpathlineto{\pgfqpoint{1.327293in}{0.172193in}}%
\pgfpathlineto{\pgfqpoint{1.930745in}{0.787290in}}%
\pgfpathlineto{\pgfqpoint{0.860366in}{1.098328in}}%
\pgfpathlineto{\pgfqpoint{0.195235in}{0.540142in}}%
\pgfpathclose%
\pgfusepath{stroke,fill}%
}%
\begin{pgfscope}%
\pgfsys@transformshift{0.000000in}{0.000000in}%
\pgfsys@useobject{currentmarker}{}%
\end{pgfscope}%
\end{pgfscope}%
\begin{pgfscope}%
\pgfpathrectangle{\pgfqpoint{0.000000in}{0.000000in}}{\pgfqpoint{2.100000in}{2.100000in}}%
\pgfusepath{clip}%
\pgfsetbuttcap%
\pgfsetroundjoin%
\pgfsetlinewidth{1.003750pt}%
\definecolor{currentstroke}{rgb}{0.941176,0.125490,0.000000}%
\pgfsetstrokecolor{currentstroke}%
\pgfsetdash{{3.700000pt}{1.600000pt}}{0.000000pt}%
\pgfpathmoveto{\pgfqpoint{1.257502in}{1.386277in}}%
\pgfpathlineto{\pgfqpoint{1.251262in}{0.690527in}}%
\pgfusepath{stroke}%
\end{pgfscope}%
\end{pgfpicture}%
\makeatother%
\endgroup%
  \end{center}

  \caption{Least squares regression finds the orthogonal projection of $\vy$ onto $\mathrm{span}\{\mX\}$ (here illustrated as the plane).}
  \label{fig:linear_regression_projection}
\end{marginfigure}

\subsection{Maximum Likelihood Estimation}\label{sec:least_squares_as_mle}

Since our function class comprises linear functions of the form $\transpose{\vw} \vx$, the observation model from \cref{eq:data} simplifies to \begin{align}
  y_i = \transpose{{\opt{\vw}}} \vx_i + \varepsilon_i \label{eq:linear_data}
\end{align} for some weight vector $\vw$, where for the purpose of this chapter we will additionally assume that $\varepsilon_i \sim \N{0}{\sigman^2}$ is homoscedastic Gaussian noise.\footnote[][0.5\baselineskip]{$\varepsilon_i$ is called \idx{additive white Gaussian noise}.}
This observation model is equivalently characterized by the Gaussian likelihood, \begin{align}
  y_i \mid \vx_i, \vw \sim \N{\transpose{\vw} \vx_i}{\sigman^2}. \label{eq:blr_likelihood} \margintag{using \cref{eq:cond_linear_gaussian}}
\end{align}
Based on this likelihood we can compute the MLE \eqref{eq:mle} of the weights: \begin{align*}
  \vwhat_\MLE &= \argmax_{\vw \in \R^d} \sum_{i=1}^n \log p(y_i \mid \vx_i, \vw) = \argmin_{\vw \in \R^d} \sum_{i=1}^n (y_i - \transpose{\vw} \vx_i)^2. \margintag{plugging in the Gaussian likelihood and simplifying}
\end{align*}
Note that therefore $\vwhat_\MLE = \vwhat_{\ls}$.

In practice, the noise variance $\sigman^2$ is typically unknown and also has to be determined, for example, through maximum likelihood estimation. It is a straightforward exercise to check that the MLE of $\sigman^2$ given fixed weights $\vw$ is $\hat{\sigma}_{\mathrm{n}}^2 = \frac{1}{n} \sum_{i=1}^n (y_i - \transpose{\vw} \vx_i)^2$ \exerciserefmark{noise_variance_mle}.

\section{Weight-space View}\pidx{weight-space view}\label{sec:blr:weight_space_view}

The most immediate and natural probabilistic interpretation of linear regression is to quantify uncertainty about the weights $\vw$.
Recall that probabilistic inference requires specification of a generative model comprised of prior and likelihood.
Throughout this chapter, we will use the Gaussian prior, \begin{align}
  \vw \sim \N{\vzero}{\sigmap^2 \mI},
\end{align} and the Gaussian likelihood from \cref{eq:blr_likelihood}.
We will discuss possible (probabilistic) strategies for choosing hyperparameters such as the prior variance $\sigmap^2$ and the noise variance $\sigman^2$ in \cref{sec:gp:model_selection}.

\begin{marginfigure}
  \incfig{blr_dgm}
  \caption{Directed graphical model of Bayesian linear regression in plate notation.}\label{fig:blr}
\end{marginfigure}

\begin{rmk}{Why a Gaussian prior?}{max_entropy_principle}
  The choice of using a Gaussian prior may seem somewhat arbitrary at first sight, except perhaps for the nice analytical properties of Gaussians that we have seen in \cref{sec:fundamentals:gaussians} and which will prove useful.
  The maximum entropy principle (cf. \cref{sec:fundamentals:inference:priors}) provides a more fundamental justification for Gaussian priors since turns out that $\mathcal{N}$ has the maximum entropy among all distributions on $\R^d$ with known mean and variance \exerciserefmark{maximum_entropy_principle}.
\end{rmk}

Next, let us derive the posterior distribution over the weights. \begin{align}
  &\log p(\vw \mid \vx_{1:n}, y_{1:n}) \nonumber \\[3pt]
  &= \log p(\vw) + \log p(y_{1:n} \mid \vx_{1:n}, \vw) + \const \margintag{by Bayes' rule \eqref{eq:bayes_rule}} \nonumber \\
  &= \log p(\vw) + \sum_{i=1}^n \log p(y_i \mid \vx_i, \vw) + \const \margintag{using independence of the samples} \nonumber \\
  &= -\frac{1}{2} \brackets*{\sigmap^{-2} \norm{\vw}_2^2 + \sigman^{-2} \sum_{i=1}^n (y_i - \transpose{\vw} \vx_i)^2} + \const \margintag{using the Gaussian prior and likelihood} \nonumber \\
  &= -\frac{1}{2} \brackets*{\sigmap^{-2} \norm{\vw}_2^2 + \sigman^{-2} \norm{\vy - \mX\vw}_2^2} + \const \margintag{using $\sum_{i=1}^n (y_i - \transpose{\vw} \vx_i)^2 = \norm{\vy - \mX\vw}_2^2$} \nonumber \\
  &= -\frac{1}{2} \brackets*{\sigmap^{-2} \transpose{\vw}\vw + \sigman^{-2} \parentheses*{\transpose{\vw}\transpose{\mX}\mX\vw - 2\transpose{\vy}\mX\vw + \transpose{\vy}\vy}} + \const \nonumber \\
  &= -\frac{1}{2} \brackets*{\transpose{\vw}(\sigman^{-2}\transpose{\mX}\mX + \sigmap^{-2}\mI)\vw - 2\sigman^{-2}\transpose{\vy}\mX\vw} + \const.
\end{align}
Observe that the log-posterior is a quadratic form in $\vw$, so the posterior distribution must be Gaussian: \begin{subequations}\begin{align}
  \vw &\mid \vx_{1:n}, y_{1:n} \sim \N{\vmu}{\mSigma} \margintag{see \cref{eq:gaussian_propto}}
  \intertext{where we can read off the mean and variance to be}
  \vmu &\defeq \sigman^{-2} \mSigma \transpose{\mX} \vy, \\
  \mSigma &\defeq \inv{\parentheses*{\sigman^{-2} \transpose{\mX} \mX + \sigmap^{-2} \mI}}. \label{eq:blr_posterior_var}
\end{align}\label{eq:blr_posterior}\end{subequations}
This also shows that Gaussians with known variance and linear likelihood are self-conjugate, a property that we had hinted at in \cref{sec:fundamentals:bayesian:conjugacy}.
It can be shown more generally that Gaussians with known variance are self-conjugate to any Gaussian likelihood \citep{murphy2007conjugate}.
For other generative models, the posterior can typically not be expressed in closed-form --- this is a very special property of Gaussians!

\subsection{Maximum a Posteriori Estimation}

Computing the MAP estimate for the weights, \begin{align}
  \vwhat_\MAP &= \argmax_\vw \log p(y_{1:n} \mid \vx_{1:n}, \vw) + \log p(\vw) \nonumber \\
  &= \argmin_\vw \norm{\vy - \mX \vw}_2^2 + \frac{\sigman^2}{\sigmap^2} \norm{\vw}_2^2, \margintag{using that the likelihood and prior are Gaussian} \label{eq:blr_as_ridge}
\end{align}
we observe that this is identical to ridge regression with weight decay ${\lambda \defeq \sigman^2 / \sigmap^2}$: $\vwhat_\MAP = \vwhat_{\ridge}$.
\Cref{eq:blr_as_ridge} is simply the MLE loss with an additional $L_2$-regularization (originating from the prior) that encourages keeping weights small.
Recall that the MAP estimate corresponds to the mode of the posterior distribution, which in the case of a Gaussian is simply its mean $\vmu$.
As to be expected, $\vmu$ coincides with the analytical solution to ridge regression from \cref{eq:ridge_regression}.

\begin{marginfigure}[2\baselineskip]
  \begin{center}
    \import{./plots/output/}{regularization.pgf}
  \end{center}

  \caption{Level sets of $L_2$- (\b{blue}) and $L_1$-regularization (\r{red}), corresponding to Gaussian and Laplace priors, respectively. It can be seen that $L_1$-regularization is more effective in encouraging sparse solutions (that is, solutions where many components are set to exactly $0$).}
\end{marginfigure}

\begin{ex}{Lasso as the MAP estimate with a Laplace prior}{}
  One problem with ridge regression is that the contribution of nearly-zero weights to the $L_2$-regularization term is negligible.
  Thus, $L_2$-regularization is typically not sufficient to perform variable selection (that is, set some weights to zero entirely), which is often desirable for interpretability of the model.

  A commonly used alternative to ridge regression is the \emph{least absolute shrinkage and selection operator} (or \midx{lasso}), which regularizes with the $L_1$-norm: \begin{align}
    \vwhat_{\lasso} \defeq \argmin_{\vw \in \R^d} \norm{\vy - \mX\vw}_2^2 + \lambda \norm{\vw}_1.
  \end{align}
  It turns out that lasso can also be viewed as probabilistic inference, using a Laplace prior ${\vw \sim \Laplace{\vzero}{h}}$ with length scale $h$ instead of a Gaussian prior.

  Computing the MAP estimate for the weights yields, \begin{align}
    \vwhat_\MAP &= \argmax_\vw \log p(y_{1:n} \mid \vx_{1:n}, \vw) + \log p(\vw) \nonumber \\
    &= \argmin_\vw \sum_{i=1}^n (y_i - \transpose{\vw} \vx_i)^2 + \frac{\sigman^2}{h} \norm{\vw}_1 \margintag{using that the likelihood is Gaussian and the prior is Laplacian}
  \end{align} which coincides with the lasso with weight decay ${\lambda \defeq \sigman^2 / h}$.
\end{ex}

To make predictions at a test point $\vxs$, we define the {(model-)}predicted point $\fs \defeq \transpose{\vwhat_{\MAP}} \vxs$ and obtain the label prediction \begin{align}
  \ys \mid \vxs, \vx_{1:n}, y_{1:n} \sim \N{\fs}{\sigman^2}.
\end{align}
Here we observe that using point estimates such as the MAP estimate does not quantify uncertainty in the weights.
The MAP estimate simply collapses all mass of the posterior around its mode.
This can be harmful when we are highly unsure about the best model, e.g., because we have observed insufficient data.

\subsection{Probabilistic Inference}

Rather than selecting a single weight vector $\vwhat$ to make predictions, we can use the full posterior distribution.
This is known as \midx{Bayesian linear regression} (BLR) and illustrated with an example in \cref{fig:blr_comparison}.

\begin{figure}
  \begin{center}
    \import{./plots/output/}{blr.pgf}
  \end{center}

  \caption{Comparison of \b{\textbf{linear regression (MLE)}}, \r{\textbf{ridge regression (MAP estimate)}}, and \textbf{Bayesian linear regression} when the data is generated according to \begin{align*}
    y \mid \vw, \vx \sim \N{\transpose{\vw}\vx}{\sigman^2}.
  \end{align*} The \textbf{true mean} is shown in black, the MLE in blue, and the MAP estimate in red. The dark gray area denotes the epistemic uncertainty of Bayesian linear regression and the light gray area the additional homoscedastic noise.
  On the left, $\sigman = 0.15$. On the right, $\sigman = 0.7$.}
  \label{fig:blr_comparison}
\end{figure}

To make predictions at a test point $\vxs$, we let $\fs \defeq \transpose{\vw} \vxs$ which has the distribution \begin{align}
  \fs \mid \vxs, \vx_{1:n}, y_{1:n} \sim \N{\transpose{\vmu} \vxs}{\transpose{\vxs} \mSigma \vxs}. \margintag{using the closedness of Gaussians under linear transformations \eqref{eq:gaussian_lin_trans}}
\end{align}
Note that this does not take into account the noise in the labels $\sigman^2$.
For the label prediction $\ys$, we obtain \begin{align}
  \ys \mid \vxs, \vx_{1:n}, y_{1:n} \sim \N{\transpose{\vmu} \vxs}{\transpose{\vxs} \mSigma \vxs + \sigman^2}. \label{eq:blr_pred_posterior} \margintag{using additivity of Gaussians \eqref{eq:gaussian_additivity}}
\end{align}

\subsection{Recursive Probabilistic Inference}\label{sec:blr:online}

We have already discussed the recursive properties of probabilistic inference in \cref{sec:fundamentals:recursive_inference}.
For Bayesian linear regression with a Gaussian prior and likelihood, this principle can be used to derive an efficient online algorithm since also the posterior is a Gaussian, \begin{align}
  p^{(t)}(\vw) = \N[\vw]{\vmu^{(t)}}{\mSigma^{(t)}},
\end{align} which can be stored efficiently using only $\BigO{d^2}$ parameters.
This leads to an efficient online algorithm for Bayesian linear regression with time-independent(!) memory complexity $\BigO{d}$ and round complexity $\BigO{d^2}$ \exerciserefmark{online_blr}.
The interpretation of Bayesian linear regression as an online algorithm also highlights similarities to other sequential models such as Kalman filters, which we discuss in \cref{sec:kf}.
In \cref{ex:blr_as_kf}, we will learn that online Bayesian linear regression is, in fact, an example of a Kalman filter.

\section{Aleatoric and Epistemic Uncertainty}\label{sec:blr:uncertainty}

The predictive posterior distribution from \cref{eq:blr_pred_posterior} highlights a decomposition of uncertainty wherein $\transpose{\vxs} \mSigma \vxs$ corresponds to the uncertainty about our model due to the lack of data (commonly referred to as the \midx{epistemic uncertainty}) and $\sigman^2$ corresponds to the uncertainty about the labels that cannot be explained by the inputs and any model from the model class (commonly referred to as the \midx{aleatoric uncertainty}, ``irreducible noise'', or simply ``(label) noise'') \exerciserefmark{aleatoric_and_epistemic_uncertainty}.

A natural probabilistic approach is to represent epistemic uncertainty with a probability distribution over models.
Intuitively, the variance of this distribution measures our uncertainty about the model and its mode corresponds to our current best (point) estimate.
The distribution over weights of a linear model is one example, and we will continue to explore this approach for other models in the following chapters.\looseness=-1

It is a practical modeling choice how much inaccuracy to attribute to epistemic or aleatoric uncertainty.
Generally, when a poor model is used to explain a process, more inaccuracy has to be attributed to irreducible noise.
For example, when a linear model is used to ``explain'' a nonlinear process, most uncertainty is aleatoric as the model cannot explain the data well.
As we use more expressive models, a larger portion of the uncertainty can be explained by the data.

Epistemic and aleatoric uncertainty can be formally defined in terms of the law of total variance \eqref{eq:lotv}, \begin{align}
  \Var{\ys \mid \vxs} = \b{\underbrace{\E[\vtheta]{\Var[\ys]{\ys \mid \vxs, \vtheta}}}_{\text{aleatoric uncertainty}}} + \r{\underbrace{\Var[\vtheta]{\E[\ys]{\ys \mid \vxs, \vtheta}}}_{\text{epistemic uncertainty}}}. \label{eq:lotv_interpretation}
\end{align}
Here, the mean variability of predictions $\ys$ averaged across all models $\vtheta$ is the estimated \midx{aleatoric uncertainty}.
In contrast, the variability of the mean prediction $\ys$ under each model $\vtheta$ is the estimated \midx{epistemic uncertainty}.
This decomposition of uncertainty will appear frequently throughout this manuscript.\looseness=-1

\section{Non-linear Regression}\label{sec:blr:non_linear}

We can use linear regression not only to learn linear functions.
The trick is to apply a nonlinear transformation $\vphi : \R^d \to \R^e$ to the features $\vx_i$, where $d$ is the dimension of the input space and $e$ is the dimension of the designed \midx{feature space}.
We denote the design matrix comprised of transformed features by $\mPhi \in \R^{n \times e}$.
Note that if the feature transformation $\vphi$ is the identity function then $\mPhi = \mX$.

\begin{marginfigure}
  \begin{center}
    \import{./plots/output/}{poly_linear_regression.pgf}
  \end{center}

  \caption{Applying linear regression with a feature space of polynomials of degree 10. The \b{\textbf{least squares estimate}} is shown in blue, \r{\textbf{ridge regression}} in red, and \g{\textbf{lasso}} in green.}
\end{marginfigure}

\begin{ex}{Polynomial regression}{polynomial_regression}
  Let ${\vphi(x) \defeq [x^2, x, 1]}$ and ${\vw \defeq [a, b, c]}$.
  Then the function that our model learns is given as \begin{align*}
    f = a x^2 + b x + c.
  \end{align*}
  Thus, our model can exactly represent all polynomials up to degree 2.

  However, to learn polynomials of degree $m$ in $d$ input dimensions, we need to apply the nonlinear transformation \begin{align*}
    \vphi(\vx) = [&1, x_1, \dots, x_d, x_1^2, \dots, x_d^2, x_1 \cdot x_2, \dots, x_{d-1} \cdot x_d, \\
    &\dots, \\
    &x_{d-m+1} \cdot \dots \cdot x_d].
  \end{align*}

  Note that the feature dimension $e$ is $\sum_{i=0}^m {d+i-1 \choose i} = \Theta(d^m)$.\safefootnote{Observe that the vector contains ${d+i-1 \choose i}$ monomials of degree $i$ as this is the number of ways to choose $i$ times from $d$ items with replacement and without consideration of order. To see this, consider the following encoding: We take a sequence of $d+i-1$ spots. Selecting any subset of $i$ spots, we interpret the remaining $d-1$ spots as ``barriers'' separating each of the $d$ items. The selected spots correspond to the number of times each item has been selected. For example, if $2$ items are to be selected out of a total of $4$ items with replacement, one possible configuration is ``$\circ\mid\mid\circ\mid$'' where $\circ$ denotes a selected spot and $\mid$ denotes a barrier. This configuration encodes that the first and third item have each been chosen once. The number of possible configurations --- each encoding a unique outcome --- is therefore ${d+i-1 \choose i}$.}
  Thus, the dimension of the feature space grows exponentially in the degree of polynomials and input dimensions.
  Even for relatively small $m$ and $d$, this becomes completely unmanageable.
\end{ex}

The example of polynomials highlights that it may be inefficient to keep track of the weights $\vw \in \R^e$ when $e$ is large, and that it may be useful to instead consider a reparameterization which is of dimension~$n$ rather than of the feature dimension.

\section{Function-space View}\label{sec:blr:function_space_view}

Let us now look at Bayesian linear regression through a different lens.
Previously, we have been interpreting it as a distribution over the weights $\vw$ of a linear function $\vf = \mPhi \vw$.
The key idea is that for a finite set of inputs (ensuring that the design matrix is well-defined), we can equivalently consider a distribution directly over the estimated function values $\vf$.
We call this the \midx{function-space view} of Bayesian linear regression.\looseness=-1

Instead of considering a prior over the weights $\vw \sim \N{\vzero}{\sigmap^2 \mI}$ as we have done previously, we now impose a prior directly on the values of our model at the observations.
Using that Gaussians are closed under linear maps \eqref{eq:gaussian_lin_trans}, we obtain the equivalent prior \begin{align}
  \vf \mid \mX \sim \N{\mPhi \E{\vw}}{\mPhi \Var{\vw} \transpose{\mPhi}} = \N{\vzero} {\underbrace{\sigmap^2 \mPhi \transpose{\mPhi}}_{\mK}} \label{eq:kernel_matrix_defn}
\end{align} where $\mK \in \R^{n \times n}$ is the so-called \midx{kernel matrix}.
Observe that the entries of the kernel matrix can be expressed as $\mK(i,j) = \sigmap^2 \cdot \transpose{\vphi(\vx_i)} \vphi(\vx_j)$.

\begin{marginfigure}
  \begin{center}
    \import{./plots/output/}{function_space_view.pgf}
  \end{center}

  \caption{An illustration of the function-space view. The model is described by the points $(x_i, f_i)$.}
\end{marginfigure}

You may say that nothing has changed, and you would be right --- that is precisely the point.
Note, however, that the shape of the kernel matrix is $n \times n$ rather than the $e \times e$ covariance matrix over weights, which becomes unmanageable when $e$ is large.
The kernel matrix $\mK$ has entries only for the finite set of observed inputs.
However, in principle, we could have observed any input, and this motivates the definition of the \midx{kernel function} \begin{align}
  k(\vx, \vxp) \defeq \sigmap^2 \cdot \transpose{\vphi(\vx)} \vphi(\vxp) \label{eq:blr_kernel}
\end{align} for arbitrary inputs $\vx$ and $\vxp$.
A kernel matrix is simply a finite ``view'' of the kernel function, \begin{align}
  \mK = \begin{bmatrix}
    k(\vx_1, \vx_1) & \cdots & k(\vx_1, \vx_n) \\
    \vdots & \ddots & \vdots \\
    k(\vx_n, \vx_1) & \cdots & k(\vx_n, \vx_n) \\
  \end{bmatrix}
\end{align}

Observe that by definition of the kernel matrix in \cref{eq:kernel_matrix_defn}, the kernel matrix is a covariance matrix and the kernel function measures the covariance of the function values $f(\vx)$ and $f(\vxp)$ given inputs~$\vx$ and~$\vxp$: \begin{align}
  k(\vx, \vxp) = \Cov{f(\vx), f(\vxp)}.
\end{align}

Moreover, note that we have reformulated\footnote{we often say ``kernelized''\pidx{kernelization}} the learning algorithm such that the feature space is now \emph{implicit} in the choice of kernel, and the kernel is defined by inner products of (nonlinearly transformed) inputs.
In other words, the choice of kernel implicitly determines the class of functions that $\vf$ is sampled from (without expressing the functions explicitly in closed-form), which encodes our prior beliefs.
This is known as the \midx{kernel trick}.

\subsection{Learning and Predictions}\label{sec:blr:learning_and_inference}

We have already kernelized the Bayesian linear regression prior.
The posterior distribution $\vf \mid \mX, \vy$ is again Gaussian due to the closedness properties of Gaussians, analogously to our derivation of the prior kernel matrix in \cref{eq:kernel_matrix_defn}.

It remains to show that we can also rely on the kernel trick for predictions.
Given the test point $\vxs$, we define \begin{align*}
  \Tilde{\mPhi} \defeq \begin{bmatrix}
    \mPhi \\
    \transpose{\vphi(\vxs)} \\
  \end{bmatrix}, \quad \Tilde{\vy} \defeq \begin{bmatrix}
    \vy \\
    \ys \\
  \end{bmatrix}, \quad \Tilde{\vf} \defeq \begin{bmatrix}
    \vf \\
    \fs \\
  \end{bmatrix}.
\end{align*}
We immediately obtain $\Tilde{\vf} = \Tilde{\mPhi} \vw$.
Analogously to our analysis of predictions from the weight-space view, we add the label noise to obtain the estimate $\Tilde{\vy} = \Tilde{\vf} + \Tilde{\vvarepsilon}$ where $\Tilde{\vvarepsilon} \defeq \transpose{[\varepsilon_1 \; \cdots \; \varepsilon_n \; \opt{\varepsilon}]} \sim \N{\vzero}{\sigman^2 \mI}$ is the independent label noise.
Applying the same reasoning as we did for the prior, we obtain \begin{align}
  \Tilde{\vf} \mid \mX, \vxs \sim \N{\vzero}{\Tilde{\mK}}
\end{align} where $\Tilde{\mK} \defeq \sigmap^2 \Tilde{\mPhi} \transpose{\Tilde{\mPhi}}$.
Adding the label noise yields \begin{align}
  \Tilde{\vy} \mid \mX, \vxs \sim \N{\vzero}{\Tilde{\mK} + \sigman^2 \mI}.
\end{align}
Finally, we can conclude from the closedness of Gaussian random vectors under conditional distributions \eqref{eq:cond_gaussian} that the predictive posterior ${\ys \mid \vxs, \mX, \vy}$ follows again a normal distribution.
We will do a full derivation of the posterior and predictive posterior in \cref{sec:gp:learning_and_inference}.

\subsection{Efficient Polynomial Regression}

But how does the kernel trick address our concerns about efficiency raised in \cref{sec:blr:non_linear}?
After all, computing the kernel for a feature space of dimension $e$ still requires computing sums of length $e$ which is prohibitive when $e$ is large.
The kernel trick opens up a couple of new doors for us: \begin{enumerate}
  \item For certain feature transformations $\vphi$, we may be able to find an easier to compute expression equivalent to $\transpose{\vphi(\vx)} \vphi(\vxp)$.
  \item If this is not possible, we could approximate the inner product by an easier to compute expression.
  \item Or, alternatively, we may decide not to care very much about the exact feature transformation and simply experiment with kernels that induce \emph{some} feature space (which may even be infinitely dimensional).
\end{enumerate}
We will explore the third approach when we revisit kernels in \cref{sec:gp:kernel_functions}.
A polynomial feature transformation can be computed efficiently in closed-form.

\begin{fct}
  For the polynomial feature transformation $\vphi$ up to degree $m$ from \cref{ex:polynomial_regression}, it can be shown that up to constant factors, \begin{align}
    \transpose{\vphi(\vx)} \vphi(\vxp) = (1 + \transpose{\vx} \vxp)^m.
  \end{align}
\end{fct}
For example, for input dimension $2$, the kernel $(1 + \transpose{\vx}\vxp)^2$ corresponds to the feature vector $\vphi(\vx) = \transpose{[1 \;\; \sqrt{2} x_1 \;\; \sqrt{2} x_2 \;\; \sqrt{2} x_1 x_2 \;\; x_1^2 \;\; x_2^2]}$.

\section*{Discussion}

We have explored a probabilistic perspective on linear models, and seen that classical approaches such as least squares and ridge regression can be interpreted as approximate probabilistic inference.
We then saw that we can even perform \emph{exact} probabilistic inference efficiently if we adopt a Gaussian prior and Gaussian noise assumption.
These are already powerful tools, which are often applied also to nonlinear models if we treat the latent feature space --- which was either human-designed or learned via deep learning --- as fixed.
In the next chapter, we will digress briefly from the storyline on ``learning'' to see how we can adopt a similar probabilistic perspective to track latent states over time.
Then, in \cref{sec:gp}, we will see how we can use the function-space view and kernel trick to learn flexible nonlinear models with exact probabilistic inference, without ever explicitly representing the feature space.

\excheading

\begin{nexercise}{Closed-form linear regression}{closed_form_linear_regression}
  \begin{enumerate}
    \item Derive the unique solutions to least squares and ridge regression from \cref{eq:linear_regression,eq:ridge_regression}.

    \item For an $n \times m$ matrix $\mA$ and vector $\vx \in \R^m$, we call $\mPi_{\mA} \vx$ the orthogonal projection of $\vx$ onto $\mathrm{span}\{\mA\} = \{\mA \vxp \mid \vxp \in \R^m\}$.
    In particular, an orthogonal projection satisfies $\vx - \mPi_{\mA} \vx \perp \mA \vxp$ for all $\vxp \in \R^m$.\par
    Show that $\vwhat_{\ls}$ from \cref{eq:linear_regression} is such that $\mX \vwhat_{\ls}$ is the unique closest point to $\vy$ on $\mathrm{span}\{\mX\}$, i.e., it satisfies $\mX \vwhat_{\ls} = \mPi_{\mX} \vy$.
  \end{enumerate}
\end{nexercise}

\begin{nexercise}{MLE of noise variance}{noise_variance_mle}
  Show that the MLE of $\sigman^2$ given fixed weights $\vw$ is \begin{align}
    \hat{\sigma}_n^2 = \frac{1}{n} \sum_{i=1}^n (y_i - \transpose{\vw} \vx_i)^2.
  \end{align}
\end{nexercise}

\begin{nexercise}{Variance of least squares around training data}{variance_around_training_data}
  Show that the variance of a prediction at the point $\transpose{[1 \; \xs]}$ is smallest when $\xs$ is the mean of the training data.
  More formally, show that if inputs are of the form $\vx_i = \transpose{[1 \; x_i]}$ where $x_i \in \R$ and $\vwhat_{\ls}$ is the least squares estimate, then $\Var{\ys \mid \transpose{[1 \; \xs]}, \vwhat_{\ls}}$ is minimized for $\xs = \frac{1}{n} \sum_{i=1}^n x_i$.
\end{nexercise}

\begin{nexercise}{Bayesian linear regression}{blr}
  Suppose you are given the following observations \begin{align*}
    \mX = \begin{bmatrix}
      1 & 1 \\
      1 & 2 \\
      2 & 1 \\
      2 & 2 \\
    \end{bmatrix}, \quad \vy = \begin{bmatrix}
      2.4 \\
      4.3 \\
      3.1 \\
      4.9 \\
    \end{bmatrix}
  \end{align*} and assume the data follows a linear model with homoscedastic noise $\N{0}{\sigman^2}$ where $\sigman^2 = 0.1$.

  \begin{enumerate}
    \item Find the maximum likelihood estimate $\vwhat_\MLE$ given the data.
    \item Now assume that we have a prior $p(\vw) = \N[\vw]{\vzero}{\sigmap^2 \mI}$ with ${\sigmap^2 = 0.05}$.
    Find the MAP estimate $\vwhat_\MAP$ given the data and the prior.
    \item Use the posterior $p(\vw \mid \mX, \vy)$ to get a posterior prediction for the label $\ys$ at $\vxs = \transpose{[3 \quad 3]}$.
    Report the mean and the variance of this prediction.
    \item How would you have to change the prior $p(\vw)$ such that \begin{align*}
      \vwhat_\MAP \to \vwhat_\MLE?
    \end{align*}
  \end{enumerate}
\end{nexercise}

\begin{nexercise}{Online Bayesian linear regression}{online_blr}
  \begin{enumerate}
    \item Can you design an algorithm that updates the posterior (as opposed to recalculating it from scratch using \cref{eq:blr_posterior}) in a smarter way?
    The requirement is that the memory should not grow as $\BigO{t}$.
    \item If $d$ is large, computing the inverse every round is very expensive.
    Can you use the recursive structure you found in the previous question to bring down the computational complexity of every round to $\BigO{d^2}$?\par
  \end{enumerate} The resulting efficient online algorithm is known as \midx{online Bayesian linear regression}.
\end{nexercise}

\begin{nexercise}{Aleatoric and epistemic uncertainty of BLR}{aleatoric_and_epistemic_uncertainty}
  Prove for Bayesian linear regression that $\transpose{\vxs}\mSigma\vxs$ is the epistemic uncertainty and $\sigman^2$ the aleatoric uncertainty in $\ys$ under the decomposition of \cref{eq:lotv_interpretation}.
\end{nexercise}

\begin{nexercise}{Hyperpriors}{hyperpriors}
  We consider a dataset \(\{(\vx_i,y_i)\}_{i=1}^n\) of size \(n\), where \(\vx_i \in \R^d\) denotes the feature vector and \(y_i \in \R\) denotes the label of the \(i\)-th data point.
  Let \(\varepsilon_i\) be i.i.d.~samples from the Gaussian distribution \(\N{0}{\lambda^{-1}}\) for a given \(\lambda > 0\).
  We collect the labels in a vector \(\vy \in \R^n\), the features in a matrix \(\mX \in \R^{n \times d}\), and the noise in a vector \(\vvarepsilon \in \R^n\).
  The labels are generated according to \(\vy = \mX \vw + \vvarepsilon.\)\par
  To perform Bayesian Linear Regression, we consider the prior distribution over the parameter vector \(\vw\) to be \( \N{\vmu}{\lambda^{-1} \mI_d}\), where \(\mI_d\) denotes the \(d\)-dimensional identity matrix and \(\vmu \in \R^d\) is a hyperparameter.

  \begin{enumerate}
    \item Given this Bayesian data model, what is the conditional covariance matrix \( \mSigma_{\vy} \defeq \Var{\vy}[\mX, \vmu, \lambda] \)?

    \item Calculate the maximum likelihood estimate of the hyperparameter~\( \vmu \).

    \item Since we are unsure about the hyperparameter \( \vmu \), we decide to model our uncertainty about \( \vmu \) by placing the ``\midx{hyperprior}'' \({\vmu \sim \N{\vzero}{\mI_d}}\). Is the posterior distribution \(p(\vmu \mid \mX, \vy, \lambda)\) a Gaussian distribution? If yes, what are its mean vector and covariance matrix?

    \item What is the posterior distribution \( p(\lambda \mid \mX, \vy, \vmu) \)?\par
    \textit{Hint: For any \( a \in \R, \mA \in \R^{n \times n} \) it holds that \( \det{a \mA} = a^n \det{\mA} \).}
  \end{enumerate}
\end{nexercise}

  \chapter{Filtering}\label{sec:kf}\pidx{filtering}[idxpagebf]

Before we continue in \cref{sec:gp} with the function-space view of regression, we want to look at a seemingly different but very related problem.
We will study Bayesian learning and inference in the \midx{state space model}, where we want to keep track of the state of an agent over time based on noisy observations.
In this model, we have a sequence of (hidden) states $(\rX_t)_{t\in\Nat_0}$ where $\rX_t$ is in $\R^d$ and a sequence of observations $(\rY_t)_{t\in\Nat_0}$ where $\rY_t$ is in $\R^m$.%

\begin{figure}
  \incfig[0.75\textwidth]{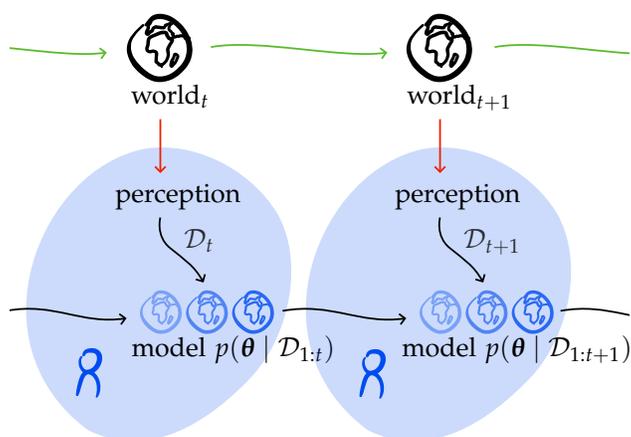}
  \caption{Schematic view of Bayesian filtering: An agent perceives the current state of the world and updates its belief accordingly.}
  \label{fig:bayesian_filtering}
\end{figure}

The process of keeping track of the state using noisy observations is also known as \midx{Bayesian filtering} or \midx{recursive Bayesian estimation}.
\Cref{fig:bayesian_filtering} illustrates this process, where an agent perceives the current state of the world and then updates its beliefs about the state based on this observation.

We will discuss Bayesian filtering more broadly in the next section.
A Kalman filter is an important special case of a Bayes' filter, which uses a Gaussian prior over the states and conditional linear Gaussians to describe the evolution of states and observations.
Analogously to the previous chapter, we will see that inference in this model is tractable due to the closedness properties of Gaussians.

\begin{defn}[Kalman filter]\pidx{Kalman filter}
  A \emph{Kalman filter} is specified by a Gaussian prior over the states, \begin{align}
    \rX_0 \sim \N{\vmu}{\mSigma},
  \end{align} and a conditional linear Gaussian \emph{motion model} and \emph{sensor model}, \begin{align}
    \rX_{t+1} &\defeq \mF \rX_t + \vvarepsilon_t \quad &&\mF \in \R^{d \times d}, \vvarepsilon_t \sim \N{\vzero}{\mSigma_x}, \label{eq:kf_motion_model} \\
    \rY_t     &\defeq \mH \rX_t + \veta_t \quad     &&\mH \in \R^{m \times d}, \veta_t \sim \N{\vzero}{\mSigma_y}, \label{eq:kf_sensor_model}
  \end{align} respectively.
  The motion model is sometimes also called \emph{transition model} or \emph{dynamics model}.
  Crucially, Kalman filters assume that $\mF$ and $\mH$ are known.
  In general, $\mF$ and $\mH$ may depend on $t$.
  Also, $\vvarepsilon$ and $\veta$ may have a non-zero mean, commonly called a ``drift''.
\end{defn}

\begin{marginfigure}
  \incfig{kf}
  \caption{Directed graphical model of a Kalman filter with hidden states $\rX_t$ and observables $\rY_t$.}\label{fig:kf}
\end{marginfigure}

Because Kalman filters use conditional linear Gaussians, which we have already seen in \cref{eq:cond_linear_gaussian}, their joint distribution (over all variables) is also Gaussian.
This means that predicting the future states of a Kalman filter is simply inference with multivariate Gaussians.
In Bayesian filtering, however, we do not only want to make predictions occasionally.
In Bayesian filtering, we want to \emph{keep track} of states, that is, predict the current state of an agent online.\footnote{Here, \idx{online} is common terminology to say that we want to perform inference at time $t$ without exposure to times $t+1, t+2, \dots$, so in ``real-time''.}
To do this efficiently, we need to update our \midx{belief} about the state of the agent recursively, similarly to our recursive Bayesian updates in Bayesian linear regression (see \cref{sec:blr:online}).

From the directed graphical model of a Kalman filter shown in \cref{fig:kf}, we can immediately gather the following conditional independence relations,\footnote{Alternatively, they follow from the definition of the motion and sensor models as linear updates.}\looseness=-1 \begin{align}
  \rX_{t+1} &\perp \rX_{1:t-1}, \rY_{1:t-1} \mid \rX_t, \label{eq:kf_cond_ind1} \\
  \rY_t     &\perp \rX_{1:t-1} \mid \rX_t \label{eq:kf_cond_ind2} \\
  \rY_t     &\perp \rY_{1:t-1} \mid \rX_{t-1}. \label{eq:kf_cond_ind3}
\end{align}
The first conditional independence property is also known as the \midx{Markov property}, which we will return to later in our discussion of Markov chains and Markov decision processes.
This characterization of the Kalman filter, yields the following factorization of the joint distribution:\looseness=-1 \begin{align}
  p(\vx_{1:t}, \vy_{1:t}) &= \prod_{i=1}^t p(\vx_i \mid \vx_{1:i-1}) p(\vy_i \mid \vx_{1:t}, \vy_{1:i-1}) \margintag{using the product rule \eqref{eq:product_rule}} \nonumber \\
  &= p(\vx_1) p(\vy_1 \mid \vx_1) \prod_{i=2}^t p(\vx_i \mid \vx_{i-1}) p(\vy_i \mid \vx_i). \margintag{using the conditional independence properties from \eqref{eq:kf_cond_ind1}, \eqref{eq:kf_cond_ind2}, and \eqref{eq:kf_cond_ind3}}
\end{align}

\section{Conditioning and Prediction}\label{sec:kf:bayesian_filtering}

We can describe Bayesian filtering by the following recursive scheme with the two phases, \emph{conditioning} (also called ``update'') and \emph{prediction}:

\begin{algorithm}
  \caption{Bayesian filtering}
  start with a prior over initial states $p(\vx_0)$\;
  \For{$t = 1$ \KwTo $\infty$}{
    assume we have $p(\vx_t \mid \vy_{1:t-1})$\;
    \textbf{conditioning}: compute $p(\vx_t \mid \vy_{1:t})$ using the new observation $\vy_t$\;
    \textbf{prediction}: compute $p(\vx_{t+1} \mid \vy_{1:t})$\;
  }
\end{algorithm}

Let us consider the conditioning step first: \begin{align}
  p(\vx_t \mid \vy_{1:t}) &= \frac{1}{Z} p(\vx_t \mid \vy_{1:t-1}) p(\vy_t \mid \vx_t, \vy_{1:t-1}) \margintag{using Bayes' rule \eqref{eq:bayes_rule}} \nonumber \\
  &= \frac{1}{Z} p(\vx_t \mid \vy_{1:t-1}) p(\vy_t \mid \vx_t). \margintag{using the conditional independence structure \eqref{eq:kf_cond_ind3}} \label{eq:bf_conditioning}
\end{align}
For the prediction step, we obtain, \begin{align}
  p(\vx_{t+1} \mid \vy_{1:t}) &= \int p(\vx_{t+1}, \vx_t \mid \vy_{1:t}) \,d\vx_t \margintag{using the sum rule \eqref{eq:sum_rule}} \nonumber \\
  &= \int p(\vx_{t+1} \mid \vx_t, y_{1:t}) p(\vx_t \mid y_{1:t}) \,d\vx_t \margintag{using the product rule \eqref{eq:product_rule}} \nonumber \\
  &= \int p(\vx_{t+1} \mid \vx_t) p(\vx_t \mid y_{1:t}) \,d\vx_t. \margintag{using the conditional independence structure \eqref{eq:kf_cond_ind1}} \label{eq:bf_prediction}
\end{align}
In general, these distributions can be very complicated, but for Gaussians (i.e., Kalman filters) they can be expressed in closed-form.

\begin{rmk}{Bayesian smoothing}{bs}
  \midx{Bayesian smoothing} is a closely related task to Bayesian filtering.
  While Bayesian filtering methods estimate the current state based only on observations obtained before and at the current time step, Bayesian smoothing computes the distribution of ${\rX_k \mid \vy_{1:t}}$ where ${t > k}$.
  That is Bayesian smoothing estimates $\rX_k$ based on data until \emph{and beyond} time $k$.
  Note that if ${k = t}$, then Bayesian smoothing coincides with Bayesian filtering.

  Analogously to \cref{eq:bf_conditioning}, \begin{align}
    p(\vx_k \mid \vy_{1:t}) \propto p(\vx_k \mid \vy_{1:k}) p(\vy_{k+1:t} \mid \vx_k). \label{eq:bs}
  \end{align}
  If we assume a Gaussian prior and conditional Gaussian transition and dynamics models (this is called \midx{Kalman smoothing}), then by the closedness properties of Gaussians, $\rX_k \mid \vy_{1:t}$ is a Gaussian.
  Indeed, all terms of \cref{eq:bs} are Gaussian PDFs and as seen in \cref{eq:gaussian_pdf_product}, the product of two Gaussian PDFs is again proportional to a Gaussian PDF.\looseness=-1

  The first term, $\rX_k \mid \vy_{1:k}$, is the marginal posterior of the hidden states of the Kalman filter which can be obtained with Bayesian filtering.\looseness=-1

  By conditioning on $\rX_{k+1}$, we have for the second term, \begin{align}
    p(\vy_{k+1:t} \mid \vx_k) &= \int p(\vy_{k+1:t} \mid \vx_k, \vx_{k+1}) p(\vx_{k+1} \mid \vx_k) \,d\vx_{k+1} \margintag{using the sum rule \eqref{eq:sum_rule} and product rule \eqref{eq:product_rule}} \nonumber\\
    &= \int p(\vy_{k+1:t} \mid \vx_{k+1}) p(\vx_{k+1} \mid \vx_k) \,d\vx_{k+1} \margintag{using the conditional independence structure \eqref{eq:kf_cond_ind2}} \nonumber\\
    &= \int p(\vy_{k+1} \mid \vx_{k+1}) p(\vy_{k+2:t} \mid \vx_{k+1}) p(\vx_{k+1} \mid \vx_k) \,d\vx_{k+1} \margintag{using the conditional independence structure \eqref{eq:kf_cond_ind3}}
  \end{align}
  Let us have a look at the terms in the product: \begin{itemize}
    \item $p(\vy_{k+1} \mid \vx_{k+1})$ is obtained from the sensor model,
    \item $p(\vx_{k+1} \mid \vx_k)$ is obtained from the transition model, and
    \item ${p(\vy_{k+2:t} \mid \vx_{k+1})}$ can be computed recursively backwards in time.
  \end{itemize}
  This recursion results in linear equations resembling a Kalman filter running backwards in time.

  Thus, in the setting of Kalman smoothing, both factors of \cref{eq:bs} can be computed efficiently: one using a (forward) Kalman filter; the other using a ``backward'' Kalman filter.
  More concretely, in time $\BigO{t}$, we can compute the two factors for all $k \in [t]$.
  This approach is known as \midx{two-filter smoothing} or the \midx{forward-backward algorithm}.\looseness=-1
\end{rmk}

\section{Kalman Filters}

Let us return to the setting of Kalman filters where priors and likelihoods are Gaussian.
Here, we will see that the update and prediction steps can be computed in closed form.

\subsection{Conditioning}

The conditioning operation in Kalman filters is also called the Kalman update.
Before introducing the general Kalman update, let us consider a simpler example:\looseness=-1

\begin{ex}{Random walk in 1d}{kf_rand_walk_1d}
  We use the simple motion and sensor models,\safefootnote{This corresponds to $\mF = \mH = \mI$ and a drift of $0$.} \begin{subequations}\begin{align}
    X_{t+1} \mid x_t &\sim \N{x_t}{\sigma_x^2}, \label{eq:kf_1d_motion_model} \\
    Y_t \mid x_t     &\sim \N{x_t}{\sigma_y^2}. \label{eq:kf_1d_sensor_model}
  \end{align}\end{subequations}
  Let ${X_t \mid y_{1:t} \sim \N{\mu_t}{\sigma_t^2}}$ be our belief at time $t$.
  It can be shown that Bayesian filtering yields the belief ${X_{t+1} \mid y_{1:t+1} \sim \N{\mu_{t+1}}{\sigma_{t+1}^2}}$ at time $t+1$ where \exerciserefmark{kf_predictive_distr} \begin{align}
    \mu_{t+1} \defeq \frac{\sigma_y^2 \mu_t + (\sigma_t^2 + \sigma_x^2) y_{t+1}}{\sigma_t^2 + \sigma_x^2 + \sigma_y^2}, \quad \sigma_{t+1}^2 \defeq \frac{(\sigma_t^2 + \sigma_x^2) \sigma_y^2}{\sigma_t^2 + \sigma_x^2 + \sigma_y^2}. \label{eq:kalman_posterior}
  \end{align}
  Although looking intimidating at first, this update has a very natural interpretation.
  Let us define the following quantity, \begin{align}
    \lambda \defeq \frac{\sigma_t^2 + \sigma_x^2}{\sigma_t^2 + \sigma_x^2 + \sigma_y^2} = 1 - \frac{\sigma_y^2}{\sigma_t^2 + \sigma_x^2 + \sigma_y^2} \in [0,1]. \label{eq:kalman_gain_1d}
  \end{align}
  Using $\lambda$, we can write the updated mean as a convex combination of the previous mean and the observation, \begin{align}
    \mu_{t+1} &= (1-\lambda)\mu_t + \lambda y_{t+1} \\
    &= \mu_t + \lambda(y_{t+1} - \mu_t). \label{eq:kf_update_1d}
  \end{align}
  Intuitively, $\lambda$ is a form of ``gain'' that influences how much of the new information should be incorporated into the updated mean.
  For this reason, $\lambda$ is also called \midx{Kalman gain}.

  The updated variance can similarly be rewritten, \begin{align}
    \sigma_{t+1}^2 = \lambda \sigma_y^2 = (1-\lambda)(\sigma_t^2 + \sigma_x^2). \label{eq:kalman_variance_1d}
  \end{align}

  In particular, observe that if ${\mu_t = y_{t+1}}$ (i.e., we observe our prediction), we have ${\mu_{t+1} = \mu_t}$ as there is no new information.
  Similarly, for $\sigma_y^2 \to \infty$ (i.e., we do not trust our observations), we have \begin{align*}
    \lambda \to 0, \quad \mu_{t+1} = \mu_t, \quad \sigma_{t+1}^2 = \sigma_t^2 + \sigma_x^2.
  \end{align*}
  In contrast, for $\sigma_y^2 \to 0$, we have \begin{align*}
    \lambda \to 1, \quad \mu_{t+1} = y_{t+1}, \quad \sigma_{t+1}^2 = 0.
  \end{align*}
\end{ex}

\begin{marginfigure}[-45\baselineskip]
  \begin{center}
    \import{./plots/output/}{1d_random_walk.pgf}
  \end{center}

  \caption{Hidden states during a random walk in one dimension.}
\end{marginfigure}

The general formulas for the \midx{Kalman update} follow the same logic as in the above example of a one-dimensional random walk.
Given the prior belief ${\rX_t \mid \vy_{1:t} \sim \N{\vmu_t}{\mSigma_t}}$, we have \begin{subequations}\begin{align}
  \rX_{t+1} \mid \vy_{1:t+1} &\sim \N{\vmu_{t+1}}{\mSigma_{t+1}} \quad\text{where} \\
  \vmu_{t+1} &\defeq \mF \vmu_t + \mK_{t+1}(\vy_{t+1} - \mH \mF \vmu_t), \\
  \mSigma_{t+1} &\defeq (\mI - \mK_{t+1} \mH)(\mF \mSigma_t \transpose{\mF} + \mSigma_x). \label{eq:kalman_update_covariance_matrix}
  \intertext{Hereby, $\mK_{t+1}$ is the \midx{Kalman gain}[idxpagebf],}
  \mK_{t+1} &\defeq (\mF \mSigma_t \transpose{\mF} + \mSigma_x) \transpose{\mH} \inv{(\mH (\mF \mSigma_t \transpose{\mF} + \mSigma_x) \transpose{\mH} + \mSigma_y)} \in \R^{d \times m}.
\end{align}\label{eq:kalman_update}\end{subequations}
Note that $\mSigma_t$ and $\mK_t$ can be computed offline as they are independent of the observation $\vy_{t+1}$. $\mF\vmu_t$ represents the expected state at time $t+1$, and hence, $\mH\mF\vmu_t$ corresponds to the expected observation. Therefore, the term $\vy_{t+1} - \mH\mF\vmu_t$ measures the error in the predicted observation and the Kalman gain $\mK_{t+1}$ appears as a measure of relevance of the new observation compared to the prediction.

\begin{ex}{Bayesian linear regression as a Kalman filter}{blr_as_kf}
  Even though they arise from a rather different setting, it turns out that Kalman filters are a generalization of Bayesian linear regression!
  To see this, recall the online Bayesian linear regression algorithm from \cref{sec:blr:online}.
  Observe that by keeping attempting to estimate the (hidden)  weights $\opt{\vw}$ from sequential noisy observations $y_t$, this algorithm performs Bayesian filtering!
  Moreover, we have used a Gaussian prior and likelihood.
  This is precisely the setting of a Kalman filter!

  Concretely, we are estimating the constant (i.e., $\mF = \mI$, $\vvarepsilon = \vzero$) hidden state $\vx_t = \vw^{(t)}$ with prior $\vw^{(0)} \sim \N{\vzero}{\sigmap^2 \mI}$.

  Our sensor model is time-dependent, since in each iteration we observe a different input $\vx_t$.
  Furthermore, we only observe a scalar-valued label $y_t$.\safefootnote{That is, $m = 1$ in our general Kalman filter formulation from above.}
  Formally, our sensor model is characterized by $\vh_t = \transpose{\vx_t}$ and noise $\eta_t = \varepsilon_t$ with $\varepsilon_t \sim \N{0}{\sigman^2}$.

  You will show in \exerciserefmark{blr_as_kf} that the Kalman update~\eqref{eq:kalman_update} is the online equivalent to computing the posterior of the weights in Bayesian linear regression.
\end{ex}

\subsection{Predicting}

Using now that the marginal posterior of $\rX_t$ is a Gaussian due to the closedness properties of Gaussians, we have \begin{align}
  \rX_{t+1} \mid \vy_{1:t} \sim \N{\vmuhat_{t+1}}{\hat{\mSigma}_{t+1}},
\end{align} and it suffices to compute the prediction mean $\vmuhat_{t+1}$ and covariance matrix~$\hat{\mSigma}_{t+1}$.\looseness=-1

For the mean, \begin{align}
  \vmuhat_{t+1} &= \E{\vx_{t+1} \mid \vy_{1:t}} \nonumber\\
  &= \E{\mF\vx_t + \vvarepsilon_t \mid \vy_{1:t}} \margintag{using the motion model \eqref{eq:kf_motion_model}} \nonumber\\
  &= \mF\E{\vx_t \mid \vy_{1:t}} \margintag{using linearity of expectation \eqref{eq:linearity_expectation} and $\E{\vvarepsilon_t} = \vzero$} \nonumber\\[5pt]
  &= \mF\vmu_t. \margintag{using the mean of the Kalman update} \label{eq:kf_pred_mean}
\end{align}

For the covariance matrix, \begin{align}
  \hat{\mSigma}_{t+1} &= \E{(\vx_{t+1} - \vmuhat_{t+1})\transpose{(\vx_{t+1} - \vmuhat_{t+1})}}[\vy_{1:t}] \margintag{using the definition of the covariance matrix \eqref{eq:covariance_matrix}} \nonumber\\[3pt]
  &= \mF \E{(\vx_t - \vmu_t)\transpose{(\vx_t - \vmu_t)}}[\vy_{1:t}] \transpose{\mF} + \E{\vvarepsilon_t\transpose{\vvarepsilon_t}} \margintag{using \eqref{eq:kf_pred_mean}, the motion model \eqref{eq:kf_motion_model} and that $\vvarepsilon_t$ is independent of the observations} \nonumber\\
  &= \mF \mSigma_t \transpose{\mF} + \mSigma_x. \label{eq:kf_prediction_cov_mat}
\end{align}

\begin{oreadings}
  Kalman filters and related models are often called \midx{temporal models}.
  For a broader look at such models, read chapter 15 of \icite{aimodernapproach}.
\end{oreadings}

\section*{Discussion}

In this chapter, we have introduced Kalman filters as a special case of probabilistic filtering where probabilistic inference can be performed in closed form.
Similarly to Bayesian linear regression, probabilistic inference is tractable due to assuming Gaussian priors and likelihoods.
Indeed, learning linear models and Kalman filters are very closely related as seen in \cref{ex:blr_as_kf}, and we will further explore this relationship in \cref{exercise:kf_as_gp}.
We will refer back to filtering in the second part of this manuscript when we discuss sequential decision-making with partial observability of the state space.
Next, we return to the storyline on ``learning'' using exact probabilistic inference.

\excheading

\begin{nexercise}{Kalman update}{kf_predictive_distr}
  Derive the predictive distribution $X_{t+1} \mid y_{1:t+1}$ \eqref{eq:kalman_posterior} of the Kalman filter described in the above example using your knowledge about multivariate Gaussians from \cref{sec:fundamentals:gaussians}. \\
  \textit{Hint: First compute the predictive distribution $X_{t+1} \mid y_{1:t}$.}
\end{nexercise}

\begin{nexercise}{Bayesian linear regression as a Kalman filter}{blr_as_kf}
  Recall the specific Kalman filter from \cref{ex:blr_as_kf}.
  With this model the Kalman update~\eqref{eq:kalman_update} simplifies to \begin{subequations}\begin{align}
    \vk_t &= \frac{\mSigma_{t-1} \vx_t}{\transpose{\vx_t} \mSigma_{t-1} \vx_t + \sigman^2}, \\
    \vmu_t &= \vmu_{t-1} + \vk_t (y_t - \transpose{\vx_t}\vmu_{t-1}), \\
    \mSigma_t &= \mSigma_{t-1} - \vk_t \transpose{\vx_t}\mSigma_{t-1},
  \end{align}\end{subequations} with $\vmu_0 = \vzero$ and $\mSigma_0 = \sigmap^2 \mI$.
  Note that the Kalman gain~$\vk_t$ is a vector in~$\R^{d}$.
  We assume $\sigman^2 = \sigmap^2 = 1$ for simplicity.\par
  Prove by induction that the $(\vmu_t, \mSigma_t)$ produced by the Kalman update are equivalent to $(\vmu, \mSigma)$ from the posterior of Bayesian linear regression~\eqref{eq:blr_posterior} given $\vx_{1:t}, y_{1:t}$.
  You may use that $\inv{\mSigma_t} \vk_t = \vx_t$.\par
  \textit{Hint: In the inductive step, first prove the equivalence of $\mSigma_t$ and then expand $\inv{\mSigma_t} \vmu_t$ to prove the equivalence of $\vmu_t$.}
\end{nexercise}

\begin{nexercise}{Parameter estimation using Kalman filters}{parameter_estimation_with_kf}
  Suppose that we want to estimate the value of an unknown constant~$\pi$ using uncorrelated measurements \begin{align*}
    y_t = \pi + \eta_t, \quad \eta_t \sim \N{0}{\sigma_y^2}.
  \end{align*}

  \begin{enumerate}
    \item How can this problem be formulated as a Kalman filter?
    Compute closed form expressions for the Kalman gain and the variance of the estimation error $\sigma_t^2$ in terms of $t$, $\sigma_y^2$, and $\sigma_0^2$.
    \item What is the Kalman filter when $t \to \infty$?
    \item Suppose that one has no prior assumptions on $\pi$, meaning that $\mu_0 = 0$ and $\sigma_0^2 \to \infty$.
    Which well-known estimator does the Kalman filter reduce to in this case?
  \end{enumerate}
\end{nexercise}

  \chapter{Gaussian Processes}\label{sec:gp}

Let us remember our first attempt from~\cref{sec:blr} at scaling up Bayesian linear regression to nonlinear functions.
We saw that we can model nonlinear functions by transforming the input space to a suitable higher-dimensional feature space, but found that this approach scales poorly if we require a large number of features.
We then found something remarkable: by simply changing our perspective from a weight-space view to a function-space view, we could implement Bayesian linear regression without ever needing to compute the features explicitly.
Under the function-space view, the key object describing the class of functions we can model is not the features~$\vphi(\vx)$, but instead the kernel function which only implicitly defines a feature space.
Our key observation in this chapter is that we can therefore stop reasoning about feature spaces, and instead directly work with kernel functions that describe ``reasonable'' classes of functions.

We are still concerned with the problem of estimating the value of a function $f : \spX \to \R$ at arbitrary points $\vxs \in \spX$ given training data $\{\vx_i,y_i\}_{i=1}^n$, where the labels are assumed to be corrupted by homoscedastic Gaussian noise with variance $\sigman^2$, \begin{align*}
  y_i = f(\vx_i) + \varepsilon_i,\quad \varepsilon_i \sim \N{0}{\sigman^2}.
\end{align*}
As in \cref{sec:blr} on Bayesian linear regression, we denote by $\mX$ the design matrix (collection of training inputs) and by $\vy$ the vector of training labels.
We will represent the unknown function value at a point $\vx \in \spX$ by the random variable $f_{\vx} \defeq f(\vx)$.
The collection of these random variables is then called a Gaussian process if any finite subset of them is jointly Gaussian:

\begin{defn}[Gaussian process, GP]\pidx{Gaussian process}
  A \emph{Gaussian process} is an infinite set of random variables such that any finite number of them are jointly Gaussian and such that they are consistent under marginalization.\footnote{That is, if you take a joint distribution for $n$ variables and marginalize out one of them, you should recover the joint distribution for the remaining $n-1$ variables.}
\end{defn}

The fact that with a Gaussian process, any finite subset of the random variables is jointly Gaussian is the key property allowing us to perform exact probabilistic inference.
Intuitively, a Gaussian process can be interpreted as a normal distribution over functions --- and is therefore often called an ``infinite-dimensional Gaussian''.

\begin{marginfigure}
  \begin{center}
    \import{./plots/output/}{gp.pgf}
  \end{center}

  \caption{A Gaussian process can be interpreted as an infinite-dimensional Gaussian over functions.
  At any location $x$ in the domain, this yields a distribution over values $f(x)$ shown in red.
  The blue line corresponds to the MAP estimate (i.e., mean function of the Gaussian process), the dark gray region corresponds to the epistemic uncertainty and the light gray region denotes the additional aleatoric uncertainty.}
\end{marginfigure}

A Gaussian process is characterized by a \midx{mean function} $\mu : \spX \to \R$ and a \midx{covariance function} (or \midx{kernel function}) $k : \spX \times \spX \to \R$ such that for any set of points $\sA \defeq \{\vx_1, \dots, \vx_m\} \subseteq \spX$, we have \begin{align}
  \vf_\sA \defeq \transpose{[f_{\vx_1} \; \cdots \; f_{\vx_m}]} \sim \N{\vmu_\sA}{\mK_{\sA\sA}} \label{eq:gp_joint_distr}
\end{align} where \begin{align}
  \vmu_\sA \defeq \begin{bmatrix}
    \mu(\vx_1) \\
    \vdots \\
    \mu(\vx_m) \\
  \end{bmatrix}, \quad \mK_{\sA\sA} \defeq \begin{bmatrix}
    k(\vx_1, \vx_1) & \cdots & k(\vx_1, \vx_m) \\
    \vdots & \ddots & \vdots \\
    k(\vx_m, \vx_1) & \cdots & k(\vx_m, \vx_m) \\
  \end{bmatrix}.
\end{align}
We write $f \sim \GP{\mu}{k}$.
In particular, given a mean function, covariance function, and using the homoscedastic noise assumption, \begin{align}
  \ys \mid \vxs \sim \N{\mu(\vxs)}{k(\vxs, \vxs) + \sigman^2}. \label{eq:gp_predictive_distr}
\end{align}
Commonly, for notational simplicity, the mean function is taken to be zero.
Note that for a fixed mean this is not a restriction, as we can simply apply the zero-mean Gaussian process to the difference between the mean and the observations.\footnote{For alternative ways of representing a mean function, refer to section 2.7 of \icite{gpml}.}

\section{Learning and Inference}\label{sec:gp:learning_and_inference}

First, let us look at learning and inference in the context of Gaussian processes.
With slight abuse of our previous notation, let us denote the set of observed points by $\sA \defeq \{\vx_1, \dots, \vx_n\}$.
Given a prior $f \sim \GP{\mu}{k}$ and the noisy observations ${y_i = f(\vx_i) + \varepsilon_i}$ with ${\varepsilon_i \sim \N{0}{\sigman^2}}$, we can then write the joint distribution of the observations $y_{1:n}$ and the noise-free prediction $\fs$ at a test point $\vxs$ as \begin{align}
  &\begin{bmatrix}
    \vy \\
    \fs \\
  \end{bmatrix} \mid \vxs, \mX \sim \N{\Tilde{\vmu}}{\Tilde{\mK}}, \quad\text{where} \\
  &\Tilde{\vmu} \defeq \begin{bmatrix}
    \vmu_\sA \\
    \mu(\vxs) \\
  \end{bmatrix}, \quad \Tilde{\mK} \defeq \begin{bmatrix}
    \mK_{\sA\sA} + \sigman^2 \mI & \vk_{\vxs,\sA} \\
    \transpose{\vk_{\vxs,\sA}} & k(\vxs, \vxs) \\
  \end{bmatrix}, \quad \vk_{\vx,\sA} \defeq \begin{bmatrix}
    k(\vx,\vx_1) \\ \vdots \\ k(\vx, \vx_n)
  \end{bmatrix}.
\end{align}
Deriving the conditional distribution using \eqref{eq:cond_gaussian}, we obtain that the Gaussian process posterior is given by \begin{align}
  f \mid \vx_{1:n}, y_{1:n} &\sim \GP{\mu'}{k'}, \quad\text{where} \label{eq:gp_posterior} \\
  \mu'(\vx) &\defeq \mu(\vx) + \transpose{\vk_{\vx,\sA}} \inv{(\mK_{\sA\sA} + \sigman^2 \mI)} (\vy_\sA - \vmu_\sA), \\
  k'(\vx, \vxp) &\defeq k(\vx, \vxp) - \transpose{\vk_{\vx,\sA}} \inv{(\mK_{\sA\sA} + \sigman^2 \mI)} \vk_{\vxp,\sA}.
\end{align}
Observe that analogously to Bayesian linear regression, the posterior covariance can only decrease when conditioning on additional data, and is independent of the observations $y_i$.

We already studied inference in the function-space view of Bayesian linear regression, but did not make the predictive posterior explicit.
Using \cref{eq:gp_posterior}, the predictive posterior at~$\vxs$ is simply \begin{align}
  \fs \mid \vxs, \vx_{1:n}, y_{1:n} &\sim \N{\mu'(\vxs)}{k'(\vxs,\vxs)}.
\end{align}

\section{Sampling}\label{sec:gp:sampling}

Often, we are not interested in the full predictive posterior distribution, but merely want to obtain samples of our Gaussian process model.
We will briefly examine two approaches.

\begin{enumerate}
  \item For the first approach, consider a discretized subset of points \begin{align*}
    \vf \defeq [f_1, \dots, f_n]
  \end{align*} that we want to sample.\footnote{For example, if we want to render the function, the length of this vector could be guided by the screen resolution.}
  Note that $\vf \sim \N{\vmu}{\mK}$.
  We have already seen in \cref{eq:gaussian_affine_transformation} that \begin{align}
    \vf = \msqrt{\mK} \vvarepsilon + \vmu
  \end{align} where $\msqrt{\mK}$ is the square root of $\mK$ and $\vvarepsilon \sim \SN$ is standard Gaussian noise.\footnote{We discuss square roots of matrices in \cref{sec:fundamentals:qf}.}
  However, computing the square root of $\mK$ takes $\BigO{n^3}$ time.

  \item For the second approach, recall the product rule \eqref{eq:product_rule}, \begin{align*}
    p(f_1, \dots, f_n) = \prod_{i=1}^n p(f_i \mid f_{1:i-1}).
  \end{align*}
  That is the joint distribution factorizes neatly into a product where each factor only depends on the ``outcomes'' of preceding factors.
  We can therefore obtain samples one-by-one, each time conditioning on one more observation: \begin{align}\begin{split}
    f_1 &\sim p(f_1) \\
    f_2 &\sim p(f_2 \mid f_1) \\
    f_3 &\sim p(f_3 \mid f_1, f_2) \\
    &\;\:\,\vdots
  \end{split}\end{align}
  This general approach is known as \midx{forward sampling}.
  Due to the matrix inverse in the formula of the GP posterior \eqref{eq:gp_posterior}, this approach also takes $\BigO{n^3}$ time.
\end{enumerate}

We will discuss more efficient approximate sampling methods in \cref{sec:gp:approximations}.

\section{Kernel Functions}\label{sec:gp:kernel_functions}

We have seen that kernel functions are the key object describing the class of functions a Gaussian process can model.
Depending on the kernel function, the ``shape'' of functions that are realized from a Gaussian process varies greatly.
Let us recap briefly from \cref{sec:blr:function_space_view} what a kernel function is:

\begin{defn}[Kernel function]\pidx{kernel function}[idxpagebf]
  A \emph{kernel function} $k : \spX \times \spX \to \R$ satisfies \begin{itemize}
    \item $k(\vx, \vxp) = k(\vxp, \vx)$ for any $\vx, \vxp \in \spX$ (symmetry), and
    \item $\mK_{\sA\sA}$ is positive semi-definite for any $\sA \subseteq \spX$.
  \end{itemize}

  The two defining conditions ensure that for any $\sA \subseteq \spX$, $\mK_{\sA\sA}$ is a valid covariance matrix.
  We say that a kernel function is \midx<positive definite>{positive definite kernel} if $\mK_{\sA\sA}$ is positive definite for any $\sA \subseteq \spX$.

\end{defn}

Intuitively, the kernel function evaluated at locations $\vx$ and $\vxp$ describes how $f(\vx)$ and $f(\vxp)$ are related, which we can express formally as \begin{align}
  k(\vx, \vxp) = \Cov{f(\vx), f(\vxp)}.
\end{align}
If $\vx$ and $\vxp$ are ``close'', then $f(\vx)$ and $f(\vxp)$ are usually taken to be positively correlated, encoding a ``smooth'' function.

In the following, we will discuss some of the most common kernel functions, how they can be combined to create ``new'' kernels, and how we can characterize the class of functions they can model.

\subsection{Common Kernels}
\begin{marginfigure}
  \begin{center}
    \import{./plots/output/}{kernel_linear.pgf}
  \end{center}

  \caption{Functions sampled according to a Gaussian process with a linear kernel and $\phi = \id$.}
\end{marginfigure}
\begin{marginfigure}
  \begin{center}
    \import{./plots/output/}{kernel_linear_transformed.pgf}
  \end{center}

  \caption{Functions sampled according to a Gaussian process with a linear kernel and $\vphi(x) = [1, x, x^2]$ (left) and $\phi(x) = \sin(x)$ (right).}
\end{marginfigure}

First, we look into some of the most commonly used kernels. Often an additional factor $\sigma^2$ (\midx{output scale}) is added, which we assume here to be $1$ for simplicity.

\begin{enumerate}
  \item The \midx{linear kernel} is defined as \begin{align}
    k(\vx, \vxp; \vphi) \defeq \transpose{\vphi(\vx)} \vphi(\vxp) \label{eq:linear_kernel}
  \end{align} where $\vphi$ is a nonlinear transformation as introduced in \cref{sec:blr:non_linear} or the identity.

  \begin{rmk}{GPs with linear kernel and BLR}{}
    A Gaussian process with a linear kernel is equivalent to Bayesian linear regression.
    This follows directly from the function-space view of Bayesian linear regression (see \cref{sec:blr:function_space_view}) and comparing the derived kernel function \eqref{eq:blr_kernel} with the definition of the linear kernel \eqref{eq:linear_kernel}.
  \end{rmk}

  \item The \midx{Gaussian kernel} (also known as \midx{squared exponential kernel} or \midx<radial basis function (RBF) kernel>{radial basis function kernel}) is defined as \begin{align}
    k(\vx, \vxp; h) \defeq \exp\parentheses*{-\frac{\norm{\vx-\vxp}_2^2}{2 h^2}} \label{eq:gaussian_kernel}
  \end{align} where $h$ is its \midx{length scale}. The larger the length scale $h$, the smoother the resulting functions.\footnote[][-1\baselineskip]{As the length scale is increased, the exponent of the exponential increases, resulting in a higher dependency between locations.}
  Furthermore, it turns out that the feature space (think back to \cref{sec:blr:function_space_view}!) corresponding to the Gaussian kernel is ``infinitely dimensional'', as you will show in \exerciserefmark{gaussian_kernel_feature_space}.
  So the Gaussian kernel already encodes a function class that we were not able to model under the weight-space view of Bayesian linear regression.

  \begin{figure}
  \begin{center}
    \import{./plots/output/}{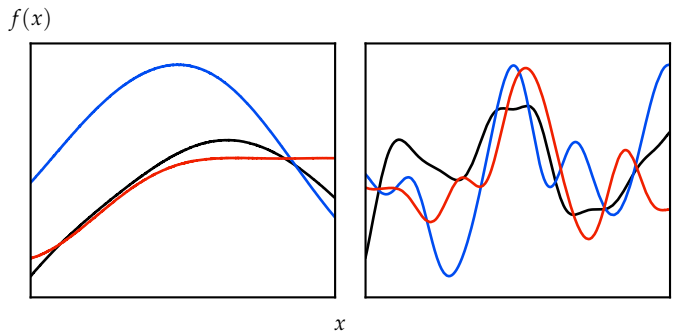}
  \end{center}

    \caption{Functions sampled according to a Gaussian process with a Gaussian kernel and length scales $h = 5$ (left) and $h = 1$ (right).}
  \end{figure}

  \begin{marginfigure}
  \begin{center}
    \import{./plots/output/}{kernel_f_gaussian.pgf}
  \end{center}

    \caption{Gaussian kernel with length scales $\b{h=1}$, $h=0.5$, and $\r{h=0.2}$.}
  \end{marginfigure}

  \item The \midx{Laplace kernel} (also known as \midx{exponential kernel}) is defined as \begin{align}
    k(\vx, \vxp; h) \defeq \exp\parentheses*{-\frac{\norm{\vx-\vxp}_2}{h}}. \label{eq:laplace_kernel}
  \end{align}
  As can be seen in \cref{fig:gp_samples_laplace}, samples from a GP with Laplace kernel are non-smooth as opposed to the samples from a GP with Gaussian kernel.

  \begin{marginfigure}[10\baselineskip]
  \begin{center}
    \import{./plots/output/}{kernel_f_laplace.pgf}
  \end{center}

    \caption{Laplace kernel with length scales $\b{h=1}$, $h=0.5$, and $\r{h=0.2}$.}
  \end{marginfigure}

  \begin{figure}
  \begin{center}
    \import{./plots/output/}{kernel_laplace.pgf}
  \end{center}

    \caption{Functions sampled according to a Gaussian process with a Laplace kernel and length scales $h = 10\,000$ (left) and $h = 10$ (right).}\label{fig:gp_samples_laplace}
  \end{figure}

  \item The \midx{Matérn kernel} trades the smoothness of the Gaussian and the Laplace kernels.
  As such, it is frequently used in practice to model ``real world'' functions that are relatively smooth.
  It is defined as \begin{align}
    k(\vx, \vxp; \nu, h) \defeq \frac{2^{1-\nu}}{\Gamma(\nu)} \parentheses*{\frac{\sqrt{2 \nu} \norm{\vx-\vxp}_2}{h}}^\nu K_\nu \parentheses*{\frac{\sqrt{2 \nu} \norm{\vx-\vxp}_2}{h}}
  \end{align} where $\Gamma$ is the Gamma function, $K_\nu$ the modified Bessel function of the second kind, and $h$ a length scale parameter.
  For ${\nu = \nicefrac{1}{2}}$, the Matérn kernel is equivalent to the Laplace kernel.
  For ${\nu \to \infty}$, the Matérn kernel is equivalent to the Gaussian kernel.
  The resulting functions are $\ceil{\nu} - 1$ times mean square differentiable.\footnote{Refer to \cref{rmk:mean_square_convergence} for the definitions of mean square continuity and differentiability.}
  In particular, GPs with a Gaussian kernel are infinitely many times mean square differentiable whereas GPs with a Laplace kernel are mean square continuous but not mean square differentiable.
\end{enumerate}

\subsection{Composing Kernels}

Given two kernels $k_1 : \spX \times \spX \to \R$ and $k_2 : \spX \times \spX \to \R$, they can be composed to obtain a new kernel $k : \spX \times \spX \to \R$ in the following ways: \begin{itemize}
  \item $k(\vx, \vxp) \defeq k_1(\vx, \vxp) + k_2(\vx, \vxp)$,
  \item $k(\vx, \vxp) \defeq k_1(\vx, \vxp) \cdot k_2(\vx, \vxp)$,
  \item $k(\vx, \vxp) \defeq c \cdot k_1(\vx, \vxp)$ for any $c > 0$,
  \item $k(\vx, \vxp) \defeq f(k_1(\vx, \vxp))$ for any polynomial $f$ with positive coefficients or $f = \exp$.
\end{itemize}

For example, the additive structure of a function $f(\vx) \defeq f_1(\vx) + f_2(\vx)$ can be easily encoded in GP models. Suppose that $f_1 \sim \GP{\mu_1}{k_1}$ and $f_2 \sim \GP{\mu_2}{k_2}$, then the distribution of the sum of those two functions $f = f_1 + f_2 \sim \GP{\mu_1 + \mu_2}{k_1 + k_2}$ is another GP.\footnote{We use $f \defeq f_1 + f_2$ to denote the function $f(\cdot) = f_1(\cdot) + f_2(\cdot)$.}

Whereas the addition of two kernels $k_1$ and $k_2$ can be thought of as an \emph{OR} operation (i.e., the kernel has high value if either $k_1$ or $k_2$ have high value), the multiplication of $k_1$ and $k_2$ can be thought of as an \emph{AND} operation (i.e., the kernel has high value if both $k_1$ and $k_2$ have high value). For example, the product of two linear kernels results in functions which are quadratic.

As mentioned previously, the constant $c$ of a scaled kernel function $k'(\vx, \vxp) \defeq c \cdot k(\vx, \vxp)$ is generally called the \midx{output scale}[idxpagebf] of a kernel, and it scales the variance $\Var{f(\vx)} = c \cdot k(\vx, \vx)$ of the predictions $f(\vx)$ from $\GP{\mu}{k'}$.

\begin{oreadings}
  For a broader introduction to how kernels can be used and combined to model certain classes of functions, read \begin{itemize}
    \item chapter 2 of \icite{duvenaud2014automatic} also known as the ``kernel cookbook'',
    \item chapter 4 of \icite{gpml}.
  \end{itemize}
\end{oreadings}

\subsection{Stationarity and Isotropy}

Kernel functions are commonly classified according to two properties:

\begin{defn}[Stationarity and isotropy]
A kernel $k : \R^d \times \R^d \to \R$ is called \begin{itemize}
  \item \midx<stationary>{stationary kernel} (or \midx<shift-invariant>{shift-invariant kernel}) if there exists a function $\tilde{k}$ such that $\tilde{k}(\vx - \vxp) = k(\vx, \vxp)$, and
  \item \midx<isotropic>{isotropic kernel} if there exists a function $\tilde{k}$ such that $\tilde{k}(\norm{\vx-\vxp}) = k(\vx, \vxp)$ with $\norm{\cdot}$ any norm.
\end{itemize}
\end{defn}

Note that stationarity is a necessary condition for isotropy. In other words, isotropy implies stationarity.

\begin{ex}{Stationarity and isotropy of kernels}{}
  \begin{center}
    \begin{tabular}{lrr}
      \toprule
      & stationary & isotropic \\
      \midrule
      linear kernel & no & no \\
      \addlinespace
      Gaussian kernel & yes & yes \\
      \addlinespace
      \makecell[l]{$k(\vx, \vxp) \defeq \exp(-\norm{\vx-\vxp}_\mM^2)$ \\ where $\mM$ is positive semi-definite} & yes & no \margintag{$\norm{\cdot}_\mM$ denotes the Mahalanobis norm induced by matrix $\mM$} \\
      \bottomrule
    \end{tabular}
  \end{center}

  For ${\vxp = \vx}$, stationarity implies that the kernel must only depend on $\vzero$.
  In other words, a stationary kernel must depend on relative locations only.
  This is clearly not the case for the linear kernel, which depends on the absolute locations of $\vx$ and $\vxp$.
  Therefore, the linear kernel cannot be isotropic either.

  For the Gaussian kernel, isotropy follows immediately from its definition.

  The last kernel is clearly stationary by definition, but not isotropic for general matrices $\mM$.
  Note that for ${\mM = \mI}$ it is indeed isotropic.
\end{ex}

Stationarity encodes the idea that relative location matters more than absolute location: the process ``looks the same'' no matter where we shift it in the input space.
This is often appropriate when we believe the same statistical behavior holds across the entire domain (e.g., no region is special).
Isotropy goes one step further by requiring that the kernel depends only on the distance between points, so that all directions in the space are treated equally.
In other words, there is no preferred orientation or axis.
This is especially useful in settings where we expect uniform behavior in every direction (as with the Gaussian kernel).
Such kernels are simpler to specify and interpret since we only need a single ``scale'' (like a length scale) rather than multiple parameters or directions.

\subsection{Reproducing Kernel Hilbert Spaces}\label{sec:rkhs}

We can characterize the precise class of functions that can be modeled by a Gaussian process with a given kernel function.
This corresponding function space is called a \midx{reproducing kernel Hilbert space} (RKHS), and we will discuss it briefly in this section.

Recall that Gaussian processes keep track of a posterior distribution $f \mid \vx_{1:n}, y_{1:n}$ over functions.
We will in fact show later that the corresponding MAP estimate $\hat{f}$ corresponds to the solution to a regularized optimization problem in the RKHS space of functions.
This duality is similar to the duality between the MAP estimate of Bayesian linear regression and ridge regression we observed in \cref{sec:blr}.
So what is the reproducing kernel Hilbert space of a kernel function $k$?

\begin{defn}[Reproducing kernel Hilbert space, RKHS]
  Given a kernel $k : \spX \times \spX \to \R$, its corresponding \emph{reproducing kernel Hilbert space} is the space of functions $f$ defined as \begin{align}
    \spH_k(\spX) \defeq \braces*{f(\cdot) = \sum_{i=1}^n \alpha_i k(\vx_i, \cdot) : n \in \Nat, \vx_i \in \spX, \alpha_i \in \R}.
  \end{align}
  The inner product of the RKHS is defined as \begin{align}
    \ip{f,g}_k \defeq \sum_{i=1}^n \sum_{j=1}^{n'} \alpha_i \alpha_j' k(\vx_i,\vx_j'),
  \end{align} where $g(\cdot) = \sum_{j=1}^{n'} \alpha_j' k(\vx_j',\cdot)$, and induces the norm ${\norm{f}_k = \sqrt{\ip{f,f}_k}}$.
  You can think of the norm as measuring the ``smoothness'' or ``complexity'' of $f$. \exerciserefmark{reproducing_kernel_hilbert_space_properties}[2]
\end{defn}
It is straightforward to check that for all $\vx \in \spX$, $k(\vx, \cdot) \in \spH_k(\spX)$.
Moreover, the RKHS inner product $\ip{\cdot,\cdot}_k$ satisfies for all $\vx \in \spX$ and $f \in \spH_k(\spX)$ that $f(\vx) = \ip{f(\cdot), k(\vx, \cdot)}_k$ which is also known as the \midx{reproducing property} \exerciserefmark{reproducing_kernel_hilbert_space_properties}[1].
That is, evaluations of RKHS functions $f$ are inner products in $\spH_k(\spX)$ parameterized by the ``feature map'' $k(\vx,\cdot)$.

The \midx{representer theorem} \citep{scholkopf2001generalized} characterizes the solution to regularized optimization problems in RKHSs:\looseness=-1

\begin{thm}[Representer theorem]
  \exerciserefmark{representer_theorem}
  Let $k$ be a kernel and let $\lambda > 0$.
  For $f \in \spH_k(\spX)$ and training data $\{(\vx_i, f(\vx_i))\}_{i=1}^n$, let ${\spL(f(\vx_1), \dots, f(\vx_n))} \in {\R \cup \{\infty\}}$ denote any loss function which depends on $f$ only through its evaluation at the training points.
  Then, any minimizer \begin{align}
    \hat{f} \in \argmin_{f \in \spH_k(\spX)} \spL(f(\vx_1), \dots, f(\vx_n)) + \lambda \norm{f}_k^2 \label{eq:representer_theorem_obj}
  \end{align} admits a representation of the form \begin{align}
    \hat{f}(\vx) = \transpose{\valphahat} \vk_{\vx,\{\vx_i\}_{i=1}^n} = \sum_{i=1}^n \hat{\alpha}_i k(\vx, \vx_i) \quad\text{for some $\valphahat \in \R^n$}. \label{eq:representer_theorem}
  \end{align}
\end{thm}

This statement is remarkable: the solutions to general regularized optimization problems over the generally infinite-dimensional space of functions~$\spH_k(\spX)$ can be represented as a linear combination of the kernel functions evaluated at the training points.
The representer theorem can be used to show that the MAP estimate of a Gaussian process corresponds to the solution of a regularized linear regression problem in the RKHS of the kernel function, namely, \exerciserefmark{mle_and_map_of_gps} \begin{align}
  \hat{f} \defeq \argmin_{f \in \spH_k(\spX)} - \log p(y_{1:n} \mid \vx_{1:n}, f) + \frac{1}{2} \norm{f}_k^2. \label{eq:map_of_gp}
\end{align}
Here, the first term corresponds to the likelihood, measuring the ``quality of fit''.
The regularization term limits the ``complexity'' of $\hat{f}$.
Regularization is necessary to prevent overfitting since in an expressive RKHSs, there may be many functions that interpolate the training data perfectly.
This shows the close link between Gaussian process regression and Bayesian linear regression, with the kernel function $k$ generalizing the inner product of feature maps to feature spaces of possibly ``infinite dimensionality''.
Because solutions can be represented as linear combinations of kernel evaluations at the training points, Gaussian processes remain computationally tractable even though they can model functions over ``infinite-dimensional'' feature spaces.

\section{Model Selection}\label{sec:gp:model_selection}

We have not yet discussed how to pick the hyperparameters $\vtheta$ (e.g., parameters of kernels).
A common technique in supervised learning is to select hyperparameters $\vtheta$, such that the resulting function estimate $\hat{f}_\vtheta$ leads to the most accurate predictions on hold-out validation data.
After reviewing this approach, we contrast it with a probabilistic approach to model selection, which avoids using point estimates of $\hat{f}_\vtheta$ and rather utilizes the full posterior.

\subsection{Optimizing Validation Set Performance}

A common approach to model selection is to split our data $\spD$ into separate training set $\spD^\train \defeq \{(\vx_i^\train, y_i^\train)\}_{i=1}^n$ and validation sets $\spD^\val \defeq \{(\vx_i^\val, y_i^\val)\}_{i=1}^m$.
We then optimize the model for a parameter candidate $\vtheta_j$ using the training set.
This is usually done by picking a point estimate (like the MAP estimate), \begin{align}
  \hat{f}_j \defeq \argmax_{f } p(f \mid \vx_{1:n}^\train, y_{1:n}^\train).
\end{align}
Then, we score $\vtheta_j$ according to the performance of $\hat{f}_j$ on the validation set,\looseness=-1 \begin{align}
  \vthetahat \defeq \argmax_{\vtheta_j} p(y_{1:m}^\val \mid \vx_{1:m}^\val, \hat{f_j}). \label{eq:cross_validation}
\end{align}
This ensures that $\hat{f}_j$ does not depend on $\spD^\val$.

\begin{rmk}{Approximating population risk}{}
  Why is it useful to separate the data into a training and a validation set?
  Recall from \cref{sec:fundamentals:supervised_learning:risk} that minimizing the empirical risk without separating training and validation data may lead to overfitting as both the loss and $\hat{f}_j$ depend on the same data~$\spD$.
  In contrast, using independent training and validation sets, $\hat{f}_j$ does not depend on $\spD^\val$, and we have that \begin{align}
    \frac{1}{m} \sum_{i=1}^m \ell(y_i^\val \mid \vx_i^\val, \hat{f}_j) \approx \E[(\vx,y)\sim\spP]{\ell(y \mid \vx, \hat{f}_j)},
  \end{align} using Monte Carlo sampling.\safefootnote{We generally assume $\spD \iid \spP$, in particular, we assume that the individual samples of the data are i.i.d.. Recall that in this setting, Hoeffding's inequality \eqref{eq:hoeffdings_inequality} can be used to gauge how large $m$ should be.}
  In words, for reasonably large $m$, minimizing the empirical risk as we do in \cref{eq:cross_validation} approximates minimizing the population risk.
\end{rmk}

While this approach often is quite effective at preventing overfitting as compared to using the same data for training and picking $\vthetahat$, it still collapses the uncertainty in $f$ into a point estimate.
Can we do better?

\subsection{Maximizing the Marginal Likelihood}\label{sec:gp:model_selection:marginal_likelihood}

We have already seen for Bayesian linear regression, that picking a point estimate loses a lot of information.
Instead of optimizing the effects of $\vtheta$ for a specific point estimate $\hat{f}$ of the model $f$, \midx{maximizing the marginal likelihood} optimizes the effects of $\vtheta$ across all realizations of $f$.
In this approach, we obtain our hyperparameter estimate via \begin{align}
  \vthetahat_\MLE &\defeq \argmax_{\vtheta} p(y_{1:n} \mid \vx_{1:n}, \vtheta) \margintag{using the definition of marginal likelihood in Bayes' rule \eqref{eq:bayes_rule}} \label{eq:maximizing_marginal_likelihood} \\
  &= \argmax_{\vtheta} \int p(y_{1:n}, f \mid \vx_{1:n}, \vtheta) \,d f \margintag{by conditioning on $f$ using the sum rule \eqref{eq:sum_rule}} \nonumber \\
  &= \argmax_{\vtheta} \int p(y_{1:n} \mid \vx_{1:n}, f, \vtheta) p(f \mid \vtheta) \,d f. \margintag{using the product rule \eqref{eq:product_rule}}
\end{align}
Remarkably, this approach typically avoids overfitting even though we do not use a separate training and validation set.
The following table provides an intuitive argument for why maximizing the marginal likelihood is a good strategy.

\begin{table}[h!]
  \centering
  \begin{tabular}{lll}
  \toprule
    & likelihood & prior \\
    \midrule
    \makecell[l]{``underfit'' model \\ (too simple $\vtheta$)} & small for ``almost all'' $f$ & large \\
    \addlinespace
    \makecell[l]{``overfit'' model \\ (too complex $\vtheta$)} & \makecell[l]{large for ``few'' $f$ \\ small for ``most'' $f$} & small \\
    \addlinespace
    ``just right'' & moderate for ``many'' $f$ & moderate \\
    \bottomrule
  \end{tabular}\\[11pt]
  \caption{The table gives an intuitive explanation of effects of parameter choices $\vtheta$ on the marginal likelihood.
  Note that words in quotation marks refer to intuitive quantities, as we have infinitely many realizations of $f$.}
\end{table}

\begin{marginfigure}[7\baselineskip]
  \begin{center}
    \import{./plots/output/}{maximizing_marginal_likelihood.pgf}
  \end{center}

  \caption{A schematic illustration of the marginal likelihood of a simple, intermediate, and complex model across all possible data sets.}
\end{marginfigure}

For an ``underfit'' model, the likelihood is mostly small as the data cannot be well described, while the prior is large as there are ``fewer'' functions to choose from.
For an ``overfit'' model, the likelihood is large for ``some'' functions (which would be picked if we were only minimizing the training error and not doing cross validation) but small for ``most'' functions.
The prior is small, as the probability mass has to be distributed among ``more'' functions.
Thus, in both cases, one term in the product will be small.
Hence, maximizing the marginal likelihood naturally encourages trading between a large likelihood and a large prior.

In the context of Gaussian process regression, recall from \cref{eq:gp_predictive_distr} that \begin{align}
  y_{1:n} \mid \vx_{1:n}, \vtheta \sim \N{\vzero}{\mK_{f,\vtheta} + \sigman^2 \mI}
\end{align} where $\mK_{f,\vtheta}$ denotes the kernel matrix at the inputs $\vx_{1:n}$ depending on the kernel function parameterized by $\vtheta$.
We write $\mK_{\vy,\vtheta} \defeq \mK_{f,\vtheta} + \sigman^2 \mI$.
Continuing from \cref{eq:maximizing_marginal_likelihood}, we obtain \begin{align}
  \vthetahat_\MLE &= \argmax_{\vtheta} \N[\vy]{\vzero}{\mK_{\vy,\vtheta}} \nonumber \\
  &= \argmin_{\vtheta} \frac{1}{2} \transpose{\vy} \inv{\mK_{\vy,\vtheta}} \vy + \frac{1}{2} \log \det{\mK_{\vy,\vtheta}} + \frac{n}{2} \log 2 \pi \label{eq:gp_mle_1} \margintag{taking the negative logarithm} \\
  &= \argmin_{\vtheta} \frac{1}{2} \transpose{\vy} \inv{\mK_{\vy,\vtheta}} \vy + \frac{1}{2} \log \det{\mK_{\vy,\vtheta}} \label{eq:gp_mle_2} \margintag{the last term is independent of $\vtheta$}
\end{align}
The first term of the optimization objective describes the ``goodness of fit'' (i.e., the ``alignment'' of $\vy$ with $\mK_{\vy,\vtheta}$).
The second term characterizes the ``volume'' of the model class.
Thus, this optimization naturally trades the aforementioned objectives.

\begin{figure}
  \begin{center}
    \import{./plots/output/}{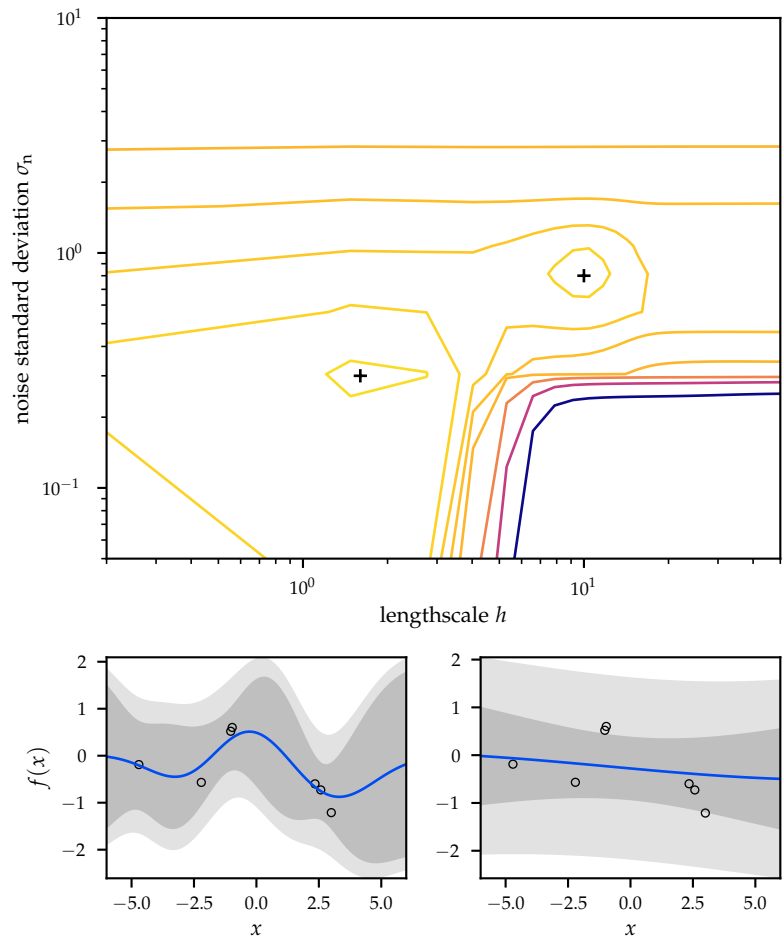}
  \end{center}

  \caption{The top plot shows contour lines of an empirical Bayes with two local optima.
  The bottom two plots show the Gaussian processes corresponding to the two optimal models.
  The left model with smaller lengthscale is chosen within a more flexible class of models, while the right model explains more observations through noise.
  Adapted from figure 5.5 of \icite{gpml}.}\label{fig:empirical_bayes}
\end{figure}

Marginal likelihood maximization is an empirical Bayes method.
Often it is simply referred to as \midx{empirical Bayes}.
It also has the nice property that the gradient of its objective (the MLL loss) can be expressed in closed-from \exerciserefmark{gradient_of_mll}, \begin{align}
  \pdv{}{\theta_j} \log p(y_{1:n} \mid \vx_{1:n}, \vtheta) = \frac{1}{2} \tr{(\valpha \transpose{\valpha} - \inv{\mK_{\vy,\vtheta}}) \pdv{\mK_{\vy,\vtheta}}{\theta_j}} \label{eq:gradient_mll}
\end{align} where $\valpha \defeq \inv{\mK_{\vy,\vtheta}} \vy$ and $\tr{\mM}$ is the trace of a matrix $\mM$.
This optimization problem is, in general, non-convex.
\Cref{fig:empirical_bayes} gives an example of two local optima according to empirical Bayes.

\begin{marginfigure}[4\baselineskip]
  \begin{center}
    \import{./plots/output/}{mll_kernel_progress.pgf}
  \end{center}

  \caption{An example of model selection by maximizing the log likelihood (without hyperpriors) using a \textbf{\b{linear}}, \textbf{\r{quadratic}}, \textbf{\purple{Laplace}}, \textbf{\g{Matérn}} ($\nu=\nicefrac{3}{2}$), and \textbf{Gaussian} kernel, respectively.
  They are used to learn the function \begin{align*}
    x \mapsto \frac{\sin(x)}{x} + \varepsilon,\quad \varepsilon \sim \N{0}{0.01}
  \end{align*} using SGD with learning rate $0.1$.}
\end{marginfigure}

Taking a step back, observe that taking a probabilistic perspective on model selection naturally led us to consider all realizations of our model $f$ instead of using point estimates.
However, we are still using point estimates for our model parameters $\vtheta$.
Continuing on our probabilistic adventure, we could place a prior $p(\vtheta)$ on them too.\footnote{Such a prior is called \idx{hyperprior}[idxpagebf].}
We could use it to obtain the MAP estimate (still a point estimate!) which adds an additional regularization term \begin{align}
  \vthetahat_\MAP &\defeq \argmax_{\vtheta} p(\vtheta \mid \vx_{1:n}, y_{1:n}) \\
  &= \argmin_{\vtheta} - \log p(\vtheta) - \log p(y_{1:n} \mid \vx_{1:n}, \vtheta). \margintag{using Bayes' rule \eqref{eq:bayes_rule} and then taking the negative logarithm}
\end{align}

An alternative approach is to consider the full posterior distribution over parameters $\vtheta$.
The resulting predictive distribution is, however, intractable, \begin{align}
  p(\ys \mid \vxs, \vx_{1:n}, y_{1:n}) = \int\int p(\ys \mid \vxs, f) \cdot p(f \mid \vx_{1:n}, y_{1:n}, \vtheta) \cdot p(\vtheta) \,d f \,d\vtheta.
\end{align}
Recall that as the mode of Gaussians coincides with their mean, the MAP estimate corresponds to the mean of the predictive posterior.

As a final note, observe that in principle, there is nothing stopping us from descending deeper in the probabilistic hierarchy.
The prior on the model parameters $\vtheta$ is likely to have parameters too.
Ultimately, we need to break out of this hierarchy of dependencies and choose a prior.\looseness=-1

\section{Approximations}\label{sec:gp:approximations}

To learn a Gaussian process, we need to invert $n \times n$ matrices, hence the computational cost is $\BigO{n^3}$.
Compare this to Bayesian linear regression which allows us to learn a regression model in $\BigO{n d^2}$ time (even online) where $d$ is the feature dimension.
It is therefore natural to look for ways of approximating a Gaussian process.

\subsection{Local Methods}

Recall that during forward sampling, we had to condition on a larger and larger number of previous samples.
When sampling at a location~$\vx$, a very simple approximation is to only condition on those samples~$\vxp$ that are ``close'' (where $\abs{k(\vx, \vxp)} \geq \tau$ for some $\tau > 0$).
Essentially, this method ``cuts off the tails'' of the kernel function $k$.
However, $\tau$ has to be chosen carefully as if $\tau$ is chosen too large, samples become essentially independent.

This is one example of a \emph{sparse approximation} of a Gaussian process.
We will discuss more advanced sparse approximations known as ``inducing point methods'' in \cref{sec:gp:approx:data_sampling}.

\subsection{Kernel Function Approximation}\label{sec:gp:approximations:kernel}

Another method is to approximate the kernel function directly.
The idea is to construct a ``low-dimensional'' feature map $\vphi : \R^d \to \R^m$ that approximates the kernel, \begin{align}
    k(\vx, \vxp) \approx \transpose{\vphi(\vx)} \vphi(\vxp).
\end{align}
Then, we can apply Bayesian linear regression, resulting in a time complexity of $\BigO{n m^2 + m^3}$.

One example of this approach are \midx{random Fourier features}, which we will discuss in the following.

\begin{marginfigure}[25\baselineskip]
  \begin{center}
    \import{./plots/output/}{eulers_formula.pgf}
  \end{center}

  \caption{Illustration of Euler's formula.
  It can be seen that $e^{i \varphi}$ corresponds to a (counter-clockwise) rotation on the unit circle as $\varphi$ varies from $0$ to $2 \pi$.}\label{fig:eulers_formula}
\end{marginfigure}

\begin{rmk}{Fourier transform}{}
  First, let us remind ourselves of Fourier transformations.
  The Fourier transform is a method of decomposing frequencies into their individual components.

  Recall \midx{Euler's formula} which states that for any $x \in \R$, \begin{align}
    e^{i x} = \cos x + i \sin x \label{eq:eulers_formula}
  \end{align} where $i$ is the imaginary unit of complex numbers.
  The formula is illustrated in \cref{fig:eulers_formula}.
  Note that $e^{- i 2 \pi x}$ corresponds to rotating clockwise around the unit circle in $\R^2$ --- completing a rotation whenever $x \in \R$ reaches the next natural number.

  We can scale $x$ by a frequency $\xi$: $e^{- i 2 \pi \xi x}$.
  If $\vx \in \R^d$, we can also scale each component $j$ of $\vx$ by a different frequency $\vxi(j)$.
  Multiplying a function $f : \R^d \to \R$ with the rotation around the unit circle with given frequencies $\vxi$, yields a quantity that describes the amplitude of the frequencies $\vxi$, \begin{align}
      \hat{f}(\vxi) \defeq \int_{\R^d} f(\vx) e^{- i 2 \pi \transpose{\vxi} \vx} \,d\vx. \label{eq:fourier_transform}
  \end{align}
  $\hat{f}$ is called the \midx{Fourier transform} of $f$.
  $f$ is called the \midx{inverse Fourier transform} of $\hat{f}$, and can be computed using \begin{align}
    f(\vx) = \int_{\R^d} \hat{f}(\vxi) e^{i 2 \pi \transpose{\vxi} \vx} \,d\vxi.
  \end{align}
  It is common to write $\vomega \defeq 2\pi\vxi$.
  See \cref{fig:fourier_transform} for an example.

  Refer to \icite{3b1bfouriertransform} for a visual introduction.
\end{rmk}

\begin{marginfigure}
  \begin{center}
    \import{./plots/output/}{fourier_transform.pgf}
  \end{center}

  \caption{The Fourier transform of a rectangular pulse, \begin{align*}
    f(x) \defeq \begin{cases}
      1 & x \in [-1, 1] \\
      0 & \text{otherwise},
    \end{cases}
  \end{align*} is given by \begin{align*}
    \hat{f}(\omega) &= \int_{-1}^1 e^{- i \omega x} \,dx = \frac{1}{i \omega} \parentheses*{e^{i \omega} - e^{- i \omega}} \\
    &= \frac{2 \sin(\omega)}{\omega}.
  \end{align*}}\label{fig:fourier_transform}
\end{marginfigure}

Because a stationary kernel $k : \R^d \times \R^d \to \R$ can be interpreted as a function in one variable, it has an associated Fourier transform which we denote by $p(\vomega)$.
That is, \begin{align}
  k(\vx-\vxp) = \int_{\R^d} p(\vomega) e^{i \transpose{\vomega} (\vx-\vxp)} \,d\vomega. \label{eq:spectral_density}
\end{align}

\begin{fct}[Bochner's theorem]\pidx{Bochner's theorem}
  A continuous stationary kernel on $\R^d$ is positive definite if and only if its Fourier transform $p(\vomega)$ is non-negative.
\end{fct}

Bochner's theorem implies that when a continuous and stationary kernel is positive definite and scaled appropriately, its Fourier transform~$p(\vomega)$ is a proper probability distribution.
In this case, $p(\vomega)$ is called the \midx{spectral density} of the kernel $k$.

\begin{rmk}{Eigenvalue spectrum of stationary kernels}{eigenvalue_spectrum_of_stationary_kernels}
  When a kernel $k$ is stationary (i.e., a univariate function of $\vx - \vxp$), its eigenfunctions (with respect to the usual Lebesgue measure) turn out to be the complex exponentials $\exp(i \transpose{\vomega} (\vx - \vxp))$.
  In simpler terms, you can think of these exponentials as ``building blocks'' at different frequencies $\vomega$.
  The spectral density $p(\vomega)$ associated with the kernel tells you how strongly each frequency contributes, i.e., how large the corresponding eigenvalue is.\looseness=-1

  A key insight of this analysis is that the rate at which these magnitudes $p(\vomega)$ decay with increasing frequency $\vomega$ reveals the smoothness of the processes governed by the kernel.
  If a kernel allocates more ``power'' to high frequencies (meaning the spectral density decays slowly), the resulting processes will appear ``rougher''.
  Conversely, if high-frequency components are suppressed, the process will appear ``smoother''.\looseness=-1

  For an in-depth introduction to the eigenfunction analysis of kernels, refer to section 4.3 of \icite{gpml}.
\end{rmk}

\begin{ex}{Spectral density of the Gaussian kernel}{}
  The Gaussian kernel with length scale $h$ has the spectral density \begin{align}
    p(\vomega) &= \int_{\R^d} k(\vx - \vxp; h) e^{-i \transpose{\vomega} (\vx-\vxp)} \,d(\vx-\vxp) \margintag{using the definition of the Fourier transform \eqref{eq:fourier_transform}} \nonumber \\
    &= \int_{\R^d} \exp\parentheses*{-\frac{\norm{\vx}_2^2}{2 h^2} - i \transpose{\vomega} \vx} \,d\vx \margintag{using the definition of the Gaussian kernel \eqref{eq:gaussian_kernel}} \nonumber \\
    &= (2 h^2 \pi)^{\nicefrac{d}{2}} \exp\parentheses*{-h^2 \frac{\norm{\vomega}_2^2}{2}}. \label{eq:spectral_density_gaussian_kernel}
  \end{align}
\end{ex}

The key idea is now to interpret the kernel as an expectation, \begin{align}
  k(\vx-\vxp) &= \int_{\R^d} p(\vomega) e^{i \transpose{\vomega} (\vx-\vxp)} \,d\vomega \margintag{from \cref{eq:spectral_density}} \nonumber \\
  &= \E[\vomega \sim p]{e^{i \transpose{\vomega}(\vx-\vxp)}} \margintag{by the definition of expectation \eqref{eq:expectation}} \nonumber \\
  &= \E[\vomega \sim p]{\cos(\transpose{\vomega}\vx-\transpose{\vomega}\vxp) + i\sin(\transpose{\vomega}\vx - \transpose{\vomega}\vxp)}. \margintag{using Euler's formula \eqref{eq:eulers_formula}} \nonumber \\
\intertext{Observe that as both $k$ and $p$ are real, convergence of the integral implies $\E[\vomega \sim p]{\sin(\transpose{\vomega}\vx - \transpose{\vomega}\vxp)} = 0$. Hence,}
  &= \E[\vomega \sim p]{\cos(\transpose{\vomega}\vx-\transpose{\vomega}\vxp)} \nonumber \\
  &= \E*[\vomega \sim p]{\E[b \sim \Unif{[0, 2\pi]}]{\cos((\transpose{\vomega}\vx + b)-(\transpose{\vomega}\vxp + b))}} \margintag{expanding with $b - b$} \nonumber \\
  &= \begin{multlined}[t]
    \E*[\vomega \sim p]{\E*[b \sim \Unif{[0, 2\pi]}]{\left[\cos(\transpose{\vomega}\vx + b) \cos(\transpose{\vomega}\vxp + b) \right. \\ \left. + \sin(\transpose{\vomega}\vx + b) \sin(\transpose{\vomega}\vxp + b)\right]}}
  \end{multlined} \margintag{using the angle subtraction identity, $\cos(\alpha - \beta) = \cos\alpha\cos\beta + \sin\alpha\sin\beta$} \nonumber \\
  &= \E*[\vomega \sim p]{\E[b \sim \Unif{[0, 2\pi]}]{2 \cos(\transpose{\vomega}\vx + b) \cos(\transpose{\vomega}\vxp + b)}} \margintag{using \begin{align*}
    &\E[b]{\cos(\alpha + b) \cos(\beta + b)} \\
    &= \E[b]{\sin(\alpha + b) \sin(\beta + b)}
  \end{align*} for $b \sim \Unif{[0, 2\pi]}$} \nonumber \\
  &= \E[\vomega \sim p, b \sim \Unif{\brackets{0, 2 \pi}}]{z_{\vomega,b}(\vx) \cdot z_{\vomega,b}(\vxp)}
\intertext{where $z_{\vomega,b}(\vx) \defeq \sqrt{2} \cos(\transpose{\vomega} \vx + b)$,}
  &\approx \frac{1}{m} \sum_{i=1}^m z_{\vomega^{(i)},b^{(i)}}(\vx) \cdot z_{\vomega^{(i)},b^{(i)}}(\vxp) \margintag{using Monte Carlo sampling to estimate the expectation, see \cref{ex:estimating_expectations}}
\intertext{for independent samples $\vomega^{(i)} \iid p$ and $b^{(i)} \iid \Unif{\brackets{0, 2 \pi}}$,}
  &= \transpose{\vz(\vx)} \vz(\vxp)
\end{align} where the (randomized) feature map of random Fourier features is \begin{align}
  \vz(\vx) \defeq \frac{1}{\sqrt{m}} \transpose{[z_{\vomega^{(1)},b^{(1)}}(\vx), \dots, z_{\vomega^{(m)},b^{(m)}}(\vx)]}.
\end{align}

Intuitively, each component of the feature map $\vz(\vx)$ projects $\vx$ onto a random direction $\vomega$ drawn from the (inverse) Fourier transform $p(\vomega)$ of $k(\vx-\vxp)$, and wraps this line onto the unit circle in $\R^2$.
After transforming two points $\vx$ and $\vxp$ in this way, their inner product is an unbiased estimator of $k(\vx - \vxp)$.
The mapping $z_{\vomega,b}(\vx) = \sqrt{2} \cos(\transpose{\vomega} \vx + b)$ additionally rotates the circle by a random amount $b$ and projects the points onto the interval $[0,1]$.

\begin{marginfigure}
  \begin{center}
    \import{./plots/output/}{rff.pgf}
  \end{center}

  \caption{Example of random Fourier features with where the number of features $m$ is $5$ (top) and $10$ (bottom), respectively.
  The noise-free true function is shown in black and the mean of the Gaussian process is shown in blue.}
\end{marginfigure}

\cite{rff} show that Bayesian linear regression with the feature map $\vz$ approximates Gaussian processes with a stationary kernel:

\begin{thm}[Uniform convergence of Fourier features]\label{thm:uniform_convergence_fourier_features}
  Suppose $\spM$ is a compact subset of $\R^d$ with diameter $\mathrm{diam}(\spM)$.
  Then for a stationary kernel $k$, the random Fourier features $\vz$, and any $\epsilon > 0$ it holds that \begin{align}
    \begin{multlined}[t]
      \Pr{\sup_{\vx,\vxp \in \spM} \abs{\transpose{\vz(\vx)} \vz(\vxp) - k(\vx - \vxp)} \geq \epsilon} \\
      \leq 2^8 \parentheses*{\frac{\sigmap \mathrm{diam}(\spM)}{\epsilon}}^2 \exp\parentheses*{-\frac{m \epsilon^2}{8(d + 2)}}
    \end{multlined}
  \end{align} where $\sigmap^2 \defeq \E[\vomega \sim p]{\transpose{\vomega} \vomega}$ is the second moment of $p$, $m$ is the dimension of~$\vz(\vx)$, and $d$ is the dimension of $\vx$. \exerciserefmark{uniform_convergence_of_fourier_features}
\end{thm}

Note that the error probability decays exponentially fast in the dimension of the Fourier feature space.

\subsection{Data Sampling}\label{sec:gp:approx:data_sampling}

Another natural approach is to only consider a (random) subset of the training data during learning.
The naive approach is to subsample uniformly at random.
Not very surprisingly, we can do much better.

\begin{marginfigure}
  \begin{center}
    \import{./plots/output/}{inducing_points.pgf}
  \end{center}

  \caption{Inducing points $\vu$ are shown as vertical dotted red lines.
  The noise-free true function is shown in black and the mean of the Gaussian process is shown in blue.
  Observe that the true function is approximated ``well'' around the inducing points.}
\end{marginfigure}

One subsampling method is the \midx{inducing points method} \citep{ip}.
The idea is to summarize the data around so-called inducing points.\footnote{The inducing points can be treated as hyperparameters.}
For now, let us consider an arbitrary set of inducing points, \begin{align*}
  \sU \defeq \{\overline{\vx}_1, \dots, \overline{\vx}_k\}.
\end{align*}

Then, the original Gaussian process can be recovered using marginalization,\looseness=-1 \begin{align}
  p(\fs, \vf) = \int_{\R^k} p(\fs, \vf, \vu) \,d\vu = \int_{\R^k} p(\fs, \vf \mid \vu) p(\vu) \,d\vu, \margintag{using the sum rule \eqref{eq:sum_rule} and product rule \eqref{eq:product_rule}}
\end{align} where $\vf \defeq \transpose{[f(\vx_1) \; \cdots \; f(\vx_n)]}$ and $\fs \defeq f(\vxs)$ at some evaluation point $\vxs \in \spX$.
We use $\vu \defeq \transpose{[f(\overline{\vx}_1) \; \cdots \; f(\overline{\vx}_k)]} \in \R^k$ to denote the predictions of the model at the inducing points $\sU$.
Due to the marginalization property of Gaussian processes \eqref{eq:gp_joint_distr}, we have that $\vu \sim \N{\vzero}{\mK_{\sU\sU}}$.
The key idea is to approximate the joint prior, assuming that $\fs$ and $\vf$ are conditionally independent given $\vu$, \begin{align}
  p(\fs, \vf) \approx \int_{\R^k} p(\fs \mid \vu) p(\vf \mid \vu) p(\vu) \,d\vu. \label{eq:inducing_points_joint_approx}
\end{align}
Here, $p(\vf \mid \vu)$ and $p(\fs \mid \vu)$ are commonly called the \midx{training conditional} and the \midx{testing conditional}, respectively.
Still denoting the observations by $\sA = \{\vx_1, \dots, \vx_n\}$ and defining $\star \defeq \{\vxs\}$, we know, using the closed-form expression for conditional Gaussians \eqref{eq:cond_gaussian}, \begin{subequations}\begin{align}
  p(\vf \mid \vu) &\sim \N[\vf]{\mK_{\sA\sU} \inv{\mK_{\sU\sU}} \vu}{\mK_{\sA\sA} - \mQ_{\sA\sA}}, \\
  p(\fs \mid \vu) &\sim \N[\fs]{\mK_{\star\sU} \inv{\mK_{\sU\sU}} \vu}{\mK_{\star\star} - \mQ_{\star\star}}
\end{align}\label{eq:inducing_point_conditionals}\end{subequations} where $\mQ_{a b} \defeq \mK_{a \sU} \inv{\mK_{\sU\sU}} \mK_{\sU b}$.
Intuitively, $\mK_{\sA\sA}$ represents the prior covariance and $\mQ_{\sA\sA}$ represents the covariance ``explained'' by the inducing points.\footnote{For more details, refer to section 2 of \icite{ip}.}\looseness=-1

Computing the full covariance matrix is expensive.
In the following, we mention two approximations to the covariance of the training conditional (and testing conditional).

\begin{ex}{Subset of regressors}{}
  The \midx{subset of regressors} (SoR) approximation is defined as \begin{subequations}\begin{align}
    q_{\mathrm{SoR}}(\vf \mid \vu) &\defeq \N[\vf]{\mK_{\sA\sU} \inv{\mK_{\sU\sU}} \vu}{\vzero}, \\
    q_{\mathrm{SoR}}(\fs \mid \vu) &\defeq \N[\fs]{\mK_{\star\sU} \inv{\mK_{\sU\sU}} \vu}{\vzero}.
  \end{align}\label{eq:sor}\end{subequations}
  Comparing to \cref{eq:inducing_point_conditionals}, SoR simply forgets about all variance and covariance.
\end{ex}

\begin{marginfigure}
  \begin{center}
    \import{./plots/output/}{inducing_points_sor_fitc.pgf}
  \end{center}

  \caption{Comparison of SoR (top) and FITC (bottom).
  The inducing points $\vu$ are shown as vertical dotted red lines.
  The noise-free true function is shown in black and the mean of the Gaussian process is shown in blue.}
\end{marginfigure}

\begin{ex}{Fully independent training conditional}{}
  The \midx{fully independent training conditional} (FITC) approximation is defined as \begin{subequations}\begin{align}
    q_{\mathrm{FITC}}(\vf \mid \vu) &\defeq \N[\vf]{\mK_{\sA\sU} \inv{\mK_{\sU\sU}} \vu}{\diag{\mK_{\sA\sA} - \mQ_{\sA\sA}}}, \\
    q_{\mathrm{FITC}}(\fs \mid \vu) &\defeq \N[\fs]{\mK_{\star\sU} \inv{\mK_{\sU\sU}} \vu}{\diag{\mK_{\star\star} - \mQ_{\star\star}}}.
  \end{align}\end{subequations}
  In contrast to SoR, FITC keeps track of the variances but forgets about the covariance.
\end{ex}

The computational cost for inducing point methods SoR and FITC is dominated by the cost of inverting $\mK_{\sU\sU}$.
Thus, the time complexity is cubic in the number of inducing points, but only linear in the number of data points.

\section*{Discussion}

This chapter introduced Gaussian processes which leverage the function-space view on linear regression to perform exact probabilistic inference with flexible, nonlinear models.
A Gaussian process can be seen as a non-parametric model since it can represent an infinite-dimensional parameter space.
Instead, as we saw with the representer theorem, such non-parametric (i.e., ``function-space'') models are directly represented as functions of the data points.
While this can make these models more flexible than a simple linear parametric model in input space, it also makes them computationally expensive as the number of data points grows.
To this end, we discussed several ways of approximating Gaussian processes.

Nevertheless, for today's internet-scale datasets, modern machine learning typically relies on large parametric models that learn features from data.
These models can effectively amortize the cost of inference during training by encoding information into a fixed set of parameters.
In the following chapters, we will start to explore approaches to approximate probabilistic inference that can be applied to such models.

\excheading

\begin{nexercise}{Feature space of Gaussian kernel}{gaussian_kernel_feature_space}
  \begin{enumerate}[label=\arabic*.]
    \item Show that the univariate Gaussian kernel with length scale $h = 1$ implicitly defines a feature space with basis vectors \begin{align*}
      \vphi(x) = \begin{bmatrix}\phi_0(x) \\ \phi_1(x) \\ \vdots\end{bmatrix} \quad\text{with}\quad \phi_j(x) = \frac{1}{\sqrt{j!}} e^{-\frac{x^2}{2}} x^j.
    \end{align*}
    \textit{Hint: Use the Taylor series expansion of the exponential function, ${e^x = \sum_{j=0}^\infty \frac{x^j}{j!}}$.}

    \item Note that the vector $\vphi(x)$ is $\infty$-dimensional.
    Thus, taking the function-space view allows us to perform regression in an infinite-dimensional feature space.
    What is the effective dimension when regressing $n$ univariate data points with a Gaussian kernel?
  \end{enumerate}\vspace{5pt}
\end{nexercise}

\begin{nexercise}{Kernels on the circle}{kernels_on_the_circle}
  Consider a dataset $\{(\vx_i, y_i)\}_{i=1}^n$ with labels $y_i \in \R$ and inputs $\vx_i$ which lie on the unit circle $\mathbb{S} \subset \R^2$.
  In particular, any element of $\mathbb{S}$ can be identified with points in $\R^2$ of form $(\cos(\theta), \sin(\theta))$ or with the respective angles $\theta \in [0, 2 \pi)$.

  You now want to use GP regression to learn an unknown mapping from $\mathbb{S}$ to $\R$ using this dataset.
  Thus, you need a valid kernel $k: \mathbb{S} \times \mathbb{S} \to \R$.
  First, we look at kernels $k$ which can be understood as analogous to the Gaussian kernel.
  \begin{enumerate}
    \item You think of the ``extrinsic'' kernel $k_{\text{e}}: \mathbb{S} \times \mathbb{S} \to \R$ defined by
    \[
    k_{\text{e}}(\theta, \theta') \defeq \exp\left(-\frac{\|\vx(\theta)-\vx(\theta')\|_2^2}{2 \kappa^2}\right),
    \]
    where $\vx(\theta) \defeq (\cos(\theta), \sin(\theta))$.
    Is $k_{\text{e}}$ positive semi-definite for all values of $\kappa > 0$?

    \item Then, you think of an ``intrinsic'' kernel $k_{\text{i}}: \mathbb{S} \times \mathbb{S} \to \R$ defined by
    \[
    k_{\text{i}}(\theta, \theta') \defeq \exp\left(-\frac{d(\theta, \theta')^2}{2 \kappa^2}\right)
    \]
    where $d(\theta, \theta') \defeq \min(|\theta - \theta'|, |\theta - \theta' - 2 \pi|, |\theta - \theta' + 2 \pi|)$ is the standard arc length distance on the circle~$\mathbb{S}$.\par
    You would now like to test whether this kernel is positive semi-definite.
    We pick $\kappa=2$ and compute the kernel matrix $\mK$ for the points corresponding to the angles $\{0, \nicefrac{\pi}{2}, \pi, \nicefrac{3 \pi}{2} \}$.
    This kernel matrix $\mK$ has eigenvectors  $(1, 1, 1, 1)$ and $(-1, 1, -1, 1)$.\par
    Now compute the eigenvalue corresponding to the eigenvector $(-1, 1, -1, 1)$.

    \item Is $k_{\text{i}}$ positive semi-definite for $\kappa = 2$?

    \item A mathematician friend of yours suggests to you yet another kernel for points on the circle $\mathbb{S}$, called the \midx{heat kernel}.
    The kernel itself has a complicated expression but can be accurately approximated by
    \[
      k_h(\theta, \theta') \defeq \frac{1}{C_{\kappa}} \left( 1 + \sum_{l=1}^{L-1} e^{-\frac{\kappa^2}{2} l^2} 2 \cos(l (\theta - \theta')) \right),
    \]
    where $L \in \Nat$ controls the quality of approximation and $C_{\kappa} > 0$ is a normalizing constant that depends only on $\kappa$.\par
    Is $k_h$ is positive semi-definite for all values of $\kappa > 0$ and $L \in \Nat$?\par
    \textit{Hint: Recall that $\cos(a - b) = \cos(a)\cos(b) + \sin(a) \sin(b)$.}
  \end{enumerate}
\end{nexercise}

\begin{nexercise}{A Kalman filter as a Gaussian process}{kf_as_gp}
  Next we will show that the Kalman filter from \cref{ex:kf_rand_walk_1d} can be seen as a Gaussian process.
  To this end, we define \begin{align}
    f : \Nat_0 \to \R, \quad t \mapsto X_t.
  \end{align}
  Assuming that $X_0 \sim \N{0}{\sigma_0^2}$ and $X_{t+1} \defeq X_t + \varepsilon_t$ with independent noise $\varepsilon_t \sim \N{0}{\sigma_x^2}$, show that \begin{align}
    f &\sim \GP{0}{k_{\mathrm{KF}}} \quad\text{where} \\
    k_{\mathrm{KF}}(t, t') &\defeq \sigma_0^2 + \sigma_x^2 \min\{t, t'\}.
  \end{align}

  This particular kernel $k(t, t') \defeq \min\{t, t'\}$ but over the continuous-time domain defines the \midx{Wiener process} (also known as Brownian motion).
\end{nexercise}

\begin{nexercise}{Reproducing property and RKHS norm}{reproducing_kernel_hilbert_space_properties}
  \begin{enumerate}
    \item Derive the reproducing property.
    \par\textit{Hint: Use $k(\vx,\vxp) = \ip{k(\vx,\cdot), k(\vxp,\cdot)}_k$.}

    \item Show that the RKHS norm $\norm{\cdot}_k$ is a measure of smoothness by proving that for any $f \in \spH_k(\spX)$ and $\vx,\vy \in \spX$ it holds that \begin{align*}
      |f(\vx) - f(\vy)| \leq \norm{f}_k \norm{k(\vx,\cdot) - k(\vy,\cdot)}_k.
    \end{align*}
  \end{enumerate}
\end{nexercise}

\begin{nexercise}{Representer theorem}{representer_theorem}
  With this, we can now derive the representer theorem~\eqref{eq:representer_theorem}.\par
  \textit{Hint: Recall \begin{enumerate}
    \item the reproducing property $f(\vx) = \ip{f, k(\vx, \cdot)}_k$ with $k(\vx, \cdot) \in \spH_k(\spX)$ which holds for all $f \in \spH_k(\spX)$ and $\vx \in \spH_k(\spX)$, and
    \item that the norm after projection is smaller or equal the norm before projection.
  \end{enumerate}
  Then decompose $f$ into parallel and orthogonal components with respect to $\mathrm{span}\{k(\vx_1, \cdot), \dots, k(\vx_n, \cdot)\}$.}
\end{nexercise}

\begin{nexercise}{MAP estimate of Gaussian processes}{mle_and_map_of_gps}
  Let us denote by $A = \{\vx_1, \dots, \vx_n\}$ the set of training points.
  We will now show that the MAP estimate of GP regression corresponds to the solution of the regularized linear regression problem in the RKHS stated in \cref{eq:map_of_gp}: \begin{align*}
    \hat{f} \defeq \argmin_{f \in \spH_k(\spX)} - \log p(y_{1:n} \mid \vx_{1:n}, f) + \frac{1}{2} \norm{f}_k^2.
  \end{align*}

  In the following, we abbreviate $\mK = \mK_{AA}$.
  We will also assume that the GP has a zero mean function.
  \begin{enumerate}
    \item Show that \cref{eq:map_of_gp} is equivalent to \begin{align}
      \valphahat \defeq \argmin_{\valpha \in \R^n} \norm{\vy - \mK \valpha}_2^2 + \lambda \norm{\valpha}_{\mK}^2 \label{eq:map_of_gp2}
    \end{align} for some $\lambda > 0$ which is also known as \midx{kernel ridge regression}.
    Determine $\lambda$.

    \item Show that \cref{eq:map_of_gp2} with the $\lambda$ determined in (1) is equivalent to the MAP estimate of GP regression.
    \par\textit{Hint: Recall from \cref{eq:gp_posterior} that the MAP estimate at a point $\vxs$ is $\E{\fs}[\vxs, \mX, \vy] = \transpose{\vk_{\vxs,A}}\inv{(\mK + \sigman^2 \mI)}\vy$.}
  \end{enumerate}
\end{nexercise}

\begin{nexercise}{Gradient of the marginal likelihood}{gradient_of_mll}
  In this exercise, we derive \cref{eq:gradient_mll}.

  Recall that we were considering a dataset $(\mX, \vy)$ of noise-perturbed evaluations $y_i = f(\vx_i) + \varepsilon_i$ where $\varepsilon_i \sim \N{0}{\sigman^2}$ and $f$ is an unknown function.
  We make the hypothesis $f \sim \GP{0}{k_{\vtheta}}$ with a zero mean function and the covariance function $k_{\vtheta}$.
  We are interested in finding the hyperparameters $\vtheta$ that maximize the marginal likelihood~${p(\vy \mid \mX, \vtheta)}$.

  \begin{enumerate}
    \item Derive \cref{eq:gradient_mll}. \\
    \textit{Hint: You can use the following identities: \begin{enumerate}
      \item for any invertible matrix $\mM$, \begin{align}
        \pdv{}{\theta_j} \inv{\mM} = -\inv{\mM} \pdv{\mM}{\theta_j} \inv{\mM} \quad\text{and} \label{eq:gradient_of_mll_hint1}
      \end{align}
      \item for any symmetric positive definite matrix $\mM$, \begin{align}
        \pdv{}{\theta_j} \log \det{\mM} = \tr{\inv{\mM} \pdv{\mM}{\theta_j}}. \label{eq:gradient_of_mll_hint2}
      \end{align}
    \end{enumerate}}

    \item Assume now that the covariance function for the noisy targets (i.e., including the noise contribution) can be expressed as \begin{align*}
      k_{\vy,\vtheta}(\vx, \vxp) = \theta_0 \tilde{k}(\vx, \vxp)
    \end{align*} where $\tilde{k}$ is a valid kernel independent of $\theta_0$.\safefootnote{That is, $\mK_{\vy,\vtheta}(i, j) = k_{\vy,\vtheta}(\vx_i, \vx_j)$.}

    Show that ${\pdv{}{\theta_0} \log p(\vy \mid \mX, \vtheta) = 0}$ admits a closed-form solution for $\theta_0$ which we denote by~$\opt{\theta_0}$.

    \item How should the optimal parameter $\opt{\theta_0}$ be scaled if we scale the labels $\vy$ by a scalar $s$?
  \end{enumerate}
\end{nexercise}

\begin{nexercise}{Uniform convergence of Fourier features}{uniform_convergence_of_fourier_features}
  In this exercise, we will prove \cref{thm:uniform_convergence_fourier_features}.

  Let $s(\vx, \vxp) \defeq \transpose{\vz(\vx)} \vz(\vxp)$ and $f(\vx, \vxp) \defeq s(\vx, \vxp) - k(\vx, \vxp)$.
  Observe that both functions are shift invariant, and we will therefore denote them as univariate functions with argument $\vDelta \equiv \vx - \vxp \in \spM_\Delta$.
  Notice that our goal is to bound the probability of the event $\sup_{\vDelta \in \spM_\Delta} |f(\vDelta)| \geq \epsilon$.

  \begin{enumerate}
    \item Show that for all $\vDelta \in \spM_\Delta$, $\Pr{|f(\vDelta)| \geq \epsilon} \leq 2 \exp\parentheses*{-\frac{m \epsilon^2}{4}}$.
  \end{enumerate}

  What we have derived in (1) is known as a \midx{pointwise convergence} guarantee.
  However, we are interested in bounding the \midx{uniform convergence} over the compact set $\spM_\Delta$.

  Our approach will be to ``cover'' the compact set $\spM_\Delta$ using $T$ balls of radius $r$ whose centers we denote by $\{\vDelta_i\}_{i=1}^T$.
  It can be shown that this is possible for some $T \leq (4 \; \mathrm{diam}(\spM) / r)^d$.
  It can furthermore be shown that \begin{align*}
    \text{$\forall i$. $|f(\vDelta_i)| < \frac{\epsilon}{2}$ and $\norm{\grad f(\opt{\vDelta})}_2 < \frac{\epsilon}{2 r}$} \implies \sup_{\vDelta \in \spM_\Delta} |f(\vDelta)| < \epsilon
  \end{align*} where $\opt{\vDelta} = \argmax_{\vDelta \in \spM_\Delta} \norm{\grad f(\vDelta)}_2$.

  \begin{enumerate}
    \setcounter{enumi}{1}
    \item Prove $\Pr{\norm{\grad f(\opt{\vDelta})}_2 \geq \frac{\epsilon}{2 r}} \leq \parentheses*{\frac{2 r \sigmap}{\epsilon}}^2$.\par
    \textit{Hint: Recall that the random Fourier feature approximation is unbiased, i.e., $\E{s(\vDelta)} = k(\vDelta)$.}

    \item Prove $\Pr{\bigcup_{i=1}^T |f(\vDelta_i)| \geq \frac{\epsilon}{2}} \leq 2 T \exp\parentheses*{-\frac{m \epsilon^2}{16}}$.

    \item Combine the results from (2) and (3) to prove \cref{thm:uniform_convergence_fourier_features}.\par
    \textit{Hint: You may use that \begin{enumerate}
      \item $\alpha r^{-d} + \beta r^2 = 2 \beta^{\frac{d}{d+2}} \alpha^{\frac{2}{d+2}}$ for $r = (\alpha / \beta)^{\frac{1}{d+2}}$ and
      \item $\frac{\sigmap \mathrm{diam}(\spM)}{\epsilon} \geq 1$.
    \end{enumerate}}

    \item Show that for the Gaussian kernel \eqref{eq:gaussian_kernel}, $\sigmap^2 = \frac{d}{h^2}$.\par
    \textit{Hint: First show $\sigmap^2 = - \tr{\hes_{\vDelta} k(\vzero)}$.}
  \end{enumerate}
\end{nexercise}

\begin{nexercise}{Subset of regressors}{sor}
  \begin{enumerate}
    \item Using an SoR approximation, prove the following: \begin{align}
      q_{\mathrm{SoR}}(\vf, \fs) &= \N*[\begin{bmatrix}
        \vf \\
        \fs
      \end{bmatrix}]{\vzero}{\begin{bmatrix}
        \mQ_{\sA\sA} & \mQ_{\sA\star} \\
        \mQ_{\star\sA} & \mQ_{\star\star} \\
      \end{bmatrix}} \\
      q_{\mathrm{SoR}}(\fs \mid \vy) &= \N[\fs]{\mQ_{\star\sA}\inv{\Tilde{\mQ}_{\sA\sA}}\vy}{\mQ_{\star\star} - \mQ_{\star\sA}\inv{\Tilde{\mQ}_{\sA\sA}}\mQ_{\sA\star}}
    \end{align} where $\Tilde{\mQ}_{a b} \defeq \mQ_{a b} + \sigman^2$.

    \item Derive that the resulting model is a degenerate Gaussian process with covariance function \begin{align}
      k_{\mathrm{SoR}}(\vx,\vxp) \defeq \transpose{\vk_{\vx,\sU}} \inv{\mK_{\sU\sU}} \vk_{\vxp,\sU}.
    \end{align}
  \end{enumerate}
\end{nexercise}

  \chapter{Variational Inference}\label{sec:approximate_inference}

We have seen how to perform (efficient) probabilistic inference with Gaussians, exploiting their closed-form formulas for marginal and conditional distributions.
But what if we work with other distributions?

In this and the following chapter, we will discuss two methods of approximate inference.
We begin by discussing variational (probabilistic) inference, which aims to find a good approximation of the posterior distribution from which it is easy to sample.
In \cref{sec:approximate_inference:mcmc}, we discuss Markov chain Monte Carlo methods, which approximate the sampling from the posterior distribution directly.

The fundamental idea behind variational inference is to approximate the true posterior distribution using a ``simpler'' posterior that is as close as possible to the true posterior:
\begin{align}
  p(\vtheta \mid \vx_{1:n}, y_{1:n}) = \frac{1}{Z} p(\vtheta, y_{1:n} \mid \vx_{1:n}) \approx q(\vtheta \mid \vlambda) \eqdef q_\vlambda(\vtheta)
\end{align} where $\vlambda$ represents the parameters of the \midx{variational posterior} $q_\vlambda$, also called \midx{variational parameters}.
In doing so, variational inference reduces probabilistic inference --- where the fundamental difficulty lies in solving high-dimensional integrals --- to an optimization problem.
Optimizing (stochastic) objectives is a well-understood problem with efficient algorithms that perform well in practice.\footnote{We provide an overview of first-order methods such as stochastic gradient descent in \cref{sec:fundamentals:optimization}.}

\section{Laplace Approximation}\label{sec:approximate_inference:laplace_approximation}

Before introducing a general framework of variational inference, we discuss a simpler method of approximate inference known as \midx{Laplace's method}.
This method was proposed as a method of approximating integrals as early as 1774 by Pierre-Simon Laplace.
The idea is to use a Gaussian approximation (that is, a second-order Taylor approximation) of the posterior distribution around its mode.
Let \begin{align}
  \psi(\vtheta) \defeq \log p(\vtheta \mid \vx_{1:n}, y_{1:n}) \label{eq:log_posterior}
\end{align} denote the log-posterior.
Then, using a second-order Taylor approximation \eqref{eq:second_order_exp} around the mode $\vthetahat$ of $\psi$ (i.e., the MAP estimate), we obtain the approximation $\hat{\psi}$ which is accurate for $\vtheta \approx \vthetahat$:\looseness=-1 \begin{align}
  \psi(\vtheta) \approx \hat{\psi}(\vtheta) &\defeq \psi(\vthetahat) + \transpose{(\vtheta - \vthetahat)} \grad \psi(\vthetahat) + \frac{1}{2} \transpose{(\vtheta - \vthetahat)} \hes_\psi(\vthetahat) (\vtheta - \vthetahat) \nonumber\\
  &= \psi(\vthetahat) + \frac{1}{2} \transpose{(\vtheta - \vthetahat)} \hes_\psi(\vthetahat) (\vtheta - \vthetahat). \margintag{using $\grad \psi(\vthetahat) = 0$}
\end{align}

Compare this expression to the log-PDF of a Gaussian: \begin{align}
  \log \N[\vtheta]{\vthetahat}{\inv{\mLambda}} = -\frac{1}{2}\transpose{(\vtheta - \vthetahat)}\mLambda(\vtheta - \vthetahat) + \const.
\end{align}
Since $\psi(\vthetahat)$ is constant with respect to $\vtheta$, \begin{align}
  \hat{\psi}(\vtheta) = \log \N[\vtheta]{\vthetahat}{- \inv{\hes_\psi(\vthetahat)}} + \const.
\end{align}
The \midx{Laplace approximation} $q$ of $p$ is \begin{subequations}\begin{align}
  q(\vtheta) &\defeq \N[\vtheta]{\vthetahat}{\inv{\mLambda}} \propto \exp(\hat{\psi}(\vtheta)) \qquad\text{where} \\
  \mLambda &\defeq - \hes_\psi(\vthetahat) = - \hes_\vtheta \log p(\vtheta \mid \vx_{1:n}, y_{1:n}) \bigl|_{\vtheta = \vthetahat}.
\end{align}\end{subequations}
Recall that for this approximation to be well-defined, the covariance matrix $\inv{\mLambda}$ (or equivalently the precision matrix $\mLambda$) needs to be symmetric and positive semi-definite.
Let us verify that this is indeed the case for sufficiently smooth $\psi$.\footnote{$\psi$ being twice continuously differentiable around $\vthetahat$ is sufficient.}
In this case, the Hessian $\mLambda$ is symmetric since the order of differentiation does not matter.
Moreover, by the second-order optimality condition, $\hes_\psi(\vthetahat)$ is negative semi-definite since $\vthetahat$ is a maximum of $\psi$, which implies that $\mLambda$ is positive semi-definite.\looseness=-1

\begin{ex}{Laplace approximation of a Gaussian}{}
  Consider approximating the Gaussian density $p(\vtheta) = \N[\vtheta]{\vmu}{\mSigma}$ using a Laplace approximation.

  We know that the mode of $p$ is $\vmu$, which we can verify by computing the gradient, \begin{align}
    \grad_\vtheta \log p(\vtheta) = -\frac{1}{2} (2 \inv{\mSigma} \vtheta - 2 \inv{\mSigma} \vmu) \overset{!}{=} 0 \iff \vtheta = \vmu.
  \end{align}
  For the Hessian of $\log p(\vtheta)$, we get \begin{align}
    \hes_\vtheta \log p(\vtheta) = \transpose{(\jac_\vtheta (\inv{\mSigma} \vmu - \inv{\mSigma} \vtheta))} = - \transpose{(\inv{\mSigma})} = - \inv{\mSigma}. \margintag{using $\transpose{(\inv{\mA})} = \inv{(\transpose{\mA})}$ and symmetry of $\mSigma$}
  \end{align}
  We see that the Laplace approximation of a Gaussian $p(\vtheta)$ is exact, which should not come as a surprise since the second-order Taylor approximation of $\log p(\vtheta)$ is exact for Gaussians.\looseness=-1
\end{ex}

\begin{marginfigure}
  \begin{center}
    \import{./plots/output/}{laplace_approx.pgf}
  \end{center}

  \caption{The Laplace approximation $q$ greedily selects the mode of the true posterior distribution $p$ and matches the curvature around the mode $\hat{p}$.
  As shown here, the Laplace approximation can be extremely overconfident when $p$ is not approximately Gaussian.}
  \label{fig:laplace_approx_overconfident}
\end{marginfigure}

The Laplace approximation matches the shape of the true posterior around its mode but may not represent it accurately elsewhere --- often leading to extremely overconfident predictions.
An example is given in \cref{fig:laplace_approx_overconfident}.
Nevertheless, the Laplace approximation has some desirable properties such as being relatively easy to apply in a post-hoc manner, that is, after having already computed the MAP estimate.
It preserves the MAP point estimate as its mean and just ``adds'' a little uncertainty around it.
However, the fact that it can be arbitrarily different from the true posterior makes it unsuitable for approximate probabilistic inference.\looseness=-1

\subsection{Example: Bayesian Logistic Regression}\label{sec:approximate_inference:bayesian_logistic_regression}

As an example, we will look at Laplace approximation in the context of Bayesian logistic regression.
Logistic regression learns a classifier that decides for a given input whether it belongs to one of two classes.
{\def\par{\let\par\endgraf}\begin{marginfigure}
  \begin{center}
    \import{./plots/output/}{logistic_function.pgf}
  \end{center}

  \caption{The logistic function squashes the linear function $\transpose{\vw}\vx$ onto the interval $(0,1)$.}\label{fig:logistic_function}
\end{marginfigure}}%
A sigmoid function, typically the \midx{logistic function}, \begin{align}
  \sigma(z) \defeq \frac{1}{1+\exp(-z)} \in (0, 1), \qquad z = \transpose{\vw} \vx, \label{eq:logistic_function}
\end{align} is used to obtain the class probabilities.
{\def\par{\let\par\endgraf}\begin{marginfigure}
  \begin{center}
    \import{./plots/output/}{logistic_regression.pgf}
  \end{center}

  \caption{Logistic regression classifies data into two classes with a linear decision boundary.}
\end{marginfigure}}%
\midx{Bayesian logistic regression} corresponds to Bayesian linear regression with a Bernoulli likelihood, \begin{align}
  y \mid \vx, \vw \sim \Bern{\sigma(\transpose{\vw} \vx)},
\end{align} where ${y \in \{-1, 1\}}$ is the binary class label.\footnote[][0.5\baselineskip]{The same approach extends to Gaussian processes where it is known as \midx{Gaussian process classification}, see \cref{exercise:gpc} and \cite{gpc}.}
Observe that given a data point $(\vx,y)$, the probability of a correct classification is \begin{align}
  p(y \mid \vx, \vw) = \begin{cases}
    \sigma(\transpose{\vw} \vx) & \text{if $y = 1$} \\
    1 - \sigma(\transpose{\vw} \vx) & \text{if $y = -1$}
  \end{cases} = \sigma(y \transpose{\vw} \vx) \label{eq:logistic_regression_prob},
\end{align} as the logistic function $\sigma$ is symmetric around $0$.
Also, recall that Bayesian linear regression used the prior \begin{align*}
  p(\vw) = \N[\vw]{\vzero}{\sigmap^2 \mI} \propto \exp\parentheses*{-\frac{1}{2 \sigmap^2} \norm{\vw}_2^2}.
\end{align*}

Let us first find the posterior mode, that is, the MAP estimate of the weights:\looseness=-1 \begin{align}
  \vwhat &= \argmax_\vw p(\vw \mid \vx_{1:n}, y_{1:n}) \nonumber \\
  &= \argmax_\vw p(\vw) p(y_{1:n} \mid \vx_{1:n}, \vw) \margintag{using Bayes' rule \eqref{eq:bayes_rule}} \nonumber \\
  &= \argmax_\vw \log p(\vw) + \log p(y_{1:n} \mid \vx_{1:n}, \vw) \margintag{taking the logarithm} \nonumber \\
  &= \argmax_\vw - \frac{1}{2 \sigmap^2} \norm{\vw}_2^2 + \sum_{i=1}^n \log \sigma(y_i \transpose{\vw} \vx_i) \margintag{using independence of the observations and \cref{eq:logistic_regression_prob}} \nonumber \\
  &= \argmin_\vw \frac{1}{2 \sigmap^2} \norm{\vw}_2^2 + \sum_{i=1}^n \log(1 + \exp(-y_i \transpose{\vw} \vx_i)). \margintag{using the definition of $\sigma$ \eqref{eq:logistic_function}} \label{eq:logistic_regression}
\end{align}
Note that for $\lambda = \nicefrac{1}{2 \sigmap^2}$, the above optimization is equivalent to standard (regularized) logistic regression where \begin{align}
  \ell_{\mathrm{log}}(\transpose{\vw} \vx; y) \defeq \log(1 + \exp(-y \transpose{\vw} \vx)) \label{eq:logistic_loss}
\end{align} is called \midx{logistic loss}.
The gradient of the logistic loss is given by \exerciserefmark{logistic_loss_gradient}[1] \begin{align}
  \grad_\vw \ell_{\mathrm{log}}(\transpose{\vw} \vx; y) = - y \vx \cdot \sigma(-y \transpose{\vw} \vx). \label{eq:logistic_loss_gradient}
\end{align}
Recall that due to the symmetry of $\sigma$ around $0$, $\sigma(-y \transpose{\vw} \vx)$ is the probability that $\vx$ was \emph{not} classified as $y$.
Intuitively, if the model is ``surprised'' by the label, the gradient is large.

We can therefore use SGD with the (regularized) gradient step and with batch size $1$, \begin{align}
  \vw \gets \vw(1-2 \lambda \eta_t) + \eta_t y \vx \sigma(-y \transpose{\vw} \vx),
\end{align} for the data point $(\vx,y)$ picked uniformly at random from the training data.
Here, $2 \lambda \eta_t$ is due to the gradient of the regularization term, in effect, performing weight decay.

\begin{ex}{Laplace approx. of Bayesian logistic regression}{bayesian_logistic_regression}
  We have already found the mode of the posterior distribution, $\vwhat$.

  Let us denote by \begin{align}
    \pi_i \defeq \Pr{y_i = 1 \mid \vx_i, \vwhat} = \sigma(\transpose{\vwhat} \vx_i) \label{eq:log_loss_hes_pi}
  \end{align} the probability of $\vx_i$ belonging to the positive class under the model given by the MAP estimate of the weights.
  For the precision matrix, we then have \begin{align}
    \mLambda &= - \left. \hes_\vw \log p(\vw \mid \vx_{1:n}, y_{1:n}) \right\rvert_{\vw = \vwhat} \nonumber \\
    &= - \left. \hes_\vw \log p(y_{1:n} \mid \vx_{1:n}, \vw) \right\rvert_{\vw = \vwhat} - \left. \hes_\vw \log p(\vw) \right\rvert_{\vw = \vwhat} \nonumber \\
    &= \sum_{i=1}^n \left. \hes_\vw \ell_{\mathrm{log}}(\transpose{\vw} \vx_i; y_i) \right\rvert_{\vw = \vwhat} + \sigmap^{-2}\mI \margintag{using the definition of the logistic loss \eqref{eq:logistic_loss}} \nonumber \\
    &= \sum_{i=1}^n \vx_i \transpose{\vx_i} \pi_i(1-\pi_i) + \sigmap^{-2}\mI \margintag{using the Hessian of the logistic loss \eqref{eq:logistic_loss_hes} which you derive in \textbf{\cref{exercise:logistic_loss_gradient} (2)}} \nonumber \\
    &= \transpose{\mX} \diag[i\in[n]]{\pi_i(1-\pi_i)} \mX + \sigmap^{-2}\mI. \label{eq:prec_mat_lap_approx_blr}
  \end{align}
  Observe that $\pi_i(1-\pi_i) \approx 0$ if $\pi_i \approx 1$ or $\pi_i \approx 0$.
  That is, if a training example is ``well-explained'' by $\vwhat$, then its contribution to the precision matrix is small.
  In contrast, we have $\pi_i(1-\pi_i) = 0.25$ for $\pi_i = 0.5$.
  Importantly, $\mLambda$ does not depend on the normalization constant of the posterior distribution which is hard to compute.

  In summary, we have that ${\N{\vwhat}{\inv{\mLambda}}}$ is the Laplace approximation of ${p(\vw \mid \vx_{1:n}, y_{1:n})}$.
\end{ex}

\section{Predictions with a Variational Posterior}

How can we make predictions using our variational approximation?
We simply approximate the (intractable) true posterior with our variational posterior: \begin{align}
  p(\ys \mid \vxs, \vx_{1:n}, y_{1:n}) &= \int p(\ys \mid \vxs, \vtheta) p(\vtheta \mid \vx_{1:n}, y_{1:n}) \,d\vtheta \margintag{using the sum rule \eqref{eq:sum_rule}} \nonumber \\
  &\approx \int p(\ys \mid \vxs, \vtheta) q_\vlambda(\vtheta) \,d\vtheta. \label{eq:var_prediction0}
  \intertext{A straightforward approach is to observe that \cref{eq:var_prediction0} can be viewed as an expectation over the variational posterior $q_\vlambda$ and approximated via Monte Carlo sampling:}
  &= \E[\vtheta \sim q_\vlambda]{p(\ys \mid \vxs, \vtheta)} \\
  &\approx \frac{1}{m} \sum_{j=1}^m p(\ys \mid \vxs, \vtheta_j)
\end{align} where $\vtheta_j \iid q_\vlambda$.

\begin{ex}{Predictions in Bayesian logistic regression}{}
  In the case of Bayesian logistic regression with a Gaussian approximation of the posterior, we can obtain more accurate predictions.

  Observe that the final prediction $\ys$ is conditionally independent of the model parameters $\vw$ given the ``latent value'' $\fs = \transpose{\vw} \vxs$:\begin{align}
    p(\ys \mid \vxs, \vx_{1:n}, y_{1:n}) &\approx \int p(\ys \mid \vxs, \vw) q_\vlambda(\vw) \,d\vw \nonumber \\
    &= \int \int p(\ys \mid \fs) p(\fs \mid \vxs, \vw) q_\vlambda(\vw) \,d\vw \,d\fs \margintag{once more, using the sum rule \eqref{eq:sum_rule}} \nonumber \\
    &= \int p(\ys \mid \fs) \int p(\fs \mid \vxs, \vw) q_\vlambda(\vw) \,d\vw \,d\fs. \margintag{rearranging terms} \label{eq:var_prediction}
  \end{align}
  The outer integral can be readily approximated since it is only one-dimensional!
  The challenging part is the inner integral, which is a high-dimensional integral over the model weights $\vw$.
  Since the posterior over weights $q_\vlambda(\vw) = \N[\vw]{\vwhat}{\inv{\mLambda}}$ is a Gaussian, we have due to the closedness properties of Gaussians \eqref{eq:gaussian_lin_trans} that \begin{align}
    \int p(\fs \mid \vxs, \vw) q_\vlambda(\vw) \,d\vw = \N{\transpose{\vwhat} \vxs}{\transpose{\vxs} \inv{\mLambda} \vxs}. \label{eq:var_prediction3}
  \end{align}
  Crucially, this is a one-dimensional Gaussian in function-space as opposed to the $d$-dimensional Gaussian $q_\vlambda$ in weight-space!

  As we have seen in \cref{eq:logistic_regression_prob}, for Bayesian logistic regression, the prediction $\ys$ depends deterministically on the predicted latent value $\fs$: $p(\ys \mid \fs) = \sigma(\ys \fs)$.
  Combining \cref{eq:var_prediction,eq:var_prediction3}, we obtain \begin{align}
    p(\ys \mid \vxs, \vx_{1:n}, y_{1:n}) \approx \int \sigma(\ys \fs) \cdot \N[\fs]{\transpose{\vwhat} \vxs}{\transpose{\vxs} \inv{\mLambda} \vxs} \,d\fs.
  \end{align}
  We have replaced the high-dimensional integral over the model parameters $\vtheta$ by the one-dimensional integral over the prediction of our variational posterior $\fs$.
  While this integral is generally still intractable, it can be approximated efficiently using numerical quadrature methods such as the Gauss-Legendre quadrature or alternatively with Monte Carlo sampling.
\end{ex}

\section{Blueprint of Variational Inference}\label{sec:vi:blueprint}\pidx{variational inference}[idxpagebf]%

General probabilistic inference poses the challenge of approximating the posterior distribution with limited memory and computation, resource constraints also present in humans and other intelligent systems.
These resource constraints require information to be compressed, and as we will see, such a compression poses a fundamental tradeoff between model accuracy (on the observed data) and model complexity (to avoid overfitting).

Laplace approximation approximates the true (intractable) posterior with a simpler one, by greedily matching mode and curvature around it.
Can we find ``less greedy'' approaches?
We can view variational probabilistic inference more generally as a family of approaches aiming to approximate the true posterior distribution by one that is closest (according to some criterion) among a ``simpler'' class of distributions.
To this end, we need to fix a class of distributions and define suitable criteria, which we can then optimize numerically.
The key benefit is that we can reduce the (generally intractable) problem of high-dimensional integration to the (often much more tractable) problem of optimization.\looseness=-1

\begin{defn}[Variational family]\pidx{variational family}
  Let $\spP$ be the class of all probability distributions.
  A \emph{variational family} $\spQ \subseteq \spP$ is a class of distributions such that each distribution $q \in \spQ$ is characterized by unique variational parameters $\vlambda \in \Lambda$.
\end{defn}

\begin{marginfigure}
  \incfig{variational_families}
  \caption{An illustration of variational inference in the space of distributions $\spP$.
  The variational distribution $\qs \in \spQ$ is the optimal approximation of the true posterior $p$.}
  \label{fig:variational_families}
\end{marginfigure}

\begin{ex}{Family of independent Gaussians}{var_family_diag_gaussian}
  A straightforward example for a variational family is the family of independent Gaussians, \begin{align}
    \spQ \defeq \braces*{q(\vtheta) = \N[\vtheta]{\vmu}{\diag[i\in[d]]{\sigma_i^2}}},
  \end{align} which is parameterized by ${\vlambda \defeq [\mu_{1:d}, \sigma_{1:d}^2]}$.
  Such a multivariate distribution where all variables are independent is called a \midx{mean-field distribution}.
  Importantly, this family of distributions is characterized by only $2d$ parameters!
\end{ex}

Note that \cref{fig:variational_families} is a generalization of the canonical distinction between estimation error and approximation error from \cref{fig:estimation_approximation_error}, only that here, we operate in the space of distributions over functions as opposed the space of functions.
A common notion of distance between two distributions $q$ and $p$ is the Kullback-Leibler divergence $\KL{q}{p}$ which we will define in the next section.
Using this notion of distance, we need to solve the following optimization problem: \begin{align}
  \qs \defeq \argmin_{q \in \spQ} \KL{q}{p} = \argmin_{\vlambda \in \Lambda} \KL{q_\vlambda}{p}. \label{eq:var_optimization}
\end{align}

In \cref{sec:approximate_inference:information_theory}, we introduce information theory and the Kullback-Leibler divergence.
Then, in \cref{sec:approximate_inference:variational_inference:elbo}, we discuss how the optimization problem of \cref{eq:var_optimization} can be solved efficiently.

\section{Information Theoretic Aspects of Uncertainty}\label{sec:approximate_inference:information_theory}

One of our main objectives throughout this manuscript is to capture the ``uncertainty'' about events $A$ in an appropriate probability space.
One very natural measure of uncertainty is their probability, $\Pr{A}$.
In this section, we will introduce an alternative measure of uncertainty, namely the so-called ``surprise'' about the event $A$.

\subsection{Surprise}

The \midx{surprise} about an event with probability $u$ is defined as \begin{align}
  \S{u} \defeq - \log u.
\end{align}
Observe that the surprise is a function from $\Rzero$ to $\R$, where we let~${\S{0} \equiv \infty}$.
Moreover, for a discrete random variable $X$, we have that $p(x) \leq 1$, and hence, $\S{p(x)} \geq 0$.
But why is it reasonable to measure surprise by $- \log u$?

\begin{marginfigure}[-4\baselineskip]
  \begin{center}
    \import{./plots/output/}{surprise.pgf}
  \end{center}

  \caption{Surprise $\S{u}$ associated with an event of probability $u$.}
\end{marginfigure}

Remarkably, it can be shown that the following natural axiomatic characterization leads to exactly this definition of surprise.

\begin{thm}[Axiomatic characterization of surprise]
  The axioms \begin{enumerate}
    \item $\S{u} > \S{v} \implies u < v$ (anti-monotonicity) \margintag{we are more surprised by unlikely events}
    \item $\fnS$ is continuous, \margintag{no jumps in surprise for infinitesimal changes of probability} \\[-5pt]
    \item $\S{uv} = \S{u} + \S{v}$ for independent events, \margintag{the surprise of independent events is additive}
  \end{enumerate} characterize $\fnS$ up to a positive constant factor.
\end{thm}
\begin{proof}
  Observe that the third condition looks similar to the product rule of logarithms: $\log(uv) = \log v + \log v$.
  We can formalize this intuition by remembering Cauchy's functional equation, $f(x + y) = f(x) + f(y)$, which has the unique family of solutions $\{f : x \mapsto c x : c \in \R\}$ if $f$ is required to be continuous.
  Such a solution is called an ``additive function''.
  Consider the function $g(x) \defeq f(e^x)$. Then, $g$ is additive if and only if \begin{align*}
    f(e^x e^y) = f(e^{x+y}) = g(x + y) = g(x) + g(y) = f(e^x) + f(e^y).
  \end{align*}
  This is precisely the third axiom of surprise for $f = \fnS$ and $e^x = u$!
  Hence, the second and third axioms of surprise imply that $g$ must be additive and that $g(x) = \S{e^x} = c x$ for any $c \in \R$.
  If we replace $e^x$ by $u$, we obtain $\S{u} = c \log u$.
  The first axiom of surprise implies that $c < 0$, and thus, $\S{u} = - c' \log u$ for any $c' > 0$.
\end{proof}

Importantly, surprise offers a different perspective on uncertainty as opposed to probability: the uncertainty about an event can either be interpreted in terms of its probability or in terms of its surprise, and the two ``spaces of uncertainty'' are related by a log-transform.
This relationship is illustrated in \cref{fig:surprise_space}.
Information theory is the study of uncertainty in terms of surprise.

\begin{marginfigure}
  \incfig{surprise_space}
  \caption{Illustration of the probability space and the corresponding ``surprise space''.\looseness=-1}\label{fig:surprise_space}
\end{marginfigure}

Throughout this manuscript we will see many examples where modeling uncertainty in terms of surprise (i.e., the information-theoretic interpretation of uncertainty) is useful.
One example where we have already encountered the ``surprise space'' was in the context of likelihood maximization (cf. \cref{sec:fundamentals:parameter_estimation:mle}) where we used that the log-transform linearizes products of probabilities.
We will see later in \cref{sec:approximate_inference:mcmc} that in many cases the surprise $\S{p(x)}$ can also be interpreted as a ``cost'' or ``energy'' associated with the state $x$.

\subsection{Entropy}

\begin{marginfigure}[5\baselineskip]
  \begin{center}
    \import{./plots/output/}{entropy.pgf}
  \end{center}

  \caption{Entropy of a Bernoulli experiment with success probability $p$.}
\end{marginfigure}

The \midx{entropy}[idxpagebf] of a distribution $p$ is the average surprise about samples from $p$.
In this way, entropy is a notion of uncertainty associated with the distribution $p$: if the entropy of $p$ is large, we are more uncertain about $x \sim p$ than if the entropy of $p$ were low.
Formally, \begin{align}
  \H{p} \defeq \E[x \sim p]{\S{p(x)}} = \E[x \sim p]{- \log p(x)}. \label{eq:entropy}
\end{align}
When $\rX \sim p$ is a random vector distributed according to $p$, we write $\H{\rX} \defeq \H{p}$.
Observe that by definition, if $p$ is discrete then $\H{p} \geq 0$ as $p(x) \leq 1 \; (\forall x)$.\footnote{The entropy of a continuous distribution can be negative.
For example, \begin{align*}
  \H{\Unif{[a,b]}} &= -\int \frac{1}{b-a} \log\frac{1}{b-a} \,dx \\
  &= \log(b-a)
\end{align*} which is negative if $b-a < 1$.}
For discrete distributions it is common to use the logarithm with base $2$ rather than the natural logarithm:\footnote{Recall that $\log_2 x = \frac{\log x}{\log 2}$, that is, logarithms to a different base only differ by a constant factor.} \begin{subequations}\begin{align}
  \H{p} &= - \sum_{x} p(x) \log_2 p(x) &&\text{(if $p$ is discrete)}, \label{eq:entropy_discrete} \\
  \H{p} &= - \int p(\vx) \log p(\vx) \,d\vx  &&\text{(if $p$ is continuous)}. \label{eq:entropy_cont}
\end{align}\end{subequations}

Let us briefly recall Jensen's inequality, which is a useful tool when working with expectations of convex functions such as entropy:\footnote{The surprise $\S{u}$ is convex in $u$.}

\begin{fct}[Jensen's Inequality]\pidx{Jensen's inequality}\exerciserefmark{jensen}[1]
  Given a random variable $X$ and a convex function $g : \R \to \R$, we have \begin{align}
    g(\E{X}) \leq \E{g(X)}. \label{eq:jensen}
  \end{align}
\end{fct}

\begin{marginfigure}[3\baselineskip]
  \begin{center}
    \import{./plots/output/}{jensen.pgf}
  \end{center}

  \caption{An illustration of Jensen's inequality.
  Due to the convexity of $g$, we have that $g$ evaluated at $\E{X}$ will always be below the average of evaluations of $g$.}
\end{marginfigure}

\begin{ex}{Examples of entropy}{}
  \begin{itemize}
    \item \emph{Fair Coin} \quad $\H{\Bern{0.5}} = -2 (0.5 \log_2 0.5) = 1.$
    \item \emph{Unfair Coin} \quad \begin{align*}
      \H{\Bern{0.1}} = -0.1 \log_2 0.1 - 0.9 \log_2 0.9 \approx0.469.
    \end{align*}
    \item \emph{Uniform Distribution} \quad \begin{align*}
      \H{\Unif{\{1, \dots, n\}}} = - \sum_{i=1}^n \frac{1}{n} \log_2 \frac{1}{n} = \log_2 n.
    \end{align*}
    The uniform distribution has the maximum entropy among all discrete distributions supported on ${\{1, \dots, n\}}$ \exerciserefmark{jensen}[2].
    Note that a fair coin corresponds to a uniform distribution with ${n = 2}$.
    Also observe that $\log_2 n$ corresponds to the number of bits required to encode the outcome of the experiment.
  \end{itemize}

  In general, the entropy $\H{p}$ of a discrete distribution $p$ can be interpreted as the average number of bits required to encode a sample $x \sim p$, or in other words, the average ``information'' carried by a sample $x$.
\end{ex}

\begin{ex}{Entropy of a Gaussian}{}
  Let us derive the entropy of a univariate Gaussian.
  Recall the PDF, \begin{align*}
    \N[x]{\mu}{\sigma^2} = \frac{1}{Z} \exp\parentheses*{-\frac{(x-\mu)^2}{2 \sigma^2}}
  \end{align*} where ${Z = \sqrt{2 \pi \sigma^2}}$.
  Using the definition of entropy \eqref{eq:entropy_cont}, we obtain, \begin{align}
    \H{\N{\mu}{\sigma^2}} \nonumber &= \begin{multlined}[t]
      - \int \frac{1}{Z} \exp\parentheses*{-\frac{(x-\mu)^2}{2 \sigma^2}} \\ \cdot \log\parentheses*{\frac{1}{Z} \exp\parentheses*{-\frac{(x-\mu)^2}{2 \sigma^2}}} \,dx
    \end{multlined} \nonumber \\
    &= \begin{multlined}[t]
    \log Z \underbrace{\int \frac{1}{Z} \exp\parentheses*{-\frac{(x-\mu)^2}{2 \sigma^2}} \,dx}_{1} \\
    + \int \frac{1}{Z} \exp\parentheses*{-\frac{(x-\mu)^2}{2 \sigma^2}} \frac{(x-\mu)^2}{2 \sigma^2} \,dx
    \end{multlined} \nonumber \\
    &= \log Z + \frac{1}{2 \sigma^2} \E{(x - \mu)^2} \margintag{using LOTUS \eqref{eq:lotus}} \nonumber \\
    &= \log(\sigma \sqrt{2 \pi}) + \frac{1}{2} \margintag{using $\E{(x - \mu)^2} = \Var{x} = \sigma^2$ \eqref{eq:variance}} \nonumber \\
    &= \log(\sigma \sqrt{2 \pi e}). \margintag{using $\log \sqrt{e} = \nicefrac{1}{2}$} \label{eq:entropy_gaussian_univ}
  \end{align}

  In general, the entropy of a Gaussian is \begin{align}
    \H{\N{\vmu}{\mSigma}} = \frac{1}{2} \log \det{2 \pi e \mSigma} = \frac{1}{2} \log \parentheses*{(2 \pi e)^d \det{\mSigma}}. \label{eq:entropy_gaussian}
  \end{align}
  Note that the entropy is a function of the determinant of the covariance matrix $\mSigma$.
  In general, there are various ways of ``scalarizing'' the notion of uncertainty for a multivariate distribution.
  The determinant of $\mSigma$ measures the volume of the credible sets around the mean~$\vmu$, and is also called the \midx{generalized variance}.
  Next to entropy and generalized variance (which are closely related for Gaussians), a common scalarization is the trace of~$\mSigma$, which is also called the \midx{total variance}.
\end{ex}

\subsection{Cross-Entropy}

How can we use entropy to measure our average surprise when assuming the data follows some distribution $q$ but in reality the data follows a different distribution $p$?

\begin{defn}[Cross-entropy]
  The \midx{cross-entropy} of a distribution $q$ relative to the distribution $p$ is \begin{align}
    \crH{p}{q} \defeq \E[x \sim p]{\S{q(x)}} = \E[x \sim p]{- \log q(x)}. \label{eq:cross_entropy}
  \end{align}
\end{defn}

Cross-entropy can also be expressed in terms of the KL-divergence (cf. \cref{sec:vi:kl}) $\KL{p}{q}$ which measures how ``different'' the distribution $q$ is from a reference distribution $p$, \begin{align}
  \crH{p}{q} = \H{p} + \KL{p}{q} \geq \H{p}. \margintag{$\KL{p}{q} \geq 0$ is shown in \cref{exercise:gibbs_ineq}} \label{eq:cross_entropy_decomp}
\end{align}
Quite intuitively, the average surprise in samples from $p$ with respect to the distribution $q$ is given by the inherent uncertainty in $p$ and the additional surprise that is due to us assuming the wrong data distribution $q$.
The ``closer'' $q$ is to the true data distribution $p$, the smaller is the additional average surprise.

\subsection{Kullback-Leibler Divergence}\label{sec:vi:kl}

As mentioned, the Kullback-Leibler divergence is a (non-metric) measure of distance between distributions.
It is defined as follows.

\begin{defn}[Kullback-Leibler divergence, KL-divergence]\pidx{Kullback-Leibler divergence}
  Given two distributions $p$ and $q$, the \emph{Kullback-Leibler divergence} (or \midx{relative entropy}) of $q$ with respect to $p$, \begin{align}
    \KL{p}{q} &\defeq \crH{p}{q} - \H{p} \label{eq:kl} \\
    &= \E[\vtheta \sim p]{\S{q(\vtheta)} - \S{p(\vtheta)}} \\
    &= \E[\vtheta \sim p]{\log \frac{p(\vtheta)}{q(\vtheta)}},
  \end{align} measures how different $q$ is from a reference distribution $p$.
\end{defn}
In words, $\KL{p}{q}$ measures the \emph{additional} expected surprise when observing samples from $p$ that is due to assuming the (wrong) distribution $q$ and which not inherent in the distribution $p$ already.\footnote[][-\baselineskip]{The KL-divergence only captures the additional expected surprise since the surprise inherent in $p$ (as measured by $\H{p}$) is subtracted.}

The KL-divergence has the following properties: \begin{itemize}
  \item $\KL{p}{q} \geq 0$ for any distributions $p$ and $q$ \exerciserefmark{gibbs_ineq}[1],
  \item $\KL{p}{q} = 0$ if and only if $p = q$ almost surely \exerciserefmark{gibbs_ineq}[2], and
  \item there exist distributions $p$ and $q$ such that $\KL{p}{q} \neq \KL{q}{p}$.
\end{itemize}

The KL-divergence can simply be understood as a shifted version of cross-entropy, which is zero if we consider the divergence between two identical distributions.

We will briefly look at another interpretation for how KL-divergence measures ``distance'' between distributions.
Suppose we are presented with a sequence $\vtheta_1, \dots, \vtheta_n$ of independent samples from either a distribution $p$ or a distribution $q$, both of which are known.
Which of $p$ or $q$ was used to generate the data is, however, unknown to us, and we would like to find out.
A natural approach is to choose the distribution whose data likelihood is larger. That is, we choose $p$ if $p(\vtheta_{1:n}) > q(\vtheta_{1:n})$ and vice versa. Assuming that the samples are independent and rewriting the inequality slightly, we choose $p$ if \begin{align}
  \prod_{i=1}^n \frac{p(\vtheta_i)}{q(\vtheta_i)} > 1, \quad\text{or equivalently if}\quad
  \sum_{i=1}^n \log \frac{p(\vtheta_i)}{q(\vtheta_i)} > 0. \label{eq:kl_decision_criterion} \margintag{taking the logarithm}
\end{align}
Assume without loss of generality that $\vtheta_i \sim p$. Then, using the law of large numbers \eqref{eq:slln}, \begin{align}
  \frac{1}{n} \sum_{i=1}^n \log \frac{p(\vtheta_i)}{q(\vtheta_i)} \almostsurely \E[\vtheta \sim p]{\log \frac{p(\vtheta)}{q(\vtheta)}} = \KL{p}{q}
\end{align} as $n \to \infty$.
Plugging this into our decision criterion from \cref{eq:kl_decision_criterion}, we find that \begin{align}
  \E{\sum_{i=1}^n \log \frac{p(\vtheta_i)}{q(\vtheta_i)}} = n \KL{p}{q}.
\end{align}
In this way, $\KL{p}{q}$ measures the observed ``distance'' between $p$ and~$q$.
Recall that assuming $p \neq q$ we have that $\KL{p}{q} > 0$ with probability $1$, and therefore we correctly choose $p$ with probability $1$ as~${n \to \infty}$.
Moreover, Hoeffding's inequality \eqref{eq:hoeffdings_inequality} can be used to determine ``how quickly'' samples converge to this limit, that is, how quickly we can distinguish between $p$ and $q$.

\begin{ex}{KL-divergence of Bernoulli random variables}{}
  Suppose we are given two Bernoulli distributions $\Bern{p}$ and $\Bern{q}$.
  Then, their KL-divergence is \begin{align}
    \KL{\Bern{p}}{\Bern{q}} &= \sum_{x \in \{0,1\}} \Bern[x]{p} \log \frac{\Bern[x]{p}}{\Bern[x]{q}} \nonumber\\
    &= p \log \frac{p}{q} + (1-p) \log \frac{(1-p)}{(1-q)}.
  \end{align}
  Observe that $\KL{\Bern{p}}{\Bern{q}} = 0$ if and only if $p = q$.
\end{ex}

\begin{ex}{KL-divergence of Gaussians}{}
  Suppose we are given two Gaussian distributions ${p \defeq \N{\vmu_p}{\mSigma_p}}$ and ${q \defeq \N{\vmu_q}{\mSigma_q}}$ with dimension $d$.
  The KL-divergence of $p$ and~$q$ is given by \exerciserefmark{kl_div_of_gaussians} \begin{align}
    \begin{multlined}[t]
    \KL{p}{q} = \frac{1}{2} \left(\mathrm{tr}(\inv{\mSigma_q} \mSigma_p) + \transpose{(\vmu_p - \vmu_q)} \inv{\mSigma_q} (\vmu_p - \vmu_q) \phantom{\frac12}\right. \\ \left. - d + \log \frac{\det{\mSigma_q}}{\det{\mSigma_p}}\right).
    \end{multlined} \label{eq:kl_gaussian}
  \end{align}

  For independent Gaussians with unit variance, $\mSigma_p = \mSigma_q = \mI$, the expression simplifies to the squared Euclidean distance, \begin{align}
    \KL{p}{q} = \frac{1}{2} \norm{\vmu_q - \vmu_p}_2^2.
  \end{align}

  If we approximate independent Gaussians with variances $\sigma_i^2$, \begin{align*}
    p \defeq \N{\vmu}{\diag{\sigma_1^2, \dots, \sigma_d^2}},
  \end{align*} by a standard normal distribution, ${q \defeq \SN}$, the expression simplifies to \begin{align}
    \KL{p}{q} = \frac{1}{2} \sum_{i=1}^d (\sigma_i^2 + \mu_i^2 - 1 - \log \sigma_i^2). \label{eq:kl_brl}
  \end{align}
  Here, the term $\mu_i^2$ penalizes a large mean of $p$, the term $\sigma_i^2$ penalizes a large variance of $p$, and the term $- \log \sigma_i^2$ penalizes a small variance of $p$.
  As expected, $\KL{p}{q}$ is proportional to the amount of information we lose by approximating $p$ with the simpler distribution $q$.
\end{ex}

\subsection{Forward and Reverse KL-divergence}\label{sec:vi:kl:forward_reverse}

\begin{marginfigure}[-15\baselineskip]
  \begin{center}
    \import{./plots/output/}{kl_divergence_1.pgf}
  \end{center}

  \begin{center}
    \import{./plots/output/}{kl_divergence_2.pgf}
  \end{center}

  \caption{Comparison of the \textbf{\r{forward}} KL-divergence $\r{\qs_1}$ and the \textbf{\b{reverse}} KL-divergence $\b{\qs_2}$ when used to approximate the \textbf{true posterior} $p$.
  The first plot shows the PDFs in a one-dimensional feature space where $p$ is a mixture of two univariate Gaussians.
  The second plot shows contour lines of the PDFs in a two-dimensional feature space where the non-diagonal Gaussian $p$ is approximated by diagonal Gaussians $\qs_1$ and $\qs_2$.
  It can be seen that $\qs_1$ selects the variance and $\qs_2$ selects the mode of $p$.
  The approximation $\qs_1$ is more conservative than the (overconfident) approximation~$\qs_2$.}\label{fig:forward_reverse_kl}
\end{marginfigure}

$\KL{p}{q}$ is also called the \midx<forward>{forward KL-divergence} (or \midx<inclusive>{inclusive KL-divergence}) KL-divergence. In contrast, $\KL{q}{p}$ is called the \midx<reverse>{reverse KL-divergence} (or \midx<exclusive>{exclusive KL-divergence}) KL-divergence.
\Cref{fig:forward_reverse_kl} shows the approximations of a general Gaussian obtained when $\spQ$ is the family of diagonal (independent) Gaussians.
Thereby, \begin{align*}
  \qs_1 \defeq \argmin_{q \in \spQ} \KL{p}{q} \quad\text{and}\quad \qs_2 \defeq \argmin_{q \in \spQ} \KL{q}{p}.
\end{align*} $\qs_1$ is the result when using the forward KL-divergence and $\qs_2$ is the result when using reverse KL-divergence.
It can be seen that the reverse KL-divergence tends to greedily select the mode and underestimating the variance which, in this case, leads to an overconfident prediction.
The forward KL-divergence, in contrast, is more conservative and yields what one could consider the ``desired'' approximation.

Recall that in the blueprint of variational inference \eqref{eq:var_optimization} we used the reverse KL-divergence.
This is for computational reasons.
Observe that to approximate the KL-divergence $\KL{p}{q}$ using Monte Carlo sampling, we would need to obtain samples from $p$ yet $p$ is the intractable posterior distribution which we were trying to approximate in the first place.
Crucially, observe that if the true posterior $p(\cdot \mid \vx_{1:n}, y_{1:n})$ is in the variational family $\spQ$, then \begin{align}\begin{split}
  \argmin_{q \in \spQ} \KL{q}{p(\cdot \mid \vx_{1:n}, y_{1:n})} \eqalmostsurely p(\cdot \mid \vx_{1:n}, y_{1:n}), \margintag{as $\min_{q \in \spQ} \KL{q}{p(\cdot \mid \vx_{1:n}, y_{1:n})} \eqalmostsurely 0$}
\end{split}\end{align} so minimizing reverse-KL still recovers the true posterior almost surely.

\begin{rmk}{Greediness of reverse-KL}{}
  As in the previous example, consider the independent Gaussians \begin{align*}
    p \defeq \N{\vmu}{\diag[i\in[d]]{\sigma_i^2}},
  \end{align*} which we seek to approximate by a standard normal distribution ${q \defeq \SN}$.
  Using \eqref{eq:kl_gaussian}, we obtain for the reverse KL-divergence, \begin{align}
    \KL{q}{p} &= \begin{multlined}[t]
      \frac{1}{2} \left(\tr{\diag{\sigma_i^{-2}}} + \transpose{\vmu}\diag{\sigma_i^{-2}}\vmu - d \right. \\ \left. + \log\det{\diag{\sigma_i^2}}\right)
    \end{multlined} \nonumber \\
    &= \frac{1}{2} \sum_{i=1}^d \parentheses*{\sigma_i^{-2} + \frac{\mu_i^2}{\sigma_i^2} - 1 + \log \sigma_i^2}.
  \end{align}
  Here, $\sigma_i^{-2}$ penalizes small variance, $\nicefrac{\mu_i^2}{\sigma_i^2}$ penalizes a large mean, and $\log \sigma_i^2$ penalizes large variance.
  Compare this to the expression for the forward KL-divergence $\KL{p}{q}$ that we have seen in \cref{eq:kl_brl}. In particular, observe that reverse-KL penalizes large variance less strongly than forward-KL.

  Note, however, that reverse-KL is not greedy in the same sense as Laplace approximation, as it does still take the variance into account and does not \emph{purely} match the mode of $p$.
\end{rmk}

\subsection{Interlude: Minimizing Forward KL-Divergence}\label{sec:vi:kl:forward}

Before completing the blueprint of variational inference in \cref{sec:approximate_inference:variational_inference:elbo} by showing how \emph{reverse-KL} can be efficiently minimized, we will digress briefly and relate minimizing \emph{forward-KL} to two other well-known inference algorithms.
This discussion will deepen our understanding of the KL-divergence and its role in probabilistic inference, but feel free to skip ahead to \cref{sec:approximate_inference:variational_inference:elbo} if you are eager to complete the blueprint.

\paragraph{Minimizing forward-KL as maximum likelihood estimation:}
First, we observe that minimizing the forward KL-divergence is equivalent to maximum likelihood estimation on an infinitely large sample size.
The classical application of this result is in the setting where $p(\vx)$ is a generative model, and we aim to estimate its density with the parameterized model $q_\vlambda$.

\begin{lem}[Forward KL-divergence as MLE]
  Given some generative model $p(\vx)$ and a likelihood $q_\vlambda(\vx) = q(\vx \mid \vlambda)$ (that we use to approximate the true data distribution), we have \begin{align}
    \argmin_{\vlambda \in \Lambda} \KL{p}{q_\vlambda} \eqalmostsurely \argmax_{\vlambda \in \Lambda} \lim_{n\to\infty} \frac{1}{n} \sum_{i=1}^n \log q(\vx_i \mid \vlambda), \label{eq:forward_kl_as_mle}
  \end{align} where $\vx_i \iid p$ are independent samples from the true data distribution.
\end{lem}
\begin{proof}
\begin{align*}
  \KL{p}{q_\vlambda} &= \crH{p}{q_\vlambda} - \H{p} \margintag{using the definition of KL-divergence \eqref{eq:kl}} \\[6pt]
  &= \E[(\vx, y) \sim p]{- \log q(\vx \mid \vlambda)} + \const \margintag{dropping $\H{p}$ and using the definition of cross-entropy \eqref{eq:cross_entropy}} \\
  &\eqalmostsurely - \lim_{n\to\infty} \frac{1}{n} \sum_{i=1}^n \log q(\vx_i \mid \vlambda) + \const \margintag{using Monte Carlo sampling, i.e., the law of large numbers \eqref{eq:slln}}
\end{align*} where $\vx_i \iid p$ are independent samples.
\end{proof}

This tells us that \emph{any} maximum likelihood estimate $q_\vlambda$ minimizes the forward KL-divergence to the empirical data distribution.
Note that here, we aim to learn model parameters $\vlambda$ for estimating the probability of $\vx$, whereas in the setting of variational probabilistic inference, we want to learn parameters $\vlambda$ of a distribution over $\vtheta$ and $\vtheta$ parameterizes a \emph{distribution over $\vx$}.
This interpretation is therefore not immediately useful for probabilistic inference (i.e., in the setting where $p$ is a posterior distribution over model parameters $\vtheta$) as a maximum likelihood estimate requires i.i.d. samples from $p$ which we cannot easily obtain in this case.\footnote{It is possible to obtain ``approximate'' samples using Markov chain Monte Carlo (MCMC) methods which we discuss in \cref{sec:approximate_inference:mcmc}.}

\begin{ex}{Minimizing cross-entropy}{}
  Minimizing the KL-divergence between $p$ and $q_{\vlambda}$ is equivalent to minimizing cross-entropy since $\KL{p}{q_\vlambda} = \crH{p}{q_\vlambda} - \H{p}$ and $\H{p}$ is constant with respect to $p$.

  Lets consider an example in a binary classification problem with the label $y \in \{0, 1\}$ and predicted class probability $\hat{y} \in [0, 1]$ for some fixed input.
  It is natural to use cross-entropy as a measure of dissimilarity between $y$ and $\hat{y}$, \begin{align}\begin{split}
    \ell_{\mathrm{bce}}(\hat{y}; y) &\defeq \crH{\Bern{y}}{\Bern{\hat{y}}} \\
    &= -\sum_{x \in \{0, 1\}} \Bern[x]{y} \log \Bern[x]{\hat{y}} \\
    &= -y \log \hat{y} - (1-y) \log (1-\hat{y}).
  \end{split}\label{eq:bce_loss}\end{align}
  This loss function is also known as the \midx{binary cross-entropy loss} and we will discuss it in more detail in \cref{sec:bnn:mle} in the context of neural networks.
\end{ex}

\paragraph{Minimizing forward-KL as moment matching:}
Now to a second interpretation of minimizing forward-KL.
\midx<Moment matching>{moment matching}[idxpagebf] (also known as the \midx{method of moments}) is a technique for approximating an unknown distribution $p$ with a parameterized distribution $q_\vlambda$ where $\vlambda$ is chosen such that $q_\vlambda$ matches the (estimated) moments of $p$.
For example, given the estimates $\va$ and $\mB$ of the first and second moment of $p$,\safefootnote{These estimates are computed using the samples from $p$. For example, using a sample mean and a sample variance to compute the estimates of the first and second moment.} and if $q_\vlambda$ is a Gaussian with parameters $\vlambda = \{\vmu, \mSigma\}$, then moment matching chooses $\vlambda$ as the solution to \begin{align*}
  \E[p]{\vtheta} \approx \va &\overset{!}{=} \vmu = \E[q_\vlambda]{\vtheta} \\
  \E[p]{\vtheta \transpose{\vtheta}} \approx \mB &\overset{!}{=} \mSigma + \vmu \transpose{\vmu} = \E[q_\vlambda]{\vtheta \transpose{\vtheta}}. \margintag{using the definition of variance \eqref{eq:variance2}}
\end{align*}
In general the number of moments to be matched (i.e., the number of equations) is adjusted such that it is equal to the number of parameters to be estimated.
We will see now that the ``matching'' of moments is also ensured when $q_\vlambda$ is obtained by minimizing the forward KL-divergence within the family of Gaussians.

The Gaussian PDF can be expressed as \begin{align}
  \N[\vtheta]{\vmu}{\mSigma} &= \frac{1}{Z(\vlambda)} \exp(\transpose{\vlambda} \vs(\vtheta)) \quad\text{where} \label{eq:exponential_family_of_distributions} \\
  \vlambda &\defeq \begin{bmatrix}
    \inv{\mSigma}\vmu \\
    \vecop{\inv{\mSigma}}
  \end{bmatrix} \label{eq:gaussian_nat_params} \\
  \vs(\vtheta) &\defeq \begin{bmatrix}
    \vtheta \\
    \vecop{-\frac{1}{2}\vtheta\transpose{\vtheta}}
  \end{bmatrix} \label{eq:gaussian_suff_stats}
\end{align} and $Z(\vlambda) \defeq \int \exp(\transpose{\vlambda} \vs(\vtheta)) \,d\vtheta$, and we will confirm this in just a moment.\safefootnote{Given a matrix $\mA \in \R^{n \times m}$, we use \begin{align*}
  \vecop{\mA} \in \R^{n \cdot m}
\end{align*} to denote the row-by-row concatenation of $\mA$ yielding a vector of length $n \cdot m$.}
The family of distributions with densities of the form \eqref{eq:exponential_family_of_distributions} --- with an additional scaling constant $h(\vtheta)$ which is often $1$ --- is called the \midx{exponential family} of distributions.
Here, $\vs(\vtheta)$ are the \midx<sufficient statistics>{sufficient statistic}, $\vlambda$ are called the \midx{natural parameters}, and $Z(\vlambda)$ is the normalizing constant.
In this context, $Z(\vlambda)$ is often called the \midx{partition function}.

To see that the Gaussian is indeed part of the exponential family as promised in \cref{eq:gaussian_nat_params,eq:gaussian_suff_stats}, consider \begin{align*}
  \N[\vtheta]{\vmu}{\mSigma} &\propto \exp\parentheses*{-\frac{1}{2}\transpose{(\vtheta - \vmu)}\inv{\mSigma}(\vtheta - \vmu)} \\
  &\propto \exp\parentheses*{\tr{-\frac{1}{2}\transpose{\vtheta}\inv{\mSigma}\vtheta} + \transpose{\vtheta}\inv{\mSigma}\vmu} \margintag{expanding the inner product and using that $\tr{x} = x$ for all $x \in \R$} \\
  &= \exp\parentheses*{\tr{-\frac{1}{2}\vtheta\transpose{\vtheta}\inv{\mSigma}} + \transpose{\vtheta}\inv{\mSigma}\vmu} \margintag{using that the trace is invariant under cyclic permutations} \\
  &= \exp\parentheses*{\transpose{\vecop{-\tfrac{1}{2}\vtheta\transpose{\vtheta}}}\vecop{\inv{\mSigma}} + \transpose{\vtheta}\inv{\mSigma}\vmu}. \margintag{using $\tr{\mA\mB} = \transpose{\vecop{\mA}}\vecop{\mB}$ for any $\mA, \mB \in \R^{n \times n}$}
\end{align*}
This allows us to express the forward KL-divergence as \begin{align*}
  \KL{p}{q_\vlambda} &= \int p(\vtheta) \log\frac{p(\vtheta)}{q_\vlambda(\vtheta)}\,d\vtheta \\
  &= -\int p(\vtheta) \cdot \transpose{\vlambda} \vs(\vtheta) \,d\vtheta + \log Z(\vlambda) + \const. \margintag{using that $\int p(\vtheta) \log p(\vtheta) \,d \vtheta$ is constant}
\end{align*} Differentiating with respect to the natural parameters $\vlambda$ gives \begin{align*}
  \grad_\vlambda \KL{p}{q_\vlambda} &= -\int p(\vtheta) \vs(\vtheta) \,d\vtheta + \frac{1}{Z(\vlambda)} \int \vs(\vtheta) \exp(\transpose{\vlambda} \vs(\vtheta)) \,d\vtheta \\
  &= -\E[\vtheta \sim p]{\vs(\vtheta)} + \E[\vtheta \sim q_\vlambda]{\vs(\vtheta)}.
\end{align*} Hence, for any minimizer of $\KL{p}{q_\vlambda}$, we have that the sufficient statistics under $p$ and $q_\vlambda$ match: \begin{align}
  \E[p]{\vs(\vtheta)} = \E[q_\vlambda]{\vs(\vtheta)}.
\end{align}
Therefore, in the Gaussian case, \begin{align*}
  \E[p]{\vtheta} = \E[q_\vlambda]{\vtheta} \quad\text{and}\quad \E[p]{-\tfrac{1}{2}\vtheta\transpose{\vtheta}} = \E[q_\vlambda]{-\tfrac{1}{2}\vtheta\transpose{\vtheta}},
\end{align*} implying that \begin{align}
  \E[p]{\vtheta} = \vmu \quad\text{and}\quad \Var[p]{\vtheta} = \E[p]{\vtheta\transpose{\vtheta}} - \E[p]{\vtheta}\cdot\transpose{\E[p]{\vtheta}} = \mSigma \margintag{using \cref{eq:variance2}}
\end{align} where $\vmu$ and $\mSigma$ are the mean and variance of the approximation $q_\vlambda$, respectively.
That is, a Gaussian $q_\vlambda$ minimizing $\KL{p}{q_\vlambda}$ has the same first and second moment as $p$.
Combining this insight with our observation from \cref{eq:forward_kl_as_mle} that minimizing forward-KL is equivalent to maximum likelihood estimation, we see that if we use MLE to fit a Gaussian to given data, this Gaussian will eventually match the first and second moments of the data distribution.

\section{Evidence Lower Bound}\label{sec:approximate_inference:variational_inference:elbo}

Let us return to the blueprint of variational inference from \cref{sec:vi:blueprint}.
To complete this blueprint, it remains to show that the reverse KL-divergence can be minimized efficiently.
We have \begin{align*}
  \KL{q}{p(\cdot \mid \vx_{1:n}, y_{1:n})} &= \E[\vtheta \sim q]{\log \frac{q(\vtheta)}{p(\vtheta \mid \vx_{1:n}, y_{1:n})}} \margintag{using the definition of the KL-divergence \eqref{eq:kl}} \\
  &= \E[\vtheta \sim q]{\log \frac{p(y_{1:n} \mid \vx_{1:n}) q(\vtheta)}{p(y_{1:n}, \vtheta \mid \vx_{1:n})}} \margintag{using the definition of conditional probability \eqref{eq:cond_prob}} \\
  &= \begin{multlined}[t]
    \log p(y_{1:n} \mid \vx_{1:n}) \\ \underbrace{- \E[\vtheta \sim q]{\log p(y_{1:n}, \vtheta \mid \vx_{1:n})} - \H{q}}_{- L(q, p; \spD_n)}
  \end{multlined} \margintag{using linearity of expectation \eqref{eq:linearity_expectation}}
\end{align*} where $L(q, p; \spD_n)$ is called the \emph{evidence lower bound} (ELBO) given the data $\spD_n = \{(\vx_i, y_i)\}_{i=1}^n$.
This gives the relationship \begin{thmb}\begin{align}
  L(q, p; \spD_n) = \underbrace{\log p(y_{1:n} \mid \vx_{1:n})}_{\const} -\ \KL{q}{p(\cdot \mid \vx_{1:n}, y_{1:n})}. \label{eq:elbo_reverse_kl_relationship}
\end{align}\end{thmb}
Thus, maximizing the ELBO coincides with minimizing reverse-KL.

Maximizing the ELBO, \begin{align}
  L(q, p; \spD_n) &\defeq \E[\vtheta \sim q]{\log p(y_{1:n}, \vtheta \mid \vx_{1:n})} + \H{q}, \label{eq:elbo}
\end{align} selects $q$ that has large joint likelihood $p(y_{1:n}, \vtheta \mid \vx_{1:n})$ and large entropy $\H{q}$.
The ELBO can also be expressed in various other forms: \vspace{-\baselineskip}\begin{subequations}\begin{align}
  L(q, p; \spD_n) &= \E[\vtheta \sim q]{\log p(y_{1:n}, \vtheta \mid \vx_{1:n}) - \log q(\vtheta)} \margintag{using the definition of entropy \eqref{eq:entropy}} \\
  &= \E[\vtheta \sim q]{\log p(y_{1:n} \mid \vx_{1:n}, \vtheta) + \log p(\vtheta) - \log q(\vtheta)} \margintag{using the product rule \eqref{eq:product_rule}} \label{eq:elbo_expanded} \\
  &= \underbrace{\E[\vtheta \sim q]{\log p(y_{1:n} \mid \vx_{1:n}, \vtheta)}}_{\text{log-likelihood}}\ \underbrace{-\ \KL{q}{p(\cdot)}}_{\text{proximity to prior}}. \margintag{using the definition of KL-divergence \eqref{eq:kl}} \label{eq:elbo2}
\end{align}\end{subequations} where we denote by $p(\cdot)$ the prior distribution.
\Cref{eq:elbo2} highlights the connection to probabilistic inference, namely that maximizing the ELBO selects a variational distribution $q$ that is close to the prior distribution $p(\cdot)$ while also maximizing the \emph{average} likelihood of the data $p(y_{1:n} \mid \vx_{1:n}, \vtheta)$ for $\vtheta \sim q$.
This is in contrast to maximum a posteriori estimation, which picks a \emph{single} model $\vtheta$ that maximizes the likelihood and proximity to the prior.
As an example, let us look at the case where the prior is noninformative, i.e., $p(\cdot) \propto 1$.
In this case, the ELBO simplifies to $\E[\vtheta \sim q]{\log p(y_{1:n}, \mid \vx_{1:n}, \vtheta)} + \H{q} + \const$.
That is, maximizing the ELBO maximizes the average likelihood of the data under the variational distribution $q$ while regularizing $q$ to have high entropy.
Why is it reasonable to maximize the entropy of $q$?
Consider two distributions $q_1$ and $q_2$ under which the data is ``equally'' likely and which are ``equally'' close to the prior.
Maximizing the entropy selects the distribution that exhibits the most uncertainty which is in accordance with the maximum entropy principle.\footnote{In \cref{sec:fundamentals:inference:priors}, we discussed the maximum entropy principle and the related principle of indifference at length. In simple terms, the maximum entropy principle states that without information, we should choose the distribution that is maximally uncertain.}

Recalling that KL-divergence is non-negative, it follows from \cref{eq:elbo_reverse_kl_relationship} that the evidence lower bound is a (uniform\footnote{That is, the bound holds for any variational distribution $q$ (with full support).}) lower bound to the \midx{evidence} $\log p(y_{1:n} \mid \vx_{1:n})$: \begin{align}
  \log p(y_{1:n} \mid \vx_{1:n}) &\geq L(q, p; \spD_n). \label{eq:log_evidence_bound}
\end{align}
This indicates that maximizing the evidence lower bound is an adequate method of model selection which can be used instead of maximizing the evidence (marginal likelihood) directly (as was discussed in \cref{sec:gp:model_selection:marginal_likelihood}).
Note that this inequality lower bounds the logarithm of an integral by an expectation of a logarithm over some variational distribution $q$.
Hence, the ELBO is a family of lower bounds --- one for each variational distribution.
Such inequalities are called variational inequalities.\looseness=-1

\begin{ex}{Gaussian VI vs Laplace approximation}{}
  Consider maximizing the ELBO within the variational family of Gaussians $\spQ \defeq \braces*{q(\vtheta) = \N[\vtheta]{\vmu}{\mSigma}}$.
  How does this relate to Laplace approximation which also fits a Gaussian approximation to the posterior \begin{align*}
    p(\vtheta \mid \spD) \propto p(y_{1:n}, \vtheta \mid \vx_{1:n})?
  \end{align*}
  It turns out that both approximations are closely related.
  Indeed, it can be shown that while the Laplace approximation is fitted \emph{locally} at the MAP estimate $\vthetahat$, satisfying \begin{align}\begin{split}
    \vzero &= \grad_{\vtheta} \log p(y_{1:n}, \vtheta \mid \vx_{1:n}) \\
    \inv{\mSigma} &= - \left. \hes_{\vtheta} \log p(y_{1:n}, \vtheta \mid \vx_{1:n}) \right\rvert_{\vtheta = \vthetahat},
  \end{split}\label{eq:posterior_laplace_approx}\end{align} Gaussian variational inference satisfies the conditions of the Laplace approximation \emph{on average} with respect to the approximation $q$ \exerciserefmark{gaussian_vi_vs_laplace}: \begin{align}\begin{split}
    \vzero &= \E[\vtheta \sim q]{\grad_{\vtheta} \log p(y_{1:n}, \vtheta \mid \vx_{1:n})} \\
    \inv{\mSigma} &= - \E[\vtheta \sim q]{\hes_{\vtheta} \log p(y_{1:n}, \vtheta \mid \vx_{1:n})}.
  \end{split}\label{eq:gaussian_vi_conditions}\end{align}
  For this reason, the Gaussian variational approximation does not suffer from the same overconfidence as the Laplace approximation.\safefootnote{see \cref{fig:laplace_approx_overconfident}}\looseness=-1
\end{ex}

\begin{ex}{ELBO for Bayesian logistic regression}{}
  Recall that Bayesian logistic regression uses the prior distribution ${\vw \sim \N{\vzero}{\mI}}$.\safefootnote{We omit the scaling factor $\sigmap^2$ here for simplicity.}

  Suppose we use the variational family $\spQ$ of all diagonal Gaussians from \cref{ex:var_family_diag_gaussian}.
  We have already seen in \cref{eq:kl_brl} that for a prior ${p \sim \SN}$ and a variational distribution \begin{align*}
    q_\vlambda \sim \N{\vmu}{\diag[i\in[d]]{\sigma_i^2}},
  \end{align*} we have \begin{align*}
    \KL{q}{p(\cdot)} = \frac{1}{2} \sum_{i=1}^d (\sigma_i^2 + \mu_i^2 - 1 - \log \sigma_i^2).
  \end{align*}

  It remains to find the expected likelihood under models from our approximate posterior: \begin{align}
    \E[\vw \sim q_\vlambda]{\log p(y_{1:n} \mid \vx_{1:n}, \vw)} &= \E[\vw \sim q_\vlambda]{\sum_{i=1}^n \log p(y_i \mid \vx_i, \vw)} \margintag{using independence of the data} \nonumber \\
    &= \E[\vw \sim q_\vlambda]{- \sum_{i=1}^n \ell_{\mathrm{log}}(\vw; \vx_i, y_i)}. \margintag{substituting the logistic loss \eqref{eq:logistic_loss}}
  \end{align}
\end{ex}

\subsection{Gradient of Evidence Lower Bound}\label{sec:approximate_inference:variational_inference:gradient_of_elbo}

We have yet to discuss how the optimization problem of maximizing the ELBO can be solved efficiently.
A suitable tool is stochastic gradient descent (SGD), however, SGD requires unbiased gradient estimates of the loss $\ell(\vlambda; \spD_n) \defeq - L(q_\vlambda, p; \spD_n)$.
That is, we need to obtain gradient estimates of \begin{align}
  \grad_\vlambda L(q_\vlambda, p; \spD_n) = \grad_\vlambda \E[\vtheta \sim q_\vlambda]{\log p(y_{1:n} \mid \vx_{1:n}, \vtheta)} - \grad_\vlambda \KL{q_\vlambda}{p(\cdot)}. \margintag{using the definition of the ELBO \eqref{eq:elbo2}}
\end{align}
Typically, the KL-divergence (and its gradient) can be computed exactly for commonly used variational families.
For example, we have already seen a closed-form expression of the KL-divergence for Gaussians in \cref{eq:kl_gaussian}.

Obtaining the gradient of the expected log-likelihood is more difficult.
This is because the expectation integrates over the measure $q_\vlambda$, which depends on the variational parameters $\vlambda$.
Thus, we cannot move the gradient operator inside the expectation as commonly done (cf. \cref{sec:background:probability:gradients_of_expectations}).
There are two main techniques which are used to rewrite the gradient in such a way that Monte Carlo sampling becomes possible.\looseness=-1

One approach is to use \midx<score gradients>{score gradient estimator} via the ``score function trick'': \begin{align}
  \begin{multlined}
    \grad_{\vlambda} \E[\vtheta \sim q_\vlambda]{\log p(y_{1:n} \mid \vx_{1:n}, \vtheta)} \\ = \E*[\vtheta \sim q_\vlambda]{[\log p(y_{1:n} \mid \vx_{1:n}, \vtheta) \underbrace{\grad_{\vlambda} \log q_{\vlambda}(\vtheta)}_{\text{\midx{score function}[idxpagebf]}}]},
  \end{multlined}
\end{align} which we introduce in \cref{sec:mfarl:policy_approximation:reinforce} in the context of reinforcement learning.
More common in the context of variational inference is the so-called ``reparameterization trick''.

\begin{thm}[Reparameterization trick]\pidx{reparameterization trick}[idxpagebf]\label{thm:reparameterization_trick}
  Given a random variable $\vvarepsilon \sim \phi$ (which is independent of $\vlambda$) and given a differentiable and invertible function $\vg : \R^d \to \R^d$.
  We let $\vtheta \defeq \vg(\vvarepsilon; \vlambda)$.
  Then, \begin{align}
    q_\vlambda(\vtheta) &= \phi(\vvarepsilon) \cdot \inv{\abs{\det{\jac_\vvarepsilon \vg(\vvarepsilon; \vlambda)}}}, \label{eq:reparameterization_trick1} \\
    \E[\vtheta \sim q_\vlambda]{\vf(\vtheta)} &= \E[\vvarepsilon \sim \phi]{\vf(\vg(\vvarepsilon; \vlambda))} \label{eq:reparameterization_trick2}
  \end{align} for a ``nice'' function $\vf : \R^d \to \R^e$.
\end{thm}
\begin{proof}
  By the change of variables formula \eqref{eq:change_of_variables} and using $\vvarepsilon = \inv{\vg}(\vtheta; \vlambda)$, \begin{align*}
    q_\vlambda(\vtheta) &= \phi(\vvarepsilon) \cdot \abs{\det{\jac_\vtheta \inv{\vg}(\vtheta; \vlambda)}} \\
    &= \phi(\vvarepsilon) \cdot \abs{\det{\inv{(\jac_\vvarepsilon \vg(\vvarepsilon; \vlambda))}}} \margintag{by the inverse function theorem, $\jac\inv{\vg}(\vy) = \inv{\jac\vg(\vx)}$} \\[6pt]
    &= \phi(\vvarepsilon) \cdot \inv{\abs{\det{\jac_\vvarepsilon \vg(\vvarepsilon; \vlambda)}}}. \margintag{using $\det{\inv{\mA}} = \inv{\det{\mA}}$}
  \end{align*}
  \Cref{eq:reparameterization_trick2} is a direct consequence of the law of the unconscious statistician \eqref{eq:lotus}.
\end{proof}

In other words, the reparameterization trick allows a change of ``densities'' by finding a function $\vg(\cdot; \vlambda)$ and a reference density $\phi$ such that $q_{\vlambda} = \pf{\vg(\cdot; \vlambda)}{\phi}$ is the pushforward of $\phi$ under perturbation $\vg$.
Applying the reparameterization trick, we can swap the order of gradient and expectation, \begin{align}
  \grad_\vlambda \E[\vtheta \sim q_\vlambda]{\vf(\vtheta)} = \E[\vvarepsilon \sim \phi]{\grad_\vlambda \vf(\vg(\vvarepsilon; \vlambda))}. \margintag{using \cref{eq:swap_grad_exp_order}}
\end{align}
We call a distribution $q_\vlambda$ \midx<reparameterizable>{reparameterizable distribution} if it admits reparameterization, i.e., if we can find $\vg$ and a suitable reference density $\phi$ which is independent of $\vlambda$.

\begin{ex}{Reparameterization trick for Gaussians}{reparameterization_trick_gaussian}
  Suppose we use a Gaussian variational approximation, \begin{align*}
    q_\vlambda(\vtheta) \defeq \N[\vtheta]{\vmu}{\mSigma},
  \end{align*} where we assume $\mSigma$ to have full rank (i.e., be invertible).
  We have seen in \cref{eq:gaussian_lin_trans} that a Gaussian random vector ${\vvarepsilon \sim \SN}$ following a standard normal distribution can be transformed to follow the Gaussian distribution $q_\vlambda$ by using the linear transformation,\looseness=-1 \begin{align}
    \vtheta = \vg(\vvarepsilon; \vlambda) \defeq \msqrt{\mSigma} \vvarepsilon + \vmu. \label{eq:reparameterization_trick_gaussian}
  \end{align}

  In particular, we have \begin{align}
    \vvarepsilon &= \inv{\vg}(\vtheta; \vlambda) = \mSigma^{-\nicefrac{1}{2}}(\vtheta - \vmu) \quad \text{and} \margintag{by solving \cref{eq:reparameterization_trick_gaussian} for $\vvarepsilon$} \\
    \phi(\vvarepsilon) &= q_\vlambda(\vtheta) \cdot \abs{\det{\msqrt{\mSigma}}}. \margintag{using the reparameterization trick (i.e., the change of variables formula) \eqref{eq:reparameterization_trick1}}
  \end{align}
\end{ex}

In the following, we write $\mC \defeq \msqrt{\mSigma}$.
Let us now derive the gradient estimate for the evidence lower bound assuming the Gaussian variational approximation from \cref{ex:reparameterization_trick_gaussian}.
This approach extends to any reparameterizable distribution.
\begin{align}
  &\grad_\vlambda \E[\vtheta \sim q_\vlambda]{\log p(y_{1:n} \mid \vx_{1:n}, \vtheta)} \nonumber \\
  &= \grad_{\mC,\vmu} \E[\vvarepsilon \sim \SN]{\left. \log p(y_{1:n} \mid \vx_{1:n}, \vtheta) \right\rvert_{\vtheta = \mC \vvarepsilon + \vmu}} \label{eq:reparameterization_trick_application} \margintag{using the reparameterization trick \eqref{eq:reparameterization_trick2}} \\
  &= n \cdot \grad_{\mC,\vmu} \E[\vvarepsilon \sim \SN]{\frac{1}{n} \sum_{i=1}^n \left. \log p(y_i \mid \vx_i, \vtheta) \right\rvert_{\vtheta = \mC \vvarepsilon + \vmu}} \margintag{using independence of the data and extending with $\nicefrac{n}{n}$} \nonumber \\
  &= n \cdot \grad_{\mC,\vmu} \E*[\vvarepsilon \sim \SN]{\E[i \sim \Unif{\brackets{n}}]{\left. \log p(y_i \mid \vx_i, \vtheta) \right\rvert_{\vtheta = \mC \vvarepsilon + \vmu}}} \margintag{interpreting the sum as an expectation} \nonumber \\
  &= n \cdot \E*[\vvarepsilon \sim \SN]{\E[i \sim \Unif{\brackets{n}}]{\grad_{\mC,\vmu} \left. \log p(y_i \mid \vx_i, \vtheta) \right\rvert_{\vtheta = \mC \vvarepsilon + \vmu}}} \margintag{using \cref{eq:swap_grad_exp_order}} \\
  &\approx n \cdot \frac{1}{m} \sum_{j=1}^m \grad_{\mC,\vmu} \left. \log p(y_{i_j} \mid x_{i_j}, \vtheta) \right\rvert_{\vtheta = \mC \vvarepsilon_j + \vmu} \margintag{using Monte Carlo sampling}
\end{align} where $\vvarepsilon_j \iid \SN$ and $i_j \iid \Unif{[n]}$.
This yields an unbiased gradient estimate, which we can use with stochastic gradient descent to maximize the evidence lower bound.
We have successfully recast the difficult problems of learning and inference as an optimization problem!\looseness=-1

The procedure of approximating the true posterior using a variational posterior by maximizing the evidence lower bound using stochastic optimization is also called \midx{black box stochastic variational inference} \citep{ranganath2014black,titsias2014doubly,duvenaud2015black}.
The only requirement is that we can obtain unbiased gradient estimates from the evidence lower bound (and the likelihood).
We have just discussed one of many approaches to obtain such gradient estimates \citep{mohamed2020monte}.
If we use the variational family of diagonal Gaussians, we only require twice as many parameters as other inference techniques like MAP estimation.
The performance can be improved by using natural gradients and variance reduction techniques for the gradient estimates such as control variates.\looseness=-1

\subsection{Minimizing Surprise via Exploration and Exploitation}\label{sec:free_energy}

Now that we have established a way to optimize the ELBO, let us dwell a bit more on its interpretation.
Observe that the evidence can also be interpreted as the negative surprise about the observations under the prior distribution $p(\vtheta)$, and by negating \cref{eq:log_evidence_bound}, we obtain the variational upper bound \begin{align}\begin{split}
  \S{p(y_{1:n} \mid \vx_{1:n})} &\leq \underbrace{\E[\vtheta \sim q]{\S{p(y_{1:n}, \vtheta \mid \vx_{1:n})}}}_{\text{called \emph{energy}}} - \H{q} \\
  &= - L(q, p; \spD_n).
\end{split}\end{align}
Here, $-L(q, p; \spD_n)$ is commonly called the \midx<(variational) free energy>{free energy} with respect to $q$.
Free energy can also be characterized as \begin{align}
  - L(q, p; \spD_n) &= \underbrace{\E[\vtheta \sim q]{\S{p(y_{1:n} \mid \vx_{1:n}, \vtheta)}}}_{\text{average surprise}} + \underbrace{\KL{q}{p(\cdot)}}_{\text{proximity to prior}}, \margintag{analogously to \cref{eq:elbo2}}
\end{align} and therefore its minimization is minimizing the average surprise about the data under the variational distribution $q$ while maximizing proximity to the prior $p(\cdot)$.
Systems that minimize the surprise in their observations are widely studied in many areas of science.\footnote{Refer to the \idx{free energy principle}[idxpagebf] which was originally introduced by the neuroscientist Karl Friston \citep{friston2010free}.}

To minimize surprise, the free energy makes apparent a natural tradeoff between two extremes:
On the one hand, models $q(\vtheta)$ that ``overfit'' to observations (e.g., by using a point estimate of $\vtheta$), and hence, result in a large surprise when new observations deviate from this specific belief.
On the other hand, models $q(\vtheta)$ that ``underfit'' to observations (e.g., by expecting outcomes with equal probability), and hence, any observation results in a non-negligible surprise.
Either of these extremes is undesirable.

As we have alluded to previously when introducing the ELBO, the two terms constituting free energy map neatly onto this tradeoff.
Therein, the \emph{entropy} (which is maximized) encourages $q$ to have uncertainty, in other words, to ``explore'' beyond the finite data.
In contrast, the \emph{energy} (which is minimized) encourages $q$ to fit the observed data closely, in other words, to ``exploit'' the finite data.
This tradeoff is ubiquitous in approximations to probabilistic inference that deal with limited computational resources and limited time, and we will encounter it many times more.
We will point out these connections as we go along.

Since this tradeoff is so fundamental and appears in many branches of science under different names, it is difficult to give it an appropriate unifying name.
The essence of this tradeoff can be captured as a \midx{principle of curiosity and conformity}[idxpagebf], which suggests that reasoning under uncertainty requires curiosity to entertain and pursue alternative explanations of the data \emph{and} conformity to make consistent predictions.

\section*{Discussion}

We have explored variational inference, where we approximate the intractable posterior distribution of probabilistic inference with a simpler distribution.
We operationalized this idea by turning the inference problem, which requires computing high-dimensional integrals, into a tractable optimization problem.
Gaussians are frequently used as variational distributions due to their versatility and compact representation.

Nevertheless, recall from \cref{fig:variational_families} that while the estimation error in variational inference can be small, choosing a variational family that is too simple can lead to a large approximation error.
We have seen that for posteriors that are multimodal or have heavy tails, Gaussians may not provide a good approximation.
In the next chapter, we will explore alternative techniques for approximate inference that can handle more complex posteriors.

\excheading

\begin{nexercise}{Logistic loss}{logistic_loss_gradient}
  \begin{enumerate}
    \item Derive the gradient of $\ell_\mathrm{log}$ as given in \cref{eq:logistic_loss_gradient}.
    \item Show that \begin{align}
      \hes_\vw \ell_\mathrm{log}(\transpose{\vw} \vx; y) = \vx \transpose{\vx} \cdot \sigma(\transpose{\vw} \vx) \cdot (1 - \sigma(\transpose{\vw} \vx)). \label{eq:logistic_loss_hes}
    \end{align}
    \textit{Hint: Begin by deriving the first derivative of the logistic function, and use the chain rule of multivariate calculus, \begin{align}
      \underbrace{\mD_\vx (\vf \circ \vg)}_{\R^n \to \R^{m \times n}} = \underbrace{(\mD_{\vg(\vx)} \vf \circ \vg) \cdot \mD_\vx \vg}_{\R^n \to \R^{m \times k} \cdot \R^{k \times n}} \label{eq:chain_rule}
    \end{align} where $\vg : \R^n \to \R^k$ and $\vf : \R^k \to \R^m$.}
    \item Is the logistic loss $\ell_\mathrm{log}$ convex in $\vw$?
  \end{enumerate}
\end{nexercise}

\begin{nexercise}{Gaussian process classification}{gpc}
  In this exercise, we will study the use of Gaussian processes for classification tasks, commonly called \midx{Gaussian process classification}[idxpagebf] (GPC).
  Linear logistic regression is extended to GPC by replacing the Gaussian prior over weights with a GP prior on $f$, \begin{align}
    f \sim \GP{0}{k}, \quad y \mid \vx, f \sim \Bern{\sigma(f(\vx))} \label{eq:gpc1}
  \end{align} where $\sigma : \mathbb{R} \to (0,1)$ is some logistic-type function.
  Note that Bayesian logistic regression is the special case where $k$ is the linear kernel and~$\sigma$ is the logistic function.
  This is analogous to the relationship of Bayesian linear regression and Gaussian process regression.

  In the GP regression setting of \cref{sec:gp}, $y_i$ was assumed to be a noisy observation of $f(\vx_i)$.
  In the classification setting, we now have that $y_i \in \{-1,+1\}$ is a binary class label and $f(\vx_i) \in \R$ is a latent value.
  We study the setting where $\sigma(z) = \Phi(z; 0, \sigman^2)$ is the CDF of a univariate Gaussian with mean $0$ and variance $\sigman^2$, also called a \midx{probit likelihood}.

  To make probabilistic predictions for a query point $\vxs$, we first compute the distribution of the latent variable $\fs$,
  \begin{align}
    p(\fs \mid \vx_{1:n}, y_{1:n}, \vxs) = \int p(\fs \mid \vx_{1:n}, \vxs, \vf) p(\vf \mid x_{1:n}, y_{1:n}) \,d \vf \label{eq:gpc_latent_predictive_posterior} \margintag{using sum rule \eqref{eq:sum_rule} and product rule \eqref{eq:product_rule}, and $\fs \perp y_{1:n} \mid \vf$}
  \end{align} where $p(\vf \mid \vx_{1:n}, y_{1:n})$ is the posterior over the latent variables.

  \begin{enumerate}
    \item Assuming that we can efficiently compute $p(\fs \mid \vx_{1:n}, y_{1:n}, \vxs)$ (approximately), describe how we can find the predictive posterior $p(\ys = +1 \mid \vx_{1:n}, y_{1:n}, \vxs)$.

    \item The posterior over the latent variables is not a Gaussian as we used a non-Gaussian likelihood, and hence, the integral of the latent predictive posterior \eqref{eq:gpc_latent_predictive_posterior} is analytically intractable.
    A common technique is to approximate the latent posterior $p(\vf \mid \vx_{1:n}, y_{1:n})$ with a Gaussian using a Laplace approximation $q \defeq \N{\vfhat}{\inv{\mLambda}}$.
    It is generally not possible to obtain an analytical representation of the mode of the Laplace approximation $\vfhat$. Instead, $\vfhat$ is commonly found using a second-order optimization scheme such as Newton's method.\looseness=-1
    \begin{enumerate}
      \item Find the precision matrix $\mLambda$ of the Laplace approximation.

      \textit{Hint: Observe that for a label $y_i \in \{-1,+1\}$, the probability of a correct classification given the latent value $f_i$ is $p(y_i \mid f_i) = \sigma(y_i f_i)$, where we use the symmetry of the probit likelihood around $0$.}

      \item Assume that $k(\vx, \vxp) = \transpose{\vx}\vxp$ is the linear kernel ($\sigmap = 1$) and that $\sigma$ is the logistic function \eqref{eq:logistic_function}.
      Show for this setting that the matrix $\mLambda$ derived in (a) is equivalent to the precision matrix $\mLambda'$ of the Laplace approximation of Bayesian logistic regression \eqref{eq:prec_mat_lap_approx_blr}.\safefootnote{This should not be surprising since --- as already mentioned --- Gaussian process classification is a generalization of Bayesian logistic regression.}
      You may assume that $\hat{f}_i = \transpose{\vwhat} \vx_i$.

      \textit{Hint: First derive under which condition $\mLambda$ and $\mLambda'$ are ``equivalent''.}

      \item Observe that the (approximate) latent predictive posterior \begin{align*}
        q(\fs \mid \vx_{1:n}, y_{1:n}, \vxs) \defeq \int p(\fs \mid \vx_{1:n}, \vxs, \vf) q(\vf \mid x_{1:n}, y_{1:n}) \,d \vf
      \end{align*} which uses the Laplace approximation of the latent posterior is Gaussian.\safefootnote{Using the Laplace-approximated latent posterior, $[\fs \; \vf]$ are jointly Gaussian. Thus, it directly follows from \cref{fct:marginal_and_cond_gaussian} that the marginal distribution over $\fs$ is also a Gaussian.}
      Determine its mean and variance.

      \textit{Hint: Condition on the latent variables $\vf$ using the laws of total expectation and variance.}

      \item Compare the prediction $p(\fs \mid \vx_{1:n}, y_{1:n}, \vxs)$ you obtained in (1) (but now using the Laplace-approximated latent predictive posterior) to the prediction $\sigma(\E[\fs \sim q]{\fs})$. Are they identical? If not, describe how they are different.
    \end{enumerate}

    \item The use of the probit likelihood may seem arbitrary.
    Consider the following model which may be more natural, \begin{align}
      f \sim \GP{0}{k}, \quad y = \Ind{\underbrace{f(\vx) + \varepsilon}_{\text{GP regression}} \geq 0}, \quad \varepsilon \sim \N{0}{\sigman^2}. \label{eq:gpc2}
    \end{align}
    Show that the model from \cref{eq:gpc1} using a noise-free latent process with probit likelihood $\Phi(z; 0, \sigman^2)$ is equivalent (in expectation over $\varepsilon$) to the model from \cref{eq:gpc2}.
  \end{enumerate}
\end{nexercise}

\begin{nexercise}{Jensen's inequality}{jensen}
  \begin{enumerate}
    \item Prove the finite form of Jensen's inequality.

    \begin{thm}[Jensen's inequality, finite form]
      Let $f : \R^n \to \R$ be a convex function. Suppose that $\vx_1, \dots, \vx_k \in \R^n$ and $\theta_1, \dots, \theta_k \geq 0$ with $\theta_1 + \dots + \theta_k = 1$. Then, \begin{align}
        f(\theta_1 \vx_1 + \dots + \theta_k \vx_k) \leq \theta_1 f(\vx_1) + \dots + \theta_k f(\vx_k).
      \end{align}
    \end{thm}

    Observe that if $X$ is a random variable with finite support, the above two versions of Jensen's inequality are equivalent.

    \item Show that for any discrete distribution $p$ supported on a finite domain of size $n$, $\H{p} \leq \log_2 n$.
    This implies that the uniform distribution has maximum entropy.
  \end{enumerate}
\end{nexercise}

\begin{nexercise}{Binary cross-entropy loss}{bce_loss}
  Show that the logistic loss \eqref{eq:logistic_loss} is equivalent to the binary cross-entropy loss with $\hat{y} = \sigma(\hat{f})$.
  That is, \begin{align}
    \ell_\mathrm{log}(\hat{f}; y) = \ell_\mathrm{bce}(\hat{y}; y).
  \end{align}
\end{nexercise}

\begin{nexercise}{Gibbs' inequality}{gibbs_ineq}
  \begin{enumerate}
    \item Prove $\KL{p}{q} \geq 0$ which is also known as \midx{Gibbs' inequality}.
    \item Let $p$ and $q$ be discrete distributions with finite identical support $\sA$.
    Show that $\KL{p}{q} = 0$ if and only if $p \equiv q$.\par
    \textit{Hint: Use that if a function $f : \R^n \to \R$ is strictly convex and $\vx_1, \dots, \vx_k \in \R^n$, $\theta_1, \dots, \theta_k \geq 0$, $\theta_1 + \dots + \theta_k = 1$, we have that \begin{align}
      f(\theta_1 \vx_1 + \dots + \theta_k \vx_k) = \theta_1 f(\vx_1) + \dots + \theta_k f(\vx_k)
    \end{align} iff $\vx_1 = \dots = \vx_k$. This is a slight adaptation of Jensen's inequality in finite-form, which you proved in \cref{exercise:jensen}.}
  \end{enumerate}
\end{nexercise}

\begin{nexercise}{Maximum entropy principle}{maximum_entropy_principle}
  In this exercise we will prove that the normal distribution is the distribution with maximal entropy among all (univariate) distributions supported on $\R$ with fixed mean $\mu$ and variance $\sigma^2$.
  Let ${g(x) \defeq \N[x]{\mu}{\sigma^2}}$, and $f(x)$ be any distribution on~$\R$ with mean~$\mu$ and variance~$\sigma^2$.
  \begin{enumerate}
    \item Prove that $\KL{f}{g} = \H{g} - \H{f}$.\par
    \textit{Hint: Equivalently, show that $\crH{f}{g} = \H{g}$.
    That is, the expected surprise evaluated based on the Gaussian $g$ is invariant to the true distribution $f$.}
    \item Conclude that $\H{g} \geq \H{f}$.
  \end{enumerate}
\end{nexercise}

\begin{nexercise}{Probabilistic inference as a consequence of the maximum entropy principle}{mep_and_posteriors}
  Consider the family of generative models of the random vectors $\rX$ in~$\spX$ and $\rY$ in~$\spY$: \begin{align*}
    \Delta^{\spX \times \spY} = \left\{q : \spX \times \spY \to \Rzero\ \middle|\ \text{$\int_{\spX\times\spY} q(\vx,\vy) \,d\vx \,d\vy = 1$}\right\}.
  \end{align*}
  Suppose that we observe $\rY$ to be $\vyp$, and are looking for a (new) generative model that is consistent with this information, that is, \begin{align*}
    q(\vy) = \int_\spX q(\vx,\vy) \,d\vx = \delta_{\vyp}(\vy) \margintag{using the sum rule \eqref{eq:sum_rule}}
  \end{align*} where $\delta_{\vyp}$ denotes the point density at $\vyp$.
  The product rule \eqref{eq:product_rule} implies that $q(\vx,\vy) = \delta_{\vyp}(\vy) \cdot q(\vx \mid \vy)$, but any choice of $q(\vx \mid \vy)$ is possible.\looseness=-1

  We will derive that given any fixed generative model $p_{\rX,\rY}$, the ``posterior'' distribution $q_{\rX}(\cdot) = p_{\rX\mid\rY}(\cdot \mid \vyp)$ minimizes the relative entropy $\KL{q_{\rX,\rY}}{p_{\rX,\rY}}$ subject to the constraint $\rY = \vyp$.
  In other words, among all distributions $q_{\rX,\rY}$ that are consistent with the observation $\rY = \vyp$, the posterior distribution $q_{\rX}(\cdot) = p_{\rX\mid\rY}(\cdot \mid \vyp)$ is the one with ``maximum entropy''.

  \begin{enumerate}
    \item Show that the optimization problem \begin{align*}
      &\argmin_{q \in \Delta^{\spX \times \spY}} \KL{q_{\rX,\rY}}{p_{\rX,\rY}} \\
      &\text{subject to}\ q(\vy) = \delta_{\vyp}(\vy) \quad \forall \vy \in \spY
    \end{align*} is solved by $q(\vx,\vy) = \delta_{\vyp}(\vy) \cdot p(\vx \mid \vy)$. \\
    \textit{Hint: Solve the dual problem.}

    \item Conclude that $q(\vx) = p(\vx \mid \vyp)$.
  \end{enumerate}
\end{nexercise}

\begin{nexercise}{KL-divergence of Gaussians}{kl_div_of_gaussians}
  Derive \cref{eq:kl_gaussian}.

  \textit{Hint: If $\rX \sim \N{\vmu}{\mSigma}$ in $d$ dimensions, then we have that for any $\vm \in \R^d$ and $\mA \in \R^{d \times d}$, \begin{align}
    \E{\transpose{(\rX - \vm)}\mA(\rX - \vm)} = \transpose{(\vmu - \vm)}\mA(\vmu - \vm) + \tr{\mA\mSigma} \label{eq:kl_div_of_gaussians_hint}
  \end{align}}
\end{nexercise}

\begin{nexercise}{Forward vs reverse KL}{forward_vs_reverse_kl}
  \begin{enumerate}
    \item Consider a factored approximation $q(x, y) = q(x)q(y)$ to a joint distribution $p(x, y)$.
    Show that to minimize the forward $\KL{p}{q}$ we should set $q(x) = p(x)$ and $q(y) = p(y)$, i.e., the optimal approximation is a product of marginals.

    \item Consider the following joint distribution $p$, where the rows represent $y$ and the columns $x$:

    \begin{center}
      \vspace{5pt}
      \begin{tabular}{l|cccc}
        \toprule
        & $1$ & $2$ & $3$ & $4$ \\
        \midrule
        $1$ & $\nicefrac{1}{8}$ & $\nicefrac{1}{8}$ & $0$ & $0$ \\
        $2$ & $\nicefrac{1}{8}$ & $\nicefrac{1}{8}$ & $0$ & $0$ \\
        $3$ & $0$ & $0$ & $\nicefrac{1}{4}$ & $0$ \\
        $4$ & $0$ & $0$ & $0$ & $\nicefrac{1}{4}$ \\
        \bottomrule
      \end{tabular}
      \vspace{5pt}
    \end{center}

    Show that the reverse $\KL{q}{p}$ for this $p$ has three distinct minima.
    Identify those minima and evaluate $\KL{q}{p}$ at each of them.

    \item What is the value of $\KL{q}{p}$ if we use the approximation ${q(x, y) = p(x) p(y)}$?
  \end{enumerate}
\end{nexercise}

\begin{nexercise}{Gaussian VI vs Laplace approximation}{gaussian_vi_vs_laplace}
  In this exercise, we compare the Laplace approximation from \cref{sec:approximate_inference:laplace_approximation} to variational inference with the variational family of Gaussians,\looseness=-1 \begin{align*}
    \spQ \defeq \braces*{q(\vtheta) = \N[\vtheta]{\vmu}{\mSigma}}.
  \end{align*}
  \begin{enumerate}
    \item Let $p$ be any distribution on $\R$, and let $\opt{q} = \argmin_{q \in \spQ} \KL{p}{q}$.
    Show that $\opt{q}$ differs from the Laplace approximation of $p$.
  \end{enumerate}

  Minimizing forward-KL is typically intractable, and we have seen that it is therefore common to minimize the reverse-KL instead: \begin{align*}
    \tilde{q} = \argmin_{q \in \spQ} \KL{q}{p(\cdot \mid \spD_n)}.
  \end{align*}
  \begin{enumerate}
    \setcounter{enumi}{1}
    \item Show that $\tilde{q} = \N{\vmu}{\mSigma}$ satisfies \cref{eq:gaussian_vi_conditions}: \begin{align*}
      \vzero &= \E[\vtheta \sim \tilde{q}]{\grad_{\vtheta} \log p(y_{1:n}, \vtheta \mid \vx_{1:n})} \\
      \inv{\mSigma} &= - \E[\vtheta \sim \tilde{q}]{\hes_{\vtheta} \log p(y_{1:n}, \vtheta \mid \vx_{1:n})}.
    \end{align*}

    \textit{Hint 1: For any positive definite and symmetric matrix $\mA$, it holds that $\grad_{\mA} \log\det{\mA} = \inv{\mA}$.}

    \textit{Hint 2: For any function $f$ and Gaussian $p = \N{\vmu}{\mSigma}$, \begin{align}\begin{split}
      \grad_{\vmu} \E[\vx \sim p]{f(\vx)} &= \E[\vx \sim p]{\grad_{\vx} f(\vx)} \\
      \grad_{\mSigma} \E[\vx \sim p]{f(\vx)} &= \frac{1}{2} \E[\vx \sim p]{\hes_{\vx} f(\vx)}.
    \end{split}\label{eq:gradient_of_gaussian_parameters}\end{align}}
  \end{enumerate}

  Recall the conditions satisfied by the Laplace approximation of the posterior $p(\vtheta \mid \spD) \propto p(y_{1:n}, \vtheta \mid \vx_{1:n})$ as detailed in \cref{eq:posterior_laplace_approx}.
  The Laplace approximation is fitted \emph{locally} at the MAP estimate $\vthetahat$.
  Comparing \cref{eq:posterior_laplace_approx} to \cref{eq:gaussian_vi_conditions}, we see that Gaussian variational inference satisfies the conditions of the Laplace approximation \emph{on average}.
  For more details, refer to \cite{opper2009variational}.
\end{nexercise}

\begin{nexercise}{Gradient of reverse-KL}{gradient_of_reverse_kl}
  Suppose $p \defeq \N{\vzero}{\sigmap^2 \mI}$ and a tractable distribution described by \begin{align*}
    q_\vlambda \defeq \N{\vmu}{\diag{\sigma_1^2, \dots, \sigma_d^2}}
  \end{align*} where $\vmu \defeq [\mu_1 \; \cdots \; \mu_d]$ and $\vlambda \defeq [\mu_1 \; \cdots \; \mu_d \; \sigma_1 \; \cdots \; \sigma_d]$.
  Show that the gradient of $\KL{q_\vlambda}{p(\cdot)}$ with respect to $\vlambda$ is given by \begin{subequations}\begin{align}
    \grad_\vmu \KL{q_\vlambda}{p(\cdot)} &= \sigmap^{-2} \vmu, \quad\text{and} \\
    \grad_{[\sigma_1 \; \cdots \; \sigma_d]} \KL{q_\vlambda}{p(\cdot)} &= \begin{bmatrix}
      \frac{\sigma_1}{\sigmap^2} - \frac{1}{\sigma_1} & \dots & \frac{\sigma_d}{\sigmap^2} - \frac{1}{\sigma_d}
    \end{bmatrix}.
  \end{align}\end{subequations}
\end{nexercise}

\begin{nexercise}{Reparameterizable distributions}{reparameterizable_distributions}
  \begin{enumerate}
    \item Let $X \sim \Unif{[a,b]}$ for any $a \leq b$.
    That is, \begin{align}
      p_X(x) = \begin{cases}
        \frac{1}{b-a} & \text{if $x \in [a,b]$} \\
        0 & \text{otherwise}. \\
      \end{cases}
    \end{align}
    Show that $X$ can be reparameterized in terms of $\Unif{[0,1]}$.
    \textit{Hint: You may use that for any $Y \sim \Unif{[a,b]}$ and $c \in \R$, \begin{itemize}
      \item $Y + c \sim \Unif{[a + c, b + c]}$ and
      \item $cY \sim \Unif{[c \cdot a, c \cdot b]}$.
    \end{itemize}}

    \item Let $Z \sim \N{\mu}{\sigma^2}$ and $X \defeq e^Z$. That is, $X$ is logarithmically normal distributed with parameters $\mu$ and $\sigma^2$. Show that $X$ can be reparameterized in terms of $\N{0}{1}$.

    \item Show that $\Cauchy{0}{1}$ can be reparameterized in terms of $\Unif{[0,1]}$.
  \end{enumerate}
  Finally, let us apply the reparameterization trick to compute the gradient of an expectation.
  \begin{enumerate}
    \setcounter{enumi}{3}
    \item Let $\mathrm{ReLU}(z) \defeq \max\{0, z\}$ and $w > 0$.
    Show that \begin{align*}
      \odv{}{\mu} \E*[x \sim \N{\mu}{1}]{\mathrm{ReLU}(w x)} = w \Phi(\mu)
    \end{align*} where $\Phi$ denotes the CDF of the standard normal distribution.
  \end{enumerate}
\end{nexercise}

  \chapter{Markov Chain Monte Carlo Methods}\label{sec:approximate_inference:mcmc}

Variational inference approximates the entire posterior distribution.
However, note that the key challenge in probabilistic inference is not learning the posterior distribution, but using the posterior distribution for predictions, \begin{align}
  p(\ys \mid \vxs, \vx_{1:n}, y_{1:n}) &= \int p(\ys \mid \vxs, \vtheta) p(\vtheta \mid \vx_{1:n}, y_{1:n}) \,d\vtheta.
  \intertext{This integral can be interpreted as an expectation over the posterior distribution,\looseness=-1}
  &= \E[\vtheta \sim p(\cdot \mid \vx_{1:n}, y_{1:n})]{p(\ys \mid \vxs, \vtheta)}.
  \intertext{Observe that the likelihood $f(\vtheta) \defeq p(\ys \mid \vxs, \vtheta)$ is easy to evaluate. The difficulty lies in sampling from the posterior distribution. Assuming we can obtain independent samples from the posterior distribution, we can use Monte Carlo sampling to obtain an unbiased estimate of the expectation,}
  &\approx \frac{1}{m} \sum_{i=1}^m f(\vtheta^{(i)})
\end{align} for independent $\vtheta^{(i)} \iid p(\cdot \mid \vx_{1:n}, y_{1:n})$.
The law of large numbers \eqref{eq:slln} and Hoeffding's inequality \eqref{eq:hoeffdings_inequality} imply that this estimator is consistent and sharply concentrated.\footnote{For more details, see \cref{sec:fundamentals:mc_approx}.}

Obtaining samples of the posterior distribution is therefore sufficient to perform approximate inference.
Recall that the difficulty of computing the posterior $p$ exactly, was in finding the normalizing constant~$Z$,\looseness=-1 \begin{align}
  p(x) = \frac{1}{Z} q(x). \label{eq:mcmc_posterior}
\end{align}
The joint likelihood $q$ is typically easy to obtain.
Note that $q(x)$ is proportional to the probability density associated with $x$, but $q$ does not integrate to 1. Such functions are also called a \midx{finite measure}.
Without normalizing $q$, we cannot directly sample from it.

\begin{rmk}{The difficulty of sampling --- even with a PDF}{}
  Even a decent approximation of $Z$ does not yield a general efficient sampling method.
  For example, one very common approach to sampling is \midx{inverse transform sampling} (cf. \cref{sec:inverse_transform_sampling}) which requires an (approximate) quantile function.
  Computing the quantile function given an arbitrary PDF requires solving integrals over the domain of the PDF which is what we were trying to avoid in the first place.
\end{rmk}

The key idea of Markov chain Monte Carlo methods is to construct a Markov chain, which is efficient to simulate and has the stationary distribution $p$.

\section{Markov Chains}

To start, let us revisit the fundamental theory behind Markov chains.\looseness=-1

\begin{defn}[Markov chain]\pidx{Markov chain}
  A \emph{(finite and discrete-time) Markov chain} over the state space \begin{align}
    \sS \defeq \{0, \dots, n-1\}
  \end{align} is a stochastic process\footnote{A \idx{stochastic process} is a sequence of random variables.} $(X_t)_{t \in \Nat_0}$ valued in $\sS$ such that the \midx{Markov property}[idxpagebf] is satisfied: \begin{align}
    X_{t+1} \perp X_{0:t-1} \mid X_t. \label{eq:markov_property}
  \end{align}
\end{defn}

\begin{marginfigure}
  \incfig{mc}
  \caption{Directed graphical model of a Markov chain.
  The random variable $X_{t+1}$ is conditionally independent of the random variables $X_{0:t-1}$ given $X_t$.}
\end{marginfigure}

Intuitively, the Markov property states that future behavior is independent of past states given the present state.

\begin{rmk}{Generalizations of Markov chains}{}
  One can also define continuous-state Markov chains (for example, where states are vectors in $\R^d$) and the results which we state for (finite) Markov chains will generally carry over.
  For a survey, refer to \icite{roberts2004general}.

  Moreover, one can also consider continuous-time Markov chains.
  One example of such a continuous-space and continuous-time Markov chain is the \midx{Wiener process} (cf. \cref{rmk:wiener_process}).
\end{rmk}

We restrict our attention to \midx<time-homogeneous>{time-homogeneous process} Markov chains,\footnote{That is, the transition probabilities do not change over time.} which can be characterized by a \midx{transition function}, \begin{align}
  p(x' \mid x) \defeq \Pr{X_{t+1} = x' \mid X_t = x}.
\end{align}
As the state space is finite, we can describe the transition function by the \midx{transition matrix}, \begin{align}
  \mP \defeq \begin{bmatrix}
    p(x_1 \mid x_1) & \cdots & p(x_n \mid x_1) \\
    \vdots & \ddots & \vdots \\
    p(x_1 \mid x_n) & \cdots & p(x_n \mid x_n) \\
  \end{bmatrix} \in \R^{n \times n}. \label{eq:transition_matrix}
\end{align}
Note that each row of $\mP$ must always sum to $1$.
Such matrices are also called \midx<stochastic>{stochastic matrix}.

The \midx{transition graph} of a Markov chain is a directed graph consisting of vertices $\sS$ and weighted edges represented by the adjacency matrix $\mP$.

The current state of the Markov chain at time $t$ is denoted by the probability distribution $q_t$ over states $\sS$, that is, $X_t \sim q_t$. In the finite setting, $q_t$ is a PMF, which is often written explicitly as the row vector~${\vq_t \in \R^{1 \times \card{\sS}}}$.
The initial state (or prior) of the Markov chain is given as~${X_0 \sim q_0}$.
One iteration of the Markov chain can then be expressed as follows:~\exerciserefmark{mc_update} \begin{align}
  \vq_{t+1} = \vq_t \mP. \label{eq:mc_update}
\end{align}
It is implied directly that we can write the state of the Markov chain at time $t + k$ as \begin{align}
  \vq_{t+k} = \vq_t \mP^k.
\end{align}
The entry $\mP^k(x,x')$ corresponds to the probability of transitioning from state $x \in \sS$ to state $x' \in \sS$ in exactly $k$ steps \exerciserefmark{mc_multi_step_transitions}.
We denote this entry by $p^{(k)}(x' \mid x)$.

In the analysis of Markov chains, there are two main concepts of interest: stationarity and convergence.
We begin by introducing stationarity.\looseness=-1

\subsection{Stationarity}

\begin{defn}[Stationary distribution]\pidx{stationary distribution}
  A distribution $\pi$ is \emph{stationary} with respect to the transition function $p$ iff \begin{align}
    \pi(x) = \sum_{x' \in S} p(x \mid x') \pi(x')
  \end{align} holds for all $x \in S$.
  It follows from \cref{eq:mc_update} that equivalently, $\pi$ is stationary w.r.t. a transition matrix $\mP$ iff \begin{align}
    \vpi = \vpi \mP. \label{eq:stationarity_vector_form}
  \end{align}
\end{defn}

After entering a stationary distribution $\pi$, a Markov chain will always remain in the stationary distribution.
In particular, suppose that $X_t$ is distributed according to $\pi$, then for all $k \geq 0$, $X_{t+k} \sim \pi$.

\begin{rmk}{When does a stationary distribution exist?}{}
  In general, there are Markov chains with infinitely many stationary distributions or no stationary distribution at all.
  You can find some examples in \cref{fig:mc_examples}.

  It can be shown that there exists a unique stationary distribution~$\pi$ if the Markov chain is \midx<irreducible>{irreducible Markov chain}, that is, if every state is reachable from every other state with a positive probability when the Markov chain is run for enough steps. Formally, \begin{align}
    \forall x,x' \in \sS. \, \exists k \in \Nat. \, p^{(k)}(x' \mid x) > 0.
  \end{align}
  Equivalently, a Markov chain is irreducible iff its transition graph is strongly connected.
\end{rmk}

\subsection{Convergence}\label{sec:mcmc:mc:convergence}

Let us now consider Markov chains with a unique stationary distribution.\footnote{Observe that the stationary distribution of an irreducible Markov chain must have full support, that is, assign positive probability to every state.}
A natural next question is whether this Markov chain converges to its stationary distribution.
We say that a Markov chain converges to its stationary distribution iff we have \begin{align}
  \lim_{t\to\infty} q_{t} = \pi,
\end{align} irrespectively of the initial distribution $q_0$.

\begin{rmk}{When does a Markov chain converge?}{mc_convergence}
  Even if a Markov chain has a unique stationary distribution, it does not have to converge to it.
  Consider example (3) in \cref{fig:mc_examples}.
  Clearly, $\pi = (\frac{1}{2}, \frac{1}{2})$ is the unique stationary distribution.
  However, observe that if we start with a suitable initial distribution such as $q_0 = (1,0)$, at no point in time will the probability of all states be positive, and in particular, the chain will not converge to $\pi$.
  Instead, the chain behaves periodically, i.e., its state distributions are $q_{2t} = (0,1)$ and $q_{2t+1} = (1,0)$ for all $t \in \Nat$.
  It turns out that if we exclude such ``periodic'' Markov chains, then the remaining (irreducible) Markov chains will always converge to their stationary distribution.

  Formally, a Markov chain is \midx<aperiodic>{aperiodic Markov chain} if for all states $x \in \sS$, \begin{align}
    \exists k_0 \in \Nat. \, \forall k \geq k_0. \, p^{(k)}(x \mid x) > 0.
  \end{align}
  In words, a Markov chain is aperiodic iff for every state $x$, the transition graph has a closed path from $x$ to $x$ with length $k$ for all $k \in \Nat$ greater than some $k_0 \in \Nat$.
\end{rmk}

This additional property leads to the concept of ergodicity.

\begin{defn}[Ergodicity]\pidx{ergodicity}
  A Markov chain is \emph{ergodic} iff there exists a $t \in \Nat_0$ such that for any $x, x' \in S$ we have \begin{align}
    p^{(t)}(x' \mid x) > 0,
  \end{align} whereby $p^{(t)}(x' \mid x)$ is the probability to reach $x'$ from $x$ in exactly $t$ steps.
  Equivalent conditions are \begin{enumerate}
    \item that there exists some $t \in \Nat_0$ such that all entries of $\mP^t$ are strictly positive; and
    \item that it is irreducible and aperiodic.
  \end{enumerate}
\end{defn}

\begin{marginfigure}
  \incfig{mc_examples}
  \caption{Transition graphs of Markov chains: (1) is not ergodic as its transition diagram is not strongly connected; (2) is not ergodic for the same reason; (3) is irreducible but periodic and therefore not ergodic; (4) is ergodic with stationary distribution $\pi(1) = \nicefrac{2}{3}, \pi(2) = \nicefrac{1}{3}$.}\label{fig:mc_examples}
\end{marginfigure}

\begin{ex}{Making a Markov chain ergodic}{}
  A commonly used strategy to ensure that a Markov chain is ergodic is to add ``self-loops'' to every vertex in the transition graph. That is, to ensure that at any point in time, the Markov chain remains with positive probability in its current state.

  Take a (not necessarily ergodic) but irreducible Markov chain with transition matrix $\mP$. We define the new Markov chain \begin{align}
    \mP' \defeq \frac{1}{2}\mP + \frac{1}{2}\mI.
  \end{align}
  It is a simple exercise to confirm that $\mP'$ is stochastic, and hence a valid transition matrix.
  Also, it follows directly that $\mP'$ is irreducible (as $\mP$ is irreducible) and aperiodic as every vertex has a closed path of length $1$ to itself, and therefore the chain is ergodic.

  Take now $\vpi$ to be a stationary distribution of $\mP$. We have that $\vpi$ is also a stationary distribution of $\mP'$ as \begin{align}
    \vpi\mP' = \frac{1}{2}\vpi\mP + \frac{1}{2}\vpi\mI = \frac{1}{2}\vpi + \frac{1}{2}\vpi = \vpi. \margintag{using \eqref{eq:stationarity_vector_form}}
  \end{align}
\end{ex}

\begin{fct}[Fundamental theorem of ergodic Markov chains, theorem~4.9 of \cite{levin2017markov}]\pidx{fundamental theorem of ergodic Markov chains}
  An ergodic Markov chain has a unique stationary distribution $\pi$ (with full support) and \begin{align}
    \lim_{t\to\infty} q_t = \pi \label{eq:fund_thm_of_ergodic_mc}
  \end{align} irrespectively of the initial distribution $q_0$.
\end{fct}

This naturally suggests constructing an ergodic Markov chain such that its stationary distribution coincides with the posterior distribution.
If we then sample ``sufficiently long'', $X_t$ is drawn from a distribution that is ``very close'' to the posterior distribution.

\begin{rmk}{How quickly does a Markov chain converge?}{}
  The convergence speed of Markov chains is a rich field of research.
  ``Sufficiently long'' and ``very close'' are commonly made precise by the notions of \emph{rapidly mixing} Markov chains and \emph{total variation distance}.\looseness=-1

  \begin{defn}[Total variation distance]\pidx{total variation distance}
    The total variation distance between two probability distributions $\mu$ and $\nu$ on $\spA$ is defined by \begin{align}
      \norm{\mu - \nu}_{\mathrm{TV}} \defeq 2 \sup_{\sA \subseteq \spA} \abs{\mu(\sA) - \nu(\sA)}.
    \end{align}
    It defines the distance between $\mu$ and $\nu$ to be the maximum difference between the probabilities that $\mu$ and $\nu$ assign to the same event.\looseness=-1
  \end{defn}
  As opposed to the KL-divergence \eqref{eq:kl}, the total variation distance is a metric.
  In particular, it is symmetric and satisfies the triangle inequality.
  It can be shown that \begin{align}
    \norm{\mu - \nu}_{\mathrm{TV}} \leq \sqrt{2 \KL{\mu}{\nu}} \label{eq:pinsker}
  \end{align} which is known as \midx{Pinsker's inequality}.
  Moreover, if $\mu$ and $\nu$ are discrete distributions over the set $\sS$, it can be shown that \begin{align}
    \norm{\mu - \nu}_{\mathrm{TV}} = \sum_{i \in \sS} \abs{\mu(i) - \nu(i)}.
  \end{align}

  \begin{defn}[Mixing time]
    For a Markov chain with stationary distribution $\pi$, its \midx{mixing time} with respect to the total variation distance for any $\epsilon > 0$ is \begin{align}
      \tau_{\mathrm{TV}}(\epsilon) \defeq \min\{t \mid \forall q_0 : \norm{q_t - \pi}_{\mathrm{TV}} \leq \epsilon\}.
    \end{align}
    Thus, the mixing time measures the time required by a Markov chain for the distance to stationarity to be small.
    A Markov chain is typically said to be \midx<rapidly mixing>{rapidly mixing Markov chain} if for any $\epsilon > 0$, \begin{align}
      \tau_{\mathrm{TV}}(\epsilon) \in \BigO{\mathrm{poly}(n, \log(1 / \epsilon))}.
    \end{align}
    That is, a rapidly mixing Markov chain on $n$ states needs to be simulated for at most $\mathrm{poly}(n)$ steps to obtain a ``good'' sample from its stationary distribution $\pi$.
  \end{defn}

  You can find a thorough introduction to mixing times in chapter 4 of \icite{levin2017markov}. Later chapters introduce methods for showing that a Markov chain is rapidly mixing.
\end{rmk}

\subsection{Detailed Balance Equation}

How can we confirm that the stationary distribution of a Markov chain coincides with the posterior distribution?
The detailed balance equation yields a very simple method.

\begin{defn}[Detailed balance equation / reversibility]\pidx{detailed balance equation}
  A Markov chain satisfies the \emph{detailed balance equation} with respect to a distribution~$\pi$ iff \begin{align}
    \pi(x) p(x' \mid x) = \pi(x') p(x \mid x') \label{eq:dbe}
  \end{align} holds for any $x, x' \in S$.
  A Markov chain that satisfies the detailed balance equation with respect to $\pi$ is called \midx<reversible>{reversible Markov chain} with respect to~$\pi$.
\end{defn}

\begin{lem}
  Given a finite Markov chain, if the Markov chain is reversible with respect to $\pi$ then $\pi$ is a stationary distribution.\footnote{Note that reversibility of $\pi$ is only a sufficient condition for stationarity of $\pi$, it is not necessary! In particular, there are irreversible ergodic Markov chains.}
\end{lem}\vspace{-10pt}
\begin{proof}
  Let $\pi \defeq q_t$. We have, \begin{align*}
    q_{t+1}(x) &= \sum_{x' \in S} p(x \mid x') q_t(x') \margintag{using the Markov property \eqref{eq:markov_property}} \\
    &= \sum_{x' \in S} p(x \mid x') \pi(x') \\
    &= \sum_{x' \in S} p(x' \mid x) \pi(x) \margintag{using the detailed balance equation \eqref{eq:dbe}} \\
    &= \pi(x) \sum_{x' \in S} p(x' \mid x) \\
    &= \pi(x). \qedhere \margintag{using that $\sum_{x' \in S} p(x' \mid x) = 1$}
  \end{align*}
\end{proof}

That is, if we can show that the detailed balance equation \eqref{eq:dbe} holds for some distribution $q$, then we know that $q$ is the stationary distribution of the Markov chain.

Next, reconsider our posterior distribution $p(x) = \frac{1}{Z} q(x)$ from \cref{eq:mcmc_posterior}.
If we substitute the posterior for $\pi$ in the detailed balance equation, we obtain\looseness=-1 \begin{align}
  \frac{1}{Z} q(x) p(x' \mid x) &= \frac{1}{Z} q(x') p(x \mid x'),
  \intertext{or equivalently,}
  q(x) p(x' \mid x) &= q(x') p(x \mid x').
\end{align}
In words, we do not need to know the true posterior $p$ to check that the stationary distribution of our Markov chain coincides with $p$, it suffices to know the finite measure $q$!

\subsection{Ergodic Theorem}\label{sec:approximate_inference:mcmc:ergodic_theorem}

If we now suppose that we can construct a Markov chain whose stationary distribution coincides with the posterior distribution --- we will see later that this is possible --- it is not apparent that this allows us to estimate expectations over the posterior distribution.
Note that although constructing such a Markov chain allows us to obtain samples from the posterior distribution, they are \emph{not} independent.
In fact, due to the structure of a Markov chain, by design, they are strongly dependent.
Thus, the law of large numbers and Hoeffding's inequality do not apply.
By itself, it is not even clear that an estimator relying on samples from a single Markov chain will be unbiased.

Theoretically, we could simulate many Markov chains separately and obtain one sample from each of them. This, however, is extremely inefficient.
It turns out that there is a way to generalize the (strong) law of large numbers to Markov chains.

\begin{thm}[Ergodic theorem, appendix~C of \cite{levin2017markov}]\pidx{ergodic theorem}
  Given an ergodic Markov chain $(X_t)_{t \in \Nat_0}$ over a finite state space $\sS$ with stationary distribution $\pi$ and a function $f : \sS \to \R$, \begin{align}
    \frac{1}{n} \sum_{i=1}^n f(x_i) \almostsurely \sum_{x \in S} \pi(x) f(x) = \E[x \sim \pi]{f(x)} \label{eq:ergodic_theorem}
  \end{align} as $n\to\infty$ where $x_i \sim X_i \mid x_{i-1}$.
\end{thm}

This result is the fundamental reason for why Markov chain Monte Carlo methods are possible. There are analogous results for continuous domains.

\begin{marginfigure}
  \begin{center}
    \import{./plots/output/}{burn_in_time.pgf}
  \end{center}

  \caption{Illustration of the ``burn-in'' time $t_0$ of a Markov chain approximating the posterior $p(\ys = 1 \mid \mX, \vy)$ of Bayesian logistic regression.
  The true posterior $p$ is shown in gray.
  The distribution of the Markov chain at time $t$ is shown in red.}
\end{marginfigure}

Note, however, that the ergodic theorem only tells us that simulating a single Markov chain yields an unbiased estimator.
It does not tell us anything about the rate of convergence and variance of such an estimator.
The convergence rate depends on the mixing time of the Markov chain, which is difficult to establish in general.

In practice, one observes that Markov chain Monte Carlo methods have a so-called \midx<``burn-in'' time>{burn-in time} during which the distribution of the Markov chain does not yet approximate the posterior distribution well.
Typically, the first $t_0$ samples are therefore discarded, \begin{align}
  \E{f(X)} \approx \frac{1}{T - t_0} \sum_{t=t_0+1}^T f(X_t). \label{eq:mcmc}
\end{align}
It is not clear in general how $T$ and $t_0$ should be chosen such that the estimator is unbiased, rather they have to be tuned.

Another widely used heuristic is to first find the mode of the posterior distribution and then start the Markov chain at that point.
This tends to increase the rate of convergence drastically, as the Markov chain does not have to ``walk to the location in the state space where most probability mass will be located''.

\section{Elementary Sampling Methods}

We will now examine methods for constructing and sampling from a Markov chain with the goal of approximating samples from the posterior distribution $p$.
Note that in this setting the state space of the Markov chain is $\R^n$ and a single state at time $t$ is described by the random vector $\rX \defeq [X_1, \dots, X_n]$.

\subsection{Metropolis-Hastings Algorithm}

Suppose we are given a \midx{proposal distribution} $r(\vxp \mid \vx)$ which, given we are in state $\vx$, proposes a new state $\vxp$.
Metropolis and Hastings showed that using the \midx{acceptance distribution} $\Bern{\alpha(\vxp \mid \vx)}$ where \begin{align}
  \alpha(\vxp \mid \vx) &\defeq \min \braces*{1, \frac{p(\vxp) r(\vx \mid \vxp)}{p(\vx) r(\vxp \mid \vx)}} \\
  &= \min \braces*{1, \frac{q(\vxp) r(\vx \mid \vxp)}{q(\vx) r(\vxp \mid \vx)}} \margintag{similarly to the detailed balance equation, the normalizing constant $Z$ cancels} \label{eq:mh_acc_distr}
\end{align} to decide whether to follow the proposal yields a Markov chain with stationary distribution $p(\vx) = \frac{1}{Z} q(\vx)$.

\begin{algorithm}[H]
  \caption{Metropolis-Hastings algorithm}\pidx{Metropolis-Hastings algorithm}
  initialize $\vx \in \R^n$\;
  \For{$t = 1$ \KwTo $T$}{
    sample $\vxp \sim r(\vxp \mid \vx)$\;
    sample $u \sim \Unif{[0, 1]}$\;
    \lIf{$u \leq \alpha(\vxp \mid \vx)$}{update $\vx \gets \vxp$}
    \lElse{update $\vx \gets \vx$}
  }
\end{algorithm}

Intuitively, the acceptance distribution corrects for the bias in the proposal distribution.
That is, if the proposal distribution $r$ is likely to propose states with low probability under~$p$, the acceptance distribution will reject these proposals frequently.
The following theorem formalizes this intuition.

\begin{thm}[Metropolis-Hastings theorem]\label{thm:metropolis_hastings}\pidx{Metropolis-Hastings theorem}
  Given an arbitrary proposal distribution $r$ whose support includes the support of $q$, the stationary distribution of the Markov chain simulated by the Metropolis-Hastings algorithm is $p(\vx) = \frac{1}{Z} q(\vx)$.
\end{thm}
\begin{proof}
  First, let us define the transition probabilities of the Markov chain.
  The probability of transitioning from a state $\vx$ to a state $\vxp$ is given by $r(\vxp \mid \vx) \alpha(\vxp \mid \vx)$ if $\vx \neq \vxp$ and the probability of proposing to remain in state $\vx$, $r(\vx \mid \vx)$, plus the probability of denying the proposal, otherwise. \begin{align}
    p(\vxp \mid \vx) = \begin{cases}
      r(\vxp \mid \vx) \alpha(\vxp \mid \vx) & \text{if}\ \vx \neq \vxp \\
      r(\vx \mid \vx) + \sum_{\vx'' \neq \vx} r(\vx'' \mid \vx)(1 - \alpha(\vx'' \mid \vx)) & \text{otherwise}.
    \end{cases} \label{eq:mh_trans_prob}
  \end{align}

  We will show that the stationary distribution is $p$ by showing that~$p$ satisfies the detailed balance equation \eqref{eq:dbe}.
  Let us fix arbitrary states $\vx$ and $\vxp$. First, observe that if $\vx = \vxp$, then the detailed balance equation is trivially satisfied.
  Without loss of generality we assume \begin{align*}
    \alpha(\vx \mid \vxp) = 1, \quad \alpha(\vxp \mid \vx) = \frac{q(\vxp) r(\vx \mid \vxp)}{q(\vx) r(\vxp \mid \vx)}.
  \end{align*}
  For $\vx \neq \vxp$, we then have, \begin{align*}
    p(\vx) \cdot p(\vxp \mid \vx) &= \frac{1}{Z} q(\vx) p(\vxp \mid \vx) \margintag{using the definition of the distribution $p$} \\
    &= \frac{1}{Z} q(\vx) r(\vxp \mid \vx) \alpha(\vxp \mid \vx) \margintag{using the transition probabilities of the Markov chain} \\
    &= \frac{1}{Z} q(\vx) r(\vxp \mid \vx) \frac{q(\vxp) r(\vx \mid \vxp)}{q(\vx) r(\vxp \mid \vx)} \margintag{using the definition of the acceptance distribution $\alpha$} \\
    &= \frac{1}{Z} q(\vxp) r(\vx \mid \vxp) \\
    &= \frac{1}{Z} q(\vxp) r(\vx \mid \vxp) \alpha(\vx \mid \vxp) \margintag{using the definition of the acceptance distribution $\alpha$} \\
    &= \frac{1}{Z} q(\vxp) p(\vx \mid \vxp) \margintag{using the transition probabilities of the Markov chain} \\
    &= p(\vxp) \cdot p(\vx \mid \vxp). \qedhere \margintag{using the definition of the distribution $p$}
  \end{align*}
\end{proof}

Note that by the fundamental theorem of ergodic Markov chains \eqref{eq:fund_thm_of_ergodic_mc}, for convergence to the stationary distribution, it is sufficient for the Markov chain to be ergodic.
Ergodicity follows immediately when the transition probabilities $p(\cdot \mid \vx)$ have full support. For example, if the proposal distribution $r(\cdot \mid \vx)$ has full support, the full support of $p(\cdot \mid \vx)$ follows immediately from \cref{eq:mh_trans_prob}. The rate of convergence of Metropolis-Hastings depends strongly on the choice of the proposal distribution, and we will explore different choices of proposal distribution in the following.

\subsection{Gibbs Sampling}

A popular example of a Metropolis-Hastings algorithm is Gibbs sampling as presented in \cref{alg:gibbs_sampling}.

\begin{algorithm}
  \caption{Gibbs sampling}\pidx{Gibbs sampling}\label{alg:gibbs_sampling}
  initialize $\vx = [x_1, \dots, x_n] \in \R^n$\;
  \For{$t = 1$ \KwTo $T$}{
    pick a variable $i$ uniformly at random from $\{1, \dots, n\}$\;
    set $\vx_{-i} \defeq [x_1, \dots, x_{i-1}, x_{i+1}, \dots, x_n]$\;
    update $x_i$ by sampling according to the posterior distribution $p(x_i \mid \vx_{-i})$\;
  }
\end{algorithm}

Intuitively, by re-sampling single coordinates according to the posterior distribution given the other coordinates, Gibbs sampling finds states that are successively ``more'' likely.
Selecting the index $i$ uniformly at random ensures that the underlying Markov chain is ergodic provided the conditional distributions $p(\cdot \mid \vx_{-i})$ have full support.

\begin{thm}[Gibbs sampling as Metropolis-Hastings]\label{thm:gibbs_sampling_as_mh}
  Gibbs sampling is a Metropolis-Hastings algorithm.
  For any fixed $i \in [n]$, it has proposal distribution \begin{align}
    r_i(\vxp \mid \vx) \defeq \begin{cases}
      p(x'_i \mid \vxp_{-i}) & \text{$\vxp$ differs from $\vx$ only in entry $i$} \\
      0 & \text{otherwise}
    \end{cases} \label{eq:gibbs_prop_distr}
  \end{align} and acceptance distribution $\alpha_i(\vxp \mid \vx) \defeq 1$.
\end{thm}
\begin{proof}
  We show that $\alpha_i(\vxp \mid \vx) = 1$ follows from the definition of an acceptance distribution in Metropolis-Hastings \eqref{eq:mh_acc_distr} and the choice of proposal distribution \eqref{eq:gibbs_prop_distr}.

  By \eqref{eq:mh_acc_distr}, \begin{align*}
    \alpha_i(\vxp \mid \vx) &= \min \braces*{1, \frac{p(\vxp) r_i(\vx \mid \vxp)}{p(\vx) r_i(\vxp \mid \vx)}}
  \intertext{Note that $p(\vx) = p(x_i, \vx_{-i}) = p(x_i \mid \vx_{-i}) p(\vx_{-i})$ using the product rule \eqref{eq:product_rule}. Therefore,}
    &= \min \braces*{1, \frac{p(x'_i \mid \vxp_{-i}) p(\vxp_{-i}) r_i(\vx \mid \vxp)}{p(x_i \mid \vx_{-i}) p(\vx_{-i}) r_i(\vxp \mid \vx)}} \\
    &= \min \braces*{1, \frac{p(x'_i \mid \vxp_{-i}) p(\vxp_{-i}) p(x_i \mid \vx_{-i})}{p(x_i \mid \vx_{-i}) p(\vx_{-i}) p(x'_i \mid \vxp_{-i})}} \margintag{using the proposal distribution \eqref{eq:gibbs_prop_distr}} \\
    &= \min \braces*{1, \frac{p(\vxp_{-i})}{p(\vx_{-i})}} \\
    &= 1. \qedhere \margintag{using that $p(\vxp_{-i}) = p(\vx_{-i})$}
  \end{align*}
\end{proof}

If the index $i$ is chosen uniformly at random as in \cref{alg:gibbs_sampling}, then the proposal distribution is $r(\vxp \mid \vx) = \frac{1}{n} \sum_{i=1}^n r_i(\vxp \mid \vx)$, which analogously to \cref{thm:gibbs_sampling_as_mh} has the associated acceptance distribution \begin{align*}
  \alpha(\vxp \mid \vx) = \min \braces*{1, \frac{p(\vxp) r(\vx \mid \vxp)}{p(\vx) r(\vxp \mid \vx)}} = \min \braces*{1, \frac{\sum_{i=1}^n p(\vxp) r_i(\vx \mid \vxp)}{\sum_{i=1}^n p(\vx) r_i(\vxp \mid \vx)}} = 1.
\end{align*}

\begin{cor}[Convergence of Gibbs sampling]
  As Gibbs sampling is a specific example of an MH-algorithm, the stationary distribution of the simulated Markov chain is $p(\vx)$.
\end{cor}

Note that for the proposals of Gibbs sampling, we have \begin{align}
  p(x_i \mid \vx_{-i}) = \frac{p(x_i, \vx_{-i})}{\sum_{x_i} p(x_i, \vx_{-i})} = \frac{q(x_i, \vx_{-i})}{\sum_{x_i} q(x_i, \vx_{-i})}. \margintag{using the definition of condition probability \eqref{eq:cond_prob} and the sum rule \eqref{eq:sum_rule}}
\end{align}
Under many models, this probability can be efficiently evaluated due to the conditioning on the remaining coordinates $\vx_{-i}$. If $X_i$ has finite support, the normalizer can be computed exactly.

\section{Sampling using Gradients}\label{sec:approximate_inference:mcmc:langevin_dynamics}

In this section, we discuss more advanced sampling methods.
The main idea that we will study is the interpretation of sampling as an optimization problem.
We will build towards an optimization view of sampling step-by-step, and first introduce what is commonly called the ``energy'' of a distribution.

\subsection{Gibbs Distributions}\label{sec:mcmc:uninformed:gibbs:distr}

Gibbs distributions are a special class of distributions that are widely used in machine learning, and which are characterized by an energy.

\begin{defn}[Gibbs distribution]\pidx{Gibbs distribution}
  Formally, a \emph{Gibbs distribution} (also called a \midx{Boltzmann distribution}) is a continuous distribution $p$ whose PDF is of the form \begin{align}
    p(\vx) = \frac{1}{Z} \exp(- f(\vx)).
  \end{align} $f : \R^n \to \R$ is also called an \midx{energy function}.
  When the energy function~$f$ is convex, its Gibbs distribution is called \midx<log-concave>{log-concave distribution}.
\end{defn}

A useful property is that Gibbs distributions always have full support.\footnote{This can easily be seen as $\exp(\cdot) > 0$.}
It is often easier to reason about ``energies'' rather than probabilities as they are neither restricted to be non-negative nor do they have to integrate to $1$.
Note that the Gibbs distribution belongs to the exponential family \eqref{eq:exponential_family_of_distributions} with sufficient statistic $-f(\vx)$.

Observe that the posterior distribution can always be interpreted as a Gibbs distribution as long as prior and likelihood have full support, \begin{align}
  p(\vtheta \mid \vx_{1:n}, y_{1:n}) &= \frac{1}{Z} p(\vtheta) p(y_{1:n} \mid \vx_{1:n}, \vtheta) \margintag{using Bayes' rule \eqref{eq:bayes_rule}} \nonumber \\
  &= \frac{1}{Z} \exp(-[-\log p(\vtheta) - \log p(y_{1:n} \mid \vx_{1:n}, \vtheta)]).
\end{align}
Thus, defining the energy function \begin{align}
  f(\vtheta) &\defeq -\log p(\vtheta) - \log p(y_{1:n} \mid \vx_{1:n}, \vtheta) \\
  &= -\log p(\vtheta) - \sum_{i=1}^n \log p(y_i \mid \vx_i, \vtheta),
\end{align} yields \begin{align}
  p(\vtheta \mid \vx_{1:n}, y_{1:n}) = \frac{1}{Z} \exp(- f(\vtheta)).
\end{align}
Note that $f$ coincides with the loss function used for MAP estimation \eqref{eq:map}.
For a noninformative prior, the regularization term vanishes and the energy reduces to the negative log-likelihood $\ell_\mathrm{nll}(\vtheta; \spD)$ (i.e., the loss function of maximum likelihood estimation \eqref{eq:mle}).

Using that the posterior is a Gibbs distribution, we can rewrite the acceptance distribution of Metropolis-Hastings, \begin{align}
  \alpha(\vxp \mid \vx) = \min \braces*{1, \frac{r(\vx \mid \vxp)}{r(\vxp \mid \vx)} \exp(f(\vx) - f(\vxp))}. \margintag{this is obtained by substituting the PDF of a Gibbs distribution for the posterior}
\end{align}

\begin{ex}{Metropolis-Hastings with Gaussian proposals}{metropolis_hastings_gaussian_proposals}
  Let us consider Metropolis-Hastings with the Gaussian proposal distribution, \begin{align}
    r(\vxp \mid \vx) \defeq \N[\vxp]{\vx}{\tau \mI}.
  \end{align}
  Due to the symmetry of Gaussians, we have \begin{align*}
    \frac{r(\vx \mid \vxp)}{r(\vxp \mid \vx)} = \frac{\N[\vx]{\vxp}{\tau \mI}}{\N[\vxp]{\vx}{\tau \mI}} = 1.
  \end{align*}
  Hence, the acceptance distribution is defined by \begin{align}
    \alpha(\vxp \mid \vx) = \min \braces*{1, \exp(f(\vx) - f(\vxp))}.
  \end{align}
  Intuitively, when a state with lower energy is proposed, that is ${f(\vxp) \leq f(\vx)}$, then the proposal will always be accepted.
  In contrast, if the energy of the proposed state is higher, the acceptance probability decreases exponentially in the difference of energies ${f(\vx) - f(\vxp)}$.
  Thus, Metropolis-Hastings minimizes the energy function, which corresponds to minimizing the negative log-likelihood and negative log-prior.
  The variance in the proposals~$\tau$ helps in getting around local optima, but the search direction is uniformly random (i.e., ``uninformed'').
\end{ex}

\begin{marginfigure}
  \begin{center}
    \import{./plots/output/}{langevin.pgf}
  \end{center}

  \caption{Metropolis-Hastings and Langevin dynamics minimize the energy function $f(\theta)$ shown in blue.
  Suppose we start at the black dot $\theta_0$, then the black and red arrows denote possible subsequent samples.
  Metropolis-Hastings uses an ``uninformed'' search direction, whereas Langevin dynamics uses the gradient of $f(\theta)$ to make ``more promising'' proposals.
  The random proposals help get past local optima.}
\end{marginfigure}

\subsection{From Energy to Surprise (and back)}\label{sec:mcmc:energy_surprise}

Energy-based models are a well-known class of models in machine learning where an energy function $f$ is learned from data.
These energy functions do not need to originate from a probability distribution, yet they induce a probability distribution via their Gibbs distribution $p(\vx) \propto \exp(-f(\vx))$.
As we will see in \cref{exercise:maximum_entropy_property_of_gibbs_distribution}, this Gibbs distribution is the associated maximum entropy distribution.
Observe that the surprise about $\vx$ under this distribution is given by \begin{align}
  \S{p(\vx)} = f(\vx) + \log Z.
\end{align}
That is, up to a constant shift, the energy of $\vx$ \emph{coincides} with the surprise about $\vx$.
Energies are therefore sufficient for comparing the ``likelihood'' of points, and they do not require normalization.\footnote{Intuitively, an energy can be used to compare the ``likelihood'' of two points $\vx$ and $\vxp$ whereas the probability $\vx$ makes a statement about the ``likelihood'' of $\vx$ relative to \emph{all} other points.}

What kind of energies could we use?
In \cref{sec:mcmc:uninformed:gibbs:distr}, we discussed the use of the negative log-posterior or negative log-likelihood as energies.
In general, \emph{any} loss function $\ell(\vx)$ can be thought of as an energy function with an associated maximum entropy distribution $p(\vx) \propto \exp(-\ell(\vx))$.

\subsection{Langevin Dynamics}

Until now, we have looked at Metropolis-Hastings algorithms with proposal distributions that do not explicitly take into account the curvature of the energy function around the current state. Langevin dynamics adapts the Gaussian proposals of the Metropolis-Hastings algorithm we have seen in \cref{ex:metropolis_hastings_gaussian_proposals} to search the state space in an ``informed'' direction.
The simple idea is to bias the sampling towards states with lower energy, thereby making it more likely that a proposal is accepted.

A natural idea is to shift the proposal distribution perpendicularly to the gradient of the energy function.
This yields the following proposal distribution, \begin{align}
  r(\vxp \mid \vx) = \N[\vxp]{\vx - \eta_t \grad f(\vx)}{2 \eta_t \mI}. \label{eq:mala}
\end{align}
The resulting variant of Metropolis-Hastings is known as the \midx{Metropolis adjusted Langevin algorithm} (MALA) or \midx{Langevin Monte Carlo} (LMC).
It can be shown that, as $\eta_t \to 0$, we have for the acceptance probability $\alpha(\vxp \mid \vx) \to 1$ using that the acceptance probability is $1$ if $\vxp = \vx$.
Hence, the Metropolis-Hastings acceptance step can be omitted once the rejection probability becomes negligible.
The algorithm which always accepts the proposal of \cref{eq:mala} is known as the \midx{unadjusted Langevin algorithm} (ULA).

Observe that if the stationary distribution is the posterior distribution \begin{align*}
  p(\vtheta \mid \vx_{1:n}, y_{1:n}) = \frac{1}{Z} \exp\underbrace{\parentheses*{\log p(\vtheta) + \sum_{i=1}^n \log p(y_i \mid \vx_i, \vtheta)}}_{-f(\vtheta)}
\end{align*} with energy $f$ as we discussed in \cref{sec:mcmc:uninformed:gibbs:distr}, then the proposal $\vtheta'$ of MALA/LMC can be equivalently formulated as \begin{align}\begin{split}
  \vtheta' &\gets \vtheta - \eta_t \grad f(\vtheta) + \vvarepsilon \\
  &= \vtheta + \eta_t \parentheses*{\grad \log p(\vtheta) + \sum_{i=1}^n \grad \log p(y_i \mid \vx_i, \vtheta)} + \vvarepsilon \label{eq:lmc}
\end{split}\end{align} where $\vvarepsilon \sim \N{\vzero}{2\eta_t\mI}$.

\begin{rmk}{ULA is a discretized diffusion}{wiener_process}
  The unadjusted Langevin algorithm can be seen as a discretization of Langevin dynamics, which is a continuous-time stochastic process with a drift and with random stationary and independent Gaussian increments.
  The randomness is modeled by a Wiener process.

  \begin{defn}[Wiener process]
    The \midx{Wiener process}[idxpagebf] (also known as \midx{Brownian motion}) is a sequence of random vectors $\{\rW_t\}_{t \geq 0}$ such that\looseness=-1 \begin{enumerate}
      \item $\rW_0 = \vzero$,
      \item with probability $1$, $\rW_t$ is continuous in $t$,
      \item the process has independent increments,\safefootnote{That is, the ``future'' increments $\rW_{t+u} - \rW_t$ for $u \geq 0$ are independent of past values $\rW_s$ for $s < t$.} and
      \item $\rW_{t+u} - \rW_t \sim \N{\vzero}{u \mI}$.
    \end{enumerate}
  \end{defn}

  Consider the continuous-time stochastic process $\vtheta$ defined by the stochastic differential equation (SDE) \begin{align}
    d \vtheta_t = \underbrace{- \grad f(\vtheta_t) \,d t}_{\text{drift}} + \underbrace{\sqrt{2} \,d \rW_t}_{\text{noise}}, \label{eq:cont_time_langevin}
  \end{align}
  Such a stochastic process is also called a \midx<diffusion (process)>{diffusion process} and \cref{eq:cont_time_langevin} specifically is called \midx{Langevin dynamics}.
  Here, the first term is called the ``drift'' of the process, and the second term is called its ``noise''.
  Note that if the noise term is zero then \cref{eq:cont_time_langevin} is simply an ordinary differential equation (ODE).

  A diffusion can be discretized using the Euler-Maruyama method (also called ``forward Euler'') to obtain a discrete approximation $\vtheta_k \approx \vtheta(\tau_k)$ where $\tau_k$ denotes the $k$-th time step.
  Choosing the time steps such that $\Delta t_k \defeq \tau_{k+1} - \tau_{k} = \eta_k$ yields the approximation \begin{align}
    \vtheta_{k+1} = \vtheta_k - \grad f(\vtheta_k) \,\Delta t_k + \sqrt{2} \,\Delta \rW_k
  \end{align} where $\Delta \rW_k \defeq \rW_{\tau_{k+1}} - \rW_{\tau_k} \sim \N{\vzero}{\Delta t_k \mI}$.
  Observe that this coincides with the update rule of Langevin dynamics from \cref{eq:lmc}.
\end{rmk}

The appearance of drift and noise is closely related to the \midx{principle of curiosity and conformity}, we encountered in the previous chapter on variational inference.
The noise term (also called the \midx{diffusion}) leads to exploration of the state space (i.e., curiosity about alternative explanations of the data), whereas the drift term (also called the \midx{distillation}) leads to minimizing the energy or loss (i.e., conformity to the data).
Interestingly, this same principle appears in both variational inference and MCMC, two very different approaches to approximate probabilistic inference.
In the remainder of this manuscript we will find this to be a reoccurring theme.

For log-concave distributions, the mixing time of the Markov chain underlying Langevin dynamics can be shown to be polynomial in the dimension $n$ \citep{vempala2019rapid}.
You will prove this result for strongly log-concave distributions in \exerciserefmark{langevin_dynamics_convergence} and see that the analysis is analogous to the canonical convergence analysis of classical optimization schemes.

\subsection{Stochastic Gradient Langevin Dynamics}

Note that computing the gradient of the energy function, which corresponds to computing exact gradients of the log-prior and log-likelihood, in every step can be expensive.
The proposal step of MALA/LMC can be made more efficient by approximating the gradient with an unbiased gradient estimate, leading to \emph{stochastic gradient Langevin dynamics} (SGLD) shown in \cref{alg:sgld} \citep{welling2011bayesian}.
Observe that SGLD \eqref{eq:sgld} differs from MALA/LMC \eqref{eq:lmc} only by using a sample-based approximation of the gradient.

\begin{algorithm}
  \caption{Stochastic gradient Langevin dynamics, SGLD}\pidx{stochastic gradient Langevin dynamics}\label{alg:sgld}
  initialize $\vtheta$\;
  \For{$t = 1$ \KwTo $T$}{
    sample $i_1, \dots, i_m \sim \Unif{\{1, \dots, n\}}$ independently\;
    sample $\vvarepsilon \sim \N{\vzero}{2 \eta_t \mI}$\;
    $\vtheta \gets \vtheta + \eta_t \parentheses*{\grad \log p(\vtheta) + \frac{n}{m} \sum_{j=1}^m \grad \log p(y_{i_j} \mid \vx_{i_j}, \vtheta)} + \vvarepsilon$ \algeq{eq:sgld}
  }
\end{algorithm}

Intuitively, in the initial phase of the algorithm, the stochastic gradient term dominates, and therefore, SGLD corresponds to a variant of stochastic gradient ascent. In the later phase, the update rule is dominated by the injected noise $\vvarepsilon$, and will effectively be Langevin dynamics. SGLD transitions smoothly between the two phases.

Under additional assumptions, SGLD is guaranteed to converge to the posterior distribution for decreasing learning rates ${\eta_t = \Theta(t^{-\nicefrac{1}{3}})}$ \citep{raginsky2017non,xu2018global}.
SGLD does not use the acceptance step from Metropolis-Hastings as asymptotically, SGLD corresponds to Langevin dynamics and the Metropolis-Hastings rejection probability goes to zero for a decreasing learning rate.

\subsection{Hamiltonian Monte Carlo}

As MALA and SGLD can be seen as a sampling-based analogue of GD and SGD, a similar analogue for (stochastic) gradient descent with momentum is the (stochastic gradient) \midx{Hamiltonian Monte Carlo} (HMC) algorithm, which we discuss in the following \citep{duane1987hybrid,sghmc}.\looseness=-1

\begin{figure}
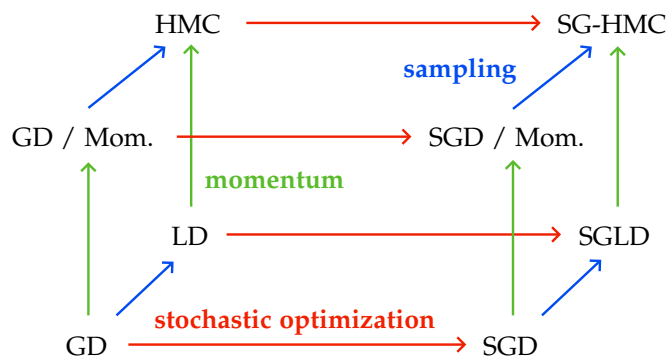

  \incfig{optimization_algorithms}
  \caption{A commutative diagram of sampling and optimization algorithms.
  Langevin dynamics (LD) is the non-stochastic variant of SGLD.}
\end{figure}

We have seen that if we want to sample from a distribution \begin{align*}
  p(\vx) \propto \exp(-f(\vx))
\end{align*} with energy function $f$, we can construct a Markov chain whose distribution converges to $p$.
We have also seen that for this approach to work, the chain must move through all areas of significant probability with reasonable speed.

If one is faced with a distribution $p$ which is multimodal (i.e., that has several ``peaks''), one has to ensure that the chain will explore all modes, and can therefore ``jump between different areas of the space''.

So in general, \emph{local} updates are doomed to fail.
Methods such as Metropolis-Hastings with Gaussian proposals, or even Langevin Monte Carlo might face this issue, as they do not jump to distant areas of the state space with significant acceptance probability.
It will therefore take a long time to move from one peak to another.

The HMC algorithm is an instance of Metropolis-Hastings which uses momentum to propose distant points that conserve energy, with high acceptance probability.
The general idea of HMC is to \emph{lift} samples $\vx$ to a higher-order space by considering an auxiliary variable $\vy$ with the same dimension as $\vx$.
We also lift the distribution $p$ to a distribution on the $(\vx, \vy)$-space by defining a distribution $p(\vy \mid \vx)$ and setting $p(\vx, \vy) \defeq p(\vy \mid \vx) p(\vx)$.
It is common to pick $p(\vy \mid \vx)$ to be a Gaussian with zero mean and variance $m\mI$.
Hence, \begin{align}
  p(\vx, \vy) \propto \exp\parentheses*{-\frac{1}{2m}\norm{\vy}_2^2 - f(\vx)}.
\end{align}
Physicists might recognize the above as the canonical distribution of a Newtonian system if one takes $\vx$ as the position and $\vy$ as the momentum.
$H(\vx, \vy) \defeq \frac{1}{2m}\norm{\vy}_2^2 + f(\vx)$ is called the \midx{Hamiltonian}.
HMC then takes a step in this higher-order space according to the Hamiltonian dynamics,\footnote{That is, HMC follows the trajectory of these dynamics for some time.}\looseness=-1 \begin{align}
  \odv{\vx}{t} = \grad_\vy H, \quad \odv{\vy}{t} = - \grad_\vx H, \label{eq:hamiltonian_dynamics}
\end{align} reaching some new point $(\vxp, \vyp)$ and \emph{projecting} back to the state space by selecting $\vxp$ as the new sample.
This is illustrated in \cref{fig:hmc}.
In the next iteration, we resample the momentum $\vyp \sim p(\cdot \mid \vxp)$ and repeat the procedure.

\begin{figure}
  \begin{center}
    \import{./plots/output/}{hmc.pgf}
  \end{center}

  \caption{Illustration of Hamiltonian Monte Carlo. Shown is the contour plot of a distribution $p$, which is a mixture of two Gaussians, in the $(x,y)$-space.

  First, the initial point in the state space is lifted to the $(x,y)$-space. Then, we move according to Hamiltonian dynamics and finally project back onto the state space.}\label{fig:hmc}
\end{figure}

In an implementation of this algorithm, one has to solve \cref{eq:hamiltonian_dynamics} numerically rather than exactly.
Typically, this is done using the \midx{Leapfrog method}, which for a step size $\tau$ computes \begin{subequations}\begin{align}
  \vy(t + \nicefrac{\tau}{2}) &= \vy(t) - \frac{\tau}{2} \grad_\vx f(\vx(t)) \label{eq:leapfrog_y1} \\
  \vx(t + \tau) &= \vx(t) + \frac{\tau}{m} \vy(t + \nicefrac{\tau}{2}) \label{eq:leapfrog_x} \\
  \vy(t + \tau) &= \vy(t + \nicefrac{\tau}{2}) - \frac{\tau}{2} \grad_\vx f(\vx(t + \tau)).
\end{align}\label{eq:leapfrog}\end{subequations}
Then, one repeats this procedure $L$ times to arrive at a point $(\vxp, \vyp)$.
To correct for the resulting discretization error, the proposal is either accepted or rejected in a final Metropolis-Hastings acceptance step.
If the proposal distribution is symmetric (which we will confirm in a moment), the acceptance probability is \begin{align}
  \alpha((\vxp, \vyp) \mid (\vx, \vy)) \defeq \min\{1, \exp(H(\vxp, \vyp) - H(\vx, \vy))\}. \label{eq:hmc_acceptance_prob}
\end{align}
It follows that $p(\vx, \vy)$ is the stationary distribution of the Markov chain underlying HMC.
Due to the independence of $\vx$ and $\vy$, this also implies that the projection to $\vx$ yields a Markov chain with stationary distribution $p(\vx)$.

So why is the proposal distribution symmetric?
This follows from the time-reversibility of Hamiltonian dynamics.
It is straightforward to check that the dynamics from \cref{eq:hamiltonian_dynamics} are identical if we replace $t$ with $-t$ and $\vy$ with $-\vy$.
Intuitively, unlike the position $\vx$, the momentum $\vy$ is reversed when time is reversed as it depends on the velocity which is the time-derivative of the position.\footnote{The momentum of the $i$-th coordinate is $y_i = m_i v_i$ where $m_i$ is the mass and $v_i = \odv{x_i}{t}$ is the velocity.}
In simpler terms, time-reversibility states that if we observe the evolution of a system (e.g., two billiard balls colliding), we cannot distinguish whether we are observing the system evolve forward or backward in time.
The Leapfrog method maintains the time-reversibility of the dynamics.

Symmetry of the proposal distribution is ensured by proposing the point $(\vxp, -\vyp)$.\footnote{More formally, the proposal distribution is the Dirac delta at $(\vxp, -\vyp)$.}
Intuitively, this is simply to ensure that the system is run backward in time as often as it is run forward in time.
Recall that the momentum is resampled before each iteration (i.e., the proposed momentum is ``discarded'') and observe that $p(\vxp, -\vyp) = p(\vxp, \vyp)$,\footnote{We use here that $p(\vy \mid \vx)$ was chosen to be symmetric around zero.} so we can safely disregard the direction of time when computing the acceptance probability in \cref{eq:hmc_acceptance_prob}.

\section*{Discussion}

To summarize, Markov chain Monte Carlo methods use a Markov chain to approximately sample from an intractable distribution.
Note that unlike for variational inference, the convergence of many methods can be guaranteed.
Moreover, for log-concave distributions (e.g., with Bayesian logistic regression), the underlying Markov chain converges quickly to the stationary distribution.
Methods such as Langevin dynamics and Hamiltonian Monte Carlo aim to accelerate mixing by proposing points with a higher acceptance probability than Metropolis-Hastings with ``undirected'' Gaussian proposals.
Nevertheless, in general, the convergence (mixing time) may be slow, meaning that, in practice, accuracy and efficiency have to be traded.

\begin{oreadings}
  \begin{itemize}
    \item \pcite{ma2019sampling}
    \item \pcite{teh2016consistency}
    \item \pcite{sghmc}
  \end{itemize}
\end{oreadings}

\excheading

\begin{nexercise}{Markov chain update}{mc_update}
  Prove \cref{eq:mc_update}, i.e., that one iteration of the Markov chain can be expressed as ${\vq_{t+1} = \vq_t \mP}$.
\end{nexercise}

\begin{nexercise}{$k$-step transitions}{mc_multi_step_transitions}
  Prove that the entry $\mP^k(x,x')$ corresponds to the probability of transitioning from state $x \in \sS$ to state $x' \in \sS$ in exactly $k$ steps.
\end{nexercise}

\begin{nexercise}{Finding stationary distributions}{finding_stationary_distributions}
  A news station classifies each day as ``good'', ``fair'', or ``poor'' based on its daily ratings which fluctuate with what is occurring in the news.
  Moreover, the following table shows the probabilistic relationship between the type of the current day and the probability of the type of the next day conditioned on the type of the current day.

  \vspace{5pt}
  \begin{center}
    \begin{tabular}{ll|ccc}
      \toprule
      && \multicolumn{3}{c}{next day} \\
      && good & fair & poor \\
      \midrule
      \multirow{3}{*}{current day} & good & $0.60$ & $0.30$ & $0.10$ \\
      & fair & $0.50$ & $0.25$ & $0.25$ \\
      & poor & $0.20$ & $0.40$ & $0.40$ \\
      \bottomrule
    \end{tabular}
  \end{center}
  \vspace{5pt}

  In the long run, what percentage of news days will be classified as ``good''?\looseness=-1
\end{nexercise}

\begin{nexercise}{Example of Metropolis-Hastings}{metropolis_hastings}
  Consider the state space $\{0,1\}^n$ of binary strings having length $n$.
  Let the proposal distribution be $r(x' \mid x) = 1/n$ if $x'$ differs from $x$ in exactly one bit and $r(x' \mid x) = 0$ otherwise.
  Suppose we desire a stationary distribution $p$ for which $p(x)$ is proportional to the number of ones that occur in the bit string $x$.
  For example, in the long run, a random walk should visit a string having five $1$s five times as often as it visits a string having only a single $1$.
  Provide a general formula for the acceptance probability $\alpha(x' \mid x)$ that would be used if we were to obtain the desired stationary distribution used the Metropolis-Hastings algorithm.
\end{nexercise}

\begin{marginbox}[30\baselineskip]{Gamma distribution}
  The PDF of the \midx{gamma distribution} $\GammaDistr{\alpha}{\beta}$ is defined as \begin{align*}
    \GammaDistr[x]{\alpha}{\beta} \propto x^{\alpha-1} e^{- \beta x}, \quad x \in \R_{>0}.
  \end{align*}
  A random variable $X \sim \GammaDistr{\alpha}{\beta}$ measures the waiting time until $\alpha > 0$ events occur in a Poisson process with rate $\beta > 0$.
  In particular, when $\alpha = 1$ then the gamma distribution coincides with the exponential distribution with rate $\beta$.
\end{marginbox}

\begin{nexercise}{Practical examples of Gibbs sampling}{gibbs_sampling}
  In this exercise, we look at some examples where Gibbs sampling is useful.\looseness=-1

  \begin{enumerate}
    \item Consider the distribution \begin{align*}
      p(x, y) \defeq {n \choose x}y^{x + \alpha - 1}(1-y)^{n - x + \beta - 1}, \quad x \in [n], y \in [0,1].
    \end{align*}
    Convince yourself that it is hard to sample directly from $p$ and prove that it is an easy task if one uses Gibbs sampling. That is, show that the conditional distributions $p(x \mid y)$ and $p(y \mid x)$ are easy to sample from.\par
    \textit{Hint: Take a look at the Beta distribution \eqref{eq:beta_distr}.}

    \item Consider the following generative model $p(\mu, \lambda, x_{1:n})$ given by the likelihood $x_{1:n} \mid \mu, \lambda \iid \N{\mu}{\inv{\lambda}}$ and the independent priors \begin{align*}
      \mu \sim \N{\mu_0}{\inv{\lambda_0}} \quad\text{and}\quad \lambda \sim \GammaDistr{\alpha}{\beta}.
    \end{align*}
    We would like to sample from the posterior $p(\mu, \lambda \mid x_{1:n})$.
    Show that \begin{align*}
      \mu \mid \lambda, x_{1:n} \sim \N{m_\lambda}{\inv{l_\lambda}} \quad\text{and}\quad \lambda \mid \mu, x_{1:n} \sim \GammaDistr{a_\mu}{b_\mu},
    \end{align*} and derive $m_\lambda, l_\lambda, a_\mu, b_\mu$.
    Such a prior is called a \midx{semi-conjugate prior} to the likelihood, as the prior on $\mu$ is conjugate for any fixed value of $\lambda$ and vice-versa.

    \item Let us assume that $x_{1:n} \mid \alpha, c \iid \Pareto{\alpha}{c}$ and assume the improper prior $p(\alpha, c) \propto \Ind{\alpha, c > 0}$ which corresponds to a noninformative prior.
    Derive the posterior $p(\alpha, c \mid x_{1:n})$.
    Then, also derive the conditional distributions $p(\alpha \mid c, x_{1:n})$ and $p(c \mid \alpha, x_{1:n})$, and observe that they correspond to known distributions / are easy to sample from.
  \end{enumerate}
\end{nexercise}

\begin{nexercise}{Energy function of Bayesian logistic regression}{bayesian_logistic_regression_energy_function}
  Recall from \cref{eq:logistic_regression} that the energy function of Bayesian logistic regression is \begin{align}
    f(\vw) = \lambda \norm{\vw}_2^2 + \sum_{i=1}^n \log(1 + \exp(-y_i \transpose{\vw} \vx_i)),
  \end{align} which coincided with the standard optimization objective of (regularized) logistic regression.

  Show that the posterior distribution of Bayesian logistic regression is log-concave.\looseness=-1
\end{nexercise}

\begin{nexercise}{Maximum entropy property of Gibbs distribution}{maximum_entropy_property_of_gibbs_distribution}
  \begin{enumerate}
    \item Let $X$ be a random variable supported on the finite set $\spT \subset \R$.\footnote{The same result can be shown to hold for arbitrary compact subsets.}
    Show that the Gibbs distribution with energy function $\frac{1}{T} f(x)$ for some temperature scalar $T \in \R$ is the distribution with maximum entropy of all distributions supported on $\spT$ that satisfy the constraint $\E{f(X)} < \infty$.\par
    \textit{Hint: Solve the dual problem (analogously to \cref{exercise:mep_and_posteriors}).}

    \item What happens for $T \to \{0, \infty\}$?
  \end{enumerate}
\end{nexercise}

\begin{nexercise}{Energy reduction of Gibbs sampling}{gibbs_sampling_energy_reduction}
  Let $p(\vx)$ be a probability density over $\mathbb{R}^d$, which we want to sample from.
  Assume that $p$ is a Gibbs distribution with energy function $f : \mathbb{R}^d \to \mathbb{R}$.

  In this exercise, we will study a single round of Gibbs sampling with initial state $\vx$ and final state $\vxp$ where \begin{align*}
    x'_j = \begin{cases}
      x'_i & \text{if $j = i$} \\
      x_j & \text{otherwise} \\
    \end{cases}
  \end{align*} for some fixed index $i$ and $x'_i \sim p(\cdot \mid \vx_{-i})$.

  Show that \begin{align}
    \E[x'_i \sim p(\cdot \mid \vx_{-i})]{f(\vxp)} \leq f(\vx) - \S{p(x_i \mid \vx_{-i})} + \H{p(\cdot \mid \vx_{-i})}.
  \end{align}
  That is, the energy is expected to decrease if the surprise of $x_i$ given $\vx_{-i}$ is larger than the expected surprise of the new $x'_i$ given~$\vx_{-i}$, i.e., $\S{p(x_i \mid \vx_{-i})} \geq \H{p(\cdot \mid \vx_{-i})}$.

  \textit{Hint: Recall the framing of Gibbs sampling as a variant of Metropolis-Hastings and relate this to the acceptance distribution of Metropolis-Hastings when $p$ is a Gibbs distribution.}
\end{nexercise}

\begin{nexercise}{Mixing time of Langevin dynamics}{langevin_dynamics_convergence}
  In this exercise, we will show that for certain Gibbs distributions, ${p(\vtheta) \propto \exp(-f(\vtheta))}$, Langevin dynamics is rapidly mixing.
  To do this, we will observe that Langevin dynamics can be seen as a continuous-time optimization algorithm in the space of distributions.

  First, we consider a simpler and more widely-known optimization algorithm, namely the \midx{gradient flow} \begin{align}
    d \vx_t = - \grad f(\vx_t) \,d t. \label{eq:gradient_flow}
  \end{align}
  Note that gradient descent is simply the discrete-time approximation of gradient flow just as ULA is the discrete-time approximation of Langevin dynamics.
  In the analysis of ODEs such as the gradient flow, so-called \midx<Lyapunov functions>{Lyapunov function} are commonly used to prove convergence of $\vx_t$ to a fixed point (also called an \midx{equilibrium}).

  Let us assume that $f$ is $\alpha$-strongly convex for some $\alpha > 0$, that is, \begin{align}
    f(\vy) \geq f(\vx) + \transpose{\grad f(\vx)}(\vy - \vx) + \frac{\alpha}{2} \norm{\vy - \vx}_2^2 \quad \forall \vx, \vy \in \R^n. \label{eq:strongly_convex}
  \end{align}
  In words, $f$ is lower bounded by a quadratic function with curvature~$\alpha$.
  Moreover, assume w.l.o.g. that $f$ minimized at $f(\vzero) = 0$.\footnote{This can always be achieved by shifting the coordinate system and subtracting a constant from $f$.}
  \begin{enumerate}
    \item Show that $f$ satisfies the \midx<Polyak-Łojasiewicz (PL) inequality>{Polyak-Łojasiewicz inequality}, i.e., \begin{align}
      f(\vx) \leq \frac{1}{2 \alpha} \norm{\grad f(\vx)}_2^2 \quad \forall \vx \in \R^n. \label{eq:polyak_lojasiewicz}
    \end{align}

    \item Prove $\odv{}{t} f(\vx_t) \leq - 2 \alpha f(\vx_t)$.
  \end{enumerate}
  Thus, $\vzero$ is the fixed point of \cref{eq:gradient_flow} and the Lyapunov function~$f$ is monotonically decreasing along the trajectory of $\vx_t$.
  We recall \midx{Grönwall's inequality} which states that for any real-valued continuous functions $g(t)$ and $\beta(t)$ on the interval $[0, T] \subset \R$ such that ${\odv{}{t} g(t) \leq \beta(t) g(t)}$ for all $t \in [0, T]$ we have \begin{align}
    g(t) \leq g(0) \exp\parentheses*{\int_0^t \beta(s) \,d s} \quad \forall t \in [0, T]. \label{eq:groenwall}
  \end{align}
  \begin{enumerate}
    \setcounter{enumi}{2}
    \item Conclude that $f(\vx_t) \leq e^{-2 \alpha t} f(\vx_0)$.
  \end{enumerate} \vspace{\baselineskip}

  Now that we have proven the convergence of gradient flow using $f$ as Lyapunov function, we will follow the same template to prove the convergence of Langevin dynamics to the distribution $p(\vtheta) \propto \exp(-f(\vtheta))$.
  We will use that the evolution of $\{\vtheta_t\}_{t \geq 0}$ following the Langevin dynamics \eqref{eq:cont_time_langevin} is equivalently characterized by their densities $\{q_t\}_{t \geq 0}$ following the \midx{Fokker-Planck equation} \begin{align}
    \pdv{q_t}{t} = \dive (q_t \grad f) + \lapl q_t. \label{eq:fokker_planck}
  \end{align}
  Here, $\dive$ and $\lapl$ are the divergence and Laplacian operators, respectively.\footnote{For ease of notation, we omit the explicit dependence of $q_t$, $p$, and $f$ on $\vtheta$.}
  Intuitively, the first term of the Fokker-Planck equation corresponds to the drift and its second term corresponds to the diffusion~(i.e., the Gaussian noise).

  \begin{rmk}{Intuition on vector calculus}{}
    Recall that the divergence $\dive \vF$ of a vector field $\vF$ measures the change of volume under the flow of $\vF$. That is, if in the small neighborhood of a point $\vx$, $\vF$ points towards $\vx$, then the divergence at $\vx$ is negative as the volume shrinks. If $\vF$ points away from $\vx$, then the divergence at $\vx$ is positive as the volume increases. \\
    The Laplacian $\lapl \varphi = \dive (\grad \varphi)$ of a scalar field $\varphi$ can be understood intuitively as measuring ``heat dissipation''. That is, if $\varphi(\vx)$ is smaller than the average value of $\varphi$ in a small neighborhood of $\vx$, then the Laplacian at $\vx$ is positive.

    Regarding the Fokker-Planck equation \eqref{eq:fokker_planck}, the second term $\lapl q_t$ can therefore be understood as locally dissipating the probability mass of $q_t$ (which is due to the diffusion term in the SDE).
    On the other hand, the term $\dive (q_t \grad f)$ can be understood as a Laplacian of $f$ ``weighted'' by $q_t$. Intuitively, the vector field $\grad f$ moves flow in the direction of high energy, and hence, its divergence is larger in regions of lower energy and smaller in regions of higher energy. This term therefore corresponds to a drift from regions of high energy to regions of low energy.
  \end{rmk}\vspace{2ex}

  \begin{enumerate}
    \setcounter{enumi}{3}
    \item Show that $\lapl q_t = \dive (q_t \grad \log q_t)$, implying that the Fokker-Planck equation simplifies to \begin{align}
      \pdv{q_t}{t} = \dive \parentheses*{q_t \grad \log \frac{q_t}{p}}. \label{eq:simp_fokker_planck}
    \end{align}
    \textit{Hint: The Laplacian of a scalar field $\varphi$ is $\lapl \varphi \defeq \dive (\grad \varphi)$.}
  \end{enumerate}
  Observe that the Fokker-Planck equation already implies that $p$ is indeed a stationary distribution, as if $q_t = p$ then $\pdv{q_t}{t} = 0$.
  Moreover, note the similarity of the integrand of $\KL{q_t}{p}$, $q_t \log \frac{q_t}{p}$, to \cref{eq:simp_fokker_planck}.
  We will therefore use the KL-divergence as Lyapunov function.
  \begin{enumerate}
    \setcounter{enumi}{4}
    \item Prove $\odv{}{t} \KL{q_t}{p} = - \Fisher{q_t}{p}$.
    Here, \begin{align}
      \Fisher{q_t}{p} \defeq \E[\vtheta \sim q_t]{\norm{\grad \log \frac{q_t(\vtheta)}{p(\vtheta)}}_2^2}
    \end{align} denotes the \midx{relative Fisher information} of $p$ with respect to $q_t$.
    \textit{Hint: For any distribution $q$ on $\R^n$, \begin{align}
      \int_{\R^n} (\dive q \vF) \varphi \,d \vx = - \int_{\R^n} q \; \grad \varphi \cdot \vF \,d \vx \label{eq:divergence_theorem_hint}
    \end{align} follows for any vector field $\vF$ and scalar field $\varphi$ from the divergence theorem and the product rule of the divergence operator.}
  \end{enumerate}
  Thus, the relative Fisher information can be seen as the negated time-derivative of the KL-divergence, and as $\Fisher{q_t}{p} \geq 0$ it follows that the KL-divergence is decreasing along the trajectory.

  The \midx{log-Sobolev inequality} (LSI) is satisfied by a distribution $p$ with a constant $\alpha > 0$ if for all $q$: \begin{align}
    \KL{q}{p} \leq \frac{1}{2 \alpha} \Fisher{q}{p}.
  \end{align}
  It is a classical result that if $f$ is $\alpha$-strongly convex then $p$ satisfies the LSI with constant $\alpha$ \citep{bakry2006diffusions}.
  \begin{enumerate}
    \setcounter{enumi}{5}
    \item Show that if $f$ is $\alpha$-strongly convex for some $\alpha > 0$ (we say that $p$ is ``strongly log-concave''), then $\KL{q_t}{p} \leq e^{-2 \alpha t} \KL{q_0}{p}$.
    \item Conclude that under the same assumption on $f$, Langevin dynamics is rapidly mixing, i.e., $\tau_{\mathrm{TV}}(\epsilon) \in \BigO{\mathrm{poly}(n, \log(1 / \epsilon))}$.
  \end{enumerate}

  To summarize, we have seen that Langevin dynamics is an optimization scheme in the space of distributions, and that its convergence can be analyzed analogously to classical optimization schemes.
  Notably, in this exercise we have studied continuous-time Langevin dynamics.
  Convergence guarantees for discrete-time approximations can be derived using the same techniques.
  If this interests you, refer to \icite{vempala2019rapid}.
\end{nexercise}

\begin{nexercise}{Hamiltonian Monte Carlo}{hmc}
  \begin{enumerate}
    \item Prove that if the dynamics are solved exactly (as opposed to numerically using the Leapfrog method), then the acceptance probability of the MH-step is always $1$.
    \item Prove that the Langevin Monte Carlo algorithm from \eqref{eq:mala} can be seen as a special case of HMC if only one Leapfrog step is used ($L = 1$) and $m = 1$.
  \end{enumerate}
\end{nexercise}

  \chapter{Deep Learning}\label{sec:bdl}

We began our journey through probabilistic machine learning with linear models.
In most practical applications, however, it is seen that models perform better when labels may nonlinearly depend on the inputs, and for this reason linear models are often used in conjunction with ``hand-designed'' nonlinear features.
Designing these features is costly, time-consuming, and requires expert knowledge.
Moreover, the design of such features is inherently limited by human comprehension and ingenuity.

\section{Artificial Neural Networks}

One widely used family of nonlinear functions
are \midx<artificial ``deep'' neural networks>{neural network},\footnote{In the following, we will refrain from using the characterizations ``artificial'' and ``deep'' for better readability.}
\begin{align}
  \vf : \R^d \to \R^k, \quad \vf(\vx; \vtheta) \defeq \vvarphi(\mW_L \vvarphi(\mW_{L-1} ( \cdots \vvarphi(\mW_1 \vx))))
\end{align} where $\vtheta \defeq [\mW_1, \dots, \mW_L]$ is a vector of \emph{weights} (written as matrices ${\mW_l \in \R^{n_l \times n_{l-1}}}$)\footnote{where $n_0 = d$ and $n_L = k$} and $\varphi : \R \to \R$ is a component-wise nonlinear function.
Thus, a deep neural network can be seen as nested (``deep'') linear functions composed with nonlinearities.
This simple kind of neural network is also called a \midx{multilayer perceptron}.

In this chapter, we will be focusing mostly on performing probabilistic inference with a \emph{given} neural network architecture.
To this end, understanding the basic architecture of a multilayer perceptron will be sufficient for us.
For a more thorough introduction to the field of deep learning and various architectures, you may refer to the books of \cite{goodfellow2016deep} and \cite{prince2023understanding}.

\begin{figure}
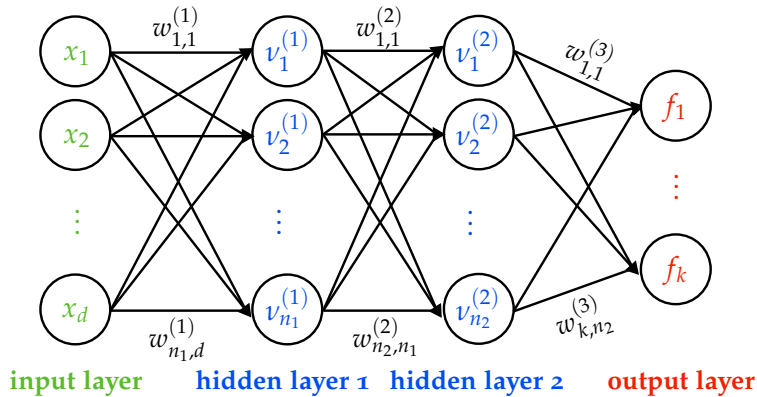

  \incfig{ann}
  \caption{Computation graph of a neural network with two hidden layers.}\label{fig:computation_graph}
\end{figure}

A neural network can be visualized by a \midx{computation graph}.
An example for such a computation graph is given in \cref{fig:computation_graph}.
The columns of the computation graph are commonly called \midx<layers>{layer}, whereby the left-most column is the \midx{input layer}, the right-most column is the \midx{output layer}, and the remaining columns are the \midx<hidden layers>{hidden layer}.
The inputs are (as we have previously) referred to as $\vx \defeq [x_1, \dots, x_d]$.
The outputs (i.e., vertices of the output layer) are often referred to as \midx{logits} and named $\vf \defeq [f_1, \dots, f_k]$.
The \midx<activations>{activation} of an individual (hidden) layer $l$ of the neural network are described by \begin{align}
  \vnu^{(l)} \defeq \vvarphi(\mW_l \vnu^{(l-1)})
\end{align} where $\vnu^{(0)} = \vx$.
The activation of the $i$-th node is $\nu_i^{(l)} = \vnu^{(l)}(i)$.

\subsection{Activation Functions}

The nonlinearity $\varphi$ is called an \midx{activation function}.
The following two activation functions are particularly common: \begin{enumerate}
  \item The \midx{hyperbolic tangent} (Tanh) is defined as \begin{align}
    \mathrm{Tanh}(z) \defeq \frac{\exp(z) - \exp(-z)}{\exp(z) + \exp(-z)} \in (-1, 1).
  \end{align}
  Tanh is a scaled and shifted variant of the sigmoid function \eqref{eq:logistic_function} which we have previously seen in the context of logistic regression as $\mathrm{Tanh}(z) = 2 \sigma(2 z) - 1$.

  \begin{marginfigure}
  \begin{center}
    \import{./plots/output/}{activation_functions.pgf}
  \end{center}

    \caption{The Tanh and ReLU activation functions, respectively.}
  \end{marginfigure}

  \item The \midx{rectified linear unit} (ReLU) is defined as \begin{align}
    \mathrm{ReLU}(z) \defeq \max \{z, 0\} \in [0, \infty).
  \end{align}
  In particular, the ReLU activation function leads to ``sparser'' gradients as it selects the halfspace of inputs with positive sign.
  Moreover, the gradients of ReLU do not ``vanish'' as $z \to \pm \infty$ which can lead to faster training.
\end{enumerate}

It is important that the activation function is nonlinear because otherwise, any composition of layers would still represent a linear function.
Non-linear activation functions allow the network to represent arbitrary functions.
This is known as the \midx{universal approximation theorem}, and it states that any artificial neural network with just a single hidden layer (with arbitrary width) and non-polynomial activation function $\varphi$ can approximate any continuous function to an arbitrary accuracy.

\subsection{Classification}\label{sec:bdl:ann:classification}

\begin{marginfigure}
  \begin{center}
    \import{./plots/output/}{softmax.pgf}
  \end{center}

  \caption{Softmax $\sigma_1(f_1, f_2)$ for a binary classification problem.
  Blue denotes a small probability and yellow denotes a large probability of belonging to class $1$, respectively.}
\end{marginfigure}

Although we mainly focus on regression, neural networks can equally well be used for classification.
If we want to classify inputs into $c$ separate classes, we can simply construct a neural network with $c$ outputs, $\vf = [f_1, \dots, f_c]$, and normalize them into a probability distribution.
Often, the \midx{softmax function} is used for normalization, \begin{align}
  \sigma_i(\vf) \defeq \frac{\exp(f_i)}{\sum_{j=1}^c \exp(f_j)} \label{eq:softmax_function}
\end{align} where $\sigma_i(\vf)$ corresponds to the probability mass of class $i$.
The softmax is a generalization of the logistic function \eqref{eq:logistic_function} to more than two classes~\exerciserefmark{softmax_and_logistic_function}.
Note that the softmax corresponds to a Gibbs distribution with energies $-\vf$.

\subsection{Maximum Likelihood Estimation}\label{sec:bnn:mle}

We will study neural networks under the lens of supervised learning~(cf. \cref{sec:fundamentals:supervised_learning}) where we are provided some independently-sampled (noisy) data $\spD = \{(\vx_i, \vy_i)\}_{i=1}^n$ generated according to an unknown process $\vy \sim p(\cdot \mid \vx, \opt{\vtheta})$, which we wish to approximate.

Upon initialization, the network does generally not approximate this process well, so a key element of deep learning is ``learning'' a parameterization $\vtheta$ that is a good approximation.
To this end, one typically considers a loss function $\ell(\vtheta; \vy)$ which quantifies the ``error'' of the network outputs $\vf(\vx; \vtheta)$.
In the classical setting of regression, i.e., $y \in \R$ and $k = 1$, $\ell$ is often taken to be the \midx<(empirical) mean squared error>{mean squared error}, \begin{align}
  \ell_\MSE(\vtheta; \spD) \defeq \frac{1}{n} \sum_{i=1}^n (f(\vx_i; \vtheta) - y_i)^2.
\end{align}
As we have already seen in \cref{sec:least_squares_as_mle} in the context of linear regression, minimizing mean squared error corresponds to maximum likelihood estimation under a Gaussian likelihood.

In the setting of classification where $\vy \in \{0,1\}^c$ is a one-hot encoding of class membership,\footnote{That is, exactly one component of $\vy$ is $1$ and all others are $0$, indicating to which class the given example belongs.} it is instead common to interpret the outputs of a neural network as probabilities akin to our discussion in \cref{sec:bdl:ann:classification}.
We denote by $q_\vtheta(\cdot \mid \vx)$ the resulting probability distribution over classes with PMF $[\sigma_1(\vf(\vx; \vtheta)), \dots, \sigma_c(\vf(\vx; \vtheta))]$, and aim to find $\vtheta$ such that $q_\vtheta(\vy \mid \vx) \approx p(\vy \mid \vx)$.
In this context, it is common to minimize the cross-entropy, \begin{align}
  \crH{p}{q_\vtheta} &= \E[(\vx, \vy) \sim p]{- \log q_\vtheta(\vy \mid \vx)} \margintag{using the definition of cross-entropy \eqref{eq:cross_entropy}} \nonumber \\
  &\approx \underbrace{- \frac{1}{n} \sum_{i=1}^n \log q_\vtheta(\vy_i \mid \vx_i)}_{\defeq \ell_\mathrm{ce}(\vtheta; \spD)} \margintag{using Monte Carlo sampling} \label{eq:ce_loss}
  \intertext{which can be understood as minimizing the surprise about the training data under the model.
  $\ell_\mathrm{ce}$ is called the \midx{cross-entropy loss}.
  Disregarding the constant $1/n$, we can rewrite the cross-entropy loss as}
  &\propto - \sum_{i=1}^n \log q_\vtheta(\vy_i \mid \vx_i) = \ell_\mathrm{nll}(\vtheta; \spD)
\end{align}
Recall that $\ell_\mathrm{nll}(\vtheta; \spD)$ is the \midx{negative log-likelihood}[idxpagebf] of the training data, and thus, empirically minimizing cross-entropy can equivalently be interpreted as maximum likelihood estimation.\footnote{We have previously seen this equivalence of MLE and empirically minimizing KL-divergence in \cref{sec:vi:kl:forward} (minimizing the cross-entropy $\crH{p}{q_\vtheta}$ is equivalent to minimizing forward-KL $\KL{p}{q_\vtheta}$). Note that this interpretation is not exclusive to the canonical cross-entropy loss from \cref{eq:ce_loss}, but holds for any MLE. For example, minimizing mean squared error corresponds to empirically minimizing the KL-divergence with a Gaussian likelihood.}
Furthermore, recall from \cref{exercise:logistic_loss_gradient} that for a two-class classification problem the cross-entropy loss is equivalent to the logistic loss.

\subsection{Backpropagation}\label{sec:bdl:ann:backprop}

A crucial property of neural networks is that they are differentiable.
That is, we can compute gradients $\grad_\vtheta \ell$ of $\vf$ with respect to the parameterization of the model $\vtheta = \mW_{1:L}$ and some loss function $\ell(\vf; \vy)$.
Being able to obtain these gradients efficiently allows for ``learning'' a particular function from data using first-order optimization methods.

The algorithm for computing gradients of a neural network is called \midx{backpropagation} and is essentially a repeated application of the chain rule.
Note that using the chain rule for every path through the network is computationally infeasible, as this quickly leads to a combinatorial explosion as the number of hidden layers is increased.
The key insight of backpropagation is that we can use the \midx<feed-forward>{feed-forward neural network} structure of our neural network to memoize computations of the gradient, yielding a linear time algorithm.
Obtaining gradients by backpropagation is often called \midx{automatic differentiation (auto-diff)}.
For more details, refer to \cite{goodfellow2016deep}.

Computing the exact gradient for each data point is still fairly expensive when the size of the neural network is large.
Typically, stochastic gradient descent is used to obtain unbiased gradient estimates using batches of only $m$ of the $n$ data points, where $m \ll n$.

\section{Bayesian Neural Networks}

\begin{marginfigure}[5\baselineskip]
  \incfig{bnn}
  \caption{Bayesian neural networks model a distribution over the weights of a neural network.}
\end{marginfigure}

How can we perform probabilistic inference in neural networks? We adopt the same strategy which we already used for Bayesian linear regression, we impose a Gaussian prior on the weights, \begin{align}
  \vtheta \sim \N{\vzero}{\sigmap^2 \mI}.
\end{align}
Similarly, we can use a Gaussian likelihood to describe how well the data is described by the model $\vf$, \begin{align}
  y \mid \vx, \vtheta \sim \N{f(\vx; \vtheta)}{\sigman^2}. \label{eq:bnn_likelihood}
\end{align}
Thus, instead of considering weights as point estimates which are learned exactly, \midx<Bayesian neural networks>{Bayesian neural network} learn a distribution over the weights of the network.
In principle, other priors and likelihoods can be used, yet Gaussians are typically chosen due to their closedness properties, which we have seen in \cref{sec:fundamentals:gaussians} and many times since.

\subsection{Maximum a Posteriori Estimation}\label{sec:bnn:map_inference}

Before studying probabilistic inference, let us first consider MAP estimation in the context of neural networks.

The MAP estimate of the weights is obtained by \begin{align}
  \vthetahat_\MAP = \argmax_\vtheta \log p(\vtheta) + \sum_{i=1}^n \log p(y_i \mid \vx_i, \vtheta). \label{eq:bnn:map}
\end{align}
In \cref{sec:bnn:mle}, we have seen that the negative log-likelihood under a Gaussian likelihood \eqref{eq:bnn_likelihood} is the squared error between label and prediction,\looseness=-1 \begin{align}
  \log p(y_i \mid \vx_i, \vtheta) = - \frac{(y_i - f(\vx_i; \vtheta))^2}{2 \sigman^2} + \const.
\end{align}
Obtaining the MAP estimate instead, simply corresponds to adding an $L_2$-regularization term to the squared error loss, \begin{align}
  \vthetahat_\MAP = \argmin_\vtheta \frac{1}{2 \sigmap^2} \norm{\vtheta}_2^2 + \frac{1}{2 \sigman^2} \sum_{i=1}^n (y_i - f(\vx_i; \vtheta))^2. \margintag{using that for an isotropic Gaussian prior, $\log p(\vtheta) = - \frac{1}{2 \sigmap^2} \norm{\vtheta}_2^2 + \const$}
\end{align}
As we have already seen in \cref{rmk:weight_decay} and the context of Bayesian linear regression (and ridge regression), using a Gaussian prior is equivalent to applying weight decay.\footnote{Recall that weight decay regularizes weights by shrinking them towards zero.}
Using gradient ascent, we obtain the following update rule, \begin{align}
  \vtheta \gets \vtheta(1 - \lambda \eta_t) + \eta_t \sum_{i=1}^n \grad \log p(y_i \mid \vx_i, \vtheta) \label{eq:bdl_map_update}
\end{align} where $\lambda \defeq \nicefrac{1}{\sigmap^2}$.
The gradients of the likelihood can be obtained using automatic differentiation.

\subsection{Heteroscedastic Noise}\label{sec:bdl:noise}

\Cref{eq:bnn_likelihood} uses the scalar parameter $\sigman^2$ to model the aleatoric uncertainty (label noise), similarly to how we modeled the label noise in Bayesian linear regression and Gaussian processes.
Such a noise model is called \midx{homoscedastic noise} as the noise is assumed to be uniform across the domain.
In many settings, however, the noise is inherently \midx{heteroscedastic noise}.
That is, the noise varies depending on the input and which ``region'' of the domain the input is from.
This behavior is visualized in \cref{fig:heteroscedastic_noise}.

\begin{marginfigure}
  \begin{center}
    \import{./plots/output/}{heteroscedastic_noise.pgf}
  \end{center}

  \caption{Illustration of data with variable (heteroscedastic) noise.
  The noise increases as the inputs increase in magnitude.
  The noise-free function is shown in black.}\label{fig:heteroscedastic_noise}
\end{marginfigure}

There is a natural way of modeling heteroscedastic noise with Bayesian neural networks.
We use a neural network with two outputs $f_1$ and $f_2$ and define \begin{subequations}\begin{align}
  y \mid \vx, \vtheta &\sim \N{\mu(\vx; \vtheta)}{\sigma^2(\vx; \vtheta)} \quad\text{where} \label{eq:bnn_likelihood_het} \\
  \mu(\vx; \vtheta) &\defeq f_1(\vx; \vtheta), \\
  \sigma^2(\vx; \vtheta) &\defeq \exp(f_2(\vx; \vtheta)). \margintag{we exponentiate $f_2$ to ensure non-negativity of the variance}
\end{align}\end{subequations}

Using this model, the likelihood term from \cref{eq:bnn:map} is \begin{align}
  \log p(y_i \mid \vx_i, \vtheta) &= \log \N[y_i]{\mu(\vx_i; \vtheta)}{\sigma^2(\vx_i; \vtheta)} \nonumber \\
  &= \log \frac{1}{\sqrt{2 \pi \sigma^2(\vx_i; \vtheta)}} - \frac{(y_i - \mu(\vx_i; \vtheta))^2}{2 \sigma^2(\vx_i; \vtheta)} \margintag{note that the normalizing constant depends on the noise model!} \nonumber \\
  &= \underbrace{\log \frac{1}{\sqrt{2 \pi}}}_{\const} - \frac{1}{2} \brackets*{\log \sigma^2(\vx_i; \vtheta) + \frac{(y_i - \mu(\vx_i; \vtheta))^2}{\sigma^2(\vx_i; \vtheta)}}. \label{eq:bdl_map}
\end{align}
Hence, the model can either explain the label $y_i$ by an accurate model $\mu(\vx_i; \vtheta)$ or by a large variance $\sigma^2(\vx_i; \vtheta)$, yet, it is penalized for choosing large variances.
Intuitively, this allows to attenuate losses for some data points by attributing them to large variance (when no model reflecting all data points simultaneously can be found).
This allows the model to ``learn'' its aleatoric uncertainty.
However, recall that the MAP estimate still corresponds to a point estimate of the weights, so we forgo modeling the epistemic uncertainty.

\section{Approximate Probabilistic Inference}

Naturally, we want to understand the epistemic uncertainty of our model too.
However, learning and inference in Bayesian neural networks are generally intractable (even when using a Gaussian prior and likelihood) when the noise is not assumed to be homoscedastic and known.\footnote{In this case, the conjugate prior to a Gaussian likelihood is not a Gaussian. See, e.g., \url{https://en.wikipedia.org/wiki/Conjugate_prior}.}
Thus, we are led to the techniques of approximate inference, which we discussed in the previous two chapters.

\subsection{Variational Inference}\label{sec:bnn:approximate_inference:vi}\pidx{variational inference}

As we have discussed in \cref{sec:approximate_inference}, we can apply black box stochastic variational inference which --- in the context of neural networks --- is also known as \midx{Bayes by Backprop}.
As variational family, we use the family of independent Gaussians which we have already encountered in \cref{ex:var_family_diag_gaussian}.\footnote{Independent Gaussians are useful because they can be encoded using only $2d$ parameters, which is crucial when the size of the neural network is large.}
Recall the fundamental objective of variational inference~\eqref{eq:var_optimization},\looseness=-1 \begin{align*}
  &\argmin_{q \in \spQ} \KL{q}{p(\cdot \mid \vx_{1:n}, y_{1:n})} \\
  &= \argmax_{q \in \spQ} L(q, p; \spD) \margintag{using \cref{eq:elbo_reverse_kl_relationship}} \\
  &= \argmax_{q \in \spQ} \E[\vtheta \sim q]{\log p(y_{1:n} \mid \vx_{1:n}, \vtheta)} - \KL{q}{p(\cdot)}. \margintag{using \cref{eq:elbo2}}
\end{align*}
The KL-divergence $\KL{q}{p(\cdot)}$ can be expressed in closed-form for Gaussians.\footnote{see \cref{eq:kl_gaussian}}
Recall that we can obtain unbiased gradient estimates of the expectation using the reparameterization trick \eqref{eq:reparameterization_trick_application}, \begin{align*}
  \E[\vtheta \sim q]{\log p(y_{1:n} \mid \vx_{1:n}, \vtheta)} = \E[\vvarepsilon \sim \SN]{\left. \log p(y_{1:n} \mid \vx_{1:n}, \vtheta) \right\rvert_{\vtheta = \msqrt{\mSigma} \vvarepsilon + \vmu}}.
\end{align*}
As $\mSigma$ is the diagonal matrix ${\diag{\sigma_1^2, \dots, \sigma_d^2}}$, ${\msqrt{\mSigma} = \diag{\sigma_1, \dots, \sigma_d}}$.
The gradients of the likelihood can be obtained using backpropagation.
We can now use the variational posterior $q_\vlambda$ to perform approximate inference,\looseness=-1 \begin{align}
  p(\ys \mid \vxs, \vx_{1:n}, \vy_{1:n}) &= \int p(\ys \mid \vxs, \vtheta) p(\vtheta \mid \vx_{1:n}, y_{1:n}) \,d\vtheta \margintag{using the sum rule \eqref{eq:sum_rule} and product rule \eqref{eq:product_rule}} \nonumber \\
  &= \E[\vtheta \sim p(\cdot \mid \vx_{1:n}, y_{1:n})]{p(\ys \mid \vxs, \vtheta)} \margintag{interpreting the integral as an expectation over the posterior} \nonumber \\[5pt]
  &\approx \E[\theta \sim q_\vlambda]{p(\ys \mid \vxs, \vtheta)} \margintag{approximating the posterior with the variational posterior $q_\vlambda$} \nonumber \\
  &\approx \frac{1}{m} \sum_{i=1}^m p(\ys \mid \vxs, \vtheta^{(i)}) \margintag{using Monte Carlo sampling} \label{eq:bnn_vi}
  \intertext{for independent samples $\vtheta^{(i)} \iid q_\vlambda$,}
  &= \frac{1}{m} \sum_{i=1}^m \N[\ys]{\mu(\vxs; \vtheta^{(i)})}{\sigma^2(\vxs; \vtheta^{(i)})}. \label{eq:bnn_posterior_vi} \margintag{using the neural network}
\end{align}
Intuitively, variational inference in Bayesian neural networks can be interpreted as averaging the predictions of multiple neural networks drawn according to the variational posterior $q_\vlambda$.

Using the Monte Carlo samples $\vtheta^{(i)}$, we can also estimate the mean of our predictions, \begin{align}
  \E{\ys \mid \vxs, \vx_{1:n}, y_{1:n}} \approx \frac{1}{m} \sum_{i=1}^m \mu(\vxs; \vtheta^{(i)}) \eqdef \mean{\mu}(\vxs),
\end{align} and the variance of our predictions, \begin{align*}
  \Var{\ys \mid \vxs, \vx_{1:n}, y_{1:n}} &= \b{\E[\vtheta]{\Var[\ys]{\ys \mid \vxs, \vtheta}}} + \r{\Var[\vtheta]{\E[\ys]{\ys \mid \vxs, \vtheta}}} \margintag{using the law of total variance \eqref{eq:lotv}} \\
  &= \b{\E[\vtheta]{\sigma^2(\vxs; \vtheta)}} + \r{\Var[\vtheta]{\mu(\vxs; \vtheta)}}. \margintag{using the likelihood \eqref{eq:bnn_likelihood_het}}
\end{align*}
Recall from \cref{eq:lotv_interpretation} that the first term corresponds to the \b{aleatoric uncertainty} of the data and the second term corresponds to the \r{epistemic uncertainty} of the model.
We can approximate them using the Monte Carlo samples $\vtheta^{(i)}$, \begin{align}
  \Var{\ys \mid \vxs, \vx_{1:n}, y_{1:n}} \approx \begin{multlined}[t]\b{\frac{1}{m} \sum_{i=1}^m \sigma^2(\vxs; \vtheta^{(i)})} \\ + \r{\frac{1}{m-1} \sum_{i=1}^m (\mu(\vxs; \vtheta^{(i)}) - \mean{\mu}(\vxs))^2}\end{multlined}
\end{align} using a sample mean \eqref{eq:sample_mean} and sample variance \eqref{eq:sample_variance}.

\subsection{Markov Chain Monte Carlo}

As we have discussed in \cref{sec:approximate_inference:mcmc}, an alternative method to approximating the full posterior distribution is to sample from it directly.
By the ergodic theorem \eqref{eq:ergodic_theorem}, we can use any of the discussed sampling strategies to obtain samples $\vtheta^{(t)}$ such that \begin{align*}
  p(\ys \mid \vxs, \vx_{1:n}, \vy_{1:n}) \approx \frac{1}{T} \sum_{t = 1}^T p(\ys \mid \vxs, \vtheta^{(t)}). \label{eq:bnn_posterior_mcmc} \margintag{see \eqref{eq:mcmc}}
\end{align*}
Here, we omit the offset $t_0$ which is commonly used to avoid the ``burn-in'' period for simplicity.
Algorithms such as SGLD or SG-HMC are often used as they rely only on stochastic gradients of the loss function which can be computed efficiently using automatic differentiation.

Typically, for large networks, we cannot afford to store all $T$ samples of models.
Thus, we need to summarize the iterates.\footnote{That is, combine the individual samples of weights $\vtheta^{(i)}$.}
One approach is to keep track of $m$ \midx<snapshots>{snapshot} of weights $[\vtheta^{(1)}, \dots, \vtheta^{(m)}]$ according to some schedule and use those for inference (e.g., by averaging the predictions of the corresponding neural networks).
This approach of sampling a subset of some data is generally called \midx{subsampling}.

Another approach is to summarize (that is, approximate) the weights using sufficient statistics (e.g., a Gaussian).\footnote{A statistic is \idx<sufficient>{sufficient statistic} for a family of probability distributions if the samples from which it is calculated yield no more information than the statistic with respect to the learned parameters. We provide a formal definition in \cref{sec:mutual_information}.}
In other words, we learn the Gaussian approximation, \begin{subequations}\begin{align}
  \vtheta &\sim \N{\vmu}{\mSigma}, \quad\text{where} \\
  \vmu &\defeq \frac{1}{T} \sum_{i=1}^T \vtheta^{(i)}, \margintag{using a sample mean \eqref{eq:sample_mean}} \\
  \mSigma &\defeq \frac{1}{T - 1} \sum_{i=1}^T (\vtheta^{(i)} - \vmu) \transpose{(\vtheta^{(i)} - \vmu)}, \margintag{using a sample variance \eqref{eq:sample_variance}}
\end{align}\label{eq:iteraties_gaussian_sufficient_statistics}\end{subequations} using sample means and sample (co)variances.
This can be implemented efficiently using running averages of the first and second moments,\looseness=-1 \begin{align}
  \vmu \gets \frac{1}{T+1} (T \vmu + \vtheta) \quad\text{and}\quad \mA \gets \frac{1}{T+1} (T \mA + \vtheta \transpose{\vtheta})
\end{align} upon observing the fresh sample $\vtheta$.
$\mSigma$ can easily be calculated from these moments, \begin{align}
  \mSigma = \frac{T}{T-1} (\mA - \vmu \transpose{\vmu}). \margintag{using the characterization of sample variance in terms of estimators of the first and second moment \eqref{eq:sample_variance2}}
\end{align}
To predict, we can sample weights $\vtheta$ from the learned Gaussian.

\begin{rmk}{Stochastic weight averaging}{}
  It turns out that this approach works well even without injecting additional Gaussian noise during training, e.g., when using SGD rather than SGLD.
  Simply taking the mean of the iterates of SGD is called \midx{stochastic weight averaging} (SWA).
  Describing the iterates of SGD by Gaussian sufficient statistics (analogously to \cref{eq:iteraties_gaussian_sufficient_statistics}), is known as the \midx{stochastic weight averaging-Gaussian} (SWAG) method~\citep{izmailov2018averaging}.
\end{rmk}

\subsection{Dropout and Dropconnect}\label{sec:bnn:approximate_inference:dropout}

We will now discuss two approximate inference techniques that are tailored to neural networks.
The first is ``dropout''/``dropconnect'' regularization.
Traditionally, \midx{dropout regularization} \citep{hinton2012improving,srivastava2014dropout} randomly omits vertices of the computation graph during training, see \cref{fig:dropout_regularization_as_vi}.
In contrast, \midx{dropconnect regularization} \citep{wan2013regularization} randomly omits edges of the computation graph.
The key idea that we will present here is to interpret this type of regularization as performing variational inference.

\begin{marginfigure}[-8\baselineskip]
  \incfig{dropout}
  \caption{Illustration of dropout regularization. Some vertices of the computation graph are randomly omitted. In contrast, dropconnect regularization randomly omits edges of the computation graph.}\label{fig:dropout_regularization}
\end{marginfigure}
\begin{marginfigure}
  \begin{center}
    %% Creator: Matplotlib, PGF backend
%%
%% To include the figure in your LaTeX document, write
%%   \input{<filename>.pgf}
%%
%% Make sure the required packages are loaded in your preamble
%%   \usepackage{pgf}
%%
%% Also ensure that all the required font packages are loaded; for instance,
%% the lmodern package is sometimes necessary when using math font.
%%   \usepackage{lmodern}
%%
%% Figures using additional raster images can only be included by \input if
%% they are in the same directory as the main LaTeX file. For loading figures
%% from other directories you can use the `import` package
%%   \usepackage{import}
%%
%% and then include the figures with
%%   \import{<path to file>}{<filename>.pgf}
%%
%% Matplotlib used the following preamble
%%   \usepackage{amsmath}\usepackage{amssymb}
%%
\begingroup%
\makeatletter%
\begin{pgfpicture}%
\pgfpathrectangle{\pgfpointorigin}{\pgfqpoint{1.900000in}{1.725999in}}%
\pgfusepath{use as bounding box, clip}%
\begin{pgfscope}%
\pgfsetbuttcap%
\pgfsetmiterjoin%
\definecolor{currentfill}{rgb}{1.000000,1.000000,1.000000}%
\pgfsetfillcolor{currentfill}%
\pgfsetlinewidth{0.000000pt}%
\definecolor{currentstroke}{rgb}{1.000000,1.000000,1.000000}%
\pgfsetstrokecolor{currentstroke}%
\pgfsetdash{}{0pt}%
\pgfpathmoveto{\pgfqpoint{0.000000in}{0.000000in}}%
\pgfpathlineto{\pgfqpoint{1.900000in}{0.000000in}}%
\pgfpathlineto{\pgfqpoint{1.900000in}{1.725999in}}%
\pgfpathlineto{\pgfqpoint{0.000000in}{1.725999in}}%
\pgfpathlineto{\pgfqpoint{0.000000in}{0.000000in}}%
\pgfpathclose%
\pgfusepath{fill}%
\end{pgfscope}%
\begin{pgfscope}%
\pgfsetbuttcap%
\pgfsetmiterjoin%
\definecolor{currentfill}{rgb}{1.000000,1.000000,1.000000}%
\pgfsetfillcolor{currentfill}%
\pgfsetlinewidth{0.000000pt}%
\definecolor{currentstroke}{rgb}{0.000000,0.000000,0.000000}%
\pgfsetstrokecolor{currentstroke}%
\pgfsetstrokeopacity{0.000000}%
\pgfsetdash{}{0pt}%
\pgfpathmoveto{\pgfqpoint{0.153558in}{0.457988in}}%
\pgfpathlineto{\pgfqpoint{1.800000in}{0.457988in}}%
\pgfpathlineto{\pgfqpoint{1.800000in}{1.476317in}}%
\pgfpathlineto{\pgfqpoint{0.153558in}{1.476317in}}%
\pgfpathlineto{\pgfqpoint{0.153558in}{0.457988in}}%
\pgfpathclose%
\pgfusepath{fill}%
\end{pgfscope}%
\begin{pgfscope}%
\pgfpathrectangle{\pgfqpoint{0.153558in}{0.457988in}}{\pgfqpoint{1.646442in}{1.018329in}}%
\pgfusepath{clip}%
\pgfsetbuttcap%
\pgfsetroundjoin%
\definecolor{currentfill}{rgb}{0.941176,0.125490,0.000000}%
\pgfsetfillcolor{currentfill}%
\pgfsetlinewidth{1.003750pt}%
\definecolor{currentstroke}{rgb}{0.941176,0.125490,0.000000}%
\pgfsetstrokecolor{currentstroke}%
\pgfsetdash{}{0pt}%
\pgfsys@defobject{currentmarker}{\pgfqpoint{-0.020833in}{-0.020833in}}{\pgfqpoint{0.020833in}{0.020833in}}{%
\pgfpathmoveto{\pgfqpoint{0.000000in}{-0.020833in}}%
\pgfpathcurveto{\pgfqpoint{0.005525in}{-0.020833in}}{\pgfqpoint{0.010825in}{-0.018638in}}{\pgfqpoint{0.014731in}{-0.014731in}}%
\pgfpathcurveto{\pgfqpoint{0.018638in}{-0.010825in}}{\pgfqpoint{0.020833in}{-0.005525in}}{\pgfqpoint{0.020833in}{0.000000in}}%
\pgfpathcurveto{\pgfqpoint{0.020833in}{0.005525in}}{\pgfqpoint{0.018638in}{0.010825in}}{\pgfqpoint{0.014731in}{0.014731in}}%
\pgfpathcurveto{\pgfqpoint{0.010825in}{0.018638in}}{\pgfqpoint{0.005525in}{0.020833in}}{\pgfqpoint{0.000000in}{0.020833in}}%
\pgfpathcurveto{\pgfqpoint{-0.005525in}{0.020833in}}{\pgfqpoint{-0.010825in}{0.018638in}}{\pgfqpoint{-0.014731in}{0.014731in}}%
\pgfpathcurveto{\pgfqpoint{-0.018638in}{0.010825in}}{\pgfqpoint{-0.020833in}{0.005525in}}{\pgfqpoint{-0.020833in}{0.000000in}}%
\pgfpathcurveto{\pgfqpoint{-0.020833in}{-0.005525in}}{\pgfqpoint{-0.018638in}{-0.010825in}}{\pgfqpoint{-0.014731in}{-0.014731in}}%
\pgfpathcurveto{\pgfqpoint{-0.010825in}{-0.018638in}}{\pgfqpoint{-0.005525in}{-0.020833in}}{\pgfqpoint{0.000000in}{-0.020833in}}%
\pgfpathlineto{\pgfqpoint{0.000000in}{-0.020833in}}%
\pgfpathclose%
\pgfusepath{stroke,fill}%
}%
\begin{pgfscope}%
\pgfsys@transformshift{0.336496in}{0.712571in}%
\pgfsys@useobject{currentmarker}{}%
\end{pgfscope}%
\begin{pgfscope}%
\pgfsys@transformshift{1.068248in}{1.221735in}%
\pgfsys@useobject{currentmarker}{}%
\end{pgfscope}%
\end{pgfscope}%
\begin{pgfscope}%
\pgfsetbuttcap%
\pgfsetroundjoin%
\definecolor{currentfill}{rgb}{0.000000,0.000000,0.000000}%
\pgfsetfillcolor{currentfill}%
\pgfsetlinewidth{0.803000pt}%
\definecolor{currentstroke}{rgb}{0.000000,0.000000,0.000000}%
\pgfsetstrokecolor{currentstroke}%
\pgfsetdash{}{0pt}%
\pgfsys@defobject{currentmarker}{\pgfqpoint{0.000000in}{-0.048611in}}{\pgfqpoint{0.000000in}{0.000000in}}{%
\pgfpathmoveto{\pgfqpoint{0.000000in}{0.000000in}}%
\pgfpathlineto{\pgfqpoint{0.000000in}{-0.048611in}}%
\pgfusepath{stroke,fill}%
}%
\begin{pgfscope}%
\pgfsys@transformshift{0.336496in}{0.457988in}%
\pgfsys@useobject{currentmarker}{}%
\end{pgfscope}%
\end{pgfscope}%
\begin{pgfscope}%
\definecolor{textcolor}{rgb}{0.000000,0.000000,0.000000}%
\pgfsetstrokecolor{textcolor}%
\pgfsetfillcolor{textcolor}%
\pgftext[x=0.336496in,y=0.360766in,,top]{\color{textcolor}\rmfamily\fontsize{6.940000}{8.328000}\selectfont \(\displaystyle 0\)}%
\end{pgfscope}%
\begin{pgfscope}%
\pgfsetbuttcap%
\pgfsetroundjoin%
\definecolor{currentfill}{rgb}{0.000000,0.000000,0.000000}%
\pgfsetfillcolor{currentfill}%
\pgfsetlinewidth{0.803000pt}%
\definecolor{currentstroke}{rgb}{0.000000,0.000000,0.000000}%
\pgfsetstrokecolor{currentstroke}%
\pgfsetdash{}{0pt}%
\pgfsys@defobject{currentmarker}{\pgfqpoint{0.000000in}{-0.048611in}}{\pgfqpoint{0.000000in}{0.000000in}}{%
\pgfpathmoveto{\pgfqpoint{0.000000in}{0.000000in}}%
\pgfpathlineto{\pgfqpoint{0.000000in}{-0.048611in}}%
\pgfusepath{stroke,fill}%
}%
\begin{pgfscope}%
\pgfsys@transformshift{1.068248in}{0.457988in}%
\pgfsys@useobject{currentmarker}{}%
\end{pgfscope}%
\end{pgfscope}%
\begin{pgfscope}%
\definecolor{textcolor}{rgb}{0.000000,0.000000,0.000000}%
\pgfsetstrokecolor{textcolor}%
\pgfsetfillcolor{textcolor}%
\pgftext[x=1.068248in,y=0.360766in,,top]{\color{textcolor}\rmfamily\fontsize{6.940000}{8.328000}\selectfont \(\displaystyle \lambda_j\)}%
\end{pgfscope}%
\begin{pgfscope}%
\definecolor{textcolor}{rgb}{0.000000,0.000000,0.000000}%
\pgfsetstrokecolor{textcolor}%
\pgfsetfillcolor{textcolor}%
\pgftext[x=0.976779in,y=0.210304in,,top]{\color{textcolor}\rmfamily\fontsize{8.330000}{9.996000}\selectfont \(\displaystyle \theta_j\)}%
\end{pgfscope}%
\begin{pgfscope}%
\definecolor{textcolor}{rgb}{0.000000,0.000000,0.000000}%
\pgfsetstrokecolor{textcolor}%
\pgfsetfillcolor{textcolor}%
\pgftext[x=0.153558in,y=1.527234in,,bottom]{\color{textcolor}\rmfamily\fontsize{8.330000}{9.996000}\selectfont \(\displaystyle q_j\)}%
\end{pgfscope}%
\begin{pgfscope}%
\pgfpathrectangle{\pgfqpoint{0.153558in}{0.457988in}}{\pgfqpoint{1.646442in}{1.018329in}}%
\pgfusepath{clip}%
\pgfsetbuttcap%
\pgfsetroundjoin%
\pgfsetlinewidth{1.003750pt}%
\definecolor{currentstroke}{rgb}{0.941176,0.125490,0.000000}%
\pgfsetstrokecolor{currentstroke}%
\pgfsetdash{}{0pt}%
\pgfpathmoveto{\pgfqpoint{0.336496in}{0.457988in}}%
\pgfpathlineto{\pgfqpoint{0.336496in}{0.712571in}}%
\pgfusepath{stroke}%
\end{pgfscope}%
\begin{pgfscope}%
\pgfpathrectangle{\pgfqpoint{0.153558in}{0.457988in}}{\pgfqpoint{1.646442in}{1.018329in}}%
\pgfusepath{clip}%
\pgfsetbuttcap%
\pgfsetroundjoin%
\pgfsetlinewidth{1.003750pt}%
\definecolor{currentstroke}{rgb}{0.941176,0.125490,0.000000}%
\pgfsetstrokecolor{currentstroke}%
\pgfsetdash{}{0pt}%
\pgfpathmoveto{\pgfqpoint{1.068248in}{0.457988in}}%
\pgfpathlineto{\pgfqpoint{1.068248in}{1.221735in}}%
\pgfusepath{stroke}%
\end{pgfscope}%
\begin{pgfscope}%
\pgfsetrectcap%
\pgfsetmiterjoin%
\pgfsetlinewidth{0.803000pt}%
\definecolor{currentstroke}{rgb}{0.000000,0.000000,0.000000}%
\pgfsetstrokecolor{currentstroke}%
\pgfsetdash{}{0pt}%
\pgfpathmoveto{\pgfqpoint{0.153558in}{0.457988in}}%
\pgfpathlineto{\pgfqpoint{0.153558in}{1.476317in}}%
\pgfusepath{stroke}%
\end{pgfscope}%
\begin{pgfscope}%
\pgfsetrectcap%
\pgfsetmiterjoin%
\pgfsetlinewidth{0.803000pt}%
\definecolor{currentstroke}{rgb}{0.000000,0.000000,0.000000}%
\pgfsetstrokecolor{currentstroke}%
\pgfsetdash{}{0pt}%
\pgfpathmoveto{\pgfqpoint{1.800000in}{0.457988in}}%
\pgfpathlineto{\pgfqpoint{1.800000in}{1.476317in}}%
\pgfusepath{stroke}%
\end{pgfscope}%
\begin{pgfscope}%
\pgfsetrectcap%
\pgfsetmiterjoin%
\pgfsetlinewidth{0.803000pt}%
\definecolor{currentstroke}{rgb}{0.000000,0.000000,0.000000}%
\pgfsetstrokecolor{currentstroke}%
\pgfsetdash{}{0pt}%
\pgfpathmoveto{\pgfqpoint{0.153558in}{0.457988in}}%
\pgfpathlineto{\pgfqpoint{1.800000in}{0.457988in}}%
\pgfusepath{stroke}%
\end{pgfscope}%
\begin{pgfscope}%
\pgfsetrectcap%
\pgfsetmiterjoin%
\pgfsetlinewidth{0.803000pt}%
\definecolor{currentstroke}{rgb}{0.000000,0.000000,0.000000}%
\pgfsetstrokecolor{currentstroke}%
\pgfsetdash{}{0pt}%
\pgfpathmoveto{\pgfqpoint{0.153558in}{1.476317in}}%
\pgfpathlineto{\pgfqpoint{1.800000in}{1.476317in}}%
\pgfusepath{stroke}%
\end{pgfscope}%
\begin{pgfscope}%
\definecolor{textcolor}{rgb}{0.941176,0.125490,0.000000}%
\pgfsetstrokecolor{textcolor}%
\pgfsetfillcolor{textcolor}%
\pgftext[x=0.409671in,y=0.692204in,left,base]{\color{textcolor}\rmfamily\fontsize{6.940000}{8.328000}\selectfont \(\displaystyle p\)}%
\end{pgfscope}%
\begin{pgfscope}%
\definecolor{textcolor}{rgb}{0.941176,0.125490,0.000000}%
\pgfsetstrokecolor{textcolor}%
\pgfsetfillcolor{textcolor}%
\pgftext[x=1.141423in,y=1.201368in,left,base]{\color{textcolor}\rmfamily\fontsize{6.940000}{8.328000}\selectfont \(\displaystyle 1-p\)}%
\end{pgfscope}%
\end{pgfpicture}%
\makeatother%
\endgroup%
  \end{center}

  \caption{Interpretation of dropconnect regularization as variational inference.
  The only coordinates where the variational posterior $q_j$ has positive probability are $0$ and $\lambda_j$.}\label{fig:dropout_regularization_as_vi}
\end{marginfigure}

In our exposition, we will focus on dropconnect, but the same approach also applies to dropout \citep{gal2016dropout}.
Suppose that we omit an edge of the computation graph (i.e., set its weight to zero) with probability $p$.
Then our variational posterior is given by \begin{align}
  q(\vtheta \mid \vlambda) \defeq \prod_{j=1}^d q_j(\theta_j \mid \lambda_j)
\end{align} where $d$ is the number of weights of the neural network and \begin{align}
  q_j(\theta_j \mid \lambda_j) \defeq p \delta_0(\theta_j) + (1-p)\delta_{\lambda_j}(\theta_j). \label{eq:dropconnect_var_family}
\end{align}
Here, $\delta_\alpha$ is the Dirac delta function with point mass at $\alpha$.\footnote{see \cref{sec:background:probability:dirac_delta}}
The variational parameters $\vlambda$ correspond to the ``original'' weights of the network.
In words, the variational posterior expresses that the $j$-th weight has value $0$ with probability $p$ and value $\lambda_j$ with probability $1-p$.
For fixed weights $\vlambda$, sampling from the variational posterior $q_{\vlambda}$ corresponds to sampling a vector $\vz$ with entries $\vz(i) \sim \Bern{p}$, yielding $\vz \odot \vlambda$ which is one of $2^d$ possible subnetworks.\footnote{$\mA \odot \mB$ denotes the Hadamard (element-wise) product.}

The weights $\vlambda$ can be learned by maximizing the ELBO, analogously to black-box variational inference.
The KL-divergence term of the ELBO is not tractable for the variational family described by \cref{eq:dropconnect_var_family}, instead a common approach is to use a mixture of Gaussians: \begin{align}
  q_j(\theta_j \mid \lambda_j) \defeq p \N[\theta_j]{0}{1} + (1-p) \N[\theta_j]{\lambda_j}{1}.
\end{align}
In this case, it can be shown that $\KL{q_{\vlambda}}{p(\cdot)} \approx \frac{p}{2} \norm{\vlambda}_2^2$ for sufficiently large $d$ \citep[proposition 1]{gal2015dropout}.
Thus, \begin{align}
  &\argmax_{\vlambda \in \Lambda} L(q_\vlambda, p; \spD) \nonumber \\
  &= \argmax_{\vlambda \in \Lambda} \E[\vtheta \sim q_\vlambda]{\log p(y_{1:n} \mid \vx_{1:n}, \vtheta)} - \KL{q}{p(\cdot)} \margintag{using \cref{eq:elbo2}} \nonumber \\
  &\approx \argmin_{\vlambda \in \Lambda} - \frac{1}{m} \sum_{i=1}^m \log p(y_{1:n} \mid \vx_{1:n}, \vtheta) \bigr\rvert_{\vtheta = \vz^{(i)} \odot \vlambda + \vvarepsilon^{(i)}} + \frac{p}{2} \norm{\vlambda}_2^2 \margintag{using Monte Carlo sampling} \label{eq:dropconnect_vi}
\end{align} where we reparameterize $\vtheta \sim q_\vlambda$ by $\vtheta = \vz \odot \vlambda + \vvarepsilon$ with $\vz(i) \sim \Bern{p}$ and $\vvarepsilon \sim \N{\vzero}{\mI}$.
\Cref{eq:dropconnect_vi} is the standard $L_2$-regularized loss function of a neural network with weights $\vlambda$ and dropconnect, and it is straightforward to obtain unbiased gradient estimates by automatic differentiation.\looseness=-1

Crucially, for the interpretation of dropconnect regularization as variational inference to be valid, we also need to perform dropconnect regularization during inference, \begin{align}
  p(\ys \mid \vxs, \vx_{1:n}, \vy_{1:n}) &\approx \E[\vtheta \sim q_\vlambda]{p(\ys \mid \vxs, \vtheta)} \nonumber \\
  &\approx \frac{1}{m} \sum_{i=1}^m p(\ys \mid \vxs, \vtheta^{(i)}) \label{eq:bnn_posterior_dr} \margintag{using Monte Carlo sampling}
\end{align} where $\vtheta^{(i)} \iid q_\vlambda$ are independent samples.
This coincides with our earlier discussion of variational inference for Bayesian neural networks in \cref{eq:bnn_vi}.
In words, we average the predictions of $m$ neural networks for each of which we randomly ``drop out'' weights.

\begin{rmk}{Masksembles}{masksembles}
  A practical problem of dropout is that for any reasonable choice of dropout probability, the dropout masks~$\vz^{(i)}$ will overlap significantly, which tends to make the predictions~$p(\ys \mid \vxs, \vtheta^{(i)})$ highly correlated.
  This can lead to underestimation of epistemic uncertainty.
  \midx<Masksembles>{masksemble}~\citep{durasov2021masksembles} mitigate this issue by choosing a fixed set of pre-defined dropout masks, which have controlled overlap, and alternating between them during training.
  In the extreme case of ``infinitely many'' masks, masksembles are equivalent to dropout since each mask is only seen once.
\end{rmk}

\subsection{Probabilistic Ensembles}\label{sec:bdl:approximate_inference:probabilistic_ensembles}

We have seen that variational inference in the context of Bayesian neural networks can be interpreted as averaging the predictions of $m$ neural networks drawn according to the variational posterior.

A natural adaptation of this idea is to immediately learn the weights of $m$ neural networks.
The idea is to randomly choose $m$ training sets by sampling uniformly from the data with replacement. Then, using our analysis from \cref{sec:bnn:map_inference}, we obtain $m$ MAP estimates of the weights~$\vtheta^{(i)}$, yielding the approximation \begin{align}
  p(\ys \mid \vxs, \vx_{1:n}, \vy_{1:n}) &= \E[\vtheta \sim p(\cdot \mid \vx_{1:n}, y_{1:n})]{p(\ys \mid \vxs, \vtheta)} \nonumber \\
  &\approx \frac{1}{m} \sum_{i=1}^m p(\ys \mid \vxs, \vtheta^{(i)}). \label{eq:bnn_posterior_pe} \margintag{using bootstrapping}
\end{align}
Here, the randomly chosen $m$ training sets induce ``diversity'' of the models.
In practice, in the context of deep neural networks where the global minimizer of the loss can rarely be identified, it is common to use the full training data to train each of the $m$ neural networks.
Random initialization and random shuffling of the training data is typically enough to ensure some degree of diversity between the individual models \citep{lakshminarayanan2017simple}.
We can connect ensembles to the other approximate inference techniques we have discussed:
First, ensembles can be seen as a specific kind of masksemble, where the masks are non-overlapping,\footnote{That is, the $m$ models do not share any of their parameters. Ensembles and dropout lie on opposite ends of this spectrum.} which mitigates the issue of correlated predictions from \cref{rmk:masksembles}.
Second, ensembling can be combined with other approximate inference techniques such as variational inference, Laplace approximation, or SWAG to get a mixture of Gaussians as the posterior approximation.

Note that \cref{eq:bnn_posterior_pe} is not equivalent to Monte Carlo sampling, although it looks very similar.
The key difference is that this approach does not sample from the true posterior distribution $p$, but instead from the empirical posterior distribution $\hat{p}$ given the (re-sampled) MAP estimates.
Intuitively, this can be understood as the difference between sampling from a distribution $p$ directly (Monte Carlo sampling) versus sampling from an approximate (empirical) distribution $\hat{p}$ (corresponding to the training data), which itself is constructed from samples of the true distribution $p$.
This approach is known as \midx{bootstrapping} or \midx{bagging} (short for \emph{b}ootstrap \emph{agg}regat\emph{ing}) and plays a central role in model-free reinforcement learning. We will return to this concept in~\cref{sec:tabular_rl:model_free:on_policy_value_estimation}.

\subsection{Diverse Probabilistic Ensembles}

Probabilistic ensembles can be loosely interpreted as randomly initializing $m$ ``particles''~$\{\vtheta^{(i)}\}_{i=1}^m$ and then pushing each particle towards regions of high posterior probability.
A potential issue with this approach is that if the initialization of particles is not sufficiently diverse, the particles may ``collapse'' since every particle eventually converges to a local optimum of the loss function.
A natural approach to mitigate this issue is to alter the objective of each particle from simply aiming to minimize the loss $- \log p(\vtheta \mid \vx_{1:n}, y_{1:n})$, which we will abbreviate by $- \log p(\vtheta)$, to also ``push'' the particles away from each other.
We will see next that this can be interpreted as a form of variational inference under a very flexible variational family.

In our discussion of \midx{variational inference}, we have seen that minimizing reverse-KL is equivalent to maximizing the evidence lower bound, and used this to derive an optimization algorithm to compute approximate posteriors.
We will discuss the alternative approach of directly computing the gradient of the KL-divergence.
Consider the variational family of all densities that can be expressed as smooth transformations of points sampled from a reference density $\phi$. That is, we consider $\spQ_\phi \defeq \{\pf{\vT}{\phi} \mid \text{$\vT : \Theta \to \Theta$ is smooth}\}$ where $\pf{\vT}{\phi}$ is the density of the random vector $\vtheta' = \vT(\vtheta)$ with $\vtheta \sim \phi$.\footnote{Refer back to \cref{sec:fundamentals:probability:cov} for a recap on pushforwards.}
The density $\phi$ can be thought of as the initial distribution of the particles, and the smooth map $\vT$ as the dynamics that push the particles towards the target density $p$.
It can be shown that for ``almost any'' reference density $\phi$, this variational family $\spQ_\phi$ is expressive enough to closely approximate ``almost arbitrary'' distributions.\footnote{For a more detailed discussion, refer to \icite{liu2016stein}.}
A natural approach is therefore to \emph{learn} the appropriate smooth map $\vT$ between the reference density $\phi$ and the target density $p$.

\begin{ex}{Gaussian variational inference}{}
  We have seen in \cref{sec:approximate_inference:variational_inference:elbo} that if $\phi$ is standard normal and $\vT(\vx; \{\vmu, \msqrt{\mSigma}\}) = \vmu + \msqrt{\mSigma} \vx$ an affine transformation then the ELBO can be maximized efficiently using stochastic gradient descent.
  However, in this case, $\spQ_\phi$ can only approximate Gaussian-like distributions since the expressivity of the map $\vT$ is limited under the fixed reference density $\phi$.
\end{ex}

An alternative approach to Gaussian variational inference is the following algorithm known as \midx{Stein variational gradient descent} (SVGD), where we recursively apply carefully chosen smooth maps to the current variational approximation: \begin{align}
  q_0 \xrightarrow{\opt{\vT_0}} q_1 \xrightarrow{\opt{\vT_1}} q_2 \xrightarrow{\opt{\vT_2}} \cdots \qquad\text{where $q_{t+1} \defeq \pf{\opt{\vT_t}}{q_t}$}. \label{eq:svgd_sketch}
\end{align}
We consider maps $\vT = \rvec{id} + \vf$ where $\rvec{id}(\vtheta) \defeq \vtheta$ denotes the identity map and $\vf(\vx)$ represents a (small) perturbation.
Recall that at time $t$ we seek to minimize $\KL{\pf{\vT}{q_t}}{p}$, so we choose the smooth map as \begin{align}
  \opt{\vT_t} \defeq \rvec{id} - \eta_t\! \left. \grad_\vf \KL{\pf{\vT}{q_t}}{p} \right\rvert_{\vf = \vzero} \label{eq:svgd_update}
\end{align} where $\eta_t$ is a step size.
In this way, the SVGD update \eqref{eq:svgd_update} can be interpreted as a step of ``functional'' gradient descent.

To be able to compute the gradient of the KL-divergence, we need to make some structural assumptions on the perturbation $\vf$.
SVGD assumes that $\vf = \transpose{[f_1 \; \cdots \; f_d]}$ with $f_i \in \spH_k(\Theta)$ is from some reproducing kernel Hilbert space $\spH_k(\Theta)$ of a positive definite kernel $k$; we say $\vf \in \spH_k^d(\Theta)$.
Within the RKHS, we can compute the gradient of the KL-divergence exactly.
\cite{liu2016stein} show that in this case, the functional gradient of the KL-divergence can be expressed in closed-form as $\left. -\grad_\vf \KL{\pf{\vT}{q}}{p} \right\rvert_{\vf = \vzero} = \opt{\vvarphi_{q,p}}$ where \begin{align}
  \opt{\vvarphi_{q,p}}(\cdot) \defeq \E[\vtheta \sim q]{k(\cdot,\vtheta) \grad_{\vtheta} \log p(\vtheta) + \grad_{\vtheta} k(\cdot,\vtheta)}.
\end{align}
SVGD then approximates $q$ using the particles $\{\vtheta^{(i)}\}_{i=1}^m$ as follows:

\begin{algorithm}
  \caption{Stein variational gradient descent, SVGD}\label{alg:svgd}
  initialize particles $\{\vtheta^{(i)}\}_{i=1}^m$\;
  \Repeat{converged}{
    \For{each particle $i \in [m]$}{
      $\vtheta^{(i)} \gets \vtheta^{(i)} + \eta_t \opt{\vvarphihat_{q,p}}(\vtheta^{(i)})$ where $\opt{\vvarphihat_{q,p}}(\vtheta) \defeq \frac{1}{m} \sum_{j=1}^m \Big[k(\vtheta,\vtheta^{(j)}) \grad_{\vtheta} \log p(\vtheta) + \grad_{\vtheta^{(j)}} k(\vtheta,\vtheta^{(j)})\Big]$\;
    }
  }
\end{algorithm}

Often, a Gaussian kernel~\eqref{eq:gaussian_kernel} with length scale $h$ is used to model the perturbations, in which case the repulsion term is \begin{align}
  \grad_{\vtheta^{(j)}} k(\vtheta,\vtheta^{(j)}) = \frac{1}{h^2} (\vtheta - \vtheta^{(j)}) k(\vtheta,\vtheta^{(j)})
\end{align} and the negative functional gradient simplifies to \begin{align}
  \opt{\vvarphihat_{q,p}}(\vtheta) = \frac{1}{m} \sum_{j=1}^m k(\vtheta,\vtheta^{(j)}) \Big[ \underbrace{\grad_{\vtheta} \log p(\vtheta)}_{\text{drift}} + \underbrace{h^{-2} (\vtheta - \vtheta^{(j)})}_{\text{repulsion}} \Big].
\end{align}

Note that SVGD has similarities to Langevin dynamics, which as seen in \exerciserefmark{langevin_dynamics_convergence} can also be interpreted as following a gradient of the KL-divergence.
Whereas Langevin dynamics perturbs particles according to a drift towards regions of high posterior probability and some random diffusion (cf.~\cref{eq:cont_time_langevin}), the first term of $\opt{\vvarphihat_{q,p}}(\vtheta)$ perturbs particles to drift towards regions of high posterior probability while the second term leads to a mutual ``repulsion'' of particles.
Notably, the perturbations of Langevin dynamics are noisy, while SVGD perturbs particles deterministically and the randomness is exclusively in the initialization of particles.
The repulsion term prevents particles from collapsing to a single mode of the posterior distribution, which is a possible failure mode of other particle-based posterior approximations such as ensembles.\looseness=-1

Note that the above decomposition of $\opt{\vvarphihat_{q,p}}(\vtheta)$ is once more an example of the \midx{principle of curiosity and conformity} which we have seen to be a recurring theme in approaches to approximate inference.
The repulsion term leads to exploration of the particles (i.e., ``curiosity'' about alternative explanations), while the drift term leads to minimization of the loss (i.e., ``conformity'' to the data).\looseness=-1

\cite{lu2019scaling} show that under some assumptions, SVGD converges asymptotically to the target density $p$ as $\eta_t \to 0$.
SVGD's name originates from \midx{Stein's method} which is a general-purpose approach for characterizing convergence in distribution.\footnote{For an introduction to Stein's method, read \icite{gorham2015measuring}.}

\section{Calibration}\pidx{calibration}

A key challenge of Bayesian deep learning (and also other probabilistic methods) is the calibration of models.
We say that a model is \emph{well-calibrated} if its confidence coincides with its accuracy across many predictions.
Consider a classification model that predicts that the label of a given input belongs to some class with probability $80\%$.
If the model is well-calibrated, then the prediction is correct about $80\%$ of the time.
In other words, during calibration, we adjust the probability estimation of the model.

We will first mention two methods of estimating the calibration of a model, namely the marginal likelihood and reliability diagrams.
Then, in \cref{sec:bdl:calibration:improving}, we survey commonly used heuristics for empirically improving the calibration.

\subsection{Evidence}

A popular method (which we already encountered multiple times) is to use the evidence of a validation set $\vx_{1:m}^\val$ of size $m$ given the training set $\vx_{1:n}^\train$ of size $n$ for estimating the model calibration.
Here, the evidence can be understood as describing how well the validation set is described by the model trained on the training set.
We obtain, \begin{align}
  &\log p(y_{1:m}^\val \mid \vx_{1:m}^\val, \vx_{1:n}^\train, y_{1:n}^\train) \nonumber \\
  &= \log \int p(y_{1:m}^\val \mid \vx_{1:m}^\val, \vtheta) p(\vtheta \mid \vx_{1:n}^\train, y_{1:n}^\train) \,d\vtheta \margintag{using the sum rule \eqref{eq:sum_rule} and product rule \eqref{eq:product_rule}} \nonumber \\
  &\approx \log \int p(y_{1:m}^\val \mid \vx_{1:m}^\val, \vtheta) q_\vlambda(\vtheta) \,d\vtheta \margintag{approximating with the variational posterior} \nonumber \\
  &= \log \int \prod_{i=1}^m p(y_i^\val \mid \vx_i^\val, \vtheta) q_\vlambda(\vtheta) \,d\vtheta \margintag{using the independence of the data}
  \intertext{The resulting integrals are typically very small which leads to numerical instabilities. Therefore, it is common to maximize a lower bound to the evidence instead,}
  &= \log \E[\vtheta \sim q_\vlambda]{\prod_{i=1}^m p(y_i^\val \mid \vx_i^\val, \vtheta)} \margintag{interpreting the integral as an expectation over the variational posterior} \nonumber \\
  &\geq \E[\vtheta \sim q_\vlambda]{\sum_{i=1}^m \log p(y_i^\val \mid \vx_i^\val, \vtheta)} \margintag{using Jensen's inequality \eqref{eq:jensen}} \\
  &\approx \frac{1}{k} \sum_{j=1}^k \sum_{i=1}^m \log p(y_i^\val \mid \vx_i^\val, \vtheta^{(j)}) \margintag{using Monte Carlo sampling}
\end{align} for independent samples $\vtheta^{(j)} \iid q_\vlambda$.

\subsection{Reliability Diagrams}\pidx{reliability diagram}

Reliability diagrams take a frequentist perspective to estimate the calibration of a model.
For simplicity, we assume a calibration problem with two classes, $1$ and $-1$ (similarly to logistic regression).\footnote{Reliability diagrams generalize beyond this restricted example.}

We group the predictions of a validation set into $M$ interval bins of size $\nicefrac{1}{M}$ according to the class probability predicted by the model, $\Pr{Y_i = 1 \mid \vx_i}$.
We then compare within each bin, how often the model thought the inputs belonged to the class (confidence) with how often the inputs actually belonged to the class (frequency).
Formally, we define $\sB_m$ as the set of samples falling into bin $m$ and let \begin{align}
  \mathrm{freq}(\sB_m) \defeq \frac{1}{\card{\sB_m}} \sum_{i \in \sB_m} \Ind{Y_i = 1}
\end{align} be the proportion of samples in bin $m$ that belong to class $1$ and let \begin{align}
  \mathrm{conf}(\sB_m) \defeq \frac{1}{\card{\sB_m}} \sum_{i \in \sB_m} \Pr{Y_i = 1 \mid \vx_i}
\end{align} be the average confidence of samples belonging to class $1$ within the bin~$m$.\looseness=-1

\begin{marginfigure}
  \begin{center}
    \import{./plots/output/}{reliability_diagrams.pgf}
  \end{center}

  \caption{Examples of reliability diagrams with ten bins.
  A perfectly calibrated model approximates the diagonal dashed red line.
  The first reliability diagram shows a well-calibrated model.
  In contrast, the second reliability diagram shows an overconfident model.}\label{fig:reliability_diagram}
\end{marginfigure}

Thus, a model is well calibrated if $\mathrm{freq}(B_m) \approx \mathrm{conf}(B_m)$ for each bin $m \in [M]$.
There are two common metrics of calibration that quantify how ``close'' a model is to being well calibrated.

\begin{enumerate}
  \item The \midx{expected calibration error} (ECE) is the average deviation of a model from perfect calibration, \begin{align}
    \ell_{\mathrm{ECE}} \defeq \sum_{m=1}^M \frac{\card{\sB_m}}{n} \abs{\mathrm{freq}(\sB_m) - \mathrm{conf}(\sB_m)}
  \end{align} where $n$ is the size of the validation set.

  \item The \midx{maximum calibration error} (MCE) is the maximum deviation of a model from perfect calibration among all bins, \begin{align}
    \ell_{\mathrm{MCE}} \defeq \max_{m \in [M]} \abs{\mathrm{freq}(\sB_m) - \mathrm{conf}(\sB_m)}.
  \end{align}
\end{enumerate}

\subsection{Heuristics for Improving Calibration}\label{sec:bdl:calibration:improving}

We now survey a few heuristics which can be used empirically to improve model calibration. \begin{enumerate}
  \item \midx<Histogram binning>{histogram binning} assigns a calibrated score $q_m \defeq \mathrm{freq}(\sB_m)$ to each bin during validation.
  Then, during inference, we return the calibrated score $q_m$ of the bin corresponding to the prediction of the model.

  \item \midx<Isotonic regression>{isotonic regression} extends histogram binning by using variable bin boundaries.
  We find a piecewise constant function $\vf \defeq [f_1, \dots, f_M]$ that minimizes the bin-wise squared loss, \begin{subequations}\begin{align}
    \min_{M, \vf, \va} \quad&\sum_{m=1}^M \sum_{i=1}^n \Ind{a_m \leq \Pr{Y_i = 1 \mid \vx_i} < a_{m+1}} (f_m - y_i)^2 \\
    \text{subject to} \quad&0 = a_1 \leq \cdots \leq a_{M+1} = 1, \\
    &f_1 \leq \cdots \leq f_M
  \end{align}\end{subequations} where $\vf$ are the calibrated scores and $\va \defeq [a_1, \dots, a_{M+1}]$ are the bin boundaries.
  We then return the calibrated score $f_m$ of the bin corresponding to the prediction of the model.

  \item \midx{Platt scaling} adjusts the logits $z_i$ of the output layer to \begin{align}
    q_i \defeq \sigma(a z_i + b)
  \end{align} and then learns parameters $a, b \in \R$ to maximize the likelihood.

  \item \midx<Temperature scaling>{temperature scaling} is a special and widely used instance of Platt scaling where $a \defeq \nicefrac{1}{T}$ and $b \defeq 0$ for some temperature scalar ${T > 0}$,\looseness=-1 \begin{align}
    q_i \defeq \sigma\parentheses*{\frac{z_i}{T}}.
  \end{align}
  Intuitively, for a larger temperature $T$, the probability is distributed more evenly among the classes (without changing the ranking), yielding a more uncertain prediction.
  In contrast, for a lower temperature $T$, the probability is concentrated more towards the top choices, yielding a less uncertain prediction.
  As seen in \cref{exercise:maximum_entropy_property_of_gibbs_distribution}, temperature scaling can be motivated as tuning the mean of the softmax distribution.
\end{enumerate}
\vspace{\baselineskip}

\begin{marginfigure}
  \begin{center}
    \import{./plots/output/}{temperature_scaling.pgf}
  \end{center}

  \caption{Illustration of temperature scaling for a classifier with three classes.
  On the top, we have a prediction with a high temperature, yielding a very uncertain prediction (in favor of class $A$).
  Below, we have a prediction with a low temperature, yielding a prediction that is strongly in favor of class $A$.
  Note that the ranking ($A \succ C \succ B$) is preserved.}
\end{marginfigure}

\begin{oreadings}
  \begin{itemize}
    \item \pcite{guo2017calibration}
    \item \pcite{blundell2015weight}
    \item \pcite{kendall2017uncertainties}
  \end{itemize}
\end{oreadings}

\section*{Discussion}

This chapter concludes our discussion of (approximate) probabilistic inference.
Across the last three chapters, we have seen numerous methods for approximating the posterior distributions of deep neural networks: \begin{itemize}
  \item Methods such as dropout and stochastic weight averaging are frequently used in practice.
  Other particle-based approaches such as ensembles and SVGD are used less frequently since they are computationally more expensive to train, but are some of the most effective methods in estimating uncertainty.

  \item Recently, Laplace approximations regained interest since they can be applied ``post-hoc'' after training simply by computing or approximating the Hessian of the loss function~\citep{daxberger2021laplace,antoran2022adapting}.
  Still, Laplace approximations come with the limitations inherent to unimodal Gaussian approximations.

  \item Other work, particularly in fine-tuning, has explored approximating the posterior distribution of deep neural networks by treating them as linear functions in a fixed learned feature space, in which case one can use the tools for exact probabilistic inference from \cref{sec:blr,sec:gp}~\citep[e.g.,][]{hubotter2025efficiently}.
\end{itemize}
Despite large progress in approximate inference over the past decade, efficient and reliable uncertainty estimation of large models remains an important open challenge.

\excheading

\begin{nexercise}{Softmax is a generalization of the logistic function}{softmax_and_logistic_function}
  Show that for a two-class classification problem (i.e., $c = 2$), the softmax function is equivalent to the logistic function \eqref{eq:logistic_function} for the univariate model $f \defeq f_1 - f_0$.
  That is, $\sigma_1(\vf) = \sigma(f)$ and $\sigma_0(\vf) = 1 - \sigma(f)$.

  Thus, the softmax function is a generalization of the logistic function to more than two classes.
\end{nexercise}

  \part{Sequential Decision-Making}\label{part2}
  \chapter*{Preface to \Cref{part2}}

In the first part of the \course, we have learned about how we can build machines that are capable of updating their beliefs and reducing their epistemic uncertainty through probabilistic inference.
We have also discussed ways of keeping track of the world through noisy sensory information by filtering.
An important aspect of intelligence is to use this acquired knowledge for making decisions and taking actions that have a positive impact on the world.

\begin{figure}
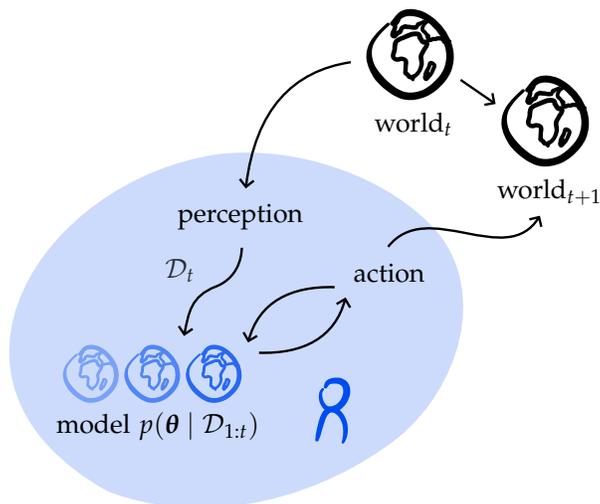

  \incfig{perception_action_diagram}
	\caption{An illustration of the perception-action loop. This is a straightforward extension of our view of probabilistic inference from \cref{fig:perception_diagram} with the addition of an ``action'' component which is capable of ``adaptively'' interacting with the outside world and the internal world model.}
  \label{fig:preception_action_loop}
\end{figure}

Already today, we are surrounded by machines that make decisions and take actions; that is, exhibit some degree of agency.
Be it a search engine producing a list of search results, a chatbot answering a question, or a driving-assistance system steering a car: these systems are all perceiving the world, making decisions, and then taking actions that in turn have an effect on the world.
\Cref{fig:preception_action_loop} illustrates this \midx{perception-action loop}.

In the second part of this \course, we will discuss the underpinning principles of building machines that are capable of making sequential decisions.
We will see that decision-making itself can be cast as probabilistic inference, obeying the same mechanisms that we used in the first part to build learning systems.

We discuss various ways of addressing the question: \begin{center}
  \emph{How to act, given that computational resources and time are limited?}
\end{center}
One approach is to act with the aim to reduce epistemic uncertainty, which is the topic of active learning.
Another approach is to act with the aim to maximize some reward signal, which is the topic of bandits, Bayesian optimization, and reinforcement learning.

This surfaces the exploration-exploitation dilemma where the agent has to prioritize either maximizing its immediate rewards or reducing its uncertainty about the world which might pay off in the future.
We discuss that this dilemma is, in fact, in direct correspondence to the principle of curiosity and conformity which we discussed extensively throughout \cref{part1}.

Since time is limited, it is critical to be sample-efficient when learning the most important aspects of the world.
At the same time, interactions with the world are often complex, and some interactions might even be harmful.
We discuss how an agent can use its epistemic uncertainty to guide the exploration of its environment while mitigating risks and reasoning about safety.

  \chapter{Active Learning}\label{sec:active_learning}

By now, we have seen that probabilistic machine learning is very useful for estimating the uncertainty in our models (epistemic uncertainty) and in the data (aleatoric uncertainty).
We have been focusing on the setting of supervised learning\pidx{supervised learning} where we are given a set $\spD = \{(\vx_i, y_i)\}_{i=1}^n$ of labeled data, yet we often encounter settings where we have only little data and acquiring new data is costly.

In this chapter --- and in the following chapter on Bayesian optimization --- we will discuss how one can use uncertainty to effectively collect more data.
In other words, we want to figure out where in the domain we should sample to obtain the most useful information.
Throughout most of this chapter, we focus on the most common way of quantifying ``useful information'', namely the expected reduction in entropy which is also called the \midx{mutual information}.

\section{Conditional Entropy}

We begin by introducing the notion of conditional entropy.
Recall that the entropy $\H{\rX}$ of a random vector $\rX$ can be interpreted as our average surprise when observing realizations $\vx \sim \rX$.
Thus, entropy can be considered as a quantification of the uncertainty about a random vector (or equivalently, its distribution).\footnote{We discussed entropy extensively in \cref{sec:approximate_inference:information_theory}.}

A natural extension is to consider the entropy of $\rX$ given the occurrence of an event corresponding to another random variable (e.g., $\rY = \vy$ for a random vector $\rY$), \begin{align}
  \H{\rX \mid \rY = \vy} \defeq \E[\vx \sim p(\vx \mid \vy)]{- \log p(\vx \mid \vy)}.
\end{align}
Instead of averaging over the surprise of samples from the distribution~$p(\vx)$ (like the entropy $\H{\rX}$), this quantity simply averages over the surprise of samples from the conditional distribution $p(\vx \mid \vy)$.

\begin{defn}[Conditional entropy]\pidx{conditional entropy}
  The \emph{conditional entropy} of a random vector $\rX$ given the random vector $\rY$ is defined as \begin{align}
    \H{\rX \mid \rY} &\defeq \E[\vy \sim p(\vy)]{\H{\rX \mid \rY = \vy}} \label{eq:cond_entropy} \\
    &= \E[(\vx, \vy) \sim p(\vx, \vy)]{- \log p(\vx \mid \vy)}.
  \end{align}
\end{defn}
Intuitively, the conditional entropy of $\rX$ given $\rY$ describes our average surprise about realizations of~$\rX$ given a particular realization of $\rY$, averaged over all such possible realizations of~$\rY$.
In other words, conditional entropy corresponds to the expected remaining uncertainty in~$\rX$ after we observe~$\rY$.
Note that, in general, $\H{\rX \mid \rY} \neq \H{\rY \mid \rX}$.

It is crucial to stress the difference between $\H{\rX \mid \rY = \vy}$ and the conditional entropy $\H{\rX \mid \rY}$.
The former simply corresponds to a probabilistic update of our uncertainty in $\rX$ after we have observed the realization $\vy \sim \rY$.
In contrast, conditional entropy \emph{predicts} how much uncertainty will remain about $\rX$ (in expectation) after we \emph{will} observe~$\rY$.\looseness=-1

\begin{defn}[Joint entropy]\pidx{joint entropy}
  One can also define the \emph{joint entropy} of random vectors $\rX$ and $\rY$, \begin{align}
    \H{\rX, \rY} &\defeq \E[(\vx, \vy) \sim p(\vx, \vy)]{- \log p(\vx, \vy)},
  \end{align} as the combined uncertainty about $\rX$ and $\rY$.
  Observe that joint entropy is symmetric.
\end{defn}

This gives the \midx{chain rule for entropy}, \begin{align}
  \H{\rX, \rY} &= \H{\rY} + \H{\rX \mid \rY} \label{eq:chain_rule_entropy} \margintag{using the product rule \eqref{eq:product_rule} and the definition of conditional entropy \eqref{eq:cond_entropy}} \\[5pt]
  &= \H{\rX} + \H{\rY \mid \rX}. \label{eq:chain_rule_entropy2} \margintag{using symmetry of joint entropy}
\end{align}
That is, the joint entropy of $\rX$ and $\rY$ is given by the uncertainty about~$\rX$ and the additional uncertainty about $\rY$ given $\rX$.
Moreover, this also yields \midx{Bayes' rule for entropy}, \begin{align}
  \H{\rX \mid \rY} &= \H{\rY \mid \rX} + \H{\rX} - \H{\rY}. \label{eq:bayes_rule_entropy} \margintag{using the chain rule for entropy \eqref{eq:chain_rule_entropy} twice}
\end{align}

A very intuitive property of entropy is its monotonicity: when conditioning on additional observations the entropy can never increase, \begin{align}
  \H{\rX \mid \rY} \leq \H{\rX}. \label{eq:information_never_hurts}
\end{align}
Colloquially, this property is also called the ``\midx{information never hurts}[idxpagebf]'' principle.
We will derive a proof in the following section.

\section{Mutual Information}\label{sec:mutual_information}

Recall that our fundamental objective is to reduce entropy, as this corresponds to reduced uncertainty in the variables, which we want to predict.
Thus, we are interested in how much information we ``gain'' about the random vector $\rX$ by choosing to observe a random vector $\rY$.
By our interpretation of conditional entropy from the previous section, this is described by the following quantity.

\begin{marginfigure}[-\baselineskip]
  \begin{center}
    \import{./plots/output/}{information_gain.pgf}
  \end{center}

  \caption{Information gain.
  The first graph shows the prior.
  The second graph shows a selection of samples with large information gain (large uncertainty reduction).
  The third graph shows a selection of samples with small information gain (small uncertainty reduction).}
\end{marginfigure}

\begin{defn}[Mutual information, MI]\pidx{mutual information}[idxpagebf]
  The \emph{mutual information} of $\rX$ and $\rY$ (also known as the \midx{information gain}) is defined as \begin{align}
    \I{\rX}{\rY} &\defeq \H{\rX} - \H{\rX \mid \rY} \label{eq:mi} \\
    &= \H{\rX} + \H{\rY} - \H{\rX, \rY}.
  \end{align}
\end{defn}

In words, we subtract the uncertainty left about $\rX$ after observing $\rY$ from our initial uncertainty about $\rX$.
This measures the reduction in our uncertainty in $\rX$ (as measured by entropy) upon observing $\rY$.
Unlike conditional entropy, it follows from the definition that mutual information is symmetric.
That is, \begin{align}
  \I{\rX}{\rY} = \I{\rY}{\rX}. \label{eq:mi_symmetry}
\end{align}

Thus, the mutual information between $\rX$ and $\rY$ can be understood as the approximation error (or information loss) when assuming that $\rX$ and $\rY$ are independent.

\begin{marginfigure}[0.5\baselineskip]
  \incfig{information_gain_venn}
  \vspace{-10pt}
  \caption{Relationship between mutual information and entropy, expressed as a Venn diagram.}
\end{marginfigure}

In particular, using Gibbs' inequality (cf. \cref{exercise:gibbs_ineq}), this relationship shows that $\I{\rX}{\rY} \geq 0$ with equality when $\rX$ and $\rY$ are independent, and also proves the \midx{information never hurts} principle \eqref{eq:information_never_hurts} as \begin{align}
  0 \leq \I{\rX}{\rY} = \H{\rX} - \H{\rX \mid \rY}. \label{eq:mi_non_neg}
\end{align}

\begin{ex}{Mutual information of Gaussians}{mi_of_gaussians}
  Given the Gaussian random vector ${\rX \sim \N{\vmu}{\mSigma}}$ and the noisy observation ${\rY = \rX + \vvarepsilon}$ where ${\vvarepsilon \sim \N{\vzero}{\sigman^2 \mI}}$, we want to find the information gain of $\rX$ when observing $\rY$.
  Using our definitions from this chapter, we obtain \begin{align}
    \I{\rX}{\rY} &= \I{\rY}{\rX} \margintag{using symmetry \eqref{eq:mi_symmetry}} \nonumber \\
    &= \H{\rY} - \H{\rY \mid \rX} \margintag{by mutual information \eqref{eq:mi}} \nonumber \\
    &= \H{\rY} - \H{\vvarepsilon} \margintag{given $\rX$, the only randomness in $\rY$ originates from $\vvarepsilon$} \nonumber \\
    &= \frac{1}{2} \log\parentheses*{(2 \pi e)^d \det{\mSigma + \sigman^2 \mI}} - \frac{1}{2} \log\parentheses*{(2 \pi e)^d \det{\sigman^2 \mI}} \margintag{using the entropy of Gaussians \eqref{eq:entropy_gaussian}} \nonumber \\
    &= \frac{1}{2} \log \frac{\det{\mSigma + \sigman^2 \mI}}{\det{\sigman^2 \mI}} \nonumber \\
    &= \frac{1}{2} \log \det{\mI + \sigman^{-2} \mSigma}. \label{eq:mi_cond_linear_gaussians}
  \end{align}
  Intuitively, the larger the noise $\sigman^2$ in relation to the covariance of~$\rX$, the smaller the information gain.
\end{ex}

\subsection{Synergy and Redundancy}\label{sec:synergy_and_redundancy}

It is sometimes useful to write down the mutual information of~$\rX$ and~$\rY$ conditioned (in expectation) on a third random vector~$\rZ$.
This leads us to the following definition.

\begin{defn}[Conditional mutual information]\pidx{conditional mutual information}
  The \emph{conditional mutual information} of $\rX$ and $\rY$ given $\rZ$ is defined as \begin{align}
    \I{\rX}{\rY}[\rZ] &\defeq \H{\rX \mid \rZ} - \H{\rX \mid \rY, \rZ}. \label{eq:cond_mi} \\
    &= \H{\rX, \rZ} + \H{\rY, \rZ} - \H{\rZ} - \H{\rX, \rY, \rZ} \margintag{using the relationship of joint and conditional entropy \eqref{eq:chain_rule_entropy}} \\
    &=\I{\rX}{\rY, \rZ} - \I{\rX}{\rZ}. \label{eq:cond_mi_joint_mi}
  \end{align}
\end{defn}
Thus, the conditional mutual information corresponds to the reduction of uncertainty in $\rX$ when observing $\rY$, given we already observed $\rZ$.
It also follows that conditional mutual information is symmetric: \begin{align}
  \I{\rX}{\rY}[\rZ] = \I{\rY}{\rX}[\rZ]. \label{eq:cond_mi_symmetry}
\end{align}
We have seen in this chapter that entropy is monotonically decreasing as we condition on new information, and called this the ``information never hurts'' principle \eqref{eq:information_never_hurts}.
However, the same does not hold for mutual information!
That is, information about a random vector $\rZ$ may reduce the mutual information between random vectors $\rX$ and $\rY$ \exerciserefmark{cond_mi_non_monotonicity}.\looseness=-1

\begin{rmk}{Sufficient statistics and data processing inequality}{}
  A related concept is the data processing inequality \eqref{eq:data_processing_inequality} which you prove in \exerciserefmark{cond_mi_non_monotonicity}[3] and which allows us to formalize a concept which we have seen multiple times already, namely the notion of a sufficient statistic.
  Consider the Markov chain $\vlambda \to \rX \to \vs(\rX)$, for example, $\vlambda$ may be parameters of the distribution of $\rX$.
  By the data processing inequality \eqref{eq:data_processing_inequality}, $\I{\vlambda}{\vs(\rX)} \leq \I{\vlambda}{\rX}$.
  If the data processing inequality is satisfied with equality then $\vs(\rX)$ is called a \midx{sufficient statistic}[idxpagebf] of $\rX$ for the inference of $\vlambda$.
\end{rmk}

To understand the behavior of mutual information under conditioning, it is helpful to consider the \midx{interaction information} \begin{align}
  \I{\rX}{\rY ; \rZ} \defeq \I{\rX}{\rY} - \I{\rX}{\rY \mid \rZ}. \label{eq:interaction_information}
\end{align}
If the interaction is positive then some information about $\rX$ that is provided by $\rY$ is also provided by $\rZ$ (i.e., conditioning on $\rZ$ decreases MI between $\rX$ and $\rY$), and we say that there is \midx{redundancy} between $\rY$ and $\rZ$ (with respect to $\rX$).
Conversely, if the interaction is negative then learning about $\rZ$ increases what can be learned from $\rY$ about $\rX$, and we say that there is \midx{synergy} between $\rY$ and $\rZ$.
We will see later in this chapter that the absence of synergies can lead to efficient algorithms for maximizing mutual information.

\subsection{Mutual Information as Utility Function}

Following our introduction of mutual information, it is natural to answer the question ``where should I collect data?'' by saying ``wherever mutual information is maximized''.
More concretely, assume we are given a set $\spX$ of possible observations of $f$, where $y_\vx$ denotes a single such observation at $\vx \in \spX$, \begin{align}
  y_\vx \defeq f_\vx + \varepsilon_\vx,
\end{align} $f_\vx \defeq f(\vx)$, and $\varepsilon_\vx$ is some zero-mean Gaussian noise.
For a set of observations $\sS = \{\vx_1, \dots, \vx_n\}$, we can write $\vy_\sS = \vf_\sS + \vvarepsilon$ where \begin{align*}
  \vy_\sS \defeq \begin{bmatrix}
    y_{\vx_1} \\ \vdots \\ y_{\vx_n}
  \end{bmatrix}, \quad \vf_\sS \defeq \begin{bmatrix}
    f_{\vx_1} \\ \vdots \\ f_{\vx_n}
  \end{bmatrix}, \quad\text{and}\quad \vvarepsilon \sim \N{\vzero}{\sigman^2 \mI}.
\end{align*}
Note that both $\vy_\sS$ and $\vf_\sS$ are random vectors.
Our goal is then to find a subset $\sS \subseteq \spX$ of size $n$ maximizing the information gain between our model $f$ and $\vy_\sS$.

This yields the maximization objective, \begin{align}
  I(\sS) \defeq \I{\vf_\sS}{\vy_\sS} = \H{\vf_\sS} - \H{\vf_\sS \mid \vy_\sS}. \label{eq:mi_optimization}
\end{align}
Here, $\Hsm{\vf_\sS}$ corresponds to the uncertainty about $\vf_\sS$ before obtaining the observations $\vy_\sS$ and $\Hsm{\vf_\sS \mid \vy_\sS}$ corresponds to the uncertainty about $\vf_\sS$, in expectation, after obtaining the observations $\vy_\sS$.

\begin{rmk}{Making optimal decisions with intrinsic rewards}{}
  Note that this objective function maps neatly onto our initial consideration of making optimal decisions under uncertainty from \cref{sec:decision_theory}.
  In fact, you can think of maximizing $I(\sS)$ simply as computing the optimal decision rule for the utility \begin{align}
    r(\vy_\sS, \sS) = \H{\vf_\sS} - \H{\vf_\sS \mid \rY_\sS = \vy_\sS}, \margintag{with $\rY_\sS = \vy_\sS$ denoting an event}
  \end{align} with $I(\sS) = \E[{\vy_{\sS}}]{r(\vy_\sS, \sS)}$ measuring the expected utility of observations~$\vy_\sS$.
  Such a utility or reward function is often called an \midx{intrinsic reward} since it does not measure an ``objective'' external quantity, but instead a ``subjective'' quantity that is internal to the model of~$f$.\looseness=-1
\end{rmk}

Observe that picking a subset of points $\sS \subseteq \spX$ from the domain $\spX$ is a combinatorial problem.
That is to say, we are optimizing a function over discrete sets.
In general, such combinatorial optimization problems tend to be very difficult.
It can be shown that maximizing mutual information is $\mathcal{NP}$-hard.

\section{Submodularity of Mutual Information}

 We will look at optimizing mutual information in the following section.
 First, we want to introduce the notion of submodularity which is important in the analysis of discrete functions.

\begin{defn}[Marginal gain]\pidx{marginal gain}
  Given a (discrete) function ${F : \pset{\spX} \to \R}$, the \emph{marginal gain} of $\vx \in \spX$ given $\sA \subseteq \spX$ is defined as \begin{align}
    \Delta_F(\vx \mid \sA) \defeq F(\sA \cup \{\vx\}) - F(\sA). \label{eq:marginal_gain}
  \end{align}
  Intuitively, the marginal gain describes how much ``adding'' the additional $\vx$ to $\sA$ increases the value of $F$.
\end{defn}
When maximizing mutual information, the marginal gain is \exerciserefmark{marginal_gain_mi} \begin{align}
  \Delta_I(\vx \mid \sA) &= \Ism{f_\vx}{y_\vx}[\vy_\sA] \label{eq:mg_mi} \\
  &= \H{y_{\vx} \mid \vy_\sA} - \H{\varepsilon_\vx}. \label{eq:mg_decomp}
\end{align}
That is, when maximizing mutual information, the marginal gain corresponds to the difference between the uncertainty after observing $\vy_\sA$ and the entropy of the noise $\H{\varepsilon_\vx}$.
Altogether, the marginal gain represents the reduction in uncertainty by observing $\{\vx\}$.

\begin{marginfigure}
  \incfig{submodularity}
  \caption{Monotone submodularity.
  The effect of ``adding'' $\vx$ to the smaller set $\sA$ is larger than the effect of adding $\vx$ to the larger set $\sB$.}
\end{marginfigure}

\begin{defn}[Submodularity]\pidx{submodularity}
  A (discrete) function $F : \pset{\spX} \to \R$ is \emph{submodular} iff for any $\vx \in \spX$ and any $\sA \subseteq \sB \subseteq \spX$ it satisfies \begin{align}
    F(\sA \cup \{\vx\}) - F(\sA) \geq F(\sB \cup \{\vx\}) - F(\sB). \label{eq:submodularity}
  \end{align}
\end{defn}

Equivalently, using our definition of marginal gain, we have that $F$ is submodular iff for any $\vx \in \spX$ and any $\sA \subseteq \sB \subseteq \spX$, \begin{align}
  \Delta_F(\vx \mid \sA) \geq \Delta_F(\vx \mid \sB). \label{eq:submodularity_mg}
\end{align}
That is, ``adding'' $\vx$ to the smaller set $\sA$ yields more marginal gain than adding $\vx$ to the larger set $\sB$.
In other words, the function $F$ has ``diminishing returns''.
In this way, submodularity can be interpreted as a notion of ``concavity'' for discrete functions.

\begin{defn}[Monotone submodularity]\pidx{monotone submodularity}\label{defn:monotone_submodularity}
  A function $F : \pset{\spX} \to \R$ is called \emph{monotone} iff for any $\sA \subseteq \sB \subseteq \spX$ it satisfies \begin{align}
    F(\sA) \leq F(\sB). \label{eq:monotonicity}
  \end{align}
  If $F$ is also submodular, then $F$ is called \emph{monotone submodular}.
\end{defn}

\begin{thm}\label{thm:monotone_submodularity_of_mi}
  The objective $I$ is monotone submodular.
\end{thm}\vspace{-10pt}
\begin{proof}
  We fix arbitrary subsets $\sA \subseteq \sB \subseteq \spX$ and any $\vx \in \spX$. We have, \begin{align*}
    \text{$I$ is submodular} \iff&& \Delta_I(\vx \mid \sA) &\geq \Delta_I(\vx \mid \sB) \margintag{by submodularity in terms of marginal gain \eqref{eq:submodularity_mg}} \\[8pt]
    \iff&& \H{y_{\vx} \mid \vy_\sA} - \H{\varepsilon_\vx} &\geq \H{y_{\vx} \mid \vy_\sB} - \H{\varepsilon_\vx} \margintag{using \cref{eq:mg_decomp}} \\
    \iff&& \H{y_{\vx} \mid \vy_\sA} &\geq \H{y_{\vx} \mid \vy_\sB}. \margintag{$\H{\varepsilon_\vx}$ cancels}
  \end{align*}
  Due to the ``information never hurts'' principle \eqref{eq:information_never_hurts} of entropy and as $\sA \subseteq \sB$, this is always true.
  Moreover, \begin{align*}
    \text{$I$ is monotone} \iff&& I(\sA) &\leq I(\sB) \margintag{by the definition of monotinicity \eqref{eq:monotonicity}} \\
    \iff&& \I{\vf_\sA}{\vy_\sA} &\leq \I{\vf_\sB}{\vy_\sB} \margintag{using the definition of $I$ \eqref{eq:mi_optimization}} \\
    \iff&& \I{\vf_\sB}{\vy_\sA} &\leq \I{\vf_\sB}{\vy_\sB} \margintag{using $\I{\vf_\sB}{\vy_\sA} = \I{\vf_\sA}{\vy_\sA}$ as $\vy_\sA \perp \vf_\sB \mid \vf_\sA$} \\[8pt]
    \iff&& \H{\vf_\sB} - \H{\vf_\sB \mid \vy_\sA} &\leq \H{\vf_\sB} - \H{\vf_\sB \mid \vy_\sB} \margintag{using the definition of MI \eqref{eq:mi}} \\
    \iff&& \H{\vf_\sB \mid \vy_\sA} &\geq \H{\vf_\sB \mid \vy_\sB}, \margintag{$\H{\vf_\sB}$ cancels}
  \end{align*} which is also satisfied due to the ``information never hurts'' principle~\eqref{eq:information_never_hurts}.\looseness=-1
\end{proof}

The submodularity of $I$ can be interpreted from the perspective of information theory.
It turns out that submodularity is equivalent to the absence of synergy between observations \exerciserefmark{submodularity_and_no_synergy}.
Intuitively, without synergies, acting greedily is enough to find a near-optimal solution.
If there are synergies, then the combinatorial search problem is much harder, because single-step optimal actions do not necessarily lead us to the optimal solution.
Consider the extreme case of having to solve a ``needle in a haystack'' problem, where only a single subset of $\spX$ with size $k$ achieves objective value $1$, with all other subsets achieving objective value $0$.
In this case, we can do nothing but exhaustively search through all $\abs{\spX}^k$ combinations to find the optimal solution.

\section{Maximizing Mutual Information}

As we cannot efficiently pick a set $\sS \subseteq \spX$ to maximize mutual information but know that $I$ is submodular, a natural approach is to maximize mutual information greedily.
That is, we pick the locations $\vx_1$ through $\vx_n$ individually by greedily finding the location with the maximal mutual information.
The following general result for monotone submodular function maximization shows that, indeed, this greedy approach provides a good approximation.

\begin{thm}[Greedy submodular function maximization]
  If the set function ${F : \spP(\spX) \to \Rzero}$ is monotone submodular, then greedily maximizing $F$ is a $(1 - \nicefrac{1}{e})$-approximation:\footnote{$1 - \nicefrac{1}{e} \approx 0.632$} \begin{align}
    F(\sS_n) \geq \parentheses*{1 - \frac{1}{e}} \max_{\substack{\sS \subseteq \spX \\ \card{\sS} = n}} F(\sS).
  \end{align}
\end{thm}\vspace{-10pt}
\begin{proof}
  Fix any $n \geq 1$. Let $\sSs \in \argmax\{F(\sS) \mid \sS \in \spX, \card{\sS} \leq n\}$.
  We can assume $\card{\sSs} = n$ due to the monotonicity \eqref{eq:monotonicity} of $F$.
  We write, $\{\opt{\vx_1}, \dots, \opt{\vx_n}\} \defeq \sSs$.
  We have, \begin{align*}
    F(\sSs) &\leq F(\sSs \cup \sS_t) \margintag{using monotonicity \eqref{eq:monotonicity}} \\
    &= F(\sS_t) + \sum_{i=1}^n \Delta_F(\opt{\vx_i} \mid \sS_t \cup \{\opt{\vx_1}, \dots, \opt{\vx_{i-1}}\}) \margintag{using the definition of marginal gain \eqref{eq:marginal_gain}} \\
    &\leq F(\sS_t) + \sum_{\vxs \in \sSs} \Delta_F(\vxs \mid \sS_t) \margintag{using submodularity \eqref{eq:submodularity_mg}} \\
    &\leq F(\sS_t) + n \parentheses*{F(\sS_{t+1}) - F(\sS_t)}. \margintag{using that $\sS_{t+1} = \sS_t \cup \{\vx\}$ is chosen such that $\Delta_F(\vx \mid \sS_t)$ is maximized \eqref{eq:greedy_mi}}
  \end{align*}

  Let $\delta_t \defeq F(\sSs) - F(\sS_t)$. Then, \begin{align*}
    \delta_t = F(\sSs) - F(\sS_t) &\leq n \parentheses*{F(\sS_{t+1}) - F(\sS_t)} = n \parentheses*{\delta_t - \delta_{t+1}}.
  \end{align*}
  This implies $\delta_{t+1} \leq (1 - \nicefrac{1}{n}) \delta_t$ and $\delta_n \leq (1 - \nicefrac{1}{n})^n \delta_0 \leq \nicefrac{\delta_0}{e}$, using the well-known inequality $1 - x \leq e^{-x}$ for all $x \in \R$.

  Finally, observe that $\delta_0 = F(\sSs) - F(\emptyset) \leq F(\sSs)$ due to the non-negativity of $F$.
  We obtain, \begin{align*}
    \delta_n = F(\sSs) - F(\sS_n) \leq \frac{\delta_0}{e} \leq \frac{F(\sSs)}{e}.
  \end{align*}
  Rearranging the terms yields the theorem.
\end{proof}

\begin{oreadings}
  The original proof of greedy maximization for submodular functions was given by \icite{greedy_mi}.

  For more background on maximizing submodular functions, see \icite{krause2014submodular}.\looseness=-1
\end{oreadings}

Now that we have established that greedy maximization of mutual information is a decent approximation to maximizing the joint information of data, we will look at how this optimization problem can be solved in practice.\looseness=-1

\subsection{Uncertainty Sampling}\label{sec:active_learning:optimizing_mutual_information:uncertainty_sampling}

When maximizing mutual information, at time $t$ when we have already picked $\sS_t = \{\vx_1, \dots, \vx_t\}$, we need to solve the following optimization problem, \begin{align}
  \vx_{t+1} &\defeq \argmax_{\vx \in \spX} \Delta_I(\vx \mid \sS_t) \label{eq:greedy_mi} \\
  &= \argmax_{\vx \in \spX} \Ism{f_\vx}{y_{\vx} \mid \vy_{\sS_t}}. \margintag{using \cref{eq:mg_mi}}
  \intertext{Note that $f_\vx$ and $y_{\vx}$ are univariate random variables.
  Thus, using our formula for the mutual information of conditional linear Gaussians~\eqref{eq:mi_cond_linear_gaussians}, we can simplify to,}
  &= \argmax_{\vx \in \spX} \frac{1}{2} \log\parentheses*{1 + \frac{\sigma_t^2(\vx)}{\sigman^2}} \label{eq:uncertainty_sampling1}
  \intertext{where $\sigma_t^2(\vx)$ is the (remaining) variance at $\vx$ after observing $\sS_t$. Assuming the label noise is independent of $\vx$ (i.e., homoscedastic),}
  &= \argmax_{\vx \in \spX} \sigma_t^2(\vx).
\end{align}
Therefore, if $f$ is modeled by a Gaussian and we assume homoscedastic noise, greedily maximizing mutual information corresponds to simply picking the point $\vx$ with the largest variance.
This algorithm is also called \midx{uncertainty sampling}.

\subsection{Heteroscedastic Noise}\label{sec:active_learning:optimizing_mutual_information:heteroscedastic_noise}

\begin{marginfigure}
  \begin{center}
    \import{./plots/output/}{uncertainty_sampling_het_noise.pgf}
  \end{center}

  \caption{Uncertainty sampling with heteroscedastic noise.
  The epistemic uncertainty of the model is shown in a dark gray.
  The aleatoric uncertainty of the data is shown in a light gray.
  Uncertainty sampling would repeatedly pick points around $\xs$ as they maximize the epistemic uncertainty, even though the aleatoric uncertainty at $\xs$ is much larger than at the boundary.}\label{fig:uncertainty_sampling_heterscedastic_noise}
\end{marginfigure}

Uncertainty sampling is clearly problematic if the noise is heteroscedastic.
If there are a particular set of inputs with a large aleatoric uncertainty dominating the epistemic uncertainty, uncertainty sampling will continuously choose those points even though the epistemic uncertainty will not be reduced substantially (cf.~\cref{fig:uncertainty_sampling_heterscedastic_noise}).

Looking at \cref{eq:uncertainty_sampling1} suggests a natural fix.
Instead of only considering the epistemic uncertainty $\sigma_t^2(\vx)$, we can also consider the aleatoric uncertainty $\sigman^2(\vx)$ by modeling heteroscedastic noise, yielding \begin{align}
  \vx_{t+1} \defeq \argmax_{\vx \in \spX} \frac{1}{2} \log\parentheses*{1 + \frac{\sigma_t^2(\vx)}{\sigman^2(\vx)}} = \argmax_{\vx \in \spX} \frac{\sigma_t^2(\vx)}{\sigman^2(\vx)}.
\end{align}

Thus, we choose locations that trade large epistemic uncertainty with large aleatoric uncertainty.
Ideally, we find a location where the epistemic uncertainty is large, and the aleatoric uncertainty is low, which promises a significant reduction of uncertainty around this location.

\subsection{Classification}\label{sec:active_learning:optimizing_mutual_information:classification}

While we focused on regression, one can apply active learning also for other settings, such as (probabilistic) classification.
In this setting, for any input $\vx$, a model produces a categorical distribution over labels~$y_{\vx}$.\footnote{see \cref{sec:fundamentals:supervised_learning}}
Here, uncertainty sampling corresponds to selecting samples that maximize the entropy of the predicted label $y_{\vx}$, \begin{align}
  \vx_{t+1} &\defeq \argmax_{\vx \in \spX} \H{y_{\vx} \mid \vx_{1:t}, y_{1:t}}. \label{eq:uncertainty_sampling_classification1}
\end{align}
The entropy of a categorical distribution is simply a finite sum of weighted surprise terms.

\begin{figure}
  \begin{center}
    \import{./plots/output/}{uncertainty_sampling_classification_working.pgf}
    \import{./plots/output/}{uncertainty_sampling_classification.pgf}
  \end{center}
  \caption{Uncertainty sampling in classification.
  The area with high uncertainty (as measured by entropy) is highlighted in yellow.
  The shown figures display each a sequence of model updates, each after one new observation.
  In the left figure, the classes are well-separated and uncertainty is dominated by epistemic uncertainty, whereas in the right figure the uncertainty is dominated by noise.
  In the latter case, if we mostly choose points $\vx$ in the area of highest uncertainty (i.e., close to the decision boundary) to make observations, the label noise results in frequently changing models.\looseness=-1}
\end{figure}

This approach generally leads to sampling points that are close to the decision boundary.
Often, the uncertainty is mainly dominated by label noise rather than epistemic uncertainty, and hence, we do not learn much from our observations.
This is a similar problem to the one we encountered with uncertainty sampling in the setting of heteroscedastic noise.\looseness=-1

This naturally suggests distinguishing between the aleatoric and epistemic uncertainty of the model $f$ (parameterized by $\vtheta$).
To this end, mutual information can be used similarly as we have done with uncertainty sampling for regression, \begin{align}
  \vx_{t+1} &\defeq \argmax_{\vx \in \spX} \I{\vtheta}{y_{\vx} \mid \vx_{1:t}, y_{1:t}} \\
  &= \argmax_{\vx \in \spX} \I{y_{\vx}}{\vtheta \mid \vx_{1:t}, y_{1:t}} \margintag{using symmetry \eqref{eq:mi_symmetry}} \nonumber \\
  &= \argmax_{\vx \in \spX} \H{y_{\vx} \mid \vx_{1:t}, y_{1:t}} - \H{y_{\vx} \mid \vtheta, \vx_{1:t}, y_{1:t}} \margintag{using the definition of mutual information \eqref{eq:mi}} \nonumber \\
  &= \argmax_{\vx \in \spX} \H{y_{\vx} \mid \vx_{1:t}, y_{1:t}} - \b{\E[\vtheta \mid \vx_{1:t}, y_{1:t}]{\H{y_{\vx} \mid \vtheta, \vx_{1:t}, y_{1:t}}}} \label{eq:uncertainty_sampling_classification2} \margintag{using the definition of conditional entropy \eqref{eq:cond_entropy}} \\
  &= \argmax_{\vx \in \spX} \underbrace{\H{y_{\vx} \mid \vx_{1:t}, y_{1:t}}}_{\substack{\text{entropy of} \\ \text{predictive posterior}}} - \E*[\vtheta \mid \vx_{1:t}, y_{1:t}]{\underbrace{\H{y_{\vx} \mid \vtheta}}_{\substack{\text{entropy of} \\ \text{likelihood}}}}. \margintag{using the definition of entropy \eqref{eq:entropy} and assuming $y_{\vx} \perp \vx_{1:t}, y_{1:t} \mid \vtheta$}
\end{align}

The first term measures the entropy of the averaged prediction while the second term measures the average entropy of predictions.
Thus, the first term looks for points where the average prediction is not confident.
In contrast, the second term penalizes points where many of the sampled models are not confident about their prediction, and thus looks for points where the models are confident in their predictions.
This identifies those points~$\vx$ where the models \emph{disagree} about the label~$y_{\vx}$ (that is, each model is ``confident'' but the models predict different labels).
For this reason, this approach is known as \midx{Bayesian active learning by disagreement} (BALD).

Note that the second term of the difference acts as a regularizer when compared to \cref{eq:uncertainty_sampling_classification1}. The second term mirrors our description of \b{aleatoric uncertainty} from \cref{sec:blr:uncertainty}.
Recall that we interpreted aleatoric uncertainty as the average uncertainty for all models.
Crucially, here we use entropy to ``measure'' uncertainty, whereas previously we have been using variance.
Therefore, intuitively, \cref{eq:uncertainty_sampling_classification2} subtracts the aleatoric uncertainty from the total uncertainty about the label.

Observe that both terms require approximate forms of the posterior distribution.
In \cref{sec:approximate_inference,sec:approximate_inference:mcmc}, we have seen various approaches from variational inference and MCMC methods which can be used here.
The first term can be approximated by the predictive distribution of an approximated posterior which is obtained, for example, using variational inference.
The nested entropy of the second term is typically easy to compute, as it corresponds to the entropy of the (discrete) likelihood distribution $p(y \mid \vx, \vtheta)$ of the model $\vtheta$.
The outer expectation of the second term may be approximated using (approximate) samples from the posterior distribution obtained via variational inference, MCMC, or some other method.\looseness=-1

\begin{oreadings}
  \begin{itemize}
    \item \pcite{gal2017deep}
  \end{itemize}
\end{oreadings}

\section{Learning Locally: Transductive Active Learning}

So far we have explored how to select observations that provide us with the best predictor $f(\vx)$ across the entire domain $\vx \in \spX$.
However, we typically utilize predictors to make predictions at a particular location $\vxs$.
It is therefore a natural question to ask how to select observations that lead to the best individual prediction $f(\vxs)$ at the prediction target $\vxs$.
The distinction between these two settings is closely related to the distinction between two general approaches to learning: \midx{inductive learning} and \midx{transductive learning}.
Inductive learning aims to extract general rules from the data, that is, to extract and compress the most bits of information.
In contrast, transductive learning aims to make the best prediction at a particular location, that is, to extract the most bits of information \emph{relevant to the prediction} at $\vxs$.
The concept of transduction was developed by Vladimir Vapnik in the 1980s who described its essence as follows: \begin{quote}
  ``When solving a problem of interest, do not solve a more general problem as an intermediate step.
  Try to get the answer that you really need but not a more general one.'' --- Vladimir Vapnik
\end{quote}

\begin{rmk}{What are the prediction targets $\vxs$?}{}
  Typically, in transductive learning, we cannot directly observe the value $f(\vxs)$ at the prediction target $\vxs$, for example, when learning from a fixed dataset~$\spX' \subset \spX$.
  If we \emph{could} observe $f(\vxs)$ directly (or perturbed by noise), solving the learning task would only require memorization.
  Instead, most interesting learning tasks require generalizing $f(\vxs)$ from the behavior of $f$ at other locations.
  Therefore, transductive learning becomes interesting precisely when we cannot directly observe $f(\vxs)$.\safefootnote{We will discuss an example of this kind in \cref{exercise:transductive_active_learning}.}

  Note that this is similar to the inductive setting where we, in principle, could observe $f(\vx)$ at any location $\vx$, but practically, we can only observe $f(\vx)$ at a finite number of locations.
  Since such an inductive model is then used to make predictions at any location~$\vxs$, it also needs to generalize from the observations.
\end{rmk}

Following the transductive paradigm, when already knowing that our goal is to predict $f(\vxs)$ our objective is to select observations that provide the most information about $f(\vxs)$.
We can express this objective elegantly using the probabilistic framework from this chapter \citep{mackay1992information,hubotter2024transductive}: \begin{align}
  \vx_{t+1} \defeq \argmax_{\vx \in \spX'} \Ism{f_{\vxs}}{y_{\vx}}[\vx_{1:t}, y_{1:t}] = \argmin_{\vx \in \spX'} \Hsm{f_{\vxs}}[\vx_{1:t}, y_{1:t}, y_{\vx}]. \label{eq:transductive_mi}
\end{align}
\cite{hubotter2024transductive} show that this objective leads to a remarkably different selection of observations compared to the inductive active learning objective we discussed earlier.
Indeed, while the inductive objective focuses on selecting a \emph{diverse} set of examples, the transductive objective also takes into account the \emph{relevance} of the examples to the prediction target $\vxs$.
We can see this tradeoff between diversity and relevance by rewriting the transductive objective as \begin{align}
  \Ism{f_{\vxs}}{y_{\vx}}[\vx_{1:t}, y_{1:t}] = \underbrace{\Ism{f_{\vxs}}{y_{\vx}}}_{\text{relevance}} - \underbrace{\Ism{f_{\vxs}}{y_{\vx} ; y_{1:t}}[\vx_{1:t}]}_{\text{redundancy}} \margintag{using the definition of interaction information~\eqref{eq:interaction_information}}
\end{align} where the first term measures the information gain of $y_{\vx}$ about $f_{\vxs}$ while the second term is the interaction information which measures the redundancy of the information in $y_{\vx}$ and $y_{1:t}$ about~$f_{\vxs}$.
In this way, transductive active learning describes a middle ground between traditional search and retrieval methods that focus on relevance on the one hand and inductive active learning which focuses on diversity on the other hand.

\begin{oreadings}
  \begin{itemize}
    \item \pcite{hubotter2024transductive}
  \end{itemize}
  In modern machine learning, one often differentiates between a ``pre-training'' and a ``fine-tuning'' stage.
  During pre-training, a model is trained on a large dataset to extract general knowledge without a specific task in mind.
  Then, during fine-tuning, the model is adapted to a specific task by training on a smaller dataset.
  Whereas (inductive) active learning is closely linked to the pre-training stage, transductive active learning has been shown to be useful for task-specific fine-tuning:
  \begin{itemize}
    \item \pcite{hubotter2025efficiently}
    \item \pcite{bagatella2024active}
  \end{itemize}
\end{oreadings}

\section*{Discussion}

We have discussed how to select the most informative data.
Thereby, we focused mostly on \emph{inductive} active learning which is applicable to ``pre-training'' when we aim to extract general knowledge from data, but also explored \emph{transductive} active learning which is useful for ``fine-tuning'' when we aim to adapt a model to a specific task.

We focused on quantifying the ``informativeness'' of data by its information gain, which is a common approach, though many other viable criteria exist:

\begin{rmk}{Beyond mutual information}{beyond_mi}
  The problem of identifying which experiments to conduct in order to maximize learning is studied extensively in the field of \midx{experimental design} where a set of observations $\sS$ is commonly called a \midx{design}.
  In the presence of a prior and likelihood, and a different possible posterior distribution for each design $\sS$, the field is also called \midx{Bayesian experimental design} \citep{chaloner1995bayesian,rainforth2024modern,mutny2024modern}.

  As we highlighted in the beginning of this chapter, how we measure the utility (i.e., the informativeness) of a design $\sS$ is crucial.
  Such a measure is called a \midx{design criterion} and a design which is optimal with respect to a design criterion is called an \midx{optimal design}.
  The literature studies various design criteria beyond maximizing mutual information (i.e., minimizing posterior entropy).
  One popular alternative is to select the observations $\sS$ that minimize the trace of the posterior covariance matrix, which corresponds to minimizing the posterior average variance.

\end{rmk}

Next, we will move to the topic of optimization and ask which data we should select to find the optimum of an unknown function as quickly as possible.
In the following chapter, we will focus on ``Bayesian optimization'' (also called ``bandit optimization'') where our aim is to \emph{find and sample} the optimal point.
A related task that is slightly more related to active learning is the ``best-arm identification'' problem where we aim only to \emph{identify} the optimal point without sampling it.
This problem is closely related transductive active learning (with the local task being defined by the location of the maximum) and so-called \midx{entropy search} methods that minimize the entropy of the posterior distribution over the location or value of the maximum (akin to \cref{eq:transductive_mi}) are often used to solve this problem~\citep{hennig2012entropy,wang2017max,hvarfner2022joint}.\looseness=-1

\excheading

\begin{nexercise}{Mutual information and KL divergence}{mi_and_kl}
  Show that by expanding the definition of mutual information, \begin{align}
    \I{\rX}{\rY} &= \E[(\vx,\vy) \sim p]{\log\frac{p(\vx,\vy)}{p(\vx) p(\vy)}} \nonumber \\
    &= \KL{p(\vx,\vy)}{p(\vx) p(\vy)} \\
    &= \E[\vy \sim p]{\KL{p(\vx \mid \vy)}{p(\vx)}},
  \end{align} where $p(\vx,\vy)$ denotes the joint distribution and $p(\vx),p(\vy)$ denote the marginal distributions of $\rX$ and $\rY$.
\end{nexercise}

\begin{nexercise}{Non-monotonicity of cond. mutual information}{cond_mi_non_monotonicity}
  \begin{enumerate}
    \item Show that if $\rX \perp \rZ$ then $\I{\rX}{\rY \mid \rZ} \geq \I{\rX}{\rY}$.

    \item Show that if $\rX \perp \rZ \mid \rY$ then $\I{\rX}{\rY \mid \rZ} \leq \I{\rX}{\rY}$.

    \item Note that the condition $\rX \perp \rZ \mid \rY$ is the Markov property, namely, $\{\rX, \rY, \rZ\}$ form a Markov chain with graphical model $\rX \to \rY \to \rZ$.
    This situation often occurs when data is processed sequentially.
    Prove \begin{align}
      \I{\rX}{\rZ} \leq \I{\rX}{\rY}. \label{eq:data_processing_inequality}
    \end{align} which is also known as the \midx{data processing inequality}, and which says that processing cannot increase the information contained in a signal.
  \end{enumerate}
\end{nexercise}

\begin{nexercise}{Interaction information}{interaction_information}
  \begin{enumerate}
    \item Show that interaction information is symmetric.
    \item Let $X_1, X_2 \sim \Bern{p}$ for some $p \in (0,1)$ and independent.
    We denote by $Y \defeq X_1 \oplus X_2$ their XOR.
    Compute $\I{Y ; X_1}{X_2}$.
  \end{enumerate}
\end{nexercise}

\begin{nexercise}{Marginal gain of maximizing mutual information}{marginal_gain_mi}
  Show that in the context of maximizing mutual information, the marginal gain is \begin{align*}
    \Delta_I(\vx \mid \sA) &= \Ism{f_\vx}{y_\vx}[\vy_\sA] = \H{y_{\vx} \mid \vy_\sA} - \H{\varepsilon_\vx}.
  \end{align*}
\end{nexercise}

\begin{nexercise}{Submodularity means no synergy}{submodularity_and_no_synergy}
  Show that submodularity is equivalent to the absence of synergy between observations.
  That is, show that for all $\sA \subseteq \sB$, \begin{align}
    \Ism{f_\vx}{y_\vx ; \vy_{\sB \setminus \sA}}[\vy_\sA] \geq 0.
  \end{align}
\end{nexercise}

\begin{nexercise}{Transductive active learning}{transductive_active_learning}

  Consider the prior distribution \(X_i \sim \mathcal{N}(0,1)\) with all \(X_i\) independent and consider the ``output'' variable \begin{align*}
    Z \defeq \sum_{i=1}^{100} i \cdot X_i.
  \end{align*}
  Our observation when \(X_{i_t}\) is selected in round \(t\) is generated by \begin{align*}
    Y_t \defeq X_{i_t} + \varepsilon_t \quad\text{with } \varepsilon_t \iid \mathcal{N}(0,1).
  \end{align*}
  For a set of inputs \(S = \{i_1, \dots, i_t\} \subseteq \{1, \dots, 100\}\), we define the \emph{information gain} of \(S\) as \[
      F(S) \defeq \I{Z}{Y_{1:t}} = \H{Z} - \H{Z}[Y_{1:t}]
  \] where \(\H{Z}\) is the entropy of \(Z\) according to the prior and \(\H{Z}[Y_{1:t}]\) is the conditional entropy after the observations \(Y_1, \dots, Y_t\).\par
  \emph{Note: The random vector \((X_1, \dots, X_{100}, Z)\) is jointly Gaussian due to the closedness of Gaussians under linear transformations.}

  \begin{enumerate}
    \item We define the \emph{marginal information gain} \(\Delta(j \mid S)\) for a new observation \(Y_j\) as \[
      \Delta(j \mid S) \defeq F(S \cup \{j\}) - F(S).
    \]
    Does \({\Delta(j \mid S) \geq 0}\) hold for all $j$ and $S$?

    \item Is maximizing the marginal information gain \(\Delta(i \mid S)\) equivalent to picking the point \(i \in \{1, \dots, 100\}\) with maximum variance under the current posterior over \(X_i\), that is, ``equivalent to uncertainty sampling''?

    \item Let us consider the alternative prior where \(X_i \sim \Bern{0.5}\) are fair coin flips which we observe directly (i.e., \(Y_t = X_{i_t}\)).
    For which of the following definitions of \(Z\) is the acquisition function \(S \mapsto F(S)\) submodular? \begin{enumerate}
        \item \(Z \defeq X_1 \land \cdots \land X_{100}\) with \(\land\) denoting the logical AND.
        \item \(Z \defeq X_1 \lor \cdots \lor X_{100}\) with \(\lor\) denoting the logical OR.
        \item \(Z \defeq X_1 \oplus \cdots \oplus X_{100}\) with \(\oplus\) denoting the logical XOR, i.e., the exclusive OR which returns \(1\) iff exactly one of the \(X_i\) is \(1\) and \(0\) otherwise.
    \end{enumerate}
  \end{enumerate}
\end{nexercise}

  \chapter{Bayesian Optimization}\label{sec:bayesian_optimization}

Often, obtaining data is costly.
In the previous chapter, this led us to investigate how we can optimally improve our understanding (i.e., reduce uncertainty) of the process we are trying to model.
However, purely improving our understanding is often not good enough.
In many cases, we want to use our improving understanding \emph{simultaneously} to reach certain goals.
This is a very common problem in artificial intelligence and will concern us for the rest of this manuscript.
One common instance of this problem is the setting of optimization.

\begin{marginfigure}
  \incfig{bayesian_optimization}
  \caption{Illustration of Bayesian optimization. We pass an input $\vx_t$ into the unknown function $\opt{f}$ to obtain noisy observations $y_t$.}
\end{marginfigure}

Given some function $\opt{f} : \spX \to \R$, suppose we want to find the \begin{align}
  \argmax_{\vx \in \spX} \opt{f}(\vx).
\end{align}
Now, contrary to classical optimization, we are interested in the setting where the function $\opt{f}$ is unknown to us (like a ``black-box'').
We are only able to obtain noisy observations of $\opt{f}$, \begin{align}
  y_t = \opt{f}(\vx_t) + \varepsilon_t.
\end{align}
Moreover, these noise-perturbed evaluations are costly to obtain.
We will assume that similar alternatives yield similar results,\footnote{That is, $\opt{f}$ is ``smooth''. We will be more precise in the subsequent parts of this chapter. If this were not the case, optimizing the function without evaluating it everywhere would not be possible. Fortunately, many interesting functions obey by this relatively weak assumption.} which is commonly encoded by placing a Gaussian process prior on $\opt{f}$.
This assumed correlation is fundamentally what will allow us to learn a model of $\opt{f}$ from relatively few samples.\footnote{There are countless examples of this problem in the ``real world''. Instances are \begin{itemize}
  \item drug trials
  \item chemical engineering --- the development of physical products
  \item recommender systems
  \item automatic machine learning --- automatic tuning of model \& hyperparameters
  \item and many more...
\end{itemize}}

\section{Exploration-Exploitation Dilemma}\label{sec:bayesian_optimization:exploration_exploitation}

In Bayesian optimization, we want to learn a model of $\opt{f}$ and use this model to optimize $\opt{f}$ simultaneously.
These goals are somewhat contrary.
Learning a model of $\opt{f}$ requires us to explore the input space while using the model to optimize $\opt{f}$ requires us to focus on the most promising well-explored areas.
This trade-off is commonly known as the \midx{exploration-exploitation dilemma}[idxpagebf], whereby \begin{itemize}
  \item \emph{exploration} refers to choosing points that are ``informative'' with respect to the unknown function. For example, points that are far away from previously observed points (i.e., have high posterior variance);\footnote{We explored this topic (with strategies like uncertainty sampling) in the previous chapter.} and
  \item \emph{exploitation} refers to choosing promising points where we expect the function to have high values. For example, points that have a high posterior mean and a low posterior variance.
\end{itemize}
In other words, the exploration-exploitation dilemma refers to the challenge of learning enough to understand $\opt{f}$, but not learning too much to lose track of the objective --- optimizing $\opt{f}$.

The exploration-exploitation dilemma is yet another example of the \midx{principle of curiosity and conformity} which we introduced in \cref{sec:free_energy} and encountered many times since in our study of approximate probabilistic inference.
We will see in subsequent chapters that sequential decision-making is intimately related to probabilistic inference, and there we will also make this correspondence more precise.

\section{Online Learning and Bandits}

Bayesian optimization is closely related to a form of online learning.
In \midx{online learning} we are given a set of possible inputs $\spX$ and an unknown function $\opt{f} : \spX \to \R$.
We are now asked to choose a sequence of inputs $\vx_1, \dots \vx_T$ online,\footnote{\emph{Online} is best translated as ``sequential''. That is, we need to pick $\vx_{t+1}$ based only on our prior observations $y_1, \dots, y_t$.} and our goal is to maximize our cumulative reward $\sum_{t=1}^T \opt{f}(\vx_t)$.
Depending on what we observe about $\opt{f}$, there are different variants of online learning.
Bayesian optimization is closest to the so-called (stochastic) ``bandit'' setting.

\subsection{Multi-Armed Bandits}\label{sec:bayesian_optimization:online_learning:mab}

The ``\midx{multi-armed bandits}'' (MAB) problem is a classical, canonical formalization of the exploration-exploitation dilemma.
In the MAB  problem, we are provided with $k$ possible actions (arms) and want to maximize our reward online within the time horizon $T$.
We do not know the reward distributions of the actions in advance, however, so we need to trade learning the reward distribution with following the most promising action.
Bayesian optimization can be interpreted as a variant of the MAB problem where there can be a potentially infinite number of actions (arms), but their rewards are correlated (because of the smoothness of the Gaussian process prior).

\begin{marginfigure}
  \incfig{multi_armed_bandits}
  \caption{Illustration of a multi-armed bandit with four arms, each with a different reward distribution.
  The agent tries to identify the arm with the most beneficial reward distribution shown in green.}
\end{marginfigure}

There exists a large body of work on this and similar problems in online decision-making.
Much of this work develops theory on how to explore and exploit in the face of uncertainty.
The shared prevalence of the exploration-exploitation dilemma signals a deep connection between online learning and Bayesian optimization (and --- as we will later come to see --- reinforcement learning).
Many of the approaches which we will encounter in the context of these topics are strongly related to methods in online learning.

One of the key principles of the theory on multi-armed bandits and reinforcement learning is the principle of \midx<``optimism in the face of uncertainty''>{optimism in the face of uncertainty}[idxpagebf], which suggests that it is a good guideline to explore where we can hope for the best outcomes.
We will frequently come back to this general principle in our discussion of algorithms for Bayesian optimization and reinforcement learning.

\subsection{Regret}

The key performance metric in online learning is the regret.

\begin{defn}[Regret]\pidx{regret}
  The \emph{(cumulative) regret} for a time horizon $T$ associated with choices $\{\vx_t\}_{t=1}^T$ is defined as \begin{align}
    R_T &\defeq \sum_{t=1}^T \underbrace{\parentheses*{\max_\vx \opt{f}(\vx) - \opt{f}(\vx_t)}}_{\text{\midx{instantaneous regret}}} \\
    &= T \max_\vx \opt{f}(\vx) - \sum_{t=1}^T \opt{f}(\vx_t). \label{eq:regret2}
  \end{align}
\end{defn}
The regret can be interpreted as the additive loss with respect to the \emph{static} optimum $\max_\vx \opt{f}(\vx)$.

The goal is to find algorithms that achieve \midx<sublinear>{sublinear regret} regret, \begin{align}
  \lim_{T\to\infty} \frac{R_T}{T} = 0.
\end{align}
Importantly, if we use an algorithm which explores \emph{forever}, e.g., by going to a random point $\tilde{\vx}$ with a constant probability $\epsilon$ in each round, then the regret will grow linearly with time.
This is because the instantaneous regret is at least $\epsilon (\max_\vx \opt{f}(\vx) - \opt{f}(\tilde{\vx}))$ and non-decreasing.
Conversely, if we use an algorithm which \emph{never} explores, then we might never find the static optimum, and hence, also incur constant instantaneous regret in each round, implying that regret grows linearly with time.
Thus, achieving sublinear regret requires \emph{balancing} exploration and exploitation.

Typically, online learning (and Bayesian optimization) consider stationary environments, hence the comparison to the static optimum.
Dynamic environments are studied in \midx{online algorithms} (see metrical task systems\footnote[][-16\baselineskip]{Metrical task systems are a classical example in online algorithms. Suppose we are moving in a (finite) decision space~$\spX$. In each round, we are given a ``task'' $f_t : \spX \to \R$ which is more or less costly depending on our state~${\vx_t \in \spX}$. In many contexts, it is natural to assume that it is also costly to move around in the decision space. This cost is modeled by a metric $d(\cdot, \cdot)$ on $\spX$. In \idx{metrical task systems}, we want to minimize our total cost, \begin{align*}
  \sum_{t=1}^T f_t(\vx_t) + d(\vx_t, \vx_{t-1}).
\end{align*} That is, we want to trade completing our tasks optimally with moving around in the state space. Crucially, we do not know the sequence of tasks~$f_t$ in advance. Due to the cost associated with moving in the decision space, previous choices affect the future!}, convex function chasing\footnote{\idx<Convex function chasing>{convex function chasing} (or \idx{convex body chasing}) generalize metrical task systems to continuous domains~$\spX$. To make any guarantees about the performance in these settings, one typically has to assume that the tasks~$f_t$ are convex. Note that this mirrors our assumption in Bayesian optimization that similar alternatives yield similar results.}, and generalizations of multi-armed bandits to changing reward distributions) and reinforcement learning.
When operating in dynamic environments, other metrics such as the competitive ratio,\footnote{To assess the performance in dynamic environments, we typically compare to a dynamic optimum. As these problems are difficult (we are usually not able to guarantee convergence to the dynamic optimum), one considers a multiplicative performance metric similar to the approximation ratio, the \idx{competitive ratio}, \begin{align*}
  \mathrm{cost}(\mathrm{ALG}) \leq \alpha \cdot \mathrm{cost}(\mathrm{OPT}),
\end{align*} where $\mathrm{OPT}$ corresponds to the dynamic optimal choice (in hindsight).} which compares against the best \emph{dynamic} choice, are useful.
As we will later come to see in \cref{sec:mbarl:planning} in the context of reinforcement learning, operating in dynamic environments is deeply connected to a rich field of research called \midx{control}.

\section{Acquisition Functions}

It is common to use a so-called \midx{acquisition function}[idxpagebf] to greedily pick the next point to sample based on the current model.

Throughout our description of acquisition functions, we will focus on a setting where we model $\opt{f}$ using a Gaussian process which we denote by $f$.
The methods generalize to other means of learning $\opt{f}$ such as Bayesian deep learning.
The various acquisition functions $F$ are used in the same way as is illustrated in \cref{alg:bo}.

\begin{algorithm}
  \caption{Bayesian optimization (with GPs)}\pidx{Bayesian optimization}\label{alg:bo}
  initialize $f \sim \GP{\mu_0}{k_0}$\;
  \For{$t = 1$ \KwTo $T$}{
    choose $\vx_t = \argmax_{\vx \in \spX} F(\vx; \mu_{t-1}, k_{t-1})$\;
    observe $y_t = f(\vx_t) + \epsilon_t$\;
    perform a probabilistic update to obtain $\mu_t$ and $k_t$\;
  }
\end{algorithm}

\begin{rmk}{Model selection}{}
  Selecting a model of $\opt{f}$ in sequential decision-making is much harder than in the i.i.d. data setting of supervised learning.
  There are mainly the following two dangers: \begin{itemize}
    \item the data sets collected in active learning and Bayesian optimization are \emph{small}; and
    \item the data points are selected \emph{dependently} on prior observations.
  \end{itemize}
  This leads to a specific danger of overfitting.
  In particular, due to feedback loops between the model and the acquisition function, one may end up sampling the same point repeatedly.

  One approach to reduce the chance of overfitting is the use of hyperpriors\pidx{hyperprior} which we mentioned previously in \cref{sec:gp:model_selection:marginal_likelihood}.
  Another approach that often works fairly well is to occasionally (according to some schedule) select points uniformly at random instead of using the acquisition function.
  This tends to prevent getting stuck in suboptimal parts of the state space.
\end{rmk}

One possible acquisition function is uncertainty sampling \eqref{eq:uncertainty_sampling1}, which we discussed in the previous chapter.
However, this acquisition function does not at all take into account the objective of maximizing $\opt{f}$ and focuses solely on exploration.

Suppose that our model $f$ of $\opt{f}$ is well-calibrated, in the sense that the true function lies within its confidence bounds.
Consider the best lower bound, that is, the maximum of the lower confidence bound.
Now, if the true function is really contained in the confidence bounds, it must hold that the optimum is somewhere above this best lower bound.
In particular, we can exclude all regions of the domain where the upper confidence bound (the optimistic estimate of the function value) is lower than the best lower bound.
This is visualized in \cref{fig:optimistic_bo}.

\begin{marginfigure}
  \begin{center}
    \import{./plots/output/}{bayesian_optimization_optimism.pgf}
  \end{center}

  \caption{Optimism in Bayesian optimization.
  The \textbf{unknown function} is shown in black, our \textbf{\b{model}} in blue with gray confidence bounds.
  The dotted black line denotes the maximum lower bound.
  We can therefore focus our exploration to the yellow regions where the upper confidence bound is higher than the maximum lower bound.}\label{fig:optimistic_bo}
\end{marginfigure}

Therefore, we only really care how the function looks like in the regions where the upper confidence bound is larger than the best lower bound.
The key idea behind the methods that we will explore is to focus exploration on these plausible maximizers.

Note that it is crucial that our uncertainty about $f$ reflects the ``fit'' of our model to the unknown function.
If the model is not well calibrated or does not describe the underlying function at all, these methods will perform poorly.
This is where we can use the Bayesian philosophy by imposing a prior belief that may be conservative.

\subsection{Upper Confidence Bound}\label{sec:bayesian_optimization:acquisition_functions:ucb}

The principle of \midx{optimism in the face of uncertainty} suggests picking the point where we can hope for the optimal outcome.
In this setting, this corresponds to simply maximizing the \midx{upper confidence bound} (UCB), \begin{align}
  \vx_{t+1} \defeq \argmax_{\vx \in \spX} \mu_{t}(\vx) + \beta_{t+1} \sigma_{t}(\vx),
\end{align} where $\sigma_t(\vx) \defeq \sqrt{k_t(\vx, \vx)}$ is the standard deviation at $\vx$ and $\beta_t$ regulates how confident we are about our model $f$ (i.e., the choice of confidence interval).

\begin{marginfigure}[-5\baselineskip]
  \begin{center}
    \import{./plots/output/}{ucb.pgf}
  \end{center}

  \caption{Plot of the UCB acquisition function for $\beta = 0.25$ and $\beta = 1$, respectively.}\label{fig:ucb}
\end{marginfigure}

This acquisition function naturally trades exploitation by preferring a large posterior mean with exploration by preferring a large posterior variance.
Note that if $\beta_t = 0$ then UCB is purely exploitative, whereas, if $\beta_t \to \infty$, UCB recovers uncertainty sampling (i.e., is purely explorative).\footnote{Due to the monotonicity of $(\cdot)^2$, it does not matter whether we optimize the variance or standard deviation at $\vx$.}
UCB is an example of an optimism-based method, as it greedily picks the point where we can hope for the best outcome.

As can be seen in \cref{fig:ucb}, the UCB acquisition function is generally non-convex.
For selecting the next point, we can use approximate global optimization techniques like Lipschitz optimization (in low dimensions) and gradient ascent with random initialization (in high dimensions).
Another widely used technique is to sample some random points from the domain, score them according to this criterion, and simply take the best one.

The choice of $\beta_t$ is crucial for the performance of UCB.
Intuitively, for UCB to work even if the unknown function $\opt{f}$ is not contained in the confidence bounds, we use $\beta_t$ to re-scale the confidence bounds to enclose $\opt{f}$ as shown in \cref{fig:ucb_rescaling_cb}.
A theoretical analysis requires that $\beta_t$ is chosen ``correctly''.
Formally, we say that the sequence $\beta_t$ is chosen correctly if it leads to \midx<well-calibrated confidence intervals>{well-calibrated confidence interval}, that is, if with probability at least $1-\delta$, \begin{align}
  \forall t \geq 1,\ \forall \vx \in \spX :\quad \opt{f}(\vx) \in \spC_t(\vx) \defeq \brackets*{\mu_{t-1}(\vx) \pm \beta_t(\delta) \cdot \sigma_{t-1}(\vx)}. \label{eq:gp_ucb_beta_calibration}
\end{align}
\begin{marginfigure}
  \begin{center}
    \import{./plots/output/}{rescaling_confidence_bounds.pgf}
  \end{center}

  \caption{Re-scaling the confidence bounds. The dotted gray lines represent updated confidence bounds.}\label{fig:ucb_rescaling_cb}
\end{marginfigure}
Bounds on $\beta_t(\delta)$ can be derived both in a ``Bayesian'' and in a ``frequentist'' setting.
In the Bayesian setting, it is assumed that $\opt{f}$ is drawn from the prior GP, i.e., $\opt{f} \sim \GP{\mu_0}{k_0}$.
However, in many cases this may be an unrealistic assumption.
In the frequentist setting, it is assumed instead that $\opt{f}$ is a fixed element of a reproducing kernel Hilbert space $\spH_k(\spX)$ which depending on the kernel $k$ can encompass a large class of functions.
We will discuss the Bayesian setting first and later return to the frequentist setting.

\begin{thm}[Bayesian confidence intervals, lemma 5.5 of \cite{srinivas2009gaussian}]\label{thm:bayesian_confidence_intervals} \exerciserefmark{bayesian_confidence_intervals}
  Let $\delta \in (0,1)$.
  Assuming $\opt{f} \sim \GP{\mu_0}{k_0}$ and Gaussian observation noise $\epsilon_t \sim \N{0}{\sigman^2}$, the sequence \begin{align}
    \beta_t(\delta) = \BigO{\sqrt{\log\parentheses*{\nicefrac{\card{\spX} t}{\delta}}}}
  \end{align} satisfies $\Pr{\forall t \geq 1, \vx \in \spX : \opt{f}(\vx) \in \spC_t(\vx)} \geq 1 - \delta$.
\end{thm}

Under the assumption of well-calibrated confidence intervals, we can bound the regret of UCB.

\begin{thm}[Regret of GP-UCB, theorem 2 of \cite{srinivas2009gaussian}]\label{thm:bayesian_regret_for_gp_ucb} \exerciserefmark{bayesian_regret_for_gp_ucb}
  If $\beta_t(\delta)$ is chosen ``correctly'' for a fixed $\delta \in (0,1)$, with probability at least $1-\delta$, greedily choosing the upper confidence bound yields cumulative regret \begin{align}
    R_T = \BigO{\beta_T(\delta) \sqrt{\gamma_T T}} \label{eq:ucb_regret}
  \end{align} where \begin{align}
    \gamma_T \defeq \max_{\substack{\sS \subseteq \spX \\ \card{\sS} = T}} \I{\vf_\sS}{\vy_\sS} = \max_{\substack{\sS \subseteq \spX \\ |\sS| = T}} \frac{1}{2} \log\det{\mI + \sigman^{-2} \mK_{\sS\sS}} \label{eq:gamma_t}
  \end{align} is the maximum information gain after $T$ rounds.
\end{thm}

Observe that if the information gain is sublinear in $T$ then we achieve sublinear regret and, in particular, converge to the true optimum.
The information gain $\gamma_T$ measures how much can be learned about $\opt{f}$ within $T$ rounds.
If the function is assumed to be smooth (perhaps even linear), then the information gain is smaller than if the function was assumed to be ``rough''.
Intuitively, the smoother the functions encoded by the prior, the smaller is the class of functions to choose from and the more can be learned from a single observation about ``neighboring'' points.

\begin{thm}[Information gain of common kernels, theorem~5 of \cite{srinivas2009gaussian} and remark~2 of \cite{vakili2021information}]\label{thm:kernel_info_gain}
  Due to submodularity, we have the following bounds on the information gain of common kernels:\looseness=-1 \begin{itemize}
    \item \emph{linear kernel} \begin{align}
      \gamma_T = \BigO{d \log T},
    \end{align}
    \item \emph{Gaussian kernel} \begin{align}
      \gamma_T = \BigO{(\log T)^{d+1}},
    \end{align}
    \item \emph{Matérn kernel} for $\nu > \frac{1}{2}$ \begin{align}
      \gamma_T = \BigO{T^{\frac{d}{2\nu + d}} (\log T)^{\frac{2\nu}{2\nu + d}}}. \label{eq:gamma_t_matern}
    \end{align}
  \end{itemize}
\end{thm}

\begin{marginfigure}
  \begin{center}
    \import{./plots/output/}{kernel_info_gain.pgf}
  \end{center}

  \caption{Information gain of \textbf{independent}, \textbf{\g{linear}}, \textbf{\b{Gaussian}}, and \r{\textbf{Matérn} ($\nu \approx 0.5$)} kernels with $d = 2$ (up to constant factors).
  The kernels with sublinear information gain have strong diminishing returns (due to their strong dependence between ``close'' points).
  In contrast, the independent kernel has no dependence between points in the domain, and therefore no diminishing returns.
  Intuitively, the ``smoother'' the class of functions modeled by the kernel, the stronger are the diminishing returns.}\label{fig:kernel_info_gain}
\end{marginfigure}

The information gain of common kernels is illustrated in \cref{fig:kernel_info_gain}.
Notably, when all points in the domain are independent, the information gain is linear in $T$.
This is because when the function $\opt{f}$ may be arbitrarily ``rough'', we cannot generalize from a single observation to ``neighboring'' points, and as there are infinitely many points in the domain $\spX$ there are no diminishing returns.
As one would expect, in this case, \cref{thm:bayesian_regret_for_gp_ucb} does not yield sublinear regret.
However, we can see from \cref{thm:kernel_info_gain} that the information gain is sublinear for linear, Gaussian, and most Matérn kernels.
Moreover, observe that unless the function is linear, the information gain grows exponentially with the dimension $d$.
This is because the number of ``neighboring'' points (with respect to Euclidean geometry) decreases exponentially with the dimension which is also known as the \midx{curse of dimensionality}.

As mentioned, the size of the confidence intervals can also be analyzed under a frequentist assumption on $\opt{f}$.

\begin{thm}[Frequentist confidence intervals, theorem 2 of \cite{chowdhury2017kernelized}]\label{thm:frequentist_confidence_intervals}
  Let $\delta \in (0,1)$.
  Assuming $\opt{f} \in \spH_k(\spX)$, we have that with probability at least $1 - \delta$, the sequence \begin{align}
    \beta_t(\delta) = \norm{\opt{f}}_k + \sigman \sqrt{2 \parentheses*{\gamma_t + \log\parentheses*{\nicefrac{1}{\delta}}}}
  \end{align} satisfies $\Pr{\forall t \geq 1, \vx \in \spX : \opt{f}(\vx) \in \spC_t(\vx)} \geq 1 - \delta$.
\end{thm}

That is, $\beta_t$ depends on the information gain of the kernel as well as on the ``complexity'' of $\opt{f}$ which is measured in terms of its norm in the underlying reproducing kernel Hilbert space $\spH_k(\spX)$.

\begin{rmk}{Bayesian vs frequentist assumption}{}
  \Cref{thm:bayesian_confidence_intervals,thm:frequentist_confidence_intervals} provide different bounds on $\beta_t(\delta)$ based on fundamentally different assumptions on the ground truth $\opt{f}$:
  The Bayesian assumption is that $\opt{f}$ is drawn from the prior GP, whereas the frequentist assumption is that $\opt{f}$ is an element of a reproducing kernel Hilbert space $\spH_k(\spX)$.
  The frequentist assumption holds uniformly for all functions $\opt{f}$ with $\norm{\opt{f}}_k < \infty$, whereas the Bayesian assumption holds only under the Bayesian ``belief'' that $\opt{f}$ is drawn from the prior GP.

  Interestingly, neither assumption encompasses the other.
  This is because if $f \sim \GP{0}{k}$ then it can be shown that almost surely $\norm{f}_k = \infty$, which implies that $f \not\in \spH_k(\spX)$ \citep{srinivas2009gaussian}.
\end{rmk}

We remark that \cref{thm:frequentist_confidence_intervals} holds also under the looser assumption that observations are perturbed by $\sigman$-sub-Gaussian noise (cf. \cref{eq:sub_gaussian}) instead of Gaussian noise.
The bound on $\gamma_T$ from \cref{eq:gamma_t_matern} for the Matérn kernel does not yield sublinear regret when combined with the standard regret bound from \cref{thm:bayesian_regret_for_gp_ucb}, however, \cite{whitehouse2023sublinear} show that the regret of GP-UCB is sublinear also in this case provided $\sigman^2$ is chosen carefully.

This concludes our discussion of the UCB algorithm.
We have seen that its regret can be analyzed under both Bayesian and frequentist assumptions on $\opt{f}$.

\subsection{Improvement}

Another well-known family of methods is based on keeping track of a running optimum $\hat{f}_t$, and scoring points according to their improvement upon the running optimum.
The \midx{improvement} of $\vx$ after round $t$ is measured by \begin{align}
  I_t(\vx) \defeq (f(\vx) - \hat{f}_t)_+ \label{eq:improvement}
\end{align} where we use $(\cdot)_+$ to denote $\max \{0, \cdot\}$.

The \midx{probability of improvement} (PI) picks the point that maximizes the probability to improve upon the running optimum, \begin{align}
  \vx_{t+1} &\defeq \argmax_{\vx \in \spX} \Pr{I_t(\vx) > 0 \mid \vx_{1:t}, y_{1:t}} \\
  &= \argmax_{\vx \in \spX} \fnPr(f(\vx) > \hat{f}_t \mid \vx_{1:t}, y_{1:t}) \label{eq:pi} \\
  &= \argmax_{\vx \in \spX} \Phi\parentheses*{\frac{\mu_{t}(\vx) - \hat{f}_t}{\sigma_{t}(\vx)}} \margintag{using linear transformations of Gaussians \eqref{eq:gaussian_lin_trans}}
\end{align} where $\Phi$ denotes the CDF of the standard normal distribution and we use that $f(\vx) \mid \vx_{1:t}, y_{1:t} \sim \N{\mu_t(\vx)}{\sigma_t^2(\vx)}$.
Probability of improvement tends to be biased in favor of exploitation, as it prefers points with large posterior mean and small posterior variance which is typically true ``close'' to the previously observed maximum $\hat{f}_t$.

\begin{marginfigure}[7\baselineskip]
  \begin{center}
    \import{./plots/output/}{pi_ei.pgf}
  \end{center}

  \caption{Plot of the PI and EI acquisition functions, respectively.}
\end{marginfigure}

Probability of improvement looks at \emph{how likely} a point is to improve upon the running optimum.
An alternative is to look at \emph{how much} a point is expected to improve upon the running optimum.
This acquisition function is called the \midx{expected improvement} (EI), \begin{align}
  \vx_{t+1} \defeq \argmax_{\vx \in \spX} \E{I_t(\vx)}[\vx_{1:t}, y_{1:t}]. \label{eq:ei}
\end{align}
Intuitively, EI seeks a large expected improvement (exploitation) while also preferring states with a large variance (exploration).
Expected improvement yields the same regret bound as UCB \citep{ei_regret}.\looseness=-1

The expected improvement acquisition function is often flat which makes it difficult to optimize in practice due to vanishing gradients.
One approach addressing this is to instead optimize the logarithm of EI \citep{ament2023unexpected}.

\begin{figure*}
  \begin{center}
    \import{./plots/output/}{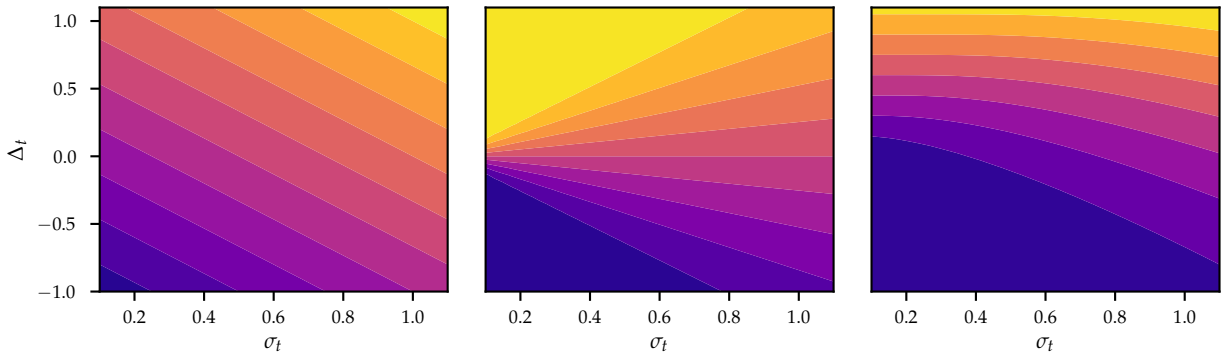}
  \end{center}

  \caption{Contour lines of acquisition functions for varying $\Delta_t = \mu_t(\vx) - \hat{f}_t$ and $\sigma_t$.
  A brighter color corresponds to a larger acquisition function value.
  The first graph shows contour lines of UCB with $\beta_t = 0.75$, the second of PI, and the third of EI.}
\end{figure*}

\subsection{Thompson Sampling}\label{sec:bayesian_optimization:acquisition_functions:thompson_sampling}

We can also interpret the principle of \midx{optimism in the face of uncertainty} in a slightly different way than we did with UCB (and EI).
Suppose we select the next point according to the probability that it is optimal (assuming that the posterior distribution is an accurate representation of the uncertainty), \begin{align}
  \pi(\vx \mid \vx_{1:t}, y_{1:t}) &\defeq \Pr[f \mid \vx_{1:t}, y_{1:t}]{f(\vx) = \max_\vxp f(\vxp)} \\
  \vx_{t+1} &\sim \pi(\cdot \mid \vx_{1:t}, y_{1:t}).
\end{align}
This approach of sampling according to the \midx{probability of maximality}~$\pi$ is called \midx{probability matching}. Probability matching is exploratory as it prefers points with larger variance (as they automatically have a larger chance of being optimal), but at the same time exploitative as it effectively discards points with low posterior mean and low posterior variance.
Unfortunately, it is generally difficult to compute $\pi$ analytically given a posterior.

Instead, it is common to use a sampling-based approximation of $\pi$. Observe that the density $\pi$ can be expressed as an expectation, \begin{align}
  \pi(\vx \mid \vx_{1:t}, y_{1:t}) &= \E*[f \mid \vx_{1:t}, y_{1:t}]{\big[\Ind{f(\vx) = \max_\vxp f(\vxp)}\big]},
\intertext{which we can approximate using Monte Carlo sampling (typically using a single sample),}
  &\approx \Ind{\Tilde{f}_{t+1}(\vx) = \max_\vxp \Tilde{f}_{t+1}(\vxp)}
\end{align} where $\Tilde{f}_{t+1} \sim p(\cdot \mid \vx_{1:t}, y_{1:t})$ is a sample from our posterior distribution.
Observe that this approximation of $\pi$ coincides with a point density at the maximizer of $\Tilde{f}_{t+1}$.

The resulting algorithm is known as \midx{Thompson sampling}[idxpagebf].
At time $t+1$, we sample a function $\Tilde{f}_{t+1} \sim p(\cdot \mid \vx_{1:t}, y_{1:t})$ from our posterior distribution.
Then, we simply maximize $\Tilde{f}_{t+1}$, \begin{align}
  \vx_{t+1} \defeq \argmax_{\vx \in \spX} \Tilde{f}_{t+1}(\vx).
\end{align}
In many cases, the randomness in the realizations of $\Tilde{f}_{t+1}$ is already sufficient to effectively trade exploration and exploitation.
Similar regret bounds to those of UCB can also be established for Thompson sampling \citep{russo2016information,kandasamy2018parallelised}.

\subsection{Information-Directed Sampling}

After having looked at multiple methods that aim to balance exploitation of the current posterior distribution over~$f$ for immediate returns with exploration to reduce uncertainty about~$f$ for future returns, we will next discuss a method that makes this tradeoff explicit.

Denoting the instantaneous regret of choosing $\vx$ as $\Delta(\vx) \defeq \max_{\vxp} \opt{f}(\vxp) - \opt{f}(\vx)$ and by $I_t(\vx)$ some function capturing the ``information gain'' associated with observing $\vx$ in iteration $t+1$, we can define the \midx{information ratio}, \begin{align}
  \Psi_t(\vx) \defeq \frac{\Delta(\vx)^2}{I_t(\vx)}, \label{eq:regret_information_ratio}
\end{align} which was originally introduced by \cite{russo2016information}.
Here exploitation reduces regret while exploration increases information gain, and hence, points $\vx$ that minimize the information ratio are those that most effectively balance exploration and exploitation.
We can make the key observation that the regret $\Delta(\cdot)$ decreases when $I_t(\cdot)$ decreases, as a small $I_t(\cdot)$ implies that the algorithm has already learned a lot about the function $\opt{f}$.
The strength of this relationship is quantified by the information ratio:\looseness=-1

\begin{thm}[Proposition 1 of \cite{russo2014learning} and theorem 8 of \cite{kirschner2018information}]\label{thm:regret_information_ratio}
  For any iteration $T \geq 1$, let $\sum_{t=1}^T I_{t-1}(\vx_t) \leq \gamma_T$ and suppose that $\Psi_{t-1}(\vx_t) \leq \overline{\Psi}_T$ for all $t \in [T]$.
  Then, the cumulative regret is bounded by \begin{align}
    R_T \leq \sqrt{\gamma_T \overline{\Psi}_T T}.
  \end{align}
\end{thm}
\begin{proof}
  By \cref{eq:regret_information_ratio}, $r_t = \Delta(\vx_t) = \sqrt{\Psi_{t-1}(\vx_t) \cdot I_{t-1}(\vx_t)}$.
  Hence, \begin{align*}
    R_T &= \sum_{t=1}^T r_t \\
    &= \sum_{t=1}^T \sqrt{\Psi_{t-1}(\vx_t) \cdot I_{t-1}(\vx_t)} \\
    &\leq \sqrt{\sum_{t=1}^T \Psi_{t-1}(\vx_t) \cdot \sum_{t=1}^T I_{t-1}(\vx_t)} \margintag{using the Cauchy-Schwarz inequality} \\
    &\leq \sqrt{\gamma_T \overline{\Psi}_T T}. \qedhere \margintag{using the assumptions on $I_t(\cdot)$ and $\Psi_t(\cdot)$}
  \end{align*}
\end{proof}

\begin{ex}{How to measure ``information gain''?}{ids_how_to_measure_information_gain}
  One possibility of measuring the ``information gain'' is \begin{align}
    I_t(\vx) \defeq \I{f_\vx}{y_\vx}[\vx_{1:t}, y_{1:t}]
  \end{align} which --- as you may recall --- is precisely the marginal gain of the utility $I(\sS) = \I{\vf_\sS}{\vy_\sS}$ we were studying in \cref{sec:active_learning}.
  In this case, \begin{align*}
    \sum_{t=1}^T I_{t-1}(\vx_t) &= \sum_{t=1}^T \I{f_{\vx_t}}{y_{\vx_t}}[\vx_{1:t-1}, y_{1:t-1}] \\
    &= \sum_{t=1}^T \I{\vf_{\vx_{1:T}}}{y_{\vx_t}}[\vx_{1:t-1}, y_{1:t-1}] \margintag{using $\I{\rX, \rZ}{\rY} \geq \I{\rX}{\rY}$ which follows from \cref{eq:cond_mi_joint_mi,eq:mi_non_neg} and is called \midx<monotonicity of MI>{monotonicity of mutual information}} \\[8pt]
    &= \I{\vf_{\vx_{1:T}}}{\vy_{\vx_{1:T}}} \margintag{by repeated application of \cref{eq:cond_mi_joint_mi}, also called the \midx<chain rule of MI>{chain rule of mutual information}} \\[8pt]
    &\leq \gamma_T. \margintag{by definition of $\gamma_T$ \eqref{eq:gamma_t}}
  \end{align*}
\end{ex}

The regret bound from \cref{thm:regret_information_ratio} suggests an algorithm which in each iteration chooses the point which minimizes the information ratio \eqref{eq:regret_information_ratio}.
However, this is not possible since $\Delta(\cdot)$ is unknown due to its dependence on $\opt{f}$.
\cite{kirschner2018information} propose to use a surrogate to the regret which is based on the current model of $\opt{f}$, \begin{align}
  \hat{\Delta}_t(\vx) \defeq \max_{\vxp \in \spX} u_{t}(\vxp) - l_{t}(\vx).
\end{align}
Here, $u_t(\vx) \defeq \mu_t(\vx) + \beta_{t+1} \sigma_t(\vx)$ and $l_t(\vx) \defeq \mu_t(\vx) - \beta_{t+1} \sigma_t(\vx)$ are the upper and lower confidence bounds of the confidence interval $\spC_t(\vx)$ of~$\opt{f}(\vx)$, respectively.
Similarly to our discussion of UCB, we make the assumption that the sequence $\beta_t$ is chosen ``correctly'' (cf. \cref{eq:gp_ucb_beta_calibration}) so that the confidence interval is well-calibrated and $\Delta(\vx) \leq \hat{\Delta}_t(\vx)$ with high probability.
The resulting algorithm \begin{align}
  \vx_{t+1} \defeq \argmin_{\vx \in \spX} \braces*{\widehat{\Psi}_t(\vx) \defeq \frac{\hat{\Delta}_t(\vx)^2}{I_t(\vx)}} \label{eq:ids}
\end{align} is known as \midx{information-directed sampling} (IDS).

\begin{thm}[Regret of IDS, lemma 8 of \cite{kirschner2018information}]\label{thm:regret_of_ids} \exerciserefmark{regret_of_ids}
  Let $\beta_t(\delta)$ be chosen ``correctly'' for a fixed $\delta \in (0,1)$.
  Then, if the measure of information gain is $I_t(\vx) = \I{f_\vx}{y_{\vx}}[\vx_{1:t}, y_{1:t}]$, with probability at least $1-\delta$, IDS has cumulative regret \begin{align}
    R_T = \BigO{\beta_T(\delta) \sqrt{\gamma_T T}}.
  \end{align}
\end{thm}

Regret bounds such as \cref{thm:regret_of_ids} can be derived also for different measures of information gain.
For example, the argument of \cref{exercise:regret_of_ids} also goes through for the ``greedy'' measure \begin{align}
  I_t(\vx) \defeq \mathrm{I}(f_{\vx_t^{\mathrm{UCB}}} ; y_\vx \mid \vx_{1:t}, y_{1:t}) \label{eq:ucb_info_gain}
\end{align} which focuses exclusively on reducing the uncertainty at $\vx_t^{\mathrm{UCB}}$ rather than globally.
We compare the two measures of information gain in \cref{fig:ids}.
Observe that the acquisition function depends critically on the choice of $I_t(\cdot)$ and is less sensitive to the scaling of confidence intervals.

\begin{marginfigure}
  \begin{center}
    \import{./plots/output/}{ids.pgf}
  \end{center}

  \caption{Plot of the surrogate information ratio $\widehat{\Psi}$: IDS selects its minimizer. The first two plots use the ``global'' information gain measure from \cref{ex:ids_how_to_measure_information_gain} with $\beta = 0.25$ and $\beta = 0.5$, respectively. The third plot uses the ``greedy'' information gain measure from \cref{eq:ucb_info_gain} and $\beta = 1$.}\label{fig:ids}
\end{marginfigure}

IDS trades exploitation and exploration by balancing the (exploitative) regret surrogate with a measure of information gain (such as those studied in \cref{sec:active_learning}) that is purely explorative.
In this way, IDS can account for kinds of information which are not addressed by alternative algorithms such as UCB or EI \citep{russo2014learning}:
Depending on the measure of information gain, IDS can select points to obtain \emph{indirect information} about other points or \emph{cumulating information} that does not immediately lead to a higher reward but only when combined with subsequent observations.
Moreover, IDS avoids selecting points which yield \emph{irrelevant information}.\looseness=-1

\subsection{Probabilistic Inference}\pidx{probability of maximality}[idxpagebf]

As we mentioned before, computing the probability of maximality~$\pi$ is generally intractable.
You can think of computing~$\pi$ as attempting to fully solve the probabilistic inference problem associated with determining the optimum of ${f \mid \vx_{1:t}, y_{1:t}}$.
In many cases, it is useful to determine the probability that a point is optimal under the current posterior distribution, and we will consider one particular example in the following.

\begin{marginfigure}[-10\baselineskip]
  \includegraphics[width=0.9\columnwidth]{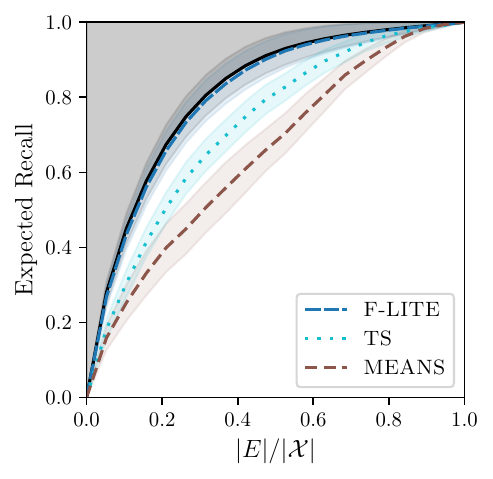}
  \caption{\cite{menet2025lite} show in an experiment that selecting $E \subseteq \spX$ according to the estimates of probability of maximality from LITE (here called F-LITE) is near-optimal. Intuitive heuristics such as Thompson sampling (TS) or selecting points with the highest posterior mean (MEANS) perform worse.}
  \label{fig:lite_recall}
\end{marginfigure}

\begin{ex}{Maximizing recall}{}
  In domains such as molecular design we often use machine learning to screen candidates for further manual testing.
  The goal here is to suggest a small set $E$ from a large domain of molecules~$\spX$, so that the probability of~$E$ containing the optimal molecule, i.e., the \midx{recall}, is maximized.
  Note that this task is quite different from online Bayesian optimization, for example, in BO we get sequential feedback that we can use to decide which inputs to query next.
  Nevertheless, we will see in this section that both tasks turn out to be closely related.

  Let us assume that the maximizer is unique almost surely, which with GPs is automatically the case if there are no same-mean, perfectly-correlated entries.
  The recall task is then intimately related to the task of determining the probability of maximality, since \begin{align}\begin{split}
    \argmax_{E \subseteq \spX : \abs{E} = k} \Pr[f\mid\vx_{1:t},y_{1:t}]{\max_{\vx \in E} f(\vx) = \max_\vxp f(\vxp)} \\
    = \argmax_{E \subseteq \spX : \abs{E} = k} \sum_{\vx \in E} \pi(\vx \mid \vx_{1:t}, y_{1:t}). \margintag{by noting that for $\vx \neq \vy$, the events ${\{f(\vx) = \max_\vxp f(\vxp)\}}$ and ${\{f(\vy) = \max_\vxp f(\vxp)\}}$ are disjoint}
  \end{split}\label{eq:recall_task}\end{align}
  Thus, to obtain the recall-optimal candidates for further testing, we need to find the probability of maximality.
  However, as mentioned, computing the probability of maximality~$\pi$ is generally intractable.

  \Cref{fig:lite_recall} depicts an example of a recall task, where the black line shows the optimal recall achieved by knowing the probability of maximality~$\pi$ exactly.
  LITE~\citep{menet2025lite} is an almost-linear time approximation of $\pi$, whereas TS selects points via Thompson sampling and MEANS selects the points with the highest posterior mean.
  Selecting $E$ according to probability of maximality achieves much higher recall than the other intuitive heuristics.
\end{ex}

So how can we estimate the probability of maximality? LITE approximates~$\pi$ by \begin{align}
  \pi(\vx \mid \vx_{1:t}, y_{1:t}) &= \Pr[f \mid \vx_{1:t}, y_{1:t}]{f(\vx) \geq \max_\vxp f(\vxp)} \nonumber \\
  &\approx \Pr[f \mid \vx_{1:t}, y_{1:t}]{f(\vx) \geq \kappa^*} \label{eq:lite_pi} \\
  &= \Phi\parentheses*{\frac{\mu_t(\vx) - \kappa^*}{\sigma_t(\vx)}} \label{eq:lite}
\end{align} with $\Phi$ denoting the CDF of the standard normal distribution and $\kappa^*$ chosen such that the approximation of $\pi$ integrates to $1$, so that it is a valid distribution.

\begin{marginfigure}
  \begin{center}
    \import{./plots/output/}{bo_prob_inference.pgf}
  \end{center}

  \caption{Plot of the probability of maximality as estimated by LITE.}
  \label{fig:bo_prob_inference}
\end{marginfigure}

Remarkably, LITE is intimately related to many of the BO methods we discussed in this chapter.
First, and most obviously, \cref{eq:lite_pi} looks similar to the PI acquisition function \eqref{eq:pi}.
But note that $\kappa^*$ is not equal to the best observed value $\hat{f}_t$ as in PI, but instead typically larger.
If $\kappa^* > \hat{f}_t$, then \cref{eq:lite_pi} is more exploratory than PI: it puts additional emphasis on points with large posterior variance and comparatively less emphasis on points with large posterior mean, which we illustrate in \cref{fig:bo_prob_inference}.

An insightful interpretation of LITE is that it balances two different kinds of exploration: \midx{optimism in the face of uncertainty} and \midx{entropy regularization}.
To see this, let us define the following variational objective:\looseness=-1 \begin{align}
  \spW(\pi) \defeq \sum_{\vx \in \spX} \pi(\vx) \cdot \Big( \underbrace{\mu_t(\vx)}_{\text{exploitation}} + \underbrace{\sqrt{2 \, \fnS'(\pi(\vx))} \cdot \sigma_t(\vx)}_{\text{exploration}} \Big) \label{eq:lite_variational}
\end{align} with the ``quasi-surprise'' $\fnS'(u) \defeq \frac{1}{2}(\phi(\inv{\Phi}(u))/u)^2$.
The quasi-surprise $\fnS'(\cdot)$ behaves similarly to the surprise $-\ln(\cdot)$.
In fact, their asymptotics coincide: \begin{align*}
  \fnS'(1) = 0 = - \ln(1) \quad\text{and}\quad \fnS'(u) \to - \ln(u) \text{ as } u \to 0^+.
\end{align*}
The objective $\spW$ is maximized for those probability distributions $\pi$ that are concentrated around points with large mean $\mu_t(\vx)$ and points with large exploration bonus.
We have seen that the uncertainty $\sigma_t(\vx)$ about $f(\vx)$ is the standard exploration bonus of UCB-style algorithms.
In \cref{eq:lite_variational}, $\sigma_t(\vx)$ is weighted by the quasi-surprise, which acts as ``entropy regularization'': it increases the entropy of $\pi$ by uniformly pushing $\pi(\vx)$ away from zero.
You can think of entropy regularization as encouraging a different kind of exploration than optimism by not making deterministic decisions like in UCB.

\cite{menet2025lite} show that LITE \eqref{eq:lite} is the solution to the variational problem \eqref{eq:lite_variational} among valid probability distributions \exerciserefmark{lite}: \begin{align}
  \argmax_{\pi \in \Delta^{\spX}} \spW(\pi) = \Phi\parentheses*{\frac{\mu_t(\cdot) - \kappa^*}{\sigma_t(\cdot)}} \label{eq:lite_variational_solution}
\end{align} with $\kappa^*$ such that the right-hand side sums to $1$.\footnote{$\Delta^{\spX}$ is the probability simplex on~$\spX$.}
This indicates that LITE and Thompson sampling, which samples from probability of maximality, achieve exploration through two means:\looseness=-1 \begin{enumerate}
  \item \textbf{Optimism:} by preferring points with large uncertainty $\sigma_t(\vx)$ about the reward value $f(\vx)$.
  \item \textbf{Decision uncertainty:} by assigning some probability mass to all $\vx$, that is, by remaining uncertain about which $\vx$ is the maximizer.
\end{enumerate}
In our discussion of balancing exploration and exploitation in reinforcement learning, we will return to this dichotomy of exploration strategies.

\begin{rmk}{But why is ``exploration'' useful for recall?}{exploration_in_recall}
  Let us pause for a moment and reflect on why this interpretation of LITE is remarkable.
  This interpretation shows that LITE is more ``exploratory'' than simply taking the highest posterior means.
  However, the recall task~\eqref{eq:recall_task} differs from the standard BO task in that we do not collect any further observations, which we may use downstream to make better decisions.
  In the recall task, we are only interested in having the best shot at including the maximizer in the set $E$, without obtaining any further information or making any subsequent decisions.
  At first sight, this seems to suggest that we should be as ``exploitative'' as possible.
  Then how can it be that ``exploration'' is useful?

  We will explore this question with the following example: You are observing a betting game.
  When placing a bet, players can either place a safe bet (``playing safe'') or a risky bet (``playing risky'').
  You now have to place a bet on their bets, and estimate which of the players will win the most money: those that play safe or those that play risky?
  Note that you do not care whether your guess ends up in 2nd place or last place among all players --- you only care about whether your guess wins.
  That is, you are interested in \emph{recalling} the best player, not in ``ranking'' the players.

  Consider three players: one that plays safe and two that play risky.
  Suppose that the safe bet has payoff $S = 1$ while each risky bet has payoff $R \sim \N{0}{100}$.
  In expectation, the safe player will win the most money.
  However, one can see with just a little bit of algebra that the probability of either of the risky players winning the most money is~$\approx\! 35\%$, whereas the safe player only wins with probability~$\approx\! 29\%$ \exerciserefmark{recall}.
  That is, it is in fact \emph{optimal} to bet on either of the risky players since the best player might have vastly outperformed their expected winnings, and performed closer to their upper confidence bound.
  In summary, maximizing recall requires us to be ``exploratory'' since it is likely that the optimum among inputs is one that has performed better than expected, not simply the one with the highest expected performance.
\end{rmk}

\section*{Discussion}

In this chapter, we have explored the exploration-exploitation dilemma in the context of optimizing black-box functions.
To this end, we have explored various methods to balance exploration and exploitation.
While computing the precise probability of maximality is generally intractable, we found that it can be understood approximately as balancing two sources of exploration: optimism in the face of uncertainty and entropy regularization.

In the following chapters, we will begin to discuss stateful settings where the black-box optimization task is known as ``reinforcement learning''.
Naturally, we will see that the exploration-exploitation dilemma is also a central challenge in reinforcement learning, and we will revisit many of the concepts we have discussed in this chapter.

\begin{oreadings}
  \begin{itemize}
    \item \pcite{srinivas2009gaussian}
    \item \pcite{golovin2017google}
    \item \pcite{romero2013navigating}
    \item \pcite{chowdhury2017kernelized}
  \end{itemize}
\end{oreadings}

\excheading

\begin{nexercise}{Convergence to the static optimum}{convergence_to_static_opt}
  Show that any algorithm where $\lim_{t\to\infty} \opt{f}(\vx_t)$ exists achieves sublinear regret if and only if it converges to the static optimum, that is, \begin{align}
    \lim_{t\to\infty} \opt{f}(\vx_t) = \max_\vx \opt{f}(\vx).
  \end{align}
  \textit{Hint: Use that if a sequence $a_n$ converges to $a$ as $n \to \infty$, then we have for the sequence \begin{align}
    b_n \defeq \frac{1}{n} \sum_{i=1}^n a_i \label{eq:cesaro_mean}
  \end{align} that $\lim_{n\to\infty} b_n = a$. This is also known as the \midx{Cesàro mean}.}
\end{nexercise}

\begin{nexercise}{Bayesian confidence intervals}{bayesian_confidence_intervals}
  In this exercise, we derive \cref{thm:bayesian_confidence_intervals}.
  \begin{enumerate}
    \item For fixed $t \geq 1$ and $\vx \in \spX$, prove \begin{align}
      \Pr{\opt{f}(\vx) \not\in \spC_t(\vx) \mid \vx_{1:t-1}, y_{1:t-1}} \leq e^{- \beta_t^2 / 2}.
    \end{align}
    \textit{Hint: Bound $\Pr{Z > c}$ for $Z \sim \N{0}{1}$ and $c > 0$.}

    \item Prove \cref{thm:bayesian_confidence_intervals}.
  \end{enumerate}
\end{nexercise}

\begin{nexercise}{Regret of GP-UCB}{bayesian_regret_for_gp_ucb}
  To develop some intuition, we will derive \cref{thm:bayesian_regret_for_gp_ucb}.

  \begin{enumerate}
    \item Show that if \cref{eq:gp_ucb_beta_calibration} holds, then for a fixed $t \geq 1$ the instantaneous regret $r_t$ is bounded by $2 \beta_t \sigma_{t-1}(\vx_t)$.

    \item Let $S_T \defeq \{\vx_t\}_{t=1}^T$, and define $\vf_T \defeq \vf_{S_T}$ and $\vy_T \defeq \vy_{s_T}$.
    Prove \begin{align}
      \I{\vf_T}{\vy_T} = \frac{1}{2} \sum_{t=1}^T \log\parentheses*{1 + \frac{\sigma_{t-1}^2(\vx_t)}{\sigman^2}}. \label{eq:bayesian_regret_for_gp_ucb_info_gain_helper}
    \end{align}

    \item\label{exercise:bayesian_regret_for_gp_ucb:3} Combine (1) and (2) to show \cref{thm:bayesian_regret_for_gp_ucb}.
    We assume w.l.o.g. that the sequence $\{\beta_t\}_t$ is monotonically increasing.\par
    \textit{Hint: If $s \in [0,M]$ for some $M > 0$ then $s \leq C \cdot \log(1 + s)$ with $C \defeq M / \log(1 + M)$.}
  \end{enumerate}
\end{nexercise}

\begin{nexercise}{Sublinear regret of GP-UCB for a linear kernel}{sublinear_regret_for_linear_kernel}
  Assume that $\opt{f} \sim \GP{0}{k}$ where $k$ is the linear kernel \begin{align*}
    k(\vx, \vxp) = \transpose{\vx}\vxp.
  \end{align*}
  In addition, we assume that for any $\vx \in \spX$, $\norm{\vx}_2 \leq 1$.
  Moreover, recall that the points in a finite set $\sS \subseteq \spX$ can be written in a matrix form (the ``design matrix'') which we denote by $\mX_\sS \in \R^{d \times \card{\sS}}$.

  \begin{enumerate}
    \item Prove that $\gamma_T = \BigO{d \log T}$.

    \item Deduce from (1) and \cref{thm:bayesian_regret_for_gp_ucb} that $\lim_{T\to\infty} R_T / T = 0$.
  \end{enumerate}
\end{nexercise}

\begin{nexercise}{Closed-form expected improvement}{closed_form_ei}
  Let us denote the acquisition function of EI from \cref{eq:ei} by $\mathrm{EI}_t(\vx)$.
  In this exercise, we derive a closed-form expression.

  \begin{enumerate}
    \item Show that \begin{align}
      \mathrm{EI}_t(\vx) = \int_{\nicefrac{(\hat{f}_t - \mu_{t}(\vx))}{\sigma_{t}(\vx)}}^{+\infty} (\mu_{t}(\vx) + \sigma_{t}(\vx) \varepsilon - \hat{f}_t) \cdot \phi(\varepsilon) \,d\varepsilon
    \end{align} where $\phi$ is the PDF of the univariate standard normal distribution \eqref{eq:univ_normal}.\par
    \textit{Hint: Reparameterize the posterior distribution \begin{align*}
      f(\vx) \mid \vx_{1:t}, y_{1:t} \sim \N{\mu_{t}(\vx)}{\sigma_{t}^2(\vx)}
    \end{align*} using a standard normal distribution.}
    \item Using the above expression, show that \begin{align}
      \mathrm{EI}_t(\vx) = (\mu_{t}(\vx) - \hat{f}_t) \Phi\parentheses*{\frac{\mu_{t}(\vx) - \hat{f}_t}{\sigma_{t}(\vx)}} + \sigma_{t}(\vx) \phi\parentheses*{\frac{\mu_{t}(\vx) - \hat{f}_t}{\sigma_{t}(\vx)}} \label{eq:ei_simp}
    \end{align} where $\Phi(u) \defeq \int_{-\infty}^u \phi(\varepsilon) \,d\varepsilon$ denotes the CDF of the standard normal distribution.
  \end{enumerate}
  Note that the first term of \cref{eq:ei_simp} encourages exploitation while the second term encourages exploration.
  EI can be seen as a special case of UCB where the confidence bounds are scaled depending on $\vx$: \begin{align*}
    \beta_t = \phi(z_t(\vx)) / \Phi(z_t(\vx))
  \end{align*} for $z_t(\vx) \defeq (\mu_{t}(\vx) - \hat{f}_t) / \sigma_{t}(\vx)$.
\end{nexercise}

\begin{nexercise}{Regret of IDS}{regret_of_ids}
  We derive \cref{thm:regret_of_ids}.
  \begin{enumerate}
    \item Prove that for all $t \geq 1$, $\hat{\Delta}_t(\vx_t^{\mathrm{UCB}}) \leq 2 \beta_{t+1} \sigma_t(\vx_t^{\mathrm{UCB}})$ where we denote by $\vx_t^{\mathrm{UCB}} \defeq \argmax_{\vx \in \spX} u_t(\vx)$ the point maximizing the upper confidence bound after iteration $t$.
    \item Using (1) and the assumption on $I_t$, bound $\widehat{\Psi}_t(\vx_t^{\mathrm{UCB}})$.\par
    \textit{Hint: You may find the hint of \cref{exercise:bayesian_regret_for_gp_ucb} (3) useful.}
    \item Complete the proof of \cref{thm:regret_of_ids}.
  \end{enumerate}
\end{nexercise}

\begin{nexercise}{Variational form of LITE}{lite}
  Let $\abs{\spX} < \infty$ be such that we can write $\spW$ as an objective function over the probability simplex $\Delta^{\abs{\spX}} \subset \R^{\abs{\spX}}$.
  Derive \cref{eq:lite_variational_solution}.\par
  \textit{Hint: First show that $\spW(\cdot)$ is concave (i.e., minimizing $-\spW(\cdot)$ is a convex optimization problem) and then use Lagrange multipliers to find the optimum.}
\end{nexercise}

\begin{nexercise}{Finding the winning player}{recall}
  Consider the betting game from \cref{rmk:exploration_in_recall} with two risky players, $R_1 \sim \N{0}{100}$ and $R_2 \sim \N{0}{100}$, and one safe player $S = 1$.
  Prove that the individual probability of any of the risky players winning the most money is larger than the winning probability of the safe player.
  We assume that players payoffs are mutually independent.

  \textit{Hint: Compute the CDF of $R \defeq \max\{R_1, R_2\}$.}
\end{nexercise}

  \chapter{Markov Decision Processes}\label{sec:mdp}

We will now turn to the topic of probabilistic planning.\pidx{planning}[idxpagebf]
Planning deals with the problem of deciding which action an agent should play in a (stochastic) environment.\footnote{An environment is \idx<stochastic>{stochastic environment} as opposed to deterministic, when the outcome of actions is random.}
A key formalism for probabilistic planning in \emph{known} environments are so-called Markov decision processes.
Starting from the next chapter, we will look at reinforcement learning, which extends probabilistic planning to unknown environments.

Consider the setting where we have a sequence of states $(X_t)_{t\in\Nat_0}$ similarly to Markov chains.
But now, the next state $X_{t+1}$ of an agent does not only depend on the previous state $X_t$ but also depends on the last action $A_t$ of this agent.

\begin{marginfigure}
  \incfig{mdp}
  \caption{Directed graphical model of a Markov decision process with hidden states $X_t$ and actions $A_t$.}
\end{marginfigure}

\begin{defn}[(Finite) Markov decision process, MDP]\pidx{Markov decision process}
  A \emph{(finite) Markov decision process} is specified by \begin{itemize}
    \item a (finite) set of \emph{states} $\sX \defeq \{1, \dots, n\}$,
    \item a (finite) set of \emph{actions} $\sA \defeq \{1, \dots, m\}$,
    \item \emph{transition probabilities} \begin{align}
      p(x' \mid x, a) \defeq \Pr{X_{t+1} = x' \mid X_t = x, A_t = a}
    \end{align} which is also called the \midx{dynamics model}, and
    \item a \emph{reward function} $r : X \times A \to \R$ which maps the current state $x$ and an action $a$ to some reward.
  \end{itemize}
\end{defn}

The reward function may also depend on the next state $x'$, however, we stick to the above model for simplicity.
Also, the reward function can be random with mean $r$.
Observe that $r$ induces the sequence of rewards $(R_t)_{t\in\Nat_0}$, where \begin{align}
  R_t \defeq r(X_t, A_t),
\end{align} which is sometimes used in the literature instead of $r$.

Crucially, we assume the dynamics model $p$ and the reward function~$r$ to be known.
That is, we operate in a known environment.
For now, we also assume that the environment is \midx<fully observable>{fully observable environment}.
In other words, we assume that our agent knows its current state.
In \cref{sec:mdp:partial_observability}, we discuss how this method can be extended to the partially observable setting.

Our fundamental objective is to learn how the agent should behave to optimize its reward.
In other words, given its current state, the agent should decide (optimally) on the action to play.
Such a decision map --- whether optimal or not --- is called a policy.

\begin{defn}[Policy]\pidx{policy}
  A \emph{policy} is a function that maps each state $x \in \sX$ to a probability distribution over the actions. That is, for any ${t > 0}$,\looseness=-1 \begin{align}
    \pi(a \mid x) \defeq \Pr{A_t = a \mid X_t = x}.
  \end{align}
  In other words, a policy assigns to each action $a \in \sA$, a probability of being played given the current state $x \in \sX$.
\end{defn}

We assume that policies are stationary, that is, do not change over time.

\begin{rmk}{Stochastic policies}{}
  We will see later in this chapter that in fully observable environments optimal policies are always deterministic.
  Thus, there is no need to consider stochastic policies in the context of Markov decision processes.
  For this chapter, you can think of a policy $\pi$ simply as a deterministic mapping $\pi : \sX \to \sA$ from current state to played action.
  In the context of reinforcement learning, we will later see in \cref{sec:mfarl:actor_critic_methods:randomized_policies} that randomized policies are important in trading exploration and exploitation.
\end{rmk}

Observe that a policy induces a Markov chain $(X_t^\pi)_{t\in\Nat_0}$ with transition probabilities,\looseness=-1 \begin{align}
  p^\pi(x' \mid x) \defeq \Pr{X_{t+1}^\pi = x' \mid X_t^\pi = x} = \sum_{a \in \sA} \pi(a \mid x) p(x' \mid x, a).
\end{align}
This is crucial: if our agent follows a fixed policy (i.e., decision-making protocol) then the evolution of the process is described fully by a Markov chain.

As mentioned, we want to maximize the reward.
There are many models of calculating a score from the infinite sequence of rewards~$(R_t)_{t\in\Nat_0}$.
For the purpose of our discussion of Markov decision processes and reinforcement learning, we will focus on a very common reward called discounted payoff.

\begin{defn}[Discounted payoff]\pidx{discounted payoff}[idxpagebf]
  The \emph{discounted payoff} (also called \emph{discounted total reward}) from time $t$ is defined as the random variable, \begin{align}
    G_t \defeq \sum_{m=0}^\infty \gamma^m R_{t+m} \label{eq:discounted_payoff}
  \end{align} where $\gamma \in [0, 1)$ is the \emph{discount factor}.
\end{defn}

\begin{rmk}{Other reward models}{other_reward_models}
  Other well-known methods for combining rewards into a score are \begin{align*}
    \overset{\text{(\midx{instantaneous})}}{\vphantom{\sum_{m=0}^M}G_t \defeq R_t},\quad \overset{\text{(\midx{finite-horizon})}}{G_t \defeq \sum_{m=0}^{T-1} R_{t+m}},\quad\text{and}\quad \overset{\text{(\midx{mean payoff})}}{G_t \defeq \liminf_{T\to\infty} \frac{1}{T} \sum_{m=0}^T R_{t+m}}.
  \end{align*}
  The methods that we will discuss can also be analyzed using these or other alternative reward models.
\end{rmk}

We now want to understand the effect of the starting state and initial action on our optimization objective $G_t$.
To analyze this, it is common to use the following two functions:

\begin{defn}[State value function]\pidx{state value function}
  The \emph{state value function},\footnote{Recall that following a fixed policy $\pi$ induces a Markov chain $(X_t^\pi)_{t\in\Nat_0}$. We define \begin{align}
    \E[\pi]{\cdot} \defeq \E[(X_t^\pi)_{t\in\Nat_0}]{\cdot} \label{eq:expectation_over_policy}
  \end{align} as an expectation over all possible sequences of states $(x_t)_{t\in\Nat_0}$ within this Markov chain.} \begin{align}
    \v[\pi]{x}[t] \defeq \E[\pi]{G_t \mid X_t = x}, \label{eq:value_function}
  \end{align} measures the average discounted payoff from time $t$ starting from state~${x \in \sX}$.\looseness=-1
\end{defn}
\begin{defn}[State-action value function]\pidx{state-action value function}
  The \emph{state-action value function} (also called \idx{Q-function}), \begin{align}
    \q[\pi]{x}{a}[t] &\defeq \E[\pi]{G_t \mid X_t = x, A_t = a} \label{eq:q_function1} \\
    &= r(x, a) + \gamma \sum_{x' \in \sX} p(x' \mid x, a) \cdot \v[\pi]{x'}[t+1], \label{eq:q_function2} \margintag{\normalfont by expanding the defition of the discounted payoff \eqref{eq:discounted_payoff}; corresponds to one step in the induced Markov chain}
  \end{align} measures the average discounted payoff from time $t$ starting from state~${x \in \sX}$ and with playing action ${a \in \sA}$.
  In other words, it combines the immediate return with the value of the next states.
\end{defn}

Note that both $\v[\pi]{x}[t]$ and $\q[\pi]{x}{a}[t]$ are deterministic scalar-valued functions.
Because we assumed stationary dynamics, rewards, and policies, the discounted payoff starting from a given state $x$ will be independent of the start time $t$.
Thus, we write ${\v[\pi]{x} \defeq \v[\pi]{x}[0]}$ and ${\q[\pi]{x}{a} \defeq \q[\pi]{x}{a}[0]}$ without loss of generality.

\section{Bellman Expectation Equation}

Let us now see how we can compute the value function, \begin{align}
  \v[\pi]{x} &= \E[\pi]{G_0 \mid X_0 = x} \margintag{using the definition of the value function \eqref{eq:value_function}} \nonumber \\
  &= \E[\pi]{\sum_{m=0}^\infty \gamma^m R_m}[X_0 = x] \margintag{using the definition of the discounted payoff \eqref{eq:discounted_payoff}} \nonumber \\
  &= \begin{multlined}[t]
  \E[\pi]{\gamma^0 R_0}[X_0 = x] + \gamma \E[\pi]{\sum_{m=0}^\infty \gamma^m R_{m+1}}[X_0 = x]
  \end{multlined} \margintag{using linearity of expectation \eqref{eq:linearity_expectation}} \nonumber \\
  &= \begin{multlined}[t]
  r(x, \pi(x)) + \gamma \E[x']{\E[\pi]{\sum_{m=0}^\infty \gamma^m R_{m+1}}[X_1 = x']}[X_0 = x]
  \end{multlined} \margintag{by simplifying the first expectation and conditioning the second expectation on $X_1$} \nonumber \\
  &= \begin{multlined}[t]
  r(x, \pi(x)) + \gamma \sum_{x' \in \sX} p(x' \mid x, \pi(x)) \E[\pi]{\sum_{m=0}^\infty \gamma^m R_{m+1}}[X_1 = x']
  \end{multlined} \margintag{expanding the expectation on $X_1$ and using conditional independence of the discounted payoff of $X_0$ given $X_1$} \nonumber \\
  &= \begin{multlined}[t]
  r(x, \pi(x)) + \gamma \sum_{x' \in \sX} p(x' \mid x, \pi(x)) \E[\pi]{\sum_{m=0}^\infty \gamma^m R_m}[X_0 = x']
  \end{multlined} \margintag{shifting the start time of the discounted payoff using stationarity} \nonumber \\
  &= r(x, \pi(x)) + \gamma \sum_{x' \in \sX} p(x' \mid x, \pi(x)) \E[\pi]{G_0 \mid X_0 = x'} \margintag{using the definition of the discounted payoff \eqref{eq:discounted_payoff}} \nonumber \\
  &= r(x, \pi(x)) + \gamma \sum_{x' \in \sX} p(x' \mid x, \pi(x)) \cdot \v[\pi]{x'}. \margintag{using the definition of the value function \eqref{eq:value_function}} \label{eq:bellman_expectation_equation} \\
  &= r(x, \pi(x)) + \gamma \E[x' \mid x, \pi(x)]{\v[\pi]{x'}}. \label{eq:bellman_expectation_equation_short} \margintag{interpreting the sum as an expectation}
\end{align}
This equation is known as the \midx{Bellman expectation equation}, and it shows a recursive dependence of the value function on itself.
The intuition is clear: the value of the current state corresponds to the reward from the next action plus the discounted sum of all future rewards obtained from the subsequent states.

For stochastic policies, the above calculation can be extended to yield, \begin{align}
  \v[\pi]{x} &= \sum_{a \in \sA} \pi(a \mid x) \parentheses*{r(x, a) + \gamma \sum_{x' \in \sX} p(x' \mid x, a) \v[\pi]{x'}} \\
  &= \E[a \sim \pi(x)]{r(x, a) + \gamma \E[x' \mid x, a]{\v[\pi]{x'}}} \\
  &= \E[a \sim \pi(x)]{\q[\pi]{x}{a}}. \label{eq:v_as_q}
\end{align}
For stochastic policies, by also conditioning on the first action, one can obtain an analogous equation for the state-action value function, \begin{align}
  \q[\pi]{x}{a} &= r(x, a) + \gamma \sum_{x' \in \sX} p(x' \mid x, a) \sum_{a' \in \sA} \pi(a' \mid x') \q[\pi]{x'}{a'} \label{eq:q_function3} \\
  &= r(x, a) + \gamma \E*[x' \mid x, a]{\E[a' \sim \pi(x')]{\q[\pi]{x'}{a'}}}.
\end{align}
Note that it does not make sense to consider a similar recursive formula for the state-action value function in the setting of deterministic policies as the action played when in state $x \in \sX$ is uniquely determined as $\pi(x)$.
In particular, \begin{align}
  \v[\pi]{x} = \q[\pi]{x}{\pi(x)}. \label{eq:svf_as_savf}
\end{align}

\begin{figure}
  \incfig{mdp_example}
  \caption{Example of an MDP, which we study in \cref{exercise:value_functions}. Suppose you are building a company. The shown MDP models ``how to become rich and famous''. Here, the action $S$ is short for \emph{saving} and the action $A$ is short for \emph{advertising}.

  Suppose you begin by being ``poor and unknown''. Then, the greedy action (i.e., the action maximizing instantaneous reward) is to save. However, within this simplified environment, saving when you are poor and unknown means that you will remain poor and unknown forever. As the potential rewards in other states are substantially larger, this simple example illustrates that following the greedy choice is generally not optimal.

  The example is adapted from Andrew Moore's lecture notes on MDPs~\citep{moore2002markov}.}\label{fig:mdp_example}
\end{figure}

\section{Policy Evaluation}\label{sec:mdp:policy_evaluation}

Bellman's expectation equation tells us how we can find the value function $\fnv[\pi]$ of a fixed policy $\pi$ using a system of linear equations!
Using,\looseness=-1 \begin{equation}\begin{gathered}
  \vv^\pi \defeq \begin{bmatrix}
    \v[\pi]{1} \\
    \vdots \\
    \v[\pi]{n}
  \end{bmatrix}, \quad \vr^\pi \defeq \begin{bmatrix}
    r(1, \pi(1)) \\
    \vdots \\
    r(n, \pi(n))
  \end{bmatrix}, \quad\text{and} \\ \mP^\pi \defeq \begin{bmatrix}
    p(1 \mid 1, \pi(1)) & \cdots & p(n \mid 1, \pi(1)) \\
    \vdots & \ddots & \vdots \\
    p(1 \mid n, \pi(n)) & \cdots & p(n \mid n, \pi(n))
  \end{bmatrix}
\end{gathered}\end{equation} and a little bit of linear algebra, the Bellman expectation equation~\eqref{eq:bellman_expectation_equation} is equivalent to \begin{align}
  &&\vv^\pi &= \vr^\pi + \gamma \mP^\pi \vv^\pi \label{eq:bellman_expectation_equation_matrix_form} \\
  \iff &&(\mI - \gamma \mP^\pi) \vv^\pi &= \vr^\pi  \nonumber \\
  \iff &&\vv^\pi &= \inv{(\mI - \gamma \mP^\pi)} \vr^\pi.
\end{align}

Solving this linear system of equations (i.e., performing matrix inversion) takes cubic time in the size of the state space.

\subsection{Fixed-point Iteration}

To obtain an (approximate) solution of $\vv^\pi$, we can use that it is the unique fixed-point of the affine mapping $\mB^\pi : \R^n \to \R^n$, \begin{align}
  \mB^\pi \vv \defeq \vr^\pi + \gamma \mP^\pi \vv. \label{eq:value_function_fixed_point_helper}
\end{align}
Using this fact (which we will prove in just a moment), we can use fixed-point iteration of $\mB^\pi$.

\begin{algorithm}[H]
  \caption{Fixed-point iteration}\label{alg:fixed_point_iteration}\pidx{fixed-point iteration}
  initialize $\vv^\pi$ (e.g., as $\vzero$)\;
  \For{$t = 1$ \KwTo $T$}{
    $\vv^\pi \gets \mB^\pi \vv^\pi = \vr^\pi + \gamma \mP^\pi \vv^\pi$\;
  }
\end{algorithm}

Fixed-point iteration has computational advantages, for example, for sparse transitions.

\begin{thm}
  $\vv^\pi$ is the unique fixed-point of $\mB^\pi$.
\end{thm}
\begin{proof}
  It is immediate from Bellman's expectation equation \eqref{eq:bellman_expectation_equation_matrix_form} and the definition of $\mB^\pi$ \eqref{eq:value_function_fixed_point_helper} that $\vv^\pi$ is a fixed-point of $\mB^\pi$.
  To prove uniqueness, we will show that $\mB^\pi$ is a contraction.

  \begin{rmk}{Contractions}{}
    A \midx{contraction} is a concept from topology.
    In a Banach space $(\spX, \norm{\cdot})$ (a metric space with a norm), $f : \spX \to \spX$ is a contraction iff there exists a $k < 1$ such that \begin{align}
      \norm{f(\vx) - f(\vy)} \leq k \cdot \norm{\vx - \vy} \label{eq:contraction}
    \end{align} for any $\vx, \vy \in \spX$.
    By the \midx{Banach fixed-point theorem}, a contraction admits a unique fixed-point.
    Intuitively, by iterating the function $f$, the distance to any fixed-point shrinks by a factor $k$ in each iteration, hence, converges to $0$.
    As we cannot converge to multiple fixed-points simultaneously, the fixed-point of a contraction $f$ must be unique.
  \end{rmk}

  Let $\vv \in \R^n$ and $\vvp \in \R^n$ be arbitrary initial guesses.
  We use the $L_\infty$ space,\footnote[][-2\baselineskip]{The \idx{$L_\infty$ norm} (also called \idx{supremum norm}) is defined as \begin{align}
    \norm{\vx}_\infty \defeq \max_i \abs{\vx(i)}. \label{eq:infinity_norm}
  \end{align}} \begin{align}
    \norm{\mB^\pi \vv - \mB^\pi \vvp}_\infty &= \norm{\vr^\pi + \gamma \mP^\pi \vv - \vr^\pi - \gamma \mP^\pi \vvp}_\infty \margintag{using the definition of $\mB^\pi$ \eqref{eq:value_function_fixed_point_helper}} \nonumber \\
    &= \gamma \norm{\mP^\pi(\vv-\vvp)}_\infty \nonumber \\
    &\leq \gamma \max_{\vx \in \sX} \sum_{x' \in \sX} p(x' \mid x, \pi(x)) \cdot \abs{\vv(x') - \vvp(x')}. \margintag{using the definition of the $L_\infty$ norm \eqref{eq:infinity_norm}, expanding the multiplication, and using $\abs{\sum_i a_i} \leq \sum_i \abs{a_i}$} \nonumber \\[5pt]
    &\leq \gamma \norm{\vv-\vvp}_\infty. \label{eq:value_function_contraction} \margintag{using $\sum_{x' \in \sX} p(x' \mid x, \pi(x)) = 1$ and $\abs{\vv(x')-\vvp(x')} \leq \norm{\vv-\vvp}_\infty$}
  \end{align}
  Thus, by \cref{eq:contraction}, $\mB^\pi$ is a contraction and by Banach's fixed-point theorem $\vv^\pi$ is its unique fixed-point.
\end{proof}

Let $\vv_t^\pi$ be the value function estimate after $t$ iterations.
Then, we have for the convergence of fixed-point iteration, \begin{align}
  \norm{\vv_t^\pi - \vv^\pi}_\infty &= \norm{\mB^\pi \vv_{t-1}^\pi - \mB^\pi \vv^\pi}_\infty \margintag{using the update rule of fixed-point iteration and $\mB^\pi \vv^\pi = \vv^\pi$} \nonumber \\[5pt]
  &\leq \gamma \norm{\vv_{t-1}^\pi - \vv^\pi}_\infty \margintag{using \eqref{eq:value_function_contraction}} \nonumber \\
  &= \gamma^t \norm{\vv_0^\pi - \vv^\pi}_\infty. \margintag{by induction} \label{eq:fixed_point_iteration_convergence}
\end{align}
This shows that fixed-point iteration converges to $\vv^\pi$ exponentially fast.\looseness=-1

\section{Policy Optimization}

Recall that our goal was to find an optimal policy, \begin{align}
  \pis \defeq \argmax_\pi \E[\pi]{G_0}.
\end{align}

We can alternatively characterize an optimal policy as follows: We define a partial ordering over policies by \begin{align}
  \pi \geq \pi' \overset{\cdot}{\iff} \v[\pi]{x} \geq \v[\pi']{x} \quad (\forall x \in \sX).
\end{align}
$\pis$ is then simply a policy which is maximal according to this partial ordering.\looseness=-1

It follows that all optimal policies have identical value functions.
Subsequently, we use $\fnv[\star] \defeq \fnv[\pis]$ and $\fnq[\star] \defeq \fnq[\pis]$ to denote the state value function and state-action value function arising from an optimal policy, respectively.
As an optimal policy maximizes the value of each state, we have that \begin{align}
  \v*{x} = \max_\pi \v[\pi]{x}, \quad \q*{x}{a} = \max_\pi \q[\pi]{x}{a}. \label{eq:opt_q_function}
\end{align}

Simply optimizing over each policy is not a good idea as there are $m^n$ deterministic policies in total.
It turns out that we can do much better.

\subsection{Greedy Policies}

Consider a policy that acts greedily according to the immediate return.
It is fairly obvious that this policy will not perform well because the agent might never get to high-reward states.
But what if someone could tell us not just the immediate return, but the long-term value of the states our agent can reach in a single step?
If we knew the value of each state our agent can reach, then we can simply pick the action that maximizes the expected value.
We will make this approach precise in the next section.

This thought experiment suggests the definition of a greedy policy with respect to a value function.

\begin{defn}[Greedy policy]\pidx{greedy policy}
  The \emph{greedy policy} with respect to a state-action value function $\fnq$ is defined as \begin{align}
    \pi_{\fnq}(x) \defeq \argmax_{a \in \sA} \q{x}{a}. \label{eq:greedy_policy1}
  \end{align}
  Analogously, we define the \emph{greedy policy} with respect to a state value function $\fnv$, \begin{align}
    \pi_{\fnv}(x) \defeq \argmax_{a \in \sA} r(x, a) + \gamma \sum_{x' \in \sX} p(x' \mid x, a) \cdot \v{x'}. \label{eq:greedy_policy2}
  \end{align}
\end{defn}
We can use $\fnv$ and $\fnq$ interchangeably \exerciserefmark{v_and_q_functions}.

\subsection{Bellman Optimality Equation}

Observe that following the greedy policy $\pi_{\fnv}$, will lead us to a new value function $\fnv[\pi_{\fnv}]$.
With respect to this value function, we can again obtain a greedy policy, of which we can then obtain a new value function.
In this way, the correspondence between greedy policies and value functions induces a cyclic dependency, which is visualized in~\cref{fig:value_function_greedy_policy_cycle}.\looseness=-1

It turns out that the optimal policy $\pis$ is a fixed-point of this dependency.
This is made precise by the following theorem.

\begin{marginfigure}
  \incfig{bellman_optimality_equation}
  \caption{Cyclic dependency between \textbf{\b{value function}} and \textbf{\r{greedy policy}}.}\label{fig:value_function_greedy_policy_cycle}
\end{marginfigure}

\begin{thm}[Bellman's theorem]\pidx{Bellman's theorem}
  A policy $\pis$ is optimal iff it is greedy with respect to its own value function.
  In other words, $\pis$ is optimal iff $\pis(x)$ is a distribution over the set $\argmax_{a \in \sA} \q*{x}{a}$.
\end{thm}

In particular, if for every state there is a unique action that maximizes the state-action value function, the policy $\pis$ is deterministic and unique, \begin{align}
  \pis(x) = \argmax_{a \in \sA} \q*{x}{a}. \label{eq:bop1}
\end{align}

\begin{proof}
  It is a direct consequence of \cref{eq:opt_q_function} that a policy is optimal iff it is greedy with respect to $\fnq[\star]$.
\end{proof}

This theorem confirms our intuition from the previous section that greedily following an optimal value function is itself optimal.
In particular, Bellman's theorem shows that there always exists an optimal policy which is deterministic and stationary.

We have seen, that $\pis$ is a fixed-point of greedily picking the best action according to its state-action value function.
The converse is also true:\looseness=-1

\begin{cor}
  The optimal value functions $\fnv[\star]$ and $\fnq[\star]$ are a fixed-point of the so-called \midx{Bellman update}, \begin{align}
    \v*{x} &= \max_{a \in \sA} \q*{x}{a}, \label{eq:bop2} \\
    &= \max_{a \in \sA} r(x, a) + \gamma \E[x' \mid x, a]{\v*{x'}} \margintag{using the definition of the q-function \eqref{eq:q_function2}} \\
    \q*{x}{a} &= r(x, a) + \gamma \E[x' \mid x, a]{\max_{a' \in \sA} \q*{x'}{a'}}. \label{eq:bop2_q}
  \end{align}
\end{cor}
\begin{proof}
  It follows from \cref{eq:v_as_q} that \begin{align}
    \v*{x} = \E[a \sim \pis(x)]{\q*{x}{a}}.
  \end{align}
  Thus, as $\pis$ is greedy with respect to $\fnq[\star]$, $\v*{x} = \max_{a \in \sA} \q*{x}{a}$.

  \Cref{eq:bop2_q} follows analogously from \cref{eq:q_function3}.
\end{proof}

These equations are also called the \midx<Bellman optimality equations>{Bellman optimality equation}.
Intuitively, the Bellman optimality equations express that the value of a state under an optimal policy must equal the expected return for the best action from that state.
Bellman's theorem is also known as \midx{Bellman's optimality principle}, which is a more general concept.

\begin{marginbox}{Bellman's optimality principle}
  Bellman's optimality equations for MDPs are one of the main settings of Bellman's optimality principle.
  However, Bellman's optimality principle has many other important applications, for example in dynamic programming.
  Broadly speaking, Bellman's optimality principle says that optimal solutions to decision problems can be decomposed into optimal solutions to sub-problems.
\end{marginbox}

The two perspectives of Bellman's theorem naturally suggest two separate ways of finding the optimal policy.
Policy iteration uses the perspective from \cref{eq:bop1} of $\pis$ as a fixed-point of the dependency between greedy policy and value function.
In contrast, value iteration uses the perspective from \cref{eq:bop2} of $\fnv[\star]$ as the fixed-point of the Bellman update.
Another approach which we will not discuss here is to use a linear program where the Bellman update is interpreted as a set of linear inequalities.

\subsection{Policy Iteration}

\begin{algorithm}
  \caption{Policy iteration}\pidx{policy iteration}[idxpagebf]\label{alg:policy_iteration}
  initialize $\pi$ (arbitrarily)\;
  \Repeat{converged}{
    compute $\fnv[\pi]$\;
    compute $\pi_{\fnv[\pi]}$\;
    $\pi \gets \pi_{\fnv[\pi]}$\;
  }
\end{algorithm}

Starting from an arbitrary initial policy, policy iteration as shown in \cref{alg:policy_iteration} uses the Bellman expectation equation to compute the value function of that policy (as we have discussed in \cref{sec:mdp:policy_evaluation}) and then chooses the greedy policy with respect to that value function as its next iterate.

Let $\pi_t$ be the policy after $t$ iterations. We will now show that policy iteration converges to the optimal policy.
The proof is split into two parts.
First, we show that policy iteration improves policies monotonically.
Then, we will use this fact to show that policy iteration converges.\looseness=-1

\begin{lem}[Monotonic improvement of policy iteration]\label{lem:policy_iteration_mon_impr}
  We have, \begin{itemize}
    \item $\v[\pi_{t+1}]{x} \geq \v[\pi_t]{x}$ for all $x \in \sX$; and
    \item $\v[\pi_{t+1}]{x} > \v[\pi_t]{x}$ for at least one $x \in \sX$, unless $\fnv[\pi_t] \equiv \fnv[\star]$.
  \end{itemize}
\end{lem}
\begin{proof}
  We consider the Bellman update from \eqref{eq:bop2} as the mapping $\mBs : \R^n \to \R^n$, \begin{align}
    (\mBs \vv)(x) \defeq \max_{a \in \sA} \q{x}{a}, \label{eq:bellman_update_operator}
  \end{align} where $\fnq$ is the state-action value function corresponding to the state value function $\vv \in \R^n$.
  Recall that after obtaining $\fnv[\pi_t]$, policy iteration first computes the greedy policy w.r.t. $\fnv[\pi_t]$, $\pi_{t+1} \defeq \pi_{\fnv[\pi_t]}$, and then computes its value function $\fnv[\pi_{t+1}]$.

  To establish the (weak) monotonic improvement of policy iteration, we consider a fixed-point iteration (cf. \cref{alg:fixed_point_iteration}) of $\vv^{\pi_{t+1}}$ initialized by~$\vv^{\pi_t}$.
  We denote the iterates by $\tilde{\vv}_\tau$, in particular, we have that $\tilde{\vv}_0 = \vv^{\pi_t}$ and $\lim_{\tau\to\infty} \tilde{\vv}_\tau = \vv^{\pi_{t+1}}$.\footnote{using the convergence of fixed-point iteration \eqref{eq:fixed_point_iteration_convergence}}
  First, observe that for the first iteration of fixed-point iteration, \begin{align*}
    \tilde{v}_1(x) &= (\mBs \vv^{\pi_t})(x) \margintag{using that $\pi_{t+1}$ is greedy wrt. $v^{\pi_t}$} \\
    &= \max_{a \in \sA} \q[\pi_t]{x}{a} \margintag{using the definition of the Bellman update \eqref{eq:bellman_update_operator}} \\
    &\geq \q[\pi_t]{x}{\pi_t(x)} \\
    &= \v[\pi_t]{x} \\
    &= \tilde{v}_0(x). \margintag{using \eqref{eq:svf_as_savf}}
  \end{align*}
  Let us now consider a single iteration of fixed-point iteration.
  We have, \begin{align*}
    \tilde{v}_{\tau+1}(x) &= r(x, \pi_{t+1}(x)) + \gamma \sum_{x' \in \sX} p(x' \mid x, \pi_{t+1}(x)) \cdot \tilde{v}_\tau(x'). \margintag{using the definition of $\tilde{\vv}_{\tau+1}$ \eqref{eq:value_function_fixed_point_helper}}
    \intertext{Using an induction on $\tau$, we conclude,}
    &\geq r(x, \pi_{t+1}(x)) + \gamma \sum_{x' \in \sX} p(x' \mid x, \pi_{t+1}(x)) \cdot \tilde{v}_{\tau-1}(x') \margintag{using the induction hypothesis, $\tilde{v}_\tau(x') \geq \tilde{v}_{\tau-1}(x')$} \\
    &= \tilde{v}_\tau(x).
  \end{align*}
  This establishes the first claim, \begin{align}
    \vv^{\pi_{t+1}} = \lim_{\tau\to\infty} \tilde{\vv}_\tau \geq \tilde{\vv}_0 = \vv^{\pi_t}. \label{eq:pi_monotonicity}
  \end{align}

  For the second claim, recall from Bellman's theorem \eqref{eq:bop2} that $\fnv[\star]$ is a (unique) fixed-point of the Bellman update $\mBs$.\footnote{We will show in \cref{eq:bellman_update_contraction} that $\mBs$ is a contraction, implying that $\fnv[\star]$ is the \emph{unique} fixed-point of $\mBs$.}
  In particular, we have $\fnv[\pi_{t+1}] \equiv \fnv[\pi_t]$ if and only if $\fnv[\pi_{t+1}] \equiv \fnv[\pi_t] \equiv \fnv[\star]$.
  In other words, if $\fnv[\pi_t] \not\equiv \fnv[\star]$ then \cref{eq:pi_monotonicity} is strict for at least one $x \in \sX$ and $\fnv[\pi_{t+1}] \not\equiv \fnv[\pi_t]$.
  This proves the strict monotonic improvement of policy iteration.
\end{proof}

\begin{thm}[Convergence of policy iteration]\label{thm:policy_iteration_convergence}
  For finite Markov decision processes, policy iteration converges to an optimal policy.
\end{thm}
\begin{proof}
  Finite Markov decision processes only have a finite number of deterministic policies (albeit exponentially many).
  Observe that policy iteration only considers deterministic policies, and recall that there is an optimal policy that is deterministic.
  As the value of policies strictly increase in each iteration until an optimal policy is found, policy iteration must converge in finite time.
\end{proof}

It can be shown that policy iteration converges to an exact solution in a polynomial number of iterations \citep{pi_convergence}.
Each iteration of policy iteration requires computing the value function, which we have seen to be of cubic complexity in the number of states.

\subsection{Value Iteration}

As we have mentioned, another natural approach of finding the optimal policy is to interpret $\fnv[\star]$ as the fixed point of the Bellman update.
Recall our definition of the Bellman update from \cref{eq:bellman_update_operator}, \begin{align*}
  (\mBs \vv)(x) = \max_{a \in \sA} \q{x}{a},
\end{align*} where $\fnq$ was the state-action value function associated with the state value function $\vv$.
The value iteration algorithm is shown in \cref{alg:value_iteration}.

\begin{algorithm}
  \caption{Value iteration}\pidx{value iteration}\label{alg:value_iteration}
  initialize $\v{x} \gets \max_{a \in \sA} r(x, a)$ for each $x \in \sX$\;
  \For{$t = 1$ \KwTo $\infty$}{
    $\v{x} \gets (\mBs \vv)(x) = \max_{a \in \sA} \q{x}{a}$ for each $x \in \sX$\;
  }
  choose $\pi_{\fnv}$\;
\end{algorithm}

We will now prove the convergence of value iteration using the fixed-point interpretation.

\begin{thm}[Convergence of value iteration]\label{thm:vi_convergence}
  Value iteration converges asymptotically to an optimal policy.
\end{thm}
\begin{proof}
  Clearly, value iteration converges if $\fnv[\star]$ is the unique fixed-point of $\mBs$.
  We already know from Bellman's theorem \eqref{eq:bop2} that $\fnv[\star]$ is a fixed-point of $\mBs$.
  It remains to show that it is indeed the unique fixed-point.\looseness=-1

  Analogously to our proof of the convergence of fixed-point iteration to the value function $\fnv[\pi]$, we show that $\mBs$ is a contraction. Fix arbitrary $\vv, \vvp \in \R^n$, then \begin{align}
    \norm{\mBs \vv - \mBs \vvp}_\infty &= \max_{x \in \sX} \abs{(\mBs \vv)(x) - (\mBs \vvp)(x)} \margintag{using the definition of the $L_\infty$ norm \eqref{eq:infinity_norm}} \nonumber \\
    &= \max_{x \in \sX} \abs{\max_{a \in \sA} \q{x}{a} - \max_{a \in \sA} \fnq'(x,a)} \margintag{using the definition of the Bellman update \eqref{eq:bellman_update_operator}} \nonumber \\
    &\leq \max_{x \in \sX} \max_{a \in \sA} \abs{\q{x}{a}-\fnq'(x,a)} \margintag{using $\abs{\max_x f(x) - \max_x g(x)} \leq \max_x \abs{f(x) - g(x)}$} \nonumber \\
    &\leq \gamma \max_{x \in \sX} \max_{a \in \sA} \sum_{x' \in \sX} p(x' \mid x, a) \abs{\vv(x')-\vvp(x')} \margintag{using the definition of the Q-function \eqref{eq:q_function2} and $\abs{\sum_i a_i} \leq \sum_i \abs{a_i}$} \nonumber \\
    &\leq \gamma \norm{\vv-\vvp}_\infty \label{eq:bellman_update_contraction} \margintag{using $\sum_{x' \in \sX}p(x' \mid x, a) = 1$ and $ \abs{\vv(x')-\vvp(x')} \leq \norm{\vv-\vvp}_\infty$}
  \end{align} where $\fnq$ and $\fnq'$ are the state-action value functions associated with $\vv$ and $\vvp$, respectively.
  By \cref{eq:contraction}, $\mBs$ is a contraction and by Banach's fixed-point theorem $\fnv[\star]$ is its unique fixed-point.
\end{proof}

\begin{rmk}{Value iteration as a dynamic program}{}
  Let us denote by $\fnv[][t]$ the value function estimate after the $t$-th iteration.
  Observe that $\v{x}[t]$ corresponds to the maximum expected reward when starting in state $x$ and the ``world ends'' after $t$ time steps.
  In particular, $\fnv[][0]$ corresponds to the maximum immediate reward.
  This suggests a different perspective on value iteration (akin to dynamic programming) where in each iteration we extend the time horizon of our approximation by one time step.
\end{rmk}

For any $\epsilon > 0$, value iteration converges to an $\epsilon$-optimal solution in polynomial time.
However, unlike policy iteration, value iteration does not generally reach the \emph{exact} optimum in a finite number of iterations.
Recalling the update rule of value iteration, its main benefit is that each iteration only requires a sum over all possible actions $a$ in state $x$ and a sum over all reachable states $x'$ from $x$.
In sparse Markov decision processes,\footnote{Sparsity refers to the interconnectivity of the state space. When only few states are reachable from any state, we call an MDP sparse.} an iteration of value iteration can be performed in (virtually) constant time.

\section{Partial Observability}\label{sec:mdp:partial_observability}

So far we have focused on the fully observable setting.
That is, at any time, our agent knows its current state.
We have seen that we can efficiently find the optimal policy (as long as the Markov decision process is finite).

We have already encountered the partially observable setting in \cref{sec:kf}, where we discussed filtering\pidx{filtering}.
In this section, we consider how Markov decision processes can be extended to a partially observable setting where the agent can only access noisy observations $Y_t$ of its state $X_t$.

\begin{marginfigure}[5\baselineskip]
  \incfig{pomdp}
  \caption{Directed graphical model of a partially observable Markov decision process with hidden states $X_t$, observables $Y_t$, and actions $A_t$.}
\end{marginfigure}

\begin{defn}[Partially observable Markov decision process, POMDP]\pidx{partially observable Markov decision process}
  Similarly to a Markov decision process, a \emph{partially observable Markov decision process} is specified by \begin{itemize}
    \item a set of \emph{states} $\sX$,
    \item a set of \emph{actions} $\sA$,
    \item \emph{transition probabilities} $p(x' \mid x, a)$, and
    \item a \emph{reward function} $r : X \times A \to \R$.
  \end{itemize} Additionally, it is specified by \begin{itemize}
    \item a set of \emph{observations} $\sY$, and
    \item \emph{observation probabilities} \begin{align}
      o(y \mid x) \defeq \Pr{Y_t = y \mid X_t = x}. \label{eq:observation_probability}
    \end{align}
  \end{itemize}
\end{defn}

Whereas MDPs are controlled Markov chains, POMDPs are controlled hidden Markov models.

\begin{marginfigure}[15\baselineskip]
  \incfig{hmm}
  \caption{Directed graphical model of a hidden Markov model with hidden states $X_t$ and observables $Y_t$.}\label{fig:hmm}
\end{marginfigure}

\begin{rmk}{Hidden Markov models}{}
  A hidden Markov model is a Markovian process with unobservable states $X_t$ and observations $Y_t$ that depend on $X_t$ in a known way.\looseness=-1

  \begin{defn}[Hidden Markov model, HMM]\pidx{hidden Markov model}
    A \emph{hidden Markov model} is specified by \begin{itemize}
      \item a set of \emph{states} $\sX$,
      \item \emph{transition probabilities} $p(x' \mid x) \defeq \Pr{X_{t+1} = x' \mid X_t = x}$ (also called \emph{motion model}), and
      \item a \emph{sensor model} $o(y \mid x) \defeq \Pr{Y_t = y \mid X_t = x}$.
    \end{itemize}
  \end{defn}

  Following from its directed graphical model shown in \cref{fig:hmm}, its joint probability distribution factorizes into \begin{align}
    \Pr{x_{1:t}, y_{1:t}} = \Pr{x_1} \cdot o(y_1 \mid x_1) \cdot \prod_{i=2}^t p(x_i \mid x_{i-1}) \cdot o(y_i \mid x_i).
  \end{align}

  Observe that a Kalman filter can be viewed as a hidden Markov model with conditional linear Gaussian motion and sensor models and a Gaussian prior on the initial state.
  In particular, the tasks of \emph{filtering}, \emph{smoothing}, and \emph{predicting} which we discussed extensively in \cref{sec:kf} are also of interest for hidden Markov models.

  A widely used application of hidden Markov models is to find the most likely sequence (also called \midx{most likely explanation}) of hidden states $x_{1:t}$ given a series of observations $y_{1:t}$,\safefootnote{This is useful in many applications such as speech recognition, decoding data that was transmitted over a noisy channel, beat detection, and many more.} that is, to find \begin{align}
    \argmax_{x_{1:t}} \Pr{x_{1:t} \mid y_{1:t}}.
  \end{align}
  This task can be solved in linear time by a simple backtracking algorithm known as the \midx{Viterbi algorithm}.
\end{rmk}

POMDPs are a very powerful model, but very hard to solve in general.
POMDPs can be reduced to a Markov decision process with an enlarged state space.
The key insight is to consider an MDP whose states are the \midx<beliefs>{belief}[idxpagebf], \begin{align}
  b_t(x) \defeq \Pr{X_t = x \mid y_{1:t}, a_{1:t-1}}, \label{eq:belief}
\end{align} about the current state in the POMDP.
In other words, the states of the MDP are probability distributions over the states of the POMDP.
We will make this more precise in the following.

Let us assume that our prior belief about the state of our agent is given by $b_0(x) \defeq \Pr{X_0 = x}$.
Keeping track of how beliefs change over time is known as \midx{filtering}, which we already encountered in \cref{sec:kf:bayesian_filtering}.
Given a prior belief $b_t$, an action taken $a_t$, and a new observation $y_{t+1}$, the belief state can be updated as follows, \begin{align}
  b_{t+1}(x) &= \Pr{X_{t+1} = x \mid y_{1:t+1}, a_{1:t}} \margintag{by the definition of beliefs \eqref{eq:belief}} \nonumber \\
  &= \frac{1}{Z} \Pr{y_{t+1} \mid X_{t+1} = x} \Pr{X_{t+1} = x \mid y_{1:t}, a_{1:t}} \margintag{using Bayes' rule \eqref{eq:bayes_rule}} \nonumber \\
  &= \frac{1}{Z} o(y_{t+1} \mid x) \Pr{X_{t+1} = x \mid y_{1:t}, a_{1:t}} \margintag{using the definition of observation probabilities \eqref{eq:observation_probability}} \nonumber \\
  &= \frac{1}{Z} o(y_{t+1} \mid x) \sum_{x' \in \sX} p(x \mid x', a_t) \Pr{X_t = x' \mid y_{1:t}, a_{1:t-1}} \margintag{by conditioning on the previous state $x'$, noting $a_t$ does not influence $X_t$} \nonumber \\
  &= \frac{1}{Z} o(y_{t+1} \mid x) \sum_{x' \in \sX} p(x \mid x', a_t) b_t(x') \margintag{using the definition of beliefs \eqref{eq:belief}}
\end{align} where \begin{align}
  Z \defeq \sum_{x \in \sX} o(y_{t+1} \mid x) \sum_{x' \in \sX} p(x \mid x', a_t) b_t(x').
\end{align}
Thus, the updated belief state is a deterministic mapping from the previous belief state depending only on the (random) observation $y_{t+1}$ and the taken action $a_t$.
Note that this obeys a Markovian structure of transition probabilities with respect to the beliefs $b_t$.

The sequence of belief-states defines the sequence of random variables~$(B_t)_{t\in\Nat_0}$,\looseness=-1 \begin{align}
  B_t \defeq X_t \mid y_{1:t}, a_{1:t-1},
\end{align} where the (state-)space of all beliefs is the (infinite) space of all probability distributions over $\sX$,\footnote{This definition naturally extends to continuous state spaces $\spX$.} \begin{align}
  \spB \defeq \Delta^{\sX} \defeq \braces*{\vb \in \R^{\card{\sX}} : \vb \geq \vzero, \textstyle\sum_{i=1}^{\card{\sX}} \vb(i) = 1}.
\end{align}
A Markov decision process, where every belief corresponds to a state is called a belief-state MDP.

\begin{defn}[Belief-state Markov decision process]\pidx{belief-state Markov decision process}
  Given a POMDP, the corresponding \emph{belief-state Markov decision process} is a Markov decision process specified by \begin{itemize}
    \item the \emph{belief space} $\spB \defeq \Delta^{\sX}$ depending on the \emph{hidden states} $\sX$,
    \item the set of \emph{actions} $\sA$,
    \item \emph{transition probabilities} \begin{align}
      \tau(b' \mid b, a) \defeq \Pr{B_{t+1} = b' \mid B_t = b, A_t = a},
    \end{align}
    \item and \emph{rewards} \begin{align}
      \rho(b, a) \defeq \E[x \sim b]{r(x, a)} = \sum_{x \in \sX} b(x) r(x, a).
    \end{align}
  \end{itemize}
\end{defn}

It remains to derive the transition probabilities $\tau$ in terms of the original POMDP.
We have, \begin{align}
  \tau(b_{t+1} \mid b_t, a_t) &= \Pr{b_{t+1} \mid b_t, a_t} \nonumber \\
  &= \sum_{y_{t+1} \in \sY} \Pr{b_{t+1} \mid b_t, a_t, y_{t+1}} \Pr{y_{t+1} \mid b_t, a_t}. \margintag{by conditioning on $y_{t+1} \in \sY$}
\end{align}
Using the Markovian structure of the belief updates, we naturally set, \begin{align}
  \Pr{b_{t+1} \mid b_t, a_t, y_{t+1}} \defeq \begin{cases}
    1 & \parbox[t]{0.5\linewidth}{if $b_{t+1}$ matches the belief update of \cref{eq:belief} given $b_t, a_t$, and $y_{t+1}$,} \\
    0 & \text{otherwise}.
  \end{cases}
\end{align}

The final missing piece is the likelihood of an observation $y_{t+1}$ given the prior belief $b_t$ and action $a_t$, which using our interpretation of beliefs corresponds to \begin{align}
  \Pr{y_{t+1} \mid b_t, a_t} &= \E[x \sim b_t]{\E[x' \mid x, a_t]{\Pr{y_{t+1} \mid X_{t+1} = x'}}} \nonumber \\
  &= \E[x \sim b_t]{\E[x' \mid x, a_t]{o(y_{t+1} \mid x')}} \margintag{using the definition of observation probabilities \eqref{eq:observation_probability}} \nonumber \\
  &= \sum_{x \in \sX} b_t(x) \sum_{x' \in \sX} p(x' \mid x, a_t) \cdot o(y_{t+1} \mid x').
\end{align}

In principle, we can now apply arbitrary algorithms for planning in MDPs to POMDPs.
Of course, the problem is that there are infinitely many beliefs, even for a finite state space $\sX$.\footnote{You can think of an $\card{\sX}$-dimensional space. Here, all points whose coordinates sum to $1$ correspond to probability distributions (i.e., beliefs) over the hidden states $\sX$. The convex hull of these points is also known as the $(\card{\sX}-1)$-dimensional \idx{probability simplex} (cf. \cref{sec:background:probability:probability_simplex}). Now, by definition of the $(\card{\sX}-1)$-dimensional probability simplex as a polytope in $\card{\sX}-1$ dimensions, we can conclude that its boundary consists of infinitely many points in $\card{\sX}$ dimensions. Noting that these points corresponded to the probability distributions on $\card{\sX}$, we conclude that there are infinitely many such distributions.}
The belief-state MDP has therefore an infinitely large belief space $\spB$.
Even when only planning over finite horizons, exponentially many beliefs can be reached.
So the belief space blows-up very quickly.

We will study MDPs with large state spaces (where transition dynamics and rewards are unknown) in \cref{sec:mfarl,sec:mbarl}.
Similar methods can also be used to approximately solve POMDPs.

A key idea in approximate solutions to POMDPs is that most belief states are never reached.
A common approach is to discretize the belief space by sampling or by applying a dimensionality reduction.
Examples are \midx{point-based value iteration} (PBVI) and \midx{point-based policy iteration}~(PBPI)~\citep{shani2013survey}.

\section*{Discussion}

Even though we focus on the fully observed setting throughout this manuscript, the partially observed setting can be reduced to the fully observed setting with very large state spaces.
In the next chapter, we will consider learning and planning in unknown Markov decision processes (i.e., reinforcement learning) for small state spaces.
The setting of small state and action spaces is also known as the \emph{tabular setting}.
Then, in the final two chapters, we will consider approximate methods for large state and action spaces.
In particular, in \cref{sec:mbarl:planning}, we will revisit the problem of probabilistic planning in known Markov decision processes, but with continuous state and action spaces.

\excheading

\begin{nexercise}{Value functions}{value_functions}
  Recall the example of ``becoming rich and famous'' from \cref{fig:mdp_example}.
  Consider the policy, $\pi \equiv S$ (i.e., to always \emph{save}) and let $\gamma = \nicefrac{1}{2}$. Show that the (rounded) state-action value function $\fnq[\pi]$ is as follows:

  \vspace{5pt}
  \begin{center}
    \begin{tabular}{lrr}
      \toprule
      & save & advertise \\
      \midrule
      poor, unknown & $\mathbf{0}$ & $0.1$ \\
      poor, famous & $\mathbf{4.4}$ & $1.2$ \\
      rich, famous & $\mathbf{17.8}$ & $1.2$ \\
      rich, unknown & $\mathbf{13.3}$ & $0.1$ \\
      \bottomrule
    \end{tabular}
  \end{center}
  \vspace{5pt}

  Shown in bold is the state value function $\fnv[\pi]$.
\end{nexercise}

\begin{nexercise}{Greedy policies}{v_and_q_functions}
  Show that if $\fnq$ and $\fnv$ arise from the same policy, that is, $\fnq$ is defined in terms of $\fnv$ as per \cref{eq:q_function2}, then \begin{align}
    \pi_{\fnv} \equiv \pi_{\fnq}.
  \end{align}
  This implies that we can use $\fnv$ and $\fnq$ interchangeably.
\end{nexercise}

\begin{nexercise}{Optimal policies}{optimal_policies}
  Again, recall the example of ``becoming rich and famous'' from \cref{fig:mdp_example}.

  \begin{enumerate}
    \item Show that the policy $\pi \equiv S$, which we considered in \cref{exercise:value_functions}, is not optimal.
    \item Instead, consider the policy \begin{align*}
      \pi' \equiv \begin{cases}
        A & \text{if poor and unknown} \\
        S & \text{otherwise} \\
      \end{cases}
    \end{align*} and let $\gamma = \nicefrac{1}{2}$. Show that the (rounded) state-action value function $\fnq[\pi']$ is as follows:

    \vspace{5pt}
    \begin{center}
      \begin{tabular}{lrr}
        \toprule
        & save & advertise \\
        \midrule
        poor, unknown & $0.8$ & $\mathbf{1.6}$ \\
        poor, famous & $\mathbf{4.5}$ & $1.2$ \\
        rich, famous & $\mathbf{17.8}$ & $1.2$ \\
        rich, unknown & $\mathbf{13.4}$ & $0.2$ \\
        \bottomrule
      \end{tabular}
    \end{center}
    \vspace{5pt}

    Shown in bold is the state value function $\fnv[\pi']$.
    \item Is the policy $\pi'$ optimal?
  \end{enumerate}
\end{nexercise}

\begin{nexercise}{Linear convergence of policy iteration}{policy_iteration_linear_convergence}
  Denote by $\pi_t$ the policy obtained by policy iteration after $t$ iterations.
  Use that the Bellman operator $\mBs$ is a contraction with the unique fixed-point $\vvs$ to show that \begin{align}
    \norm{\vv^{\pi_t} - \vvs}_\infty \leq \gamma^t \norm{\vv^{\pi_0} - \vvs}_\infty
  \end{align} where $\vv^{\pi}$ and $\vvs$ are vector representations of the functions $\fnv[\pi]$ and $\fnv[\star]$, respectively.\looseness=-1

  \textit{Hint: Recall from \cref{lem:policy_iteration_mon_impr} that $\vv^{\pi_{t+1}} \geq \mBs\vv^{\pi_t} \geq \vv^{\pi_t}$.}
\end{nexercise}

\begin{nexercise}{Reward modification}{reward_modification}
  A key technique for solving sequential decision problems is the modification of reward functions that leaves the optimal policy unchanged while improving sample efficiency or convergence rates.
  This exercise looks at simple ways of modifying rewards and understanding how these modifications affect the optimal policy.

  Consider two Markov decision processes $\sM \defeq (\sX, \sA, p, r)$ and ${\sM' \defeq (\sX, \sA, p, r')}$ where the reward function $r$ is modified to obtain $r'$, and the rewards are bounded and discounted by the discount factor ${\gamma \in [0,1)}$.
  Let $\pis_\sM$ be the optimal policy for $\sM$.

  \begin{enumerate}
    \item Suppose $r'(x) = \alpha r(x)$, where $\alpha > 0$. Show that the optimal policy~$\pis$ of~$\sM$ is also an optimal policy of~$\sM'$.
    \item Given a modification of the form $r'(x) = r(x) + c$, where $c > 0$ is a constant scalar, show that the optimal policy~$\pis_\sM$ can be different from $\pis_{\sM'}$.
    \item Another way of modifying the reward function is through \midx{reward shaping} where one supplies additional rewards to the agent to guide the learning process.
    When one has no knowledge of the underlying transition dynamics $p$, a commonly used transformation is ${r'(x, x') = r(x, x') + f(x, x')}$ where $f$ is a \emph{potential-based} shaping function defined as \begin{align}
      f(x, x') \defeq \gamma \phi(x') - \phi(x), \quad \phi : \sX \to \R.
    \end{align}

    Show that the optimal policy remains unchanged under this definition of $f$.
  \end{enumerate}
\end{nexercise}

\begin{nexercise}{A partially observable fishing problem}{fishing_problem_pomdb}
  We model an angler's decisions while fishing, where the states are partially observable.
  There are two states: (1) \textbf{Fish} ($F$): A fish is hooked on the line. (2) \textbf{No fish} ($\overline{F}$): No fish is hooked on the line.

  The angler can choose between two actions: \begin{itemize}
    \item \textbf{Pull up the rod} ($P$): If there is a fish on the line ($F$), there is a $90\%$ chance of catching it (reward $+10$, transitioning to $\overline{F}$) and a $10\%$ chance of it escaping (reward $-1$, transitioning to $\overline{F}$).
    If there is no fish ($\overline{F}$), pulling up the rod results in no catch, staying in $\overline{F}$ with a reward of $-5$.

    \item \textbf{Waiting} ($W$): All waiting actions result in a reward of $-1$.
    In state $F$, there is a $60\%$ chance of the fish staying (remaining in $F$) and a $40\%$ chance of it escaping (transitioning to $\overline{F}$).
    In state $\overline{F}$, there is a $50\%$ chance of a fish biting (transitioning to $F$) and a $50\%$ chance of no change (remaining in $\overline{F}$).
  \end{itemize}

  \textit{Suggestion: Draw the MDP transition diagram. Draw each transition with action,  associated probability, and associated reward.}

  Since the angler cannot directly observe whether there is a fish on the line, they receive a noisy observation about the state.
  This observation can be: \begin{itemize}
    \item $o_1$: The signal suggests that a fish might be on the line.
    \item $o_2$: The signal suggests that there is no fish on the line.
  \end{itemize}
  The observation model, which defines the probability of receiving each observation given the true state is as follows:

  \vspace{5pt}
  \begin{center}
    \begin{tabular}{lcc}
      \toprule
      & $\Pr{o_1 \mid \cdot}$ & $\Pr{o_2 \mid \cdot}$ \\
      \midrule
      $F$ & $0.8$ & $0.2$ \\
      $\overline{F}$ & $0.3$ & $0.7$ \\
      \bottomrule
    \end{tabular}
  \end{center}
  \vspace{5pt}

  The angler's goal is to choose actions that maximize their overall reward, balancing the chances of catching a fish against the cost of waiting and unsuccessful pulls.

  \begin{enumerate}
    \item Given an initial belief $b_0(F) = b_0(\overline{F}) = 0.5$, the angler chooses to wait and observes $o_1$.
    Compute the updated belief $b_1$ using the observation model and belief update equation~\eqref{eq:belief}.

    \item Given belief $b_1(F) \approx 0.765$ and $b_1(\overline{F}) \approx 0.235$, compute the updated belief $b_2$ for both actions $P$ (pull) and $W$ (wait), both in the case where you observe $o_1$ (fish likely) and $o_2$ (fish unlikely).
  \end{enumerate}
\end{nexercise}

  \chapter{Tabular Reinforcement Learning}\label{sec:tabular_rl}

\section{The Reinforcement Learning Problem}\pidx{reinforcement learning}

Reinforcement learning is concerned with probabilistic planning in unknown environments.\pidx{planning}
This extends our study of known environments in the previous chapter.
Those environments are still modeled by Markov decision processes, but in reinforcement learning, we do not know the dynamics $p$ and rewards $r$ in advance.
Hence, reinforcement learning is at the intersection of the theories of probabilistic planning (i.e., Markov decision processes) and learning (e.g., multi-armed bandits), which we covered extensively in the previous chapters.

\begin{marginfigure}
  \incfig{reinforcement_learning}
  \caption{In reinforcement learning, an agent interacts with its environment in a sequence of rounds.
  After playing an action~$a_t$, it observes rewards~$r_t$ and its new state~$x_{t+1}$.
  The agent then uses this information to learn how to act to maximize reward.\looseness=-1}\label{fig:rl}
\end{marginfigure}

We will continue to focus on the fully observed setting, where the agent knows its current state.
As we have seen in the previous section, the partially observed setting corresponds to a fully observed setting with an enlarged state space.
In this chapter, we will begin by considering reinforcement learning with small state and action spaces.
This setting is often called the \midx{tabular setting}, as the value functions can be computed exhaustively for all states and stored in a table.

\pidx{exploration-exploitation dilemma}%

Clearly, the agent needs to trade exploring and learning about the environment with exploiting its knowledge to maximize rewards.
Thus, the exploration-exploitation dilemma, which was at the core of Bayesian optimization (see \cref{sec:bayesian_optimization:exploration_exploitation}), also plays a crucial role in reinforcement learning.
In fact, Bayesian optimization can be viewed as reinforcement learning with a fixed state:
In each round, the agent plays an action, aiming to find the action that maximizes the reward.
However, playing the same action multiple times yields the same reward, implying that we remain in a single state.
In the context of Bayesian optimization, we used ``regret'' as performance metric: in the jargon of planning, minimizing regret corresponds to maximizing the cumulative reward.

Another key challenge of reinforcement learning is that the observed data is dependent on the played actions.
This is in contrast to the setting of supervised learning that we have been considering in earlier chapters, where the data is sampled independently.

\subsection{Trajectories}\label{sec:tabular_rl:rl:trajectories}

The data that the agent collects is modeled using so-called trajectories.

\begin{defn}[Trajectory]\pidx{trajectory}
  A \emph{trajectory} $\tau$ is a (possibly infinite) sequence,\looseness=-1 \begin{align}
    \tau \defeq (\tau_0, \tau_1, \tau_2, \dots),
  \end{align} of \midx<transitions>{transition}, \begin{align}
    \tau_i \defeq (x_i, a_i, r_i, x_{i+1}),
  \end{align} where $x_i \in \sX$ is the starting state, $a_i \in \sA$ is the played action, $r_i \in \R$ is the attained reward, and $x_{i+1} \in \sX$ is the ending state.
\end{defn}

In the context of learning a dynamics and rewards model, $x_i$ and $a_i$ can be understood as inputs, and $r_i$ and $x_{i+1}$ can be understood as labels of a regression problem.

Crucially, the newly observed states $x_{t+1}$ and the rewards $r_t$ (across multiple transitions) are conditionally independent given the previous states $x_t$ and actions $a_t$.
This follows directly from the Markovian structure of the underlying Markov decision process.\footnote{Recall the Markov property \eqref{eq:markov_property}, which assumes that in the underlying Markov decision process (i.e., in our environment) the future state of an agent is independent of past states given the agent's current state. This is commonly called a Markovian structure. From this Markovian structure, we gather that repeated encounters of state-action pairs result in independent trials of the transition model and rewards.}
Formally, we have,\looseness=-1 \begin{subequations}\label{eq:rl_cond_indep}\begin{align}
  X_{t+1} &\perp X_{t'+1} \mid X_t, X_{t'}, A_t, A_{t'}, \\
  R_t &\perp R_{t'} \mid X_t, X_{t'}, A_t, A_{t'},
\end{align}\end{subequations} for any $t, t' \in \Nat_0$.
In particular, if $x_t = x_{t'}$ and $a_t = a_{t'}$, then $x_{t+1}$ and $x_{t'+1}$ are independent samples according to the transition model $p(X_{t+1} \mid x_t, a_t)$.
Analogously, if $x_t = x_{t'}$ and $a_t = a_{t'}$, then~$r_t$ and~$r_{t'}$ are independent samples of the reward model $r(x_t, a_t)$.
As we will see later in this chapter and especially in \cref{sec:mbarl}, this independence property is crucial for being able to learn about the underlying Markov decision process.
Notably, this implies that we can apply the law of large numbers \eqref{eq:slln} and Hoeffding's inequality \eqref{eq:hoeffdings_inequality} to our estimators of both quantities.

The collection of data is commonly classified into two settings.
In the \midx{episodic setting}, the agent performs a sequence of ``training'' rounds~(called \midx<episodes>{episode}).
In the beginning of each episode, the agent is reset to some initial state.
In contrast, in the \midx{continuous setting}~(or non-episodic / online setting), the agent learns online.
Especially, every action, every reward, and every state transition counts.

The episodic setting is more applicable to an agent playing a computer game.
That is, the agent is performing in a simulated environment that is easy to reset.
The continuous setting is akin to an agent that is deployed to the ``real world''.
In principle, real-world agents can be trained in simulated environments before being deployed.
However, this bears the risk of learning to exploit or rely on features of the simulated environment that are not present in the real environment. Sometimes, using a simulated environment for training is downright impossible, as the real environment is too complex.

\subsection{On-policy and Off-policy Methods}

Another important distinction in how data is collected, is the distinction between on-policy and off-policy methods.
As the names suggest, \midx{on-policy} methods are used when the agent has control over its own actions, in other words, the agent can freely choose to follow any policy.
Being able to follow a policy is helpful, for example because it allows the agent to experiment with trading exploration and exploitation.

In contrast, \midx{off-policy} methods can be used even when the agent cannot freely choose its actions.
Off-policy methods are therefore able to make use of purely observational data.
This might be data that was collected by another agent, a fixed policy, or during a previous episode.
Off-policy methods are therefore more \emph{sample-efficient} than on-policy methods.
This is crucial, especially in settings where conducting experiments (i.e., collecting new data) is expensive.

\section{Model-based Approaches}

Approaches to reinforcement learning are largely categorized into two classes.
\midx<Model-based>{model-based reinforcement learning}[idxpagebf] approaches aim to learn the underlying Markov decision process.
More concretely, they learn models of the dynamics~$p$ and rewards~$r$.
They then use these models to perform planning~(i.e., policy optimization) in the underlying Markov decision process.
In contrast, \midx<model-free>{model-free reinforcement learning} approaches learn the value function directly.
We begin by discussing model-based approaches to the tabular setting.
In \cref{sec:tabular_rl:model_free}, we cover model-free approaches.

\subsection{Learning the Underlying Markov Decision Process}\label{sec:rl:learning_mdp}

Recall that the underlying Markov decision process was specified by its dynamics $p(x' \mid x, a)$ that correspond to the probability of entering state $x' \in \sX$ when playing action $a \in \sA$ from state $x \in \sX$, and its rewards $r(x, a)$ for playing action $a \in \sA$ in state $x \in \sX$.
A natural first idea is to use maximum likelihood estimation to approximate these quantities.\looseness=-1

We can think of each transition $x' \mid x, a$ as sampling from a categorical random variable of which we want to estimate the success probabilities for landing in each of the states.
Therefore, as we have seen in \cref{ex:mle_bern}, the MLE of the dynamics model coincides with the sample mean,\looseness=-1 \begin{align}
\hat{p}(x' \mid x, a) = \frac{N(x' \mid x, a)}{N(a \mid x)}
\end{align} where $N(x' \mid x, a)$ counts the number of transitions from state $x$ to state $x'$ when playing action $a$ and $N(a \mid x)$ counts the number of transitions that start in state $x$ and play action $a$ (regardless of the next state).
Similarly, for the rewards model, we obtain the following maximum likelihood estimate (i.e., sample mean), \begin{align}
  \hat{r}(x, a) = \frac{1}{N(a \mid x)} \sum_{\substack{t = 0 \\ x_t = x \\ a_t = a}}^\infty r_t.
\end{align}
It is immediate that both estimates are unbiased as both correspond to a sample mean.

Still, for the models of our environment to become accurate, our agent needs to visit \emph{each} state-action pair $(x,a)$ numerous times.
Note that our estimators for dynamics and rewards are only well-defined when we visit the corresponding state-action pair at least once.
However, in a stochastic environment, a single visit will likely not result in an accurate model.
We can use Hoeffding's inequality \eqref{eq:hoeffdings_inequality} to gauge how accurate the estimates are after only a limited number of visits.

\section{Balancing Exploration and Exploitation}\label{sec:tabular_rl:exploration_exploitation}

The next natural question is how to use our current model of the environment to pick actions such that exploration and exploitation are traded effectively.
This is what we will consider next.

Given the estimated MDP given by $\hat{p}$ and $\hat{r}$, we can compute the optimal policy using either policy iteration or value iteration.
For example, using value iteration, we can compute the optimal state-action value function $\opt{\fnQ}$ within the \emph{estimated} MDP, and then employ the greedy policy\looseness=-1 \begin{align}
  \pi(x) = \argmax_{a \in \sA} \Q*{x}{a}.
\end{align}
Recall from \cref{eq:bop1} that this corresponds to always picking the best action under the \emph{current} model (that is, $\pi$ is the optimal policy).
But since the model is inaccurate, while potentially quickly generating some reward, we will likely get stuck in a suboptimal state.

\subsection{$\varepsilon$-greedy}\label{sec:tabular_rl:mb:epsilon_greedy}

Consider the other extreme:
If we always pick a random action, we will eventually(!) estimate the dynamics and rewards correctly, yet we will do extremely poorly in terms of maximizing rewards along the way.
To trade exploration and exploitation, a natural idea is to balance these two extremes.

Arguably, the simplest idea is the following: At each time step, throw a biased coin.
If this coin lands heads, we pick an action uniformly at random among all actions.
If the coin lands tails, we pick the best action under our current model.
This algorithm is called \emph{$\varepsilon$-greedy}, where the probability of a coin landing heads at time $t$ is $\varepsilon_t$.

\begin{algorithm}
  \caption{$\varepsilon$-greedy}\pidx{$\varepsilon$-greedy}[idxpagebf]
  \For{$t = 0$ \KwTo $\infty$}{
    sample $u \in \Unif{[0,1]}$\;
    \lIf{$u \leq \varepsilon_t$}{pick action uniformly at random among all actions}
    \lElse{pick best action under the current model}
  }
\end{algorithm}

The $\varepsilon$-greedy algorithm provides a general framework for addressing the exploration-exploitation dilemma.
When the underlying MDP is learned using Monte Carlo estimation as we discussed in \cref{sec:rl:learning_mdp}, the resulting algorithm is known as \midx{Monte Carlo control}.
However, the same framework can also be used in the model-free setting where we pick the best action without estimating the full underlying MDP.
We discuss this approach in greater detail in \cref{sec:tabular_rl:model_free}.

Amazingly, this simple algorithm already works quite well.
Nevertheless, it can clearly be improved.
The key problem of $\varepsilon$-greedy is that it explores the state space in an uninformed manner.
In other words, it explores ignoring all past experience.
It thus does not eliminate clearly suboptimal actions.
This is a problem, especially as we typically have many state-action pairs and recalling that we have to explore each such pair many times to learn an accurate model.

\begin{rmk}{Asymptotic convergence}{glie}
  It can be shown that Monte Carlo control converges to an optimal policy (albeit slowly) almost surely when the learned policy is ``greedy in the limit with infinite exploration''.

  \begin{defn}[Greedy in the limit with infinite exploration, GLIE]
    A sequence of policies $\pi_t$ is said to be \midx{greedy in the limit with infinite exploration} if \begin{enumerate}
      \item all state-action pairs are explored infinitely many times,\safefootnote{That all state-action pairs are chosen is a fundamental requirement. There is no reason why any algorithm would converge to the true value function for \emph{all} states when it only sees some state-action pairs finitely many times, or even not at all.} \begin{align}
        \lim_{t \to \infty} N_t(x, a) = \infty \quad\text{and}
      \end{align}

      \item the policy converges to a greedy policy, \begin{align}
        \lim_{t \to \infty} \pi_t(a \mid x) = \Ind{a = \argmax_{a' \in \sA} \opt{\fnQ}_t(x, a')}
      \end{align}
    \end{enumerate} where we denote by $N_t(x, a)$ the number of transitions from state $x$ playing action $a$ until time $t$, and $\opt{\fnQ}_t$ is the optimal state-action value function in the estimated MDP at time $t$.
  \end{defn}

  Note that $\varepsilon$-greedy is GLIE with probability $1$ if the sequence $(\varepsilon_t)_{t\in\Nat_0}$ satisfies the Robbins-Monro~(RM) conditions~\eqref{eq:rm_conditions}, \begin{align*}
    \varepsilon_t \geq 0 \quad \forall t, \quad \sum_{t=0}^\infty \varepsilon_t = \infty \quad\text{and}\quad \sum_{t=0}^\infty \varepsilon_t^2 < \infty.
  \end{align*}
  The RM-conditions are satisfied, for example, if $\varepsilon_t = \nicefrac{1}{t}$.

  \begin{thm}[Convergence of Monte Carlo control]
    GLIE Monte Carlo control converges to an optimal policy with probability $1$.
  \end{thm}

  Intuitively, the probability of exploration converges to zero, and hence, the policy will ``eventually coincide'' with the greedy policy.
  Moreover, the greedy policy will ``eventually coincide'' with the optimal policy due to an argument akin to the convergence of policy iteration,\safefootnote{see \cref{lem:policy_iteration_mon_impr,thm:policy_iteration_convergence}} and using that each state-action pair is visited infinitely often.
\end{rmk}

\subsection{Softmax Exploration}

An alternative to using $\varepsilon$-greedy for trading between greedy exploitation and uniform exploration is the so-called \midx{softmax exploration} or \midx{Boltzmann exploration}.
Given the agent is in state $x$, we pick action~$a$ with probability, \begin{align}
  \pi_\lambda(a \mid x) \propto \exp\parentheses*{\frac{1}{\lambda} \Q*{x}{a}}, \label{eq:softmax_exploration}
\end{align} which is the Gibbs distribution with temperature parameter $\lambda > 0$.
Observe that for $\lambda \to 0$, softmax exploration corresponds to greedily maximizing the Q-function (i.e., greedy exploitation), whereas for~${\lambda\to\infty}$, softmax exploration explores uniformly at random.
This can outperform $\varepsilon$-greedy as the exploration is directed towards actions with larger estimated value.

\subsection{Optimism}

Recall from our discussion of multi-armed bandits in \cref{sec:bayesian_optimization:online_learning:mab} that a key principle in effectively trading exploration and exploitation is \midx{optimism in the face of uncertainty}.
Let us apply this principle to the reinforcement learning setting.
The key idea is to assume that the dynamics and rewards model ``work in our favor'' until we have learned ``good estimates'' of the true dynamics and rewards.

\begin{marginfigure}
  \incfig{rmax}
  \caption{Illustration of the fairy-tale state of $R_\mathrm{max}$.
  If in doubt, the agent believes actions from the state $x$ to lead to the fairy-tale state $\xs$ with maximal rewards.
  This encourages the exploration of unknown states.}
\end{marginfigure}

More formally, if $r(x, a)$ is unknown, we set $\hat{r}(x, a) = R_\mathrm{max}$, where $R_\mathrm{max}$ is the maximum reward our agent can attain during a single transition.
Similarly, if $p(x' \mid x, a)$ is unknown, we set $\hat{p}(\xs \mid x, a) = 1$, where $\xs$ is a ``fairy-tale state''.
The fairy-tale state corresponds to everything our agent could wish for, that is, \begin{align}
  \hat{p}(\xs \mid \xs, a) &= 1 \quad &&\forall a \in \sA, \\
  \hat{r}(\xs, a)          &= R_\mathrm{max} \quad &&\forall a \in \sA.
\end{align}
In practice, the decision of when to assume that the learned dynamics and reward models are ``good enough'' has to be tuned.

In using these optimistic estimates of $p$ and $r$, we obtain an optimistic underlying Markov decision process that exhibits a bias towards exploration.
In particular, the rewards attained in this MDP, are an upper bound of the true reward.
The resulting algorithm is known as the \emph{$R_\mathrm{max}$ algorithm}.

\begin{algorithm}
  \caption{$R_\mathrm{max}$ algorithm}\pidx{$R_\mathrm{max}$ algorithm}
  add the fairy-tale state $\xs$ to the Markov decision process\;
  set $\hat{r}(x, a) = R_\mathrm{max}$ for all $x \in \sX$ and $a \in \sA$\;
  set $\hat{p}(\xs \mid x, a) = 1$ for all $x \in \sX$ and $a \in \sA$\;
  compute the optimal policy $\hat{\pi}$ for $\hat{r}$ and $\hat{p}$\;
  \For{$t = 0$ \KwTo $\infty$}{
    execute policy $\hat{\pi}$ (for some number of steps)\;
    for each visited state-action pair $(x, a)$, update $\hat{r}(x, a)$\;
    estimate transition probabilities $\hat{p}(x' \mid x, a)$\;
    after observing ``enough'' transitions and rewards, recompute the optimal policy $\hat{\pi}$ according the current model $\hat{p}$ and $\hat{r}$.
  }
\end{algorithm}

How many transitions are ``enough''?
We can use Hoeffding's inequality \eqref{eq:hoeffdings_inequality} to get a rough idea!
The key here, is our observation from \cref{eq:rl_cond_indep} that the transitions and rewards are conditionally independent given the state-action pairs since, as we have discussed in \cref{sec:approximate_inference:mcmc:ergodic_theorem} on the ergodic theorem, Hoeffding's inequality does not hold for dependent samples.
In this case, Hoeffding's inequality tells us that for the absolute approximation error to be below $\epsilon$ with probability at least $1-\delta$, we need \begin{align}
  N(a \mid x) \geq \frac{R_\mathrm{max}^2}{2 \epsilon^2} \log \frac{2}{\delta}. \margintag{see \eqref{eq:eq:hoeffdings_inequality_sample_size}}
\end{align}

\begin{lem}[Exploration and exploitation of $R_\mathrm{max}$]
  Every $T$ time steps, with high probability, $R_\mathrm{max}$ either \begin{itemize}
    \item obtains near-optimal reward; or
    \item visits at least one unknown state-action pair.\footnote{Note that in the tabular setting, there are ``only'' polynomially many state-action pairs.}
  \end{itemize}
  Here, $T$ depends on the mixing time of the Markov chain induced by the optimal policy.
\end{lem}

\begin{thm}[Convergence of $R_\mathrm{max}$, \cite{brafman2002r}]
  With probability at least $1-\delta$, $R_\mathrm{max}$ reaches an $\epsilon$-optimal policy in a number of steps that is polynomial in $\card{\sX}$, $\card{\sA}$, $T$, $\nicefrac{1}{\epsilon}$, $\nicefrac{1}{\delta}$, and $R_\mathrm{max}$.
\end{thm}

\subsection{Challenges of Model-based Approaches}

We have seen that the $R_\mathrm{max}$ algorithm performs remarkably well in the tabular setting.
However, there are important computational limitations to the model-based approaches that we discussed so far.

First, observe that the (tabular) model-based approach requires us to store $\hat{p}(x' \mid x, a)$ and $\hat{r}(x, a)$ in a table.
This table already has $\BigO{n^2 m}$ entries.
Even though polynomial in the size of the state and action spaces, this quickly becomes unmanageable.

Second, the model-based approach requires us to ``solve'' the learned Markov decision processes to obtain the optimal policy (using policy or value iteration).
As we continue to learn over time, we need to find the optimal policy many times.
$R_\mathrm{max}$ recomputes the policy after each state-action pair is observed sufficiently often, so $\BigO{n m}$ times.

\section{Model-free Approaches}\label{sec:tabular_rl:model_free}

In the previous section, we have seen that learning and remembering the model as well as planning within the estimated model can potentially be quite expensive in the model-based approach.
We therefore turn to model-free methods that estimate the value function directly.
Thus, they require neither remembering the full model nor planning~(i.e., policy optimization) in the underlying Markov decision process.
We will, however, return to model-based methods in \cref{sec:mbarl} to see that promise lies in combining methods from model-based reinforcement learning with methods from model-free reinforcement learning.

A significant benefit to model-based reinforcement learning is that it is inherently off-policy.
That is, any trajectory regardless of the policy used to obtain it can be used to improve the model of the underlying Markov decision process.
In the model-free setting, this not necessarily true.
By default, estimating the value function according to the data from a trajectory, will yield an estimate of the value function corresponding to the policy that was used to sample the data.

We will start by discussing on-policy methods and later see how the value function can be estimated off-policy.

\subsection{On-policy Value Estimation}\label{sec:tabular_rl:model_free:on_policy_value_estimation}

Let us suppose, our agent follows a fixed policy $\pi$.
Then, the corresponding value function $\fnv[\pi]$ is given as \begin{align}
  \v[\pi]{x} &= r(x, \pi(x)) + \gamma \sum_{x' \in \sX} p(x' \mid x, \pi(x)) \cdot \v[\pi]{x'} \margintag{using the definition of the value function \eqref{eq:value_function}} \nonumber \\
  &= \E[R_0, X_1]{R_0 + \gamma \v[\pi]{X_1} \mid X_0 = x, A_0 = \pi(x)} \label{eq:td_learning_mc} \margintag{interpreting the above expression as an expectation over the random quantities $R_0$ and $X_1$}
  \intertext{Our first instinct might be to use a Monte Carlo estimate of this expectation. Due to the conditional independence of the transitions \eqref{eq:rl_cond_indep}, Monte Carlo approximation does yield an unbiased estimate,}
  &\approx r + \gamma \v[\pi]{x'}, \label{eq:bootstrapping_estimate}
\end{align} where the agent observed the transition $(x, a, r, x')$.
Note that to estimate this expectation we use a single(!) sample,\footnote{The idea is that we will use this approximation repeatedly as our agent collects new data to achieve the same effect as initially averaging over multiple samples.} unlike our previous applications of Monte Carlo sampling where we usually averaged over $m$ samples.
However, there is one significant problem in this approximation.
Our approximation of $\fnv[\pi]$ does in turn depend on the (unknown) true value of $\fnv[\pi]$!

The key idea is to use a bootstrapping estimate of the value function instead.
That is, in place of the true value function $\fnv[\pi]$, we will use a ``running estimate'' $\fnV[\pi]$.
In other words, whenever observing a new transition, we use our previous best estimate of $\fnv[\pi]$ to obtain a new estimate $\fnV[\pi]$. We already encountered bootstrapping briefly in \cref{sec:bdl:approximate_inference:probabilistic_ensembles} in the context of probabilistic ensembles in Bayesian deep learning.
More generally, \midx{bootstrapping}[idxpagebf] refers to approximating a true quantity (e.g., $\fnv[\pi]$) by using an empirical quantity (e.g., $\fnV[\pi]$), which itself is constructed using samples from the true quantity that is to be approximated.\looseness=-1

Due to its use in estimating the value function, bootstrapping is a core concept to model-free reinforcement learning.
Crucially, using a bootstrapping estimate generally results in biased estimates of the value function. Moreover, due to relying on a single sample, the estimates from \cref{eq:bootstrapping_estimate} tend to have very large variance.

The variance of the estimate is typically reduced by mixing new estimates of the value function with previous estimates using a learning rate $\alpha_t$.
This yields the \emph{temporal-difference learning} algorithm.

\begin{algorithm}
  \caption{Temporal-difference (TD) learning}\pidx{temporal-difference learning}
  initialize $\fnV[\pi]$ arbitrarily (e.g., as $\vzero$)\;
  \For{$t = 0$ \KwTo $\infty$}{
    follow policy $\pi$ to obtain the transition $(x, a, r, x')$\;
    $\V[\pi]{x} \gets (1-\alpha_t)\V[\pi]{x} + \alpha_t(r + \gamma \V[\pi]{x'})$ \algeq{eq:td_learning}
  }
\end{algorithm}

The update rule is sometimes written equivalently as \begin{align}
  \V[\pi]{x} \gets \V[\pi]{x} + \alpha_t(r + \gamma \V[\pi]{x'} - \V[\pi]{x}).
\end{align}
Thus, the update to $\V[\pi]{x}$ is proportional to the learning rate and the difference between the previous estimate and the renewed estimate using the new observation.

\begin{thm}[Convergence of TD-learning, \cite{jaakkola1993convergence}]
  If $(\alpha_t)_{t\in\Nat_0}$ satisfies the RM-conditions \eqref{eq:rm_conditions} and all state-action pairs are chosen infinitely often, then $\fnV[\pi]$ converges to $\fnv[\pi]$ with probability $1$.
\end{thm}

Importantly, note that due to the Monte Carlo approximation of \cref{eq:td_learning_mc} with respect to transitions attained by following policy $\pi$, TD-learning is fundamentally on-policy.
That is, for the estimates $\fnV[\pi]$ to converge to the true value function $\fnv[\pi]$, the transitions that are used for the estimation must follow policy $\pi$.

\subsection{SARSA: On-policy Control}

TD-learning merely estimates the value function of a fixed policy $\pi$.
To find the optimal policy $\pis$, we can use an analogue of policy iteration~(see \cref{alg:policy_iteration}).
Here, it is more convenient to use an estimate of the state-action value function $\fnq[\pi]$ which can be obtained analogously to the bootstrapping estimate of $\fnv[\pi]$ \eqref{eq:bootstrapping_estimate}, \begin{align}
  \q[\pi]{x}{a} &= r(x, a) + \gamma \sum_{x' \in \sX} p(x' \mid x, a) \sum_{a' \in \sA} \pi(a' \mid x') \q[\pi]{x'}{a'} \margintag{using Bellman's expectation equation \eqref{eq:q_function3}} \nonumber \\
  &= \E[R_0, X_1, A_1]{R_0 + \gamma \q[\pi]{X_1}{A_1} \mid X_0 = x, A_0 = a} \margintag{interpreting the above expression as an expectation over $R_0, X_1$ and $A_1$} \\[5pt]
  &\approx r + \gamma \q[\pi]{x'}{a'}, \label{eq:sarsa_bootstrapping} \margintag{Monte Carlo approximation with a single sample}
\end{align} where the agent observed transitions $(x, a, r, x')$ and $(x', a', r', x'')$.

The update rule from TD-learning is therefore adapted to\safefootnote{Note that for deterministic policies $\pi$, $\Q[\pi]{x'}{a'} = \Q[\pi]{x'}{\pi(x')} = \V[\pi]{x'}$ if the transitions are obtained by following policy $\pi$.} \begin{align}
  \Q[\pi]{x}{a} \gets (1-\alpha_t)\Q[\pi]{x}{a} + \alpha_t(r + \gamma \Q[\pi]{x'}{a'}). \label{eq:sarsa_update}
\end{align}

This algorithm is known as \midx{SARSA} (short for \emph{state-action-reward-state-action}).
Similar convergence guarantees to those of TD-learning can also be derived for SARSA.

\begin{thm}[Convergence of SARSA, \cite{singh2000convergence}]
  If $(\alpha_t)_{t\in\Nat_0}$ satisfies the RM-conditions \eqref{eq:rm_conditions} and all state-action pairs are chosen infinitely often, then $\fnQ[\pi]$ converges to $\fnq[\pi]$ with probability $1$.
\end{thm}

The policy iteration scheme to identify the optimal policy can be outlined as follows:
In each iteration $t$, we estimate the value function~$\fnq[\pi_t]$ of policy~$\pi_t$ with the estimate $\fnQ[\pi_t]$ obtained from SARSA.
We then choose the greedy policy with respect to $\fnQ[\pi_t]$ as the next policy $\pi_{t+1}$.
However, due to the on-policy nature of SARSA, we cannot reuse any data between the iterations.
Moreover, it turns out that in practice, when using only finitely many samples, this form of greedily optimizing Markov decision processes does not explore enough.
At least partially, this can be compensated for by injecting noise when choosing the next action, e.g., by following an $\varepsilon$-greedy policy or using softmax exploration.\looseness=-1

\subsection{Off-policy Value Estimation}\label{sec:tabular_rl:model_free:off_policy_value_estimation}

Consider the following slight adaptation of the derivation of SARSA~\eqref{eq:sarsa_bootstrapping},\looseness=-1 \begin{align}
  \q[\pi]{x}{a} &= r(x, a) + \gamma \sum_{x' \in \sX} p(x' \mid x, a) \sum_{a' \in \sA} \pi(a' \mid x') \q[\pi]{x'}{a'} \margintag{using Bellman's expectation equation \eqref{eq:q_function3}} \nonumber \\
  &= \E[R_0, X_1]{R_0 + \gamma \sum_{a' \in \sA} \pi(a' \mid X_1) \q[\pi]{X_1}{a'}}[X_0 = x, A_0 = a] \margintag{interpreting the above expression as an expectation over $R_0$ and $X_1$} \\
  &\approx r + \gamma \sum_{a' \in \sA} \pi(a' \mid x') \q[\pi]{x'}{a'}, \margintag{Monte Carlo approximation with a single sample}
\end{align} where the agent observed the transition $(x, a, r, x')$.
This yields the update rule, \begin{align}
  \Q[\pi]{x}{a} \gets (1-\alpha_t)\Q[\pi]{x}{a} + \alpha_t\parentheses*{r + \gamma \sum_{a' \in \sA} \pi(a' \mid x') \Q[\pi]{x'}{a'}}.
\end{align}
This adapted update rule \emph{explicitly} chooses the subsequent action $a'$ according to policy $\pi$ whereas SARSA absorbs this choice into the Monte Carlo approximation.
The algorithm has analogous convergence guarantees to those of SARSA.

Crucially, this algorithm is off-policy.
That is, we can use transitions that were obtained according to \emph{any} policy to estimate the value of a fixed policy $\pi$, which we may have never used!
Perhaps this seems contradictory at first, but it is not.
As noted, the key difference to the on-policy TD-learning and SARSA is that our estimate of the Q-function explicitly keeps track of the next-performed action.
It does so for any action in any state.
Moreover, note that the transitions that are due to the dynamics model and rewards are unaffected by the used policy.
They merely depend on the originating state-action pair.
We can therefore use the instances where other policies played action $\pi(x)$ in state $x$ to estimate the performance of $\pi$.

\subsection{Q-learning: Off-policy Control}\label{sec:tabular_rl:model_free:q_learning}

It turns out that there is a way to estimate the value function of the optimal policy directly.
Recall from Bellman's theorem \eqref{eq:bop1} that the optimal policy $\pis$ can be characterized in terms of the optimal state-action value function $\fnq[\star]$, \begin{align*}
  \pis(x) = \argmax_{a \in \sA} \q*{x}{a}.
\end{align*} $\pis$ corresponds to greedily maximizing the value function.

Analogously to our derivation of SARSA \eqref{eq:sarsa_bootstrapping}, only using Bellman's theorem \eqref{eq:bop2} in place of Bellman's expectation equation \eqref{eq:q_function3}, we obtain,\looseness=-1 \begin{align}
  \q*{x}{a} &= r(x, a) + \gamma \sum_{x' \in \sX} p(x' \mid x, a) \max_{a' \in \sA} \q*{x'}{a'} \margintag{using that the Q-function is a fixed-point of the Bellman update, see Bellman's theorem \eqref{eq:bop2}} \nonumber \\
  &= \E[R_0, X_1]{R_0 + \gamma \max_{a' \in \sA} \q*{X_1}{a'}}[X_0 = x, A_0 = a] \label{eq:q_learning_mc} \margintag{interpreting the above expression as an expectation over $R_0$ and $X_1$} \\
  &\approx r + \gamma \max_{a' \in \sA} \q*{x'}{a'}, \label{eq:q_learning_bootstrapping} \margintag{Monte Carlo approximation with a single sample}
\end{align} where the agent observed the transition $(x, a, r, x')$.
Using a bootstrapping estimate $\fnQ[\star]$ for $\fnq[\star]$, we obtain a structurally similar algorithm to TD-learning and SARSA --- only for estimating the optimal Q-function directly!
This algorithm is known as \emph{Q-learning}.
Whereas we have seen that the optimal policy can be found using SARSA in a policy-iteration-like scheme, Q-learning is conceptually similar to value iteration.

\begin{algorithm}[H]
  \caption{Q-learning}\pidx{Q-learning}
  initialize $\Q*{x}{a}$ arbitrarily (e.g., as $\vzero$)\;
  \For{$t = 0$ \KwTo $\infty$}{
    observe the transition $(x, a, r, x')$\;
    $\Q*{x}{a} \gets (1-\alpha_t)\Q*{x}{a} + \alpha_t(r + \gamma \max_{a' \in \sA} \Q*{x'}{a'})$ \algeq{eq:q_learning}
  }
\end{algorithm}

Similarly to TD-learning, the update rule can also be expressed as \begin{align}
  \Q*{x}{a} \gets \Q*{x}{a} + \alpha_t\parentheses*{r + \gamma \max_{a' \in \sA} \Q*{x'}{a'} - \Q*{x}{a}}.
\end{align}
Crucially, the Monte Carlo approximation of \cref{eq:q_learning_mc} does not depend on the policy.
Thus, Q-learning is an off-policy method.

\begin{thm}[Convergence of Q-learning, \cite{jaakkola1993convergence}]\label{thm:q_learning_convergence}
  If $(\alpha_t)_{t\in\Nat_0}$ satisfies the RM-conditions \eqref{eq:rm_conditions} and all state-action pairs are chosen infinitely often (that is, the sequence of policies used to obtain the transitions is GLIE), then $\fnQ[\star]$ converges to $\fnq[\star]$ with probability $1$.
\end{thm}

It can be shown that with probability at least $1-\delta$, Q-learning converges to an $\epsilon$-optimal policy in a number of steps that is polynomial in $\log |X|$, $\log |A|$, $\nicefrac{1}{\epsilon}$ and $\log \nicefrac{1}{\delta}$ \citep{even2003learning}.

\subsection{Optimistic Q-learning}

The next natural question is how to effectively trade exploration and exploitation to both visit all state-action pairs many times, but also attain a high reward.

However, as we have seen in \cref{sec:tabular_rl:exploration_exploitation}, random ``uninformed'' exploration like $\varepsilon$-greedy and softmax exploration explores the state space very slowly.
We therefore return to the principle of \midx{optimism in the face of uncertainty}, which already led us to the $R_\mathrm{max}$ algorithm in the model-based setting.
We will now additionally assume that the rewards are non-negative, that is, $0 \leq r(x, a) \leq R_\mathrm{max} \; (\forall x \in \sX, a \in \sA)$.
It turns out that a similar algorithm to $R_\mathrm{max}$ also exists for (model-free) Q-learning: it is called \emph{optimistic Q-learning} and shown in \cref{alg:opt_q_learning}.

\begin{algorithm}[H]
  \caption{Optimistic Q-learning}\label{alg:opt_q_learning}\pidx{optimistic Q-learning}
  initialize $\Q*{x}{a} = V_\mathrm{max} \prod_{t=1}^{T_\mathrm{init}} (1 - \alpha_t)^{-1}$\;
  \For{$t = 0$ \KwTo $\infty$}{
    pick action $a_t = \argmax_{a \in \sA} \Q*{x}{a}$ and observe the transition $(x, a_t, r, x')$\;
    $\Q*{x}{a_t} \gets (1-\alpha_t)\Q*{x}{a_t} + \alpha_t(r + \gamma \max_{a' \in \sA} \Q*{x'}{a'})$ \algeq{eq:optimistic_q_learning}
  }
\end{algorithm}

Here, \begin{align*}
  V_\mathrm{max} \defeq \frac{R_\mathrm{max}}{1-\gamma} \geq \max_{x \in \sX, a \in \sA} \q*{x}{a},
\end{align*} is an upper bound on the discounted return and $T_\mathrm{init}$ is some initialization time.
Intuitively, the initialization of $\fnQ[\star]$ corresponds to the best-case long-term reward, assuming that all individual rewards are upper bounded by $R_\mathrm{max}$. This is shown by the following lemma.

\begin{lem}\label{lem:optimistic_q_learning}
  Denote by $\fnQ[\star][t]$, the approximation of $\fnq[\star]$ attained in the $t$-th iteration of optimistic Q-learning.
  Then, for any state-action pair $(x,a)$ and iteration $t$ such that $N(a \mid x) \leq T_\mathrm{init}$,\footnote{$N(a \mid x)$ is the number of times action $a$ is performed in state $x$.} \begin{align}
    \Q*{x}{a}[t] \geq V_\mathrm{max} \geq \q*{x}{a}.
  \end{align}
\end{lem}
\begin{proof}
  We write $\beta_\tau \defeq \prod_{i=1}^\tau (1-\alpha_i)$ and $\eta_{i,\tau} \defeq \alpha_i \prod_{j=i+1}^\tau (1-\alpha_j)$.
  Using the update rule of optimistic Q-learning \eqref{eq:optimistic_q_learning}, we have \begin{align}
    \Q*{x}{a}[t] = \beta_{N(a \mid x)} \Q*{x}{a}[0] + \sum_{i=1}^{N(a \mid x)} \eta_{i,N(a \mid x)} (r + \gamma \max_{a_i \in \sA} \Q*{x_i}{a_i}[t_i]) \label{eq:opt_q_learning_char2}
  \end{align} where $x_i$ is the next state arrived at time $t_i$ when action $a$ is performed the $i$-th time in state $x$.

  Using the assumption that the rewards are non-negative, from \cref{eq:opt_q_learning_char2} and $\Q*{x}{a}[0] = \nicefrac{V_\mathrm{max}}{\beta_{T_\mathrm{init}}}$, we immediately have \begin{align*}
    \Q*{x}{a}[t] &\geq \frac{\beta_{N(a \mid x)}}{\beta_{T_\mathrm{init}}} V_\mathrm{max} \\
    &\geq V_\mathrm{max}. \margintag{using $N(a \mid x) \leq T_\mathrm{init}$} \qedhere
  \end{align*}
\end{proof}

Now, if $T_\mathrm{init}$ is chosen large enough, it can be shown that optimistic Q-learning converges quickly to an optimal policy.

\begin{thm}[Convergence of optimistic Q-learning, \cite{even2001convergence}]\label{thm:optimistic_q_learning}
  With probability at least $1-\delta$, optimistic Q-learning obtains an $\epsilon$-optimal policy after a number of steps that is polynomial in $\card{\sX}$, $\card{\sA}$, $\nicefrac{1}{\epsilon}$, $\log \nicefrac{1}{\delta}$, and $R_\mathrm{max}$ where the initialization time $T_\mathrm{init}$ is upper bounded by a polynomial in the same coefficients.
\end{thm}

Note that for Q-learning, we still need to store $\Q*{x}{a}$ for any state-action pair in memory.
Thus, Q-learning requires $\BigO{n m}$ memory. During each transition, we need to compute \begin{align*}
  \max_{a \in \sA} \Q*{x'}{a'}
\end{align*} once.
If we run Q-learning for $T$ iterations, this yields a time complexity of $\BigO{T m}$.
Crucially, for sparse Markov decision processes where, in most states, only few actions are permitted, each iteration of Q-learning can be performed in (virtually) constant time.
This is a big improvement of the quadratic (in the number of states) performance of the model-based $R_\mathrm{max}$ algorithm.

\section*{Discussion}

We have seen that both the model-based $R_\mathrm{max}$ algorithm and the model-free Q-learning take time polynomial in the number of states $\card{\sX}$ and the number of actions $\card{\sA}$ to converge.
While this is acceptable in small grid worlds, this is completely unacceptable for large state and action spaces.\looseness=-1

Often, domains are continuous, for example when modeling beliefs about states in a partially observable environment.
Also, in many structured domains (e.g., chess or multiagent planning), the size of the state and action space is exponential in the size of the input.
In the final two chapters, we will therefore explore how model-free and model-based methods can be used (approximately) in such large domains.

\excheading

\begin{nexercise}{Q-learning}{q_learning}
  Assume the following grid world, where from state $A$ the agent can go to the right and down, and from state $B$ to the left and down. From states $G_1$ and $G_2$ the only action is to exit.
  The agent receives a reward ($+10$ or $+1$) only when exiting.

  \begin{center}
    \begin{tabular}{|c|c|}
      \multicolumn{2}{c}{Rewards} \\
      \hline
      $0$ & $0$ \\
      \hline
      $+10$ & $+1$ \\
      \hline
    \end{tabular}
    \begin{tabular}{|c|c|}
      \multicolumn{2}{c}{States} \\
      \hline
      $A$ & $B$ \\
      \hline
      $G_1$ & $G_2$ \\
      \hline
    \end{tabular}
  \end{center}

  We assume the discount factor $\gamma = 1$ and that all actions are deterministic.\looseness=-1

  \begin{enumerate}
    \item We observe the following two episodes:

    \vspace{5pt}
    \begin{center}
      \begin{tabular}{|cccc|}
        \multicolumn{4}{c}{Episode 1} \\
        \hline
        $x$ & $a$ & $x'$ & $r$ \\
        \hline
        $A$ & $\downarrow$ & $G_1$ & $0$ \\
        $G_1$ & exit & & $10$ \\
        \hline
      \end{tabular}
      \begin{tabular}{|cccc|}
        \multicolumn{4}{c}{Episode 2} \\
        \hline
        $x$ & $a$ & $x'$ & $r$ \\
        \hline
        $B$ & $\leftarrow$ & $A$ & $0$ \\
        $A$ & $\downarrow$ & $G_1$ & $0$ \\
        $G_1$ & exit & & $10$ \\
        \hline
      \end{tabular}
      \vspace{5pt}
    \end{center}
    \vspace{5pt}

    Assume $\alpha = 0.3$, and that Q-values of all non-terminal states are initialized to $0.5$.
    What are the Q-values \begin{align*}
      \Q*{A}{\downarrow}, \quad \Q*{G_1}{\text{exit}}, \quad \Q*{G_2}{\text{exit}}
    \end{align*} learned by executing Q-learning with the above episodes?

    \item Will Q-learning converge to $\fnq[\star]$ for all state-action pairs $(x, a)$ if we repeat episode 1 and episode 2 infinitely often?
    If not, design a sequence of episodes that leads to convergence.

    \item How does the choice of initial Q-values influence the convergence of the Q-learning algorithm when episodes are obtained off-policy?

    \item Determine $\fnv[\star]$ for all states.
  \end{enumerate}
\end{nexercise}

  \chapter{Model-free Reinforcement Learning}\label{sec:mfarl}

In the previous chapter, we have seen methods for tabular settings.
Our goal now is to extend the model-free methods like TD-learning and Q-learning to large state-action spaces $\spX$ and $\spA$.
We have seen that a crucial bottleneck of these methods is the parameterization of the value function.
If we want to store the value function in a table, we need at least $\BigO{\card{\spX}}$ space.
If we learn the Q function, we even need $\BigO{\card{\spX} \cdot \card{\spA}}$ space.
Also, for large state-action spaces, the time required to compute the value function for every state-action pair exactly will grow polynomially in the size of the state-action space. Hence, a natural idea is to learn \emph{approximations} of these functions with a low-dimensional parameterization.
Such approximations were the focus of the first few chapters and are, in fact, the key idea behind methods for reinforcement learning in large state-action spaces.

\section{Tabular Reinforcement Learning as Optimization}\label{sec:mfarl:tabular_rl_as_optimization}

To begin with, let us reinterpret the model-free methods from the previous section, TD-learning and Q-learning, as solving an optimization problem, where each iteration corresponds to a single gradient update.
We will focus on TD-learning here, but the same interpretation applies to Q-learning.
Recall the update rule of TD-learning \eqref{eq:td_learning}, \begin{align*}
  \V[\pi]{x} \gets (1-\alpha_t) \V[\pi]{x} + \alpha_t (r + \gamma \V[\pi]{x'}).
\end{align*}
Note that this looks just like the update rule of an optimization algorithm!
We can parameterize our estimates $\fnV[\pi]$ with parameters $\vtheta$ that are updated according to the gradient of some loss function, assuming fixed bootstrapping estimates.
In particular, in the tabular setting (i.e., over a finite domain), we can parameterize the value function exactly by learning a separate parameter for each state, \begin{align}
  \vtheta \defeq [\vtheta(1), \dots, \vtheta(n)], \quad \V[\pi]{x; \vtheta} \defeq \vtheta(x).
\end{align}

To re-derive the above update rule as a gradient update, let us consider the following loss function, \begin{align}
  \mean{\ell}(\vtheta; x, r) &\defeq \frac{1}{2}\parentheses*{\v[\pi]{x} - \vtheta(x)}^2 \\
  &= \frac{1}{2}\parentheses*{r + \gamma \E[x' \mid x, \pi(x)]{\v[\pi]{x'}} - \vtheta(x)}^2 \margintag{using Bellman's expectation equation \eqref{eq:bellman_expectation_equation_short}}
\end{align}
Note that this loss corresponds to a standard squared loss of the difference between the parameter $\vtheta(x)$ and the label $\v[\pi]{x}$ we want to learn.\looseness=-1

We can now find the gradient of this loss. Elementary calculations yield,\looseness=-1 \begin{align}
  \grad_{\vtheta(x)} \mean{\ell}(\vtheta; x, r) = \vtheta(x) - \parentheses*{r + \gamma \E[x' \mid x, \pi(x)]{\v[\pi]{x'}}}. \label{eq:l_2_bar_gradient}
\end{align}
Now, we cannot compute this derivative because we cannot compute the expectation.
Firstly, the expectation is over the true value function which is unknown to us.
Secondly, the expectation is over the transition model which we are trying to avoid in model-free methods.

To resolve the first issue, analogously to TD-learning, instead of learning the true value function $\fnv[\pi]$ which is unknown, we learn the bootstrapping estimate $\fnV[\pi]$.
Recall that the core principle behind bootstrapping as discussed in \cref{sec:tabular_rl:model_free:on_policy_value_estimation} is that this bootstrapping estimate~$\fnV[\pi]$ is \emph{treated} as if it were independent of the current estimate of the value function $\vtheta$.
To emphasize this, we write $\V[\pi]{x; \old{\vtheta}} \approx \v[\pi]{x}$ where $\old{\vtheta} = \vtheta$ but $\old{\vtheta}$ is treated as a constant with respect to $\vtheta$.\footnote{That is, the bootstrapping estimate $\V[\pi]{x; \old{\vtheta}}$ is assumed to be constant with respect to $\vtheta(x)$ in the same way that $\v[\pi]{x}$ is constant with respect to $\vtheta(x)$.

If we were not using the bootstrapping estimate, the following derivation of the gradient of the loss would not be this simple.}

To resolve the second issue, analogously to the introduction of TD-learning in the previous chapter, we will use a Monte Carlo estimate using a single sample.
Recall that this is only possible because the transitions are conditionally independent given the state-action pair.

\begin{rmk}{Sample (in)efficiency of model-free methods}{}
  Taking these two shortcuts are two of the main reasons why model-free methods such as TD-learning and Q-learning are usually \emph{sample inefficient}.
  This is because using a bootstrapping estimate leads to ``(initially) incorrect'' and ``unstable'' targets of the optimization problem,\safefootnote{We explore this in some more capacity in \cref{sec:mfarl:value_function_approximation:heuristics} where we cover heuristic approaches to alleviate this problem to some degree.} and Monte Carlo estimation with a single sample leads to a large variance.
  Recall that the theoretical guarantees for model-free methods in the tabular setting therefore required that all state-action pairs are visited infinitely often.
\end{rmk}

Using the aforementioned shortcuts, let us define the loss $\ell$ after observing the single transition $(x, a, r, x')$, \begin{align}
  \ell(\vtheta; x, r, x') \defeq \frac{1}{2}\parentheses*{r + \gamma \old{\vtheta}(x') - \vtheta(x)}^2.
\end{align}
We define the gradient of this loss with respect to $\vtheta(x)$ as \begin{align}
  \delta_\mathrm{TD} &\defeq \grad_{\vtheta(x)} \ell(\vtheta; x, r, x') \nonumber \\
  &= \vtheta(x) - \parentheses*{r + \gamma \old{\vtheta}(x')}. \label{eq:td_error}
\end{align}
This error term is also called \midx<temporal-difference (TD) error>{temporal-difference error}.
The temporal difference error compares the previous estimate of the value function to the bootstrapping estimate of the value function.
We know from the law of large numbers \eqref{eq:slln} that Monte Carlo averages are unbiased.\footnote{Crucially, the samples are unbiased with respect to the approximate label in terms of the bootstrapping estimate only.
Due to bootstrapping the value function, the estimates are not unbiased with respect to the true value function.
Moreover, the variance of each individual estimation of the gradient is large, as we only consider a single transition.}
We therefore have, \begin{align}
  \E[x' \mid x, \pi(x)]{\delta_\mathrm{TD}} \approx \grad_{\vtheta(x)} \mean{\ell}(\vtheta; x, r). \label{eq:td_error_approx}
\end{align}

Naturally, we can use these unbiased gradient estimates with respect to the loss $\mean{\ell}$ to perform stochastic gradient descent.
This yields the update rule, \begin{align}
  \V[\pi]{x; \vtheta} = \vtheta(x) &\gets \vtheta(x) - \alpha_t \delta_\mathrm{TD} \label{eq:td_learning_gradient_update} \margintag{using stochastic gradient descent with learning rate $\alpha_t$, see \cref{alg:sgd}} \\[5pt]
  &= (1 - \alpha_t) \vtheta(x) + \alpha_t \parentheses*{r + \gamma \old{\vtheta}(x')} \margintag{using the definition of the temporal difference error \eqref{eq:td_error}} \nonumber \\
  &= (1-\alpha_t) \V[\pi]{x; \vtheta} + \alpha_t \parentheses*{r + \gamma \V[\pi]{x'; \old{\vtheta}}}. \margintag{substituting $\V[\pi]{x; \vtheta}$ for $\vtheta(x)$} \label{eq:td_learning_as_optimization}
\end{align}
Observe that this gradient update coincides with the update rule of TD-learning \eqref{eq:td_learning}.
Therefore, TD-learning is essentially performing stochastic gradient descent using the TD-error as an unbiased gradient estimate.\footnote{An alternative interpretation is that TD-learning performs gradient descent with respect to the loss $\ell$.}
Crucially, TD-learning performs stochastic gradient descent with respect to the bootstrapping estimate of the value function $\fnV[\pi]$ and not the true value function $\fnv[\pi]$!
Stochastic gradient descent with a bootstrapping estimate is also called \midx{stochastic semi-gradient descent}.
Importantly, the optimization target $r + \gamma \V[\pi]{x'; \old{\vtheta}}$ from the loss $\ell$ is now \emph{moving} between iterations which introduces some practical challenges we will discuss in \cref{sec:mfarl:value_function_approximation:heuristics}.
We have seen in the previous chapter that using a bootstrapping estimate still guarantees (asymptotic) convergence to the true value function.

\section{Value Function Approximation}\label{sec:mfarl:value_function_approximation}

To scale to large state spaces, it is natural to approximate the value function using a parameterized model, $\V{\vx; \vtheta}$ or $\Q{\vx}{\va; \vtheta}$.
You may think of this as a regression problem where we map state(-action) pairs to a real number.
Recall from the previous section that this is a strict generalization of the tabular setting, as we could use a separate parameter to learn the value function for each individual state-action pair.
Our goal for large state-action spaces is to exploit the smoothness properties\footnote{That is, the value function takes similar values in ``similar'' states.} of the value function to condense the representation.

A straightforward approach is to use a linear function approximation with the hand-designed feature map $\vphi$, \begin{align}
  \Q{\vx}{\va; \vtheta} \defeq \transpose{\vtheta} \vphi(\vx, \va). \label{eq:q_linear_approx}
\end{align}
A common alternative is to use a deep neural network to learn these features instead.
Doing so is also known as \midx{deep reinforcement learning}.\footnote{Note that often non-Bayesian deep learning (i.e., point estimates of the weights of a neural network) is applied. In the final chapter, \cref{sec:mbarl}, we will explore the benefits of using Bayesian deep learning.}

We will now apply the derivation from the previous section directly to Q-learning.
For Q-learning, after observing the transition $(\vx, \va, r, \vxp)$, the loss function is given as \begin{align}
  \ell(\vtheta; \vx, \va, r, \vxp) \defeq \frac{1}{2}\parentheses*{r + \gamma \max_{\vap \in \spA} \Q*{\vxp}{\vap; \old{\vtheta}} - \Q*{\vx}{\va; \vtheta}}^2. \label{eq:q_loss}
\end{align}
Here, we simply use Bellman's optimality equation \eqref{eq:bop2} to estimate $\q*{\vx}{\va}$, instead of the estimation of $\v[\pi]{\vx}$ using Bellman's expectation equation for TD-learning.
The difference between the current approximation and the optimization target, \begin{align}
  \delta_\mathrm{B} \defeq r + \gamma \max_{\vap \in \spA} \Q*{\vxp}{\vap; \old{\vtheta}} - \Q*{\vx}{\va; \vtheta}, \label{eq:bellman_error}
\end{align} is called the \midx{Bellman error}[idxpagebf].
Analogously to TD-learning,\footnote{compare to \cref{eq:td_learning_gradient_update}} we obtain the gradient update, \begin{align}
  \vtheta &\gets \vtheta - \alpha_t \grad_\vtheta \ell(\vtheta; \vx, \va, r, \vxp) \\
  &= \vtheta - \alpha_t \grad_\vtheta \frac{1}{2}\parentheses*{r + \gamma \max_{\vap \in \spA} \Q*{\vxp}{\vap; \old{\vtheta}} - \Q*{\vx}{\va; \vtheta}}^2 \margintag{using the definition of $\ell$ \eqref{eq:q_loss}} \nonumber \\
  &= \vtheta + \alpha_t \delta_\mathrm{B} \grad_\vtheta \Q*{\vx}{\va; \vtheta}. \label{eq:q_learning_gradient_update_general} \margintag{using the chain rule}
  \intertext{When using a neural network to learn $\fnQ[\star]$, we can use automatic differentiation to obtain unbiased gradient estimates. In the case of linear function approximation, we can compute the gradient exactly,}
  &= \vtheta + \alpha_t \delta_\mathrm{B} \grad_\vtheta \transpose{\vtheta} \vphi(\vx, \va) \margintag{using the linear approximation of $\fnQ[\star]$ \eqref{eq:q_linear_approx}} \nonumber \\
  &= \vtheta + \alpha_t \delta_\mathrm{B} \vphi(\vx, \va). \label{eq:q_learning_gradient_update}
\end{align}

In the tabular setting, this algorithm is identical to Q-learning and, in particular, converges to the true Q-function $\fnq[\star]$.\footnote{see \cref{thm:q_learning_convergence}}
There are few such results in the approximate setting. Usage in practice indicates that using an approximation of the value function ``should be fine'' when a ``rich-enough'' class of functions is used.

\subsection{Heuristics}\label{sec:mfarl:value_function_approximation:heuristics}

The vanilla stochastic semi-gradient descent is very slow.
In this subsection, we will discuss some ``tricks of the trade'' to improve its performance.\looseness=-1

\paragraph{Stabilizing optimization targets:}

There are mainly two problems.
The first problem is that, as mentioned previously, the bootstrapping estimate changes after each iteration.
As we are trying to learn an approximate value function that depends on the bootstrapping estimate, this means that the optimization target is ``moving'' between iterations.
In practice, moving targets lead to stability issues.
The first family of techniques we discuss here aim to ``stabilize'' the optimization targets.

One such technique is called \midx{neural fitted Q-iteration} or \midx{deep Q-networks} (DQN) \citep{mnih2015human}.
DQN updates the neural network used for the approximate bootstrapping estimate infrequently to maintain a constant optimization target across multiple episodes.
How this is implemented exactly varies.
One approach is to clone the neural network and maintain one changing neural network (``online network'') for the most recent estimate of the Q-function which is parameterized by $\vtheta$, and one fixed neural network (``target network'') used as the target which is parameterized by $\old{\vtheta}$ and which is updated infrequently.

This can be implemented by maintaining a data set $\spD$ of observed transitions (the so-called \midx{replay buffer}) and then ``every once in a while'' (e.g., once $\card{\spD}$ is large enough) solving a regression problem, where the labels are determined by the target network.
This yields a loss term where the target is fixed across all transitions in the replay buffer $\spD$, \begin{align}
  \ell_{\mathrm{DQN}}(\vtheta; \spD) \defeq \frac{1}{2} \sum_{(\vx, \va, r, \vxp) \in \spD} \parentheses*{r + \gamma \max_{\vap \in \spA} \Q*{\vxp}{\vap; \old{\vtheta}} - \Q*{\vx}{\va; \vtheta}}^2. \label{eq:dqn} \margintag{compare to the Q-learning loss \eqref{eq:q_loss}}
\end{align}
The loss can also be interpreted (in an online sense) as performing regular Q-learning with the modification that the target network $\old{\vtheta}$ is not updated to $\vtheta$ after every observed transition, but instead only after observing $\card{\spD}$-many transitions.
This technique is known as \midx{experience replay}.
Another approach is \midx{Polyak averaging} where the target network is gradually ``nudged'' by the neural network used to estimate the Q-function.\looseness=-1

\paragraph{Maximization bias:}

\begin{figure*}
  \begin{center}
    \import{./plots/output/}{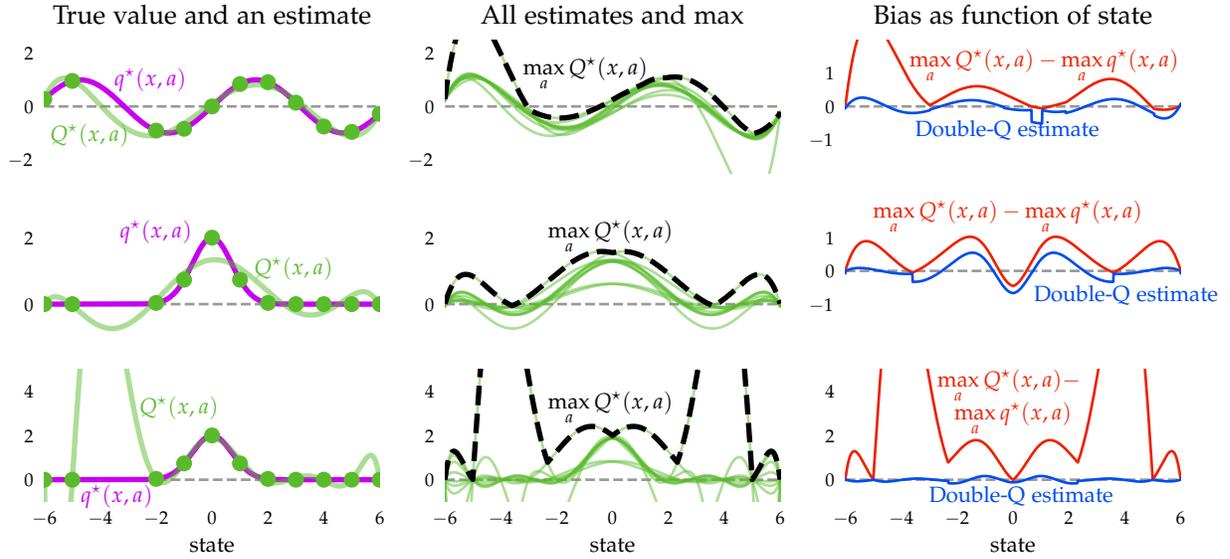}
  \end{center}

  \caption{Illustration of overestimation during learning. In each state (x-axis), there are 10 actions. The left column shows the true values $\v[\star]{x}$ (purple line). All true action values are defined by $\q[\star]{x}{a} \defeq \v[\star]{x}$. The green line shows estimated values $\Q[\star]{x}{a}$ for one action as a function of state, fitted to the true value at several sampled states (green dots). The middle column plots show all the estimated values (green), and the maximum of these values (dashed black). The maximum is higher than the true value (purple, left plot) almost everywhere. The right column plots show the difference in red. The blue line in the right plots is the estimate used by Double Q-learning with a second set of samples for each state. The blue line is much closer to zero, indicating less bias. The three rows correspond to different true functions (left, purple) or capacities of the fitted function (left, green). Reproduced with permission from \icite{van2016deep}.}\label{fig:maximization_bias}
\end{figure*}

Now, observe that the estimates $\fnQ[\star]$ are noisy estimates of $\fnq[\star]$ and consider the term, \begin{align*}
  \max_{\vap \in \spA} \q*{\vxp}{\vap} \approx \max_{\vap \in \spA} \Q*{\vxp}{\vap; \old{\vtheta}},
\end{align*} from the loss function \eqref{eq:dqn}.
This term maximizes a \emph{noisy} estimate of~$\fnq[\star]$, which leads to a \emph{biased} estimate of $\max \fnq[\star]$ as can be seen in \cref{fig:maximization_bias}.
The fact that the update rules of Q-learning and DQN are affected by inaccuracies (i.e., noise in the estimates) of the learned Q-function is known as the ``\midx{maximization bias}''.

\midx<Double DQN>{double DQN} (DDQN) is an algorithm that addresses this maximization bias \citep{van2016deep}.
Instead of picking the optimal action with respect to the old network, it picks the optimal action with respect to the new network, \begin{align}
  &\ell_{\mathrm{DDQN}}(\vtheta; \spD) \defeq \frac{1}{2} \sum_{(\vx, \va, r, \vxp) \in \spD} \parentheses*{r + \gamma \Q*{\vxp}{\vas(\vxp; \vtheta); \old{\vtheta}} - \Q*{\vx}{\va; \vtheta}}^2 \label{eq:ddqn} \\
  &\text{where} \quad \vas(\vxp; \vtheta) \defeq \argmax_{\vap \in \spA} \Q*{\vxp}{\vap; \vtheta}.
\end{align}
Intuitively, this change ensures that the evaluation of the target network is consistent with the updated Q-function, which makes the algorithm more robust to noise.

Similarly to DQN, this can also be interpreted as the online update, \begin{align}
  \vtheta \gets \vtheta + \alpha_t \parentheses*{r + \gamma \Q*{\vxp}{\vas(\vxp; \vtheta); \old{\vtheta}} - \Q*{\vx}{\va; \vtheta}} \grad_\vtheta \Q*{\vx}{\va; \vtheta} \label{eq:ddqn_single_update}
\end{align} after observing a single transition $(\vx, \va, r, \vxp)$ where while differentiating, $\vas(\vxp; \vtheta)$ is treated as constant with respect to $\vtheta$.
$\old{\vtheta}$ is then updated to $\vtheta$ after observing $\card{\spD}$-many transitions.

\section{Policy Approximation}\label{sec:mfarl:policy_approximation}

Q-learning defines a policy implicitly by \begin{align}
  \opt{\vpi}(\vx) \defeq \argmax_{\va \in \spA} \Q*{\vx}{\va}. \label{eq:q_learning_policy}
\end{align}
Q-learning also maximizes over the set of all actions in its update step while learning the Q-function.
This is intractable for large and, in particular, continuous action spaces.
A natural idea to escape this limitation is to immediately learn an approximate parameterized policy, \begin{align}
  \opt{\vpi}(\vx) \approx \vpi(\vx; \vvarphi) \eqdef \vpi_\vvarphi(\vx).
\end{align}
Methods that find an approximate policy are also called \midx<policy search methods>{policy search method} or \midx<policy gradient methods>{policy gradient method}.

Whereas with Q-learning, exploration can be encouraged by using an $\varepsilon$-greedy policy, softmax exploration, or an optimistic initialization, we will see later that policy gradient methods fundamentally rely on randomized policies for exploration.

\begin{rmk}{Notation}{}
  We refer to deterministic policies $\vpi$ in bold, as they can be interpreted as vector-valued functions from $\spX$ to $\spA$.
  We still refer to randomized policies by $\pi$, as for each state $\vx \in \spX$ they are represented as a PDF over actions $\spA$.

  In particular, we denote by $\pi_\vvarphi(\va \mid \vx)$ the probability (density) of playing action $\va$ when in state $\vx$ according to $\pi_\vvarphi$.
\end{rmk}

\subsection{Estimating Policy Values}

We will begin by attributing a ``value'' to a policy.
Recall the definition of the discounted payoff $G_t$ from time $t$, which we are aiming to maximize,\looseness=-1 \begin{align*}
  G_t = \sum_{m=0}^\infty \gamma^m R_{t+m}. \margintag{see \cref{eq:discounted_payoff}}
\end{align*}
We define $G_{t:T}$ to be the \midx{bounded discounted payoff} until time $T$, \begin{align}
  G_{t:T} \defeq \sum_{m=0}^{T-1-t} \gamma^m R_{t+m}. \label{eq:bounded_discounted_payoff}
\end{align}
Based on these two random variables, we can define the policy value function:\looseness=-1

\begin{defn}[Policy value function]\pidx{policy value function}
  The \emph{policy value function}, \begin{align}
    \j{\pi} \defeq \E[\pi]{G_0} = \E[\pi]{\sum_{t=0}^\infty \gamma^t R_t},
  \end{align} measures the expected discounted payoff of policy $\pi$.\footnote{We neglect here that implicitly one also averages over the initial state if this state is not fixed.}
  We also define the bounded variant, \begin{align}
    \j{\pi}[T] \defeq \E[\pi]{G_{0:T}} = \E[\pi]{\sum_{t=0}^{T-1} \gamma^t R_t}.
  \end{align}
\end{defn}
For simplicity, we will abbreviate $\j{\vvarphi} \defeq \j{\pi_\vvarphi}$.

\begin{rmk}{Notation of policy value function}{}
  We adopt the more common notation $\j{\pi}$ for the policy value function, as opposed to $j(\pi)$, which would be consistent with our notation of the (true) value functions $\fnv[\pi], \fnq[\pi]$.
  So don't be confused by this: just like the value functions $\fnv[\pi], \fnq[\pi]$, the policy value function $\j{\pi}$ is a deterministic object, measuring the mean discounted payoff.
  We will use $\J{\pi}$ to refer to our estimates of the policy value function.
\end{rmk}

Naturally, we want to maximize $\j{\vvarphi}$. That is, we want to solve \begin{align}
  \opt{\vvarphi} \defeq \argmax_\vvarphi \j{\vvarphi} \label{eq:policy_optimization}
\end{align} which is a non-convex optimization problem.
Let us see how $\j{\vvarphi}$ can be evaluated to understand the optimization problem better.
We will again use a Monte Carlo estimate.
Recall that a fixed $\vvarphi$ induces a unique Markov chain, which can be simulated.
In the episodic setting, each episode (also called \midx{rollout}) of length $T$ yields an independently sampled trajectory, \begin{align}
  \tau^{(i)} \defeq ((\vx_0^{(i)}, \va_0^{(i)}, r_0^{(i)}, \vx_1^{(i)}), (\vx_1^{(i)}, \va_1^{(i)}, r_1^{(i)}, \vx_2^{(i)}), \dots)
\end{align}
Simulating $m$ rollouts yields the samples $\tau^{(1)}, \dots, \tau^{(m)} \sim \altpi_\vvarphi$, where $\altpi_\vvarphi$ is the distribution of all trajectories (i.e., rollouts) of the Markov chain induced by policy $\pi_\vvarphi$.
We denote the (bounded) discounted payoff\pidx{discounted payoff} of the $i$-th rollout by \begin{align}
  g_{0:T}^{(i)} \defeq \sum_{t=0}^{T-1} \gamma^t r_t^{(i)} \label{eq:bounded_discounted_reward_sample}
\end{align} where $r_t^{(i)}$ is the reward at time $t$ of the $i$-th rollout.
Using a Monte Carlo approximation, we can then estimate $\j{\vvarphi}[T]$.
Moreover, due to the exponential discounting of future rewards, it is reasonable to approximate the policy value function using bounded trajectories, \begin{align}
  \j{\vvarphi} \approx \j{\vvarphi}[T] \approx \J{\vvarphi}[T] \defeq \frac{1}{m} \sum_{i=1}^m g_{0:T}^{(i)}.
\end{align}

\subsection{Policy Gradient}\label{sec:mfarl:policy_approximation:reinforce}

Policy gradient methods solve the above optimization problem~\eqref{eq:policy_optimization} by stochastic gradient ascent on the policy parameter $\vvarphi$: \begin{align}
  \vvarphi \gets \vvarphi + \eta \grad_\vvarphi \j{\vvarphi}. \label{eq:policy_gradient_ascent}
\end{align}

How can we compute the policy gradient?
Let us first formally define the distribution over trajectories $\altpi_\vvarphi$ that we introduced in the previous section.
We can specify the probability of a specific trajectory $\tau$ under a policy $\pi_\vvarphi$ by \begin{align}
  \altpi_\vvarphi(\tau)  = p(\vx_0) \prod_{t=0}^{T-1} \pi_\vvarphi(\va_t \mid \vx_t) p(\vx_{t+1} \mid \vx_t, \va_t). \label{eq:trajectory_distr}
\end{align}

For optimizing $\j{\vvarphi}$ we need to obtain unbiased gradient estimates: \begin{align}
  \grad_\vvarphi \j{\vvarphi} \approx \grad_\vvarphi \j{\vvarphi}[T] = \grad_\vvarphi \E[\tau \sim \altpi_\vvarphi]{G_{0:T}}.
\end{align}
Note that the expectation integrates over the measure $\altpi_\vvarphi$, which depends on the parameter $\vvarphi$.
Thus, we cannot move the gradient operator inside the expectation as we have often done previously~(cf.~\cref{sec:background:probability:gradients_of_expectations}).
This should remind you of the reparameterization trick (see \cref{eq:reparameterization_trick1}) that we used to solve a similar gradient in the context of variational inference.
In this context, however, we cannot apply the reparameterization trick.\footnote{This is because the distribution $\altpi_\vvarphi$ is generally not reparameterizable.
We will, however, see that reparameterization gradients are also useful in reinforcement learning. See, e.g., \cref{sec:mfarl:actor_critic_methods:randomized_policies,sec:mbarl:planning:stochastic_dynamics}.}
Fortunately, there is another way of estimating this gradient.

\begin{thm}[Score gradient estimator]\pidx{score gradient estimator}[idxpagebf]
  Under some regularity assumptions, we have \begin{align}
    \grad_\vvarphi \E[\tau \sim \altpi_\vvarphi]{G_0} = \E[\tau \sim \altpi_\vvarphi]{G_0 \grad_\vvarphi \log \altpi_\vvarphi(\tau)}. \label{eq:score_gradient_estimator}
  \end{align}
  This estimator of the gradient is called the \emph{score gradient estimator}.
\end{thm}
\begin{proof}
  To begin with, let us look at the so-called \midx{score function} of the distribution $\altpi_\vvarphi$, $\grad_\vvarphi \log \altpi_\vvarphi(\tau)$.
  Using the chain rule, the score function can be expressed as \begin{align}
    \grad_\vvarphi \log \altpi_\vvarphi(\tau) = \frac{\grad_\vvarphi \altpi_\vvarphi(\tau)}{\altpi_\vvarphi(\tau)} \label{eq:score_function}
  \end{align} and by rearranging the terms, we obtain \begin{align}
    \grad_\vvarphi \altpi_\vvarphi(\tau) = \altpi_\vvarphi(\tau) \grad_\vvarphi \log \altpi_\vvarphi(\tau). \label{eq:score_function_trick}
  \end{align}
  This is called the \midx{score function trick}.\footnote{We have already applied this ``trick'' in \cref{exercise:langevin_dynamics_convergence}.}

  Now, assuming that state and action spaces are continuous, we obtain \begin{align*}
    \grad_\vvarphi \E[\tau \sim \altpi_\vvarphi]{G_0} &= \grad_\vvarphi \int \altpi_\vvarphi(\tau) \cdot G_0 \,d\tau \margintag{using the definition of expectation \eqref{eq:expectation}} \\
    &= \int \grad_\vvarphi \altpi_\vvarphi(\tau) \cdot G_0 \,d\tau \margintag{using the regularity assumptions to swap gradient and integral} \\
    &= \int G_0 \cdot \altpi_\vvarphi(\tau) \grad_\vvarphi \log \altpi_\vvarphi(\tau) \,d\tau \margintag{using the score function trick \eqref{eq:score_function_trick}} \\
    &= \E[\tau \sim \altpi_\vvarphi]{G_0 \grad_\vvarphi \log \altpi_\vvarphi(\tau)}. \qedhere \margintag{interpreting the integral as an expectation over $\altpi_\vvarphi$}
  \end{align*}
\end{proof}

Intuitively, maximizing $\j{\vvarphi}$ increases the probability of policies with high returns and decreases the probability of policies with low returns.

To use the score gradient estimator for estimating the gradient, we need to compute $\grad_\vvarphi \log \altpi_\vvarphi(\tau)$. \begin{align}
  &\grad_\vvarphi \log \altpi_\vvarphi(\tau) \nonumber \\
  &= \grad_\vvarphi \parentheses*{\log p(\vx_0) + \sum_{t=0}^{T-1} \log \pi_\vvarphi(\va_t \mid \vx_t) + \sum_{t=0}^{T-1} \log p(\vx_{t+1} \mid \vx_t, \va_t)} \margintag{using the definition of the distribution over trajectories $\altpi_\vvarphi$} \nonumber \\
  &= \grad_\vvarphi \log p(\vx_0) + \sum_{t=0}^{T-1} \grad_\vvarphi \log \pi_\vvarphi(\va_t \mid \vx_t) + \sum_{t=0}^{T-1} \grad_\vvarphi \log p(\vx_{t+1} \mid \vx_t, \va_t) \nonumber \\
  &= \sum_{t=0}^{T-1} \grad_\vvarphi \log \pi_\vvarphi(\va_t \mid \vx_t). \margintag{using that the first and third term are independent of $\vvarphi$}
\end{align}
When using a neural network for the parameterization of the policy $\pi$, we can use automatic differentiation to compute the gradients.

The expectation of the score gradient estimator \eqref{eq:score_gradient_estimator} can be approximated using Monte Carlo sampling, \begin{align}
  \grad_\vvarphi \j{\vvarphi}[T] \approx \grad_\vvarphi \J{\vvarphi}[T] \defeq \frac{1}{m} \sum_{i=1}^m g_{0:T}^{(i)} \sum_{t=0}^{T-1} \grad_\vvarphi \log \pi_\vvarphi(\va_t^{(i)} \mid \vx_t^{(i)}). \label{eq:score_gradient_estimator_approx}
\end{align}
However, typically the variance of these estimates is very large.
Using so-called \midx<baselines>{baseline}, we can reduce the variance dramatically \exerciserefmark{score_gradients_with_baselines_variance}.

\begin{lem}[Score gradients with baselines]\label{lem:score_gradients_with_baselines}
  We have, \begin{align}
    \E[\tau \sim \altpi_\vvarphi]{G_0 \grad_\vvarphi \log \altpi_\vvarphi(\tau)} = \E[\tau \sim \altpi_\vvarphi]{(G_0 - b) \grad_\vvarphi \log \altpi_\vvarphi(\tau)}.
  \end{align}
  Here, $b \in \R$ is called a \emph{baseline}.
\end{lem}
\begin{proof}
  For the term to the right, we have due to linearity of expectation~\eqref{eq:linearity_expectation}, \begin{align*}
    \E[\tau \sim \altpi_\vvarphi]{(G_0 - b) \grad_\vvarphi \log \altpi_\vvarphi(\tau)} = \begin{multlined}[t]
      \E[\tau \sim \altpi_\vvarphi]{G_0 \grad_\vvarphi \log \altpi_\vvarphi(\tau)} \\ - \E[\tau \sim \altpi_\vvarphi]{b \cdot \grad_\vvarphi \log \altpi_\vvarphi(\tau)}.
    \end{multlined}
  \end{align*}
  Thus, it remains to show that the second term is zero, \begin{align*}
    \E[\tau \sim \altpi_\vvarphi]{b \cdot \grad_\vvarphi \log \altpi_\vvarphi(\tau)} &= b \cdot \int \altpi_\vvarphi(\tau) \grad_\vvarphi \log \altpi_\vvarphi(\tau) \,d\tau \margintag{using the definition of expectation \eqref{eq:expectation}} \\
    &= b \cdot \int \altpi_\vvarphi(\tau) \frac{\grad_\vvarphi \altpi_\vvarphi(\tau)}{\altpi_\vvarphi(\tau)} \,d\tau \margintag{substituting the score function \eqref{eq:score_function}, ``undoing the score function trick''} \\
    &= b \cdot \int \grad_\vvarphi \altpi_\vvarphi(\tau) \,d\tau \margintag{$\altpi_\vvarphi(\tau)$ cancels} \\
    &= b \cdot \grad_\vvarphi \int \altpi_\vvarphi(\tau) \,d\tau \\
    &= b \cdot \grad_\vvarphi 1 = 0. \qedhere \margintag{integrating a PDF over its domain is $1$ and the derivative of a constant is $0$}
  \end{align*}
\end{proof}

One can even show, that we can subtract arbitrary baselines depending on \emph{previous} states \exerciserefmark{score_gradients_state_dep_baselines}.

\begin{ex}{Downstream returns}{downstream_returns}
  A commonly used state-dependent baseline is \begin{align}
    b(\tau_{0:t-1}) \defeq \sum_{m=0}^{t-1} \gamma^m r_m.
  \end{align}
  This baseline subtracts the returns of all actions before time $t$.
  Intuitively, using this baseline, the score gradient only considers downstream returns.
  Recall from \cref{eq:bounded_discounted_payoff} that we defined $G_{t:T}$ as the bounded discounted payoff from time $t$.
  It is also commonly called the (bounded) \midx{downstream return} (or \midx{reward to go}) beginning at time $t$.

  For a fixed trajectory $\tau$ that is bounded at time $T$, we have \begin{align}
    G_0 - b(\tau_{0:t-1}) = \gamma^t G_{t:T},
  \end{align} yielding the gradient estimator, \begin{align}
    \grad_\vvarphi \j{\vvarphi} \approx \grad_\vvarphi \j{\vvarphi}[T] &= \E[\tau \sim \altpi_\vvarphi]{G_0 \grad_\vvarphi \log \altpi_\vvarphi(\tau)} \margintag{using the score gradient estimator \eqref{eq:score_gradient_estimator}} \nonumber \\
    &= \E[\tau \sim \altpi_\vvarphi]{\sum_{t=0}^{T-1} \gamma^t G_{t:T} \grad_\vvarphi \log \pi_\vvarphi(\va_t \mid \vx_t)}. \margintag{using a state-dependent baseline \eqref{eq:score_gradient_estimator_state_dep_baseline}} \label{eq:score_gradient_estimator_downstream_returns}
  \end{align}
\end{ex}

Performing stochastic gradient descent with the score gradient estimator and downstream returns is known as the \emph{REINFORCE algorithm}~\citep{williams1992simple} which is shown in~\cref{alg:reinforce}.

\begin{algorithm}[H]
  \caption{REINFORCE algorithm}\label{alg:reinforce}\pidx{REINFORCE algorithm}
  initialize policy weights $\vvarphi$\;
  \Repeat{converged}{
    generate an episode (i.e., rollout) to obtain trajectory $\tau$\;
    \For{$t = 0$ \KwTo $T-1$}{
      set $g_{t:T}$ to the downstream return from time $t$\;
      $\vvarphi \gets \vvarphi + \eta \gamma^t g_{t:T} \grad_\vvarphi \log \pi_\vvarphi(\va_t \mid \vx_t)$ \algeq{eq:reinforce}
    }
  }
\end{algorithm}

The variance of REINFORCE can be reduced further.
A common technique is to subtract a term $b_t$ from the downstream returns, \begin{align}
  \grad_\vvarphi \j{\vvarphi} = \E[\tau \sim \altpi_\vvarphi]{\sum_{t=0}^T \gamma^t (G_{t:T} - b_t) \grad_\vvarphi \log \pi_\vvarphi(\va_t \mid \vx_t)}.
\end{align}
For example, we can subtract the $t$-independent mean reward to go, \begin{align}
  b_t \defeq b = \frac{1}{T} \sum_{t'=0}^{T-1} G_{t':T}.
\end{align}

The main advantage of policy gradient methods such as REINFORCE is that they can be used in continuous action spaces.
However, REINFORCE is not guaranteed to find an optimal policy.
Even when operating in very small domains, REINFORCE can get stuck in local optima.\looseness=-1

Typically, policy gradient methods are slow due to the large variance in the score gradient estimates.
Because of this, they need to take small steps and require many rollouts of a Markov chain.
Moreover, we cannot reuse data from previous rollouts, as policy gradient methods are fundamentally on-policy.\footnote{This is because the score gradient estimator is used to obtain gradients of the policy value function with respect to the \emph{current} policy.}

Next, we will combine value approximation techniques like Q-learning and policy gradient methods, leading to an often more practical family of methods called actor-critic methods.

\section{On-policy Actor-Critics}\label{sec:mfarl:actor_critic_methods}

Actor-Critic methods reduce the variance of policy gradient estimates by using ideas from value function approximation.
They use function approximation both to approximate value functions \emph{and} to approximate policies.
The goal for these algorithms is to scale to reinforcement learning problems, where we both have large state spaces and large action spaces.

\subsection{Advantage Function}

A key concept of actor-critic methods is the advantage function.

\begin{defn}[Advantage function]\pidx{advantage function}
  Given a policy $\pi$, the \emph{advantage function},\looseness=-1 \begin{align}
    \a[\pi]{\vx}{\va} &\defeq \q[\pi]{\vx}{\va} - \v[\pi]{\vx} \label{eq:advantage_function1} \\
    &= \q[\pi]{\vx}{\va} - \E[\vap \sim \pi(\vx)]{\q[\pi]{\vx}{\vap}}, \label{eq:advantage_function2} \margintag{using \cref{eq:v_as_q}}
  \end{align} measures the advantage of picking action $\va \in \spA$ when in state $\vx \in \spX$ over simply following policy $\pi$.
\end{defn}

It follows immediately from \cref{eq:advantage_function2} that for any policy $\pi$ and state~${\vx \in \spX}$, there exists an action $\va \in \spA$ such that $\a[\pi]{\vx}{\va}$ is non-negative, \begin{align}
  \max_{\va \in \spA} \a[\pi]{\vx}{\va} \geq 0.
\end{align}
Moreover, it follows directly from Bellman's theorem \eqref{eq:bop1} that \begin{align}
  \text{$\pi$ is optimal} \iff \forall \vx \in \spX, \va \in \spA : \a[\pi]{\vx}{\va} \leq 0.
\end{align}
In other words, quite intuitively, $\pi$ is optimal if and only if there is no action that has an advantage in any state over the action that is played by~$\pi$.\looseness=-1

Finally, we can re-define the greedy policy $\pi_{\fnq}$ with respect to the state-action value function $\fnq$ as \begin{align}
  \pi_{\fnq}(\vx) \defeq \argmax_{\va \in \spA} \a{\vx}{\va}
\end{align} since \begin{align*}
  \argmax_{\va \in \spA} \a{\vx}{\va} = \argmax_{\va \in \spA} \q{\vx}{\va} - \v{\vx} = \argmax_{\va \in \spA} \q{\vx}{\va},
\end{align*} as $\v{\vx}$ is independent of $\va$.
This coincides with our initial definition of greedy policies in \cref{eq:greedy_policy1}.
Intuitively, the advantage function is a shifted version of the state-action value function $\fnq$ that is relative to~$0$.
Using this quantity rather than $\fnq$ often has numerical advantages.

\subsection{Policy Gradient Theorem}\label{sec:mfarl:actor_critic_methods:policy_gradients}

Recall the score gradient estimator \eqref{eq:score_gradient_estimator_downstream_returns} that we had introduced in the previous section, \begin{align*}
  \grad_\vvarphi \j{\vvarphi}[T] = \E[\tau \sim \altpi_\vvarphi]{\sum_{t=0}^{T-1} \gamma^t G_{t:T} \grad_\vvarphi \log \pi_\vvarphi(\va_t \mid \vx_t)}.
\end{align*}
Previously, we have approximated the policy value function $\j{\vvarphi}$ by the bounded policy value function $\j{\vvarphi}[T]$.
We said that this approximation was ``reasonable'' due to the diminishing returns.
Essentially, we have ``cut off the tails'' of the policy value function.
Let us now reinterpret score gradients while taking into account the tails of $\j{\vvarphi}$. \begin{align}
  \grad_\vvarphi \j{\vvarphi} &= \lim_{T\to\infty} \grad_\vvarphi \j{\vvarphi}[T] \\
  &= \sum_{t=0}^\infty \E[\tau \sim \altpi_\vvarphi]{\gamma^t G_t \grad_\vvarphi \log \pi_\vvarphi(\va_t \mid \vx_t)}. \margintag{substituting the score gradient estimator with downstream returns \eqref{eq:score_gradient_estimator_downstream_returns} and using linearity of expectation \eqref{eq:linearity_expectation}} \nonumber
  \intertext{Observe that because the expectations only consider downstream returns, we can disregard all data from the trajectory prior to time $t$. Let us define \begin{align}
    \tau_{t:\infty} \defeq ((\vx_t, \va_t, r_t, \vx_{t+1}), (\vx_{t+1}, \va_{t+1}, r_{t+1}, \vx_{t+2}), \dots),
  \end{align} as the trajectory from time step $t$. Then,}
  &= \sum_{t=0}^\infty \E[\tau_{t:\infty} \sim \altpi_\vvarphi]{\gamma^t G_t \grad_\vvarphi \log \pi_\vvarphi(\va_t \mid \vx_t)}. \nonumber
  \intertext{We now condition on $\vx_t$ and $\va_t$,}
  &= \sum_{t=0}^\infty \E[\vx_t, \va_t]{\gamma^t \E[r_t, \tau_{t+1:\infty}]{G_t}[\vx_t, \va_t] \grad_\vvarphi \log \pi_\vvarphi(\va_t \mid \vx_t)}. \margintag{using that $\pi_\vvarphi(a_t \mid x_t)$ is a constant given $x_t$ and $a_t$} \nonumber
  \intertext{Observe that averaging over the trajectories $\E[\tau \sim \altpi_\vvarphi]{\cdot}$ that are sampled according to policy $\pi_\vvarphi$ is equivalent to our shorthand notation $\E[\pi_\vvarphi]{\cdot}$ from \cref{eq:expectation_over_policy},}
  &= \sum_{t=0}^\infty \E[\vx_t, \va_t]{\gamma^t \E[\pi_\vvarphi]{G_t}[\vx_t, \va_t] \grad_\vvarphi \log \pi_\vvarphi(\va_t \mid \vx_t)} \nonumber \\
  &= \sum_{t=0}^\infty \E[\vx_t, \va_t]{\gamma^t \q[\pi_\vvarphi]{\vx_t}{\va_t} \grad_\vvarphi \log \pi_\vvarphi(\va_t \mid \vx_t)}. \margintag{using the definition of the Q-function \eqref{eq:q_function1}} \label{eq:policy_gradient_thm_proof}
\end{align}
It turns out that $\E[\vx_t, \va_t]{\q[\pi_\vvarphi]{\vx_t}{\va_t}}$ exhibits much less variance than our previous estimator $\E*[\vx_t, \va_t]{\E[\pi_\vvarphi]{G_t}[\vx_t, \va_t]}$.
\Cref{eq:policy_gradient_thm_proof} is known as the \midx{policy gradient theorem}.

Often, the policy gradient theorem is stated in a slightly rephrased form in terms of the \midx{discounted state occupancy measure},\footnote{Depending on the reward setting, there exist various variations of the policy gradient theorem. We derived the variant for infinite-horizon discounted payoffs. \icite{sutton2018reinforcement} derive the variant for undiscounted average rewards.} \begin{align}
  \rho_\vvarphi^\infty(\vx) \defeq (1-\gamma) \sum_{t=0}^\infty \gamma^t p_{\rX_t}\!(\vx).
\end{align}
The factor $(1-\gamma)$ ensures that $\rho_\vvarphi^\infty$ is a probability density as \begin{align*}
  \int \rho_\vvarphi^\infty(\vx) \,d\vx = (1-\gamma) \sum_{t=0}^\infty \gamma^t \int p_{\rX_t}\!(\vx) \,d\vx = (1-\gamma) \sum_{t=0}^\infty \gamma^t = 1.
\end{align*}
Intuitively, $\rho_\vvarphi^\infty(\vx)$ measures how often we visit state $\vx$ when following policy $\pi_\vvarphi$.
It can be thought of as a ``discounted frequency''.

\begin{thm}[Policy gradient theorem in terms of $\rho_\vvarphi^\infty$]
  Policy gradients can be represented in terms of the Q-function, \begin{align}
    \grad_\vvarphi \j{\vvarphi} &\propto \E*[\vx \sim \rho_\vvarphi^\infty]{\E[\va \sim \pi_\vvarphi(\cdot \mid \vx)]{\q[\pi_\vvarphi]{\vx}{\va} \grad_\vvarphi \log \pi_\vvarphi(\va \mid \vx)}}. \label{eq:policy_gradient_thm}
  \end{align}
\end{thm}
\begin{proof}
  The right hand side of \cref{eq:policy_gradient_thm_proof} can be expressed as \begin{align*}
    &\sum_{t=0}^\infty \int p_{\rX_t}\!(\vx) \cdot \E[\va_t \sim \pi_\vvarphi(\cdot \mid \vx)]{\gamma^t \q[\pi_\vvarphi]{\vx}{\va_t} \grad_\vvarphi \log \pi_\vvarphi(\va_t \mid \vx)} \,d\vx \\
    &= \frac{1}{1-\gamma} \int \rho_\vvarphi^\infty(\vx) \cdot \E[\va \sim \pi_\vvarphi(\cdot \mid \vx)]{\q[\pi_\vvarphi]{\vx}{\va} \grad_\vvarphi \log \pi_\vvarphi(\va \mid \vx)} \,d\vx
  \end{align*} where we swapped the order of sum and integral and reorganized terms.\looseness=-1
\end{proof}

Matching our intuition, according to the policy gradient theorem, maximizing $\j{\vvarphi}$ corresponds to increasing the probability of actions with a large value and decreasing the probability of actions with a small value, taking into account how often the resulting policy visits certain states.\looseness=-1

Observe that we cannot use the policy gradient to calculate the gradient exactly, as we do not know $\fnq[\pi_\vvarphi]$.
Instead, we will use bootstrapping estimates $\fnQ[\pi_\vvarphi]$ of $\fnq[\pi_\vvarphi]$.

\subsection{A First Actor-Critic}

\begin{marginfigure}
  \incfig{actor_critic}
  \caption{Illustration of one iteration of actor-critic methods.
  The dependencies between the actors and critics are shown as arrows.
  Methods differ in the exact order in which actor and critic are updated.}
\end{marginfigure}

Actor-Critic methods\pidx{actor-critic method} consist of two components: \begin{itemize}
  \item a parameterized policy, $\pi(\va \mid \vx; \vvarphi) \eqdef \pi_\vvarphi$, which is called \embeq{eq:actor} \midx{actor}; and
  \item a value function approximation, $\q[\pi_\vvarphi]{\vx}{\va} \approx \Q[\pi_\vvarphi]{\vx}{\va; \vtheta}$, which is called \midx{critic}.
  In the following, we will abbreviate $\fnQ[\pi_\vvarphi]$ by $\fnQ$. \embeq{eq:critic}
\end{itemize}
In deep reinforcement learning, neural networks are used to parameterize both actor and critic.
Therefore, in principle, the actor-critic framework allows scaling to both large state spaces and large action spaces.
We begin by discussing on-policy actor-critics.

One approach in the online setting (i.e., non-episodic setting), is to simply use SARSA for learning the critic.
To learn the actor, we use stochastic gradient descent with gradients obtained using single samples from \begin{align}
  \grad_\vvarphi \j{\vvarphi} \approx \grad_\vvarphi \J{\vvarphi} \defeq \sum_{t=0} ^\infty\E[(\vx_t,\va_t) \sim \pi_\vvarphi]{\gamma^t \Q{\vx_t}{\va_t; \vtheta} \grad_\vvarphi \log \pi_\vvarphi(\va_t \mid \vx_t)} \label{eq:policy_gradient_bootstrapping_estimate} \margintag{see \cref{eq:policy_gradient_thm_proof}}
\end{align} where $\fnQ$ is a bootstrapping estimate of $\fnq[\pi_\vvarphi]$.
This algorithm is known as \emph{online actor-critic} or \midx{Q actor-critic} and shown in \cref{alg:oac}.

\begin{algorithm}
  \caption{Online actor-critic}\pidx{online actor-critic}\label{alg:oac}
  initialize parameters $\vvarphi$ and $\vtheta$\;
  \Repeat{converged}{
    use $\pi_\vvarphi$ to obtain transition $(\vx, \va, r, \vxp)$\;
    $\delta = r + \gamma \Q{\vxp}{\pi_\vvarphi(\vxp); \vtheta} - \Q{\vx}{\va; \vtheta}$\;
    \Comment{actor update}
    $\vvarphi \gets \vvarphi + \eta \gamma^t \Q{\vx}{\va; \vtheta} \grad_{\vvarphi} \log \pi_\vvarphi(\va \mid \vx)$ \algeq{eq:online_actor_critic1}
    \Comment{critic update}
    $\vtheta \gets \vtheta + \eta \delta \grad_{\vtheta} \Q{\vx}{\va; \vtheta}$ \algeq{eq:online_actor_critic2}
  }
\end{algorithm}

Comparing to the derivation for TD-learning from \cref{eq:td_learning_gradient_update}, we observe that \cref{eq:online_actor_critic2} corresponds to the SARSA update rule.\footnote{The gradient with respect to $\vtheta_Q$ appears analogously to our derivation of approximate Q-learning \eqref{eq:q_learning_gradient_update}.}
Due to the use of SARSA for learning the critic, this algorithm is fundamentally on-policy.

Crucially, by neglecting the dependence of the bootstrapping estimate $\fnQ$ on the policy parameters $\vvarphi$, we introduce bias in the gradient estimates.
In other words, using the bootstrapping estimate $\fnQ$ means that the resulting gradient direction might not be a valid ascent direction.
In particular, the actor is not guaranteed to improve.
Still, it turns out that under strong so-called ``compatibility conditions'' that are rarely satisfied in practice, a valid ascent direction can be guaranteed.

\subsection{Improved Actor-Critics}\label{sec:mfarl:actor_critc_methods:improved}

\paragraph{Reducing variance:}

To further reduce the variance of the gradient estimates, it turns out that a similar approach to the baselines we discussed in the previous section on policy gradient methods is useful.
A common approach is to subtract the state value function from estimates of the Q-function, \begin{align}
  \vvarphi &\gets \vvarphi + \eta_t \gamma^t (\Q{\vx}{\va; \vtheta} - \V{\vx; \vtheta}) \grad_{\vvarphi} \log \pi_\vvarphi(\va \mid \vx) \\
  &= \vvarphi + \eta_t \gamma^t \A{\vx}{\va; \vtheta} \grad_{\vvarphi} \log \pi_\vvarphi(\va \mid \vx) \margintag{using the definition of the advantage function \eqref{eq:advantage_function1}}
\end{align} where $\A{\vx}{\va; \vtheta}$ is a bootstrapped estimate of the advantage function~$\fna[\pi_\vvarphi]$.
This algorithm is known as \midx{advantage actor-critic} (A2C) \citep{mnih2016asynchronous}.
Recall that the Q-function is an absolute quantity, whereas the advantage function is a relative quantity, where the sign is informative for the gradient direction.
Intuitively, an absolute value is harder to estimate than the sign.
Actor-Critic methods are therefore often implemented with respect to the advantage function rather than the Q-function.\looseness=-1

Taking a step back, observe that policy gradient methods such as REINFORCE generally have \emph{high variance} in their gradient estimates.
However, due to using Monte Carlo estimates of $G_t$, the gradient estimates are \emph{unbiased}.
In contrast, using a bootstrapped Q-function to obtain gradient estimates yields estimates with a \emph{smaller variance}, but those estimates are \emph{biased}.
We are faced with a \midx{bias-variance tradeoff}.
A natural approach is therefore to blend both gradient estimates to allow for effectively trading bias and variance.
This leads to algorithms such as \midx{generalized advantage estimation} (GAE) \citep{schulman2015high}.

\paragraph{Exploration:}

Similarly to REINFORCE, actor-critic methods typically rely on randomization in the policy to encourage exploration, the idea being that if the policy is stochastic, then the agent will visit a diverse set of states.
The inherent stochasticity of the policy is, however, often insufficient.
A common problem is that the policy quickly ``collapses'' to a deterministic policy since the objective function is greedily exploitative.
A common workaround is to use an $\varepsilon$-greedy policy (cf. \cref{sec:tabular_rl:mb:epsilon_greedy}) or to explicitly encourage the policy to exhibit uncertainty by adding an entropy term to the objective function (more on this in \cref{sec:mfarl:actor_critic_methods:entropy_regularization}).
However, note that for on-policy methods, changing the policy also changes the value function learned by the critic.

\paragraph{Improving sample efficiency:}

Actor-Critic methods typically suffer from low sample efficiency.
When additionally using an on-policy method, actor-critics often need an extremely large number of interactions before learning a near-optimal policy, because they cannot reuse past data.
Allowing to reuse past data is a major advantage of off-policy methods like Q-learning.

One well-known variant that slightly improves the sample efficiency is \midx{trust-region policy optimization} (TRPO) \citep{schulman2015trust}.
TRPO uses multiple iterations, where in each iteration a fixed critic is used to optimize the policy.\footnote{Intuitively, each iteration performs a collection of gradient ascent steps.}
During iteration $k$, we select \begin{align}
  \vvarphi_{k+1} \gets \argmax_{\vvarphi} \J{\vvarphi} \quad \text{subject to}\ \E*[\vx \sim \rho_{\vvarphi_k}^\infty]{\KL{\pi_{\vvarphi_k}(\cdot \mid \vx)}{\pi_\vvarphi(\cdot \mid \vx)}} \leq \delta \label{eq:trpo} \\[-35pt] \nonumber
\end{align} for some fixed $\delta > 0$ and where \begin{align}
  \J{\vvarphi} \defeq \E[\vx \sim \rho_{\vvarphi_k}^\infty, \va \sim \pi_{\vvarphi_k}(\cdot \mid \vx)]{w_k(\vvarphi; \vx, \va) \A[\pi_{\vvarphi_k}]{\vx}{\va}}.
\end{align}
Notably, $\fnJ$ is an expectation with respect to the \emph{previous} policy $\pi_{\vvarphi_k}$ and the previous critic $\fnA[\pi_{\vvarphi_k}]$.
TRPO uses \midx{importance sampling} where the importance weights (called ``likelihood ratios''), \begin{align*}
  w_k(\vvarphi; \vx, \va) \defeq \frac{\pi_\vvarphi(\va \mid \vx)}{\pi_{\vvarphi_k}(\va \mid \vx)},
\end{align*} are used to correct for taking the expectation over the previous policy.
When $w_k \approx 1$ the policies $\pi_{\vvarphi}$ and $\pi_{\vvarphi_k}$ are similar, whereas when $w_k~\ll~1$ or $w_k~\gg~1$, the policies differ significantly.
To be able to assume that the fixed critic is a good approximation within a certain ``trust region'' (i.e., one iteration), we impose the constraint \begin{align*}
  \E*[\vx \sim \rho_{\vvarphi_k}^\infty]{\KL{\pi_{\vvarphi_k}(\cdot \mid \vx)}{\pi_\vvarphi(\cdot \mid \vx)}} \leq \delta
\end{align*} to optimize only in the ``neighborhood'' of the current policy.
This constraint is also necessary for the importance weights not to blow up.

\begin{rmk}{Estimating the KL-divergence}{}
  Instead of naive computation with $\E[\va \sim \pi_{\vvarphi_k}(\cdot \mid \vx)]{- \log w_k(\vvarphi; \vx, \va)}$, the KL-divergence is commonly estimated by Monte Carlo samples of \begin{align*}
    \begin{multlined}[t]
      \KL{\pi_{\vvarphi_k}(\cdot \mid \vx)}{\pi_\vvarphi(\cdot \mid \vx)} \\ = \E[\va \sim \pi_{\vvarphi_k}(\cdot \mid \vx)]{w_k(\vvarphi; \vx, \va) - 1 - \log w_k(\vvarphi; \vx, \va)},
    \end{multlined}
  \end{align*} which adds the ``baseline'' $w_k(\vvarphi; \vx, \va) - 1$ with mean $0$.
  Observe that this estimator is unbiased, always non-negative since $\log(x) \leq x - 1$ for all $x$, while having a lower variance than the naive estimator.
\end{rmk}

Taking the expectation with respect to the previous policy $\pi_{\vvarphi_k}$ means that we can reuse data from rollouts within the same iteration.
That is, TRPO allows reusing past data as long as it can still be ``trusted''.
This makes TRPO ``somewhat'' off-policy.
Fundamentally, though, TRPO is still an on-policy method.

\midx<Proximal policy optimization>{proximal policy optimization} (PPO) is a family of heuristic variants of TRPO which replace the constrained optimization problem of \cref{eq:trpo} by the unconstrained optimization of a regularized objective \citep{schulman2017proximal,wang2020truly}.
PPO algorithms often work well in practice.
One canonical PPO method uses the modified objective \begin{align}
  \vvarphi_{k+1} \gets \argmax_{\vvarphi} \J{\vvarphi} - \lambda \E*[\vx \sim \rho_{\vvarphi_k}^\infty]{\KL{\pi_{\vvarphi_k}(\cdot \mid \vx)}{\pi_\vvarphi(\cdot \mid \vx)}} \label{eq:ppo}
\end{align} with some $\lambda > 0$, which regularizes towards the trust region.
Another common variant of PPO is based on controlling the importance weights directly rather than regularizing by the KL-divergence.
PPO is used, for example, to train large-scale language models such as GPT \citep{stiennon2020learning,openai2023gpt4} which we will discuss in more detail in \cref{sec:mfarl:preference_learning}.
There we will also see that the objective from \cref{eq:ppo} can be cast as performing probabilistic inference.

\paragraph{Improving computational efficiency:}
A practical problem with the above methods is that the estimation of the advantage function $\fnA(\vx, \va; \vtheta)$ requires training a separate critic, next to the policy (i.e., the actor) parameterized by $\vvarphi$.
This can be computationally expensive.
In particular, when both models are large neural networks (think multiple billions of parameters each), training both models is computationally prohibitive.
Now, recall that we introduced critics in the first place to reduce the variance of the policy gradient estimates.\footnote{That is, we moved from the REINFORCE policy update~\eqref{eq:reinforce} to the actor-critic policy update~\eqref{eq:online_actor_critic1}.}
\midx<Group relative policy optimization>{group relative policy optimization} (GRPO) replaces the critic in PPO with simple Monte Carlo estimates of the advantage function~\citep{shao2024deepseekmath}: \begin{align}
  \begin{multlined}[t]
    \J{\vvarphi} \defeq \E[\{\tau^{(i)}\}_{i=1}^m \sim \altpi_{\vvarphi_k}(\cdot \mid \vx)]{\frac{1}{m} \sum_{i=1}^m \sum_{t=1}^T w_k(\vvarphi; \va^{(i)}) \fnAhat_{t,i}^{\pi_{\vvarphi_k}}}, \\
    \text{where}\quad \fnAhat_{t,i}^{\pi_{\vvarphi_k}} \defeq \text{\small$\frac{g_{t:T}^{(i)} - \mathrm{mean}(\{\tau^{(i)}\})}{\mathrm{std}(\{\tau^{(i)}\})}$}
  \end{multlined}
\end{align} estimates the advantage of action $\va^{(i)}$ at time $t$ by comparing to the mean reward and normalizing by the standard deviation of rewards from all trajectories $\tau^{(i)}$.\footnote{In \cref{eq:score_gradient_estimator_approx}, we have already seen that using a Monte Carlo estimate of returns is a simple approach to reduce variance without needing to learn a critic.}
GRPO combines Monte Carlo sampling and baselines for variance reduction with the trust-region optimization of PPO, leading to a method that is more sample efficient than naive REINFORCE while being computationally more efficient than PPO.

\section{Off-policy Actor-Critics}\label{sec:mfarl:off_policy_actor_critic_methods}

In many applications, sample efficiency is crucial.
Either because requiring too many interactions is computationally prohibitive or because obtaining sufficiently many samples for learning a near-optimal policy is simply impossible.
We therefore now want to look at a separate family of actor-critic methods, which are off-policy, and hence, allow for the reuse of past data. These algorithms use the reparameterization gradient estimates, which we encountered before in the context of variational inference,\footnote{see \cref{sec:approximate_inference:variational_inference:gradient_of_elbo}} instead of score gradient estimators.

The on-policy methods that we discussed in the previous section can be understood as performing a variant of \midx{policy iteration}, where we use an estimate of the state-action value function of the current policy and then try to improve that policy by acting greedily with respect to this estimate.
They mostly vary in how improving the policy is traded with improving the estimate of its value.
Fundamentally, these methods rely on policy evaluation.\footnote{Policy evaluation is at the core of policy iteration. See \cref{alg:policy_iteration} for the definition of policy iteration and \cref{sec:mdp:policy_evaluation} for a summary of policy evaluation in the context of Markov decision processes.}

The techniques that we will introduce in this section are much more closely related to value iteration, essentially making use of Bellman's optimality principle to learn the optimal value function directly which characterizes the optimal policy.

To begin with, let us assume that the policy $\vpi$ is deterministic.
We will later lift this restriction in \cref{sec:mfarl:actor_critic_methods:randomized_policies}.
Recall that our initial motivation to consider policy gradient methods and then actor-critic methods was the intractability of the DQN loss \begin{align*}
  \ell_{\mathrm{DQN}}(\vtheta; \spD) = \frac{1}{2} \sum_{(\vx, \va, r, \vxp) \in \spD} \parentheses*{r + \gamma \max_{\vap \in \spA} \Q*{\vxp}{\vap; \old{\vtheta}} - \Q*{\vx}{\va; \vtheta}}^2 \margintag{see \cref{eq:dqn}}
\end{align*} when the action space $\spA$ is large.
What if we simply replace the exact maximum over actions by a parameterized policy? \begin{align}
  \ell_{\mathrm{DQN}}(\vtheta; \spD) \approx \frac{1}{2} \sum_{(\vx, \va, r, \vxp) \in \spD} \parentheses*{r + \gamma \Q*{\vxp}{\vpi_{\vvarphi}(\vxp); \old{\vtheta}} - \Q*{\vx}{\va; \vtheta}}^2.
\end{align}
We want to train our parameterized policy to learn the maximization over actions, that is, to approximate the greedy policy\footnote{Here, we already apply the improvement of DDQN to use the most-recent estimate of the Q-function for action selection (see \cref{eq:ddqn}).} \begin{align}
  \vpi_{\vvarphi}(\vx) \approx \opt{\vpi}(\vx) = \argmax_{\va \in \spA} \Q*{\vx}{\va; \vtheta}.
\end{align}
The key idea is that if we use a ``rich-enough'' parameterization of policies, selecting the greedy policy with respect to $\opt{\fnQ}$ is equivalent to \begin{align}
  \opt{\vvarphi} &= \argmax_{\vvarphi} \E[\vx \sim \mu]{\Q*{\vx}{\vpi_\vvarphi(\vx); \vtheta}} \label{eq:off_policy_actor_critics_optimization}
\end{align} where $\mu(\vx) > 0$ is an \midx{exploration distribution} over states with full support.\safefootnote{We require full support to ensure that all states are explored.}
We refer to this expectation by \begin{align}
  \J{\vvarphi; \vtheta}[\mu] \defeq \E[\vx \sim \mu]{\Q*{\vx}{\vpi_\vvarphi(\vx); \vtheta}}.
\end{align}
Commonly, the exploration distribution $\mu$ is taken to be the distribution that samples states uniformly from a replay buffer.
Note that we can easily obtain unbiased gradient estimates of $\fnJ[\mu]$ with respect to $\vvarphi$: \begin{align}
  \grad_{\vvarphi} \J{\vvarphi; \vtheta}[\mu] &= \E[\vx \sim \mu]{\grad_{\vvarphi} \Q*{\vx}{\vpi_\vvarphi(\vx); \vtheta}} \margintag{see \cref{sec:background:probability:gradients_of_expectations}}.
\end{align}

Analogously to on-policy actor-critics (see \cref{eq:policy_gradient_bootstrapping_estimate}), we use a bootstrapping estimate of $\opt{\fnQ}$.
That is, we neglect the dependence of the critic $\opt{\fnQ}$ on the actor $\vpi_\vvarphi$, and in particular, the policy parameters $\vvarphi$.
We have seen that bootstrapping works with Q-learning, so there is reason to hope that it will work in this context too.
This then allows us to use the chain rule to compute the gradient, \begin{align}
  \grad_{\vvarphi} \Q*{\vx}{\vpi_\vvarphi(\vx); \vtheta} = \jac_{\vvarphi} \vpi_\vvarphi(\vx) \cdot \grad_\va \Q*{\vx}{\va; \vtheta}\big\rvert_{\va=\vpi_\vvarphi(\vx)}. \label{eq:off_policy_acator_critic_actor_gradient}
\end{align}
This corresponds to evaluating the bootstrapping estimate of the Q-function at $\vpi_\vvarphi(\vx)$ and obtaining a gradient estimate of the policy estimate (e.g., through automatic differentiation).
Note that as $\vpi_\vvarphi$ is vector-valued, $\jac_{\vvarphi} \vpi_\vvarphi(\vx)$ is the Jacobian of $\vpi_\vvarphi$ evaluated at $\vx$.

\paragraph{Exploration:}

Now that we have estimates of the gradient of our optimization target $\fnJ[\mu]$, it is natural to ask how we should select actions (based on $\vpi_\vvarphi$) to trade exploration and exploitation.
As we have seen, policy gradient techniques rely on the randomness in the policy to explore, but here we consider deterministic policies.
As our method is off-policy, a simple idea in continuous action spaces is to add Gaussian noise to the action selected by $\vpi_\vvarphi$ --- also known as \midx{Gaussian noise ``dithering''}.\footnote{Intuitively, this adds ``additional randomness'' to the policy $\vpi_\vvarphi$.}
This corresponds to an algorithm called \emph{deep deterministic policy gradients} \citep{lillicrap2015continuous} shown in \cref{alg:ddpg}.
This algorithm is essentially equivalent to Q-learning with function approximation (e.g., DQN),\footnote{see \cref{eq:dqn}} with the only exception that we replace the maximization over actions with the learned policy $\vpi_\vvarphi$.

\begin{algorithm}
  \caption{Deep deterministic policy gradients, DDPG}\pidx{deep deterministic policy gradients}[idxpagebf]\label{alg:ddpg}
  initialize $\vvarphi, \vtheta$, a (possibly non-empty) replay buffer $\spD = \emptyset$\;
  set $\old{\vvarphi} = \vvarphi$ and $\old{\vtheta} = \vtheta$\;
  \For{$t=0$ \KwTo $\infty$}{
    observe state $\vx$, pick action $\va = \vpi_\vvarphi(\vx) + \vvarepsilon$ for $\vvarepsilon \sim \N{\vzero, \lambda \mI}$\;
    execute $\va$, observe $r$ and $\vxp$\;
    add $(\vx, \va, r, \vxp)$ to the replay buffer $\spD$\;
    \If{collected ``enough'' data}{
      \Comment{policy improvement step}
      \For{some iterations}{
        sample a mini-batch $\sB$ of $\spD$\;
        for each transition in $\sB$, compute the label $y = r + \gamma \Q*{\vxp}{\vpi(\vxp; \old{\vvarphi}); \old{\vtheta}}$\;
        \Comment{critic update}
        $\vtheta \gets \vtheta - \eta \grad_{\vtheta} \frac{1}{B} \sum_{(\vx, \va, r, \vxp, y) \in B} (y - \Q*{\vx}{\va; \vtheta})^2$\;
        \Comment{actor update}
        $\vvarphi \gets \vvarphi + \eta \grad_{\vvarphi} \frac{1}{B} \sum_{(\vx, \va, r, \vxp, y) \in B} \Q*{\vx}{\vpi(\vx; \vvarphi); \vtheta}$\;
        $\old{\vtheta} \gets (1 - \rho)\old{\vtheta} + \rho\vtheta$\;
        $\old{\vvarphi} \gets (1 - \rho)\old{\vvarphi} + \rho\vvarphi$\;
      }
    }
  }
\end{algorithm}

\midx<Twin delayed DDPG>{twin delayed DDPG} (TD3) is an extension of DDPG that uses two separate critic networks for predicting the maximum action and evaluating the policy \citep{fujimoto2018addressing}.
This addresses the maximization bias akin to Double-DQN. TD3 also applies delayed updates to the actor network, which increases stability.

\subsection{Randomized Policies}\label{sec:mfarl:actor_critic_methods:randomized_policies}

We have seen that randomized policies naturally encourage exploration.
With deterministic actor-critic methods like DDPG, we had to inject Gaussian noise to enforce sufficient exploration.
A natural question is therefore whether we can also handle randomized policies in this framework of off-policy actor-critics.

The key idea is to replace the squared loss of the critic, \begin{align*}
  \ell_{\mathrm{DQN}}(\vtheta; \spD) \approx \frac{1}{2} \parentheses*{r + \gamma \Q*{\vxp}{\vpi(\vxp; \vvarphi); \old{\vtheta}} - \Q*{\vx}{\va; \vtheta}}^2, \margintag{refer to the squared loss of Q-learning \eqref{eq:q_loss}}
\end{align*} which only considers the fixed action $\vpi(\vxp; \old{\vvarphi})$ with an expected squared loss, \begin{align}
  \ell_{\mathrm{DQN}}(\vtheta; \spD) \approx \E[\vap \sim \pi(\vxp; \vvarphi)]{\frac{1}{2} \parentheses*{r + \gamma \Q*{\vxp}{\vap; \old{\vtheta}} - \Q*{\vx}{\va;\vtheta}}^2},
\end{align} which considers a distribution over actions.

It turns out that we can still compute gradients of this expectation.
\begin{align}
  &\grad_{\vtheta} \E[\vap \sim \pi(\vxp; \vvarphi)]{\frac{1}{2} \parentheses*{r + \gamma \Q*{\vxp}{\vap; \old{\vtheta}} - \Q*{\vx}{\va;\vtheta}}^2} \nonumber \\
  &= \E[\vap \sim \pi(\vxp; \vvarphi)]{\grad_{\vtheta} \frac{1}{2} \parentheses*{r + \gamma \Q*{\vxp}{\vap; \old{\vtheta}} - \Q*{\vx}{\va;\vtheta}}^2}. \margintag{see \cref{sec:background:probability:gradients_of_expectations}} \nonumber
  \intertext{Similarly to our definition of the Bellman error \eqref{eq:bellman_error}, we define by \begin{align}
    \delta_\mathrm{B}(\vap) \defeq r + \gamma \Q*{\vxp}{\vap; \old{\vtheta}} - \Q*{\vx}{\va;\vtheta},
  \end{align} the \midx{Bellman error} for a fixed action $\vap$. Using the chain rule, we obtain}
  &= \E[\vap \sim \pi(\vxp; \vvarphi)]{\delta_\mathrm{B}(\vap) \grad_{\vtheta} \Q*{\vx}{\va;\vtheta}}. \label{eq:svg_critic}
\end{align}
Note that this is identical to the gradient in DQN \eqref{eq:q_learning_gradient_update_general}, except that now we have an expectation over actions.
As we have done many times already, we can use automatic differentiation to obtain gradient estimates of $\grad_{\vtheta} \Q*{\vx}{\va;\vtheta}$.
This provides us with a method of obtaining unbiased gradient estimates for the critic.

We also need to reconsider the actor update.
When using a randomized policy, the objective function changes to \begin{align*}
  \J{\vvarphi; \vtheta}[\mu] \defeq \E*[\vx \sim \mu]{\E[\va \sim \pi(\vx; \vvarphi)]{\Q*{\vx}{\va; \vtheta}}}.
\end{align*} of which we can obtain gradients via \begin{align}
  \grad_{\vvarphi} \J{\vvarphi; \vtheta}[\mu] = \E*[\vx \sim \mu]{\grad_{\vvarphi} \E[\va \sim \pi(\vx; \vvarphi)]{\Q*{\vx}{\va; \vtheta}}}. \margintag{see \cref{sec:background:probability:gradients_of_expectations}}
\end{align}
Note that the inner expectation is with respect to a measure that depends on the parameters $\vvarphi$, which we are trying to optimize.
We therefore cannot move the gradient operator inside the expectation.
This is a problem that we have already encountered several times.
In the previous section on policy gradients, we used the score gradient estimator.\footnote{see \cref{eq:score_gradient_estimator}}
Earlier, in \cref{sec:approximate_inference} on variational inference we have already seen reparameterization gradients.\footnote{see \eqref{eq:reparameterization_trick2}}
Here, if our policy is reparameterizable, we can use the \midx{reparameterization trick} from \cref{thm:reparameterization_trick}!

\begin{ex}{Reparameterization gradients for Gaussians}{reparameterization_gradients_gaussian}
  Suppose we use a Gaussian parameterization of policies, \begin{align*}
    \pi(\vx; \vvarphi) \defeq \N{\vmu(\vx; \vvarphi)}{\mSigma(\vx; \vvarphi)}.
  \end{align*}
  Then, using conditional linear Gaussians, our action $\va$ is given by \begin{align}
    \va = \vg(\vvarepsilon; \vx, \vvarphi) \defeq \msqrt{\mSigma}(\vx; \vvarphi) \vvarepsilon + \vmu(\vx; \vvarphi), \quad \vvarepsilon \sim \SN
  \end{align} where $\msqrt{\mSigma}(\vx; \vvarphi)$ is the square root of $\mSigma(\vx; \vvarphi)$.
  This coincides with our earlier application of the reparameterization trick to Gaussians in \cref{ex:reparameterization_trick_gaussian}.
\end{ex}

As we have seen, not only Gaussians are reparameterizable.
In general, we called a distribution (in this context, a policy) reparameterizable iff $\va \sim \pi(\vx; \vvarphi)$ is such that $\va = \vg(\vvarepsilon; \vx, \vvarphi)$, where $\vvarepsilon \sim \phi$ is an independent random variable.

Then, we have, \begin{align}
  &\grad_{\vvarphi} \E[\va \sim \pi(\vx; \vvarphi)]{\Q*{\vx}{\va; \vtheta}} \nonumber \\
  &= \E[\vvarepsilon \sim \phi]{\grad_{\vvarphi} \Q*{\vx}{\vg(\vvarepsilon; \vx, \vvarphi); \vtheta}} \label{eq:reparameterization_gradients} \margintag{using the reparameterization trick \eqref{eq:reparameterization_trick2}} \\
  &= \E[\vvarepsilon \sim \phi]{\grad_\va \Q*{\vx}{\va; \vtheta}\big\rvert_{\va=\vg(\vvarepsilon; \vx, \vvarphi)} \cdot \jac_{\vvarphi} \vg(\vvarepsilon; \vx, \vvarphi)}. \margintag{using the chain rule analogously to \cref{eq:off_policy_acator_critic_actor_gradient}}
\end{align}
In this way, we can obtain unbiased gradient estimates for reparameterizable policies.
This general technique does not only apply to continuous action spaces.
For discrete action spaces, there is the analogous so-called \midx{Gumbel-max trick}, which we will not discuss in greater detail here.\looseness=-1

The algorithm that uses \cref{eq:svg_critic} to obtain gradients for the critic and reparameterization gradients for the actor is called \midx{stochastic value gradients}[idxpagebf] (SVG) \citep{heess2015learning}.

\section{Maximum Entropy Reinforcement Learning}\label{sec:mfarl:actor_critic_methods:entropy_regularization}

In practice, algorithms like SVG often do not explore enough.
A key issue with relying on randomized policies for exploration is that they might collapse to deterministic policies.
That is, the algorithm might quickly reach a local optimum, where all mass is placed on a single action.\looseness=-1

A simple trick that encourages a little bit of extra exploration is to regularize the randomized policies ``away'' from putting all mass on a single action.
In other words, we want to encourage the policies to exhibit some uncertainty.
A natural measure of uncertainty is entropy, which we have already seen several times.\footnote{see \cref{sec:approximate_inference:information_theory}}
This approach is known as \midx{entropy regularization}[idxpagebf] or \midx{maximum entropy reinforcement learning} (MERL).
Canonically, entropy regularization is applied to finite-horizon rewards (cf. \cref{rmk:other_reward_models}), yielding the optimization problem of maximizing \begin{align}
  \j{\vvarphi}[\lambda] &\defeq \j{\vvarphi} + \lambda \H{\altpi_\vvarphi} \\
  &= \sum_{t=1}^{T} \E[(\vx_t,\va_t) \sim \altpi_\vvarphi]{r(\vx_t, \va_t) + \lambda \H{\pi_\vvarphi(\cdot \mid \vx_t)}}, \label{eq:entropy_reg_rl}
\end{align} where we have a preference for entropy in the actor distribution to encourage exploration which is regulated by the temperature parameter $\lambda$.
As $\lambda \to 0$, we recover the ``standard'' reinforcement learning objective (here for finite-horizon rewards): \begin{align}
  \j{\vvarphi} = \sum_{t=1}^{T} \E[(\vx_t,\va_t) \sim \altpi_\vvarphi]{r(\vx_t, \va_t)}.
\end{align}
Here, for notational convenience, we begin the sum with $t=1$ rather than $t=0$.

\subsection{Entropy Regularization as Probabilistic Inference}\label{sec:mfarl:entropy_regularization_as_probabilistic_inference}

The entropy-regularized objective from \cref{eq:entropy_reg_rl} leads us to a remarkable interpretation of reinforcement learning and, more generally, decision-making under uncertainty as solving an inference problem akin to variational inference.
The framing of ``control as inference'' will lead us to contemporary algorithms for reinforcement learning as well as paint a path for decision-making under uncertainty beyond stationary MDPs.

\begin{marginfigure}
  \incfig{merl}
  \caption{Directed graphical model of the underlying hidden Markov model with hidden states $X_t$, optimality variables $O_t$, and actions $A_t$.}
\end{marginfigure}

Let us denote by $\altpi_\star$ the distribution over trajectories $\tau$ under the optimal policy $\pis$.
By framing the problem of optimal control as an inference problem in a hidden Markov model with hidden ``optimality variables'' $O_t \in \{0, 1\}$ indicating whether the played action $\va_t$ was optimal we can derive $\altpi_\star$ analytically.
That is to say, when $O_t = 1$ and $O_{t+1:T} \equiv 1$ the policy from time $t$ onwards was optimal.
To simplify the notation, we will denote the event $O_t = 1$ by $\spO_t$.

We consider the HMM defined by the Gibbs distribution \begin{align}
  p(\spO_t \mid \vx_t, \va_t) \propto \exp\parentheses*{\frac{1}{\lambda} r(\vx_t, \va_t)}, \quad\text{with $\lambda > 0$} \label{eq:merl_optimality}
\end{align} which is a natural choice as we have seen in \cref{exercise:maximum_entropy_property_of_gibbs_distribution} that the Gibbs distribution maximizes entropy subject to $\E{O_t \cdot r(\vx_t, \va_t)}[\vx_t, \va_t]~<~\infty$.

The distribution over trajectories conditioned on optimality of actions~(i.e., conditioned on $\spO_{1:T}$) is given by \begin{align}
  \altpi_\star(\tau) \defeq p(\tau \mid \spO_{1:T}) = p(\vx_1) \prod_{t=1}^{T} p(\va_t \mid \vx_t, \spO_t) p(\vx_{t+1} \mid \vx_t, \va_t). \margintag{Using \cref{eq:trajectory_distr}. We assume here that the dynamics and initial state distribution are ``fixed'', that is, we assume $p(\vx_1 \mid \spO_{1:T}) = p(\vx_1)$ and $p(\vx_{t+1} \mid \vx_t, \va_t, \spO_{1:T}) = p(\vx_{t+1} \mid \vx_t, \va_t)$.} \label{eq:distr_over_optimal_trajectories}
\end{align}
It remains to determine $p(\va_t \mid \vx_t, \spO_t)$ which corresponds to the optimal policy $\opt{\pi}(\va_t \mid \vx_t)$.
It is generally useful to think of the situation where the prior policy $p(\va_t \mid \vx_t)$ is uniform on $\spA$,\footnote{This is not a restriction as any informative prior can be pushed into \cref{eq:merl_optimality}.} in which case by Bayes' rule \eqref{eq:bayes_rule}, $p(\va_t \mid \vx_t, \spO_t) \propto p(\spO_t \mid \vx_t, \va_t)$, so \begin{align}
  \altpi_\star(\tau) &\propto \brackets*{p(\vx_1) \prod_{t=1}^{T} p(\vx_{t+1} \mid \vx_t, \va_t)} \exp\parentheses*{\frac{1}{\lambda} \sum_{t=1}^{T} r(\vx_t, \va_t)}. \label{eq:opt_trajectory_distr}
\end{align}

Recall that our fundamental goal is to approximate $\altpi_\star$ with a distribution over trajectories $\altpi_\vvarphi$ under the parameterized policy $\pi_\vvarphi$.
It is therefore a natural idea to minimize $\KL{\altpi_\vvarphi}{\altpi_\star}$:\safefootnote[-2\baselineskip]{Observe that we cannot easily minimize forward-KL as we cannot sample from $\altpi_\star$. In the context of RL, it can be argued that the mode-seeking behavior of reverse-KL is preferable over the moment-matching behavior of forward-KL \citep{levine2018reinforcement}.} \begin{align}
  &\argmin_\vvarphi \KL{\altpi_\vvarphi}{\altpi_\star} \nonumber \\
  &= \argmin_\vvarphi \crH{\altpi_\vvarphi}{\altpi_\star} - \H{\altpi_\vvarphi} \margintag{using the definition of KL-divergence \eqref{eq:kl}} \nonumber \\
  &= \argmax_\vvarphi \E[\tau \sim \altpi_\vvarphi]{\log \altpi_\star(\tau) - \log \altpi_\vvarphi(\tau)} \margintag{using the definition of cross-entropy \eqref{eq:cross_entropy} and entropy \eqref{eq:entropy}} \nonumber \\
  &= \argmax_\vvarphi \E[\tau \sim \altpi_\vvarphi]{\sum_{t=1}^{T} r(\vx_t, \va_t) - \lambda \log \pi_\vvarphi(\va_t \mid \vx_t)} \margintag{using \cref{eq:opt_trajectory_distr,eq:trajectory_distr} and simplifying} \nonumber \\
  &= \argmax_\vvarphi \sum_{t=1}^{T} \E[(\vx_t, \va_t) \sim \altpi_\vvarphi]{r(\vx_t, \va_t) + \lambda \H{\pi_\vvarphi(\cdot \mid \vx_t)}}. \margintag{using the definition of entropy \eqref{eq:entropy} and linearity of expectation \eqref{eq:linearity_expectation}} \label{eq:entropy_reg_rl2}
\end{align}
That is, entropy regularization is equivalent to minimizing the KL-divergence from $\altpi_\star$ to $\altpi_\vvarphi$.
This highlights a very natural tradeoff between exploration and exploitation, wherein $\crH{\altpi_\vvarphi}{\altpi_\star}$ encourages exploitation and $\H{\altpi_\vvarphi}$ encourages exploration.

It can be shown that a ``softmax'' version of the Bellman optimality equation \eqref{eq:bop2_q} can be obtained for \cref{eq:entropy_reg_rl2} \exerciserefmark{soft_value_function}: \begin{align}
  \q*{\vx}{\va} = \frac{1}{\lambda} r(\vx, \va) + \E[\vxp \sim \vx, \va]{\log \int_{\spA} \exp\parentheses*{\q*{\vxp}{\vap}} \,d \vap} \label{eq:soft_value_function}
\end{align} with the convention that $\q*{\vx_T}{\va} = 0$ for all $\va$.\footnote{Note that \cref{eq:bop2_q} was derived in the infinite horizon setting with discounted rewards, whereas here we study the finite horizon setting. \Cref{eq:soft_value_function} is the natural extension of the standard Bellman optimality equation in the finite horizon setting where the downstream rewards are measured by a softmax rather than the greedy policy.}
Here, $\opt{\fnq}$ is called a \midx{soft value function}.
As we will see in \cref{exercise:soft_value_function}, the optimal policy has the form $\pis(\va \mid \vx) \propto \exp(\q*{\vx}{\va})$, that is, it simply corresponds to performing softmax exploration \eqref{eq:softmax_exploration} with the soft value function.
The second term of \cref{eq:soft_value_function} quantifies downstream rewards.
In comparison to the ``standard'' Bellman optimality equation \eqref{eq:bop2_q}, the soft value function is less greedy which tends to encourage robustness.

Analogously to Q-learning, the soft value function $\opt{\fnq}$ can be approximated via a bootstrapped ``critic'' $\opt{\fnQ}$ which is called \midx{soft Q-learning} \citep{levine2018reinforcement}.
Note that computing the optimal policy requires computing an integral over the action space, which is typically intractable for continuous action spaces.
As discussed in \cref{sec:mfarl:policy_approximation,sec:mfarl:actor_critic_methods,sec:mfarl:off_policy_actor_critic_methods} and analogously to actor-critic methods such as DDPG and SVG, we can learn a parameterized policy (i.e., an ``actor'') $\pi_\vvarphi$ to approximate the optimal policy $\pis$.
The resulting algorithm, \midx{soft actor critic}[idxpagebf] (SAC) \citep{haarnoja2018soft,haarnoja2018soft2}, is widely used.
Due to its off-policy nature, it is also relatively sample efficient.

\begin{figure*}
  \includegraphics[width=\textwidth]{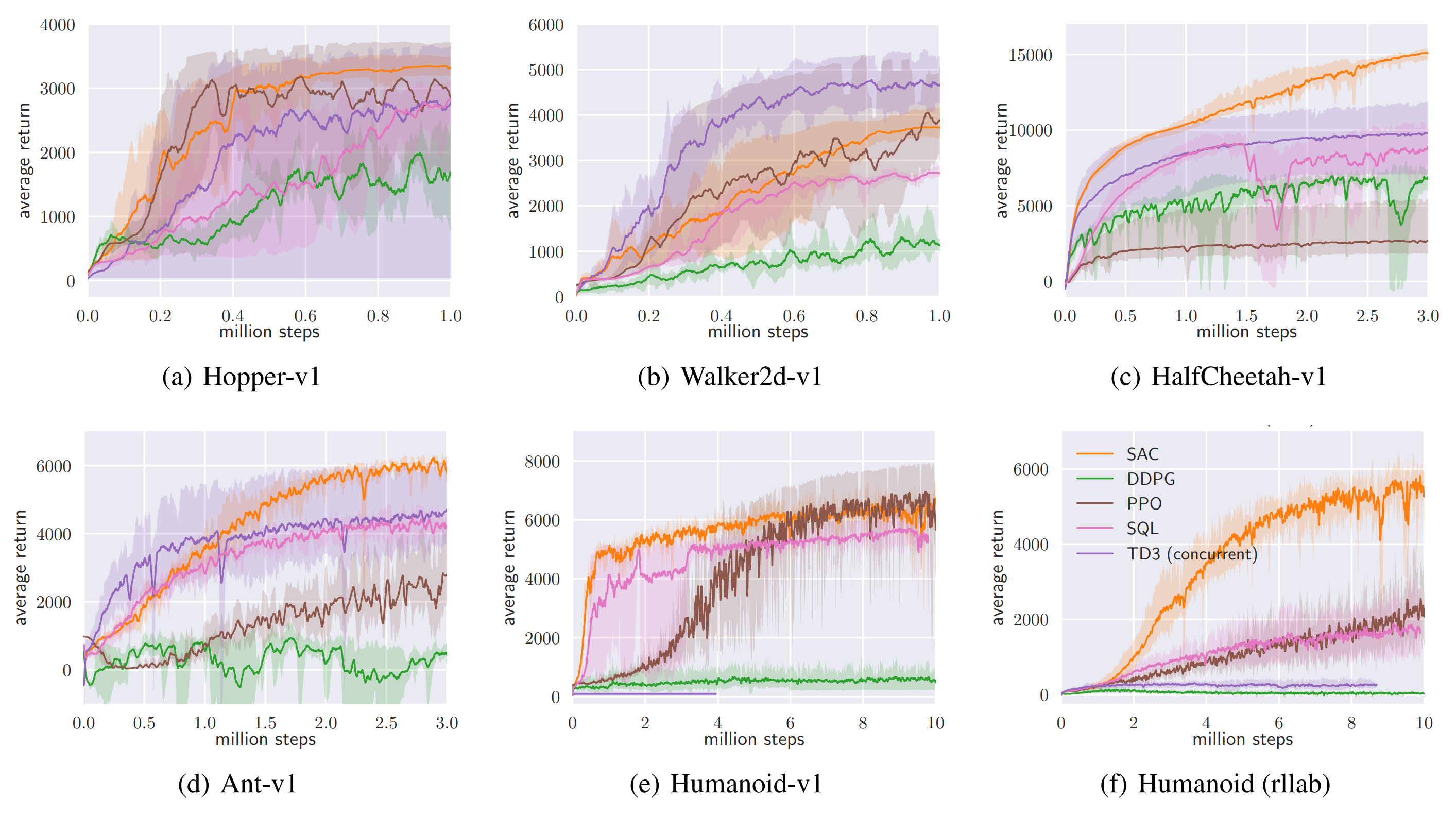}
  \caption{Comparison of training curves of a selection of on-policy and off-policy policy gradient methods.
  Reproduced with permission from \icite{haarnoja2018soft2}.}
\end{figure*}

We can also express this optimization in terms of an evidence lower bound.
The evidence lower bound for the observations $\spO_{1:T}$ is \begin{align}
  L(\altpi_\vvarphi, \altpi_\star; \spO_{1:T}) &= \E[\tau \sim \altpi_\vvarphi]{\log p(\spO_{1:T} \mid \tau) + \log \altpi(\tau) - \log \altpi_\vvarphi(\tau)} \margintag{using the definition of the ELBO \eqref{eq:elbo_expanded}}
\end{align} where $\altpi$ denotes the distribution over trajectories \eqref{eq:trajectory_distr} under the prior policy $p(\va_t \mid \vx_t)$.
This is commonly written as the variational free energy \begin{align}
  - L(\altpi_\vvarphi, \altpi_\star; \spO_{1:T}) &= \E[\tau \sim \altpi_\vvarphi]{\S{p(\spO_{1:T} \mid \tau)}} + \KL{\altpi_\vvarphi}{\altpi} \\
  &= \underbrace{\S{p(\spO_{1:T})}}_{\text{``extrinsic'' value}} + \underbrace{\KL{\altpi_\vvarphi}{\altpi_\star}}_{\text{``epistemic'' value}}. \margintag{using that $\altpi_\star(\tau) = p(\tau \mid \spO_{1:T})$}
\end{align} which we already encountered in \cref{sec:free_energy} in the context of variational inference.
Here, the ``extrinsic'' value is independent of the variational distribution $\altpi_\vvarphi$ and can be thought of as a fixed ``problem cost'', whereas the ``epistemic'' value can be interpreted as the approximation error or ``solution cost''.
To summarize, we have seen that \begin{align*}
  \argmax_\vvarphi \j{\vvarphi}[\lambda] = \argmin_\vvarphi \KL{\altpi_\vvarphi}{\altpi_\star} = \argmax_\vvarphi L(\altpi_\vvarphi, \altpi_\star; \spO_{1:T}).
\end{align*}

Recall that the free energy $-L(\altpi_\vvarphi, \altpi_\star; \spO_{1:T})$ is a variational upper bound to the surprise about observations $\S{\altpi_\star(\spO_{1:T})}$ when following an optimal policy.\footnote{see \cref{sec:free_energy}}
For example, undesirable states incur a low reward while desirable states yield a high reward, and thus, if we expect optimality of actions (i.e., $\spO_{1:T}$) paths leading to such states have high and low surprise, respectively.\safefootnote{This is because, a path leading to a low reward state will include sub-optimal actions.}

An agent which acts to minimize free energy with $\spO_{1:T}$ can be thought of as hallucinating to perform optimally, and acting to minimize the surprise about having played suboptimal actions.
Think, for example, about the robotics task of moving an arm to a new position.
Intuitively, minimizing free energy solves this task by ``hallucinating that the arm is at the goal position'', and then minimizing the surprise with respect to this perturbed world model.
In this way, MERL can be understood as identifying paths of least surprise akin to the \midx{free energy principle}.

\begin{rmk}{Towards active inference}{}
  Maximum entropy reinforcement learning makes one fundamental assumption, namely, that the ``biased'' distribution about observations specifying the underlying HMM is \begin{align*}
    p(\spO_t \mid \vx_t, \va_t) \propto \exp\parentheses*{\frac{1}{\lambda} r(\vx_t, \va_t)}.
  \end{align*}
  This assumption is key to the framing of optimal control (in an unknown MDP) with a certain reward function as an inference problem.
  One could conceive other HMMs.
  For example, \cite{fellows2019virel} propose an HMM defined in terms of the current value function of the agent: \begin{align*}
    p(\spO_t \mid \vx_t, \va_t) \propto \exp\parentheses*{\frac{1}{\lambda} \Q{\vx_t}{\va_t; \vtheta}}.
  \end{align*}

  Crucially, the choice of $p(\spO_t \mid \vx_t, \va_t)$ is the only place where the reward enters the inference problem, and one can conceive of settings where a ``stationary'' (i.e., time-independent) reward is not assumed to exist.
  The general approach to decision-making as probabilistic inference presented in this section (but for possibly reward-independent and non-stationary HMMs) is known as \midx{active inference} \citep{friston2015active,millidge2020relationship,millidge2021whence,parr2022active}.\looseness=-1
\end{rmk}

\section{Learning from Preferences}\label{sec:mfarl:preference_learning}

So far, we have been assuming that the agent is presented with a reward signal after every played action.
This is a natural assumption in domains such as games and robotics --- even though it often requires substantial ``reward engineering'' to break down complex tasks with sparse rewards to more manageable tasks (cf. \cref{sec:mbarl:exploration}).
In many other domains such as an agent learning to drive a car or a chatbot, it is unclear how one can even quantify the reward associated with an action or a sequence of actions.
For example, in the context of autonomous driving it is typically desired that agents behave ``human-like'' even though a different driving behavior may also reach the destination safely.

\begin{marginfigure}[-10\baselineskip]
  \incfig{learning_from_preferences}
  \caption{We generalize the perspective of reinforcement learning from \cref{fig:rl} by allowing the feedback to come from either the environment or an evaluation by other agents (e.g., humans), and by allowing the feedback to come in other forms than a numerical reward.}\label{fig:learning_from_preferences}
\end{marginfigure}

The task of ``aligning'' the behavior of an agent to human expectations is difficult in complex domains such as the physical world and language, yet crucial for their practical use.
To this end, one can conceive of alternative ways for presenting ``feedback'' to the agent: \begin{itemize}

  \item The classical feedback in reinforcement learning is a numerical score.
  Consider, for example, a recommender system for movies.
  The feedback is obtained after a movie was recommended to a user by a user-rating on a given scale (often $1$ to $10$).
  This rating is informative as it corresponds to an \emph{absolute value assessment}, allowing to place the recommendation in a complete ranking of all previous recommendations.
  However, numerical feedback of this type can be error-prone as it is scale-dependent (different users may ascribe different value to a recommendation rated a $7$).

  \item An alternative feedback mechanism is comparison-based.
  The user is presented with $k$ alternative actions and selects their preferred action (or alternatively returns a ranking of actions).
  This feedback is typically easy to obtain as humans are fast in making ``this-or-that'' decisions.
  However, in contrast to numerical rewards, the feedback provides information only on the user's \emph{relative preferences}.
  That is, such preference feedback encodes fewer bits of information than score feedback, and it therefore often takes longer to learn complex behavior from this feedback.
\end{itemize}

\begin{rmk}{Context-dependent feedback}{}
  We neglect here that feedback is often context-dependent \citep{lindner2022humans,casper2023open}.
  For example, if someone is asked whether they prefer ``ice cream'' over ``pizza'' the answer may depend on whether they are hungry and the weather.
\end{rmk}

\subsection{Language Models as Agents}

In the following, we discuss approaches to learning from preference feedback in the context of autoregressive\footnote{An autoregressive model predicts/generates the next ``token'' as a function of previous tokens.} large language models and chatbots.
A chatbot is an agent (often based on a transformer architecture, \cite{vaswani2017attention}) parameterized by $\vvarphi$ that given a \emph{prompt} $\vx$ returns a (stochastic) \emph{response} $\vy$.
The autoregressive generation of the response can be understood in terms of a policy $\pi_{\vvarphi}(y_{t+1} \mid \vx, y_{1:t})$ which generates the next token given the prompt and all previous tokens.\footnote{In large language models, a ``token'' is usually taken to be a letter, word, or something in-between. A special token is used to terminate the response.}
We denote the policy over complete responses (i.e., the chatbot) by\looseness=-1 \begin{align}
  \altpi_{\vvarphi}(\vy \mid \vx) = \prod_{t=0}^{T-1} \pi_{\vvarphi}(y_{t+1} \mid \vx, y_{1:t}). \label{eq:lm_trajectory_distr}
\end{align}
In RL jargon, the agents action corresponds to the choice of next token $y_{t+1}$ and the deterministic dynamics add this token to the current (incomplete) response $y_{1:t}$.
A full trajectory $\vy$ consists sequentially of all tokens $y_{1:T}$ comprising a response, and the prompt $\vx$ can be interpreted as a \midx{context} to this trajectory.
Observe that \cref{eq:lm_trajectory_distr} is derived from the general representation of the distribution over trajectories \eqref{eq:trajectory_distr}, noting that prior and dynamics are deterministic.

\begin{figure}
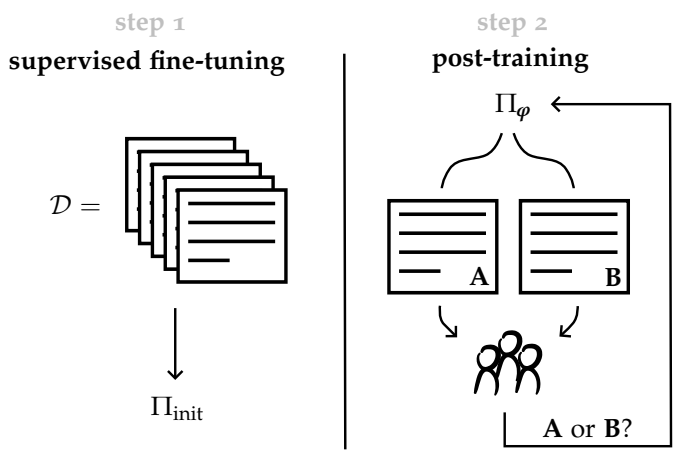

  \incfig{llm_post_training}
  \caption{Illustration of the learning pipeline of a large language model.}
  \label{fig:llm_learning_pipeline}
\end{figure}

The standard pipeline for applying pre-trained\footnote[][-3\baselineskip]{The pre-trained language model is usually obtained by self-supervised training on a large corpus of text. \midx<Self-supervised learning>{self-supervised learning} generates labeled training data from an unlabeled data source by selectively ``masking-out'' parts of the data. When training an autoregressive language model, labeled training data can be obtained by repeatedly ``masking-out'' the next word in a sentence, and training the language model to predict this word. Such large models that can be fine-tuned and ``post-trained'' to various downstream tasks are also called \midx<foundation models>{foundation model}.} large language models such as GPT \citep{openai2023gpt4} to downstream tasks consists of two main steps which are illustrated in \cref{fig:llm_learning_pipeline} \citep{stiennon2020learning}: (1) supervised fine-tuning and (2) post-training using preference feedback.\looseness=-1

The first step is to fine-tune the language model with supervised learning on high-quality data for the downstream task of interest.
For example, when the goal is to build a chatbot, this data may consist of desirable responses to some exemplary prompts.
We will denote the parameters of the fine-tuned language model by $\vvarphi^{\init}$, and its associated policy by $\altpi_{\init}$.

The second step is then to ``post-train'' the language model $\altpi_{\init}$ from the first step using human feedback.
Here, it is important that $\altpi_{\init}$ is already capable of producing sensible output (i.e., with correct spelling and grammar).
Learning this from scratch using only preference feedback would take far too long.
Instead, the post-training step is used to align the agent to the task and user preferences.

In each iteration of post-training, the model is prompted with prompts $\vx$ to produce pairs of answers $(\vy_A, \vy_B) \sim \altpi_{\vvarphi}(\cdot \mid \vx)$.
These answers are presented to human labelers who express their preference for one of the answers, denoted $\vy_A \succ \vy_B \mid \vx$.
A popular choice for modeling preferences is the \midx{Bradley-Terry model} which stipulates that the human preference distribution is given by \begin{align}
  p(\vy_A \succ \vy_B \mid \vx, r) &= \frac{\exp(r(\vy_A \mid \vx))}{\exp(r(\vy_A \mid \vx)) + \exp(r(\vy_B \mid \vx))}
  \intertext{for some unknown latent reward model $r(\vy \mid \vx)$ \citep{bradley1952rank}. This can be written in terms of the logistic function $\sigma$ \eqref{eq:logistic_function}:}
  &= \sigma(r(\vy_A \mid \vx) - r(\vy_B \mid \vx)). \margintag{as seen in \cref{exercise:softmax_and_logistic_function} this is the Gibb's distribution with energy $-r(\vy \mid \vx)$ in a binary classification problem} \label{eq:bradley_terry_model}
\end{align}

\begin{rmk}{Outcome rewards}{}
  Note that the Bradley-Terry model attributes reward only to ``complete'' responses.
  We call such a reward an \midx{outcome reward}.\safefootnote{Framing this in terms of an individual ``per-step'' rewards which we have seen so far, this corresponds to a (sparse) reward which is zero until the final action.}
  While the following discussion is on outcome rewards (which is most common in the context of language models), everything translates to individual per-step rewards.
\end{rmk}

The aggregated human feedback $\spD = \{\vy_A^{(i)} \succ \vy_B^{(i)} \mid \vx^{(i)}\}_{i=1}^n$ across $n$ different prompts is then used to update the language model $\pi_{\vvarphi}$.
In the next two sections, we discuss two standard approaches to post-training: reinforcement learning from human feedback (RLHF) and direct preference optimization (DPO).

\subsection{Reinforcement Learning from Human Feedback}\pidx{reinforcement learning from human feedback}%

RLHF separates the post-training step into two stages \citep{stiennon2020learning}.
First, the human feedback is used to learn an approximate reward model $r_{\vtheta}$.
This reward model is then used in the second stage to determine a refined policy $\altpi_{\vvarphi}$.

\begin{figure}
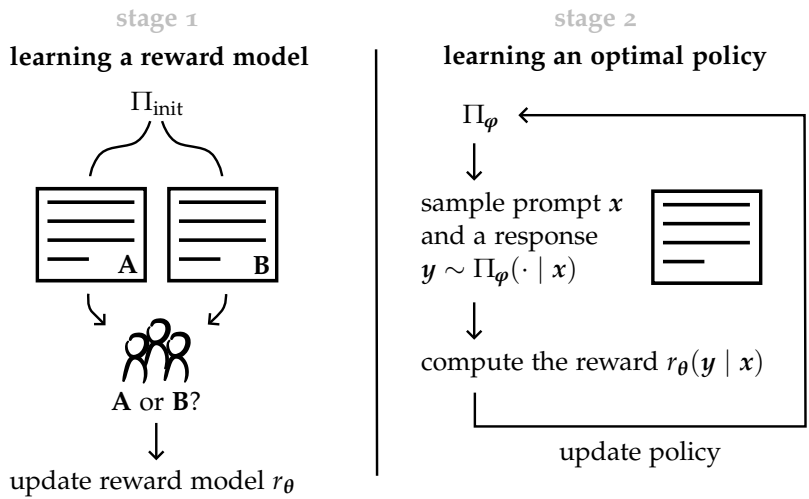

  \incfig{rlhf}
  \caption{Illustration of the post-training process of RLHF.}
  \label{fig:rlhf}
\end{figure}

\paragraph{Learning a reward model:}

During the first stage, the initial policy obtained by supervised fine-tuning $\altpi_{\init}$ is used to generate proposals~${(\vy_A, \vy_B)}$ for some exemplary prompts $\vx$, which are then ranked according to the preference of human labelers.
This preference data can be used to learn a reward model $r_{\vtheta}$ by maximum likelihood estimation (or equivalently minimizing cross-entropy): \begin{align}\begin{split}
  &\argmax_{\vtheta} p(\spD \mid r_\vtheta) \\
  &= \argmax_{\vtheta} \E[(\vy_A \succ \vy_B \mid \vx) \sim \spD]{\log \sigma\parentheses*{r_{\vtheta}(\vy_A \mid \vx) - r_{\vtheta}(\vy_B \mid \vx)}} \margintag{using \cref{eq:bradley_terry_model} and SGD} \label{eq:rlhf_reward_model}
\end{split}\end{align}%
This is analogous to the standard maximum likelihood estimation of the reward model in model-based RL with score feedback which we discussed in \cref{sec:rl:learning_mdp}.
The reward model $r_{\vtheta}$ is often initialized from the initial policy $\pi_{\init}$ by placing a linear layer producing a scalar output on top of the final transformer layer.

\paragraph{Learning an optimal policy:}

One can now employ the methods from this and previous chapters to determine the optimal policy for the approximate reward $r_{\vtheta}$.
Due to the use of an \emph{approximate} reward, however, simply maximizing $r_{\vtheta}$ surfaces the so-called ``reward gaming'' problem which is illustrated in \cref{fig:reward_gaming}.
As the responses generated by the learned policy $\pi_{\vvarphi}$ deviate from the distribution of $\spD$ (i.e., the distribution induced by the initial policy $\pi_{\init}$), the approximate reward model becomes inaccurate.
The approximate reward may severely overestimate the true reward in regions far away from the training data.
A common approach to address this problem is to regularize the policy search towards policies whose distribution does not stray away ``too far'' from the training distribution.

\begin{marginfigure}
  \begin{center}
    \import{./plots/output/}{reward_gaming.pgf}
  \end{center}

  \caption{Illustration of ``reward gaming''. Shown in black is the true reward $r(y \mid x)$ for a fixed prompt $x$. Shown in blue is the approximation based on the feedback $\spD$ to the responses shown in red. The yellow region symbolizes responses $y$ where the approximate reward can still be ``trusted''.}
  \label{fig:reward_gaming}
\end{marginfigure}

Analogously to the trust-region methods we discussed in \cref{sec:mfarl:actor_critc_methods:improved}, the deviation from the initial policy is typically controlled by maximizing the regularized objective \begin{align}
  \j{\vvarphi; \vvarphi^{\init} \mid \vx}[\lambda] &\defeq \E[\vy \sim \altpi_{\vvarphi}(\cdot \mid \vx)]{r(\vy \mid \vx)} - \lambda \KLsm{\altpi_{\vvarphi}(\cdot \mid \vx)}{\altpi_{\init}(\cdot \mid \vx)} \nonumber
  \intertext{in expectation over prompts $\vx$ sampled uniformly at random from the dataset $\spD$.
  Note that this coincides with the PPO objective from \cref{eq:ppo} with an outcome reward. We can expand the regularization term to obtain}
  &= \begin{multlined}[t]
    \underbrace{\E[\vy \sim \altpi_{\vvarphi}(\cdot \mid \vx)]{r(\vy \mid \vx)} + \lambda \Hsm{\altpi_{\vvarphi}(\cdot \mid \vx)}}_{\text{entropy-regularized RL}} \\ - \lambda \crHsm{\altpi_{\vvarphi}(\cdot \mid \vx)}{\altpi_{\init}(\cdot \mid \vx)}
  \end{multlined} \margintag{using the definition of KL-divergence \eqref{eq:kl}}
\end{align} which indicates an intimate relationship to entropy-regularized RL.\footnote{compare to \cref{eq:entropy_reg_rl}}

The optimal policy maximizing $\j{\vvarphi; \vvarphi^{\init} \mid \vx}[\lambda]$ is \exerciserefmark{ppo_as_probabilistic_inference} \begin{align}
  \altpi_\star(\vy \mid \vx) \propto \altpi_{\init}(\vy \mid \vx) \exp\parentheses*{\frac{1}{\lambda} r(\vy \mid \vx)} \label{eq:optimal_regularized_policy_rlhf}
\end{align} and can be interpreted as a probabilistic update to the prior $\altpi_{\init}$ where $\exp\parentheses{\frac{1}{\lambda} r(\vy \mid \vx)}$ corresponds to the ``likelihood of optimality''.
As $\lambda \to \infty$, $\altpi_\star \to \altpi_{\init}$, and as $\lambda \to 0$, the optimal policy reduces to deterministically picking the response with the highest reward.

In practice, sampling from $\altpi_\star$ explicitly is intractable, but note that any of the approximate inference methods discussed in \cref{part1} are applicable here.
In the context of chatbots, it is important that the sampling from the resulting policy is efficient (i.e., the chatbot should respond quickly to prompts).
Most commonly, the optimization problem of maximizing $j{\vvarphi; \vvarphi^{\init} \mid \vx}[\lambda]$ (with the estimated reward $r_{\vtheta}$) is solved approximately within a parameterized family of policies.
This is typically done using policy gradient methods such as PPO~\citep{stiennon2020learning} or GRPO~\citep{guo2025deepseek}.

\begin{rmk}{Other (non-preference) reward models}{}
  Note that the post-training pipeline we described here is agnostic to the choice of reward model.
  That is, instead of post-training our language model to comply with human preferences, we could have also post-trained it to maximize any other reward signal.
  More recently, works have explored the use of reward models for challenging ``reasoning problems'' such as mathematical calculations, where the reward is usually based on the correctness of the answer.\safefootnote{On training data where the answer to problem $\vx$ is known to be $\opt{y}(\vx)$, the reward is simply $r(\vy \mid \vx) = \Ind{\opt{y}(\vx) \in \vy}$. That is, the reward is $1$ if the response $\vy$ contains the correct answer and $0$ otherwise.}
  One prominent example for this are ``reasoning'' models such as the DeepSeek-R1 model~\citep{guo2025deepseek}, which was trained with GRPO.
\end{rmk}

\subsection{Direct Preference Optimization}\pidx{direct preference optimization}%

Observe that the reward model can be expressed in terms of its associated optimal policy: \begin{align}
  r(\vy \mid \vx) = \lambda \log\frac{\altpi_\star(\vy \mid \vx)}{\altpi_{\init}(\vy \mid \vx)} + \const. \margintag{by reorganizing the terms of \cref{eq:optimal_regularized_policy_rlhf}} \label{eq:dpo_reward_policy_relationship}
\end{align}
In particular, it follows that given a fixed prior $\altpi_{\init}$, \emph{any} policy $\altpi_\vvarphi$ has a family of associated reward models with respect to which it is optimal!
We denote by \begin{align}
  r_{[\vvarphi]}(\vy \mid \vx) \defeq \lambda \log\frac{\altpi_\vvarphi(\vy \mid \vx)}{\altpi_{\init}(\vy \mid \vx)}
\end{align} the ``simplest'' of these reward models.
Remembering the characterization of the optimal policy from \cref{eq:optimal_regularized_policy_rlhf} it follows immediately that $\altpi_\vvarphi$ is optimal with respect to $r_{[\vvarphi]}$.

Instead of first learning an approximate reward model and then finding the associated optimal policy, DPO exploits the relationship of \cref{eq:dpo_reward_policy_relationship} to learn the optimal policy directly \citep{rafailov2023direct}.
Substituting $r_\vtheta$ in the maximum likelihood estimation from \cref{eq:rlhf_reward_model}, yields the objective \begin{align}
  &\E[(\vy_A \succ \vy_B \mid \vx) \sim \spD]{\log \sigma\parentheses*{r_{[\vvarphi]}(\vy_A \mid \vx) - r_{[\vvarphi]}(\vy_B \mid \vx)}} \nonumber \\
  &= \E[(\vy_A \succ \vy_B \mid \vx) \sim \spD]{\log \sigma\parentheses*{\lambda \log\frac{\altpi_\vvarphi(\vy_A \mid \vx)}{\altpi_{\init}(\vy_A \mid \vx)} - \lambda \log\frac{\altpi_\vvarphi(\vy_B \mid \vx)}{\altpi_{\init}(\vy_B \mid \vx)}}}. \margintag{using \cref{eq:dpo_reward_policy_relationship}}
\end{align}
Gradients can be computed via automatic differentiation: \begin{align}
  \begin{multlined}
    \lambda \E*[(\vy_A \succ \vy_B \mid \vx) \sim \spD]{\Biggl[\underbrace{\sigma\parentheses*{r_{[\vvarphi]}(\vy_B \mid \vx) - r_{[\vvarphi]}(\vy_A \mid \vx)}}_{\text{weight according to error of reward estimate}} \\ \Bigl[\underbrace{\grad_\vvarphi \log \altpi_\vvarphi(\vy_A \mid \vx)}_{\text{increase likelihood of $\vy_A$}} - \underbrace{\grad_\vvarphi \log \altpi_\vvarphi(\vy_B \mid \vx)}_{\text{decrease likelihood of $\vy_B$}}\Bigr]\Biggr]}.
  \end{multlined}
\end{align}
Intuitively, DPO successively increases the likelihood of preferred responses $\vy_A$ and decreases the likelihood of dispreferred responses $\vy_B$.
Examples $(\vy_A \succ \vy_B \mid \vx)$ are weighted by the strength of regularization~$\lambda$ and by the degree to which the implicit reward model incorrectly orders the responses.

\section*{Discussion}\label{sec:mfarl:discussion}

In this chapter, we studied central ideas in actor-critic methods.
We have seen two main approaches to use policy-gradient methods.
We began, in \cref{sec:mfarl:policy_approximation}, by introducing the REINFORCE algorithm which uses policy gradients and Monte Carlo estimation, but suffered from large variance in the gradient estimates of the policy value function.
In \cref{sec:mfarl:actor_critic_methods}, we have then seen a number of actor-critic methods such as A2C and GAE behaving similarly to policy iteration that exhibit less variance, but are very sample inefficient due to their on-policy nature.
TRPO improves the sample efficiency slightly, but not fundamentally.

In \cref{sec:mfarl:off_policy_actor_critic_methods}, we discussed a second family of policy gradient techniques that generalize Q-learning and are akin to value iteration.
For reparameterizable policies, this led us to algorithms such as DDPG, TD3, SVG.
Importantly, these algorithms are significantly more sample efficient than on-policy policy gradient methods, which often results in much faster learning of a near-optimal policy.
In \cref{sec:mfarl:actor_critic_methods:entropy_regularization}, we discussed entropy regularization which frames reinforcement learning as probabilistic inference, and we derived the SAC algorithm which is widely used and works quite well in practice.

Finally, in \cref{sec:mfarl:preference_learning}, we studied two canonical approaches to learning from preference feedback: RLHF which separately learns reward model and policy and DPO which learns a policy directly.
We have seen that RLHF is akin to model-based RL as it explicitly learns a reward model through maximum likelihood estimation.
In contrast, DPO is more closely related to model-free RL and policy gradient methods as it learns the optimal policy directly.

\begin{oreadings}
  \begin{itemize}
    \item \textbf{A3C:} \pcite{mnih2016asynchronous}
    \item \textbf{GAE:} \pcite{schulman2015high}
    \item \textbf{TRPO:} \pcite{schulman2015trust}
    \item \textbf{PPO:} \pcite{schulman2017proximal}
    \item \textbf{DDPG:} \pcite{lillicrap2015continuous}
    \item \textbf{TD3:} \pcite{fujimoto2018addressing}
    \item \textbf{SVG:} \pcite{heess2015learning}
    \item \textbf{SAC:} \pcite{haarnoja2018soft}
    \item \textbf{DPO:} \pcite{rafailov2023direct}
  \end{itemize}
\end{oreadings}

\excheading

\begin{figure}
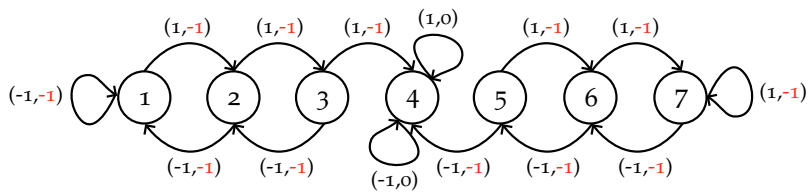

  \incfig{q_learning_example}
  \caption{MDP studied in \cref{exercise:q_learning_func_approx}. Each arrow marks a (deterministic) transition and is labeled with $(\text{action},\text{reward})$.}\label{fig:q_learning_func_approx}
\end{figure}

\begin{nexercise}{Q-learning and function approximation}{q_learning_func_approx}
  Consider the MDP of \cref{fig:q_learning_func_approx} and set $\gamma = 1$.

  \begin{enumerate}
    \item Using Bellman's theorem, prove that $\v*{x} = -\abs{x-4}$ is the optimal value function.

    \item Suppose we observe the following episode:
    \begin{center}
      \vspace{5pt}
      \begin{tabular}{cccc}
        \toprule
        $x$ & $a$ & $x'$ & $r$ \\
        \midrule
        $3$ & $-1$ & $2$ & $-1$ \\
        $2$ & $1$ & $3$ & $-1$ \\
        $3$ & $1$ & $4$ & $-1$ \\
        $4$ & $1$ & $4$ & $0$ \\
        \bottomrule
      \end{tabular}
      \vspace{5pt}
    \end{center}
    We initialize all Q-values to $0$.
    Compute the updated Q-values using Q-learning with learning rate $\alpha = \nicefrac{1}{2}$.

    \item We will now approximate the Q-function with a linear function.
    We let \begin{align*}
      \Q{x}{a; \vw} \defeq x w_0 + a w_1 + w_2
    \end{align*} where $\vw = \transpose{[w_0 \; w_1 \; w_2]} \in \R^3$.

    Suppose we have $\old{\vw} = \transpose{[1 \; -1 \; -2]}$ and $\vw = \transpose{[-1 \; 1 \; 1]}$, and we observe the transition $\tau = (2, -1, -1, 1)$.
    Use the learning rate $\alpha = \nicefrac{1}{2}$ to compute $\grad_\vw \ell(\vw; \tau)$ and the updated weights ${\vw' = \vw - \alpha \grad_\vw \ell(\vw; \tau)}$.
  \end{enumerate}
\end{nexercise}

\begin{nexercise}{Eligibility vector}{eligibility_vector}
  The vector $\grad_\vvarphi \log \pi_\vvarphi(\va_t \mid \vx_t)$ is commonly called \midx{eligibility vector}.
  In the following, we assume that the action space $\spA$ is finite and denote it by $\sA$.

  If we parameterize $\pi_\vvarphi$ as a softmax distribution \begin{align}
    \pi_\vvarphi(a \mid \vx) \defeq \frac{\exp(h(\vx, a, \vvarphi))}{\sum_{b \in \sA} \exp(h(\vx, b, \vvarphi))}
  \end{align} with linear preferences $h(\vx, a, \vvarphi) \defeq \transpose{\vvarphi} \vphi(\vx, a)$ where $\vphi(\vx, a)$ is some feature vector, what is the form of the eligibility vector?
\end{nexercise}

\begin{nexercise}{Variance of score gradients with baselines}{score_gradients_with_baselines_variance}
  In this exercise, we will see a sufficient condition for baselines to reduce the variance of score gradient estimators.

  \begin{enumerate}
    \item Suppose for a random vector $\rX$, we want to estimate $\E{f(\rX)}$ for some function $f$.
    Assume that you are given a function $g$ and also its expectation $\E{g(\rX)}$.
    Instead of estimating $\E{f(\rX)}$ directly, we will instead estimate $\E{f(\rX) - g(\rX)}$ as we know from linearity of expectation \eqref{eq:linearity_expectation} that \begin{align*}
      \E{f(\rX)} = \E{f(\rX) - g(\rX)} + \E{g(\rX)}.
    \end{align*}
    Prove that if $\frac{1}{2} \Var{g(\rX)} \leq \Cov{f(\rX), g(\rX)}$, then \begin{align}
      \Var{f(\rX) - g(\rX)} \leq \Var{f(\rX)}. \label{eq:variance_reduction}
    \end{align}

    \item Consider estimating $\grad_\vvarphi \j{\vvarphi}$.
    Prove that if $b^2 \leq 2 b \cdot r(\vx, \va)$ for every state $\vx \in \spX$ and action $\va \in \spA$, then \begin{align}
      \Var[\tau \sim \altpi_\vvarphi]{(G_0 - b) \grad_\vvarphi \log \altpi_\vvarphi(\tau)} \leq \Var[\tau \sim \altpi_\vvarphi]{G_0 \grad_\vvarphi \log \altpi_\vvarphi(\tau)}.
    \end{align}
  \end{enumerate}
\end{nexercise}

\begin{nexercise}{Score gradients with state-dependent baselines}{score_gradients_state_dep_baselines}
  For a sequence of \midx<state-dependent baselines>{state-dependent baseline} $\{b(\tau_{0:t-1})\}_{t=1}^T$ where \begin{align*}
    \tau_{0:t-1} \defeq (\tau_0, \tau_1, \dots, \tau_{t-1}),
  \end{align*} show that \begin{align}\begin{split}
    &\E[\tau \sim \altpi_\vvarphi]{G_0 \grad_\vvarphi \log \altpi_\vvarphi(\tau)} \\
    &= \E[\tau \sim \altpi_\vvarphi]{\sum_{t=0}^{T-1} (G_0 - b(\tau_{0:t-1})) \grad_\vvarphi \log \pi_\vvarphi(\va_t \mid \vx_t)}
  \end{split}\label{eq:score_gradient_estimator_state_dep_baseline}\end{align} where we write $b(\tau_{0:-1}) = 0$.
\end{nexercise}

\begin{nexercise}{Policy gradients with downstream returns}{policy_gradients_with_downstream_returns}
  Suppose we are training an agent to solve a computer game.
  There are only two possible actions, specifically: \begin{enumerate}
    \item \emph{do nothing}; and
    \item \emph{move}.
  \end{enumerate}
  Each episode lasts for four ($T = 3$) time steps.
  The policy $\pi_\theta$ is completely determined by the parameter $\theta \in [0,1]$.
  Here, for simplicity, we have assumed that the policy is independent of the current state.
  The probability of moving (action $2$) is equal to $\theta$ and the probability of doing nothing (action $1$) is $1 - \theta$.

  Initially, $\theta = 0.5$.
  One episode is played with this initial policy and the results are \begin{align*}
    \text{actions} = (1, 0, 1, 0), \quad \text{rewards} = (1, 0, 1, 1).
  \end{align*}
  Compute the policy gradient estimate with downstream returns, discount factor $\gamma = \nicefrac{1}{2}$, and the provided \emph{single} sample $\tau \sim \altpi_\theta$.
\end{nexercise}

\begin{nexercise}{Policy gradient with an exponential family}{policy_gradient_with_exponential_family}
  \begin{enumerate}
    \item Suppose, we can choose between two actions $a \in \{0, 1\}$ in each state.
    A natural stochastic policy is induced by a Bernoulli distribution, \begin{align}
      a \sim \Bern{\sigma(f_\vvarphi(\vx))},
    \end{align} where $\sigma$ is the logistic function from \cref{eq:logistic_function}.
    First, write down the expression for $\pi_\vvarphi(a \mid \vx)$.
    Then, derive the expression for $\grad_\vvarphi \j{\vvarphi}$ in terms of $\fnq[\pi_\vvarphi]$, $\sigma(f_\vvarphi(\vx))$, and $\grad_\vvarphi f_\vvarphi(\vx)$ using the policy gradient theorem.

    \item The Bernoulli distribution is part of a family of distributions that allows for a much simpler derivation of the gradient than was necessary in (1).
    A univariate \midx{exponential family} is a family of distributions whose PDF (or PMF) can be expressed in canonical form as\looseness=-1 \begin{align}
      \pi_\vvarphi(a \mid \vx) = h(a) \exp(a f_\vvarphi(\vx) - A(f_\vvarphi(\vx)))
    \end{align} where $h$, $f$, and $A$ are known functions.\safefootnote{Observe that this form is equivalent to the form introduced in \cref{eq:exponential_family_of_distributions} where $f_\vvarphi(\vx)$ is the natural parameter, and we let $A(f) = \log Z(f)$.}
    Derive the expression of the policy gradient $\grad_\vvarphi \j{\vvarphi}$ for such a distribution.

    \item Can you relate the results of the previous two exercises (1) and (2)?
    What are $h$ and $A$ in case of the Bernoulli distribution?

    \item The Gaussian distribution with unit variance $\N{f_\vvarphi(\vx)}{1}$ is of the same canonical form with \begin{align}
      A(f_\vvarphi(\vx)) = \frac{f_\vvarphi(\vx)^2}{2}.
    \end{align}
    Determine the policy gradient $\grad_\vvarphi \j{\vvarphi}$.

    \item For a Gaussian policy, can we instead apply the reparameterization trick \eqref{eq:reparameterization_trick_gaussian} that we have seen in the context of variational inference? If yes, how? If not, why?
  \end{enumerate}
\end{nexercise}

\begin{nexercise}{Soft value function}{soft_value_function}
  In this exercise, we derive the optimal policy solving \cref{eq:entropy_reg_rl2} and the soft value function from \cref{eq:soft_value_function}.
  \begin{enumerate}
    \item We let $\beta(\va_t \mid \vx_t) \defeq \exp(\frac{1}{\lambda} r(\vx_t, \va_t))$, $Z(\vx) \defeq \int_\spA \beta(\va \mid \vx) \, d \va$, and denote by $\hat{\pi}(\cdot \mid \vx)$ the policy $\beta(\cdot \mid \vx) / Z(\vx)$.
    Show that \begin{align}
      \begin{multlined}
        \KL{\altpi_\vvarphi}{\altpi_\star} \\ = \sum_{t=1}^T \E[\vx_t \sim \altpi_\vvarphi]{\KL{\pi_\vvarphi(\cdot \mid \vx_t)}{\hat{\pi}(\cdot \mid \vx_t)} - \log Z(\vx_t)}.
      \end{multlined} \label{eq:entropy_reg_rl3}
    \end{align}

    \item Show that if the space of policies parameterized by $\vvarphi$ is sufficiently expressive, $\pis(\va \mid \vx) \propto \exp(\q*{\vx}{\va})$ solves \cref{eq:entropy_reg_rl2}.
  \end{enumerate}
\end{nexercise}

\begin{nexercise}{PPO as probabilistic inference}{ppo_as_probabilistic_inference}
  \begin{enumerate}
    \item Consider the same generative model as was introduced in \cref{sec:mfarl:entropy_regularization_as_probabilistic_inference} when we interpreted entropy-regularized RL as probabilistic inference.
    Only now, assume that $T=1$ \safefootnote{since, in this section, we have been considering outcome rewards} and suppose that the prior over actions is $p(\vy \mid \vx) = \altpi_{\init}(\vy \mid \vx)$ rather than uniform.
    Define the distribution over optimal trajectories $\altpi_\star(\vy \mid \vx)$ as before in \cref{eq:distr_over_optimal_trajectories}.
    Show that for any context $\vx$, \begin{align}
      \argmin_{\vvarphi} \KL{\altpi_{\vvarphi}(\cdot \mid \vx)}{\altpi_\star(\cdot \mid \vx)} = \argmax_{\vvarphi} \j{\vvarphi; \vvarphi^{\init} \mid \vx}[\lambda].
    \end{align}

    \item Conclude that the policy maximizing $\j{\vvarphi; \vvarphi^{\init} \mid \vx}[\lambda]$ is \begin{align}
      \altpi_\star(\vy \mid \vx) \propto \altpi_{\init}(\vy \mid \vx) \exp\parentheses*{\frac{1}{\lambda} r(\vy \mid \vx)}.
    \end{align}
  \end{enumerate}
\end{nexercise}

  \chapter{Model-based Reinforcement Learning}\label{sec:mbarl}

In this final chapter, we will revisit the model-based approach to reinforcement learning.
We will see some advantages it offers over model-free methods.
In particular, we will use the machinery of probabilistic inference, which we developed in the first chapters, to quantify uncertainty about our model and use this uncertainty for planning, exploration, and reliable decision-making.

To recap, in \cref{sec:tabular_rl}, we began by discussing model-based reinforcement learning which attempts to learn the underlying Markov decision process and then use it for planning.
We have seen that in the tabular setting, computing and storing the entire model is computationally expensive.
This led us to consider the family of model-free approaches, which learn the value function directly, and as such can be considered more economical in the amount of data that they store.

In \cref{sec:mfarl}, we saw that using function approximation, we were able to scale model-free methods to very large state and action spaces.
We will now explore similar ideas in the model-based framework.
Namely, we will use function approximation to condense the representation of our model of the environment.
More concretely, we will learn an approximate dynamics model $f \approx p$ and approximate rewards $r$, which is also called a \midx{world model}.

There are a few ways in which the model-based approach is advantageous.
First, if we have an accurate model of the environment, we can use it for planning --- ideally also for interacting safely with the environment.
However, in practice, we will rarely have such a model.
In fact, the accuracy of our model will depend on the past experience of our agent and the region of the state-action space.
Understanding the uncertainty in our model of the environment is crucial for planning.
In particular, quantifying uncertainty is necessary to drive safe(!) exploration and avoid undesired states.

Moreover, modeling uncertainty in our model of the environment can be extremely useful in deciding where to explore.
Learning a model can therefore help to dramatically reduce the sample complexity over model-free techniques.
Often times this is crucial when developing agents for real-world use as in such settings, exploration is costly and potentially dangerous.

\Cref{alg:mbarl} describes the general approach to model-based reinforcement learning.

\begin{algorithm}[H]
  \caption{Model-based reinforcement learning (outline)}\pidx{model-based reinforcement learning}\label{alg:mbarl}
  start with an initial policy $\pi$ and no (or some) initial data $\spD$\;
  \For{several episodes}{
    roll out policy $\pi$ to collect data\;
    learn a model of the dynamics $f$ and rewards $r$ from data\;
    plan a new policy $\pi$ based on the estimated models\;
  }
\end{algorithm}

We face three main challenges in model-based reinforcement learning.
First, given a fixed model, we need to perform planning to decide on which actions to play.
Second, we need to learn models $f$ and $r$ accurately and efficiently.
Third, we need to effectively trade exploration and exploitation.
We will discuss these three topics in the following.

\section{Planning}\label{sec:mbarl:planning}

There exists a large literature on planning in various settings.
These settings can be mainly characterized as \begin{itemize}
  \item discrete or continuous action spaces;
  \item fully- or partially observable state spaces;
  \item constrained or unconstrained; and
  \item linear or nonlinear dynamics.
\end{itemize}
In \cref{sec:mdp}, we have already seen algorithms such as policy iteration and value iteration, which can be used to solve planning exactly in tabular settings.
In the following, we will now focus on the setting of continuous state and action spaces, fully observable state spaces, no constraints, and nonlinear dynamics.

\subsection{Deterministic Dynamics}\label{sec:mbarl:planning:determinstic_dynamics}

To begin with, let us assume that our dynamics model is deterministic and known.
That is, given a state-action pair, we know the subsequent state,\looseness=-1 \begin{align}
  \vx_{t+1} = \vf(\vx_t, \va_t).
\end{align}
We continue to focus on the setting of infinite-horizon discounted returns\pidx{discounted payoff} \eqref{eq:discounted_payoff}, which we have been considering throughout our discussion of reinforcement learning.
This yields the objective, \begin{align}
  \max_{\va_{0:\infty}} \sum_{t=0}^\infty \gamma^t r(\vx_t, \va_t) \quad \text{such that}\ \vx_{t+1} = \vf(\vx_t, \va_t).
\end{align}
Now, because we are optimizing over an infinite time horizon, we cannot solve this optimization problem directly.
This problem is studied ubiquitously in the area of \midx{optimal control}.
We will discuss one central idea from optimal control that is widely used in model-based reinforcement learning, and will later return to using this idea for learning parametric policies in \cref{sec:mbarl:planning:parametric_policies}.

\paragraph{Planning over finite horizons:}

The key idea of a classical algorithm from optimal control called \midx{receding horizon control} (RHC) or \midx{model predictive control} (MPC) is to iteratively plan over finite horizons.
That is, in each round, we plan over a finite time horizon $H$ and carry out the first action.

\begin{algorithm}[H]
  \caption{Model predictive control, MPC}
  \For{$t=0$ \KwTo $\infty$}{
    observe $\vx_t$\;
    plan over a finite horizon $H$, \vspace{-8pt}\begin{align}
      \max_{\va_{t:t+H-1}} \sum_{\tau=t}^{t+H-1} \gamma^{\tau-t} r(\vx_\tau, \va_\tau) \quad \text{such that}\ \vx_{\tau+1} = \vf(\vx_\tau, \va_\tau)
    \end{align}\vspace{-20pt}\;
    carry out action $\va_t$\;
  }
\end{algorithm}

\begin{marginfigure}
  \incfig{mpc}
  \caption{Illustration of model predictive control in a deterministic transition model.
  The agent starts in position $\vx_0$ and wants to reach $\vxs$ despite the black obstacle.
  We use the reward function $r(\vx) = -\norm{\vx-\vxs}$.
  The gray concentric circles represent the length of a single step.
  We plan with a time horizon of $H=2$. Initially, the agent does not ``see'' the black obstacle, and therefore moves straight towards the goal.
  As soon as the agent sees the obstacle, the optimal trajectory is ``replanned''.
  The dotted red line corresponds to the optimal trajectory, the agent's steps are shown in blue.}
\end{marginfigure}

Observe that the state $\vx_\tau$ can be interpreted as a deterministic function $\vx_\tau(\va_{t:\tau-1})$, which depends on all actions from time $t$ to time $\tau-1$ and the state $\vx_t$.
To solve the optimization problem of a single iteration, we therefore need to maximize, \begin{align}
  \j{\va_{t:t+H-1}}[H] \defeq \sum_{\tau=t}^{t+H-1} \gamma^{\tau-t} r(\vx_\tau(\va_{t:\tau-1}), \va_\tau).
\end{align}
This optimization problem is in general non-convex.
If the actions are continuous and the dynamics and reward models are differentiable, we can nevertheless obtain analytic gradients of $\fnJ[H]$.
This can be done using the chain rule and ``backpropagating'' through time, analogously to backpropagation in neural networks.\footnote{see \cref{sec:bdl:ann:backprop}}
Especially for large horizons $H$, this optimization problem becomes difficult to solve exactly due to local optima and vanishing/exploding gradients.

\paragraph{Tree search:}

Often, heuristic global optimization methods (also called ``search methods'') are used to optimize $\fnJ[H]$.
An example are \midx{random shooting methods}, which find the optimal choice among a set of random proposals.
Of course, obtaining a ``good'' set of randomly proposed action sequences is crucial.
A naive way of generating the proposals is to pick them uniformly at random.
This strategy, however, usually does not perform very well as it corresponds to suggesting random walks of the state space.
Alternatives are to sample from a Gaussian distribution or using the \midx{cross-entropy method} which gradually adapts the sampling distribution by reweighing samples according to the rewards they produce.

\begin{algorithm}
  \caption{Random shooting methods}
  generate $m$ sets of random samples, $\va_{t:t+H-1}^{(i)}$\;
  pick the sequence of actions $\va_{t:t+H-1}^{(\opt{i})}$ where \vspace{-8pt}\begin{align}
    \opt{i} = \argmax_{i\in[m]} \j{\va_{t:t+H-1}^{(i)}}[H]
  \end{align}\vspace{-20pt}\;
\end{algorithm}

The evolution of the state can be visualized as a tree where --- if dynamics are deterministic --- the branching is fully determined by the played actions.
For this reason, classical tree search algorithms can be employed, such as \midx{alpha-beta pruning} or \midx{Monte Carlo tree search} (MCTS), which was used for example by ``AlphaZero'' \citep{silver2016mastering,silver2017mastering} and can be viewed as an advanced variant of a shooting method.

\paragraph{Finite-horizon planning with sparse rewards:}

A common problem of finite-horizon methods is that in the setting of sparse rewards, there is often no signal that can be followed.
You can think of an agent that operates in a kitchen and tries to find a box of candy.
Yet, to get to this box, it needs to perform a large number of actions.
In particular, if this number of actions is larger than the horizon $H$, then the local optimization problem of MPC will not take into account the reward for choosing this long sequence of actions.
Thus, the box of candy will likely never be found by our agent.

\begin{marginfigure}
  \incfig{sparse_rewards}
  \caption{Illustration of finite-horizon planning with sparse rewards.
  When the finite time horizon does not suffice to ``reach'' a reward, the agent has no signal to follow.}
\end{marginfigure}

A solution to this problem is to amend a long-term value estimate to the finite-horizon sum.
The idea is to not only consider the rewards attained \emph{while} following the actions $\va_{t:t_H-1}$, but to also consider the value of the final state $\vx_{t+H}$, which estimates the discounted sum of \emph{future} rewards.
\begin{align}
  \J{\va_{t:t+H-1}}[H] \defeq \underbrace{\sum_{\tau=t}^{t+H-1} \gamma^{\tau-t} r(\vx_\tau(\va_{t:\tau-1}), \va_\tau)}_{\text{short-term}} + \underbrace{\gamma^H \V{\vx_{t+H}}}_{\text{long-term}}.
\end{align}
Intuitively, $\gamma^H \V{\vx_{t+H}}$ is estimating the tail of the infinite sum.

\begin{rmk}{Planning generalizes model-free methods!}{planning_generalizes_model_free_rl}
  Observe that for $H=1$, when we use the value function estimate associated with this MPC controller, maximizing $J_H$ coincides with using the greedy policy $\pi_V$.
  That is, \begin{align}
    \va_t = \argmax_{\va \in \spA} \J{\va}[1] = \pi_{\fnV}(\vx_t).
  \end{align}
  Thus, by looking ahead for a single time step, we recover the approaches from the model-free setting in this model-based setting!
  Essentially, if we do not plan long-term and only consider the value estimate, the model-based setting reduces to the model-free setting.
  However, in the model-based setting, we are now able to use our model of the transition dynamics to anticipate the downstream effects of picking a particular action $\va_t$.
  This is one of the fundamental reasons for why model-based approaches are typically severely more sample efficient than model-free methods.
\end{rmk}

To obtain the value estimates, we can use the approaches we discussed in detail in \cref{sec:tabular_rl:model_free}, such as TD-learning for on-policy value estimates and Q-learning for off-policy value estimates.
For large state-action spaces, we can use their approximate variants, which we discussed in \cref{sec:mfarl:tabular_rl_as_optimization} and \cref{sec:mfarl:value_function_approximation}.
To improve value estimates, we can obtain ``artificial'' data by rolling out policies within our model.
This is a key advantage over model-free methods as, once a sufficiently accurate model has been learned, data for value estimation can be generated efficiently in simulation.

\subsection{Stochastic Dynamics}\label{sec:mbarl:planning:stochastic_dynamics}

How can we extend this approach to planning to a stochastic transition model?
A natural extension of model predictive control is to do what is called \midx{stochastic average approximation} (SAA) or \midx{trajectory sampling} \citep{chua2018deep}.
Like in MPC, we still optimize over a deterministic sequence of actions, but now we will average over all resulting trajectories.\looseness=-1

\begin{algorithm}
  \caption{Trajectory sampling}
  \For{$t=0$ \KwTo $\infty$}{
    observe $\vx_t$\;
    optimize expected performance over a finite horizon $H$, \vspace{-8pt}\begin{align}
      \max_{\va_{t:t+H-1}} \E[\vx_{t+1:t+H}]{\sum_{\tau=t}^{t+H-1} \gamma^{\tau-t} r_\tau + \gamma^H \V{\vx_{t+H}}}[\va_{t:t+H-1}, f]
    \end{align}\vspace{-20pt}\;
    carry out action $\va_t$\;
  }
\end{algorithm}

\begin{marginfigure}
  \incfig{trajectory_sampling}
  \caption{Illustration of trajectory sampling.
  High-reward states are shown in brighter colors.
  The agent iteratively plans a finite number of time steps into the future and picks the best initial action.}
\end{marginfigure}

Intuitively, trajectory sampling optimizes over a much simpler object --- namely a deterministic sequence of actions of length $H$ --- than finding a policy, which corresponds to finding an optimal decision tree mapping states to actions.
Of course, using trajectory sampling (from an arbitrary starting state) implies such a policy.
However, trajectory sampling never computes this policy explicitly, and rather, in each step, only plans over a finite horizon.

Computing the expectation exactly is typically not possible as this involves solving a high-dimensional integral of nonlinear functions.
Instead, a common approach is to use Monte Carlo estimates of this expectation.
This approach is known as \midx{Monte Carlo trajectory sampling}.
The key issue with using sampling based estimation is that the sampled trajectory (i.e., sampled sequence of states) we obtain, depends on the actions we pick.
In other words, the measure we average over depends on the decision variables --- the actions.
This is a problem that we have seen several times already!
It naturally suggests using the reparameterization trick\pidx{reparameterization trick}.\footnote{see \cref{thm:reparameterization_trick} and \cref{eq:reparameterization_gradients}}

Previously, we have used the reparameterization trick to reparameterize variational distributions (see \cref{sec:approximate_inference:variational_inference:gradient_of_elbo}) and to reparameterize policies (see \cref{sec:mfarl:actor_critic_methods:randomized_policies}).
It turns out that we can use the exact same approach for reparameterizing the transition model.
We say that a (stochastic) transition model $f$ is \emph{reparameterizable} iff $\vx_{t+1} \sim f(\vx_t, \va_t)$ is such that $\vx_{t+1} = \vg(\vvarepsilon; \vx_t, \va_t)$, where $\vvarepsilon \sim \phi$ is a random variable that is independent of $\vx_t$ and $\va_t$.
We have already seen in \cref{ex:reparameterization_gradients_gaussian} (in the context of stochastic policies) how a Gaussian transition model can be reparameterized.

In this case, $\vx_\tau$ is determined recursively by $\va_{t:\tau-1}$ and $\vvarepsilon_{t:\tau-1}$, \begin{align}\begin{split}
  \vx_\tau &= \vx_\tau(\vvarepsilon_{t:\tau-1}; \va_{t:\tau-1}) \\
  &\defeq \vg(\vvarepsilon_{\tau-1}; \vg(\dots; (\vvarepsilon_{t+1}; \vg(\vvarepsilon_t; \vx_t, \va_t), \va_{t+1}), \dots), \va_{\tau-1}). \label{eq:state_encoding}
\end{split}\end{align}
This allows us to obtain unbiased estimates of $J_H$ using Monte Carlo approximation, \begin{align}\begin{split}
  \J{\va_{t:t+H-1}}[H] \approx \begin{multlined}[t]
    \frac{1}{m} \sum_{i=1}^m \Bigg( \sum_{\tau=t}^{t+H-1} \gamma^{\tau-t} r(\vx_\tau(\vvarepsilon_{t:\tau-1}^{(i)}; \va_{t:\tau-1}), \va_\tau) \\ + \gamma^H \V{\vx_{t+H}(\vvarepsilon_{t:t+H-1}^{(i)}; \va_{t:t+H-1})} \Bigg)
  \end{multlined} \label{eq:mbarl_planning_monte_carlo}
\end{split}\end{align} where $\vvarepsilon_{t:t+H-1}^{(i)} \iid \phi$ are independent samples.
To optimize this approximation we can again compute analytic gradients or use shooting methods as we have discussed in \cref{sec:mbarl:planning:determinstic_dynamics} for deterministic dynamics.\looseness=-1

\subsection{Parametric Policies}\label{sec:mbarl:planning:parametric_policies}

When using algorithms such as model predictive control for planning, planning needs to be done online before each time we take an action.
This is called \midx{closed-loop control} and can be expensive.
Especially when the time horizon is large, or we encounter similar states many times (leading to ``repeated optimization problems''), it can be beneficial to ``store'' the planning decision in a (deterministic) policy, \begin{align}
  \va_t = \vpi(\vx_t; \vvarphi) \eqdef \vpi_\vvarphi(\vx_t).
\end{align}
This policy can then be trained offline and evaluated cheaply online, which is known as \midx{open-loop control}.

This is akin to a problem that we have discussed in detail in the previous chapter when extending Q-learning to large action spaces.
There, this led us to discuss policy gradient and actor-critic methods.
Recall that in Q-learning, we seek to follow the greedy policy, \begin{align*}
  \opt{\vpi}(\vx) = \argmax_{\va \in \spA} \Q*{\vx}{\va; \vtheta}, \margintag{see \cref{eq:q_learning_policy}}
\end{align*} and therefore had to solve an optimization problem over all actions.
We accelerated this by learning an approximate policy that ``mimicked'' this optimization, \begin{align*}
  \opt{\vvarphi} = \argmax_{\vvarphi} \underbrace{\E[\vx \sim \mu]{\Q*{\vx}{\vpi_\vvarphi(\vx); \vtheta}}}_{\J{\vvarphi; \vtheta}[\mu]} \margintag{see \cref{eq:off_policy_actor_critics_optimization}}
\end{align*} where $\mu(\vx) > 0$ was some exploration distribution that has full support and thus leads to the exploration of all states.
The key idea was that if we use a differentiable approximation $Q$ and a differentiable parameterization of policies, which is ``rich enough'', then both optimizations are equivalent, and we can use the chain rule to obtain gradient estimates of the second expression.
We then used this to derive the \midx{deep deterministic policy gradients} (DDPG) and \midx{stochastic value gradients} (SVG) algorithms.
It turns out that there is a very natural analogue to DDPG/SVG for model-based reinforcement learning.

Instead of maximizing the Q-function directly, we use finite-horizon planning to estimate the immediate value of the policy within the next $H$ time steps and simply use the Q-function to approximate the terminal value (i.e., the tails of the infinite sum).
Then, our objective becomes,\looseness=-1 \begin{align}
  \J{\vvarphi; \vtheta}[\mu,H] \defeq \E[\vx_0 \sim \mu, \vx_{1:H} \mid \vpi_\vvarphi, f]{\sum_{\tau=0}^{H-1} \gamma^\tau r_\tau + \gamma^H \Q*{\vx_H}{\vpi_\vvarphi(\vx_H); \vtheta}} \label{eq:rl_objective_with_tails}
\end{align}
This approach naturally extends to randomized policies using reparameterization gradients, which we have discussed in \cref{sec:mfarl:actor_critic_methods:randomized_policies}.
Analogously to \cref{rmk:planning_generalizes_model_free_rl}, for $H=0$, this coincides with the DDPG objective!
For larger time horizons, the look-ahead takes into account the transition model for planning next time steps.
This tends to help dramatically in improving policies \emph{much more rapidly} between episodes. Instead of just gradually improving policies a little by slightly adapting the policy to the learned value function estimates (as in model-free RL), we use the model to anticipate the consequences of actions multiple time steps ahead. This is at the heart of model-based reinforcement learning.\looseness=-1

Essentially, we are using methods such as Q-learning and DDPG/SVG as subroutines within the framework of model predictive control to do much bigger steps in policy improvement than to slightly improve the next picked action.
To encourage exploration, it is common to extend the objective in \cref{eq:rl_objective_with_tails} by an additional entropy regularization term as seen in \cref{sec:mfarl:actor_critic_methods:entropy_regularization}.

\begin{rmk}{Planning as inference}{planning_as_inference}
  We have been using \midx{Monte Carlo rollouts} to estimate the expectation of \cref{eq:rl_objective_with_tails}.
  That is, we have been using a Monte Carlo approximation (e.g., a sample mean) using samples obtained by ``rolling out'' the induced Markov chain of a fixed policy.

  It would certainly be preferable to compute the expectation exactly, however, this is generally not possible as this involves solving a high dimensional integral.
  Recall that we faced the same problem when studying inference in the first half of the manuscript.
  In both problems, we need to approximate high-dimensional integrals (i.e., expectations).
  This suggests a deep connection between the problems of planning and inference.
  It is therefore not surprising that many techniques for approximate inference that we have seen earlier can also be applied to planning.

  \begin{itemize}
    \item \midx{Monte Carlo approximation} which we have been focusing on during our discussion of planning is a very simple inference algorithm --- approximating an expectation by sampling from the distribution that is averaged over.
    This allowed us to obtain unbiased gradient estimates (which may have high variance).

    \item An alternative approach is \midx{moment matching} (cf. \cref{sec:vi:kl:forward}).
    Instead of approximating the expectation, here, we approximate the distribution over trajectories using a tractable distribution (e.g., a Gaussian) and ``matching'' their moments.
    This then allows us to analytically compute gradients of the expectation in \cref{eq:rl_objective_with_tails}.
    A prominent example of this approach is \midx{probabilistic inference for learning control} (PILCO).

    \item In \cref{sec:mfarl:actor_critic_methods:entropy_regularization}, we have used \midx{variational inference} for planning, and seen that it coincides with entropy regularization as implemented by the \midx{soft actor critic} (SAC) algorithm.
  \end{itemize}
\end{rmk}

\section{Learning}

Thus far, we have considered known environments.
That is, we assumed that the transition model $f$ and the rewards $r$ are known.
In reinforcement learning, $f$ and $r$ are (of course!) not known.
Instead, we have to estimate them from data.
This will also be crucial in our later discussion of exploration in \cref{sec:mbarl:exploration} where we explore methods of driving data collection to learn what we need to learn about the world.\looseness=-1

First, let us revisit one of our key observations when we first introduced the reinforcement learning problem.
Namely, that the observed transitions $\vxp$ and rewards $r$ are conditionally independent given the state-action pairs $(\vx, \va)$.\footnote{see \cref{eq:rl_cond_indep}}
This is due to the Markovian structure of the underlying Markov decision process.

This is the key observation that allows us to treat the estimation of the dynamics and rewards as a simple regression problem (or a density estimation problem when the quantities are stochastic rather than deterministic).
More concretely, we can estimate the dynamics and rewards off-policy using the standard supervised learning techniques we discussed in earlier chapters, from a replay buffer \begin{align}
  \spD = \{(\underbrace{\vx_t, \va_t}_\text{``input''}, \underbrace{r_t, \vx_{t+1}}_\text{``label''})\}_t.
\end{align}
Here, $\vx_t$ and $\va_t$ are the ``inputs'', and $r_t$ and $\vx_{t+1}$ are the ``labels'' of the regression problem.
Due to the conditional independence of the labels given the inputs, we have independent label noise (i.e., ``independent training data'') which is the basic assumption that we have been making throughout our discussion of techniques for probabilistic machine learning in \cref{part1}.

The key difference to supervised learning is that the set of inputs depends on how we act.
That is, the current inputs arise from previous policies, and the inputs which we will observe in the future will depend on the model (and policy) obtained from the current data: we have feedback loops!
We will come back to this aspect of reinforcement learning in the next section on exploration.
For now, recall we only assume that we have used an arbitrary policy to collect some data, which we then stored in a replay buffer, and which we now want to use to learn the ``best-possible'' model of our environment.

\subsection{Probabilistic Inference}

In the following, we will discuss how we can use the techniques from probabilistic inference, which we have seen in \cref{part1}, to learn the dynamics and reward models.
Thereby, we will focus on learning the transition model $f$ as learning the reward model $r$ is completely analogous.
For learning deterministic dynamics or rewards, we can use for example Gaussian processes (cf. \cref{sec:gp}) or deep neural networks (cf. \cref{sec:bdl}).
We will now focus on the setting where the dynamics are stochastic, that is, \begin{align}
  \vx_{t+1} \sim f(\vx_t, \va_t; \vpsi).
\end{align}

\begin{ex}{Conditional Gaussian dynamics}{}
  We could, for example, use a conditional Gaussian for the transition model, \begin{align}
    \vx_{t+1} \sim \N{\vmu(\vx_t, \va_t; \vpsi)}{\mSigma(\vx_t, \va_t; \vpsi)}. \label{eq:conditional_gaussian_dynamics}
  \end{align}
  As we have seen in \cref{eq:cholesky}, we can rewrite the covariance matrix $\mSigma$ as a product of a lower-triangular matrix $\mCalL$ and its transpose using the Cholesky decomposition ${\mSigma = \mCalL \transpose{\mCalL}}$ of $\mSigma$.
  This allows us to represent the model by only $n (n + 1) / 2$ parameters.
  Moreover, we have learned that Gaussians are reparameterizable, which we have seen to be useful for planning.

  Note that this model reduces to a deterministic model if the covariance is zero.
  So the stochastic transition models encompass all deterministic models.
  Moreover, in many applications (such as robotics), it is often useful to use stochastic models to attribute slight inaccuracies and measurement noise to a small uncertainty in the model.
\end{ex}

A first approach might be to obtain a point estimate for $f$, either through maximum likelihood estimation (which we have seen to overfit easily) or through maximum a posteriori estimation.
If, for example, $f$ is represented as a deep neural network, we have already seen how to find the MAP estimate of its weights in \cref{sec:bnn:map_inference}.

\begin{rmk}{The key pitfall of point estimates}{}
  However, using point estimates leads to a \emph{key pitfall} of model-based reinforcement learning.
  Using a point estimate of the model for planning --- even if this point estimate is very accurate --- often performs \emph{very poorly}.
  The reason is that planning is very good at overfitting (i.e., exploiting) small errors in the transition model.
  Moreover, the errors in the model estimate compound over time when using a longer time horizon $H$.
  The key to remedy this pitfall lies in being robust to misestimated models.
  This naturally suggests quantifying the uncertainty in our model estimate and taking it into account during planning.\safefootnote{Quantifying the uncertainty of an estimate is a problem that we have spent the first few chapters exploring. Notably, refer to \begin{itemize}
    \item \cref{sec:blr:uncertainty} for a description of epistemic and aleatoric uncertainty;
    \item \cref{sec:gp} for our use of uncertainty estimates in the context of Gaussian processes;
    \item \cref{sec:bdl} for our use of uncertainty estimates in the context of Bayesian deep learning; and
    \item \cref{sec:active_learning,sec:bayesian_optimization} for our use of epistemic uncertainty estimates to drive exploration.
  \end{itemize}}
  In the following section, we will rediscover that estimates of epistemic uncertainty are also extremely useful for driving (safe) exploration --- something that we have already encountered in our discussion of Bayesian optimization.\looseness=-1
\end{rmk}

In the following, we will differentiate between the epistemic uncertainty and the aleatoric uncertainty.
Recall from \cref{sec:blr:uncertainty} that epistemic uncertainty corresponds to our uncertainty about the model, $\r{p(f \mid \spD)}$, while aleatoric uncertainty corresponds to the uncertainty of the transitions in the underlying Markov decision process (which can be thought of as ``irreducible'' noise), $\b{p(\vx_{t+1} \mid f, \vx_t, \va_t)}$.

Intuitively, probabilistic inference of dynamics models corresponds to learning a distribution over possible models $f$ and $r$ given prior \emph{beliefs}, where $f$ and $r$ characterize the underlying Markov decision process.
This goes to show another benefit of the model-based over the model-free approach to reinforcement learning.
Namely, that it is much easier to encode prior knowledge about the transition and rewards model.

\begin{ex}{Inference with conditional Gaussian dynamics}{}
  Let us revisit inference with our conditional Gaussian dynamics model from \cref{eq:conditional_gaussian_dynamics}, \begin{align*}
    \vx_{t+1} \sim \N{\vmu(\vx_t, \va_t; \vpsi)}{\mSigma(\vx_t, \va_t; \vpsi)}.
  \end{align*}

  Recall that in the setting of Bayesian deep learning, most approximate inference techniques represented the approximate posterior using some form of a mixture of Gaussians,\safefootnote{see \begin{itemize}
    \item \cref{eq:bnn_posterior_vi} for variational inference;
    \item \cref{eq:bnn_posterior_mcmc} for Markov chain Monte Carlo;
    \item \cref{eq:bnn_posterior_dr} for dropout regularization; and
    \item \cref{eq:bnn_posterior_pe} for probabilistic ensembles.
  \end{itemize}} \begin{align}
    p(\vx_{t+1} \mid \spD, \vx_t, \va_t) \approx \frac{1}{m} \sum_{i=1}^m \N{\vmu(\vx_t, \va_t; \vpsi^{(i)})}{\mSigma(\vx_t, \va_t; \vpsi^{(i)})}.
  \end{align}
  Hereby, the epistemic uncertainty is represented by the variance between mixture components, and the aleatoric uncertainty by the average variance within the components.\safefootnote{see \cref{sec:bnn:approximate_inference:vi}}
\end{ex}

In supervised learning, we often conflated the notions of epistemic and aleatoric uncertainty.
In the context of planning, there is an important consequence of the decomposition into epistemic and aleatoric uncertainty.
Recall that the epistemic uncertainty corresponds to a distribution over Markov decision processes $f$, whereas the aleatoric uncertainty corresponds to the randomness in the transitions within one such MDP $f$.
Crucially, this randomness in the transitions must be consistent within a single MDP!
That is, once we selected a single MDP for planning, we should disregard the epistemic uncertainty and solely focus on the randomness of the transitions.
Then, to take into account epistemic uncertainty, we should average our plan across the different realizations of $f$.
This yields the following Monte Carlo estimate of our reward $\fnJ[H]$, \begin{align}
  \J{\va_{t:t+H-1}}[H] &\approx \frac{1}{m} \sum_{i=1}^m \J{\va_{t:t+H-1}; \r{f^{(i)}}}[H] \qquad\text{where} \label{eq:learning_with_planning} \\
  \J{\va_{t:t+H-1}; f}[H] &\defeq \sum_{\tau=t}^{t+H-1} \gamma^{\tau-t} r(\vx_\tau(\b{\vvarepsilon_{t:\tau-1}^{(i)}}; \va_{t:\tau-1}, f), \va_\tau) + \gamma^H \V{\vx_{t+H}}.
\end{align}
Here, $\r{f^{(i)} \iid p(f \mid \spD)}$ are independent samples of the transition model, and $\b{\vvarepsilon_{t:t+H-1}^{(i)} \iid \phi}$ parameterizes the dynamics analogously to \cref{eq:state_encoding}: \begin{align}\begin{split}
  &\vx_\tau(\vvarepsilon_{t:\tau-1}; \va_{t:\tau-1}, f) \\
  &\defeq f(\vvarepsilon_{\tau-1}; f(\dots; (\vvarepsilon_{t+1}; f(\vvarepsilon_t; \vx_t, \va_t), \va_{t+1}), \dots), \va_{\tau-1}).
\end{split}\end{align}
Observe that the epistemic and aleatoric uncertainty are treated differently.
Within a particular MDP $f$, we ensure that randomness (i.e., aleatoric uncertainty) is simulated consistently using our previous framework from our discussion of planning.\footnote{see \cref{eq:mbarl_planning_monte_carlo}}
The Monte Carlo samples of $f$ take into account the epistemic uncertainty about the transition model.
In our previous discussion of planning, we assumed the Markov decision process $f$ to be fixed.
Essentially, in \cref{eq:learning_with_planning} we are now using Monte Carlo trajectory sampling as a subroutine and average over an ``ensemble'' of Markov decision processes.

\begin{figure}
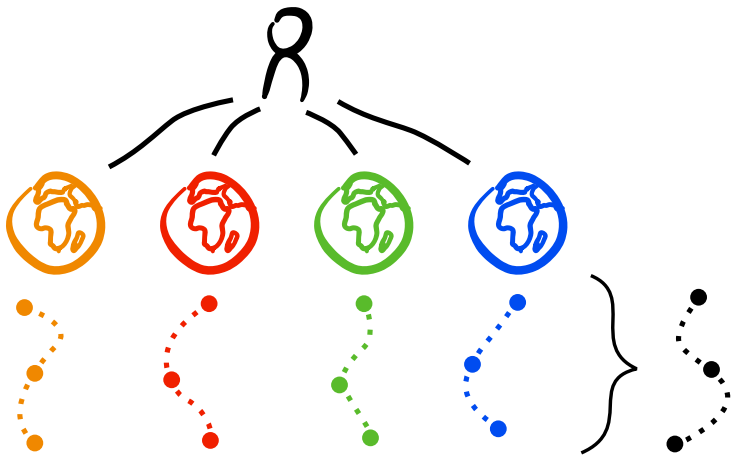

  \incfig{planning_epistemic_uncertainty}
  \caption{Illustration of planning with epistemic uncertainty and Monte Carlo sampling.
  The agent considers $m$ alternative ``worlds''. Within each world, it plans a sequence of actions over a finite time horizon.
  Then, the agent averages all optimal initial actions from all worlds.
  Crucially, each world by itself is \emph{consistent}.
  That is, its transition model (i.e., the aleatoric uncertainty of the model) is constant.}
\end{figure}

The same approach that we have seen in \cref{sec:mbarl:planning:parametric_policies} can be used to ``compile'' these plans into a parametric policy that can be trained offline, in which case, we write $\J{\pi}[H]$ instead of $\J{\va_{t:t+H-1}}[H]$.

This leads us to a first greedily exploitative algorithm for model-based reinforcement learning, which is shown in \cref{alg:rl_greedy}.
This algorithm is purely exploitative as it greedily maximizes the expected reward with respect to the transition model, taking into account epistemic uncertainty.\looseness=-1

\begin{algorithm}
  \caption{Greedy exploitation for model-based RL}\label{alg:rl_greedy}
  start with (possibly empty) data $\spD = \emptyset$ and a prior $p(f) = p(f \mid \spD)$\;
  \For{several episodes}{
    plan a new policy $\pi$ to (approximately) maximize, \vspace{-8pt}\begin{align}
      \max_\pi \E[f \sim p(\cdot \mid \spD)]{\J{\pi; f}[H]}
    \end{align}\vspace{-20pt}\;
    roll out policy $\pi$ to collect more data\;
    update posterior $p(f \mid \spD)$\;
  }
\end{algorithm}

In the context of Gaussian process models, this algorithm is called \midx{probabilistic inference for learning control}[idxpagebf] (PILCO) \citep{deisenroth2011pilco}, which was the first demonstration of how sample efficient model-based reinforcement learning can be.
As was mentioned in \cref{rmk:planning_as_inference}, the originally proposed PILCO uses moment matching instead of Monte Carlo averaging.

\begin{figure*}
  \includegraphics[width=\textwidth]{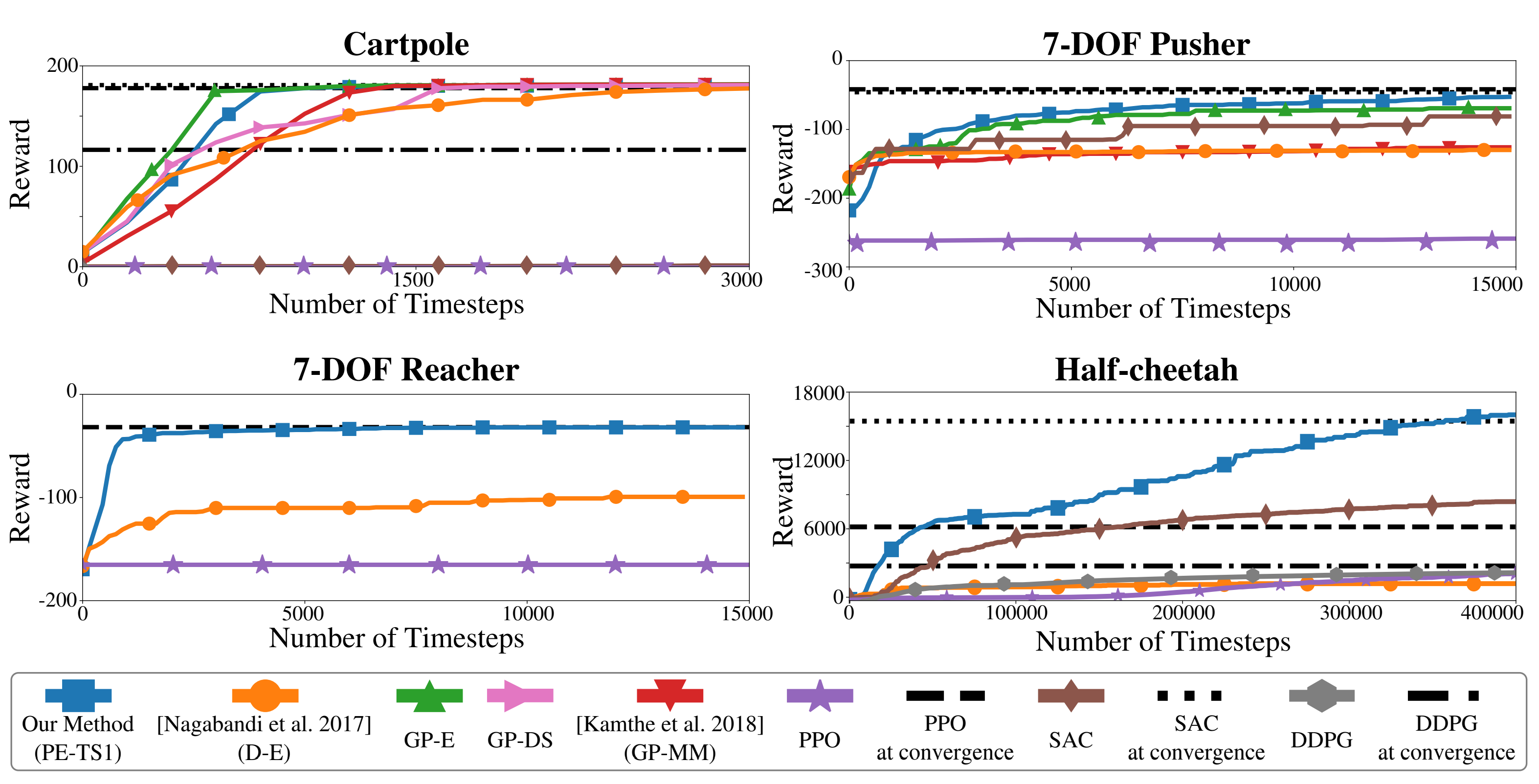}
  \caption{Sample efficiency of model-free and model-based reinforcement learning.
  DDPG and SAC are shown as constant (black) lines, because they take an order of magnitude more time steps before learning a good model.
  They also compare the PETS algorithm (blue) to deterministic ensembles (orange), where the transition models are assumed to be deterministic (or to have covariance $0$).
  Reproduced with permission from \icite{chua2018deep}.}
\end{figure*}

In the context of neural networks, this algorithm is called \midx{probabilistic ensembles with trajectory sampling} (PETS) \citep{chua2018deep}, which is one of the state-of-the-art algorithms.
PETS uses an ensemble of conditional Gaussian distributions over weights, trajectory sampling for evaluating the performance and model predictive control for planning.

Notably, PETS does not explicitly explore.
Exploration only happens due to the uncertainty in the model, which already drives exploration to some extent.
In many settings, however, this incentive is not sufficient for exploration --- especially when rewards are sparse.

\begin{oreadings}
  \begin{itemize}
    \item \textbf{PILCO:} \pcite{deisenroth2011pilco}
    \item \textbf{PETS:} \pcite{chua2018deep}
  \end{itemize}
\end{oreadings}

\subsection{Partial Observability}

Depending on the task, it may be difficult to learn the observed dynamics directly.
For example, when learning an artificial player for a computer game based only on the games' visual interface it may be difficult to predict the next frame in pixel space.
Instead, a common approach is to treat the visual interface as an observation in a POMDP with a hidden latent space (cf. \cref{sec:mdp:partial_observability}), and to learn the dynamics within the latent space and the observation model separately.

An example of this approach is the \midx{deep planning network} (PlaNet) algorithm which learns such a POMDP via variational inference and uses the cross-entropy method for closed-loop planning of action sequences \citep{hafner2019learning}.
The \midx{Dreamer} algorithm replaces the closed-loop planning of PlaNet by open-loop planning with entropy regularization \citep{hafner2019dream,hafner2020mastering}.
Notably, neither PlaNet nor Dreamer explicitly take into account epistemic model uncertainty during planning but rather use point estimates.

\section{Exploration}\label{sec:mbarl:exploration}\pidx{exploration-exploitation dilemma}%

Exploration is critical when interacting with unknown environments, such as in reinforcement learning.
We first encountered the exploration-exploitation dilemma in our discussion of Bayesian optimization, where we aimed at maximizing an unknown function in as little time as possible by making noisy observations.\footnote{see \cref{sec:bayesian_optimization}}
Within the framework of Bayesian optimization, we used so-called acquisition functions\pidx{acquisition function} for selecting the next point at which to observe the unknown function.
Observe that these acquisition functions are analogous to policies in the setting of reinforcement learning.
In particular, the policy that uses greedy exploitation like we have seen in the previous section is analogous to simply picking the point that maximizes the mean of the posterior distribution.
In the context of Bayesian optimization, we have already seen that this is insufficient for exploration and can easily get stuck in locally optimal solutions.
As reinforcement learning is a strict generalization of Bayesian optimization, there is no reason why such a strategy should be sufficient now.

Recall from our discussion of Bayesian optimization that we ``solved'' this problem by using the epistemic uncertainty in our model of the unknown function to pick the next point to explore.
This is what we will now explore in the context of reinforcement learning.

One simple strategy that we already investigated is the addition of some \midx{exploration noise}.
In other words, one follows a greedy exploitative strategy, but every once in a while, one chooses a random action (like in $\varepsilon$-greedy\pidx{$\varepsilon$-greedy}); or one adds additional noise to the selected actions (known as Gaussian noise ``dithering''\pidx{Gaussian noise ``dithering''}[idxpagebf]).
We have already seen that in difficult exploration tasks, these strategies are not systematic enough.

Two other approaches that we have considered before are Thompson sampling (cf. \cref{sec:bayesian_optimization:acquisition_functions:thompson_sampling}) and, more generally, the principle of \midx{optimism in the face of uncertainty} (cf. \cref{sec:bayesian_optimization:online_learning:mab}).

\subsection{Thompson Sampling}\pidx{Thompson sampling}

Recall from \cref{sec:bayesian_optimization:acquisition_functions:thompson_sampling} the main idea behind Thompson sampling: namely, that the randomness in the realizations of $f$ from the posterior distribution is already enough to drive exploration.
That is, instead of picking the action that performs best on average across all realizations of $f$, Thompson sampling picks the action that performs best for a \emph{single realization} of $f$.
The epistemic uncertainty in the realizations of~$f$ leads to variance in the picked actions and provides an additional incentive for exploration.
This yields \cref{alg:thompson_sampling} which is an immediate adaptation of greedy exploitation and straightforward to implement.

\begin{algorithm}
  \caption{Thompson sampling}\label{alg:thompson_sampling}
  start with (possibly empty) data $\spD = \emptyset$ and a prior $p(f) = p(f \mid \spD)$\;
  \For{several episodes}{
    sample a model $f \sim p(\cdot \mid \spD)$\;
    plan a new policy $\pi$ to (approximately) maximize, \vspace{-8pt}\begin{align}
      \max_\pi \J{\pi; f}[H]
    \end{align}\vspace{-20pt}\;
    roll out policy $\pi$ to collect more data\;
    update posterior $p(f \mid \spD)$\;
  }
\end{algorithm}

\subsection{Optimistic Exploration}\label{sec:mbarl:exploration:optimistic}\pidx{optimism in the face of uncertainty}

We have already seen in the context of Bayesian optimization and tabular reinforcement learning that optimism is a central pillar for exploration.
But how can we explore optimistically in model-based reinforcement learning?

Let us consider a set $\spM(\spD)$ of \midx<plausible models>{plausible model} given some data $\spD$.
Optimistic exploration would then optimize for the most advantageous model among all models that are plausible given the seen data.

\begin{ex}{Plausible conditional Gaussians}{}
  In the context of conditional Gaussians, we can consider the set of all models such that the prediction of a single dimension $i$ lies in some confidence interval, \begin{align}
    f_i(\vx, \va) \in \spC_{i} \defeq \begin{multlined}[t]
      [\mu_i(\vx,\va\mid\spD)-\beta_i\sigma_i(\vx,\va\mid\spD), \\ \mu_i(\vx,\va\mid\spD)+\beta_i\sigma_i(\vx,\va\mid\spD)],
    \end{multlined} \label{eq:plausible_gaussian_confidence_intervals}
  \end{align} where $\beta_i$ tunes the width of the confidence interval, analogously to \cref{sec:bayesian_optimization:acquisition_functions:ucb}.
  The set of plausible models is then given as \begin{align}
    \spM(\spD) \defeq \{f \mid \forall \vx \in \spX, \va \in \spA, i \in [d] : f_i(\vx, \va) \in \spC_{i}\}.
  \end{align}
\end{ex}

When compared to greedy exploitation, instead of taking the optimal step on average with respect to all realizations of the transition model~$f$, optimistic exploration as shown in \cref{alg:optimistic_exploration} takes the optimal step with respect to the most optimistic model among all transition models that are consistent with the data.

\begin{algorithm}
  \caption{Optimistic exploration}\pidx{optimistic exploration}\label{alg:optimistic_exploration}
  start with (possibly empty) data $\spD = \emptyset$ and a prior $p(f) = p(f \mid \spD)$\;
  \For{several episodes}{
    plan a new policy $\pi$ to (approximately) maximize, \vspace{-8pt}\begin{align}
      \max_\pi \max_{f \in \spM(\spD)} \J{\pi; f}[H]
    \end{align}\vspace{-20pt}\;
    roll out policy $\pi$ to collect more data\;
    update posterior $p(f \mid \spD)$\;
  }
\end{algorithm}

In general, the joint maximization over $\pi$ and $f$ is very difficult to solve computationally.
Yet, remarkably, it turns out that this complex optimization problem can be reduced to standard model-based reinforcement learning with a fixed model.
The key idea is to consider an agent that can control its ``luck''.
In other words, we assume that the agent believes it can control the outcome of its actions --- or rather choose which of the plausible dynamics it follows.
The ``luck'' of the agent can be considered as additional decision variables.
Consider the optimization problem, \begin{align}
  \pi_{t+1} \defeq \argmax_\pi \max_{\veta(\cdot) \in [-1,1]^d} \J{\pi; \widehat{f}_t}[H]
\end{align} with ``optimistic'' dynamics \begin{align}
  \widehat{f}_{t,i}(\vx, \va) \defeq \mu_{t, i}(\vx, \va) + \beta_{t, i} \eta_i(\vx, \va) \sigma_{t, i}(\vx, \va). \label{eq:dynamics_with_luck}
\end{align}
Here the decision variables $\eta_i$ control the variance of an action.
That is, within the confidence bounds of the transition model, the agent can freely choose the state that is reached by playing an action $\va$ from state $\vx$.
Essentially, this corresponds to maximizing expected reward in an augmented (optimistic) MDP with \emph{known} dynamics $\widehat{f}$ and with a larger action space that also includes the decision variables $\veta$.
This is a known MDP for which we can use our toolbox for planning which we developed in \cref{sec:mbarl:planning}.

\begin{figure}
  \begin{center}
    \import{./plots/output/}{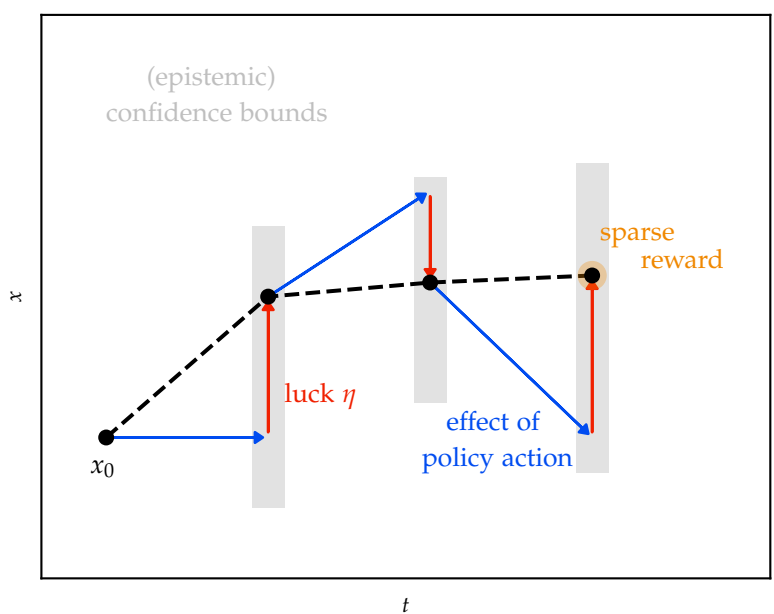}
  \end{center}

  \caption{Illustration of H-UCRL in a one-dimensional state space.
  The agent ``hallucinates'' that it takes the black trajectory when, in reality, the outcomes of its actions are as shown in blue.
  The agent can hallucinate to land anywhere within the gray confidence regions (i.e., the epistemic uncertainty in the model) using the luck decision variables $\veta$.
  This allows agents to discover long sequences of actions leading to sparse rewards.}\label{fig:hucrl}
\end{figure}

The algorithm that maximizes expected reward in this augmented MDP is called \midx{hallucinated upper confidence reinforcement learning} (H-UCRL) \citep{curi2020efficient,treven2023efficient}.
H-UCRL can be seen as the natural extension of the UCB acquisition function from Bayesian optimization to reinforcement learning.
An illustration of the algorithm is given in \cref{fig:hucrl}.

Intuitively, the agent has the tendency to believe that it can achieve much more than it actually can.
As more data is collected, the confidence bounds shrink and the optimistic policy rewards converge to the actual rewards.
Yet, crucially, we only collect data in regions of the state-action space that are more promising than the regions that we have already explored.
That is, we only collect data in the most promising regions.

Optimistic exploration yields the strongest effects for hard exploration tasks, for example, settings with large penalties associated with performing certain actions and settings with sparse rewards.\footnote{Action penalties are often used to discourage the agent from exhibiting unwanted behavior.
However, increasing the action penalty increases the difficulty of exploration.
Therefore, optimistic exploration is especially useful in settings where we want to practically disallow many actions by attributing large penalties to them.}
In those settings, most other strategies (i.e., those that are not sufficiently explorative), learn not to act at all.
However, even in settings of ``ordinary rewards'', optimistic exploration often learns good policies faster.

\subsection{Constrained Exploration}\label{sec:mbarl:exploration:constrained}

Besides making exploration more efficient, another use of uncertainty is to make exploration more safe.
Today, reinforcement learning is still far away from being deployed directly to the real world.
In practice, reinforcement learning is almost always used together with a simulator, in which the agent can ``safely'' train and explore.
Yet, in many domains, it is not possible to simulate the training process, either because we are lacking a perfect model of the environment, or because simulating such a model is too computationally inefficient.
This is where we can make use of uncertainty estimates of our model to avoid unsafe states.

\begin{marginfigure}
  \incfig{planning_with_conf_bounds}
  \caption{Illustration of planning with confidence bounds.
  The unsafe set of states is shown as the red region.
  The agent starts at the position denoted by the black dot and plans a sequence of actions.
  The confidence bounds on the subsets of states that are reached by this action sequence are shown in gray.}
\end{marginfigure}

Let us denote by $\spX_\mathrm{unsafe}$ the subset of unsafe states of our state space~$\spX$.
A natural idea is to perform planning using confidence bounds of our epistemic uncertainty.
This allows us to pessimistically forecast the plausible consequences of our actions, given what we have already learned about the transition model.
As we collect more data, the confidence bounds will shrink, requiring us to be less conservative over time.
This idea is also at the heart of fields like \midx{robust control}.

A general formalism for planning under constraints is the notion of \midx{constrained Markov decision processes} \citep{altman1999constrained}.
Given dynamics $f$, planning in constrained MDPs amounts to the following optimization problem: \begin{subequations}\begin{align}
  \max_\pi \quad&J_{\mu}(\pi; f) \defeq \E[\vx_0 \sim \mu, \vx_{1:\infty} \mid \pi, f]{\sum_{t=0}^\infty \gamma^t r_t} \\
  \text{subject to} \quad&J_{\mu}^{c}(\pi; f) \leq \delta
\end{align}\label{eq:cmdp_opt}\end{subequations} where $\mu$ is a distribution over the initial state and \begin{align}
  J_{\mu}^{c}(\pi; f) \defeq \E[\vx_0 \sim \mu, \vx_{1:\infty} \mid \pi, f]{\sum_{t=0}^\infty \gamma^t c(\vx_t)}
\end{align} are expected discounted costs with respect to a cost function ${c : \spX \to \Rzero}$.\footnote{It is straightforward to extend this framework to allow for multiple constraints.}
Observe that for the cost function $c(\vx) \defeq \Ind{\vx \in \spX_\mathrm{unsafe}}$, the value $J_{\mu}^c(\pi; f)$ can be interpreted as an upper bound on the (discounted) probability of visiting unsafe states under dynamics $f$,\footnote{This follows from a simple union bound \eqref{eq:union_bound}.} and hence, the constraint $J_{\mu}^c(\pi; f) \leq \delta$ bounds the probability of visiting an unsafe state when following policy $\pi$.\looseness=-1

The \midx{augmented Lagrangian method} can be used to solve the optimization problem of \cref{eq:cmdp_opt}.\footnote{For an introduction, read chapter 17 of \icite{wright1999numerical}.}
Thereby, one solves \begin{align}
  &\max_\pi \min_{\lambda \geq 0} \quad J_\mu(\pi; f) - \lambda (J_\mu^{c}(\pi; f) - \delta) \\
  &= \max_\pi \begin{cases}
    J_\mu(\pi; f) & \text{if $\pi$ is feasible} \\
    -\infty & \text{otherwise}.
  \end{cases}
\end{align}
Observe that if $\pi$ is feasible, then $J_\mu^{c}(\pi; f) \leq \delta$ and so the minimum over $\lambda$ is satisfied if $\lambda = 0$.
Conversely, if $\pi$ is infeasible, $\lambda$ can be made arbitrarily large to solve the optimization problem.
In practice, an additional penalty term is added to smooth the objective when transitioning between feasible and infeasible policies.

Note that solving constrained optimization problems such as \cref{eq:cmdp_opt} yields an optimal safe policy.
However, it is not ensured that constraints are not violated during the search for the optimal policy.
Generally, exterior penalty methods such as the augmented Lagrangian method allow for generating infeasible points during the search, and are therefore not suitable when constraints have to be strictly enforced at all times.
Thus, this method is more applicable in the episodic setting (e.g., when an agent is first ``trained'' in a simulated environment and then ``deployed'' to the actual task) rather than in the continuous setting where the agent has to operate in the ``real world'' from the beginning and cannot easily be reset.

\begin{rmk}{Barrier functions}{}
  The augmented Lagrangian method is merely one example of optimizing a joint objective of maximizing rewards and minimizing costs to find a policy with bounded costs.
  Another example of this approach are so-called \midx<Barrier functions>{Barrier function} which augment the reward objective with a smoothed approximation of the boundary of a set $\spX_\mathrm{unsafe}$.
  That is, one solves \begin{align}
    \max_\pi J_\mu(\pi; f) - \lambda \cdot B_\mu^c(\pi; f)
  \end{align} for some $\lambda > 0$ where the barrier function $B_\mu^c(\pi; f)$ goes to infinity as a state on the boundary of $\spX_\mathrm{unsafe}$ is approached.
  Examples for barrier functions are $- \log c(\vx)$ and $\frac{1}{c(\vx)}$.

  Barrier functions are an interior penalty method which ensure that points are feasible during the search for the optimal solution.
\end{rmk}

So far, we have assumed knowledge of the true dynamics to solve the optimization problem of \cref{eq:cmdp_opt}.
If we do not know the true dynamics but instead have access to a set of plausible models $\spM(\spD)$ given data $\spD$ (cf. \cref{sec:mbarl:exploration:optimistic}) which encompasses the true dynamics, then a natural strategy is to be \g{optimistic} with respect to future rewards and \r{pessimistic} with respect to future constraint violations.
More specifically, we solve the optimization problem, \begin{subequations}\begin{align}
  \max_\pi \g{\max_{f \in \spM(\spD)}} \quad&J_{\mu}(\pi; f) \margintag{\emph{optimistic}} \\
  \text{subject to} \quad&\r{\max_{f \in \spM(\spD)}} J_{\mu}^{c}(\pi; f) \leq \delta. \margintag{\emph{pessimistic}}
\end{align}\label{eq:lambda}\end{subequations}
Intuitively, jointly maximizing $J_\mu(\pi; f)$ with respect to $\pi$ and (plausible) $f$ can lead the agent to try behaviors with potentially high reward due to \emph{optimism} (i.e., the agent ``dreams'' about potential outcomes).
On the other hand, being \emph{pessimistic} with respect to constraint violations enforces the safety constraints (i.e., the agent has ``nightmares'' about potential outcomes).

If the policy values $J_\mu(\pi; f)$ and $J_\mu^{c}(\pi; f)$ are modeled as a distribution (e.g., using Bayesian deep learning), then the inner maximization over plausible dynamics can be approximated using samples from the posterior distributions.
Thus, the augmented Lagrangian method can also be used to solve the general optimization problem of \cref{eq:lambda}.
The resulting algorithm is known as \midx{Lagrangian model-based agent} (LAMBDA) \citep{as2022constrained}.

\subsection{Safe Exploration}\label{sec:mbarl:exploration:safe}

As noted in the previous section, in many settings we do not only want to eventually find a safe policy, but we also want to ensure that we act safely while searching for an optimal policy.
To this end, recall the general approach outlined in the beginning of the previous section wherein we plan using (pessimistic) confidence bounds of the plausible consequences of our actions.

One key challenge of this approach is that we need to forecast plausible trajectories.
The confidence bounds of such trajectories cannot be nicely represented anymore.
One approach is to over-approximate the confidence bounds over reachable states from trajectories.

\begin{thm}[informal, \cite{koller2018learning}]
  For conditional Gaussian dynamics, the reachable states of trajectories can be over-approximated with high probability.\looseness=-1
\end{thm}

\begin{marginfigure}
  \incfig{planning_long_term_consequences}
  \caption{Illustration of long-term consequence when planning a finite number of steps.
  Green dots are to denote safe states and the red dot is to denote an unsafe state.
  After performing the first action, the agent is still able to return to the previous state.
  Yet, after reaching the third state, the agent is already guaranteed to end in an unsafe state.
  When using only a finite horizon of $H=2$ for planning, the agent might make this transition regardless.}
\end{marginfigure}

Another key challenge is that actions might have consequences that exceed the time horizon used for planning.
In other words, by performing an action now, our agent might put itself into a state that is not yet unsafe, but out of which it cannot escape and which will eventually lead to an unsafe state.
You may think of a car driving towards a wall.
When a crash with the wall is designated as an unsafe state, there are quite a few states in advance at which it is already impossible to avoid the crash.
Thus, looking ahead a finite number of steps is not sufficient to prevent entering unsafe states.

It turns out that it is still possible to use the epistemic uncertainty about the model and short-term plausible behaviors to make guarantees about certain long-term consequences.
One idea is to also consider a set of safe states $\spX_\mathrm{safe}$ alongside the set of unsafe states $\spX_\mathrm{unsafe}$, for which our agent knows how to stay inside (i.e., remain safe).
In other words, for states $\vx \in \spX_\mathrm{safe}$, we know that our agent can always behave in such a way that it will not reach an unsafe state.\footnote{In the example of driving a car, the set of safe states corresponds to those states where we know that we can still safely brake before hitting the wall.}
An illustration of this approach is shown in \cref{fig:safe_learning_based_mpc}.

\begin{figure}
  \incfig{safe_learning_based_mpc}
  \caption{Illustration of long-term consequence when planning a finite number of steps.
  The unsafe set of states is shown in red and the safe set of states is shown in blue.
  The confidence intervals corresponding to actions of the performance plan and safety plan are shown in orange and green, respectively.}\label{fig:safe_learning_based_mpc}
\end{figure}

The problem is that this set of safe states might be very conservative.
That is to say, it is likely that rewards are mostly attained \emph{outside} of the set of safe states.
The key idea is to plan two sequences of actions, instead of only one.
``Plan A'' (the \midx{performance plan}) is planned with the objective to solve the task, that is, attain maximum reward.
``Plan B'' (the \midx{safety plan}) is planned with the objective to return to the set of safe states $\spX_\mathrm{safe}$.
In addition, we enforce that both plans must agree on the first action to be played.

Under the assumption that the confidence bounds are conservative estimates, this guarantees that after playing this action, our agent will still be in a state of which it can return to the set of safe states.
In this way, we can gradually increase the set of states that are safe to explore.
It can be shown that under suitable conditions, returning to the safe set can be guaranteed \citep{koller2018learning}.

\subsection{Safety Filters}

An alternative approach is to (slightly) adjust a potentially unsafe policy $\pi$ to obtain a policy $\hat{\pi}$ which avoids entering unsafe states with high probability.

Following our interpretation of constrained policy optimization in terms of optimism and pessimism from \cref{sec:mbarl:exploration:constrained}, to pessimistically evaluate the safety of a policy with respect to a cost function $c$ given a set of plausible models $\spM(\spD)$, we can use \begin{align}
  C^\pi(\vx) \defeq \max_{f \in \spM(\spD)} J_{\delta_\vx}^c(\pi; f) \margintag{$\delta_\vx$ is the point density at $\vx$} \label{eq:pessimistic_cost_estimate}
\end{align} where the initial-state distribution $\delta_\vx$ is to denote that the initial state is $\vx$.
Observe that \cref{eq:pessimistic_cost_estimate} permits a reparameterization in terms of additional decision variables $\veta$ which is analogous to our discussion in \cref{sec:mbarl:exploration:optimistic}.
Concretely, we have \begin{align}
  C^\pi(\vx) = \max_\veta J_{\delta_\vx}^c(\pi; \widehat{f})
\end{align} where $\widehat{f}$ are the adjusted dynamics \eqref{eq:dynamics_with_luck} which are based on the ``luck'' variables $\veta$.
In the context of estimating costs which we aim to minimize (as opposed to rewards which we aim to maximize), $\veta$ can be interpreted as the ``bad luck'' of the agent.

When our only objective is to act safely, that is, we only aim to minimize cost and are indifferent to rewards, then this reparameterization allows us to find a ``maximally safe'' policy, \begin{align}
  \pi_{\mathrm{safe}} \defeq \argmin_\pi \E[\vx \sim \mu]{C^\pi(\vx)} = \argmin_\pi \max_\veta J_{\mu}^c(\pi; \widehat{f}).
\end{align}
Under some conditions $\pi_{\mathrm{safe}}$ can be shown to satisfy $\pi_{\mathrm{safe}}(\vx) = \pi_{\vx}(\vx)$ for any $\vx$ where $\pi_{\vx} \defeq \argmin_\pi C^\pi(\vx)$ \citep[proposition 4.2]{curi2022safe}.
On its own, the policy $\pi_{\mathrm{safe}}$ is rather useless for exploring the state space.
In particular, when in a state that we already deem safe, following policy $\pi_{\mathrm{safe}}$, the agent aims simply to ``stay'' within the set of safe states which means that it has no incentive to explore / maximize reward.\looseness=-1

Instead, one can interpret $\pi_{\mathrm{safe}}$ as a ``backup'' policy in case we realize upon exploring that a certain trajectory is too dangerous, akin to our notion of the ``safety plan'' in \cref{sec:mbarl:exploration:safe}.
That is, given any (potentially explorative and dangerous) policy $\pi$, we can find the adjusted policy,\looseness=-1 \begin{subequations}\begin{align}
  \hat{\pi}(\vx) \defeq \argmin_{\va \in \spA} \quad&d(\pi(\vx), \va) \\
  \text{subject to} \quad&\max_\veta C^{\pi_{\mathrm{safe}}}(\widehat{f}(\vx, \va)) \leq \delta
\end{align}\end{subequations} for some metric $d(\cdot, \cdot)$ on $\spA$.
The constraint ensures that the pessimistic next state $\widehat{f}(\vx, \va)$ is recoverable by following policy $\pi_{\mathrm{safe}}$.
In this way, \begin{align}\begin{split}
  \tilde{\pi}(\vx) \defeq \begin{cases}
    \hat{\pi}(\vx) & \text{if $C^{\pi_{\mathrm{safe}}}(\vx) \leq \delta$} \\
    \pi_{\mathrm{safe}}(\vx) & \text{if $C^{\pi_{\mathrm{safe}}}(\vx) > \delta$} \\
  \end{cases}
\end{split}\end{align} is the policy ``closest'' to $\pi$ which is $\delta$-safe with respect to the pessimistic cost estimates $C^{\pi_{\mathrm{safe}}}$ \citep{curi2022safe}.\footnote{The policy $\tilde{\pi}$ is required as, a priori, it is not guaranteed that the state $\vx_t$ will satisfy $C^{\pi_{\mathrm{safe}}}(\vx_t) \leq \delta$ for all $t$ unless $\hat{\pi}$ is replanned after every step.}
This is also called a \midx{safety filter}.
Similar approaches using backup policies also allow safe exploration in the model-free setting \citep{sukhija2022scalable,widmer2023tuning}.

\begin{oreadings}
  \begin{itemize}
    \item \pcite{curi2020efficient}
    \item \pcite{berkenkamp2017safe}
    \item \pcite{koller2018learning}
    \item \pcite{as2022constrained}
    \item \pcite{curi2022safe}
    \item \pcite{turchetta2019safe}
  \end{itemize}
\end{oreadings}

\section*{Discussion}

In this final chapter, we learned that leveraging a learned ``world model'' for planning can be dramatically more sample-efficient than model-free reinforcement learning.
Additionally, such world models allow for effective multitask learning since they can be reused across tasks and only the reward function needs to be swapped out.\footnote{A similar idea, though often classified as model-free, is to learn a goal-conditioned policy which is studied extensively in \midx{goal-conditioned reinforcement learning}~\citep{andrychowicz2017hindsight,plappert2018multi,park2024ogbench}.}
Finally, we explored how to use probabilistic inference of our world model to drive exploration more effectively and safely.

\excheading

\begin{nexercise}{Bounding the regret of H-UCRL}{hucrl_regret}
  Analogously to our analysis of the UCB acquisition function for Bayesian optimization, we can use optimism to bound the regret of H-UCRL.
  We assume for simplicity that $d = 1$ and drop the index $i$ in the following.
  Let us denote by $x_{t,k}$ the $k$-th state visited during episode $t$ when following policy $\pi_t$.
  We denote by $\widehat{x}_{t,k}$ the corresponding state but under the optimistic dynamics $\widehat{f}$ rather than the true dynamics $f$.
  The instantaneous regret during episode $t$ is given by \begin{align}
    r_t = J_H(\pis; f) - J_H(\pi_t; f)
  \end{align} where we take the objective to be $J_H(\pi; f) = \sum_{k=0}^{H-1} r(x_{t,k}, \pi(x_{t,k}))$.

  You may assume that $f(x, \pi(x))$, $r(x, \pi(x))$, and $\sigma(x, \pi(x))$ are Lipschitz continuous in $x$.
  \begin{enumerate}
    \item Show by induction that for any $k \geq 0$, with high probability, \begin{align}
      \norm{\widehat{x}_{t,k} - x_{t,k}} \leq 2 \beta_t \sum_{l=0}^{k-1} \alpha_t^{k-1-l} \sigma_{t-1}(x_{t,l}, \pi_t(x_{t,l})) \label{eq:hucrl_regret:induction}
    \end{align} where $\alpha_t$ depends on the Lipschitz constants and $\beta_t$.

    \item Assuming w.l.o.g. that $\alpha_t \geq 1$, show that with high probability, \begin{align}
      r_t \leq \BigO{\beta_t H \alpha_t^{H-1} \sum_{k=0}^{H-1} \sigma_{t-1}(x_{t,k}, \pi_t(x_{t,k}))}
    \end{align}

    \item Let $\Gamma_T \defeq \max_{\pi_1, \dots \pi_T} \sum_{t=1}^T \sum_{k=0}^{H-1} \sigma_{t-1}^2(x_{t,k}, \pi_t(x_{t,k}))$.
    Analogously to \cref{exercise:bayesian_regret_for_gp_ucb}, it can be derived that $\Gamma_T \leq \BigO{H \gamma_T}$ if the dynamics are modeled by a Gaussian process.\safefootnote{For details, see appendix A of \cite{treven2023efficient}.}
    Bound the cumulative regret $R_T = \sum_{t=1}^T r_t \leq \BigO{\beta_{T} H^{\frac{3}{2}} \alpha_T^{H-1} \sqrt{T \Gamma_T}}$.
  \end{enumerate}
  Thus, if the dynamics are modeled by a Gaussian process with kernel such that $\gamma_T$ is sublinear, the regret of H-UCRL is sublinear.
\end{nexercise}

  \appendix

  \chapter{Mathematical Background}\label{sec:mathematical_background}

\section{Probability}

\subsection{Common discrete distributions}\label{sec:background:probability:common_discrete_distributions}

\begin{itemize}
  \item \midx<Bernoulli>{Bernoulli distribution} \quad $\Bern{p}$ describes (biased) coin flips.
  Its domain is ${\Omega = \{0,1\}}$ where $1$ corresponds to heads and $0$ corresponds to tails. $p \in [0,1]$ is the probability of the coin landing heads, that is, ${\Pr{X = 1} = p}$ and ${\Pr{X = 0} = 1 - p}$.
  \item \midx<Binomial>{binomial distribution} \quad $\Bin{n}{p}$ counts the number of heads in $n$ independent Bernoulli trials, each with probability of heads $p$.
  \item \midx<Categorical>{categorical distribution} \quad $\mathrm{Cat}(m, p_1, \dots, p_m)$ is a generalization of the Bernoulli distribution and represents throwing a (biased) $m$-sided die.
  Its domain is ${\Omega = [m]}$ and we have ${\Pr{X = i} = p_i}$. We require $p_i \geq 0$ and $\sum_{i \in [m]} p_i = 1$.
  \item \midx<Multinomial>{multinomial distribution} \quad $\mathrm{Mult}(n, m, p_1, \dots, p_m)$ counts the number of outcomes of each side in $n$ independent Categorical trials.
  \item \midx<Uniform>{uniform distribution} \quad $\Unif{\sS}$ assigns identical probability mass to all values in the set $\sS$. That is, $\Pr{X = x} = \frac{1}{|\sS|} \; (\forall x \in \sS)$.
\end{itemize}

\subsection{Probability simplex}\label{sec:background:probability:probability_simplex}

We denote by $\Delta^m$ the set of all categorical distributions on $m$ classes. Observe that $\Delta^m$ is an ${(m-1)}$-dimensional convex polytope.

To see this, let us consider the $m$-dimensional space of probabilities~$[0,1]^m$.
It follows from our characterization of the categorical distribution in \cref{sec:background:probability:common_discrete_distributions} that there is a one-to-one correspondence between probability distributions over $m$ classes and points in the space~$[0,1]^m$ where all coordinates sum to one.
This $(m-1)$-dimensional subspace of~$[0,1]^m$ is also known as the \midx{probability simplex}[idxpagebf].

\subsection{Universality of the Uniform and Sampling}\label{sec:inverse_transform_sampling}

Sampling from a continuous random variable is a common and non-trivial problem.
A family of techniques is motivated by the following general property of the uniform distribution, colloquially known as the ``universality of the uniform''.

\begin{marginbox}{Uniform distribution}
  The (continuous) \midx{uniform distribution}[idxpagebf] $\Unif{[a,b]}$ is the only distribution that assigns constant density to all points in the support $[a,b]$. That is, it has PDF \begin{align*}
    p(u) = \begin{cases}
      \frac{1}{b - a} & \text{if $u \in [a,b]$} \\
      0 & \text{otherwise} \\
    \end{cases}
  \end{align*} and CDF \begin{align*}
    P(u) = \begin{cases}
      \frac{u - a}{b - a} & \text{if $u \in [a,b]$} \\
      0 & \text{otherwise}. \\
    \end{cases}
  \end{align*}
\end{marginbox}

First, given a random variable $X \sim P$, we call \begin{align}
  \inv{P}(u) \defeq \min\{x \mid P(x) \geq u\} \quad\text{for all $0 < u < 1$}.
\end{align} the \midx{quantile function} of $X$.
That is, $\inv{P}(u)$ corresponds to the value $x$ such that the probability of $X$ being at most $x$ is $u$.
If the CDF $P$ is invertible, then $\inv{P}$ coincides with the inverse of $P$.

\begin{thm}[Universality of the uniform]\pidx{universality of the uniform}[idxpagebf]
  If $U \sim \Unif{[0,1]}$ and $P$ is invertible, then $\inv{P}(U) \sim P$.
\end{thm}
\begin{proof}
  Let $U \sim \Unif{[0,1]}$. Then, \begin{align*}
    \Pr{\inv{P}(U) \leq x} = \Pr{U \leq P(x)} = P(x). \qedhere \margintag{using $\Pr{U \leq u} = u$ if $U \sim \Unif{[0,1]}$}
  \end{align*}
\end{proof}

This implies that if we are able to sample from $\Unif{[0,1]}$,\safefootnote{This is done in practice using so-called \midx{pseudo-random number generators}.} then we are able to sample from any distribution with invertible CDF.
This method is known as \midx{inverse transform sampling}.

In the case of Gaussians, we learned that the CDF cannot be expressed in closed-form (and hence, is not invertible), however, for practical purposes, the quantile function of Gaussians can be approximated well.

\subsection{Point densities and Dirac's delta function}\label{sec:background:probability:dirac_delta}

The \midx{Dirac delta function} $\delta_\alpha$ is a function satisfying $\delta_\alpha(x) = 0$ for any~${x \neq \alpha}$ and $\int \delta_\alpha(x) \,dx = 1$.
$\delta_\alpha$ is also called a \midx{point density} at $\alpha$.

As an example, let us consider the random variable $Y = g(X)$, which is defined in terms of another random variable $X$ and a continuously differentiable function $g : \R \to \R$.
Using the sum rule and product rule, we can express the PDF of $Y$ as \begin{align}
  p(y) = \int p(y \mid x) \cdot p(x) \,d x = \int \delta_{g(x)}(y) \cdot p(x) \,d x.
\end{align}

In \cref{sec:fundamentals:probability:cov}, we discuss how one can obtain an explicit representation of $p_Y$ using the ``change of variables'' formula.

\subsection{Gradients of expectations}\label{sec:background:probability:gradients_of_expectations}

We often encounter gradients of expectations, $\grad_\vtheta \E[\rX]{f(\rX,\vtheta)}$.

\begin{marginbox}{Gradient}
  The \idx{gradient} of a function $f : \R^n \to \R$ at a point $\vx \in \R^n$ is \begin{align}
    \grad f(\vx) \defeq \transpose{\begin{bmatrix}
      \pdv{f(\vx)}{\vx(1)} & \cdots & \pdv{f(\vx)}{\vx(n)} \\
   \end{bmatrix}}.
  \end{align}
\end{marginbox}

\begin{fct}[Differentiation under the integral sign]\pidx{differentiation under the integral sign}
  Let $f(\vx, t)$ be differentiable in $\vx$, integrable in $t$, and be such that \begin{align*}
    \abs{\pdv{f(\vx, t)}{t}} \leq g(\vx)
  \end{align*} for some integrable function $g$.
  Then, if the distribution of $\rX$ is not parameterized by $t$,\safefootnote{That is, $\E{f(\rX, t)} = \int p(\vx) \cdot f(\vx, t) \,d\vx$ and $p(\vx)$ does not depend on $t$.} \begin{align}
    \pdv{\E{f(\rX, t)}}{t} = \E{\pdv{f(\rX, t)}{t}}.
  \end{align}
\end{fct}

Therefore, if the distribution of $\rX$ is not parameterized by $\vtheta$, \begin{align}
  \grad_\vtheta \E{f(\rX,\vtheta)} = \E{\grad_\vtheta f(\rX,\vtheta)}. \label{eq:swap_grad_exp_order}
\end{align}

\section{Quadratic Forms and Gaussians}\label{sec:fundamentals:qf}

\begin{defn}[Quadratic form]\pidx{quadratic form}
  Given a symmetric matrix $\mA \in \R^{n \times n}$, the \emph{quadratic form} induced by $\mA$ is defined as \begin{align}
    f_\mA : \R^n \to \R, \quad \vx \mapsto \transpose{\vx} \mA \vx = \sum_{i=1}^n \sum_{j=1}^n \mA(i,j) \cdot \vx(i) \cdot \vx(j). \label{eq:quadratic_form}
  \end{align}
\end{defn}

We call $\mA$ a \midx{positive definite} matrix if all eigenvalues of $\mA$ are positive.
Equivalently, we have $f_\mA(\vx) > 0$ for all $\vx \in \R^n \setminus \{\vzero\}$ and $f_\mA(\vzero) = 0$.
Similarly, $\mA$ is called \midx{positive semi-definite} if all eigenvalues of $\mA$ are non-negative, or equivalently, if $f_\mA(\vx) \geq 0$ for all $\vx \in \R^n$.
In particular, if $\mA$ is positive definite then $\sqrt{f_\mA(\vx)}$ is a norm (called the \midx{Mahalanobis norm} induced by $\mA$), and is often denoted by $\norm{\vx}_\mA$.

If $\mA$ is positive definite, then the sublevel sets of its induced quadratic form $f_\mA$ are convex ellipsoids.
Not coincidentally, the same is true for the sublevel sets of the PDF of a normal distribution $\N{\vmu}{\mSigma}$, which we have seen an example of in \cref{fig:multivariate_normal}. Hereby, the axes of the ellipsoid and their corresponding squared lengths are the eigenvectors and eigenvalues of $\mSigma$, respectively.

\begin{rmk}{Correspondence of quadratic forms and Gaussians}{qf_and_gaussians}
  Observe that the PDF of a zero-mean multivariate Gaussian $\N{\vzero}{\mSigma}$ is an exponentiated and appropriately scaled quadratic form induced by the positive definite precision matrix $\inv{\mSigma}$.
  The constant factor is chosen such that the resulting function is a valid probability density function, that is, sums to one.
\end{rmk}

One important property of positive definiteness of $\mSigma$ is that $\mSigma$ can be decomposed into the product of a lower-triangular matrix with its transpose.
This is known as the \emph{Cholesky decomposition}.

\begin{fct}[Cholesky decomposition, symmetric matrix-form]\pidx{Cholesky decomposition}
  For any symmetric and positive (semi-)definite matrix $\mA \in \R^{n \times n}$, there is a decomposition of the form \begin{align}
    \mA = \mCalL\transpose{\mCalL} \label{eq:cholesky}
  \end{align} where $\mCalL \in \R^{n \times n}$ is lower triangular and positive (semi-)definite.
\end{fct}

We will not prove this fact, but it is not hard to see that a decomposition exists (it takes more work to show that $\mCalL$ is lower-triangular).

Let $\mA$ be a symmetric and positive (semi-)definite matrix.
By the spectral theorem of symmetric matrices, $\mA = \mV\mLambda\transpose{\mV}$, where $\mLambda$ is a diagonal matrix of eigenvalues and $\mV$ is an orthonormal matrix of corresponding eigenvectors.
Consider the matrix \begin{align}
  \msqrt{\mA} \defeq \mV \msqrt{\mLambda} \transpose{\mV} \label{eq:matrix_sqrt}
\end{align} where $\msqrt{\mLambda} \defeq \diag{\sqrt{\mLambda(i, i)}}$, also called the \midx{square root} of $\mA$.
Then, \begin{align}
  \msqrt{\mA} \msqrt{\mA} &= \mV \msqrt{\mLambda} \transpose{\mV} \mV \msqrt{\mLambda} \transpose{\mV} \nonumber \\
  &= \mV \msqrt{\mLambda} \msqrt{\mLambda} \transpose{\mV} \nonumber \\
  &= \mV \mLambda \transpose{\mV} = \mA.
\end{align}
It is immediately clear from the definition that $\msqrt{\mA}$ is also symmetric and positive (semi-)definite.

Quadratic forms of positive semi-definite matrices are a generalization of the Euclidean norm, as \begin{align}
  \norm{\vx}_{\mA}^2 = \transpose{\vx} \mA \vx = \transpose{\vx} \msqrt{\mA} \msqrt{\mA} \vx = \transpose{(\msqrt{\mA} \vx)} \msqrt{\mA} \vx = \norm*{\msqrt{\mA} \vx}_2^2,
\end{align} and in particular, \begin{align}
  \log \N[\vx]{\vmu}{\mSigma} &= -\frac{1}{2}\norm*{\vx-\vmu}_{\inv{\mSigma}}^2 + \const \\
  &= -\frac{1}{2} \brackets*{\transpose{\vx}\inv{\mSigma}\vx - 2 \transpose{\vmu}\inv{\mSigma}\vx} + \const, \label{eq:gaussian_propto} \\
  \log \SN[\vx] &= -\frac{1}{2}\norm{\vx}_2^2 + \const. \label{eq:isotropic_gaussian_propto}
\end{align}

\section{Parameter Estimation}\label{sec:background:parameter_estimation}

In this section, we provide a more detailed account of parameter estimation as outlined in \cref{sec:fundamentals:supervised_learning}.

\subsection{Estimators}\label{sec:fundamentals:parameter_esitmation:estimators}

Suppose we are given a collection of independent samples $\vx_1,\dots,\vx_n$ from some random vector $\rX$.
Often, the exact distribution of $\rX$ is unknown to us, but we still want to ``estimate'' some property of this distribution, for example its mean.
We denote the property that we aim to estimate from our sample by $\opt{\vtheta}$.
For example, if our goal is estimating the mean of $\rX$, then $\opt{\vtheta} \defeq \E{\rX}$.

An \midx{estimator} for a parameter $\opt{\vtheta}$ is a random vector $\vthetahat_n$ that is a function of $n$ sample variables $\rX_1,\dots,\rX_n$ whose distribution is identical to the distribution of $\rX$.
Any concrete sample $\{\vx_i \sim \rX_i\}_{i=1}^n$ yields a concrete estimate of $\opt{\vtheta}$.

\begin{ex}{Estimating expectations}{estimating_expectations}
  The most common estimator for $\E{\rX}$ is the \midx{sample mean} \begin{align}
    \mean{\rX}_n \defeq \frac{1}{n} \sum_{i=1}^n \rX_i \label{eq:sample_mean}
  \end{align} where $\rX_i \iid \rX$.
  Using a sample mean to estimate an expectation is often also called \emph{Monte Carlo sampling}, and the resulting estimate is referred to as a \emph{Monte Carlo average}.
\end{ex}

\begin{ex}{MLE of Bernoulli random variables}{mle_bern}
  Let us consider an i.i.d. sample $x_{1:n}$ of $\Bern{p}$ distributed random variables.
  We want to estimate the parameter $p$ using a MLE.
  We have, \begin{align}
    \hat{p}_\MLE &= \argmax_p \Pr{x_{1:n} \mid p} \nonumber \\
    &= \argmax_p \sum_{i=1}^n \log \Pr{x_i \mid p} \nonumber \\
    &= \argmax_p \sum_{i=1}^n \log p^{x_i} (1-p)^{1-x_i} \margintag{using the Bernoulli PMF, see \cref{sec:background:probability:common_discrete_distributions}} \nonumber \\
    &= \argmax_p \sum_{i=1}^n x_i \log p + (1-x_i) \log(1-p). \nonumber
  \intertext{Computing the first derivative with respect to $p$, we see that the objective is maximized by}
    &= \frac{1}{n} \sum_{i=1}^n x_i.
  \end{align}
  Thus, the maximum likelihood estimate for the parameter $p$ of $X \sim \Bern{p}$ coincides with the sample mean $\mean{X}_n$.
\end{ex}

What does it mean for $\vthetahat_n$ to be a ``good estimate'' of $\opt{\vtheta}$?
There are two canonical measures of goodness of an estimator: its bias and its variance.

Clearly, we want $\E{\vthetahat_n} = \opt{\vtheta}$.
Estimators that satisfy this property are called \midx<unbiased>{unbiased estimator}.
The \midx{bias}, $\E[\vthetahat_n]{\vthetahat_n-\opt{\vtheta}}$, of an unbiased estimator is $\vzero$.
It follows directly from linearity of expectation \eqref{eq:linearity_expectation} that the sample mean is unbiased.
In \cref{sec:fundamentals:mc_approx}, we will see that the variance of the sample mean is small for reasonably large $n$ for ``light-tailed'' distributions.

\begin{ex}{Estimating variances}{estimating_variances}
  Analogously to the sample mean, the most common estimator for the covariance matrix $\Var{\rX}$ is the \midx{sample variance} (also called \midx{sample covariance matrix}) \begin{align}
    \mS_n^2 &\defeq \frac{1}{n-1} \sum_{i=1}^n (\rX_i - \mean{\rX}_n) \transpose{(\rX_i - \mean{\rX}_n)} \label{eq:sample_variance} \\
    &= \frac{n}{n-1} \parentheses*{\frac{1}{n} \sum_{i=1}^n \rX_i \transpose{{\rX_i}} - \mean{\rX}_n \transpose{\mean{\rX}_n}} \label{eq:sample_variance2}
  \end{align} where $\rX_i \iid \rX$.
  It can be shown that the sample variance is unbiased \exerciserefmark{sample_variance}.
  Intuitively, the normalizing factor is $\nicefrac{1}{n-1}$ because $\mS_n^2$ depends on the sample mean $\mean{\rX}_n$, which is obtained using the same samples.
  For this reason, $\mS_n^2$ has $n-1$ degrees of freedom.\safefootnote{That is, any sample $\rX_i$ can be recovered using all other samples and the sample mean.}
\end{ex}

The second desirable property of an estimator $\vthetahat_n$ is that its variance is small.\footnote{The variance of estimators vector-valued estimators $\vthetahat_n$ is typically studied component wise.}
A common measure for the variance of an estimator of $\theta$ is the \midx{mean squared error}[idxpagebf], \begin{align}
  \mathrm{MSE}(\hat{\theta}_n) \defeq \E[\hat{\theta}_n]{(\hat{\theta}_n-\opt{\theta})^2}. \label{eq:mse}
\end{align}
The mean squared error can be decomposed into the estimator's bias and variance: \begin{align}
  \mathrm{MSE}(\hat{\theta}_n) &= \E[\hat{\theta}_n]{(\hat{\theta}_n - \opt{\theta})^2} = \Var{\hat{\theta}_n} + (\E{\hat{\theta}_n} - \opt{\theta})^2, \margintag{using \eqref{eq:variance2} and $\Var[\hat{\theta}_n]{\hat{\theta}_n - \opt{\theta}} = \Var{\hat{\theta}_n}$} \label{eq:mse_bias_variance}
\end{align} the mean squared error of an estimator can be written as the sum of its variance and its squared bias.

A desirable property is for $\hat{\theta}_n$ to converge to $\opt{\theta}$ in probability \eqref{eq:p_convergence}: \begin{align*}
  \forall \epsilon > 0: \quad \lim_{n\to\infty} \Pr{\abs{\hat{\theta}_n - \opt{\theta}} > \epsilon} = 0.
\end{align*}
Such estimators are called \midx<consistent>{consistent estimator}, and a sufficient condition for consistency is that the mean squared error converges to zero as $n \to \infty$.\footnote{It follows from Chebyshev's inequality \eqref{eq:chebyshev_inequality} that $\Pr{\abs{\hat{\theta}_n - \opt{\theta}} > \epsilon} \leq \frac{\mathrm{MSE}(\hat{\theta}_n)}{\epsilon^2}$.}

Consistency is an asymptotic property.
In practice, one would want to know ``how quickly'' the estimator converges as $n$ grows.
To this end, an estimator $\hat{\theta}_n$ is said to be \midx{sharply concentrated} around $\opt{\theta}$ if \begin{align}
  \forall \epsilon > 0: \quad \Pr{\abs{\hat{\theta}_n - \opt{\theta}} > \epsilon} \leq \exp(-\Omega(\epsilon)),
\end{align} where $\Omega(\epsilon)$ denotes the class of functions that grow at least linearly in~$\epsilon$.\footnote{That is, $h \in \Omega(\epsilon)$ if and only if $\lim_{\epsilon \to \infty} \frac{h(\epsilon)}{\epsilon} > 0$. With slight abuse of notation, we force $h$ to be positive (so as to ensure that the argument to the exponential function is negative) whereas in the traditional definition of Landau symbols, $h$ is only required to grow linearly in absolute value.}
Thus, if an estimator is sharply concentrated, its absolute error is bounded by an exponentially quickly decaying error probability.

\subsection{Heavy Tails}\label{sec:fundamentals:parameter_esitmation:heavy_tails}

It is often said that a sharply concentrated estimator $\hat{\theta}_n$ has \midx{small tails}, where ``tails'' refer to the ``ends'' of a PDF.
Let us examine the difference between a \midx<light-tailed>{light-tailed distribution} and a \midx<heavy-tailed>{heavy-tailed distribution} distribution more closely.

\begin{marginfigure}
  \begin{center}
    \import{./plots/output/}{heavy_tails.pgf}
  \end{center}

	\caption{Shown are the right tails of the PDFs of a \textbf{Gaussian} with mean $1$ and variance $1$, a \textbf{\b{exponential distribution}} with mean $1$ and parameter $\lambda=1$, and a \textbf{\r{log-normal distribution}} with mean $1$ and variance $1$ on a log-scale.}
\end{marginfigure}

\begin{defn}
  A distribution $P_X$ is said to have a \emph{heavy (right) tail} if its \midx{tail distribution function} \begin{align}
    \mean{P}_X(x) \defeq 1 - P_X(x) = \Pr{X > x}
  \end{align} decays slower than that of the exponential distribution, that is, \begin{align}
    \limsup_{x\to\infty} \frac{\mean{P}_X(x)}{e^{-\lambda x}} = \infty
  \end{align} for all $\lambda > 0$.
  When $\limsup_{x\to\infty} \frac{\mean{P}_X(x)}{\mean{P}_Y(x)} > 1$, the (right) tail of $X$ is said to be \emph{heavier} than the (right) tail of $Y$.
\end{defn}

It is immediate from the definitions that the distribution of an unbiased estimator is light-tailed if and only if the estimator is sharply concentrated, so both notions are equivalent.

\begin{marginfigure}[13\baselineskip]
  \begin{center}
    \import{./plots/output/}{laplace_vs_normal_tails.pgf}
  \end{center}

	\caption{Shown are the right tails of the PDFs of a \textbf{\r{Laplace distribution}} with mean $0$ and length scale $1$ and a \textbf{\b{Gaussian}} with mean $0$ and variance $1$.}
\end{marginfigure}

\begin{ex}{Light- and heavy-tailed distributions}{}
  \begin{itemize}
    \item A Gaussian $X \sim \N{0}{1}$ is light-tailed since its tail distribution function is bounded by \begin{align}
      \mean{P}_X(x) = \int_x^\infty \frac{1}{\sqrt{2 \pi}} e^{- \nicefrac{t^2}{2}} \,d t \leq \int_x^\infty \frac{t}{x} \frac{1}{\sqrt{2 \pi}} e^{- \nicefrac{t^2}{2}} \,d t = \frac{e^{-\nicefrac{x^2}{2}}}{\sqrt{2\pi}x}. \margintag{using $\frac{t}{x} \geq 1$}
    \end{align}

    \item The \midx{Laplace distribution} with PDF \begin{align}
      \Laplace[x]{\mu}{h} \propto \exp\parentheses*{-\frac{\abs{x - \mu}}{h}}
    \end{align} is light-tailed, but its tails decay slower than those of the Gaussian.
    It is also ``more sharply peaked'' at its mean than the Gaussian.\looseness=-1
  \end{itemize}

  Heavy-tailed distributions frequently occur in many domains, the following just serving as a few examples.
  \begin{itemize}
    \item A well-known example of a heavy-tailed distribution is the \midx{log-normal distribution}.
    A random variable $X$ is logarithmically normal distributed with parameters $\mu$ and $\sigma^2$ if $\log X \sim \N{\mu}{\sigma^2}$.
    The log-normal arises when modeling natural growth phenomena or stock prices which are multiplicative, and hence, become additive on a logarithmic scale.

    \item The \midx{Pareto distribution} was originally used to model the distribution of wealth in a society, but it is also used to model many other phenomena such as the size of cities, the frequency of words, and the returns on stocks.
    Formally, the Pareto distribution is defined by the following PDF, \begin{align}
      \Pareto[x]{\alpha}{c} = \frac{\alpha c^\alpha}{x^{\alpha+1}} \Ind{x \geq c}, \quad x \in \R \label{eq:pareto_distr}
    \end{align} where the \midx{tail index} $\alpha > 0$ models the ``weight'' of the right tail of the distribution (larger $\alpha$ means lighter tail) and $c > 0$ corresponds to a cutoff threshold.
    The distribution is supported on $[c, \infty)$, and as $\alpha \to \infty$ it approaches a point density at $c$.
    The Pareto's right tail is $\mean{P}(x) = (\frac{c}{x})^{\alpha}$ for all $x \geq c$.

    \item The \midx{Cauchy distribution} arises in the description of numerous physical phenomena.
    The CDF of a Cauchy distribution is \begin{align}
      P_X(x) \defeq \frac{1}{\pi} \arctan\parentheses*{\frac{x - m}{\tau}} + \frac{1}{2}
    \end{align} where $X \sim \Cauchy{m}{\tau}$ with location $m$ and scale $\tau > 0$.
  \end{itemize}
\end{ex}

\begin{figure}
  \begin{center}
    \import{./plots/output/}{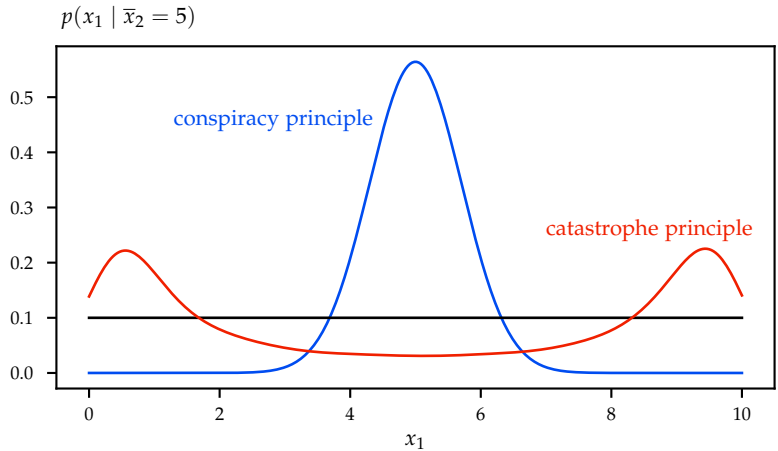}
  \end{center}

	\caption{Shown are the conditional distributions of $X_1$ given the event that the sample mean across two samples is surprisingly large: $\mean{X}_2 = 5$. We plot the examples where $X$ is a \textbf{\b{light-tailed}} Gaussian with mean $1$, $X$ is a \textbf{\r{heavy-tailed}} log-normal with mean $1$, and $X$ is \textbf{exponential} with mean $1$. When $X$ is \emph{light-tailed}, the large mean is explained by ``conspiratory'' samples $X_1$ and $X_2$. In contrast, when $X$ is \emph{heavy-tailed}, the large mean is explained by a single ``catastrophic'' event. See also \cite{nair2022fundamentals}.}
  \label{fig:conspiracy_vs_catastrophe}
\end{figure}

Light- and heavy-tailed distributions exhibit dramatically different behavior.
For example, consider the sample mean $\mean{X}_{n}$ of $n$ i.i.d. random variables $X_i$ with mean $1$, and we are told that the sample mean is $5$.
What does this tell us about the (conditional) distribution of $X_1$?
There are many possible explanations for this observation, but they can be largely grouped into two categories: \begin{enumerate}
  \item either \emph{many} $X_i$ are \emph{slightly} larger than $1$,
  \item or \emph{very few} $X_i$ are \emph{much} larger than $1$.
\end{enumerate}
\cite{nair2022fundamentals} term interpretation (1) the \midx{conspiracy principle} and interpretation (2) the \midx{catastrophe principle}.
It turns out that which of the two principles applies depends on the tails of the distribution of $X^{(1)}$.
If the tails are light, then the conspiracy principle applies, and if the tails are heavy, then the catastrophe principle applies as is illustrated in \cref{fig:conspiracy_vs_catastrophe}.

When working with light-tailed distributions, the conspiracy principle tells us that outliers are rare, and hence, we can usually ignore them.
In contrast, when working with heavy-tailed distributions, the catastrophe principle tells us that outliers are not just common but a defining feature, and hence, we need to be careful to not ignore them.

\begin{readings}
  For more details and examples of heavy-tailed distributions, refer to \icite{nair2022fundamentals}.
\end{readings}

\subsection{Mean Estimation and Concentration}\label{sec:fundamentals:mc_approx}%

We have seen that we desire two properties in estimators: namely, (1) that they are unbiased; and (2) that their variance is small.\footnote{That is, they are consistent and, ideally, their variance converges quickly.}
For estimating expectations with the sample mean $\mean{X}_n$ which is also known as a \midx{Monte Carlo approximation}[idxpagebf], we will see that both properties follow from standard results in statistics.

We have already concluded that the sample mean \eqref{eq:sample_mean} is an unbiased estimator for $\E*{X}$.
We will now see that the sample mean is also consistent, that its limiting distribution can usually be determined explicitly, and in some cases it is even sharply concentrated.
First, let us recall the common notions of convergence of sequences of random variables.

\begin{defn}[Convergence of random variables]\label{def:rv_convergence}
  Let $\{X_n\}_{n\in\Nat}$ be a sequence of random variables and $X$ another random variable. We say that, \begin{enumerate}
    \item $X_n$ converges to $X$ \midx<almost surely>{almost sure convergence} (also called \midx{convergence with probability $1$}) if \begin{align}
      \Pr{\braces*{\omega \in \Omega : \lim_{n\to\infty} X_n(\omega) = X(\omega)}} = 1, \label{eq:as_convergence}
    \end{align} and we write $X_n \almostsurely X$ as $n\to\infty$.

    \item $X_n$ converges to $X$ \midx<in probability>{convergence in probability} if for any $\epsilon > 0$, \begin{align}
      \lim_{n\to\infty} \Pr{\abs{X_n - X} > \epsilon} = 0, \label{eq:p_convergence}
    \end{align} and we write $X_n \convp X$ as $n\to\infty$.

    \item $X_n$ converges to $X$ \midx<in distribution>{convergence in distribution} if for all points $x \in X(\Omega)$ at which $P_X$ is continuous, \begin{align}
      \lim_{n\to\infty} P_{X_n}(x) = P_X(x), \label{eq:d_convergence}
    \end{align} and we write $X_n \convd X$ as $n\to\infty$.
  \end{enumerate}

  It can be shown that as $n\to\infty$, \begin{align}
    X_n \almostsurely X \implies X_n \convp X \implies X_n \convd X.
  \end{align}
\end{defn}

\begin{rmk}{Convergence in mean square}{mean_square_convergence}
  Mean square continuity and mean square differentiability are a generalization of continuity and differentiation to random processes, where the limit is generalized to a limit in the ``mean square sense''.

  A sequence of random variables $\{X_n\}_{n \in \Nat}$ is said to converge to the random variable $X$ in \midx<mean square>{mean square convergence} if \begin{align}
    \lim_{n\to\infty} \E{(X_n - X)^2} = 0.
  \end{align}
  Using Markov's inequality \eqref{eq:markov_inequality} it can be seen that convergence in mean square implies convergence in probability.
  Moreover, convergence in mean square implies \begin{align}
    \lim_{n\to\infty} \E*{X_n} &= \E*{X} \quad\text{and} \\
    \lim_{n\to\infty} \E*{X_n^2} &= \E*{X^2}.
  \end{align}

  Whereas a deterministic function $f(\vx)$ is said to be continuous at a point $\vxs$ if $\lim_{\vx\to\vxs} f(\vx) = f(\vxs)$, a random process $f(\vx)$ is \midx<mean square continuous>{mean square continuity} at $\vxs$ if \begin{align}
    \lim_{\vx\to\vxs} \E{(f(\vx) - f(\vxs))^2} = 0.
  \end{align}
  It can be shown that a random process is mean square continuous at $\vxs$ iff its kernel function $k(\vx, \vxp)$ is continuous at $\vx = \vxp = \vxs$.

  Similarly, a random process $f(\vx)$ is \midx<mean square differentiable>{mean square differentiability} at a point $\vx$ in direction $i$ if $(f(\vx + \delta \ve_i) - f(\vx)) / \delta$ converges in mean square as $\delta \to 0$ where $\ve_i$ is the unit vector in direction $i$.
  This notion can be extended to higher-order derivatives.

  For the precise notions of mean square continuity and mean square differentiability in the context of Gaussian processes, refer to section 4.1.1 of \icite{gpml}.
\end{rmk}

Given a random variable $X : \Omega \to \R$ with finite variance, it can be shown that \begin{align}
  \mean{X}_n \convp \E*{X} \label{eq:wlln}
\end{align} which is known as the \midx{weak law of large numbers} (WLLN) and which establishes consistency of the sample mean \exerciserefmark{concentration_inequalities}.
Using more advanced tools, it is possible to show almost sure convergence of the sample mean even when the variance is infinite:

\begin{thmb}
  \begin{fct}[Strong law of large numbers, SLLN]\pidx{law of large numbers}
    Given the random variable $X : \Omega \to \R$ with finite mean.
    Then, as $n\to\infty$, \begin{align}
      \mean{X}_n \almostsurely \E*{X}. \label{eq:slln}
    \end{align}
  \end{fct}
\end{thmb}

To get an idea of how quickly the sample mean converges, we can look at its variance: \begin{align}
  \Var{\mean{X}_n} = \Var{\frac{1}{n} \sum_{i=1}^n X_i} = \frac{\Var{X}}{n}.
\end{align}
Remarkably, one cannot only compute its variance, but also its limiting distribution.
This is known as the \midx{central limit theorem} (CLT) which states that the prediction error of the sample mean tends to a normal distribution as the sample size goes to infinity (even if the samples themselves are not normally distributed).

\begin{thmb}
  \begin{fct}[Central limit theorem by Lindeberg-Lévy]
    Given the random variable $X : \Omega \to \R$ with finite mean and finite variance.
    Then, \begin{align}
      \mean{X}_n \convd \N*{\E*{X}}{\frac{\Var{X}}{n}}
    \end{align} as $n\to\infty$.
  \end{fct}
\end{thmb}

The central limit theorem makes the critical assumption that the variance of $X$ is finite.
This is not the case for many heavy-tailed distributions such as the Pareto or Cauchy distributions.
One can generalize the central limit theorem to distributions with infinite variance, with the important distinction that their limiting distribution is no longer a Gaussian but also a heavy-tailed distribution with infinite variance~\citep{nair2022fundamentals}.

For a subclass of light-tailed distributions it is also possible to show that the sample mean $\mean{X}_n$ is sharply concentrated, which is a much stronger and much more practical property than consistency.
We will consider the class of sub-Gaussian random variables which encompass random variables whose tail probabilities decay at least as fast as those of a Gaussian.

\begin{defn}[Sub-Gaussian random variable]\pidx{sub-Gaussian random variable}
  A random variable $X : \Omega \to \R$ is called $\sigma$-\emph{sub-Gaussian} for $\sigma > 0$ if for all $\lambda \in \R$, \begin{align}
    \E{e^{\lambda (X - \E*{X})}} \leq \exp\parentheses*{\frac{\sigma^2 \lambda^2}{2}}. \label{eq:sub_gaussian}
  \end{align}
\end{defn}

\begin{ex}{Examples of sub-Gaussian random variables}{sub_gaussian_examples}
  \leavevmode\begin{itemize}
    \item If a random variable $X$ is $\sigma$-sub-Gaussian, then so is $-X$.

    \item If $X \sim \N{\mu}{\sigma^2}$ then a simple calculation shows that \begin{align}
      \varphi_X(\lambda) \defeq \E{e^{\lambda X}} = \exp\parentheses*{\mu \lambda + \frac{\sigma^2 \lambda^2}{2}} \qquad\text{for any $\lambda \in \R$} \label{eq:mgf_univ_gaussian}
    \end{align} which is called the ``moment-generating function'' of the normal distribution, and implying that $X$ is $\sigma$-sub-Gaussian.

    \item If the random variable $X : \Omega \to \R$ satisfies $a \leq X \leq b$ with probability $1$ then $X$ is $\frac{b-a}{2}$-sub-Gaussian.
    This is also known as \midx{Hoeffding's lemma}.
  \end{itemize}
\end{ex}

\begin{thmb}
  \begin{thm}[Hoeffding's inequality]\pidx{Hoeffding's inequality}\label{thm:hoeffdings_inequality}
    Let $X : \Omega \to \R$ be a $\sigma$-sub-Gaussian random variable.
    Then, for any $\epsilon > 0$, \begin{align}
      \Pr{\abs{\mean{X}_n - \E*{X}} \geq \epsilon} \leq 2 \exp\parentheses*{-\frac{n \epsilon^2}{2 \sigma^2}}. \label{eq:hoeffdings_inequality}
    \end{align}
  \end{thm}
\end{thmb}

In words, the absolute error of the sample mean is bounded by an exponentially quickly decaying error probability $\delta$.
Solving for $n$, we obtain that for \begin{align}
  n \geq \frac{2 \sigma^2}{\epsilon^2} \log \frac{2}{\delta} \label{eq:eq:hoeffdings_inequality_sample_size}
\end{align} the probability that the absolute error is greater than $\epsilon$ is at most $\delta$.

\begin{proof}[Proof of \cref{thm:hoeffdings_inequality}]
  Let $S_n \defeq n \mean{X}_n = X_1 + \dots + X_n$.
  We have for any $\lambda, \epsilon > 0$ that \begin{align*}
    \Pr{\mean{X}_n - \E*{X} \geq \epsilon} &= \Pr{S_n - \E*{S_n} \geq n \epsilon} \\
    &= \Pr{e^{\lambda(S_n - \E*{S_n})} \geq e^{n \epsilon \lambda}} \margintag{using that $z \mapsto e^{\lambda z}$ is increasing} \\
    &\leq e^{-n\epsilon\lambda} \E*{[e^{\lambda(S_n - \E*{S_n})}]} \margintag{using Markov's inequality \eqref{eq:markov_inequality}} \\
    &= e^{-n\epsilon\lambda} \prod_{i=1}^n \E*{[e^{\lambda(X_i - \E*{X})}]} \margintag{using independence of the $X_i$} \\
    &\leq e^{-n\epsilon\lambda} \prod_{i=1}^n e^{\sigma^2 \lambda^2 / 2} \margintag{using the characterizing property of a $\sigma$-sub-Gaussian random variable \eqref{eq:sub_gaussian}} \\
    &= \exp\parentheses*{-n\epsilon\lambda + \frac{n \sigma^2 \lambda^2}{2}}.
  \end{align*}
  Minimizing the expression with respect to $\lambda$, we set $\lambda = \epsilon /\sigma^2$, and obtain \begin{align*}
    \Pr{\mean{X}_n - \E*{X} \geq \epsilon} \leq \min_{\lambda > 0}\braces*{\exp\parentheses*{-n\epsilon\lambda + \frac{n \sigma^2 \lambda^2}{2}}} = \exp\parentheses*{-\frac{n \epsilon^2}{2 \sigma^2}}.
  \end{align*}
  The theorem then follows from \begin{align*}
    \Pr{\abs{\mean{X}_n - \E*{X}} \geq \epsilon} = \Pr{\mean{X}_n - \E*{X} \geq \epsilon} + \Pr{\mean{X}_n - \E*{X} \leq -\epsilon}
  \end{align*} and noting that the second term can be bounded analogously to the first term by considering the random variables $-X_1, \dots, -X_n$.
\end{proof}

The law of large numbers and Hoeffding's inequality tell us that when~$X$ is light-tailed, we can estimate $\E*{X}$ very precisely with ``few'' samples using a sample mean.
Crucially, the sample mean requires \emph{independent} samples $x_i$ from $X$ which are often hard to obtain.

\begin{rmk}{When the sample mean fails}{}
  While we have seen that the sample mean works well for light-tailed distributions, it fails for heavy-tailed distributions.
  For example, if we only assume that the variance of $X$ is finite, then the best known error rate is obtained by applying Chebyshev's inequality \eqref{eq:chebyshev_inequality}, \begin{align}
    \Pr{\abs{\mean{X}_n - \E*{X}} \geq \epsilon} \leq \frac{\Var{X}}{n \epsilon^2}
  \end{align} which decays only linearly in $n$.
  This is a result of the catastrophe principle, namely, that outliers may be likely and therefore need to be accounted for.

  Robust methods of estimation in the presence of outliers or corruptions have been studied extensively in \midx{robust statistics}.
  On the subject of mean estimation, approaches such as trimming outliers before taking the sample mean (called the \midx{truncated sample mean}) or using the \midx{median-of-means} which selects the median among $k$ sample means --- each computed on a subset of the data --- have been shown to yield sharply concentrated estimates, even when the variance of $X$ is infinite \citep{bubeck2013bandits}.
\end{rmk}

\subsection{Asymptotic Efficiency of Maximum Likelihood Estimation}\label{sec:fundamentals:parameter_estimation:asymptotic_efficiency}

In \cref{sec:fundamentals:parameter_estimation:mle}, we briefly discussed the asymptotic behavior of the MLE, and we mentioned that the MLE can be shown to be ``asymptotically efficient''.

Let us first address the question of how the asymptotic covariance matrix $\mS_n$ looks like?
It can be shown that $\mS_n = \inv{\mI_n(\vtheta)}$ where \begin{align}
  \mI_n(\vtheta) \defeq \E[\spD_n]{\hes_\vtheta \ell_\mathrm{nll}(\vtheta; \spD_n)}
\end{align} is the so-called \midx{Fisher information} which captures the curvature of the negative log-likelihood around $\vtheta$.
The Fisher information $\mI_n(\vtheta)$ can be used to measure the ``difficulty'' of estimating $\vtheta$ as shown by the \midx{Cramér-Rao lower bound}:

\begin{fct}[Cramér-Rao lower bound]
  Let $\vthetahat_n$ be an unbiased estimator of~$\vtheta$.
  Then,\safefootnote{We use $\mA \succeq \mB$ as shorthand for $\mA - \mB$ being positive semi-definite. This partial ordering of positive semi-definite matrices is called the \midx{Loewner order}.} \begin{align}
    \Var{\vthetahat_n} \succeq \inv{\mI_n(\vtheta)}.
  \end{align}
\end{fct}

An estimator is called \midx{efficient} if it achieves equality in the Cramér-Rao lower bound, and the MLE is therefore \emph{asymptotically efficient}.

\subsection{Population Risk and Empirical Risk}\label{sec:fundamentals:supervised_learning:risk}

The notion of ``error'' mentioned in \cref{sec:fundamentals:supervised_learning} is typically captured by a loss function $\ell(\hat{y}; y) \in \R$ which is small when the prediction $\hat{y} = \hat{f}(\vx)$ is ``close'' to the true label $y = \opt{f}(\vx)$ and large otherwise.
Let us fix a distribution $\spP_\spX$ over inputs, which together with the likelihood from \cref{eq:data} induces an unknown joint distribution over input-label pairs $(\vx, y)$ by $\spP$.
The canonical objective is to best-approximate the mappings $(\vx,y) \sim \spP$, that is, to minimize \begin{align}
  \E[(\vx, y) \sim \spP]{\ell(\hat{f}(\vx); y)}.
\end{align}
This quantity is also called the \midx{population risk}.
However, the underlying distribution $\spP$ is unknown to us.
All that we can work with is the training data for which we assume $\spD_n \iid \spP$.
It is therefore natural to consider minimizing \begin{align}
  \frac{1}{n} \sum_{i=1}^n \ell(\hat{f}(\vx_i); y_i), \quad \spD_n = \{(\vx_i, y_i)\}_{i=1}^n,
\end{align} which is known as the \midx{empirical risk}.

However, selecting $\hat{f}$ by minimizing the empirical risk can be problematic.
The reason is that in this case the model $\hat{f}$ and the empirical risk depend on the same data $\spD_n$, implying that the empirical risk will not be an unbiased estimator of the population risk.
This can result in a model which fits the training data too closely, and which is failing to generalize to unseen data --- a problem called \midx{overfitting}[idxpagebf].
We will discuss some (probabilistic) solutions to this problem in \cref{sec:gp:model_selection} when covering model selection.

\section{Optimization}\label{sec:fundamentals:optimization}

Finding parameter estimates is one of the many examples where we seek to minimize some function $\ell$.\footnote{W.l.o.g. we assume that we want to minimize $\ell$. If we wanted to maximize the objective, we can simply minimize its negation.}
The field of optimization has a rich history, which we will not explore in much detail here.
What will be important for us is that given that the function to be optimized (called the \midx<objective>{objective function} or \midx<loss>{loss function}) fulfills certain smoothness properties, optimization is a well-understood problem and can often be solved exactly (e.g., when the objective is convex) or ``approximately'' when the objective is non-convex.
In fact, we will see that it is often advantageous to frame problems as optimization problems when suitable because the machinery to solve these problems is so extensive.

\begin{readings}
  For a more thorough reminder of optimization methods, read chapter 7 of \icite{mml}.
\end{readings}

\subsection{Stationary Points}

In this section, we derive some basic facts about unconstrained optimization problems.
Given some function $f : \R^n \to \R$, we want to find\looseness=-1 \begin{align}
  \min_{\vx \in \R^n} f(\vx).
\end{align}
We say that a point $\vxs \in \R^n$ is a \midx<(global) optimum>{global optimum} of $f$ if $f(\vxs) \leq f(\vx)$ for any $\vx \in \R^n$.

Consider the more general problem of minimizing $f$ over some subset $\sS \subseteq \R^n$, that is, to minimize the function $f_\sS : \sS \to \R, \vx \mapsto f(\vx)$.
If there exists some open subset $\sS \subseteq \R^n$ including $\vxs$ such that $\vxs$ is optimal with respect to the function $f_\sS$, then $\vxs$ is called a \midx<local optimum>{local optimum} of $f$.

\begin{rmk}{Differentiability}{}
  We will generally assume that $f$ is continuously (Fréchet) differentiable on $\R^n$.
  That is, at any point $\vx \in \R^n$, there exists $\grad f(\vx)$ such that for any $\vdelta \in \R^n$, \begin{align}
    f(\vx + \vdelta) = f(\vx) + \transpose{\grad f(\vx)}\vdelta + o(\norm{\vdelta}_2), \label{eq:first_order_exp}
  \end{align} where $\lim_{\vdelta \to \vzero} \frac{o(\norm{\vdelta}_2)}{\norm{\vdelta}_2} = 0$, and $\grad f(\vx)$ is continuous on $\R^n$.

  \Cref{eq:first_order_exp} is also called a \midx{first-order expansion} of $f$ at $\vx$.
\end{rmk}

\begin{defn}[Stationary point]
  Given a function $f : \R^n \to \R$, a point $\vx \in \R^n$ where $\grad f(\vx) = \vzero$ is called a \midx{stationary point} of $f$.
\end{defn}

\begin{marginfigure}
  \begin{center}
    \import{./plots/output/}{saddle_point.pgf}
  \end{center}

  \caption{Example of a saddle point at $x = 0$.}
\end{marginfigure}

Being a stationary point is not sufficient for optimality.
Take for example the point $x \defeq 0$ of $f(x) \defeq x^3$.
Such a point that is stationary but not (locally) optimal is called a \midx{saddle point}.

\begin{thm}[First-order optimality condition]\pidx{first-order optimality condition}
  If $\vx \in \R^n$ is a local extremum of a differentiable function $f : \R^n \to \R$, then $\grad f(\vx) = \vzero$.
\end{thm}
\begin{proof}
  Assume $\vx$ is a local minimum of $f$.
  Then, for all $\vd \in \R^n$ and for all small enough $\lambda \in \R$, we have $f(\vx) \leq f(\vx + \lambda\vd)$, so \begin{align*}
    0 &\leq f(\vx + \lambda\vd) - f(\vx) \\
    &= \lambda \transpose{\grad f(\vx)}\vd + o(\lambda\norm{\vd}_2). \margintag{using a first-order expansion of $f$ around $\vx$}
  \end{align*}
  Dividing by $\lambda$ and taking the limit $\lambda \to 0$, we obtain \begin{align*}
    0 \leq \transpose{\grad f(\vx)}\vd + \lim_{\lambda \to 0}\frac{o(\lambda\norm{\vd}_2)}{\lambda} = \transpose{\grad f(\vx)}\vd.
  \end{align*}
  Take $\vd \defeq - \grad f(\vx)$.
  Then, $0 \leq - \norm{\grad f(\vx)}_2^2$, so $\grad f(\vx) = \vzero$.
\end{proof}

\subsection{Convexity}

Convex functions are a subclass of functions where finding global minima is substantially easier than for general functions.

\begin{marginfigure}
  \begin{center}
    \import{./plots/output/}{convexity_0.pgf}
  \end{center}

  \caption{Example of a convex function.
  Any line between two points on $f$ lies ``above'' $f$.}
\end{marginfigure}

\begin{defn}[Convex function]
  A function $f : \R^n \to \R$ is \midx<convex>{convex function} if \begin{align}
    \forall \vx, \vy \in \R^n : \forall \theta \in [0,1] : f(\theta\vx + (1-\theta)\vy) \leq \theta f(\vx) + (1-\theta) f(\vy). \label{eq:convexity}
  \end{align}
  That is, any line between two points on $f$ lies ``above'' $f$.
  If the inequality of \cref{eq:convexity} is strict, we say that $f$ is \midx<strictly convex>{strict convexity}.
\end{defn}

If the function $f$ is convex, we say that the function $-f$ is \midx<concave>{concave function}.
The above is also known as the \emph{$0$th-order characterization of convexity}.

\begin{thm}[First-order characterization of convexity]\label{thm:fo-characterization-convexity}\pidx{first-order characterization of convexity}
  Suppose that $f : \R^n \to \R$ is differentiable, then $f$ is convex if and only if \begin{align}
    \forall \vx, \vy \in \R^n : f(\vy) \geq f(\vx) + \transpose{\grad f(\vx)}(\vy - \vx).
  \end{align}
\end{thm}

\begin{marginfigure}
  \begin{center}
    \import{./plots/output/}{convexity_1.pgf}
  \end{center}

  \caption{The first-order characterization characterizes convexity in terms of affine lower bounds.
  Shown is an affine lower bound based at $x = -2$.}
\end{marginfigure}

Observe that the right-hand side of the inequality is an affine function with slope $\grad f(\vx)$ based at $f(\vx)$.

\begin{proof}
  In the following, we will make use of directional derivatives.
  In particular, we will use that given a function $f : \R^n \to \R$ that is differentiable at $\vx \in \R^n$, \begin{align}
    \lim_{\lambda \to 0} \frac{f(\vx + \lambda\vd) - f(\vx)}{\lambda} = \transpose{\grad f(\vx)} \vd
  \end{align} for any direction $\vd \in \R^n$ \exerciserefmark{directional_derivatives}.

  \vspace{5pt}\begin{itemize}
    \item ``$\Rightarrow$'':
    Fix any $\vx, \vy \in \R^n$.
    As $f$ is convex, \begin{align*}
      f((1-\theta)\vx + \theta\vy) \leq (1-\theta)f(\vx) + \theta f(\vy),
    \end{align*} for all $\theta \in [0,1]$.
    We can rearrange to \begin{align*}
      f(\underbrace{(1-\theta)\vx + \theta\vy}_{\vx + \theta(\vy - \vx)}) - f(\vx) \leq \theta(f(\vy) - f(\vx)).
    \end{align*}
    Dividing by $\theta$ yields, \begin{align*}
      \frac{f(\vx + \theta(\vy - \vx)) - f(\vx)}{\theta} \leq f(\vy) - f(\vx).
    \end{align*}
    Taking the limit $\theta \to 0$ on both sides gives the directional derivative at $\vx$ in direction $\vy - \vx$, \begin{align*}
      \transpose{\grad f(\vx)}(\vy - \vx) = D f(\vx)[\vy - \vx] \leq f(\vy) - f(\vx).
    \end{align*}

    \item ``$\Leftarrow$'':
    Fix any $\vx, \vy \in \R^n$ and let $\vz \defeq \theta\vy + (1-\theta)\vx$.
    We have, \begin{align*}
      f(\vy) &\geq f(\vz) + \transpose{\grad f(\vz)}(\vy - \vz), \quad\text{and} \\
      f(\vx) &\geq f(\vz) + \transpose{\grad f(\vz)}(\vx - \vz).
    \end{align*}
    We also have $\vy - \vz = (1-\theta)(\vy-\vx)$ and $\vx - \vz = \theta(\vx - \vy)$.
    Hence, \begin{align*}
      \theta f(\vy) + (1-\theta) f(\vx) &\geq f(\vz) + \transpose{\grad f(\vz)}(\underbrace{\theta(\vy - \vz) + (1-\theta)(\vx - \vz)}_{\vzero}) \\
      &= f(\theta\vy + (1-\theta)\vx). \qedhere
    \end{align*}
  \end{itemize}
\end{proof}

\begin{thmb}
  \begin{thm}
    Let $f : \R^n \to R$ be a convex and differentiable function.
    Then, if $\vxs \in \R^n$ is a stationary point of $f$, $\vxs$ is a global minimum of $f$.
  \end{thm}
\end{thmb}
\begin{proof}
  By the first-order characterization of convexity, we have for any $\vy \in \R^n$, \begin{align*}
    f(\vy) \geq f(\vxs) + \underbrace{\transpose{\grad f(\vxs)}}_{\vzero}(\vy - \vxs) = f(\vxs). &\qedhere
  \end{align*}
\end{proof}

Generally, the main difficulty in solving convex optimization problems lies therefore in finding stationary points (or points that are sufficiently close to stationary points).

\begin{rmk}{Second-order characterization of convexity}{second_order_convexity}
  We say that $f : \R^n \to \R$ is twice continuously (Fréchet) differentiable if for any point $\vx \in \R^n$, there exists $\grad f(\vx)$ and $\hes_f(\vx)$ such that for any $\vdelta \in \R^n$, \begin{align}
    f(\vx + \vdelta) = f(\vx) + \transpose{\grad f(\vx)}\vdelta + \frac{1}{2}\transpose{\vdelta}\hes_f(\vx)\vdelta + o(\norm{\vdelta}^2_2), \label{eq:second_order_exp}
  \end{align} where $\lim_{\vdelta \to \vzero} \frac{o(\norm{\vdelta}^2_2)}{\norm{\vdelta}^2_2} = 0$, and $\grad f(\vx)$ and $\hes_f(\vx)$ are continuous on $\R^n$.
  \Cref{eq:second_order_exp} is also called a \midx{second-order expansion} of $f$ at $\vx$.

  Twice continuously differentiable functions admit a natural characterization of convexity in terms of the Hessian.

  \begin{fct}[Second-order characterization of convexity]\pidx{second-order characterization of convexity}
    Consider a twice continuously differentiable function $f : \R^n \to R$. Then, $f$ is convex if and only if $\hes_f(\vx)$ is positive semi-definite for all $\vx \in \R^n$.
  \end{fct}

  It follows that $f$ is concave if and only if $\hes_f(\vx)$ is negative semi-definite for all $\vx \in \R^n$.
\end{rmk}

\begin{marginbox}[-30\baselineskip]{Jacobian}
  Given a vector-valued function, \begin{align*}
    \vg : \R^n \to \R^m,\quad \vx \mapsto \begin{bmatrix}
      g_1(\vx) \\
      \vdots \\
      g_m(\vx) \\
   \end{bmatrix},
  \end{align*} where $g_i : \R^n \to \R$, the \idx{Jacobian} of $\vg$ at $\vx \in \R^n$ is \begin{align}
    \jac{\vg(\vx)} \defeq \begin{bmatrix}
      \pdv{g_1(\vx)}{\vx(1)} & \cdots & \pdv{g_1(\vx)}{\vx(n)} \\
      \vdots & \ddots & \vdots \\
      \pdv{g_m(\vx)}{\vx(1)} & \cdots & \pdv{g_m(\vx)}{\vx(n)} \\
    \end{bmatrix}.
  \end{align}
  Observe that for a function $f : \R^n \to \R$, \begin{align}
    \jac f(\vx) = \transpose{\grad f(\vx)}.
  \end{align}
\end{marginbox}

\begin{marginbox}[-5\baselineskip]{Hessian}
  The \idx{Hessian} of a function $f : \R^n \to \R$ at a point $\vx \in \R^n$ is \begin{align}
    \hes_f(\vx) &\defeq \hes f(\vx) \nonumber \\
    &\defeq \begin{bmatrix}
      \pdv{f(\vx)}{\vx(1),\vx(1)} & \cdots & \pdv{f(\vx)}{\vx(1),\vx(n)} \\
      \vdots & \ddots & \vdots \\
      \pdv{f(\vx)}{\vx(n),\vx(1)} & \cdots & \pdv{f(\vx)}{\vx(n),\vx(n)} \\
   \end{bmatrix} \\
   &= \transpose{(\jac\grad f(\vx))} \label{eq:hessian} \\
   &\in \R^{n \times n}. \nonumber
  \end{align}
  Thus, a Hessian captures the curvature of $f$.
  If the Hessian of $f$ is positive definite at a point $\vx$, then $f$ is ``curved up around $\vx$''.

  Hessians are symmetric when the second partial derivatives are continuous, due to Schwartz's theorem.
\end{marginbox}

\subsection{Stochastic Gradient Descent}

In this \course, we primarily employ so-called first-order methods, which rely on (estimates of) the gradient of the objective function to determine a direction of local improvement.
The main idea behind these methods is to repeatedly take a step in the opposite direction of the gradient scaled by a learning rate $\eta_t$, which may depend on the current iteration $t$.

We will often want to minimize a stochastic optimization objective \begin{align}
  L(\vtheta) \defeq \E[\vx \sim p]{\ell(\vtheta; \vx)}
\end{align} where $\ell$ and its gradient $\grad \ell$ are known.

Based on our discussion in previous subsections, it is a natural first step to look for stationary points of $L$, that is, the roots of $\grad L$.

\begin{thmb}
  \begin{fct}[Robbins-Monro (RM) algorithm]
    Let $\vg : \R^n \to \R^m$ be an unknown function of which we want to find a root and suppose that we have access to unbiased noisy observations $\rG(\vtheta)$ of $\vg(\vtheta)$.
    The scheme \begin{align}
      \vtheta_{t+1} \defeq \vtheta_t - \eta_t \vg^{(t)}(\vtheta_t),
    \end{align} where $\vg^{(t)}(\vtheta_t) \sim \rG(\vtheta_t)$ are independent and unbiased estimates of $\vg(\vtheta_t)$, is known as the \midx{Robbins-Monro algorithm}.

    It can be shown that if the sequence of learning rates $\{\eta_t\}_{t=0}^\infty$ is chosen such that the \midx{Robbins-Monro conditions},\safefootnote{$\eta_t = \nicefrac{1}{t}$ is an example of a learning rate satisfying the RM-conditions.} \begin{align}
      \eta_t \geq 0 \quad \forall t, \quad \sum_{t=0}^\infty \eta_t = \infty \quad\text{and}\quad \sum_{t=0}^\infty \eta_t^2 < \infty, \label{eq:rm_conditions}
    \end{align} and additional regularity assumptions are satisfied,\safefootnote{For more details, refer to \icite{bottou1998online}.} then we have that \begin{align}
      \vg(\vtheta_t) \almostsurely \vzero \quad\text{as $t \to \infty$}.
    \end{align}
    That is, the RM-algorithm converges to a root almost surely.\safefootnote{See \cref{eq:as_convergence} for a definition of almost sure convergence of a sequence of random variables.}
  \end{fct}
\end{thmb}

Using Robbins-Monro to find a root of $\grad L$ is known as \emph{stochastic gradient descent}.
In particular, when $\ell$ is convex and the RM-conditions are satisfied, Robbins-Monro converges to a stationary point (and hence, a global minimum) of $L$ almost surely.
Moreover, it can be shown that for general $\ell$ and satisfied RM-conditions, stochastic gradient descent converges to a local minimum almost surely.
Intuitively, the randomness in the gradient estimates allows the algorithm to ``jump past'' saddle points.

A commonly used strategy to obtain unbiased gradient estimates is to take the sample mean of the gradient with respect to some set of samples $\sB$ (also called a \midx{batch}) as is shown in \cref{alg:sgd}.

\begin{algorithm}
  \caption{Stochastic gradient descent, SGD}\pidx{stochastic gradient descent}\label{alg:sgd}
  initialize $\vtheta$\;
  \Repeat{converged}{
    draw mini-batch $\sB \defeq \{\vx^{(1)}, \dots, \vx^{(m)}\}, \vx^{(i)} \sim p$\;
    $\vtheta \gets \vtheta - \eta_t \frac{1}{m} \sum_{i=1}^m \grad_{\vtheta} \ell(\vtheta; \vx^{(i)})$\;
  }
\end{algorithm}

\begin{ex}{Minimizing training loss}{}
  A common application of SGD in the context of machine learning is the following: $p \defeq \Unif{\{\vx_1, \dots, \vx_n\}}$ is a uniform distribution over the training inputs, yielding the objective \begin{align}
    L(\vtheta) = \frac{1}{n} \sum_{i=1}^n \ell(\vtheta; \vx_i),
  \end{align} where $\ell(\vtheta; \vx_i)$ is the loss of a model parameterized by $\vtheta$ for training input $\vx_i$ and $m$ is a fixed \midx{batch size}.
  Here, computing the gradients exactly (i.e., choosing $m = n$) is expensive when the size of the training set is large.

  A commonly used alternative to sampling each batch independently, is to split the training set into equally sized batches and perform a gradient descent step with respect to each of them.
  Typically, the optimum is not found after a single pass through the training data, so the same procedure is repeated multiple times. One such iteration is called an \midx{epoch}.
\end{ex}

\begin{rmk}{Regularization via weight decay}{weight_decay}
  A common technique to improve the ``stability'' of minima found through gradient-based methods is to ``regularize'' the loss by adding an explicit bias favoring ``simpler'' solutions.
  That is, given a loss $\ell$ measuring the quality of fit, we consider the regularized loss\looseness=-1 \begin{align}
    \ell'(\vtheta; \vx) \defeq \underbrace{\ell(\vtheta; \vx)}_{\text{quality of fit}} + \underbrace{r(\vtheta)}_{\text{regularization}}
  \end{align} where $r(\vtheta)$ is large for ``complex'' and small for ``simple'' choices of $\vtheta$, respectively.
  A common choice is $r(\vtheta) = \frac{\lambda}{2} \norm{\vtheta}_2^2$ for some $\lambda > 0$, which is known as \midx{$L_2$-regularization}.

  Recall from \cref{sec:fundamentals:parameter_esitmation:map} that in the context of likelihood maximization, this choice of $r$ corresponds to imposing the Gaussian prior $\N{\vzero}{\lambda\mI}$, and finding the MAP estimate.
  Imposing, for example, a Laplace prior leads to \midx{$L_1$-regularization}.

  Using $L_2$-regularization, we obtain for the gradient, \begin{align}
    \grad_\vtheta \ell'(\vtheta; \vx) = \grad_\vtheta \ell(\vtheta; \vx) + \lambda \vtheta.
  \end{align}
  Thus, a gradient descent step changes to \begin{align}
    \vtheta \gets (1 - \lambda \eta_t) \vtheta - \eta_t \grad_\vtheta \ell(\vtheta; \vx).
  \end{align}
  That is, in each step, $\vtheta$ decays towards zero at the rate $\lambda \eta_t$.
  This regularization method is also called \midx{weight decay}.
\end{rmk}

\begin{rmk}{Adaptive learning rates and momentum}{}
  Commonly, in optimization, constant learning rates are used to accelerate mixing.
  The performance can be further improved by taking an adaptive gradient step with respect to the geometry of the cost function, which is done by commonly used algorithms such as \midx{Adagrad} and \midx{Adam}.
  The methods employed by these algorithms are known as \midx<adaptive learning rates>{adaptive learning rate} and \midx{momentum}.

  For more details on momentum read section 7.1.2 of \icite{mml} and for an overview of the aforementioned optimization algorithms refer to \icite{ruder2016overview}.
\end{rmk}

\section{Useful Matrix Identities and Inequalities}\label{sec:background:identities_and_inequalities}

\begin{itemize}
  \item \midx<Woodbury's matrix identity>{Woodbury matrix identity} states that for any matrices $\mA \in \R^{n \times n}$, $\mC \in \R^{m \times m}$, and $\mU, \transpose{\mV} \in \R^{n \times m}$ where $\mA$ and $\mC$ are invertible, \begin{align}
    \inv{(\mA + \mU\mC\mV)} = \inv{\mA} - \inv{\mA}\mU\inv{(\inv{\mC} + \mV\inv{\mA}\mU)}\mV\inv{\mA}. \label{eq:general_woodbury}
  \end{align}
  The following identity, called the \midx{Sherman-Morrison formula}, is a direct consequence: \begin{subequations}\begin{align}
    \inv{(\mA + \vx\transpose{\vx})} &= \inv{\mA} - \inv{\mA}\vx\inv{(1 + \transpose{\vx}\inv{\mA}\vx)}\transpose{\vx}\inv{\mA} \\
    &= \inv{\mA} - \frac{(\inv{\mA}\vx)\transpose{(\inv{\mA}\vx)}}{1 + \transpose{\vx}\inv{\mA}\vx}
  \end{align}\label{eq:woodbury}\end{subequations} for any symmetric and invertible matrix $\mA \in \R^{n \times n}$ and $\vx \in \R^n$.

  \item The \midx{matrix inversion lemma} states that for matrices $\mA, \mB \in \R^{n \times n}$, \begin{align}
    \inv{(\mA + \mB)} = \inv{\mA} - \inv{\mA}\inv{(\inv{\mA} + \inv{\mB})} \inv{\mA}. \label{eq:matrix_inversion_lemma}
  \end{align}

  \item \midx{Hadamard's inequality} states that \begin{align}
    \det{\mM} \leq \prod_{i=0}^d \mM(i,i) \label{eq:hadamard}
  \end{align} for any positive definite matrix $\mM \in \R^{d \times d}$.

  \item The \midx{Weinstein-Aronszajn identity} for positive definite matrices $\mA \in \R^{d \times n}$ and $\mB \in \R^{n \times d}$ states that \begin{align}
    \det{\mI_{d \times d} + \mA\mB} = \det{\mI_{n \times n} + \mB\mA}. \label{eq:weinstein_aronszajn}
  \end{align}
\end{itemize}

\excheading

\begin{nexercise}{The sample variance is unbiased}{sample_variance}
  Let $X$ be a zero-mean random variable.
  Confirm that the sample variance of $X$ is indeed unbiased.
\end{nexercise}

\begin{nexercise}{Consistency of the sample mean}{concentration_inequalities}
  We will now show that the sample mean is consistent.
  To this end, let us first derive two classical concentration inequalities. \begin{enumerate}
    \item Prove \midx{Markov's inequality} which says that if $X$ is a non-negative random variable, then for any $\epsilon > 0$, \begin{align}
      \Pr{X \geq \epsilon} \leq \frac{\E*{X}}{\epsilon}. \label{eq:markov_inequality}
    \end{align}

    \item Prove \midx{Chebyshev's inequality} which says that if $X$ is a random variable with finite and non-zero variance, then for any $\epsilon > 0$, \begin{align}
      \Pr{\abs{X - \E*{X}} \geq \epsilon} \leq \frac{\Var*{X}}{\epsilon^2}. \label{eq:chebyshev_inequality}
    \end{align}

    \item Prove the weak law of large numbers from \cref{eq:wlln}.
  \end{enumerate}
\end{nexercise}

\begin{nexercise}{Directional derivatives}{directional_derivatives}
  Given a function $f : \R^n \to \R$ that is differentiable at $\vx \in \R^n$, the \midx{directional derivative} of $f$ at $\vx$ in the direction $\vd \in \R^n$ is \begin{align}
    D f(\vx)[\vd] \defeq \lim_{\lambda \to 0} \frac{f(\vx + \lambda\vd) - f(\vx)}{\lambda}.
  \end{align}

  Show that \begin{align}
    D f(\vx)[\vd] = \transpose{\grad f(\vx)} \vd.
  \end{align}
  \textit{Hint: Consider a first-order expansion of $f$ at $\vx$ in direction $\lambda\vd$.}
\end{nexercise}

  \chapter{Solutions}

\section*{\nameref{sec:fundamentals}}

\begin{solution}{properties_of_probability}
  \begin{enumerate}[beginpenalty=10000]
    \item By the third axiom and $B = A \cup (B \setminus A)$, \begin{align*}
      \Pr{B} = \Pr{A} + \Pr{B \setminus A}.
    \end{align*}
    Noting from the first axiom that $\Pr{B \setminus A} \geq 0$ completes the proof.

    \item By the second axiom, \begin{align*}
      \Pr{A \cup \compl{A}} = \Pr{\Omega} = 1
    \end{align*} and by the third axiom, \begin{align*}
      \Pr{A \cup \compl{A}} = \Pr{A} + \Pr{\compl{A}}.
    \end{align*}
    Reorganizing the equations completes the proof.

    \item Define the countable sequence of events \begin{align*}
      A'_i \defeq A_i \setminus \parentheses*{\bigcup_{j=1}^{i-1} A_j}.
    \end{align*}
    Note that the sequence of events $\{A'_i\}_i$ is disjoint. Thus, we have by the third axiom and then using (1) that \begin{align*}
      \Pr{\bigcup_{i=1}^\infty A_i} = \Pr{\bigcup_{i=1}^\infty A'_i} = \sum_{i=1}^\infty \Pr{A'_i} \leq \sum_{i=1}^\infty \Pr{A_i}.
    \end{align*}
  \end{enumerate}
\end{solution}

\begin{solution}{graph_random_walk}
  We show that any vertex $v$ is visited eventually with probability $1$.

  We denote by $w \to v$ the event that the random walk starting at vertex $w$ visits the vertex $v$ eventually, we denote by $\Gamma(w)$ the neighborhood of $w$, and we write $\deg(w) \defeq |\Gamma(w)|$.
  We have, \begin{align*}
    \Pr{w \to v} &= \sum_{w' \in \Gamma(w)} \Pr{\text{the random walk first visits $w'$}} \cdot \Pr{w' \to v} \margintag{using the law of total probability \eqref{eq:lotp}} \\
    &= \frac{1}{\deg(w)} \sum_{w' \in \Gamma(w)} \Pr{w' \to v}. \margintag{using that the random walk moves to a neighbor uniformly at random}
  \end{align*}

  Take $u$ to be the vertex such that $\Pr{u \to v}$ is minimized. Then, \begin{align*}
    \Pr{u \to v} = \frac{1}{\deg(u)} \sum_{u' \in \Gamma(u)} \underbrace{\Pr{u' \to v}}_{\geq \Pr{u \to v}} \geq \Pr{u \to v}.
  \end{align*}
  That is, for all neighbors $u'$ of $u$, $\Pr{u \to v} = \Pr{u' \to v}$.
  Using that the graph is connected and finite, we conclude $\Pr{u \to v} = \Pr{w \to v}$ for any vertex $w$.
  Finally, note that $\Pr{v \to v} = 1$, and hence, the random walk starting at any vertex $u$ visits the vertex $v$ eventually with probability $1$.
\end{solution}

\begin{solution}{lote}
  Let $\Ind{\sA_i}$ be the indicator random variable for the event $\sA_i$.
  Then, \begin{align*}
    \E{\rX \cdot \Ind{\sA_i}} &= \E{\E{\rX \cdot \Ind{\sA_i} \mid \Ind{\sA_i}}} \margintag{by the tower rule \eqref{eq:tower_rule}} \\
    &= \begin{multlined}[t]
      \E{\rX \cdot \Ind{\sA_i} \mid \Ind{\sA_i} = 1} \cdot \Pr{\Ind{\sA_i} = 1} \\ + 0 \cdot \Pr{\Ind{\sA_i} = 0}
    \end{multlined} \margintag{expanding the outer expectation} \\
    &= \E{\rX \mid \sA_i} \cdot \Pr{\sA_i}. \margintag{the event $\Ind{\sA_i} = 1$ is equivalent to $\sA_i$}
  \end{align*}
  Summing up for all $i$, the left-hand side becomes \begin{align*}
    \E{\rX} = \sum_{i=1}^k \E{\rX \cdot \Ind{\sA_i}}.
  \end{align*}
\end{solution}

\begin{solution}{cov_mat_pos_sd}
  Let $\mSigma \defeq \Var{\rX}$ be a covariance matrix of the random vector $\rX$ and fix any $\vz \in \R^n$.
  Then, \begin{align*}
    \transpose{\vz} \mSigma \vz &= \transpose{\vz} \E{(\rX - \E{\rX}) \transpose{(\rX - \E{\rX})}} \vz \margintag{using the definitiion of variance \eqref{eq:variance}} \\
    &= \E{\transpose{\vz} (\rX - \E{\rX}) \transpose{(\rX - \E{\rX})} \vz}. \margintag{using linearity of expectation \eqref{eq:linearity_expectation}}
  \intertext{Define the random variable $Z \defeq \transpose{\vz} (\rX - \E{\rX})$. Then,}
    &= \E{Z^2} \geq 0.
  \end{align*}
\end{solution}

\begin{solution}{bayes_rule}
  Let us start by defining some events that we will reason about later.
  For ease of writing, let us call the person in question X. \begin{align*}
    D &= \text{X has the disease}, \\
    P &= \text{The test shows a positive response}.
  \end{align*}
  Now we can translate the information in the question to formal statements in terms of $D$ and $P$, \begin{align*}
    \Pr{D} &= 10^{-4} \margintag{the disease is rare} \\
    \Pr{P \mid D} = \Pr{\compl{P} \mid \compl{D}} &= 0.99. \margintag{the test is accurate}
  \end{align*}
  We want to determine $\Pr{D \mid P}$. One can find this probability by using Bayes' rule \eqref{eq:bayes_rule}, \begin{align*}
    \Pr{D \mid P} = \frac{\Pr{P \mid D} \cdot \Pr{D}}{\Pr{P}}.
  \end{align*}
  From the quantities above, we have everything except for $\Pr{P}$. This, however, we can compute using the law of total probability, \begin{align*}
    \Pr{P} &= \Pr{P \mid D} \cdot \Pr{D} + \Pr{P \mid \compl{D}} \cdot \Pr{\compl{D}} \\
    &= 0.99 \cdot 10^{-4} + 0.01 \cdot (1 - 10^{-4}) \\
    &= 0.010098.
  \end{align*}
  Hence, $\Pr{D \mid P} = 0.99 \cdot 10^{-4} / 0.010098 \approx 0.0098 = 0.98\%$.
\end{solution}

\begin{solution}{zero_ev_of_cov_mats}
  \begin{enumerate}[beginpenalty=10000]
    \item Suppose that $\rX$ is not linearly independent.
    Then, there exists some $\valpha \in \R^n \setminus \{\vzero\}$ such that $\transpose{\valpha} \rX = 0$.
    We have that $\valpha$ is an eigenvector of $\Var{\rX}$ with zero eigenvalue since \begin{align*}
      \transpose{\valpha} \Var{\rX} \valpha = \Var{\transpose{\valpha} \rX} = \Var{0} = 0. \margintag{using \cref{eq:linear_map_variance}}
    \end{align*}

    \item Suppose that $\Var{\rX}$ has a zero eigenvalue.
    Let $\vlambda$ be the corresponding eigenvector and consider the suggested linear combination $Y \defeq \transpose{\vlambda} \rX$.
    We have \begin{align*}
      \Var{Y} &= \Var{\transpose{\vlambda} \rX} \\
      &= \transpose{\vlambda} \Var{\rX} \vlambda \margintag{using \cref{eq:linear_map_variance}} \\
      &= \sum_{i=1}^n \vlambda(i) \cdot 0 \margintag{using the property of a zero eigenvalue: $\transpose{\vlambda} \Var{\rX} \vlambda = 0$} \\
      &= 0
    \end{align*} which implies that $Y$ must be a constant.
  \end{enumerate}
\end{solution}

\begin{solution}{gaussian_pdf_prod}
  We need to find $\vmu \in \R^n$ and $\mSigma \in \R^{n \times n}$ such that for all $\vx \in \R^n$, \begin{align}
    \begin{multlined}[t]
      \transpose{(\vx - \vmu)} \inv{\mSigma} (\vx - \vmu) \nonumber \\ = \transpose{(\vx - \vmu_1)} \inv{\mSigma_1} (\vx - \vmu_1) + \transpose{(\vx - \vmu_2)} \inv{\mSigma_2} (\vx - \vmu_2) + \const.
    \end{multlined}\label{eq:gaussian_pdf_prod_goal}
  \end{align}

  The left-hand side of \cref{eq:gaussian_pdf_prod_goal} is equal to \begin{align*}
    \transpose{\vx} \inv{\mSigma} \vx - 2 \transpose{\vx} \inv{\mSigma} \vmu + \transpose{\vmu} \inv{\mSigma} \vmu.
  \end{align*}
  The right-hand side of \cref{eq:gaussian_pdf_prod_goal} is equal to \begin{align*}
    &\begin{multlined}[t]
      \parentheses*{\transpose{\vx} \inv{\mSigma_1} \vx + \transpose{\vx} \inv{\mSigma_2} \vx} - 2 \parentheses*{\transpose{\vx} \inv{\mSigma_1} \vmu_1 + \transpose{\vx} \inv{\mSigma_2} \vmu_2} \\ + \parentheses*{\transpose{\vmu_1} \inv{\mSigma_1} \vmu_1 + \transpose{\vmu_2} \inv{\mSigma_2} \vmu_2}
    \end{multlined} \\
    &= \begin{multlined}[t]
      \transpose{\vx} \parentheses*{\inv{\mSigma_1} + \inv{\mSigma_2}} \vx - 2 \transpose{\vx} \parentheses*{\inv{\mSigma_1} \vmu_1 + \inv{\mSigma_2} \vmu_2} \\ + \parentheses*{\transpose{\vmu_1} \inv{\mSigma_1} \vmu_1 + \transpose{\vmu_2} \inv{\mSigma_2} \vmu_2}.
    \end{multlined}
  \end{align*}
  We observe that both sides are equal up to constant terms if \begin{align*}
    \inv{\mSigma} = \inv{\mSigma_1} + \inv{\mSigma_2} \quad\text{and}\quad \inv{\mSigma} \vmu = \inv{\mSigma_1} \vmu_1 + \inv{\mSigma_2} \vmu_2.
  \end{align*}
\end{solution}

\begin{solution}{grv_uncor_indep}
  Recall that independence of any random vectors $\rX$ and $\rY$ implies that they are uncorrelated,\footnote{This is because the expectation of their product factorizes, see \cref{eq:expectation_product}.} the converse implication is a special property of Gaussian random vectors.

  Consider the joint Gaussian random vector $\rZ \defeq [\rX \; \rY] \sim \N{\vmu}{\mSigma}$ and assume that $\rX \sim \N{\vmu_{\rX}}{\mSigma_{\rX}}$ and $\rY \sim \N{\vmu_{\rY}}{\mSigma_{\rY}}$ are uncorrelated. Then, $\mSigma$ can be expressed as \begin{align*}
    \mSigma = \begin{bmatrix}
      \mSigma_{\rX} & \mzero \\
      \mzero & \mSigma_{\rY} \\
    \end{bmatrix},
  \end{align*} implying that the PDF of $\rZ$ factorizes, \begin{align*}
    &\N[[\vx \; \vy]]{\vmu}{\mSigma} \\
    &= \begin{multlined}[t]
      \frac{1}{\sqrt{\det{2\pi\mSigma_{\rX}} \cdot \det{2\pi\mSigma_{\rY}}}} \exp\left(-\frac{1}{2}\transpose{(\vx-\vmu_{\rX})}\inv{\mSigma_{\rX}}(\vx-\vmu_{\rX}) \right. \\ \left. -\frac{1}{2}\transpose{(\vy-\vmu_{\rY})}\inv{\mSigma_{\rY}}(\vy-\vmu_{\rY})\right)
    \end{multlined} \\
    &= \N[\vx]{\vmu_{\rX}}{\mSigma_{\rX}} \cdot \N[\vy]{\vmu_{\rY}}{\mSigma_{\rY}}. \qedhere
  \end{align*}
\end{solution}

\begin{solution}{marginal_and_cond_gaussian}
  Let $\vx \sim \rX$.
  The joint distribution can be expressed as \begin{align*}
    p(\vx) &= p(\vx_A, \vx_B) \\
    &= \frac{1}{Z} \exp\parentheses*{-\frac{1}{2} \transpose{\begin{bmatrix}
      \vx_A - \vmu_A \\
      \vx_B - \vmu_B
    \end{bmatrix}} \begin{bmatrix}
      \mLambda_{AA} & \mLambda_{AB} \\
      \mLambda_{BA} & \mLambda_{BB}
    \end{bmatrix} \begin{bmatrix}
      \vx_A - \vmu_A \\
      \vx_B - \vmu_B
    \end{bmatrix}}
  \end{align*} where $Z$ denotes the normalizing constant.
  To simplify the notation, we write \begin{align*}
    \begin{bmatrix}
      \vDelta_A \\
      \vDelta_B
    \end{bmatrix} \defeq \begin{bmatrix}
      \vx_A - \vmu_A \\
      \vx_B - \vmu_B
    \end{bmatrix}.
  \end{align*}

  \begin{enumerate}
    \item We obtain \begin{align*}
      &p(\vx_A) \\
      &= \frac{1}{Z} \int \exp\parentheses*{-\frac{1}{2} \transpose{\begin{bmatrix}
        \vDelta_A \\
        \vDelta_B
      \end{bmatrix}} \begin{bmatrix}
        \mLambda_{AA} & \mLambda_{AB} \\
        \mLambda_{BA} & \mLambda_{BB}
      \end{bmatrix} \begin{bmatrix}
        \vDelta_A \\
        \vDelta_B
      \end{bmatrix}} \,d\vx_B \margintag{using the sum rule \eqref{eq:sum_rule}} \\
      &= \begin{multlined}[t]
        \frac{1}{Z} \exp\parentheses*{-\frac{1}{2} \brackets*{\transpose{\vDelta_A} (\mLambda_{AA} - \mLambda_{AB} \inv{\mLambda_{BB}} \mLambda_{BA}) \vDelta_A}} \\ \cdot \int \exp\parentheses*{-\frac{1}{2} \brackets*{\transpose{(\vDelta_\sB + \inv{\mLambda_{\sB\sB}}\mLambda_{\sB\sA}\vDelta_\sA)}\mLambda_{\sB\sB}(\vDelta_\sB + \inv{\mLambda_{\sB\sB}}\mLambda_{\sB\sA}\vDelta_\sA)}} \,d\vx_B.
      \end{multlined} \margintag{using the first hint} \\
    \intertext{Observe that the integrand is an unnormalized Gaussian PDF, and hence, the integral evaluates to $\sqrt{\det{2 \pi \inv{\mLambda_{BB}}}}$ (so is constant with respect to $\vx_A$). We can therefore simplify to}
      &= \frac{1}{Z'} \exp\parentheses*{-\frac{1}{2} \brackets*{\transpose{\vDelta_A} (\mLambda_{AA} - \mLambda_{AB} \inv{\mLambda_{BB}} \mLambda_{BA}) \vDelta_A}} \\
      &= \frac{1}{Z'} \exp\parentheses*{-\frac{1}{2} \brackets*{\transpose{\vDelta_A} \inv{\mSigma_{AA}} \vDelta_A}}. \margintag{using the second hint}
    \end{align*}

    \item We obtain \begin{align*}
      &p(\vx_B \mid \vx_A) \\
      &= \frac{p(\vx_A, \vx_B)}{p(\vx_A)} \margintag{using the definition of conditional distributions \eqref{eq:cond_distr}} \\
      &= \frac{1}{Z'} \exp\parentheses*{-\frac{1}{2} \transpose{\begin{bmatrix}
        \vDelta_A \\
        \vDelta_B
      \end{bmatrix}} \begin{bmatrix}
        \mLambda_{AA} & \mLambda_{AB} \\
        \mLambda_{BA} & \mLambda_{BB}
      \end{bmatrix} \begin{bmatrix}
        \vDelta_A \\
        \vDelta_B
      \end{bmatrix}} \margintag{noting that $p(\vx_A)$ is constant with respect to $\vx_B$} \\
      &= \begin{multlined}[t]
        \frac{1}{Z'} \exp\parentheses*{-\frac{1}{2} \brackets*{\transpose{\vDelta_A} (\mLambda_{AA} - \mLambda_{AB} \inv{\mLambda_{BB}} \mLambda_{BA}) \vDelta_A}} \\ \cdot \exp\parentheses*{-\frac{1}{2} \brackets*{\transpose{(\vDelta_\sB + \inv{\mLambda_{\sB\sB}}\mLambda_{\sB\sA}\vDelta_\sA)}\mLambda_{\sB\sB}(\vDelta_\sB + \inv{\mLambda_{\sB\sB}}\mLambda_{\sB\sA}\vDelta_\sA)}}
      \end{multlined} \margintag{using the first hint} \\
      &= \frac{1}{Z''} \exp\parentheses*{-\frac{1}{2} \brackets*{\transpose{(\vDelta_\sB + \inv{\mLambda_{\sB\sB}}\mLambda_{\sB\sA}\vDelta_\sA)}\mLambda_{\sB\sB}(\vDelta_\sB + \inv{\mLambda_{\sB\sB}}\mLambda_{\sB\sA}\vDelta_\sA)}}. \margintag{observing that the first exponential is constant with respect to $\vx_B$}
    \end{align*}
    Finally, observe that \begin{align*}
      \vDelta_\sB + \inv{\mLambda_{\sB\sB}}\mLambda_{\sB\sA}\vDelta_\sA &= \vmu_B - \inv{\mLambda_{\sB\sB}}\mLambda_{\sB\sA}(\vx_\sA - \vmu_\sA) = \vmu_{\sB\mid\sA} \quad\text{and} \margintag{using the third hint} \\
      \inv{\mLambda_{\sB\sB}} &= \mSigma_{\sB\sB} - \mSigma_{\sB\sA} \inv{\mSigma_{\sA\sA}} \mSigma_{\sA\sB} = \mSigma_{\sB\mid\sA}. \margintag{using the second hint}
    \end{align*}
    Thus, $\vx_B \mid \vx_A \sim \N{\vmu_{\sB\mid\sA}}{\mSigma_{\sB\mid\sA}}$.
  \end{enumerate}
\end{solution}

\begin{solution}{gaussian_closedness}
  \begin{enumerate}[beginpenalty=10000]
    \item Let $\rY \defeq \mA \rX + \vb$ and define $\vs \defeq \transpose{\mA} \vt$.
    We have for the MGF of $\rY$ that \begin{align*}
      \varphi_\rY(\vt) &= \E*{\exp\parentheses*{\transpose{\vt} \rY}} \\
      &= \E*{\exp\parentheses*{\transpose{\vt} \mA \rX} \cdot \exp\parentheses*{\transpose{\vt} \vb}} \\
      &= \E*{\exp\parentheses*{\transpose{\vs} \rX}} \cdot \exp\parentheses*{\transpose{\vt} \vb} \\
      &= \varphi_\rX(\vs) \cdot \exp\parentheses*{\transpose{\vt} \vb} \\
      &= \exp\parentheses*{\transpose{\vs} \vmu + \frac{1}{2}\transpose{\vs}\mSigma\vs} \cdot \exp\parentheses*{\transpose{\vt} \vb} \\
      &= \exp\parentheses*{\transpose{\vt} (\mA\vmu + \vb) + \frac{1}{2}\transpose{\vt} \mA \mSigma \transpose{\mA} \vt},
    \end{align*} which implies that $\rY \sim \N{\mA \vmu + \vb}{\mA \mSigma \transpose{\mA}}$.

    \item We have for the MGF of $\rY \defeq \rX + \rX'$ that \begin{align*}
      \varphi_\rY(\vt) &= \E*{\exp\parentheses*{\transpose{\vt} \rY}} \\
      &= \E*{\exp\parentheses*{\transpose{\vt} \rX}} \cdot \E*{\exp\parentheses*{\transpose{\vt} \rX'}} \margintag{using the independence of $\rX$ and $\rX'$} \\
      &= \varphi_\rX(\vt) \cdot \varphi_{\rX'}(\vt) \\
      &= \exp\parentheses*{\transpose{\vt} \vmu + \frac{1}{2}\transpose{\vt}\mSigma\vt} \cdot \exp\parentheses*{\transpose{\vt} \vmu' + \frac{1}{2}\transpose{\vt}\mSigma'\vt} \\
      &= \exp\parentheses*{\transpose{\vt} (\vmu + \vmu') + \frac{1}{2}\transpose{\vt} (\mSigma + \mSigma') \vt},
    \end{align*} which implies that $\rY \sim \N{\vmu + \vmu'}{\mSigma + \mSigma'}$.
  \end{enumerate}
\end{solution}

\begin{solution}{expectation_and_variance_of_gaussians}
  Note that if $Y \sim \N{0}{1}$ is a (univariate) standard normal random variable, then \begin{align*}
    \E{Y} &= \frac{1}{\sqrt{2\pi}} \int_{-\infty}^\infty y \cdot \exp\parentheses*{-\frac{1}{2} y^2} \,d y \margintag{using the PDF of the standard normal distribution \eqref{eq:univ_normal}} \\
    &= \frac{1}{\sqrt{2\pi}} \parentheses*{\int_{-\infty}^0 y \cdot \exp\parentheses*{-\frac{1}{2} y^2} \,d y + \int_0^\infty y \cdot \exp\parentheses*{-\frac{1}{2} y^2} \,d y} \\
    &= \frac{1}{\sqrt{2\pi}} \parentheses*{\left[ - \exp\parentheses*{-\frac{1}{2} y^2} \right]_{-\infty}^0 + \left[ - \exp\parentheses*{-\frac{1}{2} y^2} \right]_0^\infty} \\
    &= \frac{1}{\sqrt{2\pi}} ([-1 + 0] + [0 + 1]) = 0, \\
    \Var{Y} &= \E{Y^2} - \underbrace{\E{Y}^2}_{0} \margintag{using the definition of variance \eqref{eq:variance2}} \\
    &= \frac{1}{\sqrt{2\pi}} \int_{-\infty}^\infty y^2 \cdot \exp\parentheses*{-\frac{1}{2} y^2} \,d y \margintag{using the PDF of the standard normal distribution \eqref{eq:univ_normal}} \\
    &= \frac{1}{\sqrt{2\pi}} \parentheses*{\int_{-\infty}^0 y^2 \cdot \exp\parentheses*{-\frac{1}{2} y^2} \,d y + \int_0^\infty y^2 \cdot \exp\parentheses*{-\frac{1}{2} y^2} \,d y} \\
    &= \begin{multlined}[t]
      \frac{1}{\sqrt{2\pi}} \left(\left[- y \cdot \exp\parentheses*{-\frac{1}{2} y^2} \right]_{-\infty}^0 + \int_{-\infty}^0 \exp\parentheses*{-\frac{1}{2} y^2} \,d y \right. \\ \left. + \left[- y \cdot \exp\parentheses*{-\frac{1}{2} y^2} \right]_0^\infty + \int_0^\infty \exp\parentheses*{-\frac{1}{2} y^2} \,d y\right)
    \end{multlined} \margintag{integrating by parts} \\
    &= \frac{1}{\sqrt{2\pi}} \parentheses*{\int_{-\infty}^0 \exp\parentheses*{-\frac{1}{2} y^2} \,d y + \int_0^\infty \exp\parentheses*{-\frac{1}{2} y^2} \,d y} \\
    &= \frac{1}{\sqrt{2\pi}} \int_{-\infty}^\infty \exp\parentheses*{-\frac{1}{2} y^2} \,d y = 1. \margintag{a PDF integrates to $1$}
  \end{align*}

  Recall from \cref{eq:gaussian_affine_transformation} that we can express $\rX \sim \N{\vmu}{\mSigma}$ as \begin{align*}
    \rX = \msqrt{\mSigma} \rY + \vmu
  \end{align*} where $\rY \sim \SN$.
  Using that $\rY$ is a vector of independent univariate standard normal random variables, we conclude that $\E{\rY} = \vzero$ and $\Var{\rY} = \mI$.
  We obtain \begin{align*}
    \E{\rX} &= \E{\msqrt{\mSigma} \rY + \vmu} = \msqrt{\mSigma} \underbrace{\E{\rY}}_{\vzero} + \vmu = \vmu, \quad\text{and} \margintag{using linearity of expectation \eqref{eq:linearity_expectation}} \\
    \Var{\rX} &= \Var{\msqrt{\mSigma} \rY + \vmu} = \msqrt{\mSigma} \underbrace{\Var{\rY}}_{\mI} \transpose{{\msqrt{\mSigma}}} = \mSigma. \margintag{using \cref{eq:linear_map_variance}}
  \end{align*}
\end{solution}

\begin{solution}{non_affine_transformations_of_gaussians}
  \begin{enumerate}[beginpenalty=10000]
    \item \textbf{Yes}. In the following, we give one example of a non-affine transformation that preserves the Gaussianity of a random vector.

    Let $\rX \sim \N{\vmu}{\mSigma}$ be a Gaussian random vector in $\R^d$.
    Define the coordinate-wise transformation $\vphi : \R^d \to \R^d$ as $(\vphi(\vx))_i = \phi_i(x_i)$ with \begin{align*}
      \phi_i(x_i) \defeq \begin{cases}
        - (x_i - \mu_i) + \mu_i & \abs{x_i - \mu_i} < 1 \\
        x_i & \text{otherwise}.
      \end{cases}
    \end{align*}
    Intuitively, $\phi_i$ simply flips all the ``mass'' in a neighborhood of $\mu_i$ on the $i$-th coordinate.
    Due to the symmetry of the Gaussian distribution around its mean, you could easily imagine that the transformation still preserves the Gaussian property.
    Also, this function cannot be an affine transformation since it is not continuous.

    To prove Gaussianity, we can use the change of variables formula \eqref{eq:change_of_variables}.
    Let \begin{align*}
      \rY \defeq \vphi(\rX), \quad p_{\rY}(\vy) = p_{\rX}(\vphi^{-1}(\vy)) \cdot \abs{\det{\jac \vphi^{-1}(\vy)}} \quad (\forall \vy \in \R^d).
    \end{align*}
    There are two cases for $\vy$: \begin{enumerate}
      \item If $\abs{y_i - \mu_i} < 1$, then $x_i = \phi_i^{-1}(y_i) = -(y_i - \mu_i) + \mu_i$.
      Also note that in this case, the Jacobian matrix is simply a diagonal matrix with $-1$ in the $i$-th position and $1$ elsewhere.
      Hence, its determinant is $-1$.
      We have \begin{align*}
        p_{\rY}(\vy) &= p_{\rX}(\vphi^{-1}(\vy)) \cdot \abs{\det{\jac \vphi^{-1}(\vy)}} \\
        &= p_{\rX}(\vphi^{-1}(\vy)) \\
        &= p_{\rX}\parentheses*{\begin{bmatrix}
          \vdots \\
          -(y_i - \mu_i) + \mu_i \\
          \vdots
        \end{bmatrix}} \\
        &= p_{\rX}\parentheses*{\begin{bmatrix}
          \vdots \\
          y_i \\
          \vdots
        \end{bmatrix}} \margintag{since $p_{\rX}$ is symmetric w.r.t. $\mu_i$} \\
        &= p_{\rX}(\vy).
      \end{align*}

      \item If $\abs{y_i - \mu_i} \geq 1$, then $p_{\rY}(\vy) = p_{\rX}(\vy)$ since $\vphi$ is the identity function in this case.
    \end{enumerate}\vspace{1ex}
    Thus, $\rY$ has an identical PDF to $\rX$ and is therefore Gaussian.

    \item \textbf{Yes}.
    It is difficult to calculate the distribution of $W \defeq \frac{X + YZ}{\sqrt{1 + Z^2}}$ directly since $Z$ is in the denominator.
    The trick here is to first condition $W$ on $Z$, and then integrate over the distribution of $Z$.
    For all $z \in \R$, we have \begin{align*}
      p(W \mid Z = z) &= p\left(\frac{X + YZ}{\sqrt{1 + Z^2}} \middle| Z = z\right) \\
      &= p\left(\frac{X + Yz}{\sqrt{1 + z^2}}\right) \margintag{since $X, Y \perp Z$} \\
      &= \N[W]{0}{1} \margintag{as $X + Yz \sim \N{0}{1 + z^2}$}
    \end{align*} which means that $W \mid Z = z \sim \N{0}{1}$ which is \emph{independent} of $z$!
    Using the law of total probability~\eqref{eq:lotp}, we can now write \begin{align*}
      p(W) &= \int p(W \mid Z = z) \cdot p_Z(z) \,d z \\
      &= \N[W]{0}{1} \int p_Z(z) \,d z \\
      &= \N[W]{0}{1}.
    \end{align*}
  \end{enumerate}
\end{solution}

\begin{solution}{decision_theory}
  \begin{enumerate}[beginpenalty=10000]
    \item We have \begin{align*}
      \E*[y]{[(y - a)^2 \mid \vx]} &= a^2 - 2 a \E[y]{y}[\vx] + \const,
    \end{align*} which is easily seen to be minimized when $a = \E[y]{y}[\vx]$.

    \item We have \begin{align*}
      \E[y]{r(y, a)}[\vx] = c_1 \int_a^\infty (y-a) p(y \mid \vx) \,d y + c_2 \int_{-\infty}^a (a-y) p(y \mid \vx) \,d y.
    \end{align*}
    We can differentiate this expression with respect to $a$: \begin{align*}
      \odv{}{a} \E[y]{r(y, a)}[\vx] &= - c_1 \int_a^\infty p(y \mid \vx) \,d y + c_2 \int_{-\infty}^a p(y \mid \vx) \,d y \\
      &= - c_1 \Pr{y \geq a \mid \vx} + c_2 \Pr{y \leq a \mid \vx} \\
      &= - c_1 (1 - \Pr{y \leq a \mid \vx}) + c_2 \Pr{y \leq a \mid \vx} \\
      &= - c_1 + (c_1 + c_2) \Pr{y \leq a \mid \vx}.
    \end{align*}
    Setting this equal to zero, we find the critical point condition \begin{align*}
      \Pr{y \leq a \mid \vx} \overset{!}{=} \frac{c_1}{c_1 + c_2}.
    \end{align*}
    We obtain $\opt{a}(\vx) = \mu_\vx + \sigma_\vx \cdot \inv{\Phi}\parentheses*{\frac{c_1}{c_1 + c_2}}$ by transforming to a standard normal random variable.
  \end{enumerate}
\end{solution}

\section*{\nameref{sec:blr}}

\begin{solution}{closed_form_linear_regression}
  \begin{enumerate}[beginpenalty=10000]
    \item We begin by deriving the gradient and Hessian of the least squares and ridge losses.
    For least squares, \begin{align*}
      \grad_{\vw} \norm{\vy - \mX \vw}_2^2 &= \grad_{\vw} (\transpose{\vw} \transpose{\mX} \mX \vw - 2 \transpose{\vw} \transpose{\mX} \vy + \norm{\vy}_2^2) \\
      &= 2 (\transpose{\mX} \mX \vw - \transpose{\mX} \vy), \\
      \hes_{\vw} \norm{\vy - \mX \vw}_2^2 &= 2 \transpose{\mX} \mX.
    \end{align*}
    Similarly, for ridge regression, \begin{align*}
      \grad_{\vw} \norm{\vy - \mX \vw}_2^2 + \lambda \norm{\vw}_2^2 &= 2 (\transpose{\mX} \mX \vw - \transpose{\mX} \vy) + 2 \lambda \vw, \\
      \hes_{\vw} \norm{\vy - \mX \vw}_2^2 + \lambda \norm{\vw}_2^2 &= 2 \transpose{\mX} \mX + 2 \lambda \mI.
    \end{align*}
    From the assumption that the Hessians are positive definite, we know that any minimizer is a unique globally optimal solution (due to strict convexity), and that $\transpose{\mX} \mX$ and $\transpose{\mX} \mX + \lambda \mI$ are invertible.

    Using the first-order optimality condition for convex functions, we attain the solutions to least squares and ridge regression by setting the respective gradient to $\vzero$.

    \item We choose $\vwhat_{\ls}$ such that $\mX \vwhat_{\ls} = \mPi_{\mX} \vy$.
    This implies that $\vy - \mX \vwhat_{\ls} \perp \mX \vw$ for all $\vw \in \R^d$.
    In other words, $\transpose{(\vy - \mX \vwhat_{\ls})} \mX \vw = 0$ for all $\vw$, which implies that $\transpose{(\vy - \mX \vwhat_{\ls})} \mX = \vzero$.
    By simple algebraic manipulation it can be seen that this condition is equivalent to the gradient condition of the previous problem.
  \end{enumerate}
\end{solution}

\begin{solution}{noise_variance_mle}
  To prevent confusion with the length of the data set $n$, we drop the subscript from the noise variance $\sigman^2$ in the following.
  We have \begin{align*}
    \hat{\sigma} &= \argmax_{\sigma} \sum_{i=1}^n \log p(y_i \mid \vx_i, \sigma) \\
    &= \argmin_{\sigma} \sum_{i=1}^n \frac{1}{2} \log(2 \sigma^2 \pi) + \frac{1}{2 \sigma^2} (y_i - \transpose{\vw} \vx_i)^2 \\
    &= \argmin_{\sigma} \frac{n}{2} \log \sigma^2 + \frac{1}{2 \sigma^2} \sum_{i=1}^n (y_i - \transpose{\vw} \vx_i)^2.
  \end{align*}
  We can solve this optimization problem by differentiating and setting to zero: \begin{align*}
    \pdv{}{\sigma^2} \brackets*{\frac{n}{2} \log \sigma^2 + \frac{1}{2 \sigma^2} \sum_{i=1}^n (y_i - \transpose{\vw} \vx_i)^2} &= 0 \\
    \frac{n}{2 \sigma^2} - \frac{1}{2 \sigma^4} \sum_{i=1}^n (y_i - \transpose{\vw} \vx_i)^2 &= 0 \\
    n - \frac{1}{\sigma^2} \sum_{i=1}^n (y_i - \transpose{\vw} \vx_i)^2 &= 0 \margintag{multiplying through by $2 \sigma^2$}.
  \end{align*}
  The desired result follows by solving for $\sigma^2$.
\end{solution}

\begin{solution}{variance_around_training_data}
  Let us first derive the variance of the least squares estimate: \begin{align}
    \Var{\vwhat_{\ls}}[\mX] &= \Var{\inv{(\transpose{\mX} \mX)} \transpose{\mX} \vy}[\mX] \margintag{using \cref{eq:linear_regression}} \nonumber\\
    &= \inv{(\transpose{\mX} \mX)} \transpose{\mX} \Var{\vy}[\mX] \transpose{(\inv{(\transpose{\mX} \mX)} \transpose{\mX})} \margintag{using \cref{eq:linear_map_variance}} \nonumber\\
    &= \inv{(\transpose{\mX} \mX)} \transpose{\mX} \Var{\vy}[\mX] \mX \inv{(\transpose{\mX} \mX)}. \margintag{using $\inv{(\transpose{\mA})} = \transpose{(\inv{\mA})}$}
    \intertext{Due to the Gaussian likelihood \eqref{eq:blr_likelihood}, $\Var{\vy}[\mX] = \sigman^2\mI$, so}
    &= \sigman^2 \inv{(\transpose{\mX} \mX)}. \label{eq:least_squares_variance}
  \end{align}
  In the two-dimensional setting, i.e., data is of the form $\vx_i = \transpose{[1 \; x_i]}\ (x_i \in \R)$, we have \begin{align*}
    \transpose{\mX} \mX = \begin{bmatrix}
      n & \sum x_i \\
      \sum x_i & \sum x_i^2 \\
    \end{bmatrix}.
  \end{align*}
  Thus, \begin{align*}
    \Var{\vwhat_{\ls}}[\mX] = \sigman^2 \inv{(\transpose{\mX} \mX)} = \frac{\sigman^2}{Z} \begin{bmatrix}
      \sum x_i^2 & -\sum x_i \\
      -\sum x_i & n \\
    \end{bmatrix} \margintag{using \cref{eq:least_squares_variance}}
  \end{align*} where $Z \defeq n (\sum x_i^2) - \parentheses*{\sum x_i}^2$.

  Therefore, the predictive variance at a point $\transpose{[1 \; \xs]}$ is \begin{align*}
    \begin{bmatrix}
      1 & \xs
    \end{bmatrix} \Var{\vwhat_{\ls}} \begin{bmatrix}
      1 \\
      \xs \\
    \end{bmatrix} = \frac{\sigman^2}{Z} \sum x_i^2 - 2 x_i \xs + (\xs)^2 = \frac{\sigman^2}{Z} \sum (x_i - \xs)^2.
  \end{align*}
  Thus, the predictive variance is minimized for $\xs = \frac{1}{n} \sum x_i$.
\end{solution}

\begin{solution}{blr}
  \begin{enumerate}[beginpenalty=10000]
    \item Recall from \cref{sec:least_squares_as_mle} that the MLE and least squares estimate coincide if the noise is zero-mean Gaussian. We therefore have, \begin{align*}
      \vwhat_\MLE = \vwhat_{\ls} = \inv{(\transpose{\mX}\mX)}\transpose{\mX}\vy = \begin{bmatrix}
        0.63 \\
        1.83
      \end{bmatrix}.
    \end{align*}

    \item Recall from \cref{eq:blr_posterior} that $\vw \mid \mX, \vy \sim \N{\vmu}{\mSigma}$ with \begin{align*}
      \mSigma &= \inv{\parentheses*{\sigman^{-2} \transpose{\mX}\mX + \sigmap^{-2} \mI}} \quad\text{and} \\
      \vmu &= \sigman^{-2} \mSigma \transpose{\mX} \vy.
    \end{align*}
    Inserting $\sigman^2 = 0.1$ and $\sigmap^2 = 0.05$ yields, \begin{align*}
        \mSigma = \begin{bmatrix}
        0.019 & -0.014 \\
        -0.014 & 0.019
      \end{bmatrix}.
    \end{align*}
    Then, \begin{align*}
      \vwhat_\MAP = \vmu = \begin{bmatrix}
        0.91 \\
        1.31
      \end{bmatrix}.
    \end{align*}

    \item Recall from \cref{eq:blr_pred_posterior} that $\ys \mid \mX, \vy, \vxs \sim \N{\mu_\ys}{\sigma_{\ys}^2}$ with \begin{align*}
      \mu_\ys = \transpose{\vmu} \vxs \quad\text{and}\quad \sigma_{\ys}^2 = \transpose{\vxs} \mSigma \vxs + \sigman^2.
    \end{align*}
    Inserting $\vxs = \transpose{[3 \; 3]}$, and $\sigman^2, \vmu$ and $\mSigma$ from above yields, \begin{align*}
      \mu_\ys = 6.66 \quad\text{and}\quad \sigma_{\ys}^2 = 0.186.
    \end{align*}

    \item One would have to let $\sigmap^2 \to \infty$.
  \end{enumerate}
\end{solution}

\begin{solution}{online_blr}
  We denote by $\mX_t$ the design matrix and by $\vy_t$ the vector of observations including the first $t$ data points.
  \begin{enumerate}
    \item Note that \begin{align*}
      \transpose{\mX_t} \mX_t = \sum_{i=1}^t \vx_i \transpose{\vx_i} \quad\text{and}\quad \transpose{\mX_t} \vy_t = \sum_{i=1}^t y_i \vx_i.
    \end{align*}
    This means that after observing the $(t+1)$-st data point, we have that \begin{align*}
      \transpose{\mX_{t+1}} \mX_{t+1} &= \transpose{\mX_t} \mX_t + \vx_{t+1} \transpose{\vx_{t+1}} \quad\text{and} \\
      \transpose{\mX_{t+1}} \vy_{t+1} &= \transpose{\mX_t} \vy_t + y_{t+1} \vx_{t+1}.
    \end{align*}
    Hence, by just keeping $\transpose{\mX_t} \mX_t$ (which is a $d \times d$ matrix) and $\transpose{\mX_t} \vy_t$ (which is a vector in $\R^d$) in memory, and updating them as above, we do not need to keep the whole data in memory.

    \item One has to compute $\inv{(\sigman^{-2}\transpose{\mX_t}\mX_t + \sigmap^{-2}\mI)}$ for finding $\vmu$ and $\mSigma$ in every round.
    We can write \begin{align*}
      \inv{(\sigman^{-2}\transpose{\mX_{t+1}}\mX_{t+1} + \sigmap^{-2}\mI)} &= \sigman^2\inv{(\transpose{\mX_{t+1}}\mX_{t+1} + \sigman^2\sigmap^{-2}\mI)} \\
      &= \sigman^2\inv{(\underbrace{\transpose{\mX_t}\mX_t + \sigman^2\sigmap^{-2}\mI}_{\mA_t} + \vx_{t+1} \transpose{\vx_{t+1}})}
    \end{align*} where $\mA_t \in \R^{d \times d}$.
    Using the Woodbury matrix identity \eqref{eq:woodbury} and that we know the inverse of $\mA_t$ (from the previous iteration), the computation of the inverse of $(\mA_t + \vx_{t+1} \transpose{\vx_{t+1}})$ is of $\BigO{d^2}$, which is much better than computing the inverse of $(\mA_t + \vx_{t+1} \transpose{\vx_{t+1}})$ from scratch.
  \end{enumerate}
\end{solution}

\begin{solution}{aleatoric_and_epistemic_uncertainty}
  The law of total variance \eqref{eq:lotv_interpretation} yields the following decomposition of the predictive variance, \begin{align*}
    \Var[\ys]{\ys}[\vxs, \vx_{1:n}, y_{1:n}] = \begin{multlined}[t]
      \b{\E[\vw]{\Var[\ys]{\ys}[\vxs, \vw]}[\vx_{1:n}, y_{1:n}]} \\ + \r{\Var[\vw]{\E[\ys]{\ys}[\vxs, \vw]}[\vx_{1:n}, y_{1:n}]}
    \end{multlined}
  \end{align*} wherein the first term corresponds to the \b{aleatoric uncertainty} and the second term corresponds to the \r{epistemic uncertainty}.

  The aleatoric uncertainty is given by \begin{align*}
    \E[\vw]{\Var[\ys]{\ys}[\vxs, \vw]}[\vx_{1:n}, y_{1:n}] = \E[\vw]{\sigman^2}[\vx_{1:n}, y_{1:n}] = \sigman^2 \margintag{using the definition of $\sigman^2$. \eqref{eq:blr_likelihood}}
  \end{align*}
  For the epistemic uncertainty, \begin{align*}
    \Var[\vw]{\E[\ys]{\ys}[\vxs, \vw]}[\vx_{1:n}, y_{1:n}] &= \Var[\vw]{\transpose{\vw} \vxs}[\vx_{1:n}, y_{1:n}] \margintag{using that $\ys = \transpose{\vw} \vxs + \varepsilon$ where $\varepsilon$ is zero-mean noise} \\[5pt]
    &= \transpose{\vxs} \Var[\vw]{\vw}[\vx_{1:n}, y_{1:n}] \vxs \margintag{using \cref{eq:linear_map_variance}} \\
    &= \transpose{\vxs} \mSigma \vxs \margintag{using \cref{eq:blr_posterior}}
  \end{align*} where $\mSigma$ is the posterior covariance matrix.
\end{solution}

\begin{solution}{hyperpriors}
  \begin{enumerate}[beginpenalty=10000]
    \item Let \( \vf \defeq \mX \vw \).
    Since \( \vw \mid \vmu, \lambda \sim \N{\vmu}{\lambda^{-1} \mI_d} \), we have \[
      \vf \mid \mX, \vmu, \lambda \sim \N{\mX \vmu}{\lambda^{-1} \mX \mX^\top}. \margintag{using \cref{eq:linear_map_variance}}
    \]
    Thus, \[
      \vy \mid \mX, \vmu, \lambda \sim \N{\mX \vmu}{\lambda^{-1} \mX \mX^\top + \lambda^{-1} \mI_n}.
    \]

    \item We will denote the MLE by \( \widehat{\vmu}_1 \).
    We have \[
      \widehat{\vmu}_1 = \argmax_{\vmu} p(\vy \mid \mX, \vmu, \lambda) = \argmax_{\vmu} \log \N[\vy]{\mX \vmu}{\mSigma_{\vy}}.
    \]
    We can simplify \begin{align*}
      - \log \N[\vy]{\mX \vmu}{\mSigma_{\vy}} &= \frac{1}{2} (\vy - \mX \vmu)^\top \mSigma_{\vy}^{-1} (\vy - \mX \vmu) + \const \\
      &= \frac{1}{2} \vmu^\top \mX^\top \mSigma_{\vy}^{-1} \mX \vmu - \vmu^\top \mX^\top \mSigma_{\vy}^{-1} \vy + \const.
    \end{align*}
    Taking the gradient with respect to \( \vmu \), we obtain \[
      \grad_{\vmu} \left( \frac{1}{2} \vmu^\top \mX^\top \mSigma_{\vy}^{-1} \mX \vmu - \vy^\top \mSigma_{\vy}^{-1} \mX \vmu + \const \right) = \mX^\top \mSigma_{\vy}^{-1} \mX \vmu - \mX^\top \mSigma_{\vy}^{-1} \vy.
    \]
    This is zero iff \( \vmu = (\mX^\top \mSigma_{\vy}^{-1} \mX)^{-1} \mX^\top \mSigma_{\vy}^{-1} \vy \).

    \item By Bayes' rule~\eqref{eq:bayes_rule}, \[
      p(\vmu \mid \mX, \vy, \lambda) \propto p(\vy \mid \mX, \vmu, \lambda) \cdot p(\vmu).
    \]
    Taking the negative logarithm, \begin{align*}
      - \log p(\vmu \mid \mX, \vy, \lambda) &= - \log p(\vy \mid \mX, \vmu, \lambda) - \log p(\vmu) + \const.
      \intertext{Analogously to the previous question, we can simplify}
      &= \frac{1}{2} (\vy - \mX \vmu)^\top \mSigma_{\vy}^{-1} (\vy - \mX \vmu) + \frac{1}{2} \vmu^\top \vmu + \const \\
      &= \frac{1}{2} \vmu^\top (\mX^\top \mSigma_{\vy}^{-1} \mX + \mI_d) \vmu - \vmu^\top \mX^\top \mSigma_{\vy}^{-1} \vy + \const.
    \end{align*}
    This is a quadratic in \( \vmu \), so the posterior distribution must be Gaussian.
    By matching the terms with \[ \log \N[\vx]{\vmu'}{\mSigma'} = - \frac{1}{2} \vx^\top \mSigma'^{-1} \vx + \vx^\top \mSigma'^{-1} \vmu' + \const,\] we obtain the covariance matrix \( (\mX^\top \mSigma_{\vy}^{-1} \mX + \mI_d)^{-1} \) and the mean vector \( \mSigma_{\vmu} \mX^\top \mSigma_{\vy}^{-1} \vy \).

    \item Let \( \widetilde{\mSigma}_{\vy} \defeq \lambda \mSigma_{\vy} \).
    By Bayes' rule~\eqref{eq:bayes_rule}, \begin{align*}
      p(\lambda \mid \mX, \vy, \vmu) &\propto p(\lambda) \cdot p(\vy \mid \mX, \vmu, \lambda) \\
      &= e^{-\lambda} \cdot \N[\vy]{\mX \vmu}{\mSigma_{\vy}} \\
      &\propto e^{-\lambda} \cdot \det{\mSigma_{\vy}}^{-1/2} \exp\left( -\frac{1}{2} (\vy - \mX \vmu)^\top \mSigma_{\vy}^{-1} (\vy - \mX \vmu) \right) \\
      &= e^{-\lambda} \cdot \det{\lambda^{-1} \widetilde{\mSigma}_{\vy}}^{-1/2} \exp\left( -\frac{1}{2} (\vy - \mX \vmu)^\top (\lambda^{-1} \widetilde{\mSigma}_{\vy})^{-1} (\vy - \mX \vmu) \right) \\
      &\propto e^{-\lambda} \cdot \lambda^{n / 2} \exp\left( -\frac{\lambda}{2} (\vy - \mX \vmu)^\top \widetilde{\mSigma}_{\vy}^{-1} (\vy - \mX \vmu) \right) \margintag{using that $\inv{\lambda}\widetilde{\mSigma}_{\vy}$ is independent of $\lambda$} \\
      &= \lambda^{n / 2} \exp\left( -\lambda \left[1 + \frac{1}{2} (\vy - \mX \vmu)^\top \widetilde{\mSigma}_{\vy}^{-1} (\vy - \mX \vmu) \right] \right) \\
      &\propto \GammaDistr{\alpha}{\beta}
    \end{align*} with \( \alpha = 1 + \frac{n}{2} \) and \( \beta = 1 + \frac{1}{2} (\vy - \mX \vmu)^\top \widetilde{\mSigma}_{\vy}^{-1} (\vy - \mX \vmu) \).
  \end{enumerate}
\end{solution}

\section*{\nameref{sec:kf}}

\begin{solution}{kf_predictive_distr}
  Recall from \cref{eq:bf_conditioning} that \begin{align}
    p(x_{t+1} \mid y_{1:t+1}) &= \frac{1}{Z} p(x_{t+1} \mid y_{1:t}) p(y_{t+1} \mid x_{t+1}). \label{eq:kf_predictive_distr_start}
  \end{align}

  Using the sensor model \eqref{eq:kf_1d_sensor_model}, \begin{align*}
    p(y_{t+1} \mid x_{t+1}) &= \frac{1}{Z'} \exp\parentheses*{-\frac{1}{2} \frac{(y_{t+1} - x_{t+1})^2}{\sigma_y^2}}.
  \end{align*}

  It remains to compute the predictive distribution, \begin{align*}
    p(x_{t+1} \mid y_{1:t}) &= \int p(x_{t+1} \mid x_t) p(x_t \mid y_{1:t}) \,d x_t \margintag{using \cref{eq:bf_prediction}} \\
    &= \frac{1}{Z''} \int \exp\parentheses*{-\frac{1}{2} \brackets*{\frac{(x_{t+1} - x_t)^2}{\sigma_x^2} + \frac{(x_t - \mu_t)^2}{\sigma_t^2}}} \,d x_t \margintag{using the motion model \eqref{eq:kf_1d_motion_model} and previous update} \\
    &= \frac{1}{Z''} \int \exp\parentheses*{-\frac{1}{2} \brackets*{\frac{\sigma_t^2 (x_{t+1} - x_t)^2 + \sigma_x^2 (x_t - \mu_t)^2}{\sigma_t^2 \sigma_x^2}}} \,d x_t.
  \intertext{The exponent is the sum of two expressions that are quadratic in $x_t$. \midx<Completing the square>{completing the square} allows rewriting any quadratic $a x^2 + b x + c$ as the sum of a squared term $a (x + \frac{b}{2 a})^2$ and a residual term $c - \frac{b^2}{4 a}$ that is independent of $x$. In this case, we have $a = (\sigma_t^2 + \sigma_x^2) / (\sigma_t^2 \sigma_x^2)$, $b = -2 (\sigma_t^2 x_{t+1} + \sigma_x^2 \mu_t) / (\sigma_t^2 \sigma_x^2)$, and $c = (\sigma_t^2 x_{t+1}^2 + \sigma_x^2 \mu_t^2) / (\sigma_t^2 \sigma_x^2)$. The residual term can be taken outside the integral, giving}
    &= \frac{1}{Z''} \exp\parentheses*{-\frac{1}{2} \brackets*{c - \frac{b^2}{4 a}}} \int \exp\parentheses*{-\frac{a}{2} \brackets*{x_t + \frac{b}{2 a}}^2} \,d x_t.
  \intertext{The integral is simply the integral of a Gaussian over its entire support, and thus evaluates to $1$. We are therefore left with only the residual term from the quadratic. Plugging back in the expressions for $a, b$, and $c$ and simplifying, we obtain}
    &= \frac{1}{Z''} \exp\parentheses*{-\frac{1}{2} \frac{(x_{t+1} - \mu_t)^2}{\sigma_t^2 + \sigma_x^2}}.
  \end{align*}
  That is, $X_{t+1} \mid y_{1:t} \sim \N{\mu_t}{\sigma_t^2 + \sigma_x^2}$.

  Plugging our results back into \cref{eq:kf_predictive_distr_start}, we obtain \begin{align*}
    p(x_{t+1} \mid y_{1:t+1}) &= \frac{1}{Z'''}  \exp\parentheses*{-\frac{1}{2} \brackets*{\frac{(x_{t+1} - \mu_t)^2}{\sigma_t^2 + \sigma_x^2} + \frac{(y_{t+1} - x_{t+1})^2}{\sigma_y^2}}}.
  \intertext{Completing the square analogously to our derivation of the predictive distribution yields,}
    &= \frac{1}{Z'''} \exp\parentheses*{-\frac{1}{2} \frac{\parentheses*{x_{t+1} - \frac{(\sigma_t^2 + \sigma_x^2) y_{t+1} + \sigma_y^2 \mu_t}{\sigma_t^2 + \sigma_x^2 + \sigma_y^2}}^2}{\frac{(\sigma_t^2 + \sigma_x^2) \sigma_y^2}{\sigma_t^2 + \sigma_x^2 + \sigma_y^2}}}.
  \end{align*}
  Hence, $X_{t+1} \mid y_{t+1} \sim \N{\mu_{t+1}}{\sigma_{t+1}^2}$ as defined in \cref{eq:kalman_posterior}.
\end{solution}

\begin{solution}{blr_as_kf}
  We prove the equivalence by induction.
  Note that the base case is satisfied trivially since the priors are identical.

  Assume after $t-1$ steps that $(\vmu_{t-1}, \mSigma_{t-1})$ coincide with the BLR posterior for the first $t-1$ data points.
  We will show that the Kalman filter update equations yield the BLR posterior for the first $t$ data points.

  \paragraph{Covariance update:} \begin{align*}
    \mSigma_t &= \mSigma_{t-1} - \vk_t \transpose{\vx_t} \mSigma_{t-1} \\
    &= \mSigma_{t-1} - \frac{(\mSigma_{t-1} \vx_t) \transpose{(\mSigma_{t-1} \vx_t)}}{\transpose{\vx_t}\mSigma_{t-1}\vx_t + 1} \margintag{using the symmetry of $\mSigma_{t-1}$} \\
    &= \inv{\parentheses*{\inv{\mSigma_{t-1}} + \vx_t \transpose{\vx_t}}} \margintag{using the Sherman-Morrison formula~\eqref{eq:woodbury} with $\inv{\mA} = \mSigma_{t-1}$} \\
    &= \inv{\parentheses*{\transpose{\mX_{1:t-1}} \mX_{1:t-1} + \vx_t \transpose{\vx_t} + \mI}} \margintag{by the inductive hypothesis and using \cref{eq:blr_posterior}} \\
    &= \inv{\parentheses*{\transpose{\mX_{1:t}} \mX_{1:t} + \mI}} \\
    &= \mSigma_t^{\mathrm{BLR}}. \margintag{using \cref{eq:blr_posterior}}
  \end{align*}

  \paragraph{Mean update:} \begin{align*}
    \inv{\mSigma_t} \vmu_t &= \inv{\mSigma_t} \vmu_{t-1} + \inv{\mSigma_t} \vk_t (y_t - \transpose{\vx_t} \vmu_{t-1}) \\
    &= \inv{\mSigma_{t-1}} \vmu_{t-1} + \vx_t \transpose{\vx_t} \vmu_{t-1} + \inv{\mSigma_t} \vk_t (y_t - \transpose{\vx_t} \vmu_{t-1}) \margintag{using ${\inv{\mSigma_t} = \inv{\mSigma_{t-1}} + \vx_t \transpose{\vx_t}}$} \\
    &= \inv{\mSigma_{t-1}} \vmu_{t-1} + \vx_t \transpose{\vx_t} \vmu_{t-1} + \vx_t (y_t - \transpose{\vx_t} \vmu_{t-1}) \margintag{using ${\inv{\mSigma_t} \vk_t = \vx_t}$} \\
    &= \inv{\mSigma_{t-1}} \vmu_{t-1} + \vx_t y_t \margintag{canceling terms} \\
    &= \transpose{\mX_{1:t-1}} \vy_{1:t-1} + \vx_t y_t \margintag{by the inductive hypothesis and using \cref{eq:blr_posterior}} \\
    &= \transpose{\mX_{1:t}} \vy_{1:t} \\
    &= \inv{\mSigma_t} \mSigma_t \transpose{\mX_{1:t}} \vy_{1:t} \\
    &= \inv{\mSigma_t} \mSigma_t^{\mathrm{BLR}} \transpose{\mX_{1:t}} \vy_{1:t} \margintag{using the covariance update above} \\
    &= \inv{\mSigma_t} \vmu_t^{\mathrm{BLR}}. \margintag{using \cref{eq:blr_posterior}}
  \end{align*}
  This completes the induction.
\end{solution}

\begin{solution}{parameter_estimation_with_kf}
  \begin{enumerate}[beginpenalty=10000]
    \item The given parameter estimation problem can be formulated as a Kalman filter in the following way: \begin{align*}
      x_t &= \pi &&\forall t, \\
      y_t &= x_t + \eta_t &&\eta_t \sim \N{0}{\sigma_y^2}.
    \end{align*}
    Thus, in terms of Kalman filters, this yields $f = h = 1, \varepsilon_t = \sigma_x^2 = 0$.
    Using \cref{eq:kalman_gain_1d}, the \emph{Kalman gain} is given by \begin{align*}
      k_{t+1} = \frac{\sigma_t^2}{\sigma_t^2 + \sigma_y^2},
    \end{align*} whereas the \emph{variance of the estimation error} $\sigma_t^2$ satisfies \begin{align*}
      \sigma_{t+1}^2 = \sigma_y^2 k_{t+1} = \frac{\sigma_t^2 \sigma_y^2}{\sigma_t^2 + \sigma_y^2}. \margintag{using \cref{eq:kalman_variance_1d}}
    \end{align*}
    To get the closed form, observe that \begin{align*}
      \frac{1}{\sigma_{t+1}^2} = \frac{1}{\sigma_t^2} + \frac{1}{\sigma_y^2} = \frac{1}{\sigma_{t-1}^2} + \frac{2}{\sigma_y^2} = \cdots = \frac{1}{\sigma_0^2} + \frac{t+1}{\sigma_y^2},
    \end{align*} yielding, \begin{align*}
      \sigma_{t+1}^2 = \frac{\sigma_0^2 \sigma_y^2}{(t+1)\sigma_0^2 + \sigma_y^2} \quad\text{and}\quad k_{t+1} = \frac{\sigma_0^2}{(t+1)\sigma_0^2 + \sigma_y^2}.
    \end{align*}

    \item When $t \to \infty$, we get $k_{t+1} \to 0$ and $\sigma_{t+1}^2 \to 0$, giving \begin{align*}
      \mu_{t+1} = \mu_t + k_{t+1} (y_{t+1} - \mu_t) = \mu_t, \margintag{using \cref{eq:kf_update_1d}}
    \end{align*} thus resulting in a stationary sequence.

    \item Observe that $\sigma_0^2 \to \infty$ implies $k_{t+1} = \frac{1}{t+1}$.
    Therefore, \begin{align*}
      \mu_{t+1} &= \mu_t + \frac{1}{t+1}(y_{t+1} - \mu_t) \margintag{using \cref{eq:kf_update_1d}} \\
      &= \frac{t}{t+1}\mu_t + \frac{y_{t+1}}{t+1} \\
      &= \frac{t}{t+1}\parentheses*{\frac{t-1}{t}\mu_{t-1} + \frac{y_t}{t}} + \frac{y_{t+1}}{t+1} \margintag{using \cref{eq:kf_update_1d}} \\
      &= \frac{t-1}{t+1}\mu_{t-1} + \frac{y_t + y_{t+1}}{t+1} \\
      &\;\;\vdots \\
      &= \frac{y_1 + \cdots + y_{t+1}}{t+1},
    \end{align*} which is simply the \emph{sample mean}.
  \end{enumerate}
\end{solution}

\section*{\nameref{sec:gp}}

\begin{solution}{gaussian_kernel_feature_space}
  \begin{enumerate}[beginpenalty=10000]
    \item We have for any $j \in \NatZ$ and $x, y \in \R$, \begin{align*}
      \frac{1}{\sqrt{j!}} e^{-\frac{x^2}{2}} x^j \cdot \frac{1}{\sqrt{j!}} e^{-\frac{y^2}{2}} y^j &= e^{-\frac{x^2 + y^2}{2}} \frac{(x y)^j}{j!}.
    \end{align*}
    Summing over all $j$, we obtain \begin{align*}
      \transpose{\vphi(x)} \vphi(y) &= \sum_{j=0}^\infty \phi_j(x) \phi_j(y) \\
      &= e^{-\frac{x^2 + y^2}{2}} \sum_{j=0}^\infty \frac{(x y)^j}{j!} \\
      &= e^{-\frac{x^2 + y^2}{2}} e^{x y} \margintag{using the Taylor series expansion for the exponential function} \\
      &= e^{-\frac{(x - y)^2}{2}} \\
      &= k(x, y).
    \end{align*}

    \item As we have seen in \cref{sec:blr:function_space_view}, the effective dimension is $n$.
    The crucial difference of kernelized regression (e.g., Gaussian processes) to linear regression is that the effective dimension \emph{grows} with the sample size, whereas it is fixed for linear regression.
    Models where the effective dimension may depend on the sample size are called \emph{non-parametric} models, and models where the effective dimension is fixed are called \emph{parametric} models.
  \end{enumerate}
\end{solution}

\begin{solution}{kernels_on_the_circle}
  \begin{enumerate}[beginpenalty=10000]
    \item \textbf{Yes}. This is merely a restriction of the standard Gaussian kernel in $\R^2$ onto its subset $\mathbb{S}$.

    \item We have \[
      \begin{multlined}[t]
        \underbrace{
      \begin{pmatrix}
      1 & e^{-\nicefrac{\pi^2}{32}} & e^{-\nicefrac{\pi^2}{8}} & e^{-\nicefrac{\pi^2}{32}} \\
      e^{-\nicefrac{\pi^2}{32}} & 1 & e^{-\nicefrac{\pi^2}{32}} & e^{-\nicefrac{\pi^2}{8}} \\
      e^{-\nicefrac{\pi^2}{8}} & e^{-\nicefrac{\pi^2}{32}} & 1 & e^{-\nicefrac{\pi^2}{32}} \\
      e^{-\nicefrac{\pi^2}{32}} & e^{-\nicefrac{\pi^2}{8}} & e^{-\nicefrac{\pi^2}{32}} & 1
      \end{pmatrix}
      }_{\boldsymbol{K}}
      \begin{pmatrix}
      -1 \\
      1 \\
      -1 \\
      1
      \end{pmatrix} \\
      =
      \underbrace{
      \left(
      1 + e^{-\nicefrac{\pi^2}{8}} - 2 e^{-\nicefrac{\pi^2}{32}}
      \right)}_{\approx -0.18}
      \begin{pmatrix}
      -1 \\
      1 \\
      -1 \\
      1
      \end{pmatrix}
      \end{multlined}
    \]

    \item \textbf{No}. (2) gives an example of a negative eigenvalue in a covariance matrix.
    This implies that $k_{\text{i}}$ cannot be positive semi-definite.

    \item \textbf{Yes}. We have $k_h(\theta, \theta') = \langle \vphi(\theta), \vphi(\theta') \rangle_{\mathbb{R}^{L}}$ with \[
      \vphi(\theta) \defeq \frac{1}{\sqrt{C_{\kappa}}} \begin{pmatrix}
      1 \\
      e^{-\frac{\kappa^2}{4}} \sqrt{2} \cos(\theta) \\
      e^{-\frac{\kappa^2}{4}} \sqrt{2} \sin(\theta) \\
      \vdots \\
      e^{-\frac{\kappa^2}{4} (L-1)^2} \sqrt{2} \cos((L-1) \theta) \\
      e^{-\frac{\kappa^2}{4} (L-1)^2} \sqrt{2} \sin((L-1) \theta)
      \end{pmatrix}.
    \]
  \end{enumerate}
\end{solution}

\begin{solution}{kf_as_gp}
  First we look at the mean, \begin{align*}
    \mu(t) = \E{X_t} = \E{X_{t-1} + \varepsilon_{t-1}} = \E{X_{t-1}} = \mu(t-1).
  \end{align*}
  Knowing that $\mu(0) = 0$, we can derive that $\mu(t) = 0 \; (\forall t)$.

  Now we look at the variance of $X_t$, \begin{align*}
    \Var{X_t} = \Var{X_0 + \sum_{\tau=0}^{t-1} \varepsilon_{\tau}} = \sigma_0^2 + t \sigma_x^2. \margintag{using that the noise is independent}
  \end{align*}
  Finally, we look at the distribution of $\transpose{[f(t) \; f(t')]}$ for arbitrary $t \leq t'$. \begin{align*}
    \begin{bmatrix}
      X_t \\
      X_{t'}
    \end{bmatrix} = \begin{bmatrix}
      X_t \\
      X_t
    \end{bmatrix} + \sum_{\tau = t}^{t' - 1} \begin{bmatrix}
      0 \\
      \varepsilon_{\tau}
    \end{bmatrix}.
  \end{align*}
  Therefore, we get that \begin{align*}
    \begin{bmatrix}
      X_t \\
      X_{t'}
    \end{bmatrix} &\sim \N*{\begin{bmatrix}
      0 \\
      0
    \end{bmatrix}}{\Var{X_t} \begin{bmatrix}
      1 & 1 \\
      1 & 1
    \end{bmatrix} + (t' - t) \begin{bmatrix}
      0 & 0 \\
      0 & \sigma_x^2
    \end{bmatrix}} \\
    &= \N*{\begin{bmatrix}
      0 \\
      0
    \end{bmatrix}}{\begin{bmatrix}
      \Var{X_t} & \Var{X_t} \\
      \Var{X_t} & \Var{X_t} + (t' - t) \sigma_x^2
    \end{bmatrix}}.
  \end{align*}
  We take the kernel $k_{\mathrm{KF}}(t, t')$ to be the covariance between $f(t)$ and $f(t')$, which is $\Var{X_t} = \sigma_0^2 + \sigma_x^2 t$.
  Notice, however, that we assumed $t \leq t'$.
  Thus, overall, the kernel is described by \begin{align*}
    k_{\mathrm{KF}}(t, t') = \sigma_0^2 + \sigma_x^2 \min\{t, t'\}.
  \end{align*}
\end{solution}

\begin{solution}{reproducing_kernel_hilbert_space_properties}
  \begin{enumerate}[beginpenalty=10000]
    \item As $f \in \spH_k(\spX)$ we can express $f(\vx)$ for some $\beta_i \in \R$ and $\vx_i \in \spX$ as \begin{align*}
      f(\vx) &= \sum_{i=1}^n \beta_i k(\vx, \vx_i) \\
      &= \sum_{i=1}^n \beta_i \ip{k(\vx_i,\cdot), k(\vx,\cdot)}_k \\
      &= \ip{\sum_{i=1}^n \beta_i k(\vx_i,\cdot), k(\vx,\cdot)}_k \\
      &= \ip{f(\cdot), k(\vx,\cdot)}_k.
    \end{align*}

    \item By applying Cauchy-Schwarz, \begin{align*}
      |f(\vx) - f(\vy)| &= |\ip{f, k(\vx,\cdot) - k(\vy,\cdot)}_k| \\
      &\leq \norm{f}_k \norm{k(\vx,\cdot) - k(\vy,\cdot)}_k
    \end{align*}
  \end{enumerate}
\end{solution}

\begin{solution}{representer_theorem}
  We denote by $f_{\sA} \defeq \Pi_{\sA} f$ the orthogonal projection of $f$ onto $\mathrm{span}\{k(\vx_1, \cdot), \dots, k(\vx_n, \cdot)\}$ which implies that $\hat{f}_{\sA}$ is a linear combination of $k(\vx_1, \cdot), \dots, k(\vx_n, \cdot)$.

  We then have that $f_{\sA}^\perp \defeq f - f_{\sA}$ is orthogonal to $\mathrm{span}\{k(\vx_1, \cdot), \dots, k(\vx_n, \cdot)\}$.
  Therefore, for any $i \in [n]$, \begin{align*}
    f(\vx) = \ip{f, k(\vx_i, \cdot)}_k = \ip{f_{\sA} + f_{\sA}^\perp, k(\vx_i, \cdot)}_k = \ip{f_{\sA}, k(\vx_i, \cdot)}_k = f_{\sA}(\vx)
  \end{align*} which implies \begin{align*}
    \spL(f(\vx_1), \dots, f(\vx_n)) = \spL(f_{\sA}(\vx_1), \dots, f_{\sA}(\vx_n)).
  \end{align*}

  Denoting the objective of \cref{eq:representer_theorem_obj} by $j(f)$ and noting that ${\norm{f_{\sA}}_k \leq \norm{f}_k}$, we have that $j(f_{\sA}) \leq j(f)$.
  Therefore, if $\hat{f}$ minimizes $j(f)$, then ${\hat{f}_{\sA} \defeq \Pi_{\sA} \hat{f}}$ is also a minimizer since $j(\hat{f}_{\sA}) \leq j(\hat{f})$.
  Thus, we conclude that there exists some $\valphahat \in \R^n$ such that $\hat{f}_{\sA}(\vx) = \sum_{i=1}^n \hat{\alpha}_i k(\vx_i, \vx)$ minimizes $j(f)$.
\end{solution}

\begin{solution}{mle_and_map_of_gps}
  \begin{enumerate}[beginpenalty=10000]
    \item By the representer theorem~\eqref{eq:representer_theorem}, $\hat{f}(\vx) = \transpose{\valphahat} \vk_{\vx,A}$.
    In particular, we have $\vf = \mK \valphahat$ and therefore \begin{align*}
      - \log p(y_{1:n} \mid \vx_{1:n}, f) &= \frac{1}{2 \sigman^2} \norm{\vy - \vf}_2^2 + \const \\
      &= \frac{1}{2 \sigman^2} \norm{\vy - \mK \valphahat}_2^2 + \const.
    \end{align*}
    The regularization term simplifies to \begin{align*}
      \norm{f}_k^2 = \ip{f,f}_k = \transpose{\valphahat} \mK \valphahat = \norm{\valphahat}_{\mK}^2.
    \end{align*}
    Combining, we have that \begin{align*}
      \valphahat = \argmin_{\valpha \in \R^n} \frac{1}{2 \sigman^2} \norm{\vy - \mK \valpha}_2^2 + \frac{1}{2} \norm{\valpha}_{\mK}^2
    \end{align*} as desired.
    It follows by multiplying through with $2 \sigman^2$ that $\lambda = \sigman^2$.

    \item Expanding the objective determined in (1), we are looking for the minimizer of \begin{align*}
      \transpose{\valpha} (\sigman^2 \mK + \mK^2) \valpha - 2 \transpose{\vy} \mK \valpha + \transpose{\vy} \vy.
    \end{align*}
    Differentiating with respect to the coefficients $\valpha$, we obtain the minimizer $\valphahat = \inv{(\mK + \sigman^2 \mI)} \vy$.
    Thus, the prediction at a point $\vxs$ is $\transpose{\vk_{\vxs,A}} \inv{(\mK + \sigman^2 \mI)} \vy$ which coincides with the MAP estimate.
  \end{enumerate}
\end{solution}

\begin{solution}{gradient_of_mll}
  \begin{enumerate}[beginpenalty=10000]
    \item Applying the two hints \cref{eq:gradient_of_mll_hint1,eq:gradient_of_mll_hint2} to \cref{eq:gp_mle_2} yields, \begin{align*}
      \pdv{}{\theta_j} \log p(\vy \mid \mX, \vtheta) &= \frac{1}{2} \transpose{\vy} \inv{\mK_{\vy,\vtheta}} \pdv{\mK_{\vy,\vtheta}}{\theta_j} \inv{\mK_{\vy,\vtheta}} \vy - \frac{1}{2} \tr{\inv{\mK_{\vy,\vtheta}} \pdv{\mK_{\vy,\vtheta}}{\theta_j}}.
    \intertext{We can simplify to}
      &= \frac{1}{2} \tr{\transpose{\vy} \inv{\mK_{\vy,\vtheta}} \pdv{\mK_{\vy,\vtheta}}{\theta_j} \inv{\mK_{\vy,\vtheta}} \vy} - \frac{1}{2} \tr{\inv{\mK_{\vy,\vtheta}} \pdv{\mK_{\vy,\vtheta}}{\theta_j}} \margintag{using that $\transpose{\vy} \inv{\mK_{\vy,\vtheta}} \pdv{\mK_{\vy,\vtheta}}{\theta_j} \inv{\mK_{\vy,\vtheta}} \vy$ is a scalar} \\
      &= \frac{1}{2} \tr{\inv{\mK_{\vy,\vtheta}} \vy \transpose{\vy} \inv{\mK_{\vy,\vtheta}} \pdv{\mK_{\vy,\vtheta}}{\theta_j} - \inv{\mK_{\vy,\vtheta}} \pdv{\mK_{\vy,\vtheta}}{\theta_j}} \margintag{using the cyclic property and linearity of the trace} \\
      &= \frac{1}{2} \tr{\inv{\mK_{\vy,\vtheta}} \vy \transpose{(\inv{\mK_{\vy,\vtheta}} \vy)} \pdv{\mK_{\vy,\vtheta}}{\theta_j} - \inv{\mK_{\vy,\vtheta}} \pdv{\mK_{\vy,\vtheta}}{\theta_j}} \margintag{using that $\inv{\mK_{\vy,\vtheta}}$ is symmetric} \\
      &= \frac{1}{2} \tr{(\alpha \transpose{\alpha} - \inv{\mK_{\vy,\vtheta}}) \pdv{\mK_{\vy,\vtheta}}{\theta_j}}.
    \end{align*}

    \item We denote by $\tilde{\mK}$ the covariance matrix of $\vy$ for the covariance function $\tilde{k}$, so $\mK_{\vy,\vtheta} = \theta_0 \tilde{\mK}$.
    Then, \begin{align*}
      \pdv{}{\theta_0} \log p(\vy \mid \mX, \vtheta) &= \frac{1}{2} \tr{(\theta_0^{-2} \inv{\tilde{\mK}} \vy \transpose{(\inv{\tilde{\mK}} \vy)} - \inv{\theta_0} \inv{\tilde{\mK}}) \tilde{\mK}}. \margintag{using \cref{eq:gradient_mll}}
    \end{align*}
    Simplifying the terms and using linearity of the trace, we obtain that \begin{align*}
      \pdv{}{\theta_0} \log p(\vy \mid \mX, \vtheta) = 0 \quad\iff\quad \theta_0 = \frac{1}{n} \tr{\vy \transpose{\vy} \inv{\tilde{\mK}}}.
    \end{align*}
    If we define $\tilde{\mLambda} \defeq \inv{\tilde{\mK}}$ as the precision matrix associated to $\vy$ for the covariance function $\tilde{k}$, we can express $\opt{\theta_0}$ in closed form as \begin{align}
      \opt{\theta_0} = \frac{1}{n} \sum_{i=1}^n \sum_{j=1}^n \tilde{\mLambda}(i, j) y_i y_j. \label{eq:gradient_of_mll_theta_0}
    \end{align}

    \item We immediately see from \cref{eq:gradient_of_mll_theta_0} that $\opt{\theta_0}$ scales by $s^2$ if $\vy$ is scaled by $s$.
  \end{enumerate}
\end{solution}

\begin{solution}{uniform_convergence_of_fourier_features}
  \begin{enumerate}[beginpenalty=10000]
    \item We have that $s(\cdot) \in [-\sqrt{2}, \sqrt{2}]$, and hence, $s(\vDelta_i)$ is $\sqrt{2}$-sub-Gaussian.\footnote{see \cref{ex:sub_gaussian_examples}}
    It then follows from Hoeffding's inequality \eqref{eq:hoeffdings_inequality} that \begin{align*}
      \Pr{|f(\vDelta_i)| \geq \epsilon} \leq 2 \exp\parentheses*{-\frac{m \epsilon^2}{4}}.
    \end{align*}

    \item We can apply Markov's inequality \eqref{eq:markov_inequality} to obtain \begin{align*}
      \Pr{\norm{\grad f(\opt{\vDelta})}_2 \geq \frac{\epsilon}{2 r}} &= \Pr{\norm{\grad f(\opt{\vDelta})}_2^2 \geq \parentheses*{\frac{\epsilon}{2 r}}^2} \\
      &\leq \parentheses*{\frac{2 r \E*{\norm{\grad f(\opt{\vDelta})}_2}}{\epsilon}}^2.
    \end{align*}
    It remains to bound the expectation.
    We have \begin{align*}
      \E*{\norm{\grad f(\opt{\vDelta})}_2^2} &= \E*{\norm{\grad s(\opt{\vDelta}) - \grad k(\opt{\vDelta})}_2^2} \\
      &= \E*{\norm{\grad s(\opt{\vDelta})}_2^2} - 2 \transpose{\grad k(\opt{\vDelta})} \E*{\grad s(\opt{\vDelta})} + \E*{\norm{\grad k(\opt{\vDelta})}_2^2}. \margintag{using linearity of expectation \eqref{eq:linearity_expectation}}
      \intertext{Note that $\E*{\grad s(\vDelta)} = \grad k(\vDelta)$ using \cref{eq:swap_grad_exp_order} and using that $s(\vDelta)$ is an unbiased estimator of $k(\vDelta)$. Therefore,}
      &= \E*{\norm{\grad s(\opt{\vDelta})}_2^2} - \E*{\norm{\grad k(\opt{\vDelta})}_2^2} \\
      &\leq \E*{\norm{\grad s(\opt{\vDelta})}_2^2} \\
      &\leq \E*[\vomega \sim p]{\norm{\vomega}_2^2} = \sigmap^2. \margintag{using that $s$ is the $\cos$ of a linear function in $\vomega$}
    \end{align*}

    \item Using a union bound \eqref{eq:union_bound} and then the result of (1), \begin{align*}
      \Pr{\bigcup_{i=1}^T |f(\vDelta_i)| \geq \frac{\epsilon}{2}} \leq \sum_{i=1}^T \Pr{|f(\vDelta_i)| \geq \frac{\epsilon}{2}} \leq 2 T \exp\parentheses*{-\frac{m \epsilon^2}{16}}.
    \end{align*}

    \item First, note that by contraposition, \begin{align*}
      \sup_{\vDelta \in \spM_\Delta} |f(\vDelta)| \geq \epsilon \implies \text{$\exists i$. $|f(\vDelta_i)| \geq \frac{\epsilon}{2}$ or $\norm{\grad f(\opt{\vDelta})}_2 \geq \frac{\epsilon}{2 r}$},
    \end{align*} and therefore, \begin{align*}
      \Pr{\sup_{\vDelta \in \spM_\Delta} |f(\vDelta)| \geq \epsilon} &\leq \Pr{\bigcup_{i=1}^T |f(\vDelta_i)| \geq \frac{\epsilon}{2} \cup \norm{\grad f(\opt{\vDelta})}_2 \geq \frac{\epsilon}{2 r}} \\
      &\leq \Pr{\bigcup_{i=1}^T |f(\vDelta_i)| \geq \frac{\epsilon}{2}} + \Pr{\norm{\grad f(\opt{\vDelta})}_2 \geq \frac{\epsilon}{2 r}} \margintag{using a union bound \eqref{eq:union_bound}} \\
      &\leq 2 T \exp\parentheses*{-\frac{m \epsilon^2}{2^4}} + \parentheses*{\frac{2 r \sigmap}{\epsilon}}^2 \margintag{using the results from (2) and (3)} \\
      &\leq \alpha r^{-d} + \beta r^2 \margintag{using $T \leq (4 \; \mathrm{diam}(\spM) / r)^d$}
      \intertext{with $\alpha = 2 (4 \; \mathrm{diam}(\spM))^d \exp\parentheses*{-\frac{m \epsilon^2}{2^4}}$ and $\beta = \parentheses*{\frac{2 \sigmap}{\epsilon}}^2$. Using the hint, we obtain}
      &= 2 \beta^{\frac{d}{d+2}} \alpha^{\frac{2}{d+2}} \\
      &= 2 \parentheses*{2 \parentheses*{\frac{2^3 \sigmap \mathrm{diag}(\spM)}{\epsilon}}^{2 d}}^{\frac{1}{d+2}} \exp\parentheses*{- \frac{m \epsilon^2}{2^3 (d+2)}} \\
      &\leq 2^8 \parentheses*{\frac{\sigmap \mathrm{diag}(\spM)}{\epsilon}}^2 \exp\parentheses*{- \frac{m \epsilon^2}{2^3 (d+2)}} \margintag{using $\frac{\sigmap \mathrm{diam}(\spM)}{\epsilon} \geq 1$}
    \end{align*}

    \item We have \begin{align*}
      \sigmap^2 &= \E[\vomega \sim p]{\transpose{\vomega} \vomega} \\
      &= \int \transpose{\vomega} \vomega \cdot p(\vomega) \,d\vomega \\
      &= \int \transpose{\vomega} \vomega \cdot e^{i \transpose{\vomega} \vzero} p(\vomega) \,d\vomega.
      \intertext{Now observe that $\pdv[order=2]{}{\Delta_j} \int e^{i \transpose{\vomega} \vDelta} p(\vomega) \,d\vomega = - \int \omega_j^2 e^{i \transpose{\vomega} \vDelta} p(\vomega) \,d\vomega$. Thus,}
      &= - \tr{\hes_{\vDelta} \left. \int p(\vomega) e^{i \transpose{\vomega} \vDelta} \,d\vomega \right|_{\vDelta = \vzero}} \\
      &= - \tr{\hes_{\vDelta} k(\vzero)}. \margintag{using that $p$ is the Fourier transform of $k$ \eqref{eq:spectral_density}}
    \end{align*}
    Finally, we have for the Gaussian kernel that \begin{align*}
      \pdv[order=2]{}{\Delta_j} \left. \exp\parentheses*{- \frac{\transpose{\vDelta} \vDelta}{2 h^2}} \right|_{\Delta_j = 0} = - \frac{1}{h^2}.
    \end{align*}
  \end{enumerate}
\end{solution}

\begin{solution}{sor}
  \begin{enumerate}[beginpenalty=10000]
    \item We write $\tilde{\vf} \defeq \transpose{[\vf \; \fs]}$.
    From the definition of SoR \eqref{eq:sor}, we gather \begin{align}
      q_{\mathrm{SoR}}(\tilde{\vf} \mid \vu) = \N*[\tilde{\vf}]{\begin{bmatrix}
        \mK_{\sA\sU} \inv{\mK_{\sU\sU}} \\
        \mK_{\star\sU} \inv{\mK_{\sU\sU}}
      \end{bmatrix} \vu}{\mzero}. \label{eq:sor_helper}
    \end{align}
    We know that $\tilde{\vf}$ and $\vu$ are jointly Gaussian, and hence, the marginal distribution of $\tilde{\vf}$ is also Gaussian.
    We have for the mean and variance that \begin{align*}
      \E{\tilde{\vf}} &= \E{\E{\tilde{\vf}}[\vu]} \margintag{using the tower rule \eqref{eq:tower_rule}} \\
      &= \E[\vu]{\begin{bmatrix}
        \mK_{\sA\sU} \inv{\mK_{\sU\sU}} \\
        \mK_{\star\sU} \inv{\mK_{\sU\sU}}
      \end{bmatrix} \vu} \margintag{using \cref{eq:sor_helper}} \\
      &= \begin{bmatrix}
        \mK_{\sA\sU} \inv{\mK_{\sU\sU}} \\
        \mK_{\star\sU} \inv{\mK_{\sU\sU}}
      \end{bmatrix} \E{\vu} = \vzero \margintag{using linearity of expectation \eqref{eq:linearity_expectation}} \\
      \Var{\tilde{\vf}} &= \E*{\underbrace{\Var{\tilde{\vf}}[\vu]}_{\mzero}} + \Var*{\,\E{\tilde{\vf}}[\vu]} \margintag{using the law of total variance \eqref{eq:lotv}} \\
      &= \Var[\vu]{\begin{bmatrix}
        \mK_{\sA\sU} \inv{\mK_{\sU\sU}} \\
        \mK_{\star\sU} \inv{\mK_{\sU\sU}}
      \end{bmatrix} \vu} \margintag{using \cref{eq:sor_helper}} \\
      &= \begin{bmatrix}
        \mK_{\sA\sU} \inv{\mK_{\sU\sU}} \\
        \mK_{\star\sU} \inv{\mK_{\sU\sU}}
      \end{bmatrix} \underbrace{\Var{\vu}}_{\mK_{\sU\sU}} \transpose{\begin{bmatrix}
        \mK_{\sA\sU} \inv{\mK_{\sU\sU}} \\
        \mK_{\star\sU} \inv{\mK_{\sU\sU}}
      \end{bmatrix}} \margintag{using \cref{eq:linear_map_variance}} \\
      &= \begin{bmatrix}
        \mQ_{\sA\sA} & \mQ_{\sA\star} \\
        \mQ_{\star\sA} & \mQ_{\star\star}
      \end{bmatrix}.
    \end{align*}
    Having determined $q_{\mathrm{SoR}}(\vf, \fs)$, $q_{\mathrm{SoR}}(\fs \mid \vy)$ follows directly using the formulas for finding the Gaussian process predictive posterior \eqref{eq:gp_posterior}.

    \item The given covariance function follows directly from inspecting the derived covariance matrix, $\Var{\vf} = \mQ_{\sA\sA}$.
  \end{enumerate}
\end{solution}

\section*{\nameref{sec:approximate_inference}}

\begin{solution}{logistic_loss_gradient}
  \begin{enumerate}[beginpenalty=10000]
    \item We have \begin{align*}
      \grad_\vw \ell_\mathrm{log}(\transpose{\vw} \vx; y) &= \grad_\vw \log(1 + \exp(-y \transpose{\vw} \vx)) \margintag{using the definition of the logistic loss \eqref{eq:logistic_loss}} \\
      &= \grad_\vw \log \frac{1 + \exp(y \transpose{\vw} \vx)}{\exp(y \transpose{\vw} \vx)} \\
      &= \grad_\vw \log(1 + \exp(y \transpose{\vw} \vx)) - y \transpose{\vw} \vx \\
      &= \frac{1}{1 + \exp(-y \transpose{\vw} \vx)} \cdot \exp(-y \transpose{\vw} \vx) \cdot y \vx - y \vx \margintag{using the chain rule} \\
      &= - y \vx \cdot \parentheses*{1 - \frac{\exp(y \transpose{\vw} \vx)}{1 + \exp(y \transpose{\vw} \vx)}} \\
      &= - y \vx \cdot \sigma(-y \transpose{\vw} \vx). \margintag{using the definition of the logistic function \eqref{eq:logistic_function}}
    \end{align*}

    \item As suggested in the hint, we compute the first derivative of $\sigma$, \begin{align*}
      \sigma'(z) \defeq \odv{}{z} \sigma(z) &= \odv{}{z} \frac{1}{1 + \exp(-z)} \margintag{using the definition of the logistic function \eqref{eq:logistic_function}} \\
      &= \frac{\exp(-z)}{(1 + \exp(-z))^2} \margintag{using the quotient rule} \\
      &= \sigma(z) \cdot (1 - \sigma(z)). \margintag{using the definition of the logistic function \eqref{eq:logistic_function}}
    \end{align*}
    We get for the Hessian of $\ell_\mathrm{log}$, \begin{align*}
      \hes_\vw \ell_\mathrm{log}(\transpose{\vw} \vx; y) &= \mD_\vw \grad_\vw \ell_\mathrm{log}(\transpose{\vw} \vx; y) \margintag{using the definition of a Hessian \eqref{eq:hessian} and symmetry} \\[5pt]
      &= - y \vx \; \mD_\vw \sigma(-y \transpose{\vw} \vx). \margintag{using the gradient of the logistic loss from (1)}
    \intertext{We have $\mD_\vw -y \transpose{\vw} \vx = -y \transpose{\vx}$ and $\mD_z \sigma(z) = \sigma'(z)$, and therefore, using the chain rule of multivariate calculus \eqref{eq:chain_rule},}
      &= -y \vx \cdot \sigma'(-y \transpose{\vw} \vx) \cdot (-y \transpose{\vx}) \\
      &= \vx \transpose{\vx} \cdot \sigma'(-y \transpose{\vw} \vx). \margintag{using $y^2 = 1$}
    \end{align*}
    Finally, recall that \begin{align*}
      \sigma(\transpose{\vw} \vx) &= \Pr{y = +1 \mid \vx, \vw} \quad\text{and} \\
      \sigma(-\transpose{\vw} \vx) &= \Pr{y = -1 \mid \vx, \vw}.
    \end{align*}
    Thus, \begin{align*}
      \sigma'(-y \transpose{\vw} \vx) &= \Pr{Y \neq y \mid \vx, \vw} \cdot (1 - \Pr{Y \neq y \mid \vx, \vw}) \\
      &= \Pr{Y \neq y \mid \vx, \vw} \cdot \Pr{Y = y \mid \vx, \vw} \\
      &= (1 - \Pr{Y = y \mid \vx, \vw}) \cdot \Pr{Y = y \mid \vx, \vw} \\
      &= \sigma'(\transpose{\vw} \vx).
    \end{align*}

    \item By the second-order characterization of convexity (cf. \cref{rmk:second_order_convexity}), a twice differentiable function $f$ is convex if and only if its Hessian is positive semi-definite.
    Writing $c \defeq \sigma'(\transpose{\vw} \vx)$, we have for any $\vdelta \in \R^n$ that \begin{align*}
      \transpose{\vdelta} \hes_\vw \ell_\mathrm{log}(\transpose{\vw} \vx; y) \vdelta &= \transpose{\vdelta} \parentheses*{c \cdot \vx \transpose{\vx}} \vdelta = c (\transpose{\vdelta} \vx)^2 \geq 0, \margintag{using $c \geq 0$ and $(\cdot)^2 \geq 0$}
    \end{align*} and hence, the logistic loss is convex in $\vw$.
  \end{enumerate}
\end{solution}

\begin{solution}{gpc}
  \begin{enumerate}[beginpenalty=10000]
    \item Using the law of total probability \eqref{eq:lotp}, \begin{align*}
      &p(\ys = +1 \mid \vx_{1:n}, y_{1:n}, \vxs) \\
      &= \int p(\ys = +1 \mid \fs) p(\fs \mid \vx_{1:n}, y_{1:n}, \vxs) \,d\fs \\
      &= \int \sigma(\fs) p(\fs \mid \vx_{1:n}, y_{1:n}, \vxs) \,d\fs. \margintag{using $\ys \sim \Bern{\sigma(\fs)}$}
    \end{align*}
    Due to the non-Gaussian likelihood, the integral is analytically intractable.
    However, as the integral is one-dimensional, numerical approximations such as the Gauss-Legendre quadrature can be used.

    \item \begin{enumerate}
      \item According to Bayes' rule \eqref{eq:bayes_rule}, we know that \begin{align*}
        \psi(\vf) &= \log p(\vf \mid \vx_{1:n}, y_{1:n}) \margintag{using \cref{eq:log_posterior}} \\
        &= \log p(y_{1:n} \mid \vf) + \log p(\vf \mid \vx_{1:n}) - \log p(y_{1:n} \mid \vx_{1:n}) \margintag{using Bayes' rule \eqref{eq:bayes_rule}} \\
        &= \log p(y_{1:n} \mid \vf) + \log p(\vf \mid \vx_{1:n}) + \const.
      \intertext{Note that $p(\vf \mid \vx_{1:n}) = \N[\vf]{\vzero}{\mK_{AA}}$. Plugging in this closed-form Gaussian distribution of the GP prior gives}
        &= \log p(y_{1:n} \mid \vf) - \frac{1}{2} \transpose{\vf} \inv{\mK_{\sA\sA}} \vf + \const
      \end{align*}
      Differentiating with respect to $\vf$ yields \begin{align*}
        \grad \psi(\vf) &= \grad \log p(y_{1:n} \mid \vf) - \inv{\mK_{\sA\sA}} \vf \\
        \hes \psi(\vf) &= \hes_\vf \log p(y_{1:n} \mid \vf) - \inv{\mK_{\sA\sA}}.
      \end{align*}
      Hence, $\mLambda = \inv{\mK_{\sA\sA}} + \mW$ where $\mW \defeq \left. -\hes_\vf \log p(y_{1:n} \mid \vf) \right\rvert_{\vf = \vfhat}$.

      It remains to derive $\mW$ for the probit likelihood.
      Using independence of the training examples, \begin{align*}
        \log p(y_{1:n} \mid \vf) = \sum_{i=1}^n \log p(y_i \mid f_i),
      \end{align*} and hence, the Hessian of this expression is diagonal.
      Using the symmetry of $\Phi(z; 0, \sigman^2)$ around zero, we can write \begin{align*}
        \log p(y_i \mid f_i) = \log \Phi(y_i f_i; 0, \sigman^2).
      \end{align*}
      In the following, we write $\mathcal{N}(z) \defeq \N[z]{0}{\sigman^2}$ and $\Phi(z) \defeq \Phi(z; 0, \sigman^2)$ to simplify the notation.
      Differentiating with respect to $f_i$, we obtain \begin{align*}
        \pdv{}{f_i} \log \Phi(y_i f_i) &= \frac{y_i \mathcal{N}(f_i)}{\Phi(y_i f_i)} \margintag{using $\mathcal{N}(y_i f_i) = \mathcal{N}(f_i)$ since $\mathcal{N}$ is an even function and $y_i \in \{\pm 1\}$} \\
        \pdv[order=2]{}{f_i} \log \Phi(y_i f_i) &= - \frac{\mathcal{N}(f_i)^2}{\Phi(y_i f_i)^2} - \frac{y_i f_i \mathcal{N}(f_i)}{\sigman^2 \Phi(y_i f_i)},
      \end{align*} and $\mW = \left. -\diag{\pdv[order=2]{}{f_i} \log \Phi(y_i f_i)}_{i=1}^n \right\rvert_{\vf = \vfhat}$.

      \item Note that $\mLambda'$ is a precision matrix over weights $\vw$ and $\vf = \mX \vw$, so by \cref{eq:linear_map_variance} the corresponding variance over latent values $\vf$ is $\mX \inv{\mLambda'} \transpose{\mX}$.
      The two precision matrices are therefore equivalent if $\inv{\mLambda} = \mX \inv{\mLambda'}\transpose{\mX}$.

      Analogously to \cref{ex:bayesian_logistic_regression,exercise:logistic_loss_gradient}, we have that \begin{align*}
        \mW &= \left. -\hes_\vf \log p(y_{1:n} \mid \vf) \right\rvert_{\vf = \vfhat} \\
        &= \diag[i \in [n]]{\sigma(\hat{f}_i) (1-\sigma(\hat{f}_i))} \\
        &= \diag[i \in [n]]{\pi_i (1-\pi_i)}.
      \end{align*}
      By Woodbury's matrix identity \eqref{eq:general_woodbury}, \begin{align*}
        \inv{\mLambda'} &= \inv{(\mI + \transpose{\mX} \mW \mX)} \\
        &= \mI - \transpose{\mX} \inv{(\inv{\mW} + \mX \transpose{\mX})} \mX.
      \end{align*}
      Thus, \begin{align*}
        \mX \inv{\mLambda'} \transpose{\mX} &= \mX \transpose{\mX} - \mX \transpose{\mX} \inv{(\inv{\mW} + \mX \transpose{\mX})} \mX \transpose{\mX} \\
        &= \mK_{\sA\sA} - \mK_{\sA\sA} \inv{(\inv{\mW} + \mK_{\sA\sA})} \mK_{\sA\sA} \margintag{using that $\mK_{\sA\sA} = \mX \transpose{\mX}$} \\
        &= \inv{(\inv{\mK_{\sA\sA}} + \mW)} \margintag{by the matrix inversion lemma \eqref{eq:matrix_inversion_lemma}} \\
        &= \inv{\mLambda}.
      \end{align*}

      \item Using the formulas for the conditional distribution of a Gaussian \eqref{eq:cond_gaussian}, evaluating a conditional GP at a test point $\vxs$ yields, \begin{subequations}\begin{align}
        \fs \mid \vxs, \vf &\sim \N{\opt{\mu}}{\opt{k}}, \quad\text{where} \\
        \opt{\mu} &\defeq \transpose{\vk_{\vxs,\sA}} \inv{\mK_{\sA\sA}} \vf, \label{eq:gpc_helper_mean} \\
        \opt{k} &\defeq k(\vxs, \vxs) - \transpose{\vk_{\vxs,\sA}} \inv{\mK_{\sA\sA}} \vk_{\vxs,\sA}. \label{eq:gpc_helper_var}
      \end{align}\end{subequations}
      Then, using the tower rule \eqref{eq:tower_rule}, \begin{align}
        \E[q]{\fs}[\vxs, \vx_{1:n}, y_{1:n}] &= \E[\vf \sim q]{\E[\fs \sim p]{\fs}[\vxs, \vf]}[\vx_{1:n}, y_{1:n}] \nonumber \\
        &= \transpose{\vk_{\vxs,\sA}} \inv{\mK_{\sA\sA}} \E[q]{\vf}[\vx_{1:n}, y_{1:n}] \margintag{using \cref{eq:gpc_helper_mean} and linearity of expectation \eqref{eq:linearity_expectation}} \nonumber \\
        &= \transpose{\vk_{\vxs,\sA}} \inv{\mK_{\sA\sA}} \vfhat. \label{eq:gpc_helper_cond_mean}
      \intertext{Observe that any maximum $\vfhat$ of $\psi(\vf)$ needs to satisfy $\grad \psi(\vf) = \vzero$.
      Hence, $\vfhat = \mK_{\sA\sA} (\grad_\vf \log p(y_{1:n} \mid \vf))$, and the expectation simplifies to}
        &= \transpose{\vk_{\vxs,\sA}} (\grad_\vf \log p(y_{1:n} \mid \vf)). \nonumber
      \end{align}

      Using the law of total variance \eqref{eq:lotv}, \begin{align}
        &\Var[q]{\fs}[\vxs, \vx_{1:n}, y_{1:n}] \nonumber \\
        &= \begin{multlined}[t]
          \E[\vf \sim q]{\Var[\fs \sim p]{\fs \mid \vxs, \vx_{1:n}, \vf}}[\vx_{1:n}, y_{1:n}] \\ + \Var[\vf \sim q]{\E[\fs \sim p]{\fs \mid \vxs, \vx_{1:n}, \vf}}[\vx_{1:n}, y_{1:n}]
        \end{multlined} \nonumber \\
        &= \E[q]{\opt{k}}[\vx_{1:n}, y_{1:n}] + \Var[q]{\transpose{\vk_{\vxs,\sA}} \inv{\mK_{\sA\sA}} \vf}[\vx_{1:n}, y_{1:n}] \margintag{using \cref{eq:gpc_helper_var,eq:gpc_helper_mean}} \nonumber \\
        &= \opt{k} + \transpose{\vk_{\vxs,\sA}} \inv{\mK_{\sA\sA}} \Var[q]{\vf}[\vx_{1:n}, y_{1:n}] \inv{\mK_{\sA\sA}} \vk_{\vxs,\sA}. \margintag{using that $\opt{k}$ is independent of $\vf$, \cref{eq:linear_map_variance}, and symmetry of $\inv{\mK_{\sA\sA}}$} \nonumber
      \intertext{Recall from (a) that $\Var[q]{\vf}[\vx_{1:n}, y_{1:n}] = \inv{(\inv{\mK_{\sA\sA} + \mW})}$, so}
        &= \begin{multlined}[t]
          k(\vxs, \vxs) - \transpose{\vk_{\vxs,\sA}} \inv{\mK_{\sA\sA}} \vk_{\vxs,\sA} \\ + \transpose{\vk_{\vxs,\sA}} \inv{\mK_{\sA\sA}} \inv{(\inv{\mK_{\sA\sA} + \mW})} \inv{\mK_{\sA\sA}} \vk_{\vxs,\sA}
        \end{multlined} \margintag{plugging in the expression for $\opt{k}$ \eqref{eq:gpc_helper_var}} \nonumber \\
        &= k(\vxs, \vxs) - \transpose{\vk_{\vxs,\sA}} \inv{(\mK_{\sA\sA} + \inv{\mW})} \vk_{\vxs,\sA}. \margintag{using the matrix inversion lemma \eqref{eq:matrix_inversion_lemma}} \label{eq:gpc_helper_cond_variance}
      \end{align}
      Notice the similarity of \cref{eq:gpc_helper_cond_mean,eq:gpc_helper_cond_variance} to \cref{eq:gpc_helper_mean,eq:gpc_helper_var}.
      The latter is the conditional mean and variance if $\vf$ is known whereas the former is the conditional mean and variance given the noisy observation $y_{1:n}$ of $\vf$.
      The matrix $\mW$ quantifies the noise in the observations.

      \item Recall that \begin{align*}
        p(\ys = +1 \mid \vx_{1:n}, y_{1:n}, \vxs) &\approx \int \sigma(\fs) q(\fs \mid \vx_{1:n}, y_{1:n}, \vxs) \,d\fs \margintag{using the Laplace-approximated latent predictive posterior} \\[5pt]
        &= \E[q]{\sigma(\fs)}. \margintag{using LOTUS \eqref{eq:lotus}}
      \end{align*}
      This quantity can be interpreted as the \emph{averaged prediction} over all latent predictions $\fs$.
      In contrast, $\sigma(\E[q]{\fs})$ can be understood as the \emph{``MAP'' prediction}, which is obtained using the MAP estimate of $\fs$.\footnote{As $q$ is a Gaussian, its mode (i.e., the MAP estimate) and its mean coincide.}
      As $\sigma$ is nonlinear, the two quantities are not identical, and generally the averaged prediction is preferred.
    \end{enumerate}

    \item We have \begin{align*}
      \E*[\varepsilon]{\Ind{f(\vx) + \varepsilon \geq 0}} &= \Pr[\varepsilon]{f(\vx) + \varepsilon \geq 0} \margintag{using $\E{X} = p$ if $X \sim \Bern{p}$} \\
      &= \Pr[\varepsilon]{-\varepsilon \leq f(\vx)} \\
      &= \Pr[\varepsilon]{\varepsilon \leq f(\vx)} \margintag{using that the distribution of $\varepsilon$ is symmetric around $0$} \\
      &= \Phi(f(\vx); 0, \sigman^2).
    \end{align*}
  \end{enumerate}
\end{solution}

\begin{solution}{jensen}
  \begin{enumerate}[beginpenalty=10000]
    \item Recall that as $f$ is convex, \begin{align*}
      \forall \vx_1, \vx_2, \forall \lambda \in [0,1] :\quad f(\lambda \vx_1 + (1-\lambda) \vx_2) \leq \lambda f(\vx_1) + (1-\lambda) f(\vx_2).
    \end{align*}
    We prove the statement by induction on $k$.
    The base case, $k=2$, follows trivially from the convexity of $f$.
    For the induction step, suppose that the statement holds for some fixed $k \geq 2$ and assume w.l.o.g. that $\theta_{k+1} \in (0,1)$. We then have, \begin{align*}
      \sum_{i=1}^{k+1} \theta_i f(\vx_i) &= (1-\theta_{k+1}) \parentheses*{\sum_{i=1}^k \frac{\theta_i}{1 - \theta_{k+1}} f(\vx_i)} + \theta_{k+1} f(\vx_{k+1}) \\
      &\geq (1-\theta_{k+1}) \cdot f\left(\sum_{i=1}^k \frac{\theta_i}{1 - \theta_{k+1}} \vx_i\right) + \theta_{k+1} f(\vx_{k+1}) \margintag{using the induction hypothesis} \\
      &\geq f\left(\sum_{i=1}^{k+1} \theta_i \vx_i\right). \margintag{using the convexity of $f$}
    \end{align*}

    \item Noting that $\log_2$ is concave, we have by Jensen's inequality, \begin{align*}
      \H{p} &= \E[x \sim p]{\log_2\parentheses*{\frac{1}{p(x)}}} \margintag{by definition of entropy \eqref{eq:entropy_discrete}} \\
      &\leq \log_2 \E[x \sim p]{\frac{1}{p(x)}} \margintag{by Jensen's inequality \eqref{eq:jensen}} \\
      &= \log_2 n.
    \end{align*}
  \end{enumerate}
\end{solution}

\begin{solution}{bce_loss}
  If $y = 1$ then \begin{align*}
    \ell_{\mathrm{bce}}(\hat{y}; y) = - \log \hat{y} = \log(1 + e^{-\hat{f}}) = \ell_{\mathrm{log}}(\hat{F}; y).
  \end{align*}
  If $y = -1$ then \begin{align*}
    \ell_{\mathrm{bce}}(\hat{y}; y) = - \log(1 - \hat{y}) = \log(1 + e^{\hat{f}}) = \ell_{\mathrm{log}}(\hat{f}; y).
  \end{align*}
  Here the second equality follows from the simple algebraic fact \begin{align*}
    1 - \frac{1}{1 + e^{-z}} = 1 - \frac{e^z}{e^z + 1} = \frac{1}{1 + e^z}. \margintag{multiplying by $\frac{e^z}{e^z}$}
  \end{align*}
\end{solution}

\begin{solution}{gibbs_ineq}
  \begin{enumerate}[beginpenalty=10000]
    \item Let $p$ and $q$ be two (continuous) distributions.
    The KL-divergence between $p$ and $q$ is \begin{align*}
      \KL{p}{q} &= \E[\vx \sim p]{\log \frac{p(\vx)}{q(\vx)}} \margintag{using the definition of KL-divergence \eqref{eq:kl}} \\
      &= \E[\vx \sim p]{\S{\frac{q(\vx)}{p(\vx)}}}.
    \intertext{Note that the surprise $\S{u} = - \log u$ is a convex function in $u$, and hence,}
      &\geq \S{\E[\vx \sim p]{\frac{q(\vx)}{p(\vx)}}} \margintag{using Jensen's inequality \eqref{eq:jensen}} \\
      &= \S{\int q(\vx) \,d\vx} \\
      &= \S{1} = 0. \margintag{a probability density integrates to $1$}
    \end{align*}

    \item We observe from the derivation of (1) that $\KL{p}{q} = 0$ iff equality holds for Jensen's inequality.
    Now, if $p$ and $q$ are discrete with final and identical support, we can follow from the hint that Jensen's inequality degenerates to an equality iff $p$ and $q$ are point wise identical.
  \end{enumerate}
\end{solution}

\begin{solution}{maximum_entropy_principle}
  \begin{enumerate}[beginpenalty=10000]
    \item We have \begin{align*}
      \crH{f}{g} &= - \int_\R f(x) \log g(x) \,d x \margintag{using the definition of cross-entropy \eqref{eq:cross_entropy}} \\
      &= - \int_\R f(x) \cdot \parentheses*{\log\parentheses*{\frac{1}{\sqrt{2 \pi \sigma^2}}} - \frac{(x - \mu)^2}{2 \sigma^2}} \,d x \margintag{using that $g(x) = \N[x]{\mu}{\sigma^2}$} \\
      &= - \log\parentheses*{\frac{1}{\sqrt{2 \pi \sigma^2}}} \underbrace{\int_\R f(x) \,d x}_{1} + \frac{1}{2 \sigma^2} \int_\R f(x) (x - \mu)^2 \,d x \\
      &= \log(\sigma\sqrt{2\pi}) + \frac{1}{2 \sigma^2} \underbrace{\E[x \sim f]{(x - \mu)^2}}_{\sigma^2} \\
      &= \log(\sigma\sqrt{2\pi}) + \frac{1}{2} \\
      &= \H{g}. \margintag{using the entropy of Gaussians \eqref{eq:entropy_gaussian_univ}}
    \end{align*}

    \item We have shown that \begin{align*}
      \H{g} - \H{f} = \KL{f}{g} \geq 0,
    \end{align*} and hence, $\H{g} \geq \H{f}$.
    That is, for a fixed mean $\mu$ and variance $\sigma^2$, the distribution that has maximum entropy among all distributions on $\R$ is the normal distribution.
  \end{enumerate}
\end{solution}

\begin{solution}{mep_and_posteriors}
  \begin{enumerate}[beginpenalty=10000]
    \item This sample solution follows the works of \cite{caticha2006updating,caticha2021entropy}.
    Writing down the Lagrangian with dual variables $\lambda_0$ and $\lambda_1(\vy)$ for the normalization and data constraints yields \begin{align*}
      L(q, \lambda_0, \lambda_1) &= \begin{multlined}[t]
        \int q(\vx,\vy) \log\frac{q(\vx,\vy)}{p(\vx,\vy)} \,d\vx \,d\vy \margintag{the objective, using \eqref{eq:kl}} \\ + \lambda_0 \parentheses*{1 - \int q(\vx,\vy) \,d\vx \,d\vy} \margintag{the normalization constraint} \\ + \int \lambda_1(\vy) \brackets*{\delta_{\vyp}(\vy) - \int q(\vx,\vy) \,d\vx} \,d\vy \margintag{the data constraint}
      \end{multlined} \\
      &= \int q(\vx,\vy) \brackets*{\log\frac{q(\vx,\vy)}{p(\vx,\vy)} - \lambda_0 - \lambda_1(\vy)} \,d\vx \,d\vy + \const.
    \end{align*}
    Note that $L(q, \lambda_0, \lambda_1) = \KL{q_{\rX,\rY}}{p_{\rX,\rY}}$ if the constraints are satisfied.
    Thus, we simply need to solve the (dual) optimization problem \begin{align*}
      \min_{q(\cdot,\cdot) \geq 0} \max_{\lambda_0, \lambda_1(\cdot) \in \R} L(q, \lambda_0, \lambda_1).
    \end{align*}
    We have \begin{align*}
      \pdv{L(q, \lambda_0, \lambda_1)}{{q(\vx,\vy)}} = \log\frac{q(\vx,\vy)}{p(\vx,\vy)} - \lambda_0 - \lambda_1(\vy) + 1. \margintag{using the product rule of differentiation}
    \end{align*}
    Setting the partial derivatives to zero, we obtain \begin{align*}
      q(\vx,\vy) = \exp(\lambda_0 + \lambda_1(\vy) - 1) p(\vx,\vy) = \frac{1}{Z} \exp(\lambda_1(\vy)) p(\vx,\vy)
    \end{align*} where $Z \defeq \exp(\lambda_0 - 1)$ denotes the normalizing constant.
    We can determine $\lambda_1(\vy)$ from the data constraint: \begin{align*}
      \int q(\vx,\vy) \,d\vx &= \frac{1}{Z} \exp(\lambda_1(\vy)) \int p(\vx,\vy) \,d\vx \\
      &= \frac{1}{Z} \exp(\lambda_1(\vy)) \cdot p(\vy) \margintag{using the sum rule \eqref{eq:sum_rule}} \\
      &\overset{!}{=} \delta_{\vyp}(\vy).
    \end{align*}
    It follows that $q(\vx,\vy) = \delta_{\vyp}(\vy) \cdot \frac{p(\vx,\vy)}{p(\vy)} = \delta_{\vyp}(\vy) \cdot p(\vx \mid \vy)$.\margintag{using the definition of conditional distributions \eqref{eq:cond_distr}}

    \item From the sum rule \eqref{eq:sum_rule}, we obtain \begin{align*}
      q(\vx) = \int q(\vx,\vy) \,d\vy = \int \delta_{\vyp}(\vy) \cdot p(\vx \mid \vy) \,d\vy = p(\vx \mid \vyp).
    \end{align*}
  \end{enumerate}
\end{solution}

\begin{solution}{kl_div_of_gaussians}
  We can rewrite the KL-divergence as \begin{align*}
    \KL{p}{q} &= \E[\vx \sim p]{\log p(\vx) - \log q(\vx)} \margintag{using the definition of KL-divergence \eqref{eq:kl}} \\
    &= \begin{multlined}[t]
      \E*[\vx \sim p]{\left[\frac{1}{2} \log \frac{\det{\mSigma_q}}{\det{\mSigma_p}} - \frac{1}{2} \transpose{(\vx - \vmu_p)} \inv{\mSigma_p} (\vx - \vmu_p) \right. \\ + \left. \frac{1}{2} \transpose{(\vx - \vmu_q)} \inv{\mSigma_q} (\vx - \vmu_q)\right]}
    \end{multlined} \margintag{using the Gaussian PDF \eqref{eq:normal}} \\
    &= \begin{multlined}[t]
      \frac{1}{2} \log \frac{\det{\mSigma_q}}{\det{\mSigma_p}} - \frac{1}{2} \E[\vx \sim p]{\transpose{(\vx - \vmu_p)} \inv{\mSigma_p} (\vx - \vmu_p)} \\ + \frac{1}{2} \E[\vx \sim p]{\transpose{(\vx - \vmu_q)} \inv{\mSigma_q} (\vx - \vmu_q)}
    \end{multlined} \margintag{using linearity of expectation \eqref{eq:linearity_expectation}}
  \end{align*}
  As $\transpose{(\vx - \vmu_p)} \inv{\mSigma_p} (\vx - \vmu_p) \in \R$, we can rewrite the second term as \begin{align*}
    &\frac{1}{2} \E[\vx \sim p]{\tr{\transpose{(\vx - \vmu_p)} \inv{\mSigma_p} (\vx - \vmu_p)}} \\
    &= \frac{1}{2} \E[\vx \sim p]{\tr{(\vx - \vmu_p) \transpose{(\vx - \vmu_p)} \inv{\mSigma_p}}} \margintag{using the cyclic property of the trace} \\
    &= \frac{1}{2} \tr{\E[\vx \sim p]{(\vx - \vmu_p) \transpose{(\vx - \vmu_p)}} \inv{\mSigma_p}} \margintag{using linearity of the trace and linearity of expectation \eqref{eq:linearity_expectation}} \\
    &= \frac{1}{2} \tr{\mSigma_p \inv{\mSigma_p}} \margintag{using the definition of the covariance matrix \eqref{eq:variance}} \\
    &= \frac{1}{2} \tr{\mI} = \frac{d}{2}.
  \end{align*}
  For the third term, we use the hint \eqref{eq:kl_div_of_gaussians_hint} to obtain \begin{align*}
    &\frac{1}{2} \E[\vx \sim p]{\transpose{(\vx - \vmu_q)} \inv{\mSigma_q} (\vx - \vmu_q)} \\
    &= \frac{1}{2} \parentheses*{\transpose{(\vmu_p - \vmu_q)} \inv{\mSigma_q} (\vmu_p - \vmu_q) + \tr{\inv{\mSigma_q} \mSigma_p}}.
  \end{align*}
  Putting all terms together we get \begin{align*}
    \KL{p}{q} = \begin{multlined}[t]
      \frac{1}{2} \left(\log \frac{\det{\mSigma_q}}{\det{\mSigma_p}} - d + \transpose{(\vmu_p - \vmu_q)} \inv{\mSigma_q} (\vmu_p - \vmu_q) \right. \\ \left. + \tr{\inv{\mSigma_q} \mSigma_p}\right).
    \end{multlined}
  \end{align*}
\end{solution}

\begin{solution}{forward_vs_reverse_kl}
  \begin{enumerate}[beginpenalty=10000]
    \item Let $p$ and $q$ be discrete distributions.
    The derivation is analogous if $p$ and $q$ are taken to be continuous.
    First, we write the KL-divergence between $p$ and $q$ as \begin{align*}
      &\KL{p}{q} \\
      &= \sum_x \sum_y p(x, y) \log_2 \frac{p(x, y)}{q(x) q(y)} \margintag{using the definition of KL-divergence \eqref{eq:kl}} \\
      &= \begin{multlined}[t]
        \sum_x \sum_y p(x, y) \log_2 p(x, y) - \sum_x \sum_y p(x, y) \log_2 q(x) \\ - \sum_x \sum_y p(x, y) \log_2 q(y)
      \end{multlined} \\
      &= \sum_x \sum_y p(x, y) \log_2 p(x, y) - \sum_x p(x) \log_2 q(x) - \sum_y p(y) \log_2 q(y) \margintag{using the sum rule \eqref{eq:sum_rule}} \\
      &= - \H{p(x, y)} + \crH{p(x)}{q(x)} + \crH{p(y)}{q(y)} \margintag{using the definitions of entropy \eqref{eq:entropy} and cross-entropy \eqref{eq:cross_entropy}} \\[5pt]
      &= \begin{multlined}[t]
        - \H{p(x, y)} + \H{p(x)} + \H{p(y)} + \KL{p(x)}{q(x)} \\ + \KL{p(y)}{q(y)}
      \end{multlined} \margintag{using \cref{eq:cross_entropy_decomp}} \\
      &= \KL{p(x)}{q(x)} + \KL{p(y)}{q(y)} + \const.
    \end{align*}
    Hence, to minimize $\KL{p}{q}$ with respect to the variational distributions $q(x)$ and $q(y)$ we should set $\KL{p(x)}{q(x)} = 0$ and $\KL{p(y)}{q(y)} = 0$, respectively.
    This is obtained when \begin{align*}
      q(x) = p(x) \qquad\text{and}\qquad q(y) = p(y).
    \end{align*}

    \item The reverse KL-divergence $\KL{q}{p}$ on the finite domain $x, y \in \{1, 2, 3, 4\}$ is defined as \begin{align*}
      \KL{q}{p} = \sum_x \sum_y q(x) q(y) \log_2 \frac{q(x) q(y)}{p(x, y)}.
    \end{align*}
    We can easily observe from the above formula that the support of $q$ must be a subset of the support of $p$.
    In other words, if $q(x, y)$ is positive outside the support of $p$ (i.e., when $p(x, y) = 0$) then $\KL{q}{p} = \infty$.
    Hence, the reverse KL-divergence has an infinite value except when the support of $q$ is either $\{1,2\} \times \{1,2\}$ or $\{(3,3)\}$ or $\{(4,4)\}$.
    Thus, it has three local minima.

    For the first case, the minimum is achieved when $q(x) = q(y) = (\frac{1}{2}, \frac{1}{2}, 0, 0)$.
    The corresponding KL-divergence is $\KL{q}{p} = \log_2 2 = 1$.
    For the second case and the third case, $q(x) = q(y) = (0, 0, 1, 0)$ and $q(x) = q(y) = (0, 0, 0, 1)$, respectively.
    The KL-divergence in both cases is $\KL{q}{p} = \log_2 4 = 2$.

    \item Let us compute $p(x = 4)$ and $p(y = 1)$: \begin{align*}
      p(x = 4) &= \sum_y p(x = 4, y) = \frac{1}{4}, \\
      p(y = 1) &= \sum_x p(x, y = 1) = \frac{1}{4}.
    \end{align*}
    Hence, $q(x = 4, y = 1) = p(x = 4) p(y = 1) = \frac{1}{16}$, however, $p(x = 4, y = 1) = 0$.
    We therefore have for the reverse KL-divergence that $\KL{q}{p} = \infty$.
  \end{enumerate}
\end{solution}

\begin{solution}{gaussian_vi_vs_laplace}
  \begin{enumerate}[beginpenalty=10000]
    \item Recall from \cref{sec:vi:kl:forward} that $\opt{q}$ matches the first and second \emph{moments} of $p$.
    In contrast, the Laplace approximation matches the \emph{mode} of $p$ and the second derivative of $- \log p$.
    In general, the mean is different from the mode, and so is the second moment from the second derivative.

    \item We have \begin{align*}
      &\argmin_{q \in \spQ} \KL{q}{p(\cdot \mid \spD)} \\
      &= \argmax_{q \in \spQ} L(q, p; \spD) \margintag{using \cref{eq:elbo_reverse_kl_relationship}} \\
      &= \argmax_{q \in \spQ} \E[\vtheta \sim q]{\log p(y_{1:n}, \vtheta \mid \vx_{1:n})} + \H{q} \margintag{using \cref{eq:elbo}} \\
      &= \argmax_{q \in \spQ} \E[\vtheta \sim q]{\log p(y_{1:n}, \vtheta \mid \vx_{1:n})} + \frac{n}{2} \log(2 \pi e) + \frac{1}{2} \log\det{\mSigma}. \margintag{using \cref{eq:entropy_gaussian}}
    \end{align*}
    Differentiating with respect to $\vmu$ and $\mSigma$, we have that $\tilde{q}$ must satisfy \begin{align*}
      \vzero &= \grad_{\vmu} \E[\vtheta \sim \tilde{q}]{\log p(y_{1:n}, \vtheta \mid \vx_{1:n})} \\
      \inv{\mSigma} &= -2 \grad_{\mSigma} \E[\vtheta \sim \tilde{q}]{\log p(y_{1:n}, \vtheta \mid \vx_{1:n})}. \margintag{using the first hint}
    \end{align*}
    The result follows by \cref{eq:gradient_of_gaussian_parameters}.
  \end{enumerate}
\end{solution}

\begin{solution}{gradient_of_reverse_kl}
  To simplify the notation, we write $\mSigma \defeq \diag{\sigma_1^2, \dots, \sigma_d^2}$.
  The reverse KL-divergence can be expressed as \begin{align*}
    \KL{q_\vlambda}{p(\cdot)} &= \frac{1}{2} \parentheses*{\tr{\sigmap^{-2} \mSigma} + \sigmap^{-2} \transpose{\vmu} \vmu - d + \log \frac{(\sigmap^2)^d}{\det{\mSigma}}} \margintag{using the expression for the KL-divergence of Gaussians \eqref{eq:kl_gaussian}} \\
    &= \frac{1}{2} \parentheses*{\sigmap^{-2} \sum_{i=1}^d \sigma_i^2 + \sigmap^{-2} \transpose{\vmu} \vmu - d + d \log \sigmap^2 - \sum_{i=1}^d \log \sigma_i^2}.
  \end{align*}
  It follows immediately that $\grad_\vmu \KL{q_\vlambda}{p(\cdot)} = \sigmap^{-2} \vmu$.
  Moreover, \begin{align*}
    \pdv{}{\sigma_i} \KL{q_\vlambda}{p(\cdot)} = \frac{1}{2} \bigg(\sigmap^{-2} \underbrace{\pdv{}{\sigma_i} \sigma_i^2}_{2 \sigma_i} - \underbrace{\pdv{}{\sigma_i} \log \sigma_i^2}_{\nicefrac{2}{\sigma_i}}\bigg) = \frac{\sigma_i}{\sigmap^2} - \frac{1}{\sigma_i}.
  \end{align*}
\end{solution}

\begin{solution}{reparameterizable_distributions}
  \begin{enumerate}[beginpenalty=10000]
    \item Let $Y \sim \Unif{[0,1]}$.
    Then, using the two hints, \begin{align*}
      (b - a) Y + a \sim \Unif{[a,b]}.
    \end{align*}

    \item Let $Z_1 \sim \N{\mu}{\sigma^2}$ and $Z_2 \sim \N{0}{1}$.
    We have, $X = e^{Z_1}$.
    Recall from \cref{eq:gaussian_affine_transformation} that $Z_1$ can equivalently be expressed in terms of $Z_2$ as $Z_1 = \sigma Z_2 + \mu$.
    This yields, \begin{align*}
      X = e^{Z_1} = e^{\sigma Z_2 + \mu}.
    \end{align*}

    \item Denote by $F$ the CDF of $\Cauchy{0}{1}$.
    Observe that $F$ is invertible with inverse $\inv{F}(y) = \tan(\pi (y - \frac{1}{2}))$.
    Let $Y \sim \Unif{[0,1]}$ and write $X = \inv{F}(Y)$.
    Then, \begin{align*}
        P_X(x) &= \Pr{X \leq x} \\
        &= \Pr{\inv{F}(Y) \leq x} \\
        &= \Pr{Y \leq F(x)} \\
        &= F(x).
    \end{align*}
    This reparameterization works for any distribution with invertible CDF (not just Cauchy) and is known as the \midx{universality of the uniform} (cf. \cref{sec:inverse_transform_sampling}).
    The universality of the uniform is commonly used in pseudo-random number generators as it allows ``lifting'' samples from a uniform distribution to countless other well-known distributions.

    \item The derivative of $\mathrm{ReLU}(z)$ is $\Ind{z > 0}$.
    Applying the reparameterization trick gives \begin{align*}
      \odv{}{\mu} \E*[x \sim \N{\mu}{1}]{\mathrm{ReLU}(w x)} &= \E*[\varepsilon \sim \N{0}{1}]{\odv{}{\mu} \mathrm{ReLU}(w (\mu + \varepsilon))} \\
      &= w \E*[\varepsilon]{\Ind{w(\mu + \varepsilon) > 0}} \margintag{using the chain rule} \\
      &= w \E*[\varepsilon]{\Ind{\mu + \varepsilon > 0}} \\
      &= w \Pr[\varepsilon]{\mu + \varepsilon > 0} \\
      &= w \Pr[\varepsilon]{\varepsilon < \mu} \\
      &= w \Pr[\varepsilon]{\varepsilon \leq \mu} \\
      &= w \Phi(\mu).
    \end{align*}
  \end{enumerate}
\end{solution}

\section*{\nameref{sec:approximate_inference:mcmc}}

\begin{solution}{mc_update}
  We have \begin{align*}
    q_{t+1}(x') = \Pr{X_{t+1} = x'} = \sum_x \underbrace{\Pr{X_t = x}}_{q_t(x)} \underbrace{\Pr{X_{t+1} = x' \mid X_t = x}}_{p(x' \mid x)}. \margintag{using the sum rule \eqref{eq:sum_rule}}
  \end{align*}
  Noting that $p(x' \mid x) = \mP(x, x')$, we conclude $\vq_{t+1} = \vq_t \mP$.
\end{solution}

\begin{solution}{mc_multi_step_transitions}
  It follows directly from the definition of matrix multiplication that \begin{align*}
    \mP^k(x, x') &= \sum_{x_1, \dots, x_{k-1}} \mP(x, x_1) \cdot \mP(x_1, x_2) \cdots \mP(x_{k-1}, x') \\
    &= \sum_{x_1, \dots, x_{k-1}} \Pr{X_1 = x_1 \mid X_0 = x} \cdots \Pr{X_k = x' \mid X_{k-1} = x_{k-1}} \margintag{using the definition of the transition matrix \eqref{eq:transition_matrix}} \\
    &= \sum_{x_1, \dots, x_{k-1}} \Pr{X_1 = x_1, \cdots, X_{k-1} = x_{k-1}, X_k = x' \mid X_0 = x} \margintag{using the product rule \eqref{eq:product_rule}} \\
    &= \Pr{X_k = x' \mid X_0 = x}. \margintag{using the sum rule \eqref{eq:sum_rule}}
  \end{align*}
\end{solution}

\begin{solution}{finding_stationary_distributions}
  We consider the transition matrix \begin{align*}
    \mP = \begin{bmatrix}
      0.60 & 0.30 & 0.10 \\
      0.50 & 0.25 & 0.25 \\
      0.20 & 0.40 & 0.40 \\
    \end{bmatrix}.
  \end{align*}
  We note that the entries of $\mP$ are all different from $0$, thus the Markov chain corresponding to this transition matrix is ergodic.\footnote{All elements of the transition matrix being strictly greater than $0$ is a sufficient, but not necessary, condition for ergodicity.}
  Thus, there exists a unique stationary distribution $\pi$ to which the Markov chain converges irrespectively of the distribution over initial states $q_0$.

  We know that $\transpose{\mP}\vpi = \vpi$ (where we write $\vpi$ as a column vector), therefore, to find the stationary distribution $\pi$, we need to find the normalized eigenvector associated with eigenvalue $1$ of the matrix $\transpose{\mP}$.
  That is, we want to solve $(\transpose{\mP} - \mI) \vpi = \vzero$ for $\vpi$.
  We obtain the linear system of equations, \begin{align*}
    -0.40 \pi_1 + 0.50 \pi_2 + 0.20 \pi_3 &= 0 \\
    0.30 \pi_1 - 0.75 \pi_2 + 0.40 \pi_3 &= 0 \\
    0.10 \pi_1 + 0.25 \pi_2 - 0.60 \pi_3 &= 0.
  \end{align*}
  Note that the left hand side of equation $i$ corresponds to the probability of entering state $i$ at stationarity minus $\pi_i$.
  Quite intuitively, this difference should be $0$, that is, after one iteration the random walk is at state $i$ with the same probability as before the iteration.

  Solving this system of equations, for example, using the Gaussian elimination algorithm, we obtain the normalized eigenvector \begin{align*}
    \vpi = \frac{1}{72} \begin{bmatrix}
      35 \\ 22 \\ 15
    \end{bmatrix}.
  \end{align*}
  Thus, we conclude that in the long run, the percentage of news days that will be classified as ``good'' is $\nicefrac{35}{72}$.
\end{solution}

\begin{solution}{metropolis_hastings}
  Observe that the described proposal distribution is symmetric.
  Therefore, the acceptance probability simplifies to \begin{align*}
    \alpha(x' \mid x) = \min\braces*{1, \frac{p(x')}{p(x)}}.
  \end{align*}
  If we denote the number of $1$s in a bit string by $w(x)$, we have the requirement that $p(x) \propto w(x)$.
  Therefore, the acceptance probability becomes \begin{align*}
    \alpha(x' \mid x) = \begin{cases}
      \min\braces*{1, \frac{w(x')}{w(x)}} & \text{if $w(x) \neq 0$} \\
      1 & \text{otherwise}.
    \end{cases}
  \end{align*}
\end{solution}

\begin{solution}{gibbs_sampling}
  \begin{enumerate}[beginpenalty=10000]
    \item We have to compute the conditional distributions.
    Notice that for $x \in \{0, \dots, n\}$ and $y \in [0,1]$, \begin{align*}
      p(x, y) = {n \choose x} y^x (1-y)^{n-x} \cdot y^{\alpha-1}(1-y)^{\beta - 1} = \Bin[x]{n}{y} \cdot C_y,
    \end{align*}
    where $\Bin{n}{y}$ is the PMF of the binomial distribution \eqref{eq:binomial_distr} with $n$ trials and success probability $y$, and $C_y$ is a constant depending on $y$.
    It is clear that \begin{align*}
      p(x \mid y) &= \frac{p(x,y)}{p(y)} \margintag{using the definition of conditional probability \eqref{eq:cond_prob}} \\
      &= \frac{\Bin[x]{n}{y} \cdot C_y}{p(y)} \\
      &= \Bin[x]{n}{y}. \margintag{using that $p(x \mid y)$ is a probability distribution over $x$ and $\Bin[x]{n}{y}$ already sums to $1$, so $C_y = p(y)$}
    \end{align*}
    So in short, sampling from $p(x \mid y)$ is equivalent to sampling from a binomial distribution, which amounts to $n$ times throwing a coin with bias $y$, and outputting the number of heads.

    For the other conditional distribution, recall the PMF of the beta distribution with parameters $\alpha, \beta$, \begin{align*}
      \Beta[y]{\alpha}{\beta} = C \cdot y^{\alpha-1} (1-y)^{\beta-1}
    \end{align*} where $C$ is some constant depending on $\alpha$ and $\beta$ only.
    We then have \begin{align*}
      p(x, y) = \Beta[y]{x + \alpha}{n - x + \beta} \cdot C_x,
    \end{align*} where $C_x$ is some constant depending on $x,\alpha,\beta$.
    This shows (analogously to above), that \begin{align*}
      p(y \mid x) = \Beta[y]{x + \alpha}{n - x + \beta}.
    \end{align*}
    So for sampling $y$ given $x$, one can sample from the beta distribution.
    There are several methods for sampling from a beta distribution, and we refer the reader to the corresponding \href{https://en.wikipedia.org/wiki/Beta_distribution}{Wikipedia page}.

    \item We first derive the posterior distribution of $\mu$.
    We have \begin{align*}
      \log p(\mu \mid \lambda, x_{1:n}) &= \log p(\mu) + \log p(x_{1:n} \mid \mu, \lambda) + \const \\
      &= - \frac{\lambda_0}{2} (\mu - \mu_0)^2 - \frac{\lambda}{2} \sum_{i=1}^n (x_i - \mu)^2 + \const \\
      &= - \frac{1}{2} (\underbrace{\lambda_0 + n \lambda}_{l_\lambda}) \mu^2 + \Big(\underbrace{\lambda_0 \mu_0 + \lambda \sum_{i=1}^n x_i}_{m_\lambda l_\lambda}\Big) \mu + \const. \margintag{by expanding the squares}
    \end{align*}
    It follows that $l_\lambda = \lambda_0 + n \lambda, m_\lambda = (\lambda_0 \mu_0 + \lambda \sum_{i=1}^n x_i) / l_\lambda$, and $\mu \mid \lambda, x_{1:n} \sim \N{\mu_\lambda}{\inv{l_\lambda}}$.
    That is, the posterior precision is the sum of prior precision and the precisions of each observation, and the posterior mean is a weighted average of prior mean and observations (where the weights are the precisions).

    For the posterior distribution of $\lambda$, we have \begin{align*}
      p(\lambda \mid \mu, x_{1:n}) &\propto p(\lambda) \cdot p(x_{1:n} \mid \mu, \lambda) \\
      &= \lambda^{\alpha-1} e^{- \beta \lambda} \cdot \lambda^{\frac{n}{2}} e^{- \frac{\lambda}{2} \sum_{i=1}^n (x_i - \mu)^2} \\
      &= \lambda^{\alpha + \frac{n}{2} - 1} e^{-\lambda \parentheses*{\beta + \frac{1}{2} \sum_{i=1}^n (x_i - \mu)^2}} \\
      &= \GammaDistr[\lambda]{a_\mu}{b_\mu}
    \end{align*} where $a_\mu = \alpha + \frac{n}{2}$ and $b_\mu = \beta + \frac{1}{2} \sum_{i=1}^n (x_i - \mu)^2$.

    \item We have \begin{align*}
      p(\alpha, c \mid x_{1:n}) &\propto p(\alpha, c) \cdot p(x_{1:n} \mid \alpha, c) \\
      &\propto \Ind{\alpha, c > 0} \prod_{i=1}^n \frac{\alpha c^\alpha}{x_i^{\alpha+1}} \Ind{x_i \geq c} \\
      &= \frac{\alpha^n c^{n \alpha}}{(\prod_{i=1}^n x_i)^{\alpha+1}} \Ind{c < \opt{x}} \Ind{\alpha, c > 0}
    \end{align*} where $\opt{x} \defeq \min\{x_1, \dots, x_n\}$.
    It is not clear how one could sample from this distribution directly.
    Instead, we use Gibbs sampling.

    For the posterior distribution of $\alpha$, we have \begin{align*}
      p(\alpha \mid c, x_{1:n}) &\propto p(\alpha, c \mid x_{1:n}) \\
      &\propto \frac{\alpha^n c^{n \alpha}}{(\prod_{i=1}^n x_i)^{\alpha+1}} \Ind{\alpha > 0} \\
      &= \alpha^n \exp\parentheses*{- \alpha \parentheses*{\sum_{i=1}^n \log x_i - n \log c}} \Ind{\alpha > 0} \\
      &\propto \GammaDistr[\alpha]{a}{b}
    \end{align*} where $a \defeq n+1$ and $b \defeq \sum_{i=1}^n \log x_i - n \log c$.
    On the other hand, \begin{align*}
      p(c \mid \alpha, x_{1:n}) &\propto p(\alpha, c \mid x_{1:n}) \propto c^{n \alpha} \Ind{0 < c < \opt{x}}.
    \end{align*}
    It remains to show that it is easy to sample from a random variable $X$ with $p_X(x) \propto x^a \Ind{0 < x < b}$ for $a,b > 0$.
    We have that the normalizing constant is given by $\int_0^b x^a \,d x = \frac{b^{a+1}}{a+1}$.
    Therefore, the CDF of $X$ is \begin{align*}
      P_X(x) &= \int_0^x p(y) \,d y \\
      &= \frac{a+1}{b^{a+1}} \int_0^x y^a \,d y \\
      &= \frac{a+1}{b^{a+1}} \frac{x^{a+1}}{a+1} \\
      &= \parentheses*{\frac{x}{b}}^{a+1}.
    \end{align*}
    The inverse CDF is given by $\inv{P_X}(y) = b y^{\frac{1}{a+1}}$.
    Therefore, we can sample from $X$ using inverse transform sampling (cf. \cref{sec:inverse_transform_sampling}) by sampling $Y \sim \Unif{[0,1]}$ and setting $X = b Y^{\frac{1}{a+1}}$.
  \end{enumerate}
\end{solution}

\begin{solution}{bayesian_logistic_regression_energy_function}
  First, note that the sum of convex functions is convex, hence, we consider each term individually.

  The Hessian of the regularization term is $\lambda \mI$, and thus, by the second-order characterization of convexity, this term is convex in $\vw$.

  Finally, note that the second term is a sum of logistic losses $\ell_\mathrm{log}$ \eqref{eq:logistic_loss}, and we have seen in \cref{exercise:logistic_loss_gradient} that $\ell_\mathrm{log}$ is convex in $\vw$.
\end{solution}

\begin{solution}{maximum_entropy_property_of_gibbs_distribution}
  \begin{enumerate}[beginpenalty=10000]
    \item Our goal is to solve the optimization problem \begin{align}\begin{split}
      &\max_{p \in \Delta^{\spT}} - \sum_{x \in \spT} p(x) \log_2 p(x) \\
      &\text{subject to} \sum_{x \in \spT} p(x) f(x) = \mu \label{eq:maximum_entropy_property_of_gibbs_distribution}
    \end{split}\end{align}
    for some $\mu \in \R$.
    The Lagrangian with dual variables $\lambda_0$ and $\lambda_1$ is given by \begin{align*}
      L(p, \lambda_0, \lambda_1) &= \begin{multlined}[t]
        - \sum_{x \in \spT} p(x) \log_2 p(x) + \lambda_0 \parentheses*{1 - \sum_{x \in \spT} p(x)} + \\ \lambda_1 \parentheses*{\mu - \sum_{x \in \spT} p(x) f(x)}
      \end{multlined} \\
      &= - \sum_{x \in \spT} p(x) \parentheses*{\log_2 p(x) + \lambda_0 + \lambda_1 f(x)} + \const.
    \end{align*}
    Note that $L(p, \lambda_0, \lambda_1) = \H{p}$ if the constraints are satisfied.
    Thus, we simply need to solve the (dual) optimization problem \begin{align*}
      \max_{\vp \geq \vzero} \min_{\lambda_0, \lambda_1 \in \R} L(\vp, \lambda_0, \lambda_1).
    \end{align*}
    We have \begin{align*}
      \pdv{}{p(x)} L(p, \lambda_0, \lambda_1) = - \log_2 p(x) - \lambda_0 - \lambda_1 f(x) - 1.
    \end{align*}
    Setting the partial derivatives to zero, we obtain \begin{align*}
      p(x) &= 2 \exp(- \lambda_0 - \lambda_1 f(x) - 1) \margintag{using $\log_2(\cdot) = \frac{\log(\cdot)}{\log(2)}$} \\
      &\propto \exp(- \lambda_1 f(x)).
    \end{align*}
    Clearly, $p$ is a valid probability mass function when normalized (i.e., for an appropriate choice of $\lambda_0$).
    We complete the proof by setting $T \defeq \frac{1}{\lambda_1}$.

    \item As $T \to \infty$ (and $\lambda_1 \to 0$), the optimization problem reduces to picking the maximum entropy distribution without the first-moment constraint.
    This distribution is the uniform distribution over $\spT$.
    Conversely, as $T \to 0$ (and $\lambda_1 \to \infty$), the Gibbs distribution reduces to a point density around its mode.
  \end{enumerate}
\end{solution}

\begin{solution}{gibbs_sampling_energy_reduction}
  Recall the following two facts: \begin{enumerate}
    \item Gibbs sampling is an instance of Metropolis-Hastings with proposal distribution \begin{align*}
      r(\vxp \mid \vx) = \begin{cases}
        p(x'_i \mid \vxp_{-i}) & \text{if $\vxp$ differs from $\vx$ only in entry $i$} \\
        0 & \text{otherwise} \\
      \end{cases}
    \end{align*} and acceptance distribution $\alpha(\vxp \mid \vx) = 1$.

    \item The acceptance distribution of Metropolis-Hastings where the stationary distribution $p$ is a Gibbs distribution with energy function $f$ is \begin{align*}
      \alpha(\vxp \mid \vx) = \min\left\{1, \frac{r(\vx \mid \vxp)}{r(\vxp \mid \vx)} \exp(f(\vx) - f(\vxp))\right\}.
    \end{align*}
  \end{enumerate}

  We therefore know that \begin{align*}
    \frac{r(\vx \mid \vxp)}{r(\vxp \mid \vx)} \exp(f(\vx) - f(\vxp)) \geq 1.
  \end{align*}
  We remark that this inequality even holds with equality using our derivation of \cref{thm:gibbs_sampling_as_mh}.
  Taking the logarithm and reorganizing the terms, we obtain \begin{align}
    f(\vxp) \leq f(\vx) + \log r(\vx \mid \vxp) - \log r(\vxp \mid \vx). \label{eq:gibbs_sampling_energy_reduction_helper}
  \end{align}

  By the definition of the proposal distribution of Gibbs sampling, \begin{align*}
    r(\vxp \mid \vx) = p(x'_i \mid \vx_{-i}) \quad\text{and}\quad r(\vx \mid \vxp) = p(x_i \mid \vx_{-i}). \margintag{using $\vxp_{-i} = \vx_{-i}$}
  \end{align*}

  Taking the expectation of \cref{eq:gibbs_sampling_energy_reduction_helper}, \begin{align*}
    \E[x'_i \sim p(\cdot \mid \vx_{-i})]{f(\vxp)} &\leq f(\vx) + \log p(x_i \mid \vx_{-i}) - \E[x'_i \sim p(\cdot \mid \vx_{-i})]{\log p(x'_i \mid \vx_{-i})} \\
    &= f(\vx) - \S{p(x_i \mid \vx_{-i})} + \H{p(\cdot \mid \vx_{-i})}.
  \end{align*}

  That is, the energy is expected to decrease if the expected surprise of the new sample $x'_i \mid \vx_{-i}$ is smaller than the surprise of the current sample $x_i \mid \vx_{-i}$.
\end{solution}

\begin{solution}{langevin_dynamics_convergence}
  \begin{enumerate}[beginpenalty=10000]
    \item We take the minimum of \cref{eq:strongly_convex} with respect to $\vy$ on both sides.
    The minimum of the left-hand side is $f(\vzero) = 0$.
    To find the minimum of the right-hand side, we differentiate with respect to $\vy$: \begin{align*}
      \pdv{}{\vy} \brackets*{f(\vx) + \transpose{\grad f(\vx)}(\vy - \vx) + \frac{\alpha}{2} \norm{\vy - \vx}_2^2} = \grad f(\vx) + \alpha (\vy - \vx).
    \end{align*}
    Setting the partial derivative to zero, we obtain \begin{align*}
      \vy = \vx - \frac{1}{\alpha} \grad f(\vx).
    \end{align*}
    Plugging this $\vy$ into \cref{eq:strongly_convex}, we have \begin{align*}
      0 &\geq f(\vx) - \frac{1}{\alpha} \norm{\grad f(\vx)}_2^2 + \frac{1}{2 \alpha} \norm{\grad f(\vx)}_2^2 \\
      &= f(\vx) - \frac{1}{2 \alpha} \norm{\grad f(\vx)}_2^2.
    \end{align*}

    \item Using the chain rule, \begin{align*}
      \odv{}{t} f(\vx_t) &= \transpose{\grad f(\vx_t)} \odv{}{t} \vx_t.
      \intertext{Note that $\odv{}{t} \vx_t = - \grad f(\vx_t)$ by \cref{eq:gradient_flow}, so}
      &= -\norm{\grad f(\vx_t)}_2^2 \\
      &\leq - 2 \alpha f(\vx_t)
    \end{align*} where the last step follows from the PL-inequality \eqref{eq:polyak_lojasiewicz}.

    \item Follows directly from (2) and Grönwall's inequality \eqref{eq:groenwall} by letting $g(t) = f(\vx_t)$ and noting that $- \int_0^t 2 \alpha \,d s = - 2 \alpha t$.

    \item It suffices to show $\grad q_t = q_t \grad \log q_t$.
    By the chain rule, \begin{align*}
      \grad \log q_t = \frac{\grad q_t}{q_t}.
    \end{align*}
    We obtain the desired result by rearranging the terms.

    \item We have \begin{align*}
      \odv{}{t} \KL{q_t}{p} &= \odv{}{t} \int q_t \log \frac{q_t}{p} \,d \vtheta.
      \intertext{By the chain rule,}
      &= \int \pdv{q_t}{t} \log \frac{q_t}{p} \,d \vtheta + \int q_t \pdv{}{t} \log \frac{q_t}{p} \,d \vtheta.
      \intertext{For the second term, we have \begin{align*}
        \int q_t \pdv{}{t} \log \frac{q_t}{p} \,d \vtheta = \int \pdv{q_t}{t} \,d \vtheta = \odv{}{t} \underbrace{\int q_t \,d \vtheta}_1 = 0.
      \end{align*} Plugging in the Fokker-Planck equation \eqref{eq:simp_fokker_planck} into the first term, we obtain}
      &= \int \dive \parentheses*{q_t \grad \log \frac{q_t}{p}} \log \frac{q_t}{p} \,d \vtheta.
      \intertext{Letting $\varphi \defeq \log \frac{q_t}{p}$ and $\vF \defeq \grad \varphi$, and then applying the hint, we have}
      &= \int (\dive q_t \vF) \varphi \,d \vtheta \\
      &= - \int q_t \norm{\grad \varphi}_2^2 \,d \vtheta \\
      &= - \E[\vtheta \sim q_t]{\norm{\grad \log \frac{q_t(\vtheta)}{p(\vtheta)}}_2^2} \\
      &= - \Fisher{q_t}{p}.
    \end{align*}

    \item Noting that $p$ satisfies the LSI with constant $\alpha$ and combining with (5), we have that $\odv{}{t} \KL{q_t}{p} \leq -2 \alpha \KL{q_t}{p}$.
    Observe that this result is analogous to the result derived in (2) and the LSI can intuitively be seen as the PL-inequality, but in the space of distributions.
    Analogously to (3), we obtain the desired convergence result by applying Grönwall's inequality \eqref{eq:groenwall}.

    \item By Pinsker's inequality \eqref{eq:pinsker}, $\norm{q_t - p}_{\mathrm{TV}} \leq e^{-\alpha t} \sqrt{2 \KL{q_0}{p}}$.
    It follows from elementary algebra that $\norm{q_t - p}_{\mathrm{TV}} \leq \epsilon$ if $t \geq \frac{1}{\alpha} \log
    \frac{C}{\epsilon}$ where $C \defeq \sqrt{2 \KL{q_0}{p}}$.
    Thus, $\tau_{\mathrm{TV}}(\epsilon) \in \BigO{\log(1/\epsilon)}$.
  \end{enumerate}
\end{solution}

\begin{solution}{hmc}
  \begin{enumerate}[beginpenalty=10000]
    \item We show that under the dynamics \eqref{eq:hamiltonian_dynamics}, the Hamiltonian $H(\vx, \vy)$ is a constant.
    In particular, $H(\vxp, \vyp) = H(\vx, \vy)$.
    This directly implies that \begin{align*}
      \alpha((\vxp,\vyp) \mid (\vx,\vy)) = \min\{1, \exp(H(\vxp, \vyp) - H(\vx, \vy))\} = 1.
    \end{align*}

    To see why $H(\vx, \vy)$ is constant, we compute \begin{align*}
      \odv{}{t} H(\vx, \vy) &= \grad_\vx H \cdot \odv{\vx}{t} + \grad_\vy H \cdot \odv{\vy}{t} \margintag{using the chain rule} \\
      &= \grad_\vx H \cdot \grad_\vy H - \grad_\vx H \cdot \grad_\vy H \margintag{using the Hamiltonian dynamics \eqref{eq:hamiltonian_dynamics}} \\
      &= 0.
    \end{align*}

    \item By applying one Leapfrog step, we have for $\vx$, \begin{align*}
      \vx(t + \tau) &= \vx(t) + \tau \vy(t + \nicefrac{\tau}{2}) \margintag{using \cref{eq:leapfrog_x}} \\
      &= \vx(t) + \tau \parentheses*{\vy(t) - \frac{\tau}{2} \grad_\vx f(\vx(t))} \margintag{using \cref{eq:leapfrog_y1}} \\
      &= \vx(t) - \frac{\tau^2}{2} \grad_\vx f(\vx(t)) + \tau \vy(t).
    \end{align*}
    Now observe that $\vy(t)$ is a Gaussian random variable, independent of $\vx$ (because we sample $\vy$ freshly at the beginning of the $L$ Leapfrog steps, and we are doing just one Leapfrog step).
    By renaming $\tau' \defeq \nicefrac{\tau^2}{2}$ and $\epsilon \defeq \vy(t)$, we get \begin{align*}
      \vx_{t+1} = \vx_t - \tau' \grad_\vx f(\vx_t) + \sqrt{2 \tau'} \epsilon
    \end{align*} which coincides with the proposal distribution of Langevin Monte Carlo \eqref{eq:mala}.
  \end{enumerate}
\end{solution}

\section*{\nameref{sec:bdl}}

\begin{solution}{softmax_and_logistic_function}
  We have \begin{align*}
    \sigma_1(\vf) &= \frac{\exp(f_1)}{\exp(f_0) + \exp(f_1)} \margintag{using the definition of the softmax function \eqref{eq:softmax_function}} \\
    &= \frac{1}{\exp(f_0 - f_1) + 1} \margintag{multiplying by $\frac{\exp(-f_1)}{\exp(-f_1)}$} \\
    &= \sigma(-(f_0 - f_1)). \margintag{using the definition of the logistic function \eqref{eq:logistic_function}}
  \end{align*}
  $\sigma_0(\vf) = 1 - \sigma(f)$ follows from the fact that $\sigma_0(\vf) + \sigma_1(\vf) = 1$.
\end{solution}

\section*{\nameref{sec:active_learning}}

\begin{solution}{mi_and_kl}
  We have \begin{align*}
    \I{\rX}{\rY} &= \H{\rX} -  \H{\rX \mid \rY} \margintag{using the definition of mutual information \eqref{eq:mi}} \\[5pt]
    &= \E[\vx]{-\log p(\vx)} - \E[(\vx,\vy)]{-\log p(\vx \mid \vy)} \margintag{using the definitions of entropy \eqref{eq:entropy} and conditional entropy \eqref{eq:cond_entropy}} \\[5pt]
    &= \E[(\vx, \vy)]{-\log p(\vx)} - \E[(\vx,\vy)]{-\log p(\vx \mid \vy)} \margintag{using the law of total expectation \eqref{eq:lotp}} \\
    &= \E[(\vx, \vy)]{\log \frac{p(\vx \mid \vy)}{p(\vx)}}.
  \end{align*}

  From this we get directly that \begin{align*}
    \I{\rX}{\rY} &= \E*[\vy]{\E[\vx \mid \vy]{\log \frac{p(\vx \mid \vy)}{p(\vx)}}} \\
    &= \E[\vy]{\KL{p(\vx \mid \vy)}{p(\vx)}}, \margintag{using the definition of KL-divergence \eqref{eq:kl}}
  \end{align*} and we also conclude \begin{align*}
    \I{\rX}{\rY} &= \E[(\vx, \vy)]{\log \frac{p(\vx, \vy)}{p(\vx) p(\vy)}} \margintag{using the definition of conditional probability \eqref{eq:cond_prob}} \\
    &= \KL{p(\vx, \vy)}{p(\vx) p(\vy)}. \margintag{using the definition of KL-divergence \eqref{eq:kl}}
  \end{align*}
\end{solution}

\begin{solution}{cond_mi_non_monotonicity}
  Symmetry of conditional mutual information \eqref{eq:cond_mi_symmetry} and \cref{eq:cond_mi_joint_mi} give the following relationship, \begin{align}
    \I{\rX}{\rY,\rZ} = \I{\rX}{\rY} + \I{\rX}{\rZ \mid \rY} = \I{\rX}{\rZ} + \I{\rX}{\rY \mid \rZ}. \label{eq:cond_mi_non_monotonicity_helper}
  \end{align}
  \begin{enumerate}
    \item $\rX \perp \rZ$ implies $\I{\rX}{\rZ} = 0$.
    Thus, \cref{eq:cond_mi_non_monotonicity_helper} simplifies to \begin{align*}
      \I{\rX}{\rY} + \I{\rX}{\rZ \mid \rY} = \I{\rX}{\rY \mid \rZ}.
    \end{align*}
    Using that $\I{\rX}{\rZ \mid \rY} \geq 0$, we conclude $\I{\rX}{\rY \mid \rZ} \geq \I{\rX}{\rY}$.

    \item $\rX \perp \rZ \mid \rY$ implies $\I{\rX}{\rZ \mid \rY} = 0$.
    \Cref{eq:cond_mi_non_monotonicity_helper} simplifies to \begin{align*}
      \I{\rX}{\rY} = \I{\rX}{\rZ} + \I{\rX}{\rY \mid \rZ}.
    \end{align*}
    Using that $\I{\rX}{\rZ} \geq 0$, we conclude $\I{\rX}{\rY \mid \rZ} \leq \I{\rX}{\rY}$.

    \item Again, \cref{eq:cond_mi_non_monotonicity_helper} simplifies to \begin{align*}
      \I{\rX}{\rY} = \I{\rX}{\rZ} + \I{\rX}{\rY \mid \rZ}.
    \end{align*}
    Using that $\I{\rX}{\rY \mid \rZ} \geq 0$, we conclude $\I{\rX}{\rZ} \leq \I{\rX}{\rY}$.
  \end{enumerate}
\end{solution}

\begin{solution}{interaction_information}
  \begin{enumerate}[beginpenalty=10000]
    \item Expanding the definition of interaction information, one obtains \begin{align*}
      \I{\rX}{\rY ; \rZ} = \; &(\H{\rX} + \H{\rY} + \H{\rZ}) \\
      &- (\H{\rX,\rY} + \H{\rX,\rZ} + \H{\rY,\rZ}) \\
      &+ \H{\rX,\rY,\rZ},
    \end{align*} and hence, interaction information is symmetric.

    \item Conditional on either one of $X_1$ or $X_2$, the distribution of $Y$ remains unchanged, and hence $\I{Y}{X_1 \mid X_2} = \I{Y}{X_2 \mid X_1} = 0$.
    Conversely, conditional on both $X_1$ and $X_2$, $Y$ is fully determined, and hence $\I{Y}{X_1,X_2} = 1$ noting that $Y$ encodes one bit worth of information.
    Thus, $\I{Y ; X_1}{X_2} = -1$ meaning that there is synergy between $X_1$ and $X_2$ with respect to $Y$.
  \end{enumerate}
\end{solution}

\begin{solution}{marginal_gain_mi}
  As suggested, we derive the result in two steps.
  \begin{enumerate}
    \item First, we have \begin{align*}
      \Delta_I(\vx \mid \sA) &= I(\sA \cup \{\vx\}) - I(\sA) \margintag{using the definition of marginal gain \eqref{eq:marginal_gain}} \\[5pt]
      &= \Ism{\vf_{\sA \cup \{\vx\}}}{\vy_\sA, y_\vx} - \I{\vf_\sA}{\vy_\sA} \margintag{using the definition of $I$ \eqref{eq:mi_optimization}} \\
      &= \Ism{\vf_{\sA \cup \{\vx\}}}{\vy_\sA, y_\vx} - \Ism{\vf_{\sA \cup \{\vx\}}}{\vy_\sA} \margintag{using $\vy_\sA \perp f_\vx \mid \vf_\sA$} \\
      &= \Hsm{\vf_{\sA \cup \{\vx\}} \mid \vy_\sA} - \Hsm{\vf_{\sA \cup \{\vx\}} \mid \vy_\sA, y_\vx} \margintag{using the definition of MI \eqref{eq:mi}} \\
      &= \Ism{\vf_{\sA \cup \{\vx\}}}{y_\vx}[\vy_\sA] \margintag{using the definition of cond. MI \eqref{eq:cond_mi}} \\
      &= \Ism{\vf_\sA}{y_\vx}[f_\vx, \vy_\sA] + \Ism{f_\vx}{y_\vx}[\vy_\sA] \margintag{using \cref{eq:cond_mi_joint_mi}} \\
      &= \Ism{f_\vx}{y_\vx}[\vy_\sA]. \margintag{using $\Ism{\vf_\sA}{y_\vx}[f_\vx, \vy_\sA] = 0$ as $y_\vx \perp \vf_\sA \mid f_\vx$}
    \end{align*}

    \item For the second part, we get \begin{align*}
      \Ism{f_\vx}{y_\vx}[\vy_\sA] &= \Ism{y_\vx}{f_\vx}[\vy_\sA] \margintag{using symmetry of conditional MI \eqref{eq:cond_mi_symmetry}} \\
      &= \H{y_{\vx} \mid \vy_\sA} - \Hsm{y_{\vx} \mid f_\vx, \vy_\sA} \margintag{using the definition of cond. MI \eqref{eq:cond_mi}} \\
      &= \H{y_{\vx} \mid \vy_\sA} - \H{y_{\vx} \mid f_\vx} \margintag{using that $y_\vx \perp \vvarepsilon_\sA$ so $y_\vx \perp \vy_\sA \mid f_\vx$} \\
      &= \H{y_{\vx} \mid \vy_\sA} - \H{\varepsilon_\vx}. \margintag{given $f_\vx$, the only randomness in $y_{\vx}$ originates from $\varepsilon_\vx$}
    \end{align*}
  \end{enumerate}
\end{solution}

\begin{solution}{submodularity_and_no_synergy}
  We have \begin{align*}
    \Ism{f_\vx}{y_\vx ; \vy_{\sB \setminus \sA}}[\vy_\sA] &= \Ism{f_\vx}{y_\vx}[\vy_\sA] - \Ism{f_\vx}{y_\vx}[\vy_\sB] \margintag{using the definitiion of interaction information \eqref{eq:interaction_information}} \\[5pt]
    &= \Delta_I(\vx \mid \sA) - \Delta_I(\vx \mid \sB) \margintag{using \cref{eq:mg_mi}} \\
    &\geq 0
  \end{align*} where the final inequality follows from submodularity of $I$.
\end{solution}

\begin{solution}{transductive_active_learning}
  \begin{enumerate}[beginpenalty=10000]
    \item \textbf{Yes}.
    We have \[
      \Delta(j \mid S) = F(S \cup \{j\}) - F(S) = \Hsm{Z}[Y_S] - \Hsm{Z}[Y_{S\cup\{j\}}].
    \]
    This is non-negative iff \(\Hsm{Z}[Y_S] \geq \Hsm{Z}[Y_{S\cup\{j\}}]\) which is the ``information never hurts'' property \eqref{eq:information_never_hurts} of conditional entropy.

    \item \textbf{No}, our acquisition function $F$ is not equivalent to uncertainty sampling.
    Note that all \(X_i\) have identical prior variance.
    It suffices to show that not all \(i_1 \in \{1, \dots, 100\}\) maximize the marginal information gain \(\Delta(i_1 \mid \emptyset)\).

    The prior variance of \(Z\) is \(\Var{Z} = \sum_{i=1}^{100} i^2\) since all \(X_i\) are independent.
    Since the \(x_{1:100}\) and \(Z\) are jointly Gaussian, the posterior variance of \(Z\) after observing \(Y_1\) is \[
        \Var{Z}[Y_1] = \Var{Z} - \frac{\Cov{Z, Y_1}^2}{\Var{Y_1}}.
    \]
    The marginal information gain \(\Delta(i_1 \mid \emptyset)\) is equal up to constants to \(-\frac{1}{2} \log \Var{Z \mid Y_1}\).
    The variance of the observation is \(\Var{Y_1} = \Var{X_{i_1} + \varepsilon_1} = 2\) and the covariance is \(\Cov{Z, Y_1} = i_1\).
    Therefore, \[
        \Var{Z}[Y_1] = \sum_{i=1}^{100} i^2 - \frac{1}{2} i_1,
    \] which is uniquely minimized for \(i_1 = 100\).

    \item The acquisition function is submodular if the previous observation of an input \(X_i\) does not imply that the information conveyed by observing another input \(X_j\) about the value of \(Z\) is larger than it would have been without having previously observed \(X_i\).
    Clearly, if \(i = j\) then the information conveyed by the second observation is always $0$ bit, since the first observation was noiseless.
    However, depending on \(Z\) previous observations of inputs other than \(X_j\) may be crucial for the informativeness of observing \(X_j\).
    \begin{enumerate}
        \item \textbf{Yes}. If \(X_j = 0\) then we know that \(Z = 0\) also, no matter the previously observed inputs. If \(X_j = 1\) then it becomes slightly more probable that \(Z = 1\), however, this does not depend on which other points have been observed previously.
        \item \textbf{Yes}. The case of OR is symmetric to AND. If \(X_j = 1\) then we know that \(Z = 1\) also, no matter the previously observed inputs. If \(X_j = 0\) then it becomes slightly more probable that \(Z = 0\), however, this does not depend on which other points have been observed previously.
        \item \textbf{No}. Consider the case where we have observed \(X_{1:99}\) and are now observing \(X_{100}\). The marginal information gain of all previous observations was $0$ bit, since the observed values for \(X_{1:99}\) did not affect the distribution of \(Z\): the prior of \(Z\) was \(\Bern{0.5}\) and the current posterior is still the same Bernoulli distribution.
        Now, observing \(X_{100}\) will deterministically determine the value of \(f\), therefore conveying $1$ bit of information.
        However, just observing \(X_{100}\) alone, without any prior observations, conveys no information.
        This shows that the marginal information gain can increase after conditioning on more observations, and therefore the acquisition function \(F\) is not submodular with this choice of \(Z\).
    \end{enumerate}\vspace{1ex}
    The key intuition behind the above arguments is that AND and OR are monotonic functions whereas XOR is not.
  \end{enumerate}
\end{solution}

\section*{\nameref{sec:bayesian_optimization}}

\begin{solution}{convergence_to_static_opt}
  First, observe that \begin{align}
    \lim_{T \to \infty} \frac{R_T}{T} &= \max_{\vx} \opt{f}(\vx) - \lim_{T \to \infty} \frac{1}{T} \sum_{t=1}^T \opt{f}(\vx_t) \margintag{using the definition of regret \eqref{eq:regret2}} \nonumber \\
    &= \max_{\vx} \opt{f}(\vx) - \lim_{t \to \infty} \opt{f}(\vx_t). \label{eq:sublinear_regret2} \margintag{using the Cesàro mean \eqref{eq:cesaro_mean}}
  \end{align}

  Now, suppose that the algorithm converges to the static optimum, \begin{align*}
    \lim_{t \to \infty} \opt{f}(\vx_t) = \max_{\vx} \opt{f}(\vx).
  \end{align*}
  Together with \cref{eq:sublinear_regret2} we conclude that the algorithm achieves sublinear regret.

  For the other direction, we prove the contrapositive.
  That is, we assume that the algorithm does not converge to the static optimum and show that it has (super-)linear regret.
  We distinguish between two cases.
  Our assumption is formalized by \begin{align*}
    \lim_{t \to \infty} \opt{f}(\vx_t) < \max_{\vx} \opt{f}(\vx).
  \end{align*}
  Together with \cref{eq:sublinear_regret2} we conclude $\lim_{T \to \infty} \nicefrac{R_T}{T} > 0$.
\end{solution}

\begin{solution}{bayesian_confidence_intervals}
  \begin{enumerate}[beginpenalty=10000]
    \item As suggested by the hint, we let $Z \sim \N{0}{1}$ and $c > 0$, and bound \begin{align}
      \Pr{Z > c} &= \Pr{Z - c > 0} \nonumber \\
      &= \int_0^\infty \frac{1}{\sqrt{2\pi}} e^{-(z+c)^2 / 2} \,d z. \margintag{using the PDF of the univariate normal distribution \eqref{eq:univ_normal}} \nonumber
      \intertext{Note that for $z \geq 0$, \begin{align*}
        \frac{(z+c)^2}{2} = \frac{z^2}{2} + z c + \frac{c^2}{2} \geq \frac{z^2}{2} + \frac{c^2}{2}.
      \end{align*}
      Thus,}
      &\leq e^{-c^2 / 2} \int_0^\infty \frac{1}{\sqrt{2\pi}} e^{-z^2 / 2} \,d z \nonumber \\
      &= e^{-c^2 / 2} \Pr{Z > 0} \nonumber \\
      &= \frac{1}{2} e^{-c^2 / 2}. \margintag{using symmetry of the standard normal distribution around $0$} \label{eq:bayesian_confidence_intervals_helper}
    \end{align}
    Since we made the Bayesian assumption $\opt{f}(\vx) \sim \N{\mu_{0}(\vx)}{\sigma_{0}^2(\vx)}$ and assumed Gaussian observation noise $y_t \sim \N{\opt{f}(\vx_t)}{\sigman^2}$, the posterior is also Gaussian: \begin{align*}
      \opt{f}(\vx) \mid \vx_{1:t}, y_{1:t} \sim \N{\mu_{t}(\vx)}{\sigma_{t}^2(\vx)}.
    \end{align*}
    Hence, writing $\fnPr_t(\cdot) \defeq \fnPr(\cdot \mid \vx_{1:t}, y_{1:t})$, \begin{align*}
      \fnPr_{t-1}(\opt{f}(\vx) \not\in \spC_t(\vx)) &= \fnPr_{t-1}(\abs{\opt{f}(\vx) - \mu_{t-1}(\vx)} > \beta_t \sigma_{t-1}(\vx)) \\
      &= 2 \fnPr_{t-1}\parentheses*{\frac{\opt{f}(\vx) - \mu_{t-1}(\vx)}{\sigma_{t-1}(\vx)} > \beta_t} \margintag{using symmetry of the Gaussian distribution} \\
      &\leq e^{- \beta_t^2 / 2}. \margintag{using \cref{eq:bayesian_confidence_intervals_helper} with $c = \beta_t$}
    \end{align*}

    \item We have \begin{align*}
      \Pr{\bigcup_{\vx \in \spX} \bigcup_{t \geq 1} \opt{f}(\vx) \not\in \spC_t(\vx)} &\leq \sum_{\vx \in \spX} \sum_{t \geq 1} \Pr{\opt{f}(\vx) \not\in \spC_t(\vx)} \margintag{using a union bound \eqref{eq:union_bound}} \\
      &\leq \card{\spX} \sum_{t \geq 1} e^{- \beta_t^2 / 2} \margintag{using (1)}.
      \intertext{Letting $\beta_t^2 \defeq 2 \log(\card{\spX} (\pi t)^2 / (6 \delta))$, we get}
      &= \frac{6 \delta}{\pi^2} \sum_{t \geq 1} \frac{1}{t^2} \\
      &= \delta. \margintag{using $\sum_{t \geq 1} \frac{1}{t^2} = \frac{\pi^2}{6}$}
    \end{align*}
  \end{enumerate}
\end{solution}

\begin{solution}{bayesian_regret_for_gp_ucb}
  \begin{enumerate}[beginpenalty=10000]
    \item We denote the static optimum by $\vxs$.
    By the definition of $\vx_t$, \begin{align*}
      \mu_{t-1}(\vx_t) + \beta_t \sigma_{t-1}(\vx_t) &\geq \mu_{t-1}(\vxs) + \beta_t \sigma_{t-1}(\vxs) \\
      &\geq \opt{f}(\vxs). \margintag{using \cref{eq:gp_ucb_beta_calibration}}
    \end{align*}
    Thus, \begin{align*}
      r_t &= \opt{f}(\vxs) - \opt{f}(\vx_t) \\
      &\leq \beta_t \sigma_{t-1}(\vx_t) + \mu_{t-1}(\vx_t) - \opt{f}(\vx_t) \\
      & \leq 2 \beta_t \sigma_{t-1}(\vx_t). \margintag{again using \cref{eq:gp_ucb_beta_calibration}}
    \end{align*}

    \item We have for any fixed $T$, \begin{align*}
      \I{\vf_{T+1}}{\vy_{T+1}} &= \H{\vy_{T+1}} - \H{\vvarepsilon_{T+1}} \margintag{analogously to \cref{eq:mg_decomp}} \\
      &= \H{\vy_T} - \H{\vvarepsilon_T} + \H{y_{\vx_{T+1}} \mid \vy_T} - \H{\varepsilon_{\vx_{T+1}}} \margintag{using the chain rule for entropy \eqref{eq:chain_rule_entropy} and the mutual independence of $\vvarepsilon_{T+1}$} \\[5pt]
      &= \I{\vf_T}{\vy_T} + \I{f_{\vx_{T+1}}}{y_{\vx_{T+1}} \mid \vy_T} \margintag{using the definition of MI \eqref{eq:mi}} \\
      &= \I{\vf_T}{\vy_T} + \frac{1}{2}\log\parentheses*{1 + \frac{\sigma_T^2(\vx_{T+1})}{\sigman^2}}. \margintag{using \cref{eq:mi_cond_linear_gaussians}}
    \end{align*}
    Note that $\I{\vf_0}{\vy_0} = 0$.
    The result then follows by induction.

    \item By Cauchy-Schwarz, $R_T^2 \leq T \sum_{t=1}^T r_t^2$, and hence, it suffices to show $\sum_{t=1}^T r_t^2 \leq \BigO{\beta_T^2 \gamma_T}$.
    We have \begin{align*}
      \sum_{t=1}^T r_t^2 &\leq 4 \beta_T^2 \sum_{t=1}^T \sigma_{t-1}^2(\vx_t) \margintag{using part (1)} \\
      &= 4 \sigman^2 \beta_T^2 \sum_{t=1}^T \frac{\sigma_{t-1}^2(\vx_t)}{\sigman^2}.
      \intertext{Observe that $\sigma_{t-1}^2(\vx_t) /\sigman^2$ is bounded by $M \defeq \max_{\vx \in \spX}\sigma_0^2(\vx) / \sigman^2$ as variance is monotonically decreasing (cf. \cref{sec:fundamentals:gaussians}). Applying the hint, we obtain}
      &\leq 4 C \sigman^2 \beta_T^2 \sum_{t=1}^T \log\parentheses*{1 + \frac{\sigma_{t-1}^2(\vx_t)}{\sigman^2}} \\
      &= 8 C \sigman^2 \beta_T^2 \I{\vf_T}{\vy_T} \margintag{using part (2)} \\
      &\leq 8 C \sigman^2 \beta_T^2 \gamma_T. \margintag{using the definition of $\gamma_T$ \eqref{eq:gamma_t}}
    \end{align*}
  \end{enumerate}
\end{solution}

\begin{solution}{sublinear_regret_for_linear_kernel}
  \begin{enumerate}[beginpenalty=10000]
    \item Let $\sS \subseteq \spX$ be such that $\card{\sS} \leq T$.
    Recall from \cref{eq:mi_cond_linear_gaussians} that $\I{\vf_\sS}{\vy_\sS} = \frac{1}{2} \log \det{\mI + \sigman^{-2}\mK_{\sS\sS}}$.
    Using that the kernel is linear we can rewrite $\mK_{\sS\sS} = \transpose{\mX_\sS} \mX_\sS$.
    Using Weinstein-Aronszajn's identity \eqref{eq:weinstein_aronszajn} we have \begin{align*}
      \I{\vf_\sS}{\vy_\sS} = \frac{1}{2}\log\det{\mI + \sigman^{-2} \transpose{\mX_\sS} \mX_\sS} = \frac{1}{2}\log\det{\mI + \sigman^{-2} \mX_\sS \transpose{\mX_\sS}}.
    \end{align*}
    If we define $\mM \defeq \mI + \sigman^{-2} \mX_\sS \transpose{\mX_\sS}$ as a sum of symmetric positive definite matrices, $\mM$ itself is symmetric positive definite.
    Thus, we have from Hadamard's inequality \eqref{eq:hadamard}, \begin{align*}
      \det{\mM} &\leq \det{\diag{\mI + \sigman^{-2} \mX_\sS \transpose{\mX_\sS}}} \margintag{$\diag{\mA}$ refers to the diagonal matrix whose elements are those of $\mA$} \\
      &= \det{\mI + \sigman^{-2} \diag{\mX_\sS \transpose{\mX_\sS}}}.
    \end{align*}
    Note that \begin{align*}
      \diag{\mX_\sS \transpose{\mX_\sS}}(i,i) &= \sum_{t=1}^{\card{\sS}} \vx_t(i)^2 \\
      &\leq \sum_{i=1}^d \sum_{t=1}^{\card{\sS}} \vx_t(i)^2 = \sum_{t=1}^{\card{\sS}} \underbrace{\norm{\vx_t}_2^2}_{\leq 1} \leq \card{\sS} \leq T.
    \end{align*}
    If we denote by $\lambda \leq T$ the largest term of $\diag{\mX_\sS \transpose{\mX_\sS}}$ then we have \begin{align*}
      \det{\mM} \leq (1 + \sigman^{-2} \lambda)^d \leq (1 + \sigman^{-2} T)^d,
    \end{align*} yielding, \begin{align*}
      \I{\vf_\sS}{\vy_{\sS}} \leq \frac{d}{2} \log(1 + \sigman^{-2} T)
    \end{align*} implying that $\gamma_T = \BigO{d \log T}$.

    \item Using the regret bound (cf. \cref{thm:bayesian_regret_for_gp_ucb}) and the Bayesian confidence intervals (cf. \cref{thm:bayesian_confidence_intervals}), and then $\gamma_T = \BigO{d \log T}$, we have \begin{align*}
      R_T = \BigO{\beta_T \sqrt{\gamma_T T}} = \BigOTil{\sqrt{d T}},
    \end{align*} and hence, $\lim_{T\to\infty} \frac{R_T}{T} = 0$.
  \end{enumerate}
\end{solution}

\begin{solution}{closed_form_ei}
  \begin{enumerate}[beginpenalty=10000]
    \item Note that $f$ is a Gaussian process, and hence, our posterior distribution after round $t$ is entirely defined by the mean function $\mu_{t}$ and the covariance function $k_{t}$.
    Reparameterizing the posterior distribution using a standard Gaussian \eqref{eq:gaussian_affine_transformation}, we obtain \begin{align*}
      f(\vx) \mid \vx_{1:t}, y_{1:t} = \mu_{t}(\vx) + \sigma_{t}(\vx) \varepsilon
    \end{align*} for $\varepsilon \sim \N{0}{1}$.
    We get \begin{align*}
      \mathrm{EI}_t(\vx) &= \E[f(\vx) \sim \N{\mu_{t}(\vx)}{\sigma_{t}^2(\vx)}]{I_t(\vx)} \margintag{using the definition of expected improvement \eqref{eq:ei}} \\[5pt]
      &= \E[f(\vx) \sim \N{\mu_{t}(\vx)}{\sigma_{t}^2(\vx)}]{(f(\vx) - \hat{f}_t)_+} \margintag{using the definition of improvement \eqref{eq:improvement}} \\[5pt]
      &= \E[\varepsilon \sim \N{0}{1}]{(\mu_{t}(\vx) + \sigma_{t}(\vx) \varepsilon - \hat{f}_t)_+} \margintag{using the reparameterization} \\
      &= \int_{-\infty}^{+\infty} (\mu_{t}(\vx) + \sigma_{t}(\vx) \varepsilon - \hat{f}_t)_+ \cdot \phi(\varepsilon) \,d\varepsilon.
    \end{align*}
    For $\varepsilon < \frac{\hat{f}_t - \mu_{t}(\vx)}{\sigma_{t}(\vx)} \eqdef z_t(\vx)$ we have $(\mu_{t}(\vx) + \sigma_{t}(\vx) \varepsilon - \hat{f}_t)_+ = 0$.
    Thus, we obtain \begin{align}
      \mathrm{EI}_t(\vx) &= \int_{z_t(\vx)}^{+\infty} (\mu_{t}(\vx) + \sigma_{t}(\vx) \varepsilon - \hat{f}_t) \cdot \phi(\varepsilon) \,d\varepsilon. \label{eq:ei_closed_form_helper}
    \end{align}

    \item By splitting the integral from \cref{eq:ei_closed_form_helper} into two distinct terms, we obtain \begin{align*}
      \mathrm{EI}_t(\vx) &= \begin{multlined}[t]
        (\mu_{t}(\vx) - \hat{f}_t) \int_{z_t(\vx)}^{+\infty} \phi(\varepsilon) \,d\varepsilon \\ - \sigma_{t}(\vx) \int_{z_t(\vx)}^{+\infty} (-\varepsilon) \cdot \phi(\varepsilon) \,d\varepsilon.
      \end{multlined}
    \intertext{For the first term, we use the symmetry of $\N{0}{1}$ around $0$ to write the integral in terms of the CDF.
    For the second term, we notice that $(-\varepsilon) \cdot \phi(\varepsilon) = \frac{1}{\sqrt{2\pi}} \odv{}{\varepsilon} e^{\nicefrac{-\varepsilon^2}{2}}$.
    Thus, we can derive this integral directly,}
      &= \begin{multlined}[t]
        (\mu_{t}(\vx) - \hat{f}_t) \Phi\parentheses*{-z_t(\vx)} \\ - \sigma_{t}(\vx) \parentheses*{\lim_{\varepsilon\to\infty}\phi(\varepsilon) - \phi\parentheses*{z_t(\vx)}}.
      \end{multlined}
    \end{align*}
    Using the symmetry of $\phi$ around $0$, we obtain \begin{align*}
      \mathrm{EI}_t(\vx) = (\mu_{t}(\vx) - \hat{f}_t) \Phi\parentheses*{\frac{\mu_{t}(\vx) - \hat{f}_t}{\sigma_{t}(\vx)}} + \sigma_{t}(\vx) \phi\parentheses*{\frac{\mu_{t}(\vx) - \hat{f}_t}{\sigma_{t}(\vx)}}.
    \end{align*}
  \end{enumerate}
\end{solution}

\begin{solution}{regret_of_ids}
  \begin{enumerate}[beginpenalty=10000]
    \item As $\vx_t^{\mathrm{UCB}}$ is the UCB action, \begin{align*}
      \hat{\Delta}_t(\vx_t^{\mathrm{UCB}}) = u_t(\vx_t^{\mathrm{UCB}}) - l_t(\vx_t^{\mathrm{UCB}}) = 2 \beta_{t+1} \sigma_t(\vx_t^{\mathrm{UCB}}).
    \end{align*}

    \item We first bound $I_t(\vx_t^{\mathrm{UCB}})$: \begin{align*}
      I_t(\vx_t^{\mathrm{UCB}}) &= \I{f_{\vx_t^{\mathrm{UCB}}}}{y_{\vx_t^{\mathrm{UCB}}}}[\vx_{1:t}, y_{1:t}] \\
      &= \frac{1}{2} \log\parentheses*{1 + \frac{\sigma_t^2(\vx_t^{\mathrm{UCB}})}{\sigman^2}}. \margintag{using \cref{eq:mi_cond_linear_gaussians}}
      \intertext{Note that $\sigma_t^2(\vx) / \sigman^2 \leq C$ for some constant $C$ since variance is decreasing monotonically. So, applying the hint,}
      &\geq \frac{\sigma_t^2(\vx_t^{\mathrm{UCB}})}{2 C \sigman^2}.
    \end{align*}
    Combining this with (1), we obtain \begin{align*}
      \widehat{\Psi}_t(\vx_t^{\mathrm{UCB}}) &= \frac{\hat{\Delta}_t(\vx_t^{\mathrm{UCB}})^2}{I_t(\vx_t^{\mathrm{UCB}})} \leq 8 C \sigman^2 \beta_{t+1}^2.
    \end{align*}

    \item With high probability, \begin{align*}
      \Psi_t(\vx_{t+1}) \leq \widehat{\Psi}_t(\vx_{t+1}) \leq \widehat{\Psi}_t(\vx_t^{\mathrm{UCB}}) \leq 8 C \sigman^2 \beta_{t+1}^2,
    \end{align*} where the first inequality is due to $\Delta(\vx) \leq \hat{\Delta}_t(\vx)$ with high probability, the second inequality due to the definition of the IDS algorithm \eqref{eq:ids}, and the third inequality is from (2).
    Invoking \cref{thm:regret_information_ratio} with $\overline{\Psi}_T \defeq 8 C \sigman^2 \beta_T^2$, we obtain that \begin{align*}
      R_T &\leq \sqrt{\gamma_T \overline{\Psi}_T T} = \beta_T \sqrt{8 C \sigman^2 \gamma_T T}
    \end{align*} with high probability.
  \end{enumerate}
\end{solution}

\begin{solution}{lite}
  First notice that by the definition of $\fnS'$, we obtain the simpler objective:
  \begin{equation*}
    \spW(\pi) = \sum_{x \in \spX} \pi(x) \parentheses*{ \mu_t(x) + \phi(\Phi^{-1}(\pi(x))) \sigma_t(x) }.
  \end{equation*}
  Next, we show that \(\spW(\pi)\) is concave by computing the Hessian:
  \begin{align*}
    \frac{\partial}{\partial \pi(x)} \spW(\pi) & = \mu_t(x) - \sigma_t(x) \Phi^{-1}(\pi(x)) \phi(\Phi^{-1}(\pi(x))) \frac{d}{d \pi(x)} \Phi^{-1}(\pi(x))\\
    & = \mu_t(x) - \sigma_t(x) \Phi^{-1}(\pi(x))\nonumber\\
    \frac{\partial^2}{\partial \pi(x) \partial \pi(z)} \spW(r) & = - \sigma_t(x) \Ind{x=z} \frac{1}{\phi(\Phi^{-1}(\pi(x)))} \begin{cases}
      < 0 & x = z\\
      = 0 & x \not = z
    \end{cases},
  \end{align*}
  where we used the inverse function rule twice.
  From negative definiteness of the Hessian, it follows that $\spW(\cdot)$ is strictly concave.

  We show next that the optimum lies in the relative interior of the probability simplex, \(\pi^* \in \text{relint}(\Delta^{\spX})\).
  Indeed, at the border of the probability simplex the partial derivatives explode:
  \begin{equation*}
    \frac{\partial}{\partial \pi(x)} \spW(r) = \mu_t(x) - \sigma_t(x) \Phi^{-1}(\pi(x)) = \begin{cases}
      \infty & \pi(x) \to 0^+\\
      \text{finite} & \pi(x) \in (0,1)\\
      -\infty & \pi(x) \to 1^-
    \end{cases}.
  \end{equation*}
  Together with the concavity of \(\spW(\cdot)\) this ensures that \(r^* \in \text{relint}(\Delta^{\spX})\).
  Hence, \(\pi^*\) is a local optimizer of \(\spW(r)\) on the plane defined by \({\sum_{x \in \spX} \pi(x) = 1}\).
  Consequently, we obtain the Lagrangian
  \begin{equation*}
    \spL(\pi, \kappa) : (0,1)^{|\spX|} \times \R \to \R,\quad \pi \mapsto \spW(\pi) + \kappa \parentheses*{1-\sum_{x\in\spX} \pi(x)}.
  \end{equation*}
  Setting its partial derivatives equal to zero, we derive the closed-form solution:
  \begin{align*}
    0 = \mu_t(x) - \sigma_t(x) \Phi^{-1}(\pi^*(x)) - \kappa^* \iff \pi^*(x)  = \Phi\parentheses*{\frac{\mu_t(x) - \kappa^*}{\sigma_t(x)}},
  \end{align*}
  where \(\kappa^*\) ensures a valid distribution, i.e., \(\sum_{x\in \spX} \pi^*(x) = 1\).
\end{solution}

\begin{solution}{recall}
  Since $R_1$ and $R_2$ are independent, we have \begin{align*}
    \Pr{\max\{R_1, R_2\} \leq x} &= \Pr{\{R_1 \leq x\} \cap \{R_2 \leq x\}} \\
    &= \Pr{R_1 \leq x} \cdot \Pr{R_2 \leq x} \margintag{using independence} \\
    &= F_{R_1}(x) \cdot F_{R_2}(x) \\
    &= \Phi^2\parentheses*{\frac{x}{10}}
  \end{align*} where $\Phi$ is the CDF of the standard Gaussian.
  We are looking for the probability that either $R_1$ or $R_2$ are larger than $S = 1$: \begin{align*}
    \Pr{\max\{R_1, R_2\} > 1} = 1 - \Pr{\max\{R_1, R_2\} \leq 1} = 1 - \Phi^2\parentheses*{\frac{1}{10}}.
  \end{align*}
  We have $\Phi^2(\frac{1}{10}) \approx 0.29$. Thus, the probability that either $R_1$ or $R_2$ are larger than $S = 1$ is approximately $0.71$.
  Due to the symmetry in the problem, we know \begin{align*}
    \Pr{R_1 > 1} = \Pr{R_2 > 1} \approx 0.35.
  \end{align*}
\end{solution}

\section*{\nameref{sec:mdp}}

\begin{solution}{value_functions}
  We can use \cref{eq:q_function3} to write the state-action values as a linear system of equations (i.e., as a ``table'').
  This linear system can be solved, for example, using Gaussian elimination to yield the desired result.
\end{solution}

\begin{solution}{v_and_q_functions}
  It follows directly from the definition of the state-action value function \eqref{eq:q_function2} that \begin{align*}
    \argmax_{a \in \sA} \q{x}{a} = \argmax_{a \in \sA} r(x, a) + \gamma \sum_{x' \in \sX} p(x' \mid x, a) \cdot \v{x'}.
  \end{align*}
\end{solution}

\begin{solution}{optimal_policies}
  \begin{enumerate}[beginpenalty=10000]
    \item Recall from Bellman's theorem \eqref{eq:bop1} that a policy is optimal iff it is greedy with respect to its state-action value function.
    Now, observe that in the ``poor, unknown'' state, the policy $\pi$ is not greedy.

    \item Analogously to \cref{exercise:value_functions}, we write the state-action values as a linear system of equations and solve the system using, e.g., Gaussian elimination.

    \item Observe from the result of (2) that $\pi'$ is greedy with respect to its state-action value function, and hence, it follows from Bellman's theorem that $\pi'$ is optimal.
  \end{enumerate}
\end{solution}

\begin{solution}{policy_iteration_linear_convergence}
  Using the hint and $\vvs \geq \vv^{\pi}$ for any policy $\pi$, \begin{align*}
    \norm{\vv^{\pi_t} - \vvs}_\infty &\leq \norm{\mBs \vv^{\pi_{t-1}} - \vvs}_\infty \\
    &= \norm{\mBs \vv^{\pi_{t-1}} - \mBs\vvs}_\infty \margintag{using that $\vvs$ is a fixed-point of $\mBs$, that is, $\mBs\vvs = \vvs$} \\[5pt]
    &\leq \gamma \norm{\vv^{\pi_{t-1}} - \vvs}_\infty \margintag{using that $\mBs$ is a contraction, see \cref{thm:vi_convergence}} \\[5pt]
    &\leq \gamma^t \norm{\vv^{\pi_0} - \vvs}_\infty. \margintag{by induction}
  \end{align*}
\end{solution}

\begin{solution}{reward_modification}
  \begin{enumerate}[beginpenalty=10000]
    \item Recall that the value function $\fnv[\pi][\sM]$ for an MDP $\sM$ is defined as $\v[\pi]{x}[\sM] = \E[\pi]{\sum_{t=0}^\infty \gamma^t R_t}[X_0 = x]$.
    Given an optimal policy $\pis$ for $\sM$ and any policy $\pi$, we know that for any $x \in \sX$, \begin{align*}
      &&\v[\pis]{x}[\sM] &\geq \v[\pi]{x}[\sM] \\
      \iff&&\qquad \E[\pis]{\textstyle\sum_{t=0}^\infty \gamma^t R_t}[X_0 = x] &\geq \E[\pi]{\textstyle\sum_{t=0}^\infty \gamma^t R_t}[X_0 = x] \\
      \iff&&\qquad \E[\pis]{\textstyle\sum_{t=0}^\infty \gamma^t \alpha R_t}[X_0 = x] &\geq \E[\pi]{\textstyle\sum_{t=0}^\infty \gamma^t \alpha R_t}[X_0 = x] \margintag{multiplying both sides by $\alpha$} \\
      \iff&&\qquad \E[\pis]{\textstyle\sum_{t=0}^\infty \gamma^t R'_t}[X_0 = x] &\geq \E[\pi]{\textstyle\sum_{t=0}^\infty \gamma^t R'_t}[X_0 = x] \\
      \iff&&\qquad \v[\pis]{x}[\sM'] &\geq \v[\pi]{x}[\sM'].
    \end{align*}
    Thus, $\pis$ is an optimal policy for $\sM'$.

    \item We give an example where the optimal policies differ when rewards are shifted.

    Consider an MDP with three states $\{1,2,3\}$ where $1$ is the initial state and $3$ is a terminal state.
    If one plays action $A$ in states $1$ or $2$ one transitions directly to the terminal state.
    Additionally, in state $1$ one can play action $B$ which leads to state $2$.
    Let every transition give a deterministic reward of $r \defeq -1$.
    Then it is optimal to traverse the shortest path to the terminal state, in particular, to choose action $A$ when in state $1$.

    If we consider the reward $r' \defeq r + 2 = 1$, then it is optimal to traverse the longest path to the terminal state, in particular, to choose action $B$ when in state $1$.

    \item For an MDP $\sM$, we know that is optimal state-action value function satisfies Bellman's optimality equation \eqref{eq:bop2_q}, \begin{align*}
      \q*{x}{a}[\sM] &= \E[x' \mid x, a]{r(x, x') + \gamma \max_{a' \in \sA} \q*{x'}{a'}[\sM]}.
    \end{align*}
    For the MDP $\sM'$, we have \begin{align*}
      \q*{x}{a}[\sM'] &= \E[x' \mid x, a]{r'(x, x') + \gamma \max_{a' \in \sA} \q*{x'}{a'}[\sM']} \\
      &= \E[x' \mid x, a]{r(x, x') + f(x, x') + \gamma \max_{a' \in \sA} \q*{x'}{a'}[\sM']} \\
      &= \E[x' \mid x, a]{r(x, x') + \gamma \phi(x') - \phi(x) + \gamma \max_{a' \in \sA} \q*{x'}{a'}[\sM']}.
    \end{align*}
    Reorganizing the terms, we obtain \begin{align*}
      \q*{x}{a}[\sM'] + \phi(x) = \E[x' \mid x, a]{r(x, x') + \gamma \max_{a' \in \sA} \parentheses*{\q*{x'}{a'}[\sM'] + \phi(x')}}.
    \end{align*}
    If we now define $\q{x}{a} \defeq \q*{x}{a}[\sM'] + \phi(x)$, we have \begin{align*}
      \q{x}{a} &= \E[x' \mid x, a]{r(x, x') + \gamma \max_{a' \in \sA} \q{x'}{a'}}.
    \end{align*}
    This is exactly Bellman's optimality equation for the MDP $\sM$ with reward function $r$, and hence, $\fnq \equiv \fnq[\star][\sM]$.

    If we take $\pis$ to be an optimal policy for $\sM$, then it satisfies \begin{align*}
      \pis(x) &\in \argmax_{a \in \sA} \q*{x}{a}[\sM] \\
      &= \argmax_{a \in \sA} \q*{x}{a}[\sM'] + \phi(x) \margintag{using the above characterization of $\fnq[\star][\sM]$} \\
      &= \argmax_{a \in \sA} \q*{x}{a}[\sM']. \margintag{using that $\phi(x)$ is independent of $a$}
    \end{align*}
  \end{enumerate}
\end{solution}

\begin{solution}{fishing_problem_pomdb}
  \begin{enumerate}[beginpenalty=10000]
    \item We compute the answer in three steps: \begin{enumerate}
      \item \textbf{Predict step}: We compute the predicted belief after action $W$ but before observing $o_1$. \begin{align*}
        b'_0(F) &= b_0(F) \cdot 0.6 + b_0(\overline{F}) \cdot 0.5 = 0.55 \\
        b'_0(\overline{F}) &= b_0(F) \cdot 0.4 + b_0(\overline{F}) \cdot 0.5 = 0.45.
      \end{align*}
      \item \textbf{Update step}: Using the observation $o_1$ and the observation model, we update the belief. \begin{align*}
        b_1(F) &= \frac{1}{Z} b'_0(F) \Pr{o_1 \mid F} = \frac{1}{Z} \cdot 0.55 \cdot 0.8 \\
        b_1(\overline{F}) &= \frac{1}{Z} b'_0(\overline{F}) \Pr{o_1 \mid \overline{F}} = \frac{1}{Z} \cdot 0.45 \cdot 0.3.
      \end{align*}
      \item \textbf{Normalization}: We compute the normalization constant $Z$. \begin{align*}
        Z = 0.55 \cdot 0.8 + 0.45 \cdot 0.3 = 0.575.
      \end{align*}
    \end{enumerate}
    Therefore, \begin{align*}
      b_1(F) &= \frac{0.55 \cdot 0.8}{0.575} \approx 0.765, \\
      b_1(\overline{F}) &= \frac{0.45 \cdot 0.3}{0.575} \approx 0.235.
    \end{align*}

    \item We observe that if $A_1 = P$, then $\Pr{X_2 = F} = 0$ given any starting state, which will result in a zero in the belief update formula.
    Therefore, \begin{itemize}
      \item $b_2 = (0, 1)$ given $A_1 = P$ and $O_1 = o_1$,
      \item $b_2 = (0, 1)$ given $A_1 = P$ and $O_1 = o_2$.
    \end{itemize}

    \textbf{Case with $A_1 = W$ and $O_1 = o_1$}:
    Using the belief update formula~\eqref{eq:belief}, \begin{align*}
      b_2(F) &\propto \Pr{o_1 \mid F} \cdot \parentheses*{b_1(F) \cdot p(F \mid F, W) + b_1(\overline{F}) \cdot p(F \mid \overline{F}, W)} \approx 0.461 \\
      b_2(\overline{F}) &\propto \Pr{o_1 \mid \overline{F}} \cdot \parentheses*{b_1(F) \cdot p(\overline{F} \mid F, W) + b_1(\overline{F}) \cdot p(\overline{F} \mid \overline{F}, W)} \approx 0.127.
    \end{align*}
    By rescaling the probabilities, we get $b_2 \approx (0.784, 0.216)$.

    \textbf{Case with $A_1 = W$ and $O_1 = o_2$}:
    Using the belief update formula~\eqref{eq:belief}, \begin{align*}
      b_2(F) &\propto \Pr{o_2 \mid F} \cdot \parentheses*{b_1(F) \cdot p(F \mid F, W) + b_1(\overline{F}) \cdot p(F \mid \overline{F}, W)} \approx 0.115 \\
      b_2(\overline{F}) &\propto \Pr{o_2 \mid \overline{F}} \cdot \parentheses*{b_1(F) \cdot p(\overline{F} \mid F, W) + b_1(\overline{F}) \cdot p(\overline{F} \mid \overline{F}, W)} \approx 0.296.
    \end{align*}
    By rescaling the probabilities, we get $b_2 \approx (0.279, 0.721)$.
  \end{enumerate}
\end{solution}

\section*{\nameref{sec:tabular_rl}}

\begin{solution}{q_learning}
  \begin{enumerate}[beginpenalty=10000]
    \item $\Q*{A}{\downarrow} = 1.355, \Q*{G_1}{\text{exit}} = 5.345, \Q*{G_2}{\text{exit}} = 0.5$

    \item Repeating the given episodes infinitely often will not lead to convergence to the optimal Q-function because not all state-action pairs are visited infinitely often.

    Let us assume we observe the following episode instead of the first episode.
    \begin{center}
      \begin{tabular}{|cccc|}
        \multicolumn{4}{c}{Episode 3} \\
        \hline
        $x$ & $a$ & $x'$ & $r$ \\
        \hline
        $A$ & $\rightarrow$ & $B$ & $0$ \\
        $B$ & $\downarrow$ & $G_2$ & $0$ \\
        $G_2$ & exit & & $1$ \\
        \hline
      \end{tabular}
      \vspace{5pt}
    \end{center}
    If we repeat episodes 2 and 3 infinitely often, Q-learning will converge to the optimal Q-function as all state-action pairs will be visited infinitely often.

    \item First, recall that Q-learning is an off-policy algorithm, and hence, even if episodes are obtained off-policy, Q-learning will still converge to the optimal Q-function (if the other convergence conditions are met).
    Note that it only matters which state-action pairs are observed and not which policies were followed to obtain these observations.

    The ``closer'' the initial Q-values are to the optimal Q-function, the faster the convergence of Q-learning.
    However, if the convergence conditions are met, Q-learning will converge to the optimal Q-function regardless of the initial Q-values.

    \item $\v*{A} = 10$, $\v*{B} = 10$, $\v*{G_1} = 10$, $\v*{G_2} = 1$
  \end{enumerate}
\end{solution}

\section*{\nameref{sec:mfarl}}

\begin{solution}{q_learning_func_approx}
  \begin{enumerate}[beginpenalty=10000]
    \item We have to show that \begin{align*}
      \v*{x} = \max_{a \in \sA} r(x, a) + \gamma \E[x' \mid x, a]{\v*{x'}}
    \end{align*} for every $x \in \{1, 2, \dots, 7\}$.
    We give a derivation here for $x = 1$ and $x = 2$.
    \begin{itemize}
      \item For $x = 1$, \begin{align*}
        \v*{1} &= -3 \\
        \max_{a \in \sA} r(1, a) + \gamma \E[x' \mid 1, a]{\v*{x'}} &= -3
      \end{align*} since \begin{align*}
        r(1, a) + \gamma \E[x' \mid 1, a]{\v*{x'}} = \begin{cases}
          -1 + -2 = -3 & \text{if $a = 1$} \\
          -1 + -3 = -4 & \text{if $a = -1$}.
        \end{cases}
      \end{align*}
      \item Likewise, for $x = 2$, \begin{align*}
        \v*{2} &= -2 \\
        \max_{a \in \sA} r(2, a) + \gamma \E[x' \mid 2, a]{\v*{x'}} &= -2
      \end{align*} since \begin{align*}
        r(2, a) + \gamma \E[x' \mid 2, a]{\v*{x'}} = \begin{cases}
          -1 + -1 = -2 & \text{if $a = 1$} \\
          -1 + -3 = -4 & \text{if $a = -1$}.
        \end{cases}
      \end{align*}
    \end{itemize}

    \item We have \begin{align*}
      \Q{3}{-1} &= 0 + \frac{1}{2}\parentheses*{-1 + \max_{a' \in \sA} \Q{2}{a'}} = \frac{1}{2} (-1 + 0) = - \frac{1}{2} \\
      \Q{2}{1} &= 0 + \frac{1}{2}\parentheses*{-1 + \max_{a' \in \sA} \Q{3}{a'}} = \frac{1}{2} (-1 + 0) = - \frac{1}{2} \\
      \Q{3}{1} &= 0 + \frac{1}{2}\parentheses*{-1 + \max_{a' \in \sA} \Q{4}{a'}} = \frac{1}{2} (-1 + 0) = - \frac{1}{2} \\
      \Q{4}{1} &= 0 + \frac{1}{2}\parentheses*{0 + \max_{a' \in \sA} \Q{4}{a'}} = \frac{1}{2} (0 + 0) = 0.
    \end{align*}

    \item We compute \begin{align*}
      \grad_\vw \ell(\vw; \tau) &= - \parentheses*{r + \gamma \max_{a' \in \sA} \Q{x'}{a'; \old{\vw}} - \Q{x}{a; \vw}} \begin{bmatrix}
        x \\ a \\ 1
      \end{bmatrix} \margintag{using the derivation of \cref{eq:q_learning_gradient_update}} \\
      &= - \parentheses*{-1 + \max_{a' \in \sA}\{1 - a' -2\} - (-2 - 1 + 1)} \begin{bmatrix}
        2 \\ -1 \\ 1
      \end{bmatrix} \\
      &= \begin{bmatrix}
        -2 \\ 1 \\ -1
      \end{bmatrix}.
    \end{align*}

    This gives \begin{align*}
      \vw' &= \vw - \alpha \grad_\vw \ell(\vw; \tau) \\
      &= \begin{bmatrix}
        -1 \\ 1 \\ 1
      \end{bmatrix} - \frac{1}{2} \begin{bmatrix}
        -2 \\ 1 \\ -1
      \end{bmatrix} = \begin{bmatrix}
        0 \\ \nicefrac{1}{2} \\ \nicefrac{3}{2}
      \end{bmatrix}.
    \end{align*}
  \end{enumerate}
\end{solution}

\begin{solution}{eligibility_vector}
  We have \begin{align*}
    \grad_\vvarphi \log \pi_\vvarphi(a \mid \vx) &= \frac{\grad_\vvarphi \pi_\vvarphi(a \mid \vx)}{\pi_\vvarphi(a \mid \vx)} \margintag{using the chain rule} \\
    &= \vphi(\vx, a) - \frac{\sum_{b \in \sA} \vphi(\vx, b) \exp(\transpose{\vvarphi} \vphi(\vx, b))}{\sum_{b \in \sA} \exp(\transpose{\vvarphi} \vphi(\vx, b))} \margintag{using elementary calculus} \\
    &= \vphi(\vx, a) - \sum_{b \in \sA} \pi_\vvarphi(b \mid \vx) \cdot \vphi(\vx, b).
  \end{align*}
\end{solution}

\begin{solution}{score_gradients_with_baselines_variance}
  \begin{enumerate}[beginpenalty=10000]
    \item The result follows directly using that \begin{align*}
      \Var{f(\rX) - g(\rX)} = \Var{f(\rX)} + \Var{g(\rX)} - 2 \Cov{f(\rX), g(\rX)}. \margintag{using \cref{eq:sum_variance}}
    \end{align*}

    \item Denote by $r(\tau)$ the discounted rewards attained by trajectory $\tau$.
    Let $f(\tau) \defeq r(\tau) \grad_\vvarphi \altpi_\vvarphi(\tau)$ and $g(\tau) \defeq b \grad_\vvarphi \altpi_\vvarphi(\tau)$.
    Recall that $\E[\tau \sim \altpi_\vvarphi]{g(\tau)} = 0$, implying that \begin{align*}
      \Var{g(\tau)} &= \Var{b \grad_\vvarphi \altpi_\vvarphi(\tau)} \\
      &= \E{\parentheses*{b \grad_\vvarphi \altpi_\vvarphi(\tau)}^2}. \margintag{using the definition of variance \eqref{eq:variance}}
    \end{align*}
    On the other hand, \begin{align*}
      \Cov{f(\tau), g(\tau)} &= \E{(f(\tau) - \E{f(\tau)}) g(\tau)} \margintag{using the definition of covariance \eqref{eq:covariance}} \\
      &= \E{f(\tau) g(\tau)} - \E{f(\tau)} \underbrace{\E{g(\tau)}}_{0} \margintag{using linearity of expectation \eqref{eq:linearity_expectation}} \\
      &= \E{b \cdot r(\tau) \cdot \parentheses*{\grad_\vvarphi \altpi_\vvarphi(\tau)}^2}.
    \end{align*}
    Therefore, if $b^2 \leq 2 b \cdot r(\vx, \va)$ for every state $\vx \in \spX$ and action $\va \in \spA$, then the result follows from \cref{eq:variance_reduction}.
  \end{enumerate}
\end{solution}

\begin{solution}{score_gradients_state_dep_baselines}
  First, observe that \begin{align*}
    \E[\tau \sim \altpi_\vvarphi]{G_0 \grad_\vvarphi \log \altpi_\vvarphi(\tau)} = \E[\tau \sim \altpi_\vvarphi]{\sum_{t=0}^{T-1} G_0 \grad_\vvarphi \log \pi_\vvarphi(\va_t \mid \vx_t)}, \margintag{using \cref{eq:trajectory_distr}}
  \end{align*} and hence, it suffices to show \begin{align}\begin{split}
    &\E[\tau \sim \altpi_\vvarphi]{\sum_{t=0}^{T-1} G_0 \grad_\vvarphi \log \pi_\vvarphi(\va_t \mid \vx_t)} \\
    &= \E[\tau \sim \altpi_\vvarphi]{\sum_{t=0}^{T-1} (G_0 - b(\tau_{0:t-1})) \grad_\vvarphi \log \pi_\vvarphi(\va_t \mid \vx_t)}.
  \end{split}\label{eq:score_gradient_estimator_state_dep_baseline_interm}\end{align}

  We prove \cref{eq:score_gradient_estimator_state_dep_baseline_interm} with an induction on $T$.
  The base case ($T = 0$) is satisfied trivially.
  Fixing any $T$ and assuming \cref{eq:score_gradient_estimator_state_dep_baseline_interm} holds for $T$, we have, \begin{align*}
    &\E[\tau \sim \altpi_\vvarphi]{\sum_{t=0}^{T} (G_0 - b(\tau_{0:t-1})) \grad_\vvarphi \log \pi_\vvarphi(\va_t \mid \vx_t)} \\
    &= \E[\tau_{0:T-1}]{\E[\tau_{T}]{\sum_{t=0}^{T} (G_0 - b(\tau_{0:t-1})) \grad_\vvarphi \log \pi_\vvarphi(\va_t \mid \vx_t)}[\tau_{0:T-1}]} \margintag{using the tower rule \eqref{eq:tower_rule}} \\
    &= \begin{multlined}[t]
      \mathbb{E}_{\tau_{0:T-1}}\Bigg[\sum_{t=0}^{T-1} G_0 \grad_\vvarphi \log \pi_\vvarphi(\va_t \mid \vx_t) \\
      + \E[\tau_{T}]{(G_0 - b(\tau_{0:T-1})) \grad_\vvarphi \log \pi_\vvarphi(\va_{T} \mid \vx_{T})}[\tau_{0:T-1}]\Bigg].
    \end{multlined} \margintag{using the induction hypothesis} \\
    \intertext{Using the score function trick for the score function $\grad_\vvarphi \log \pi_\vvarphi(\va_t \mid \vx_t)$ analogously to the proof of \cref{lem:score_gradients_with_baselines}, we have, \begin{align*}
      &\E[\tau_{T}]{b(\tau_{0:T-1}) \grad_\vvarphi \log \pi_\vvarphi(\va_{T} \mid \vx_{T})}[\tau_{0:T-1}] \\[3pt]
      &= \E[\va_{T}]{b(\tau_{0:T-1}) \grad_\vvarphi \log \pi_\vvarphi(\va_{T} \mid \vx_{T})}[\tau_{0:T-1}] \\
      &= b(\tau_{0:T-1}) \int \pi_\vvarphi(\va_{T} \mid \vx_{T}) \grad_\vvarphi \log \pi_\vvarphi(\va_{T} \mid \vx_{T}) \,d\va_{T} \\
      &= b(\tau_{0:T-1}) \int \grad_\vvarphi \pi_\vvarphi(\va_{T} \mid \vx_{T}) \,d\va_{T} \margintag{using the score function trick, $\grad_\vvarphi \log \pi_\vvarphi(\va_t \mid \vx_t) = \nicefrac{\grad_\vvarphi \pi_\vvarphi(\va_t \mid \vx_t)}{\pi_\vvarphi(\va_t \mid \vx_t)}$} \\
      &= 0.
    \end{align*}
    Thus,}
    &= \E[\tau_{0:T-1}]{\E[\tau_{T}]{\sum_{t=0}^{T} G_0 \grad_\vvarphi \log \pi_\vvarphi(\va_t \mid \vx_t)}[\tau_{0:T-1}]} \\
    &= \E[\tau \sim \altpi_\vvarphi]{\sum_{t=0}^{T} G_0 \grad_\vvarphi \log \pi_\vvarphi(\va_t \mid \vx_t)}. \margintag{using the tower rule \eqref{eq:tower_rule} again} \qedhere
  \end{align*}
\end{solution}

\begin{solution}{policy_gradients_with_downstream_returns}
  Each trajectory $\tau$ is described by four transitions, \begin{align*}
    \tau = (x_0, a_0, r_0, x_1, a_1, r_1, x_2, a_2, r_2, x_3, a_3, r_3, x_4).
  \end{align*}
  Moreover, we have given \begin{align*}
    \pi_\theta(2 \mid x) &= \theta, & \pdv{\pi_\theta(2 \mid x)}{\theta} &= +1 \\
    \pi_\theta(1 \mid x) &= 1 - \theta, & \pdv{\pi_\theta(1 \mid x)}{\theta} &= -1. \\
  \end{align*}

  We first compute the downstream returns for the given episode, \begin{align*}
    G_{0:4} &= r_0 + \gamma r_1 + \gamma^2 r_2 + \gamma^3 r^3 = 1 + \frac{1}{2} \cdot 0 + \frac{1}{4} \cdot 1 + \frac{1}{8} \cdot 1 = \frac{11}{8} \\
    G_{1:4} &= r_1 + \gamma r_2 + \gamma^2 r_3 = 0 + \frac{1}{2} \cdot 1 + \frac{1}{4} \cdot 1 = \frac{3}{4} \\
    G_{2:4} &= r_2 + \gamma r_3 = 1 + \frac{1}{2} \cdot 1 = \frac{3}{2} \\
    G_{3:4} &= r_3 = 1.
  \end{align*}
  Lastly, we can combine them to compute the policy gradient, \begin{align*}
    \grad_\theta j(\theta) &\approx \sum_{t=0}^3 \gamma^t G_{t:4} \grad_\theta \log\pi_\theta(a_t \mid x_t) \margintag{using Monte Carlo approximation of \cref{eq:score_gradient_estimator_downstream_returns} with a single sample} \\
    &= 1 \cdot \frac{11}{8} - 1 \cdot \frac{1}{2} \cdot \frac{3}{4} + 1 \cdot \frac{1}{4} \cdot \frac{3}{2} - 1 \cdot \frac{1}{8} \cdot 1 = \frac{5}{4}.
  \end{align*}
\end{solution}

\begin{solution}{policy_gradient_with_exponential_family}
  \begin{enumerate}[beginpenalty=10000]
    \item The distribution on actions is given by \begin{align*}
      \pi_\vvarphi(a \mid x) = \sigma(f_\vvarphi(x))^a \cdot (1 - \sigma(f_\vvarphi(x)))^{1-a}.
    \end{align*}
    To simplify the notation, we write $\mathbb{E}$ for $\mathbb{E}_{(\vx, a) \sim \pi_\vvarphi}$ and $\fnq$ for $\fnq[\pi_\vvarphi]$.
    We get \begin{align*}
      &\grad_\vvarphi j(\vvarphi) \\
      &= \E{\q{\vx}{a} \grad_\vvarphi (a \log \sigma(f_\vvarphi(\vx)) + (1-a) \log(1 - \sigma(f_\vvarphi(\vx))))} \margintag{using the policy gradient theorem \eqref{eq:policy_gradient_thm}} \\[5pt]
      &= \E{\q{\vx}{a} \grad_f (a \log \sigma(f) + (1-a) \log(1 - \sigma(f))) \grad_\vvarphi f_\vvarphi(\vx)} \margintag{using the chain rule} \\
      &= \E{\q{\vx}{a} \grad_f \parentheses*{-a \log(1 + e^{-f}) + (1-a) \log\parentheses*{\frac{e^{-f}}{1 + e^{-f}}}} \grad_\vvarphi f_\vvarphi(\vx)} \margintag{using the definition of the logistic function \eqref{eq:logistic_function}} \\
      &= \E{\q{\vx}{a} \grad_f \parentheses*{-f + a f - \log(1 + e^{-f})} \grad_\vvarphi f_\vvarphi(\vx)} \\
      &= \E{\q{\vx}{a} \parentheses*{a - 1 + \frac{e^{-f}}{1 + e^{-f}}} \grad_\vvarphi f_\vvarphi(\vx)} \\
      &= \E{\q{\vx}{a} \parentheses*{a - \sigma(f)} \grad_\vvarphi f_\vvarphi(\vx)}. \margintag{using the definition of the logistic function \eqref{eq:logistic_function}}
    \end{align*}
    The term $a - \sigma(f)$ can be understood as a residual as it corresponds to the difference between the target action $a$ and the expected action $\sigma(f)$.

    \item We have \begin{align*}
      \grad_\vvarphi j(\vvarphi) &= \E{\q{\vx}{a} \grad_f \log \pi_f(a \mid \vx) \grad_\vvarphi f(\vx)} \margintag{using \cref{eq:policy_gradient_thm} and the chain rule} \\
      &= \E{\q{\vx}{a} \grad_f (\log h(a) + a f - A(f)) \grad_\vvarphi f(\vx)} \\
      &= \E{\q{\vx}{a} (a - \grad_f A(f)) \grad_\vvarphi f(\vx)}.
    \end{align*}

    \item We have $\grad_f A(f) = \sigma(f)$.
    We are therefore looking for a function $A(f)$ whose derivative is $\sigma(f) = \frac{1}{1 + e^{-f}} = \frac{e^f}{1 + e^f}$.
    With this equality of the sigmoid we can find the integral, and we have $A(f) = \log(1 + e^f) + c$.
    Let us confirm that this gives us the Bernoulli distribution with $c = 0$: \begin{align*}
      \pi_f(a \mid \vx) &= h(a) \exp(a f - \log(1 + e^f)) \\
      &= h(a) \frac{e^{a f}}{1 + e^f} \\
      &= h(a) \begin{cases}
        \sigma(f) & \text{if $a = 1$} \\
        1 - \sigma(f) & \text{if $a = 0$}.
      \end{cases}
    \end{align*}
    This is the Bernoulli distribution with parameter $\sigma(f)$ where we have $h(a) = 1$.

    \item Using that $\grad_f A(f) = f$, we immediately get \begin{align*}
      \grad_\vvarphi j(\vvarphi) &= \E{\q{\vx}{a} (a - f) \grad_\vvarphi f(\vx)}.
    \end{align*}

    \item No, we cannot use the reparameterization trick since we do not know how the states $\vx$ depend on action $a$.
    These dependencies are determined by the unknown dynamics of the environment.
    Nonetheless, we can apply it after sampling an episode according to a policy and then updating policy parameters in hindsight.
    This is for example done by the soft actor-critic (SAC) algorithm (cf. \cref{sec:mfarl:actor_critic_methods:entropy_regularization}).
  \end{enumerate}
\end{solution}

\begin{solution}{soft_value_function}
  \begin{enumerate}[beginpenalty=10000]
    \item \Cref{eq:entropy_reg_rl2} can be written as \begin{align*}
      &\KL{\altpi_\vvarphi}{\altpi_\star} \\
      &= \sum_{t=1}^T \E[(\vx_t, \va_t) \sim \altpi_\vvarphi]{- \frac{1}{\lambda} r(\vx_t, \va_t) - \H{\pi_\vvarphi(\cdot \mid \vx_t)}} \\
      &= \sum_{t=1}^T \E[(\vx_t, \va_t) \sim \altpi_\vvarphi]{- \log \beta(\va_t \mid \vx_t) - \H{\pi_\vvarphi(\cdot \mid \vx_t)}}.
      \intertext{Adding and subtracting $\log Z(\vx_t)$ gives}
      &= \sum_{t=1}^T \E[(\vx_t, \va_t) \sim \altpi_\vvarphi]{\S{\hat{\pi}(\va_t \mid \vx_t)} - \log Z(\vx_t) - \H{\pi_\vvarphi(\cdot \mid \vx_t)}} \\
      &= \sum_{t=1}^T \E[\vx_t \sim \altpi_\vvarphi]{\crH{\pi_\vvarphi(\cdot \mid \vx_t)}{\hat{\pi}(\cdot \mid \vx_t)} - \log Z(\vx_t) - \H{\pi_\vvarphi(\cdot \mid \vx_t)}} \margintag{using the definition of cross-entropy \eqref{eq:cross_entropy}} \\
      &= \sum_{t=1}^T \E[\vx_t \sim \altpi_\vvarphi]{\KL{\pi_\vvarphi(\cdot \mid \vx_t)}{\hat{\pi}(\cdot \mid \vx_t)} - \log Z(\vx_t)}. \margintag{using the definition of KL-divergence \eqref{eq:kl}}
    \end{align*}

    \item We prove the statement by (reverse) induction on $t$.
    For the base case, note that the term \begin{align*}
      \E[\vx_T \sim \altpi_{\vvarphi}]{\KL{\pi_{\vvarphi}(\cdot \mid \vx_T)}{\hat{\pi}(\cdot \mid \vx_T)} - \log Z(\vx_T)}
    \end{align*} is minimized for $\pi_{\vvarphi} \equiv \hat{\pi}$.
    The KL-divergence then evaluates to zero, and we are left only with the $\log Z(\vx_T)$ term.

    For the inductive step, fix any $1 \leq t < T$ and $\pis(\va_t \mid \vx_t)$ must minimize the two terms \begin{align*}
      &\begin{multlined}[t]
        \E[\vx_t \sim \altpi_{\vvarphi}]{\KL{\pi_{\vvarphi}(\cdot \mid \vx_t)}{\hat{\pi}(\cdot \mid \vx_t)} - \log Z(\vx_t)} \\ + \E[(\vx_t, \va_t) \sim \altpi_{\vvarphi}]{\E[\vx_{t+1} \sim p(\cdot \mid \vx_t, \va_t)]{- \log Z(\vx_{t+1})}}
      \end{multlined}
      \intertext{where the first term stems directly from the objective \eqref{eq:entropy_reg_rl3} and the second term represents the contribution of $\pi_{\vvarphi}(\va_t \mid \vx_t)$ to all subsequent terms.
      Letting $\opt{\beta}(\va \mid \vx) \defeq \exp(\q*{\vx}{\va})$, $\opt{Z}(\vx) \defeq \int_\spA \opt{\beta}(\va \mid \vx) \, d \va$, and recalling that we denote by $\opt{\pi}(\cdot \mid \vx)$ the policy $\opt{\beta}(\cdot \mid \vx) / \opt{Z}(\vx)$, this objective can be reexpressed as}
      &= \E[\vx_t \sim \altpi_{\vvarphi}]{\KL{\pi_{\vvarphi}(\cdot \mid \vx_t)}{\opt{\pi}(\cdot \mid \vx_t)} - \log \opt{Z}(\vx_t)}
    \end{align*} which is minimized for $\pi_{\vvarphi} \equiv \opt{\pi}$, leaving only the $\log \opt{Z}(\vx_t)$ term.

    It remains only to observe that $\opt{\beta}(\va_T \mid \vx_T) = \beta(\va_T \mid \vx_T)$, so in the final state $\vx_T$, $\log \opt{Z}(\vx_T) = \log Z(\vx_T)$ and the policies $\pis$ and $\hat{\pi}$ coincide.
  \end{enumerate}
\end{solution}

\begin{solution}{ppo_as_probabilistic_inference}
  \begin{enumerate}[beginpenalty=10000]
    \item Let $\spO$ denote the event that the response $\vy$ is optimal.
    Since the prior over actions (i.e., responses) is not uniform, we have \begin{align}
      \altpi_\star(\vy \mid \vx) &= p(\vy \mid \vx, \spO) \nonumber \\
      &\propto p(\vy \mid \vx) \cdot p(\spO \mid \vx, \vy) \margintag{using Bayes' rule \eqref{eq:bayes_rule}} \nonumber \\
      &= \altpi_{\vvarphi^{\mathrm{init}}}(\vy \mid \vx) \exp\parentheses*{\frac{1}{\lambda} r(\vy \mid \vx)}. \label{eq:ppo_as_probabilistic_inference_helper}
    \end{align}
    The derivation is then analogous to \cref{eq:entropy_reg_rl2}: \begin{align*}
      &\argmin_\vvarphi \KL{\altpi_\vvarphi(\cdot \mid \vx)}{\altpi_\star(\cdot \mid \vx)} \\
      &= \argmin_\vvarphi \crH{\altpi_\vvarphi(\cdot \mid \vx)}{\altpi_\star(\cdot \mid \vx)} - \H{\altpi_\vvarphi(\cdot \mid \vx)} \margintag{using the definition of KL-divergence \eqref{eq:kl}} \\
      &= \argmax_\vvarphi \E[\vy \sim \altpi_\vvarphi(\cdot \mid \vx)]{\log \altpi_\star(\vy \mid \vx) - \log \altpi_\vvarphi(\vy \mid \vx)} \margintag{using the definition of cross-entropy \eqref{eq:cross_entropy} and entropy \eqref{eq:entropy}} \\
      &= \argmax_\vvarphi \E[\vy \sim \altpi_\vvarphi(\cdot \mid \vx)]{r(\vy \mid \vx) + \lambda \log \altpi_{\vvarphi^{\mathrm{init}}}(\vy \mid \vx) - \lambda \log \altpi_\vvarphi(\vy \mid \vx)} \margintag{using \cref{eq:ppo_as_probabilistic_inference_helper} and simplifying} \\
      &= \argmax_\vvarphi \j{\vvarphi; \vvarphi^{\init} \mid \vx}[\lambda]. \margintag{using the definition of cross-entropy \eqref{eq:cross_entropy} and entropy \eqref{eq:entropy}}
    \end{align*}

    \item Recall that the KL-divergence is minimized at $0$ if and only if the two distributions are identical.
    The desired result follows from (1) and \cref{eq:ppo_as_probabilistic_inference_helper}.
  \end{enumerate}
\end{solution}

\section*{\nameref{sec:mbarl}}

\begin{solution}{hucrl_regret}
  \begin{enumerate}[beginpenalty=10000]
    \item To simplify the notation, we use $z_{t,k}$ as shorthand for $(x_{t,k}, \pi_t(x_{t,k}))$ (and similarly $\widehat{z}_{t,k}$ for $(\widehat{x}_{t,k}, \pi_t(\widehat{x}_{t,k}))$).
    The base case is implied trivially. For the induction step, assume that \cref{eq:hucrl_regret:induction} holds at iteration $k$.
    We have \begin{align*}
      \norm{\widehat{x}_{t,k+1} - x_{t,k+1}} &= \norm{\widehat{f}_{t-1}(\widehat{z}_{t,k}) - f(z_{t,k})} \\
      & \le \begin{multlined}[t]
        \norm{\widehat{f}_{n-1}(\widehat{z}_{t,k}) - f(\widehat{z}_{n,k})} + \norm{f(\widehat{z}_{t,k}) - f(z_{t,k})}
      \end{multlined} \margintag{adding and subtracting $f(\widehat{z}_{n,k})$ and using Cauchy-Schwarz} \\
      &\le 2 \beta_{t} \sigma_{t-1}(\widehat{z}_{t,k}) + L_1 \norm{\widehat{x}_{t,k} - x_{t,k}}
      \intertext{where the final inequality follows with high probability from the assumption that the confidence intervals are well-calibrated (cf. \cref{eq:plausible_gaussian_confidence_intervals}) and the assumed Lipschitzness.}
      &= \begin{multlined}[t]
        2 \beta_{t} \brackets*{\sigma_{t-1}(z_{t,k}) + \sigma_{t-1}(\widehat{z}_{t,k}) - \sigma_{t-1}(z_{t,k})} \\ + L_1 \norm{\widehat{x}_{t,k} - x_{t,k}}
      \end{multlined}
      \intertext{Once more using Lipschitz continuity, we obtain}
      &\leq \begin{multlined}[t]
        2 \beta_{t} \brackets*{\sigma_{t-1}(z_{t,k}) + L_2 \norm{\widehat{x}_{t,k} - x_{t,k}}} \\ + L_1 \norm{\widehat{x}_{t,k} - x_{t,k}}
      \end{multlined} \\
      &= 2 \beta_{t} \sigma_{t-1}(z_{t,k}) + \alpha_t \norm{\widehat{x}_{t,k} - x_{t,k}}
      \intertext{where $\alpha_t \defeq (L_1 + 2 \beta_t L_2)$. By the induction hypothesis,}
      &\le 2 \beta_t \sum_{l=0}^{k} \alpha_t^{k-l} \sigma_{t-1}(z_{t,l}).
    \end{align*}
    This is identical to the analysis of UCB from \cref{exercise:bayesian_regret_for_gp_ucb}, only that here errors compound along the trajectory.

    \item The assumption that $\alpha_t \geq 1$ implies that \begin{align}
      \norm{\widehat{x}_{t,k} - x_{t,k}} \leq 2 \beta_t \alpha_t^{H-1} \sum_{l=0}^{k-1} \sigma_{t-1}(z_{t,l}). \label{eq:hucrl_regret:induction_result_helper}
    \end{align}
    Moreover, by definition of $\pi_t$, we have with high probability that $J_H(\pi_t; \widehat{f}) \geq J_H(\pis; f)$.
    This is because $\pi_t$ maximizes reward under \emph{optimistic} dynamics.
    Thus, \begin{align*}
      r_t &= \sum_{k=0}^{H-1} r(x_{t,k}, \pis(x_{t,k})) - r(x_{t,k}, \pi_t(x_{t,k})) \\
      &\leq \sum_{k=0}^{H-1} r(\widehat{x}_{t,k}, \pi_t(\widehat{x}_{t,k})) - r(x_{t,k}, \pi_t(x_{t,k})) \\
      &\leq L_3 \sum_{k=0}^{H-1} \norm{\widehat{x}_{t,k} - x_{t,k}} \margintag{using Lipschitz-continuity of $r$} \\
      &\leq 2 \beta_t \alpha_t^{H-1} L_3 \sum_{k=0}^{H-1} \sum_{l=0}^{k-1} \sigma_{t-1}(z_{t,l}) \margintag{using \cref{eq:hucrl_regret:induction_result_helper}} \\
      &\leq 2 \beta_t H \alpha_t^{H-1} L_3 \sum_{k=0}^{H-1} \sigma_{t-1}(z_{t,k}).
    \end{align*}

    \item Let us first bound $R_T^2$.
    By the Cauchy-Schwarz inequality,
    \begin{align*}
        R_T^2 &\le T \sum_{t=1}^T r_t^2 \\
        &\le \BigO{T \sum_{t=1}^T \beta_{t}^2 H^2 \alpha_t^{2(H-1)} \parentheses*{\sum_{k=0}^{H-1} \sigma_{t-1}(z_{t,k})}^2} \margintag{using (2)} \\
        &\le \BigO{T \beta_{T}^2 H^3 \alpha_T^{2(H-1)} \sum_{t=1}^T \sum_{k=0}^{H-1} \sigma_{t-1}^2(z_{t,k})}. \margintag{using Cauchy-Schwarz and assuming w.l.o.g. that $\beta_t$ is monotonically increasing}
    \end{align*}
    Taking the square root, we obtain
    \begin{align*}
        R_T &\le \BigO{\beta_{T} H^{\frac{3}{2}} \alpha_T^{H-1} \sqrt{T \sum_{t=1}^T \sum_{k=0}^{H-1} \sigma_{t-1}^2(z_{t,k})}} \\
        &\le \BigO{\beta_{T} H^{\frac{3}{2}} \alpha_T^{H-1} \sqrt{T \Gamma_T}}.
    \end{align*}
  \end{enumerate}
\end{solution}

\section*{\nameref{sec:mathematical_background}}

\begin{solution}{sample_variance}
  Using that $X$ is zero-mean, we have that \begin{align*}
    \E{\mean{X}_n^2} &= \Var{\mean{X}_n} = \frac{1}{n} \Var{X} \qquad\text{and} \\
    \E{\frac{1}{n} \sum_{i=1}^n {X_i}^2} &= \frac{1}{n} \sum_{i=1}^n \E{{X_i}^2} = \Var{X}.
  \end{align*}
  Thus, \begin{align*}
    \E{S_n^2} = \frac{n}{n-1} \parentheses*{\E{\frac{1}{n} \sum_{i=1}^n {X_i}^2} - \E{\mean{X}_n^2}} = \Var{X}.
  \end{align*}
\end{solution}

\begin{solution}{concentration_inequalities}
  \begin{enumerate}[beginpenalty=10000]
    \item W.l.o.g. we assume that $X$ is continuous.
    We have \begin{align*}
      \E*{X} &= \int_0^\infty x \cdot p(x) \,d x \margintag{using the definition of expectation \eqref{eq:expectation}} \\
      &= \underbrace{\int_0^\epsilon x \cdot p(x) \,d x}_{\geq 0} + \int_\epsilon^\infty x \cdot p(x) \,d x \geq \epsilon \underbrace{\int_\epsilon^\infty \cdot p(x) \,d x}_{\Pr{X \geq \epsilon}}.
    \end{align*}

    \item Consider the non-negative random variable $Y \defeq (X - \E*{X})^2$.
    We have \begin{align*}
      \Pr{\abs{X - \E*{X}} \geq \epsilon} &= \Pr{(X - \E*{X})^2 \geq \epsilon^2} \\
      &\leq \frac{\E{(X - \E*{X})^2}}{\epsilon^2} \margintag{using Markov's inequality \eqref{eq:markov_inequality}} \\
      &= \frac{\Var*{X}}{\epsilon^2}. \margintag{using the definition of variance \eqref{eq:variance}}
    \end{align*}

    \item Fix any $\epsilon > 0$.
    Applying Chebyshev's inequality and noting that $\E*{\mean{X}_n} = \E*{X}$, we obtain \begin{align*}
      \Pr{\abs{\mean{X}_n - \E*{X}} \geq \epsilon} \leq \frac{\Var*{\mean{X}_n}}{\epsilon^2}.
    \end{align*}
    We further have for the variance of the sample mean that \begin{align*}
      \Var*{\mean{X}_n} = \Var{\frac{1}{n} \sum_{i=1}^n X_i} = \frac{1}{n^2} \sum_{i=1}^n \Var{X_i} = \frac{\Var*{X}}{n}.
    \end{align*}
    Thus, \begin{align*}
      \lim_{n\to\infty} \Pr{\abs{\mean{X}_n - \E*{X}} \geq \epsilon} \leq \lim_{n\to\infty} \frac{\Var*{X}}{\epsilon^2 n} = 0
    \end{align*} which is precisely the definition of $\mean{X}_n \convp \E*{X}$.
  \end{enumerate}
\end{solution}

\begin{solution}{directional_derivatives}
  Fix a $\lambda > 0$.
  Using a first-order expansion, we have \begin{align*}
    f(\vx + \lambda \vd) = f(\vx) + \lambda \transpose{\grad f(\vx)} \vd + o(\lambda \norm{\vd}_2).
  \end{align*}
  Dividing by $\lambda$ yields, \begin{align*}
    \frac{f(\vx + \lambda \vd) - f(\vx)}{\lambda} = \transpose{\grad f(\vx)} \vd + \underbrace{\frac{o(\lambda \norm{\vd}_2)}{\lambda}}_{\to 0}.
  \end{align*}
  Taking the limit $\lambda \to 0$ gives the desired result.
\end{solution}

  \begin{fullwidth}
    \backmatter
    {\small
      \renewcommand{\leftmark}{\textsc{\bibname}}
      \renewcommand{\rightmark}{\textsc{\bibname}}
      \addcontentsline{toc}{chapter}{Bibliography}
      \bibliography{sources}
      \cleardoublepage
    }

    \ifthenelse{\boolean{manuscript}}{}{\chapter{Syllabus}

\begin{fullwidth}
  \setlength{\parindent}{0pt}
  \begin{enumerate}
    \item \emph{Introduction} \quad \cref{sec:fundamentals}
  \end{enumerate}

  \Cref{part1}: Probabilistic Machine Learning
  \begin{enumerate}
    \setcounter{enumi}{1}
    \item \emph{Bayesian Linear Regression (\& Filtering)} \quad \cref{sec:blr} through to \cref{sec:blr:uncertainty} (and \cref{sec:kf})
    \item \emph{Gaussian Processes 1} \quad rest of \cref{sec:blr} and \cref{sec:gp} through to \cref{sec:gp:kernel_functions}
    \item \emph{Gaussian Processes 2} \quad rest of \cref{sec:gp}
    \item \emph{Variational Inference} \quad \cref{sec:approximate_inference} up to \cref{sec:approximate_inference:variational_inference:gradient_of_elbo}
    \item \emph{Markov Chain Monte Carlo} \quad \cref{sec:approximate_inference:variational_inference:gradient_of_elbo} and \cref{sec:approximate_inference:mcmc}
    \item \emph{Bayesian Deep Learning} \quad \cref{sec:bdl}
  \end{enumerate}\vspace{0.7em}

  \Cref{part2}: Sequential Decision-Making
  \begin{enumerate}
    \setcounter{enumi}{7}
    \item \emph{Active Learning \& Bayesian Optimization} \quad \cref{sec:active_learning} and \cref{sec:bayesian_optimization}
    \item \emph{Markov Decision Processes} \quad \cref{sec:mdp}
    \item \emph{Reinforcement Learning 1} \quad \cref{sec:tabular_rl}
    \item \emph{Reinforcement Learning 2} \quad \cref{sec:mfarl} through to \cref{sec:mfarl:policy_approximation}
    \item \emph{Reinforcement Learning 3} \quad \cref{sec:mfarl:actor_critic_methods} through to \cref{sec:mfarl:discussion}
    \item \emph{Reinforcement Learning 4} \quad \cref{sec:mbarl}
  \end{enumerate}
\end{fullwidth}
}
    \setlength\LTleft{0pt}
\setlength\LTright{0pt}

\chapter{Summary of Notation}

We follow these general rules: \begin{itemize}
  \item uppercase italic for constants $N$
  \item lowercase italic for indices $i$ and scalar variables $x$
  \item lowercase italic bold for vectors $\vx$, entries are denoted $\vx(i)$
  \item uppercase italic bold for matrices $\mM$, entries are denoted $\mM(i,j)$
  \item uppercase italic for random variables $X$
  \item uppercase bold for random vectors $\rX$, entries are denoted $\rX(i)$
  \item uppercase italic for sets $\sA$
  \item uppercase calligraphy for spaces (usually infinite sets) $\spA$
\end{itemize}

\vspace{16ex}

\begin{longtable}{@{}p{2.5cm}l@{\extracolsep{\fill}}}
  $\defeq$ & equality by definition \\
  $\approx$ & approximately equals \\
  $\propto$ & proportional to (up to multiplicative constants), $f \propto g$ iff $\exists k.\ \forall x.\ f(x) = k \cdot g(x)$ \\
  $\const$ & an (additive) constant \\
  $\Nat$ & set of natural numbers $\{1, 2, \dots\}$ \\
  $\Nat_0$ & set of natural numbers, including $0$, $\Nat \cup \{0\}$ \\
  $\R$ & set of real numbers \\
  $[m]$ & set of natural numbers from $1$ to $m$, $\{1, 2, \dots, m-1, m\}$ \\
  $i:j$ & subset of natural numbers between $i$ and $j$, $\{i, i+1, \dots, j-1, j\}$ \\
  $(a,b]$ & real interval between $a$ and $b$ including $b$ but not including $a$ \\
  $f : \sA \to \sB$ & function $f$ from elements of set $\sA$ to elements of set $\sB$ \\
  $f \circ g$ & function composition, $f(g(\cdot))$ \\
  $(\cdot)_+$ & $\max\{0, \cdot\}$ \\
  $\log$ & logarithm with base $e$ \\
  $\pset{\sA}$ & power set (set of all subsets) of $\sA$ \\
  $\Ind{predicate}$ & indicator function ($\Ind{predicate} \defeq 1$ if the $predicate$ is true, else $0$) \\
  $\odot$ & Hadamard (element-wise) product \\
  $\gets$ & assignment \\
  \vspace{1ex} \\
  \toprule
  \caption*{\smallcaps{Analysis}} \\
  $\grad f(\vx) \in \R^n$ & gradient of a function $f : \R^n \to \R$ at a point $\vx \in \R^n$ \\
  $\jac \vg(\vx) \in \R^{m \times n}$ & Jacobian of a function $\vg : \R^n \to \R^m$ at a point $\vx \in \R^n$ \\
  $\hes f(\vx) \in \R^{n \times n}$ & Hessian of a function $f : \R^n \to \R$ at a point $\vx \in \R^n$ \\
  \addlinespace
  $\dive \vF$ & divergence operation on vector field $\vF$ \\
  $\lapl f$ & Laplacian of a scalar field $f : \R^n \to \R$ \\
  \addlinespace
  $f \in \BigO{g}$ & $f$ grows at most as fast as $g$ (up to constant factors), $0 \leq \limsup_{n\to\infty} \abs{\frac{f(n)}{g(n)}} < \infty$ \\
  $f \in \BigOTil{g}$ & $f$ grows at most as fast as $g$ up to constant and logarithmic factors \\
  \addlinespace
  $\norm{\cdot}_\alpha$ & $\alpha$-norm \\
  $\norm{\cdot}_\mA$ & Mahalanobis norm induced by matrix $\mA$ \\
  \vspace{2ex} \\
  \toprule
  \caption*{\smallcaps{Linear Algebra}} \\
  $\mI$ & identity matrix \\
  \addlinespace
  $\transpose{\mA}$ & transpose of matrix $\mA$ \\
  $\inv{\mA}$ & inverse of invertible matrix $\mA$ \\
  $\mA^{\nicefrac{1}{2}}$ & \makecell[tl]{square root of a symmetric and positive semi-definite matrix $\mA$} \\
  \addlinespace
  $\det{\mA}$ & determinant of $\mA$ \\
  $\tr{\mA}$ & trace of $\mA$, $\sum_i \mA(i,i)$ \\
  $\diag[i\in\sI]{a_i}$ & diagonal matrix with elements $a_i$, indexed according to the set $\sI$ \\
  \vspace{2ex} \\
  \toprule
  \caption*{\smallcaps{Probability}} \\
  $\Omega$ & sample space \\
  $\spA$ & event space \\
  $\fnPr$ & probability measure \\
  \addlinespace
  $X \sim P$ & random variable $X$ follows the distribution $P$ \\
  $X_{1:n} \iid P$ & \makecell[tl]{random variables $X_{1:n}$ are independent and identically distributed according to \\ distribution $P$} \\
  $x \sim P$ & value $x$ is sampled according to distribution $P$ \\
  $P_X$ & cumulative distribution function of a random variable $X$ \\
  $\overline{P}_X$ & tail distribution function of a random variable $X$ \\
  $\inv{P}_X$ & quantile function of a random variable $X$ \\
  $p_X$ & \makecell[tl]{probability mass function (if discrete) or probability density function \\ (if continuous) of a random variable $X$} \\
  $\Delta^{\spA}$ & set of all probability distributions over the set $\spA$ \\
  $\delta_\alpha$ & Dirac delta function, point density at $\alpha$ \\
  $\pf{\vg}{p}$ & pushforward of a density $p$ under perturbation $\vg$ \\
  \addlinespace
  $X \perp Y$ & random variable $X$ is independent of random variable $Y$ \\
  $X \perp Y \mid Z$ & \makecell[tl]{random variable $X$ is conditionally independent of random variable $Y$ \\ given random variable $Z$} \\
  \addlinespace
  $\E{X}$ & expected value of random variable $X$ \\
  $\E[x \sim X]{f(x)}$ & expected value of the random variable $f(X)$, $\E{f(X)}$ \\
  $\E{X \mid Y}$ & conditional expectation of random variable $X$ given random variable $Y$ \\
  $\Cov{X,Y}$ & covariance of random variable $X$ and random variable $Y$ \\
  $\Cor{X,Y}$ & correlation of random variable $X$ and random variable $Y$ \\
  $\Var{X}$ & variance of random variable $X$ \\
  $\Var{X \mid Y}$ & conditional variance of random variable $X$ given random variable $Y$ \\
  $\mSigma_\rX$ & covariance matrix of random vector $\rX$ \\
  $\mLambda_\rX$ & precision matrix of random vector $\rX$ \\
  \addlinespace
  $\mathrm{MSE}(X)$ & mean squared error of random variable $X$ \\
  $\mean{X}_n$ & sample mean of random variable $X$ with $n$ samples \\
  $S_n^2$ & sample variance of random variable $X$ with $n$ samples \\
  \addlinespace
  $X_n \almostsurely X$ & the sequence of random variables $X_n$ converges almost surely to $X$ \\
  $X_n \convp X$ & the sequence of random variables $X_n$ converges to $X$ in probability \\
  $X_n \convd X$ & the sequence of random variables $X_n$ converges to $X$ in distribution \\
  \addlinespace
  $\S{u}$ & surprise associated with an event of probability $u$ \\
  $\H{p}, \H{X}$ & entropy of distribution $p$ (or random variable $X$) \\
  $\crH{p}{q}$ & cross-entropy of distribution $q$ with respect to distribution $p$ \\
  $\KL{p}{q}$ & KL-divergence of distribution $q$ with respect to distribution $p$ \\
  $\Fisher{p}{q}$ & relative Fisher information of distribution $q$ with respect to distribution $p$ \\
  $\H{X \mid Y}$ & conditional entropy of random variable $X$ given random variable $Y$ \\
  $\H{X, Y}$ & joint entropy of random variables $X$ and $Y$ \\
  $\I{X}{Y}$ & mutual information of random variables $X$ and $Y$ \\
  $\I{X}{Y}[Z]$ & \makecell[tl]{conditional mutual information of random variables $X$ and $Y$ given random \\ variable $Z$} \\
  \addlinespace
  $\N{\vmu}{\mSigma}$ & normal distribution with mean $\vmu$ covariance $\mSigma$ \\
  $\Laplace{\vmu}{h}$ & Laplace distribution with mean $\vmu$ scale $h$ \\
  $\Unif{\sS}$ & uniform distribution on the set $\sS$ \\
  $\Bern{p}$ & Bernoulli distribution with success probability $p$ \\
  $\Bin{n}{p}$ & binomial distribution with $n$ trials and success probability $p$ \\
  $\Beta{\alpha}{\beta}$ & beta distribution with shape parameters $\alpha$ and $\beta$ \\
  $\GammaDistr{\alpha}{\beta}$ & gamma distribution with shape $\alpha$ and rate $\beta$ \\
  $\Cauchy{m}{\tau}$ & Cauchy distribution with location $m$ and scale $\tau$ \\
  $\Pareto{c}{\alpha}$ & Pareto distribution with cutoff threshold $c$ and shape $\alpha$ \\
  \vspace{2ex} \\
  \toprule
  \caption*{\smallcaps{Supervised Learning}} \\
  $\vtheta$ & parameterization of a model \\
  \addlinespace
  $\spX$ & input space \\
  $\spY$ & label space \\
  $\vx \in \spX$ & input \\
  $\epsilon(\vx)$ & zero-mean noise, sometimes assumed to be independent of $\vx$ \\
  $y \in \spY$ & (noisy) label, $f(\vx) + \epsilon(\vx)$ where $f$ is unknown \\
  $\spD \subseteq \spX \times \spY$ & labeled training data, $\{(\vx_i, y_i)\}_{i=1}^n$ \\
  $\mX \in \R^{n \times d}$ & design matrix when $\spX = \R^d$ \\
  $\mPhi \in \R^{n \times e}$ & design matrix in feature space $\R^e$ \\
  $\vy \in \R^n$ & label vector when $\spY = \R$ \\
  \addlinespace
  $p(\vtheta)$ & prior belief about $\vtheta$ \\
  $p(\vtheta \mid \vx_{1:n}, y_{1:n})$ & posterior belief about $\vtheta$ given training data \\
  $p(y_{1:n} \mid \vx_{1:n}, \vtheta)$ & likelihood of training data under the model parameterized by $\vtheta$ \\
  $p(y_{1:n} \mid \vx_{1:n})$ & marginal likelihood of training data \\
  \addlinespace
  $\vthetahat_\MLE$ & maximum likelihood estimate of $\vtheta$ \\
  $\vthetahat_\MAP$ & maximum a posteriori estimate of $\vtheta$ \\
  \addlinespace
  $\ell_\mathrm{nll}(\vtheta; \spD)$ & negative log-likelihood of the training data $\spD$ under model $\vtheta$ \\
  \vspace{2ex} \\
  \toprule
  \caption*{\smallcaps{Bayesian Linear Models}} \\
  $\vw \in \R^d$ & weights of linear function $f(\vx; \vw) = \transpose{\vw} \vx$ \\
  \addlinespace
  $\vwhat_\ls$ & least squares estimate of $\vw$ \\
  $\vwhat_\ridge$ & ridge estimate of $\vw$ \\
  $\vwhat_\lasso$ & lasso estimate of $\vw$ \\
  \addlinespace
  $\N{\vzero}{\sigmap^2 \mI}$ & prior \\
  $\N{\transpose{\vw} \vx}{\sigman^2}$ & likelihood \\
  $\vmu \in \R^d$ & posterior mean, $\sigman^{-2} \mSigma \transpose{\mX} \vy$ \\
  $\mSigma \in \R^{d \times d}$ & posterior covariance matrix, $\inv{\parentheses{\sigman^{-2} \transpose{\mX} \mX + \sigmap^{-2} \mI}}$ \\
  \addlinespace
  $\mK \in \R^{n \times n}$ & kernel matrix, $\sigmap^2 \mX \transpose{\mX}$ \\
  \addlinespace
  $\sigma$ & logistic function \\
  $\Bern{\sigma(\transpose{\vw} \vx)}$ & logistic likelihood \\
  $\ell_\mathrm{log}(\cdot; \vx, y)$ & logistic loss of a single training example $(\vx, y)$ \\
  \vspace{2ex} \\
  \toprule
  \caption*{\smallcaps{Kalman Filters}} \\
  $\rX_t$ & sequence of hidden states in $\R^d$ \\
  $\rY_t$ & sequence of observations in $\R^m$ \\
  \addlinespace
  $\mF \in \R^{d \times d}$ & motion model \\
  $\mH \in \R^{m \times d}$ & sensor model \\
  $\vvarepsilon_t$ & zero-mean motion noise with covariance matrix $\mSigma_x$ \\
  $\veta_t$ & zero-mean sensor noise with covariance matrix $\mSigma_y$ \\
  \addlinespace
  $\mK_t \in \R^{d \times m}$ & Kalman gain \\
  \vspace{2ex} \\
  \toprule
  \caption*{\smallcaps{Gaussian Processes}} \\
  $\mu : \spX \to \R$ & mean function \\
  $k : \spX \times \spX \to \R$ & kernel function / covariance function \\
  $f \sim \GP{\mu}{k}$ & $f$ is a Gaussian process with mean function $\mu$ and kernel function $k$ \\
  \addlinespace
  $\spH_k(\spX)$ & reproducing kernel Hilbert space associated with kernel function $k : \spX \times \spX \to \R$ \\
  \vspace{2ex} \\
  \toprule
  \caption*{\smallcaps{Deep Models}} \\
  $\mW_l \in \R^{n_l \times n_{l-1}}$ & weight matrix of layer $l$ \\
  $\vnu^{(l)} \in \R^{n_l}$ & activations of layer $l$ \\
  \addlinespace
  $\varphi$ & activation function \\
  $\mathrm{Tanh}$ & hyperbolic tangent activation function \\
  $\mathrm{ReLU}$ & rectified linear unit activation function \\
  \addlinespace
  $\sigma_i(\vf)$ & softmax function computing the probability mass of class $i$ given outputs $\vf$ \\
  \vspace{2ex} \\
  \toprule
  \caption*{\smallcaps{Variational Inference}} \\
  $\spQ$ & variational family \\
  $\vlambda \in \Lambda$ & variational parameters \\
  $q_\vlambda$ & variational posterior parameterized by $\vlambda$ \\
  $L(q, p; \spD)$ & evidence lower bound for data $\spD$ of variational posterior $q$ and true posterior $p(\cdot \mid \spD)$ \\
  \vspace{2ex} \\
  \toprule
  \caption*{\smallcaps{Markov Chains}} \\
  $\sS$ & set of $n$ states \\
  $X_t$ & sequence of states \\
  $p(x' \mid x)$ & transition function, probability of going from state $x$ to state $x'$ \\
  $p^{(t)}(x' \mid x)$ & probability of reaching $x'$ from $x$ in exactly $t$ steps \\
  $\mP \in \R^{n \times n}$ & transition matrix \\
  \addlinespace
  $q_t$ & distribution over states at time $t$ \\
  $\pi$ & stationary distribution \\
  \addlinespace
  $\norm{\mu - \nu}_\mathrm{TV}$ & total variation distance between two distributions $\mu$ and $\nu$ \\
  $\tau_\mathrm{TV}$ & mixing time with respect to total variation distance \\
  \vspace{7.9ex} \\
  \toprule
  \caption*{\smallcaps{Markov Chain Monte Carlo Methods}} \\
  $r(\vxp \mid \vx)$ & proposal distribution, probability of proposing $\vxp$ when in $\vx$ \\
  $\alpha(\vxp \mid \vx)$ & acceptance distribution, probability of accepting the proposal $\vxp$ when in $\vx$ \\
  \addlinespace
  $f$ & energy function \\
  \vspace{2ex} \\
  \toprule
  \caption*{\smallcaps{Active Learning}} \\
  $\sS \subseteq \spX$ & set of observations \\
  $I(\sS)$ & maximization objective, quantifying the ``information value'' of $\sS$ \\
  $\Delta_I(\vx \mid \sA)$ & marginal gain of observation $\vx$ with respect to objective $I$ given prior observations $\sA$ \\
  \vspace{2ex} \\
  \toprule
  \caption*{\smallcaps{Bayesian Optimization}} \\
  $R_T$ & cumulative regret for time horizon $T$ \\
  $F(\vx; \mu, \sigma)$ & acquisition function \\
  $\spC_t(\vx)$ & confidence interval of $\opt{f}(\vx)$ after round $t$ \\
  $\beta_t(\delta)$ & scale of confidence interval to achieve confidence level $\delta$ \\
  $\gamma_T$ & maximum information gain after $T$ rounds \\
  \vspace{2ex} \\
  \toprule
  \caption*{\smallcaps{Reinforcement Learning}} \\
  $X, \spX$ & set of states \\
  $A, \spA$ & set of actions \\
  $p(x' \mid x, a)$ & \makecell[tl]{dynamics model, probability of transitioning from state $x$ to state $x'$ when playing \\ action $a$} \\
  $r$ & reward function \\
  $X_t$ & sequence of states \\
  $A_t$ & sequence of actions \\
  $R_t$ & sequence of rewards \\
  $\pi(a \mid x)$ & policy, probability of playing action $a$ when in state $x$ \\
  $G_t$ & discounted payoff from time $t$ \\
  $\gamma$ & discount factor \\
  \addlinespace
  $\v[\pi]{x}[t]$ & state value function, average discounted payoff from time $t$ starting from state $x$ \\
  $\q[\pi]{x}{a}[t]$ & \makecell[tl]{state-action value function, average discounted payoff from time $t$ starting from state $x$ \\ playing action $a$} \\
  $\a[\pi]{x}{a}[t]$ & advantage function, $\q[\pi]{x}{a}[t] - \v[\pi]{x}[t]$ \\
  $\j{\pi}$ & policy value function, expected reward of policy $\pi$ \\
\end{longtable}

    \chapter*{Acronyms}\addcontentsline{toc}{chapter}{Acronyms}
    \begin{flushleft}
      \begin{small}
        \begin{multicols}{2}
          \setlist{itemsep=-5pt}\printacronyms[heading=none]
        \end{multicols}
      \end{small}
    \end{flushleft}

    \printindex

    \ifthenelse{\boolean{manuscript}}{}{\cleardoublepage\chapter{Errata}

\begin{longtable}{p{4cm}l}
  September 28, 2024 & fixed errors in solution to \cref{exercise:gaussian_pdf_prod} \\
  October 7, 2024 & definition of Laplace kernel \eqref{eq:laplace_kernel} was using incorrect norm \\
  January 3, 2025 & $\lambda$ in examples of regularization in \cref{sec:fundamentals:parameter_esitmation:map} was incorrect \\
  January 7, 2025 & fixed misplaced brackets in \cref{eq:mbarl_planning_monte_carlo} \\
  January 8, 2025 & fixed typo in \cref{rmk:mc_convergence} \\
\end{longtable}
}
  \end{fullwidth}
\end{document}